\documentclass[10pt,journal,compsoc]{IEEEtran}

\usepackage[nocompress]{cite}

\usepackage{url}
\usepackage{ragged2e}
\usepackage{epsfig}
\usepackage{graphicx}
\usepackage{amsmath,amssymb} 
\usepackage{mathptmx}      
\usepackage{algorithm}
\usepackage{algorithmicx}
\usepackage{algpseudocode}
\usepackage{graphics}
\usepackage{mathrsfs}
\usepackage{threeparttable}
\usepackage{color}
\usepackage[normalem]{ulem}
\usepackage{multirow}
\usepackage{float}
\usepackage{amsfonts}
\usepackage{bm}
\usepackage{array}
\usepackage[table]{xcolor}
\usepackage{colortbl}
\usepackage{pifont}
\usepackage{diagbox}
\usepackage{rotating}
\usepackage{booktabs}
\usepackage{overpic}
\usepackage{textcomp}
\usepackage{contour}
\usepackage{enumitem}
\usepackage{stfloats}
\usepackage{threeparttable}
\usepackage{makecell}
\usepackage{subcaption}
\usepackage[colorlinks=true,breaklinks=true,urlcolor=magenta]{hyperref}

\def\etal{{\em et al.~}}

\newcommand{\secref}[1]{\S\ref{#1}}

\definecolor{mygray1}{gray}{.75}

\graphicspath{{./figs/}}
\DeclareGraphicsExtensions{.jpg,.pdf,.png}

\RequirePackage{silence}
\begin{document}

\title{When SAM2 Meets Video Shadow and Mirror Detection}

\author{
Leiping Jie\\
\IEEEcompsocitemizethanks{
\IEEEcompsocthanksitem Leiping Jie is with Faculty of Mathematics and Computer Science, Guangdong Ocean University, China (leipingjie@gdou.edu.cn).
}
}


\IEEEtitleabstractindextext{%
\begin{abstract} \justifying
  As the successor to the Segment Anything Model (SAM), the Segment Anything Model 2 (SAM2) not only improves performance in image segmentation but also extends its capabilities to video segmentation. However, its effectiveness in segmenting rare objects that seldom appear in videos remains underexplored. In this study, we evaluate SAM2 on three distinct video segmentation tasks: Video Shadow Detection (VSD) and Video Mirror Detection (VMD). Specifically, we use ground truth point or mask prompts to initialize the first frame and then predict corresponding masks for subsequent frames. Experimental results show that SAM2's performance on these tasks is suboptimal, especially when point prompts are used, both quantitatively and qualitatively. Code is available at \url{https://github.com/LeipingJie/SAM2Video}.
\end{abstract}

}

\maketitle

\IEEEdisplaynontitleabstractindextext
\IEEEpeerreviewmaketitle
\IEEEraisesectionheading{\section{Introduction}\label{sec:introduction}}

\IEEEPARstart{S}{egment} Anything Model (SAM)\cite{kirillov2023segment}, a groundbreaking creation by Meta, has revolutionized the field of computer vision with its unprecedented ability to segment objects within images. Since its release in April 2023\footnote{\url{https://segment-anything.com/}}, SAM has been applied to a variety of specific object segmentation tasks, including Camouflaged Object Detection~\cite{chen2023sam, tang2023can, zhang2024towards}, Shadow Detection~\cite{jie2023sam, chen2023sam}, Medical Image Segmentation~\cite{ma2024segment, mazurowski2023segment}, and Remote Sensing Image Semantic Segmentation~\cite{wang2024samrs, osco2023segment}, among others. However, the performance reported in many of these tasks has been less than satisfactory. This limitation is likely due to the lack of relevant training samples in SAM's dataset, SA-1B, which consists of \textit{11M diverse, high-resolution, licensed, and privacy-protecting images, as well as 1.1B high-quality segmentation masks collected with its data engine}. Inspired by fine-tuning techniques~\cite{he2021towards, hu2021lora}, such as adapters—small, learnable bottleneck layers seamlessly integrated at various points within pre-trained models—several studies have proposed elaborated adapters for tasks like shadow detection~\cite{jie2023adaptershadow, chen2023sam, chen2023make}, and medical image segmentation~\cite{wu2023medical, gong20243dsam}, demonstrating superior performance.

Recently, Meta AI introduced SAM2~\cite{ravi2024sam}, which is not only more accurate but also $6x$ faster than its predecessor, SAM. SAM2 extends the original image segmentation architecture to support real-time video segmentation by incorporating a transformer-based design enhanced with streaming memory~\footnote{\url{https://ai.meta.com/sam2/}}. Subsequent evaluations of SAM2's performance have been conducted~\cite{chen2024sam2, tang2024evaluating, xiong2024sam2, zhou2024sam2, lian2024evaluation}, with results varying across different datasets. For instance, while the performance of shadow detection appears to be similar for both SAM and SAM2. In~\cite{chen2024sam2}, SAM2 outperforms SAM in polyp segmentation. However, beyond the evaluated video segmentation datasets in SAM2, there has been limited exploration of SAM2's performance on other video segmentation tasks.

In this paper, we further evaluate SAM2's performance on two distinct video segmentation datasets: ViSha~\cite{chen2021triple} and VMD~\cite{lin2023learning}, which focus on video shadow detection and video mirror detection, respectively. Specifically, we initialize the first frame of each video using either point or mask prompts, allowing SAM2 to generate the corresponding masks for subsequent frames. We then compare the results using public evaluation metrics against state-of-the-art video segmentation models. Our experimental results show the following: (1) SAM2 demonstrates excellent performance when using mask prompts, and (2) the mask prompt outperforms the point prompt, which often fails to segment objects accurately in the first frame.

\section{Experiments}\label{sec:experiment}
In this section, we first introduce the ViSha~\cite{chen2021triple} and VMD~\cite{lin2023learning} datasets, which are used to evaluate the performance of SAM2. Next, we present the evaluation metrics (\secref{sec:evaluation_metric}). Finally, we compare SAM2 with state-of-the-art (SOTA) methods for video shadow and mirror segmentation (\secref{sec:quantitative_eval}) and provide qualitative visualizations of the results on these datasets (\secref{sec:qualitative_eval}).

\subsection{Datasets}\label{sec:datasets}
We adopt two different datasets to evaluate SAM2's video segmentation performance.

\noindent\textbf{ViSha}~\cite{chen2021triple} is a video shadow detection dataset comprising 120 videos with diverse characteristics. More than half of the videos are sourced from five video tracking benchmarks, while the remaining 59 are self-captured. The dataset has a frame rate of 30 fps, totaling 11,685 frames across 390 seconds. The longest video contains 103 frames, and the shortest has 11 frames. To mitigate overfitting, the dataset is randomly split into training and testing sets at a 5:7 ratio (50 training and 70 testing video sequences).

\noindent\textbf{VMD}~\cite{lin2023learning} is a large-scale video mirror detection dataset designed to advance research in video mirror detection. It contains 269 videos and 14,988 frames with precise annotations from a variety of scenes. The videos were captured using smartphones in real-life scenarios featuring mirrors, then manually trimmed to ensure each frame contains at least one mirror region. The 269 video clips are randomly split into a training set of 143 videos (7,835 frames) and a test set of 126 videos (7,152 frames). Pixel-level mirror masks were annotated for each frame. All videos have a frame rate of 30 fps, with a total duration of 502 seconds.

\begin{table*}[hbt!]
  \caption{Quantitative comparison results on the ViSha~\cite{chen2021triple} dataset. We use two types of prompts: point and mask. When point mask is adopted, we randomly select predefined number of postive and negative points from the first frame using the corresponding ground truth mask. Here, the number inside the bracket of column two represents the predefined number. For simplicity, we set the number of positive points equal to the number of negative points. The best and second-best results are highlighted in \textcolor{red}{red} and \textcolor{blue}{blue}, respectively.}
  \label{table_quantitative_visha}
  \begin{center}
    \begin{tabular}{|c|c|c|cccc|}
      \hline
      Method & Prompt & Pub$_{year}$ & MAE$\downarrow$ & F1-score$\uparrow$ & IOU$\uparrow$ & BER$\downarrow$ \\
    \hline
    TVSD-Net~\cite{chen2021triple} & - & CVPR$_{21}$ & 0.033 & 0.757 & 0.567 & 17.70 \\
    TFW~\cite{hu2021temporal} & - & Arxiv$_{21}$ & 0.078 & 0.683 & 0.510 & 17.03 \\
    STF-Net~\cite{lin2022spatial} & - & VRCAI$_{22}$ & 0.029 & 0.761 & 0.584 & 15.31 \\
    STICT~\cite{lu2022video} & - & CVPR$_{22}$ & 0.046 & 0.702 & 0.545 & 16.60 \\
    SC-Cor~\cite{ding2022learning} & - & ECCV$_{22}$ & 0.042 & 0.762 & 0.615 & 13.61 \\
    MPL-Net~\cite{chen2022semi} & - & MM$_{22}$ & - & - & - & 10.34 \\
    SCOTCH and SODA~\cite{liu2023scotch} & - & CVPR$_{23}$ & 0.029 & 0.793 & 0.640 & 9.066 \\
    DAS~\cite{wang2023detect} & - & TCSVT$_{23}$ & 0.034 & 0.754 & 0.575 & 12.58 \\
    TBGDiff~\cite{zhou2024timeline} & - & MM$_{24}$ & 0.023 & 0.797 & 0.667 & 8.58 \\
    Lin~\etal~\cite{lin2024learning} & - & CVPR$_{18}$ & \color{blue}{0.019} & 0.817 & 0.741 & \color{blue}{7.2} \\
    CVSD~\cite{duan2024twostage} & - & ECCV$_{24}$ & 0.027 & 0.801 & 0.684 & 8.96 \\
    SSTINet~\cite{weistructure24} & - & IJCAI$_{24}$ & \color{red}{0.017} & 0.866 & 0.746 & \color{red}{6.48} \\
    \hline 
    \hline 
    SAM2-tiny~\cite{kirillov2023segment} & point(5) & - & 0.113 & 0.68 & 0.474 & 20.665 \\
    SAM2-tiny~\cite{kirillov2023segment} & point(10) & - & 0.138 & 0.676 & 0.425 & 23.582 \\
    SAM2-tiny~\cite{kirillov2023segment} & point(20) & - & 0.111 & 0.598 & 0.191 & 38.402 \\
    SAM2-tiny~\cite{kirillov2023segment} & point(30) & - & 0.094 & 0.152 & 0.039 & 48.011 \\
    SAM2-tiny~\cite{kirillov2023segment} & point(40) & - & 0.096 & 0.122 & 0.0 & 49.992 \\
    SAM2-tiny~\cite{kirillov2023segment} & point(50) & - & 0.096 & 0.122 & 0.0 & 50.0 \\
    SAM2-small~\cite{kirillov2023segment} & point(5) & - & 0.11 & 0.666 & 0.461 & 22.11 \\
    SAM2-small~\cite{kirillov2023segment} & point(10) & - & 0.105 & 0.673 & 0.5 & 17.665 \\
    SAM2-small~\cite{kirillov2023segment} & point(20) & - & 0.171 & 0.591 & 0.429 & 20.29 \\
    SAM2-small~\cite{kirillov2023segment} & point(30) & - & 0.191 & 0.605 & 0.378 & 24.554 \\
    SAM2-small~\cite{kirillov2023segment} & point(40) & - & 0.167 & 0.635 & 0.286 & 32.124 \\
    SAM2-small~\cite{kirillov2023segment} & point(50) & - & 0.09 & 0.59 & 0.2 & 38.477 \\
    SAM2-large~\cite{kirillov2023segment} & point(5) & - & 0.074 & 0.676 & 0.464 & 21.872 \\
    SAM2-large~\cite{kirillov2023segment} & point(10) & - & 0.054 & 0.718 & 0.505 & 19.489 \\
    SAM2-large~\cite{kirillov2023segment} & point(20) & - & 0.058 & 0.707 & 0.445 & 22.597 \\
    SAM2-large~\cite{kirillov2023segment} & point(30) & - & 0.057 & 0.705 & 0.416 & 24.667 \\
    SAM2-large~\cite{kirillov2023segment} & point(40) & - & 0.065 & 0.652 & 0.368 & 26.683 \\
    SAM2-large~\cite{kirillov2023segment} & point(50) & - & 0.055 & 0.671 & 0.35 & 27.842 \\
    SAM2-tiny~\cite{kirillov2023segment} & mask & - & \color{red}{0.017} & \color{red}{0.878} & \color{blue}{0.748} & 9.877 \\
    SAM2-small~\cite{kirillov2023segment} & mask & - & \color{blue}{0.019} & \color{blue}{0.874} & \color{red}{0.755} & 9.056 \\
    SAM2-large~\cite{kirillov2023segment} & mask & - & \color{blue}{0.019} & 0.868 & 0.744 & 9.543 \\
    \hline 
    \end{tabular}
  \end{center}
\end{table*}

\begin{table*}[htb!]
  \caption{Quantitative comparison results on the VMD~\cite{lin2023learning} dataset. We use two types of prompts: point and mask. When point mask is adopted, we randomly select predefined number of postive and negative points from the first frame using the corresponding ground truth mask. Here, the number inside the bracket of column two represents the predefined number. For simplicity, we set the number of positive points equal to the number of negative points. The best and second-best results are highlighted in \textcolor{red}{red} and \textcolor{blue}{blue}, respectively.}
  \label{table_quantitative_vmd}
  \begin{center}
    \begin{tabular}{|c|c|c|ccc|}
      \hline
      Method & Prompt & Pub$_{year}$ & MAE$\downarrow$ & F1-score$\uparrow$ & IOU$\uparrow$ \\
    \hline
    VMD-Net~\cite{lin2023learning} & - & CVPR$_{23}$ & 0.105 & 0.787 & 0.567 \\
    MG-VMD~\cite{Warren_2024_CVPR} & - & CVPR$_{24}$ & 0.127 & 0.869 & 0.725 \\
    ZOOM~\cite{xu2024zoom} & - & AAAI$_{24}$ & 0.387 & 0.448 & 0.294 \\
    Xu~\etal~\cite{xu2024fusion} & - & ICME$_{24}$ & 0.100 & 0.810 & 0.634 \\
    FTM-Net~\cite{xingfarther} & - & - & 0.083 & 0.833 & 0.649 \\
    \hline 
    \hline 
    SAM2-tiny~\cite{kirillov2023segment} & point(5) & - & 0.139 & 0.779 & 0.468 \\
    SAM2-tiny~\cite{kirillov2023segment} & point(10) & - & 0.159 & 0.731 & 0.389 \\
    SAM2-tiny~\cite{kirillov2023segment} & point(20) & - & 0.21 & 0.276 & 0.071 \\
    SAM2-tiny~\cite{kirillov2023segment} & point(30) & - & 0.216 & 0.264 & 0.008 \\
    SAM2-tiny~\cite{kirillov2023segment} & point(40) & - & 0.217 & 0.264 & 0.0 \\
    SAM2-tiny~\cite{kirillov2023segment} & point(50) & - & 0.217 & 0.264 & 0.0 \\
    SAM2-small~\cite{kirillov2023segment} & point(5) & - & 0.138 & 0.771 & 0.469 \\
    SAM2-small~\cite{kirillov2023segment} & point(10) & - & 0.119 & 0.794 & 0.565 \\
    SAM2-small~\cite{kirillov2023segment} & point(20) & - & 0.117 & 0.797 & 0.59 \\
    SAM2-small~\cite{kirillov2023segment} & point(30) & - & 0.115 & 0.799 & 0.537 \\
    SAM2-small~\cite{kirillov2023segment} & point(40) & - & 0.126 & 0.731 & 0.378 \\
    SAM2-small~\cite{kirillov2023segment} & point(50) & - & 0.178 & 0.458 & 0.148 \\
    SAM2-large~\cite{kirillov2023segment} & point(5) & - & 0.139 & 0.786 & 0.464 \\
    SAM2-large~\cite{kirillov2023segment} & point(10) & - & 0.146 & 0.796 & 0.484 \\
    SAM2-large~\cite{kirillov2023segment} & point(20) & - & 0.152 & 0.786 & 0.457 \\
    SAM2-large~\cite{kirillov2023segment} & point(30) & - & 0.16 & 0.751 & 0.405 \\
    SAM2-large~\cite{kirillov2023segment} & point(40) & - & 0.176 & 0.713 & 0.342 \\
    SAM2-large~\cite{kirillov2023segment} & point(50) & - & 0.179 & 0.676 & 0.297 \\
    SAM2-tiny~\cite{kirillov2023segment} & mask & - & \color{blue}{0.032} & \color{blue}{0.955} & \color{blue}{0.871} \\
    SAM2-small~\cite{kirillov2023segment} & mask & - & 0.037 & \color{blue}{0.955} & 0.857 \\
    SAM2-large~\cite{kirillov2023segment} & mask & - & \color{red}{0.029} & \color{red}{0.961} & \color{red}{0.884} \\
    \hline 
    \end{tabular}
  \end{center}
\end{table*}

\begin{figure*}[!ht]
	\centering
	\vspace*{1.3mm}
	\begin{subfigure}{0.08\textwidth}
		\includegraphics[width=\textwidth]{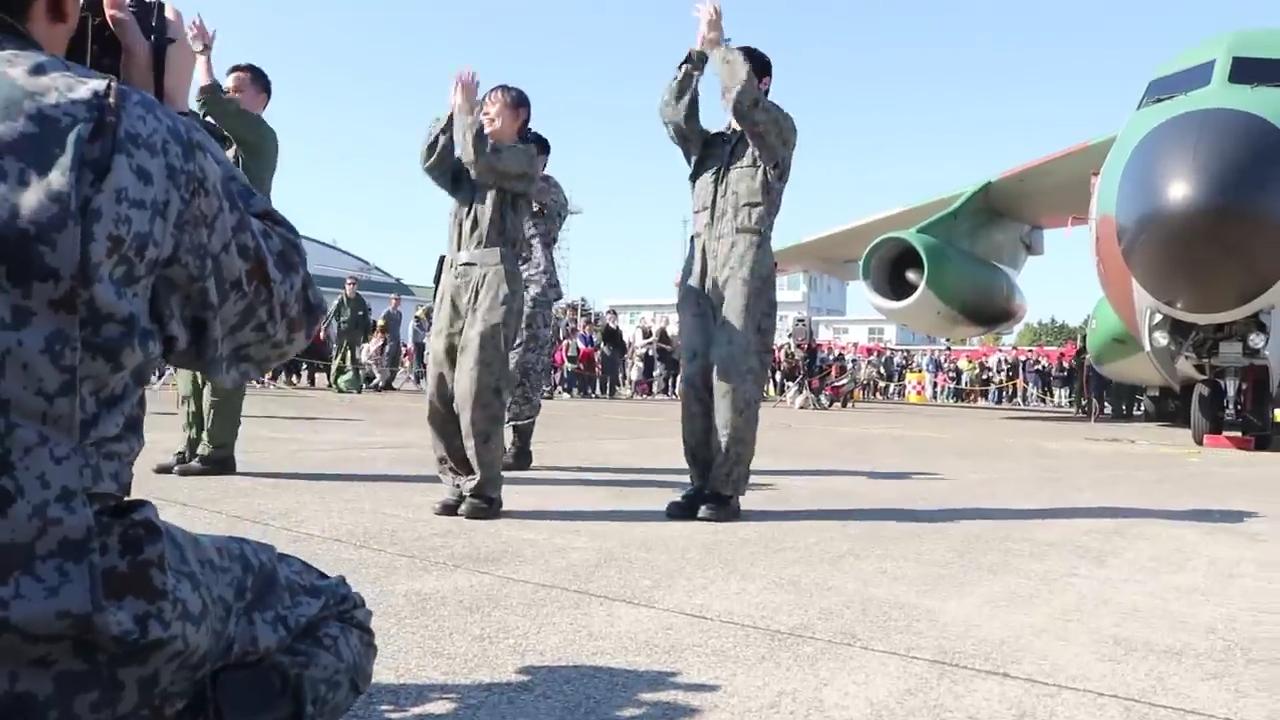}
	\end{subfigure}
	\begin{subfigure}{0.08\textwidth}
		\includegraphics[width=\textwidth]{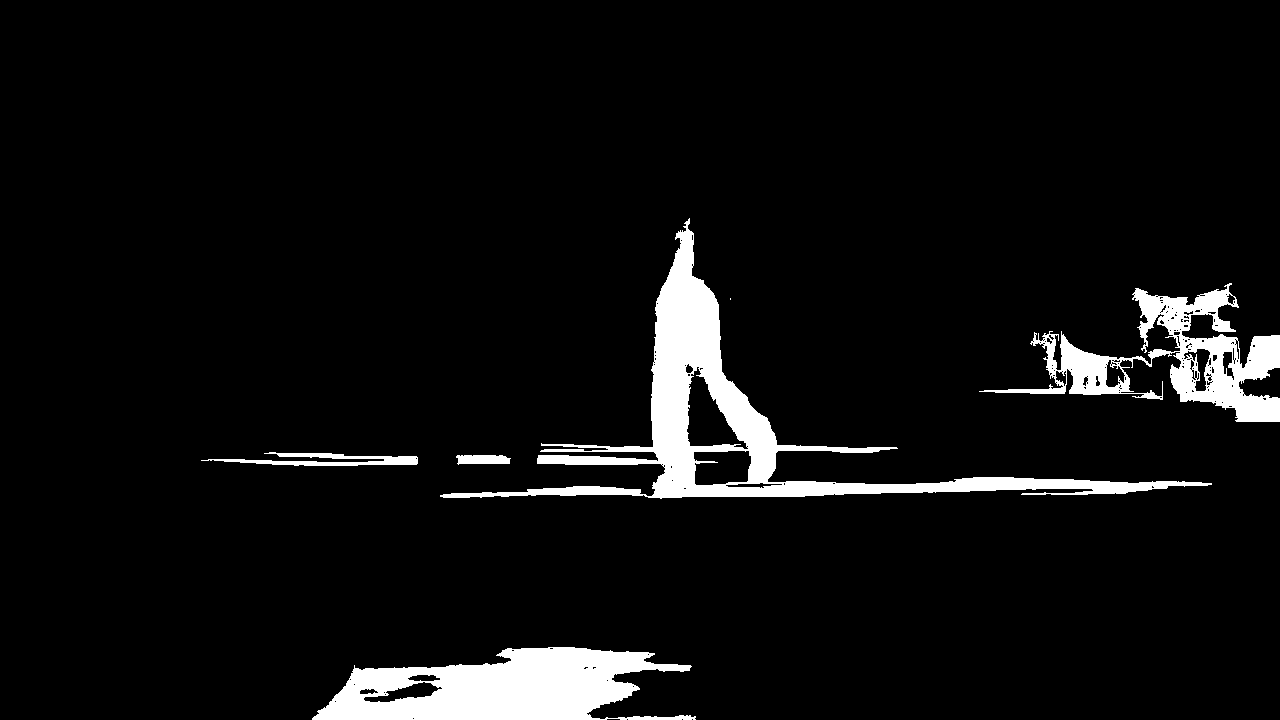}
	\end{subfigure}
	\begin{subfigure}{0.08\textwidth}
		\includegraphics[width=\textwidth]{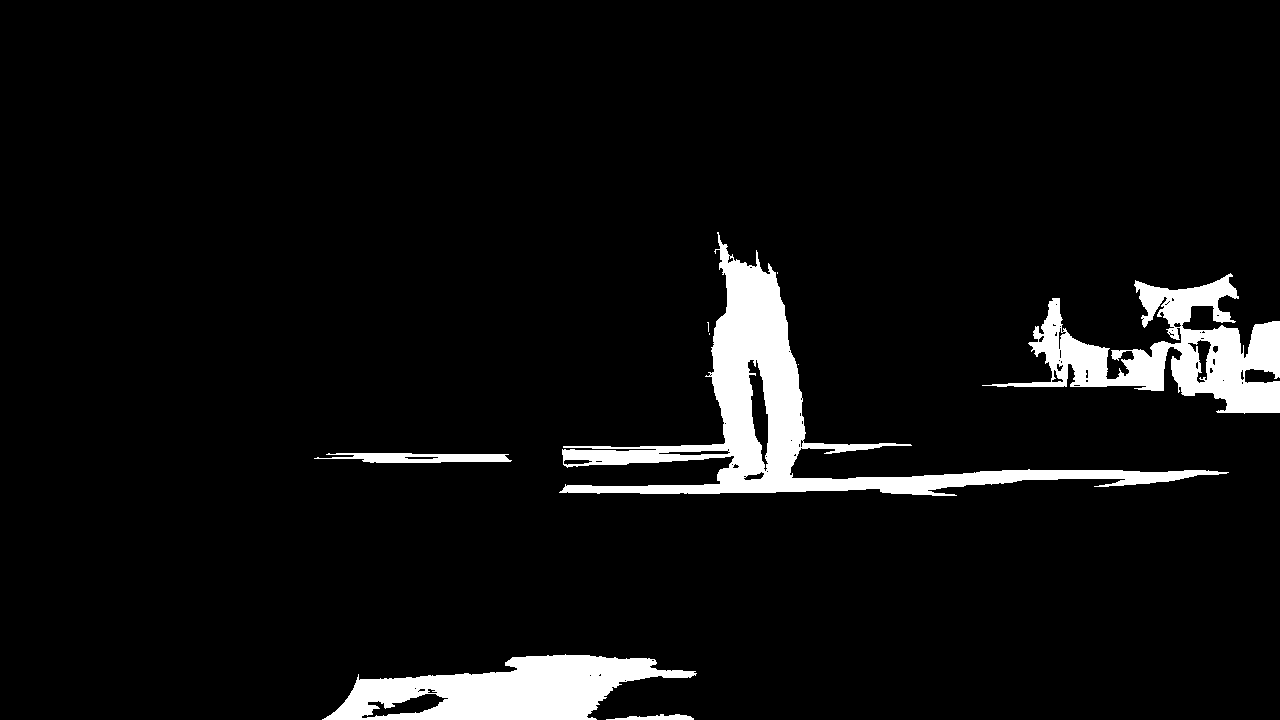}
	\end{subfigure}
	\begin{subfigure}{0.08\textwidth}
		\includegraphics[width=\textwidth]{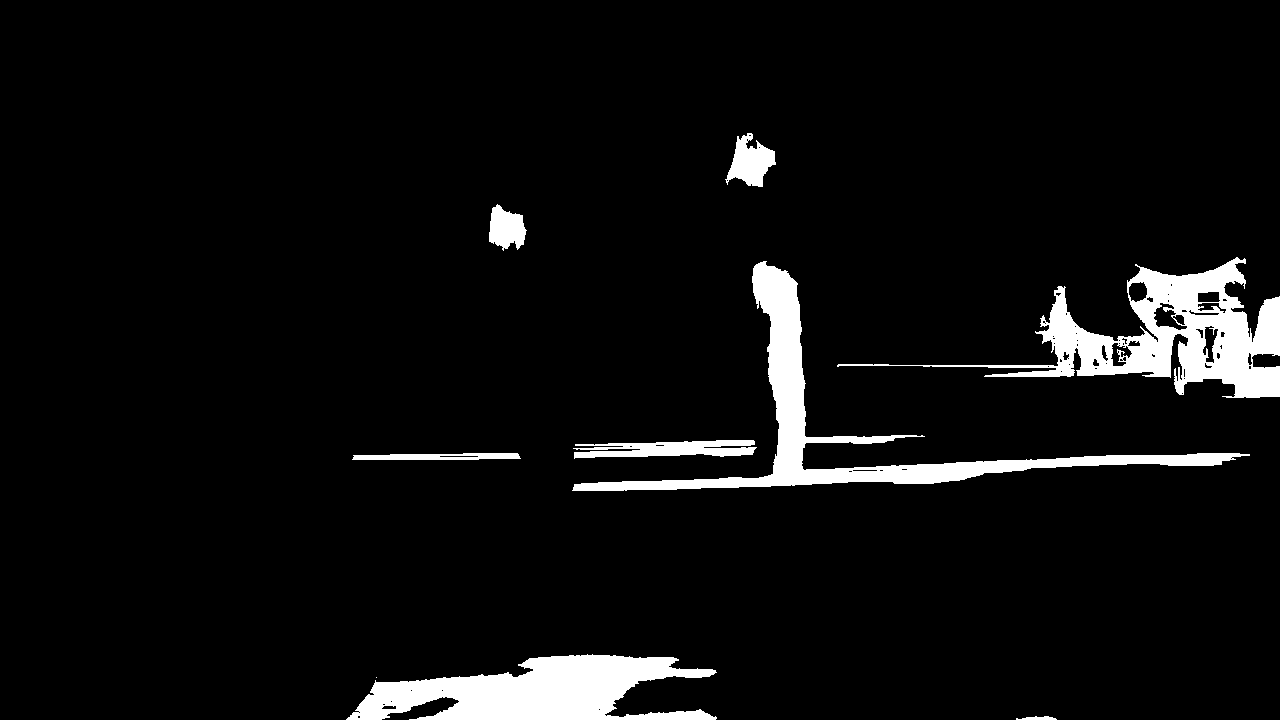}
	\end{subfigure}
	\begin{subfigure}{0.08\textwidth}
		\includegraphics[width=\textwidth]{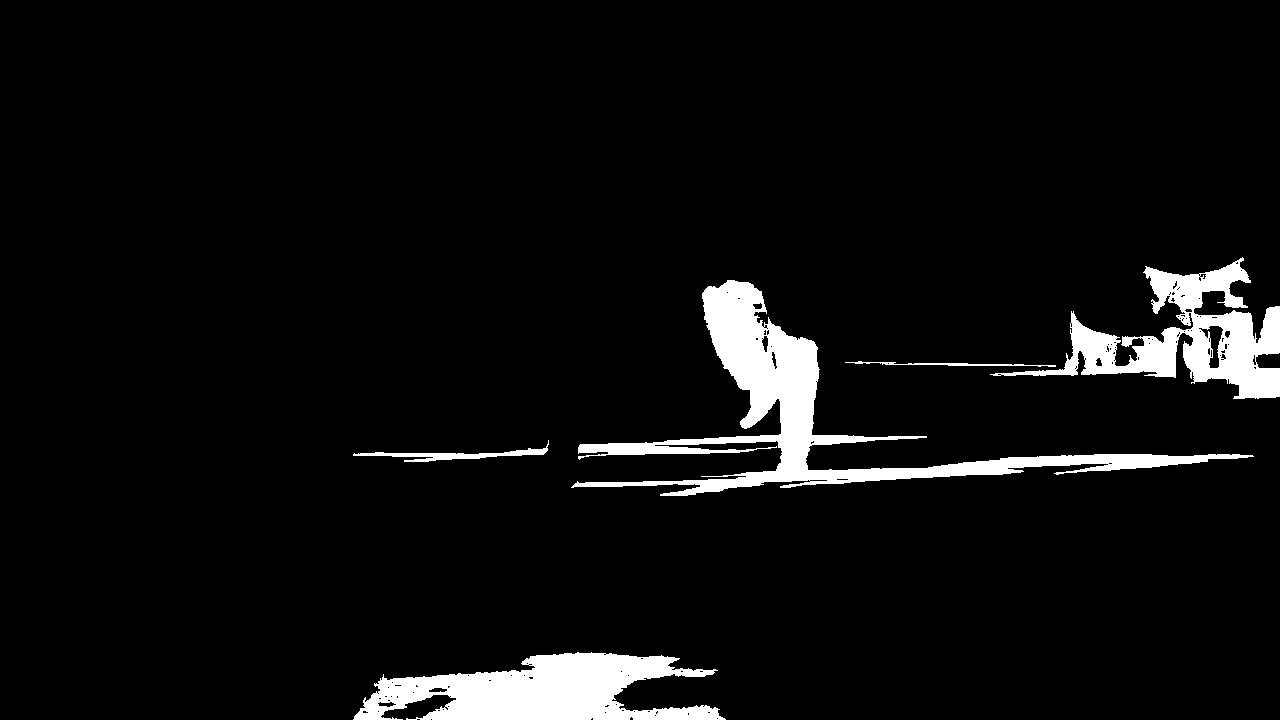}
	\end{subfigure}
	\begin{subfigure}{0.08\textwidth}
		\includegraphics[width=\textwidth]{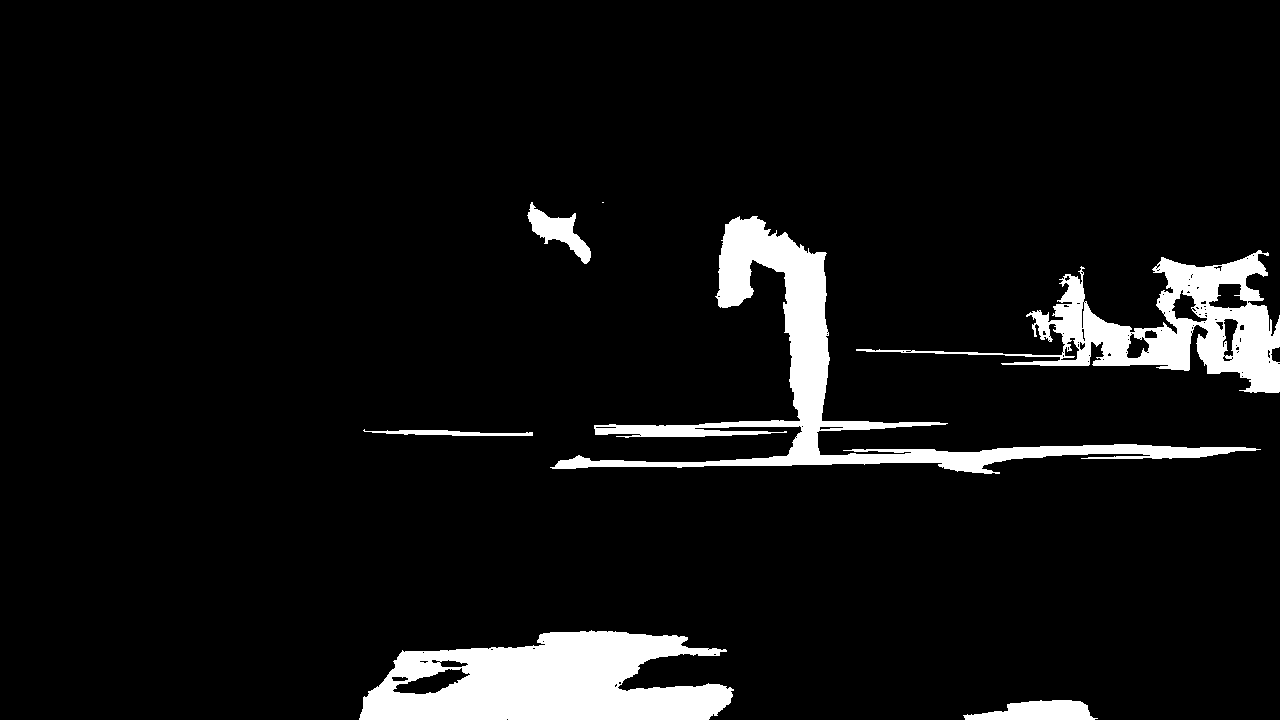}
	\end{subfigure}
	\begin{subfigure}{0.08\textwidth}
		\includegraphics[width=\textwidth]{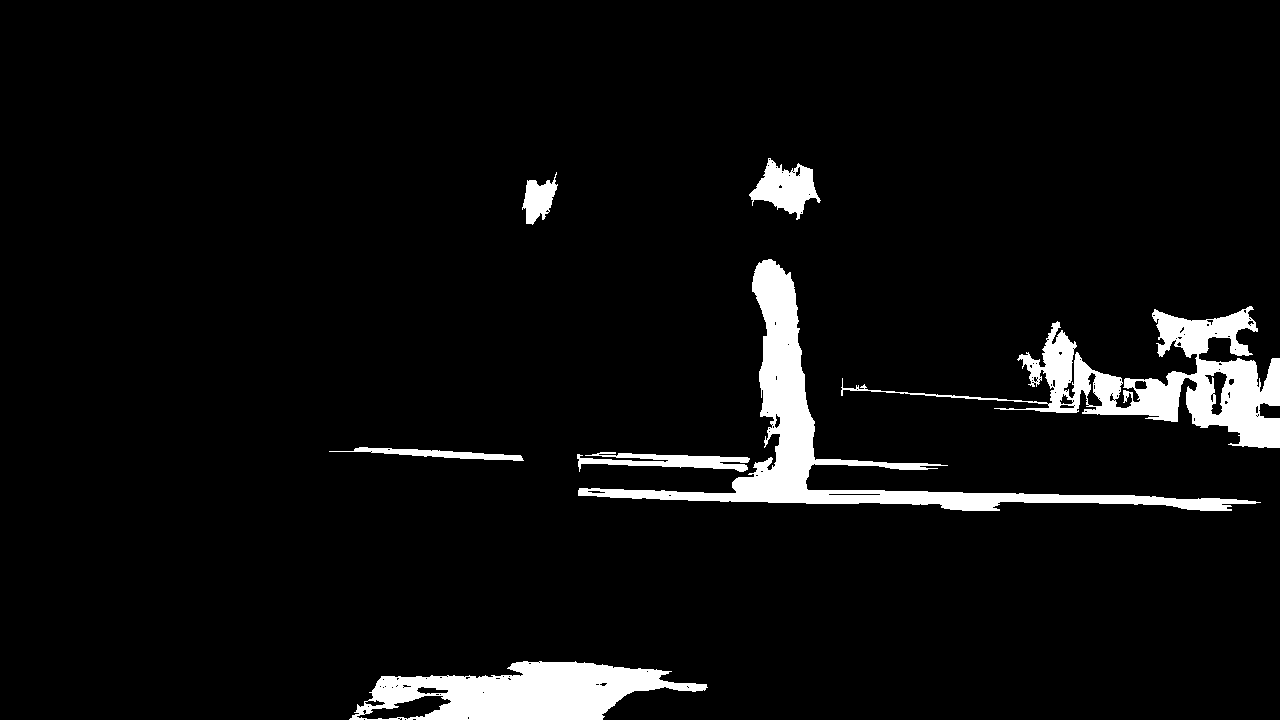}
	\end{subfigure}
	\begin{subfigure}{0.08\textwidth}
		\includegraphics[width=\textwidth]{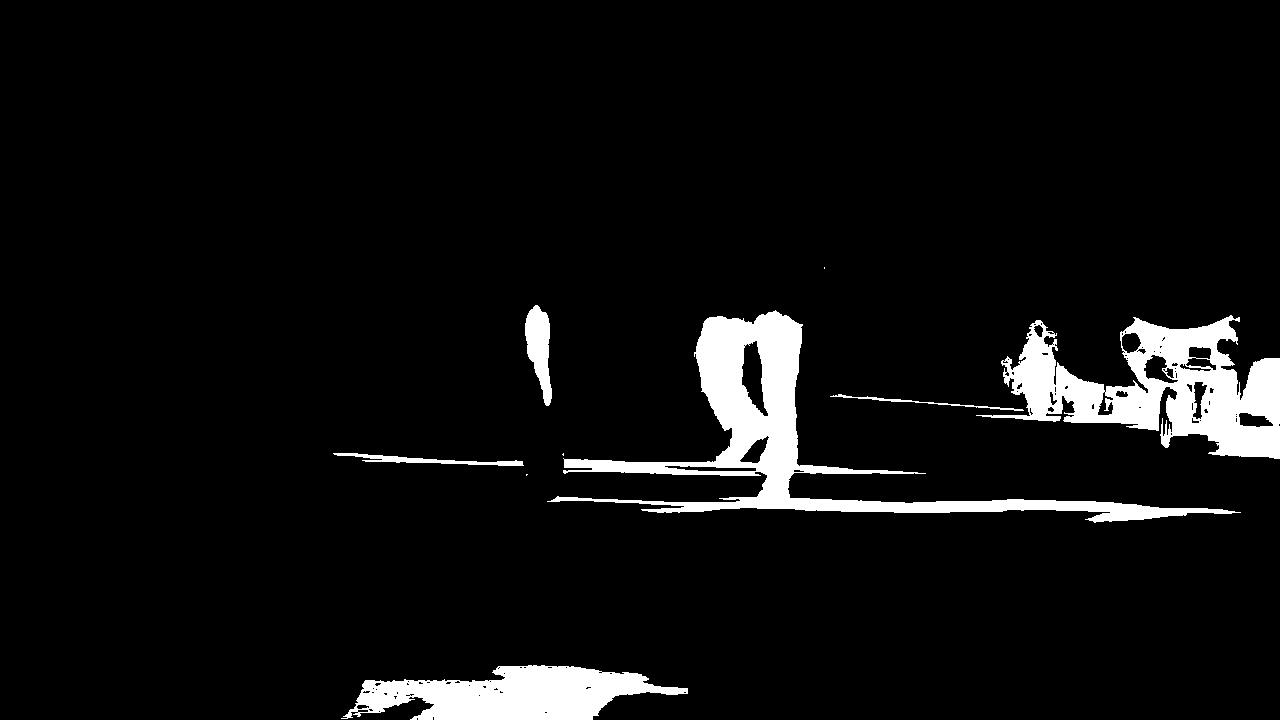}
	\end{subfigure}
	\begin{subfigure}{0.08\textwidth}
		\includegraphics[width=\textwidth]{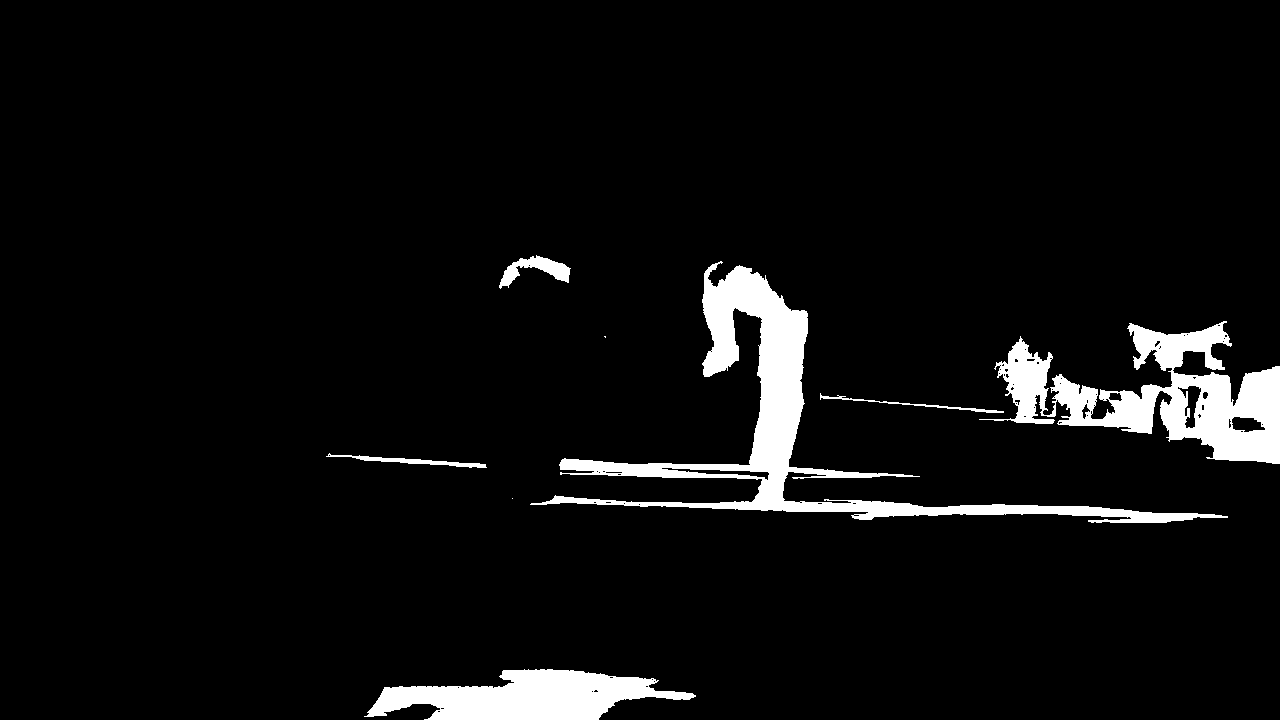}
	\end{subfigure}
	\begin{subfigure}{0.08\textwidth}
		\includegraphics[width=\textwidth]{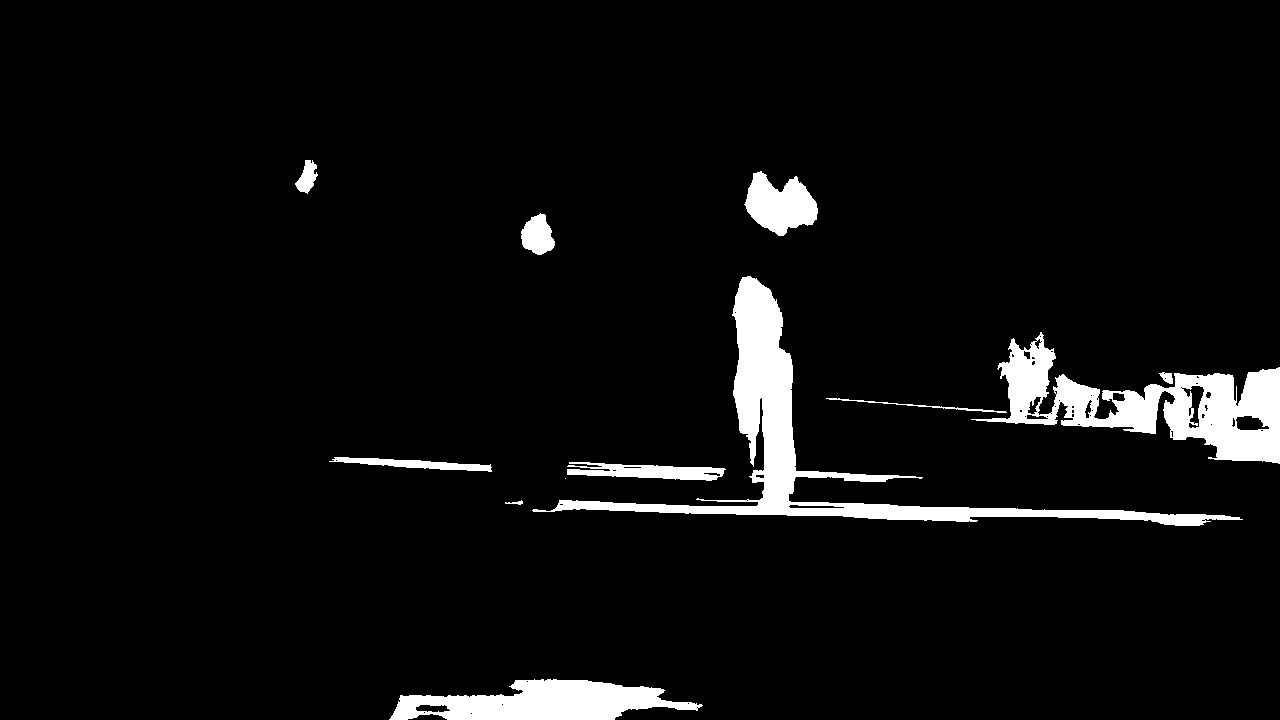}
	\end{subfigure}

	\vspace*{1.3mm}
	\begin{subfigure}{0.08\textwidth}
		\includegraphics[width=\textwidth]{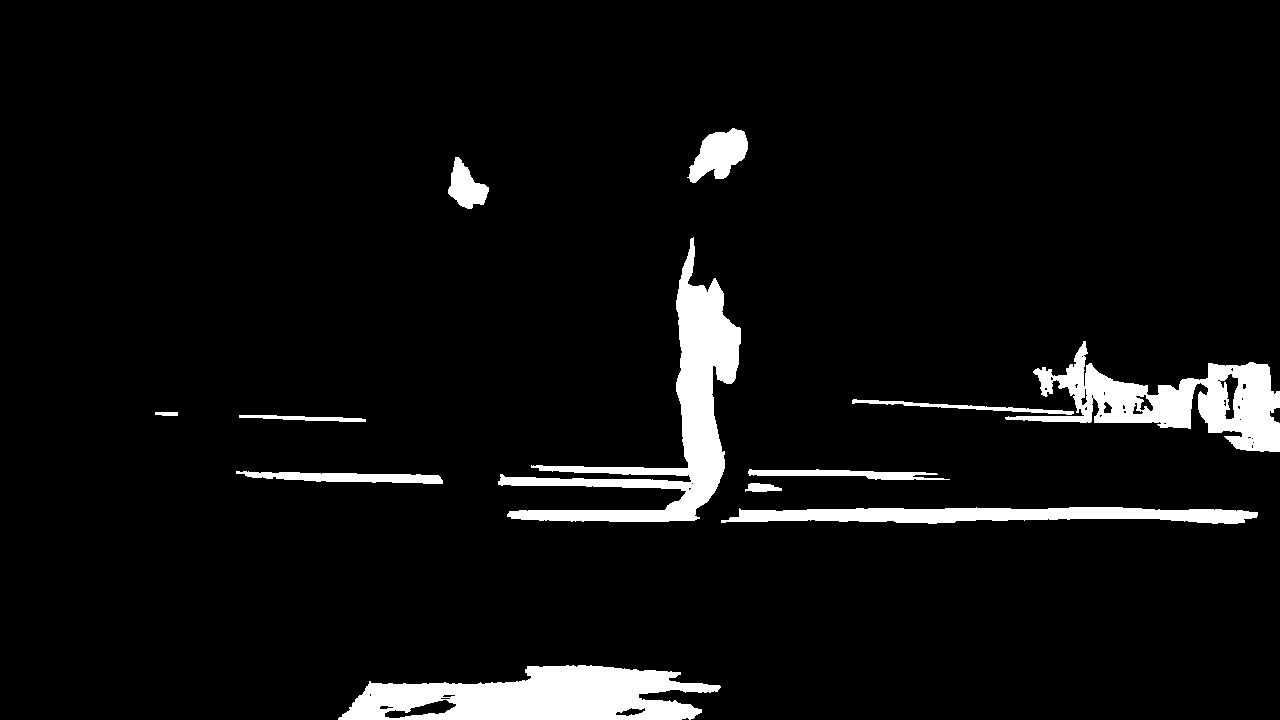}
	\end{subfigure}
	\begin{subfigure}{0.08\textwidth}
		\includegraphics[width=\textwidth]{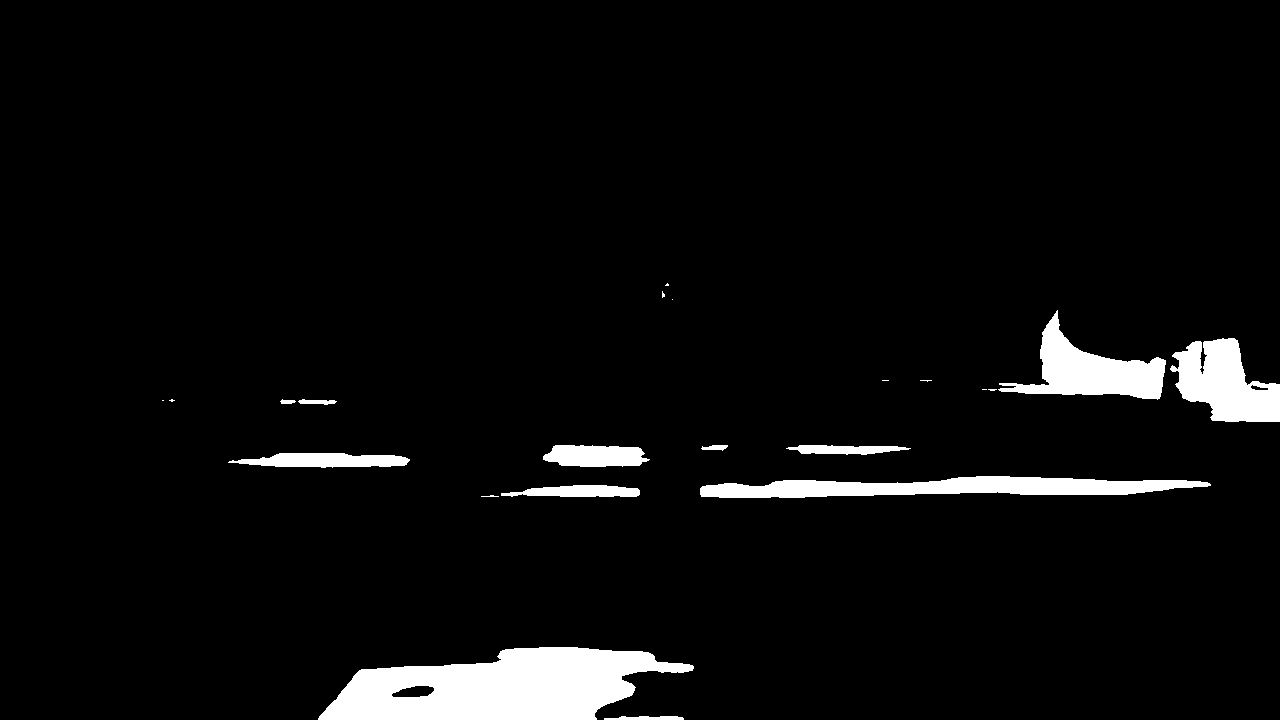}
	\end{subfigure}
	\begin{subfigure}{0.08\textwidth}
		\includegraphics[width=\textwidth]{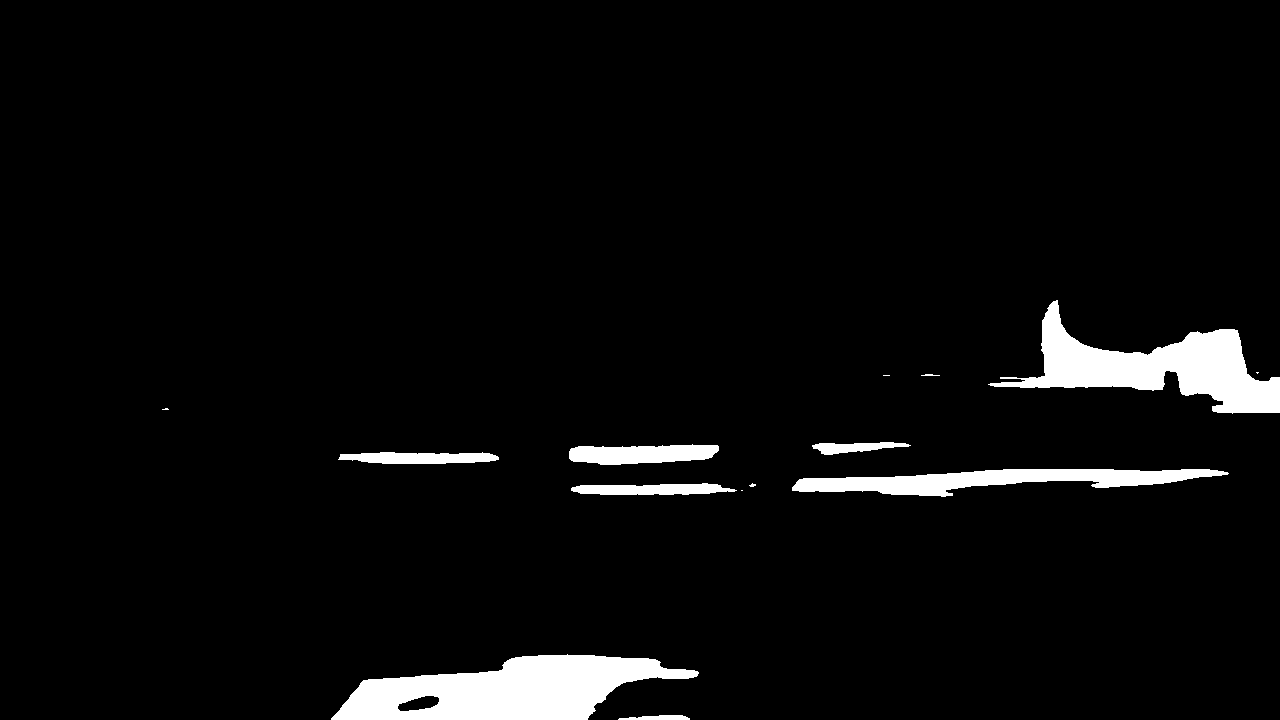}
	\end{subfigure}
	\begin{subfigure}{0.08\textwidth}
		\includegraphics[width=\textwidth]{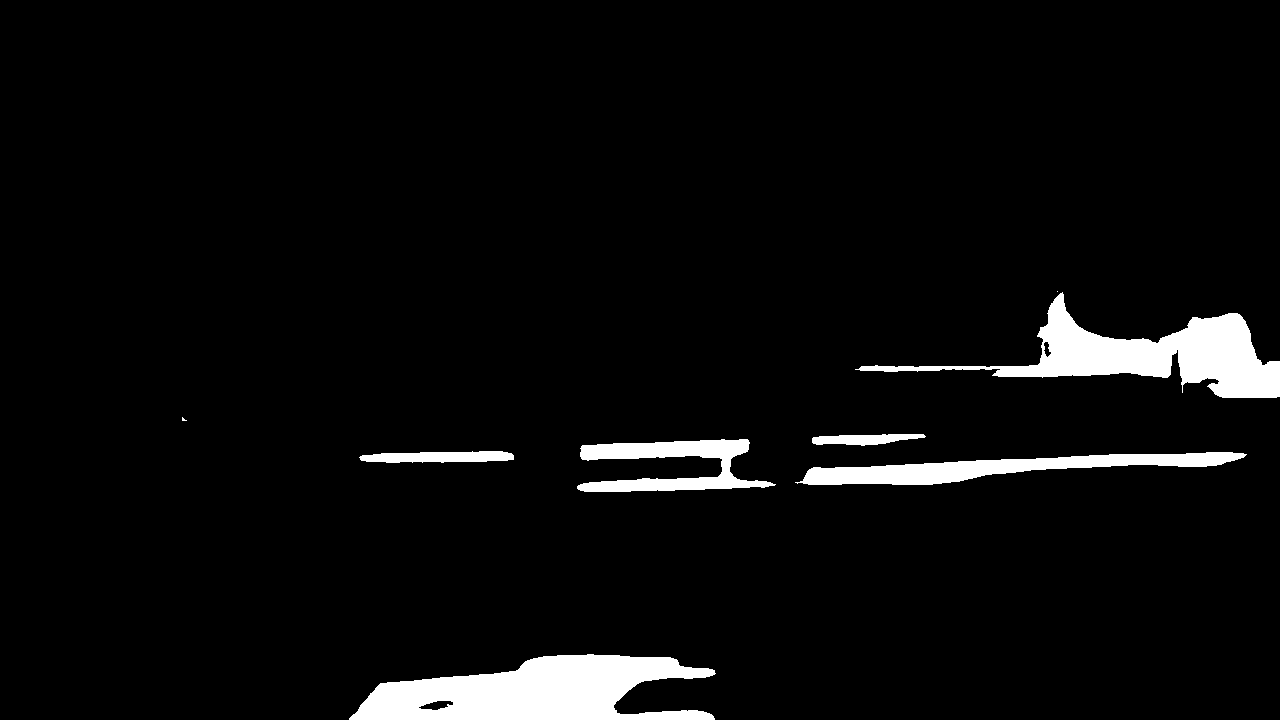}
	\end{subfigure}
	\begin{subfigure}{0.08\textwidth}
		\includegraphics[width=\textwidth]{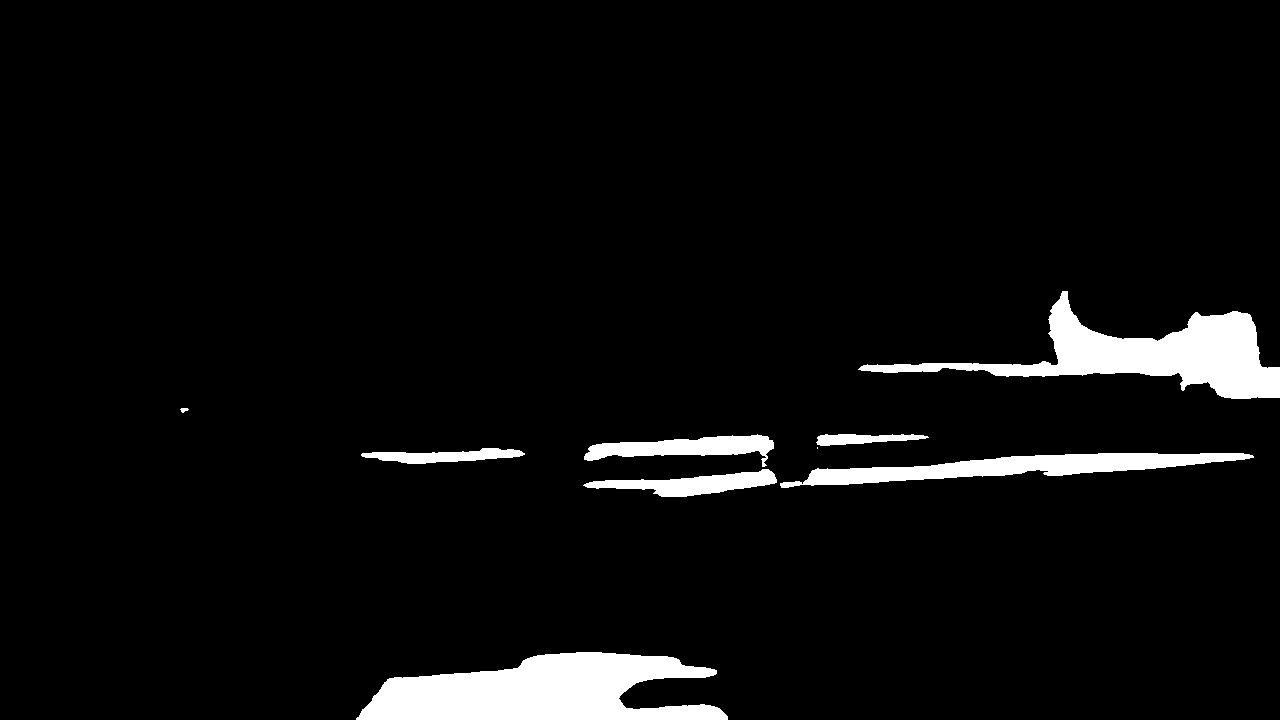}
	\end{subfigure}
	\begin{subfigure}{0.08\textwidth}
		\includegraphics[width=\textwidth]{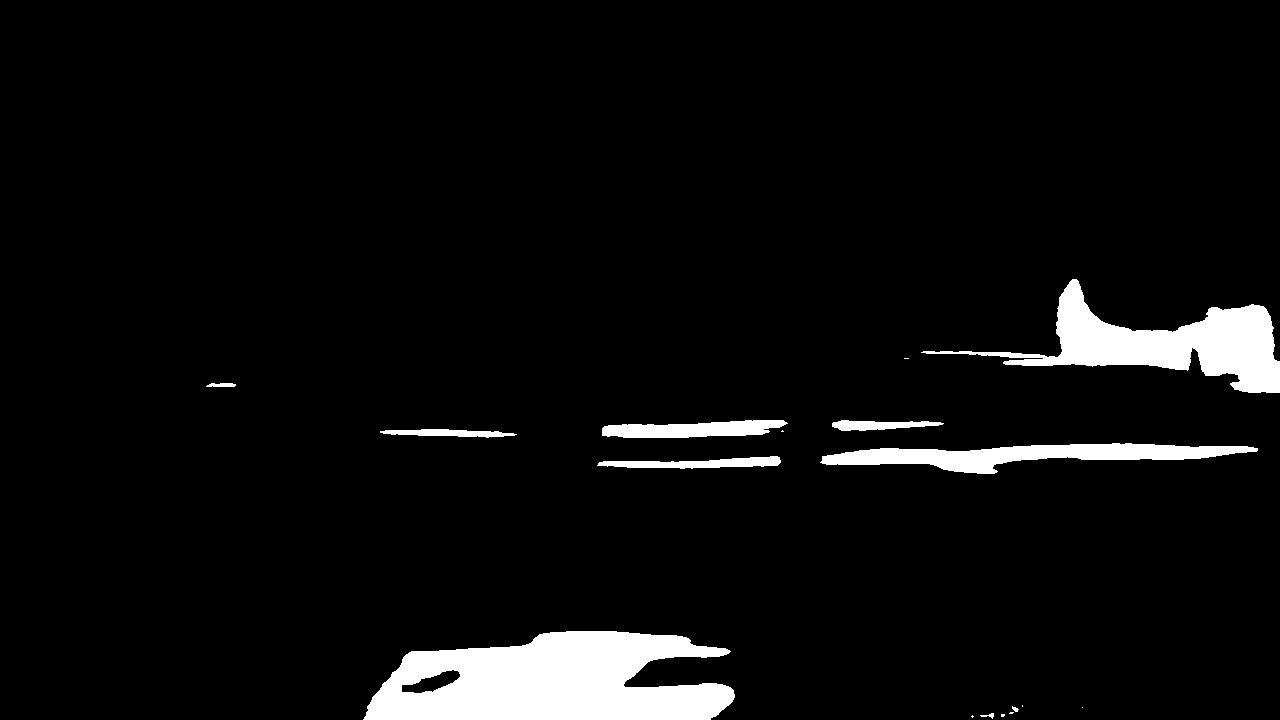}
	\end{subfigure}
	\begin{subfigure}{0.08\textwidth}
		\includegraphics[width=\textwidth]{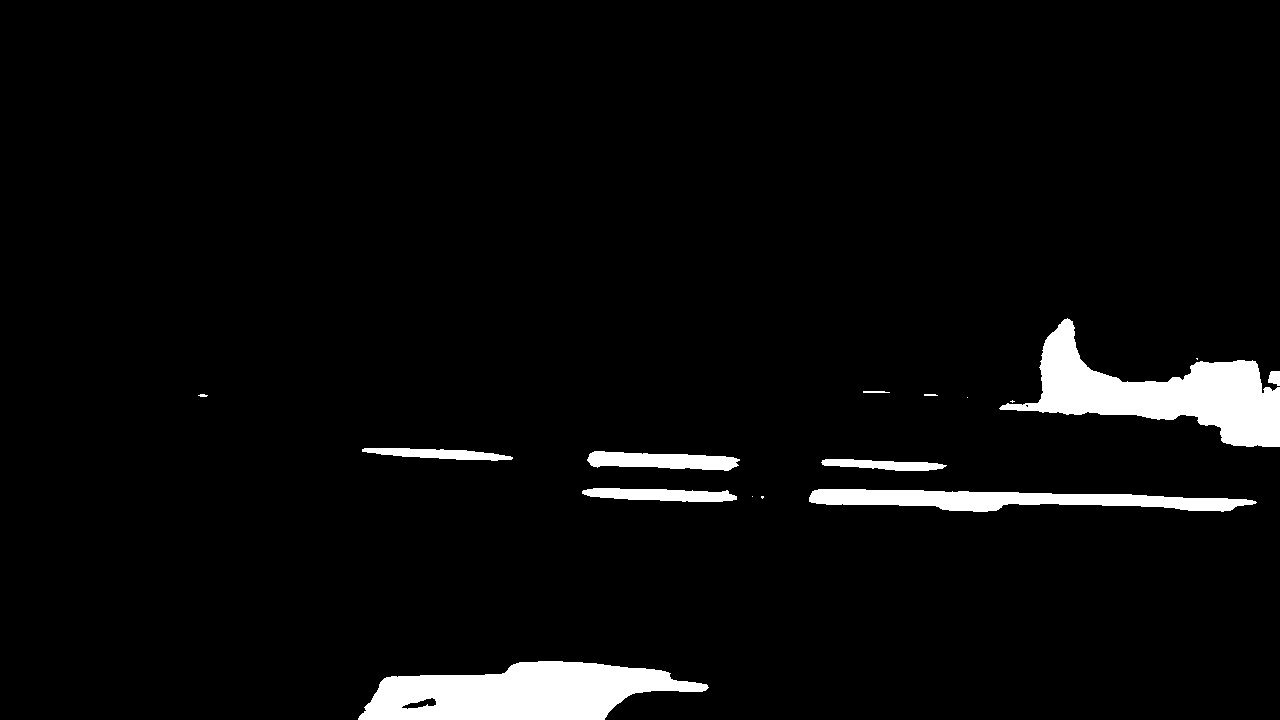}
	\end{subfigure}
	\begin{subfigure}{0.08\textwidth}
		\includegraphics[width=\textwidth]{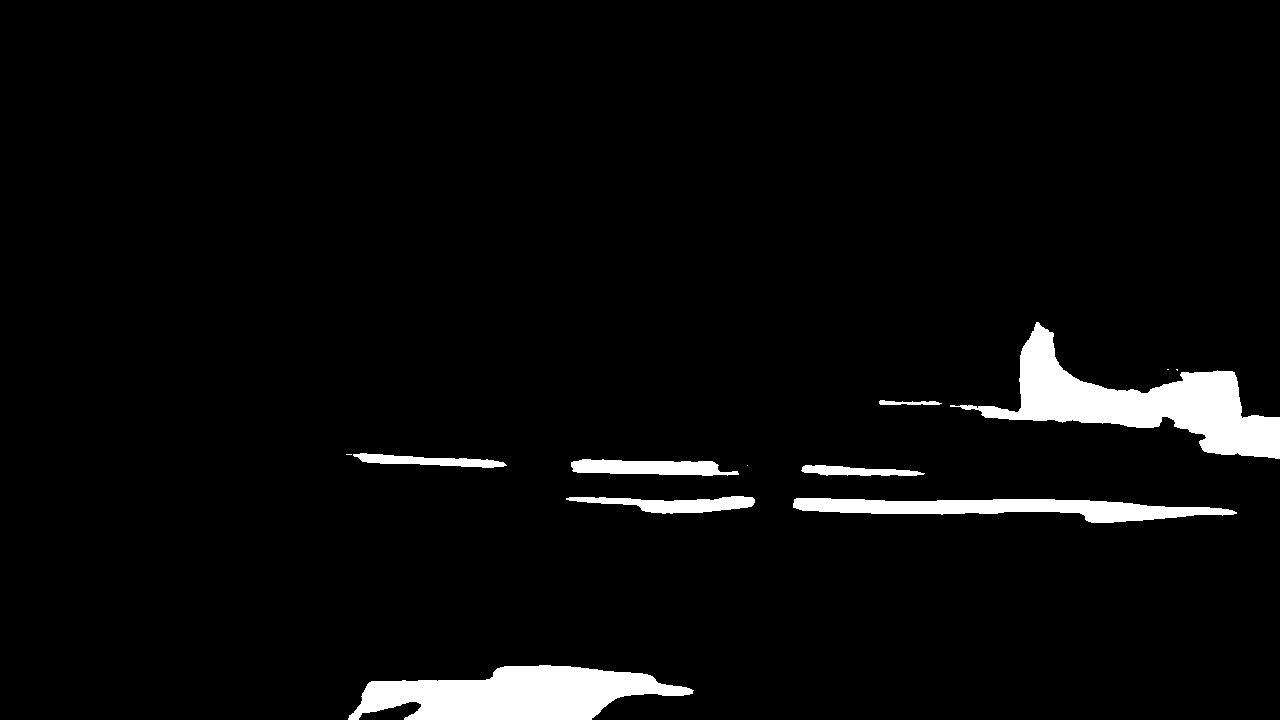}
	\end{subfigure}
	\begin{subfigure}{0.08\textwidth}
		\includegraphics[width=\textwidth]{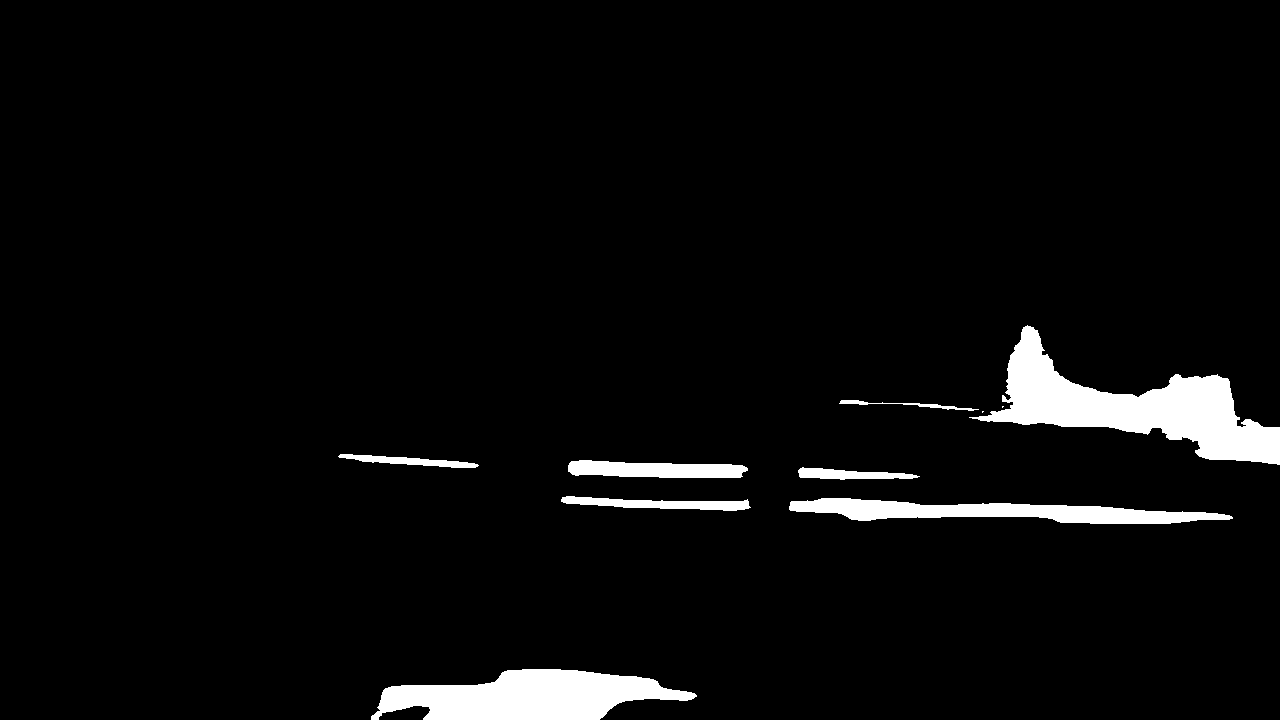}
	\end{subfigure}
	\begin{subfigure}{0.08\textwidth}
		\includegraphics[width=\textwidth]{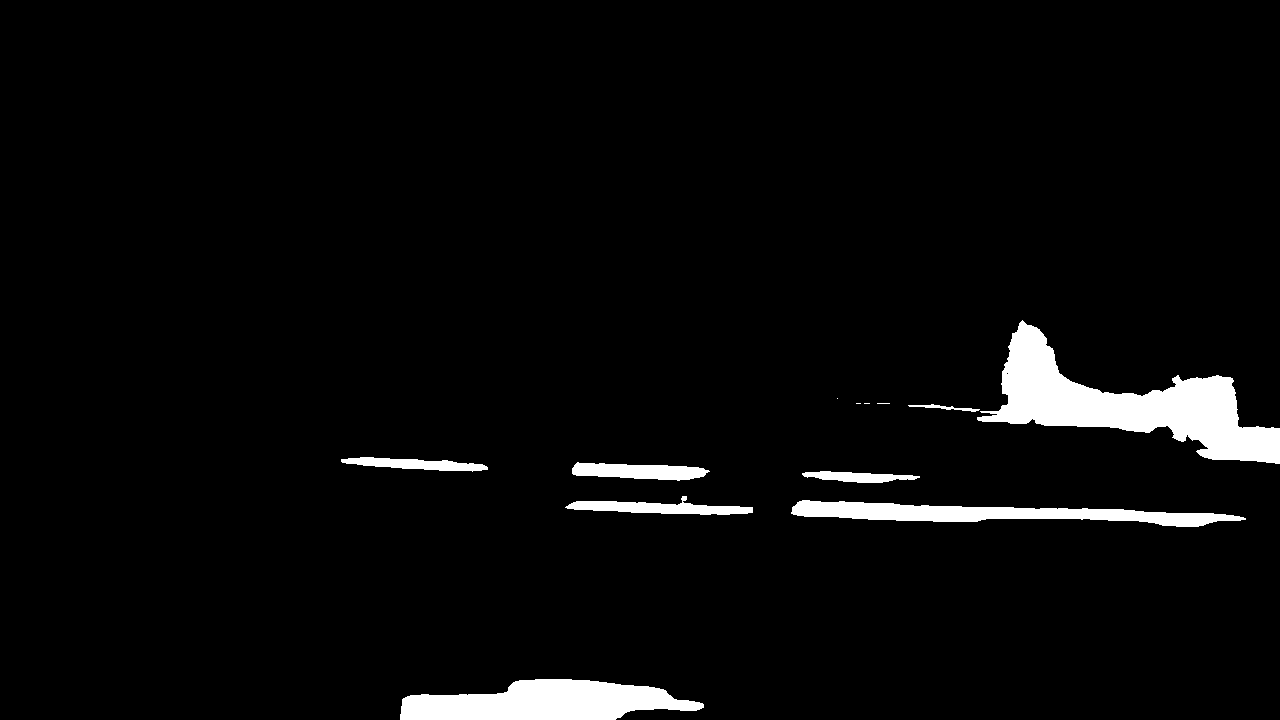}
	\end{subfigure}

	\vspace*{1.3mm}
	\begin{subfigure}{0.08\textwidth}
		\includegraphics[width=\textwidth]{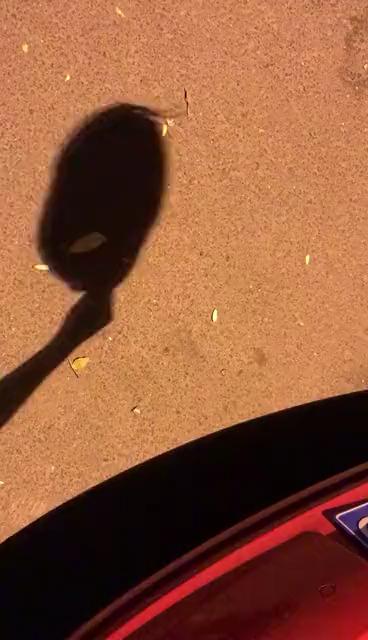}
	\end{subfigure}
	\begin{subfigure}{0.08\textwidth}
		\includegraphics[width=\textwidth]{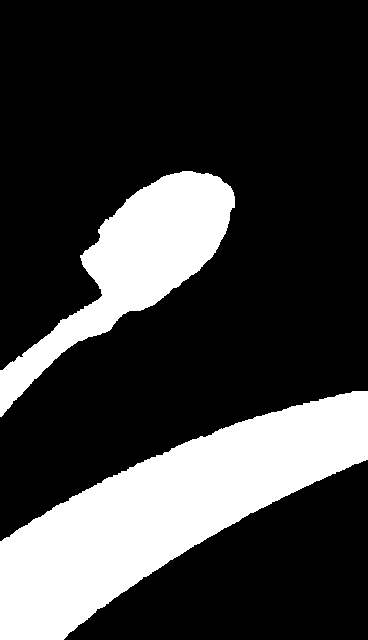}
	\end{subfigure}
	\begin{subfigure}{0.08\textwidth}
		\includegraphics[width=\textwidth]{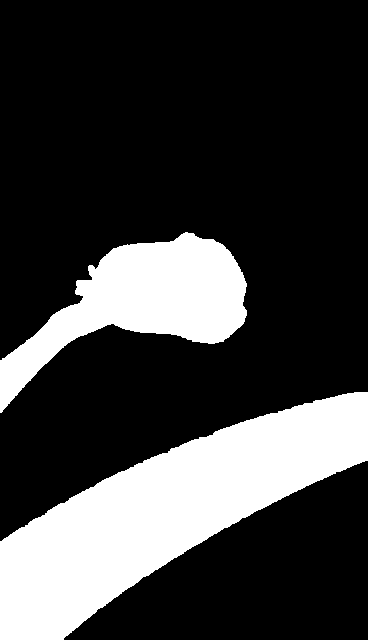}
	\end{subfigure}
	\begin{subfigure}{0.08\textwidth}
		\includegraphics[width=\textwidth]{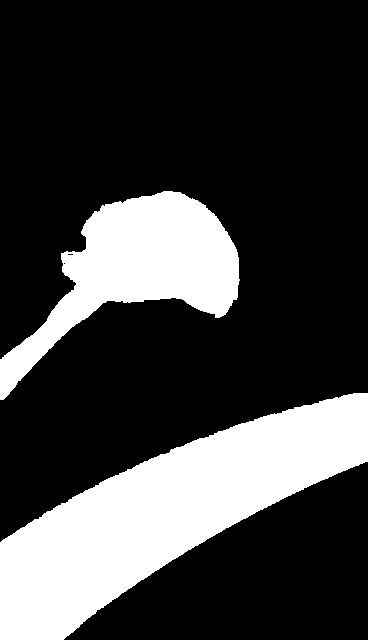}
	\end{subfigure}
	\begin{subfigure}{0.08\textwidth}
		\includegraphics[width=\textwidth]{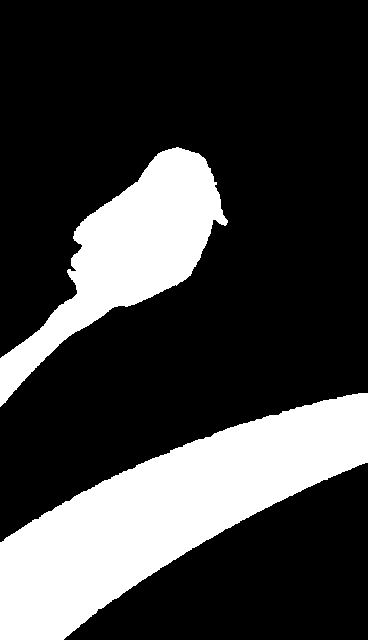}
	\end{subfigure}
	\begin{subfigure}{0.08\textwidth}
		\includegraphics[width=\textwidth]{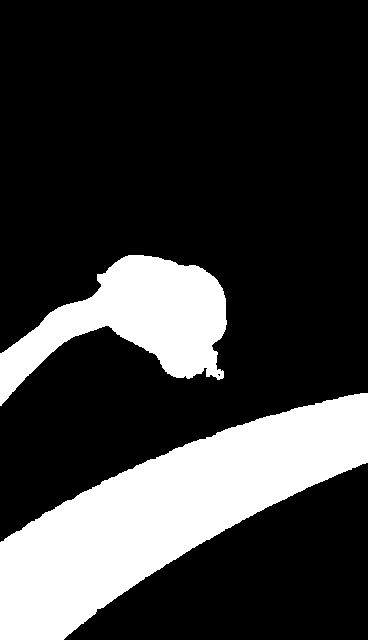}
	\end{subfigure}
	\begin{subfigure}{0.08\textwidth}
		\includegraphics[width=\textwidth]{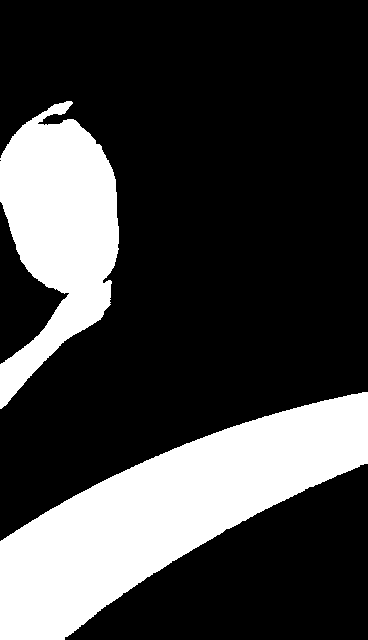}
	\end{subfigure}
	\begin{subfigure}{0.08\textwidth}
		\includegraphics[width=\textwidth]{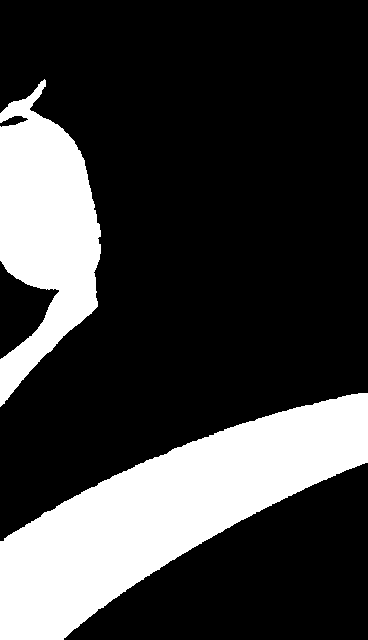}
	\end{subfigure}
	\begin{subfigure}{0.08\textwidth}
		\includegraphics[width=\textwidth]{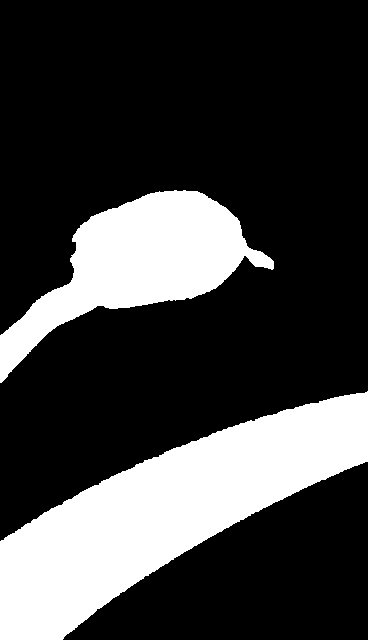}
	\end{subfigure}
	\begin{subfigure}{0.08\textwidth}
		\includegraphics[width=\textwidth]{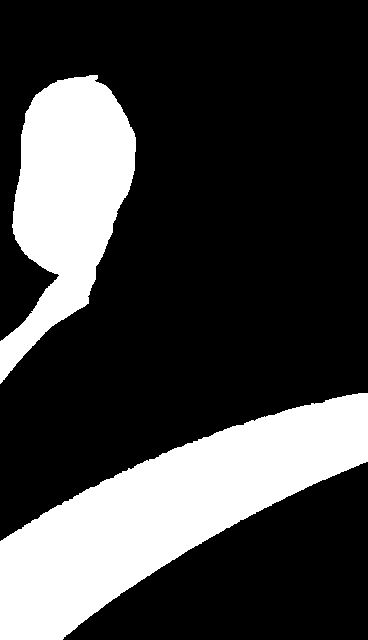}
	\end{subfigure}
	
	\vspace*{1.3mm}
	\begin{subfigure}{0.08\textwidth}
		\includegraphics[width=\textwidth]{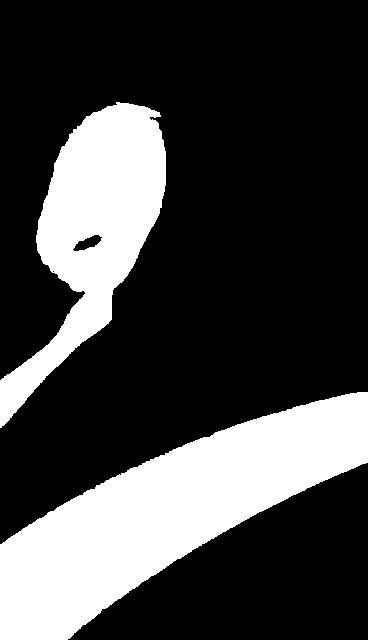}
	\end{subfigure}
	\begin{subfigure}{0.08\textwidth}
		\includegraphics[width=\textwidth]{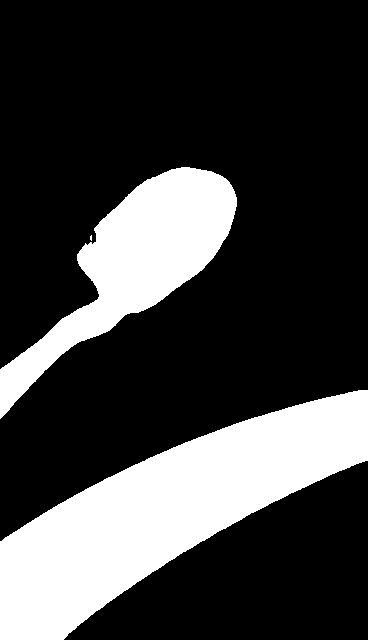}
	\end{subfigure}
	\begin{subfigure}{0.08\textwidth}
		\includegraphics[width=\textwidth]{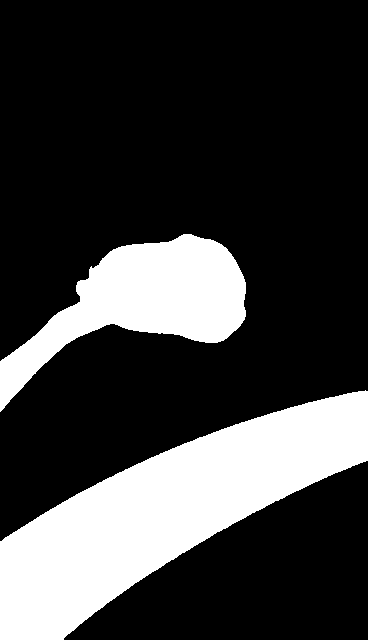}
	\end{subfigure}
	\begin{subfigure}{0.08\textwidth}
		\includegraphics[width=\textwidth]{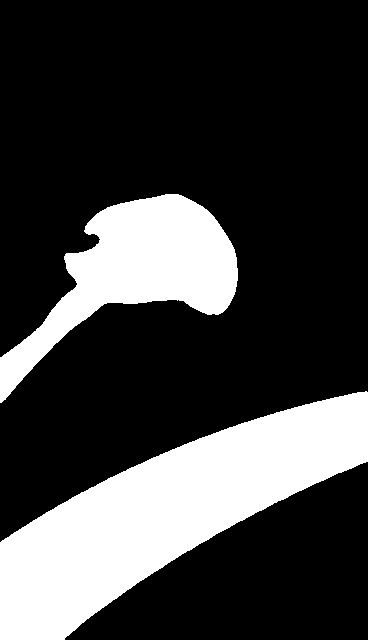}
	\end{subfigure}
	\begin{subfigure}{0.08\textwidth}
		\includegraphics[width=\textwidth]{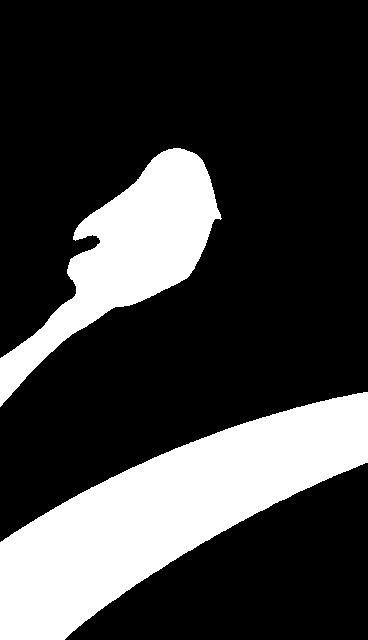}
	\end{subfigure}
	\begin{subfigure}{0.08\textwidth}
		\includegraphics[width=\textwidth]{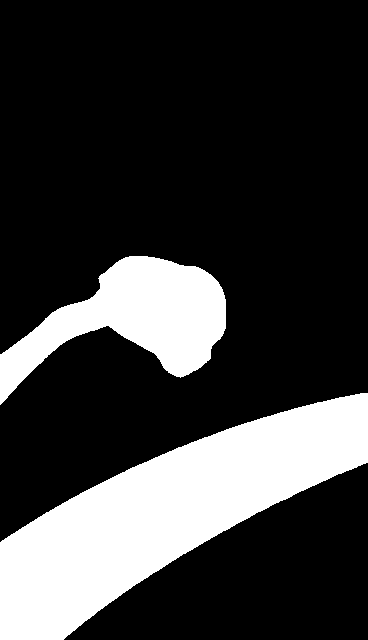}
	\end{subfigure}
	\begin{subfigure}{0.08\textwidth}
		\includegraphics[width=\textwidth]{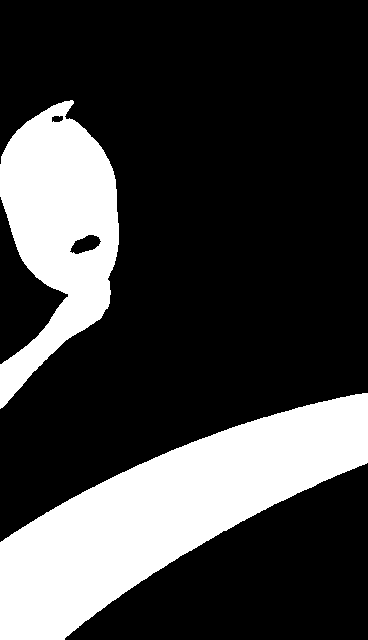}
	\end{subfigure}
	\begin{subfigure}{0.08\textwidth}
		\includegraphics[width=\textwidth]{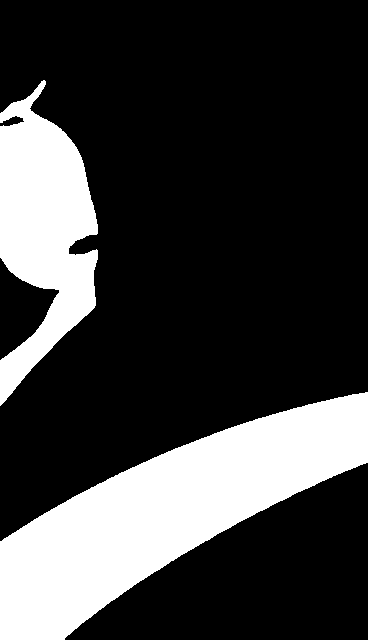}
	\end{subfigure}
	\begin{subfigure}{0.08\textwidth}
		\includegraphics[width=\textwidth]{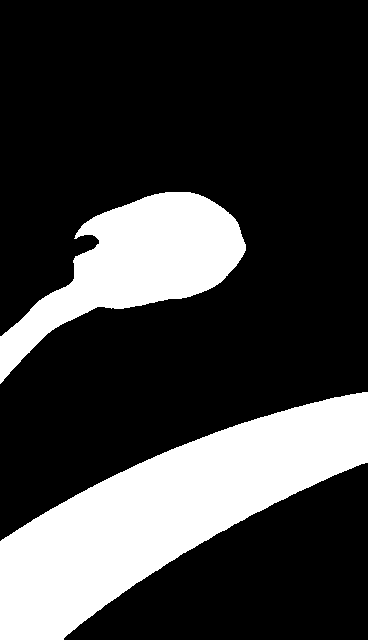}
	\end{subfigure}
	\begin{subfigure}{0.08\textwidth}
		\includegraphics[width=\textwidth]{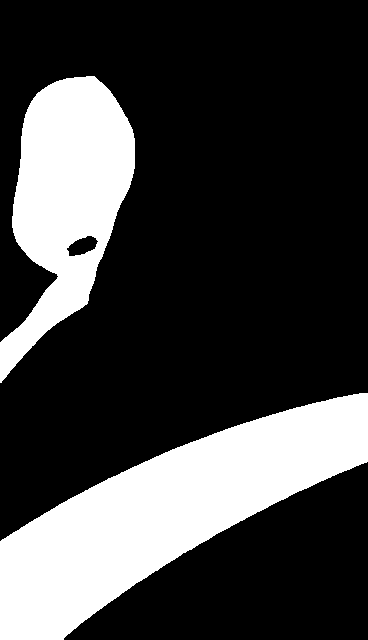}
	\end{subfigure}

	\vspace*{1.3mm}
	\begin{subfigure}{0.08\textwidth}
		\includegraphics[width=\textwidth]{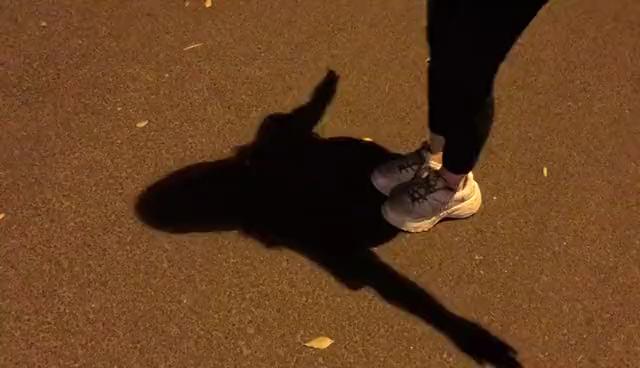}
	\end{subfigure}
	\begin{subfigure}{0.08\textwidth}
		\includegraphics[width=\textwidth]{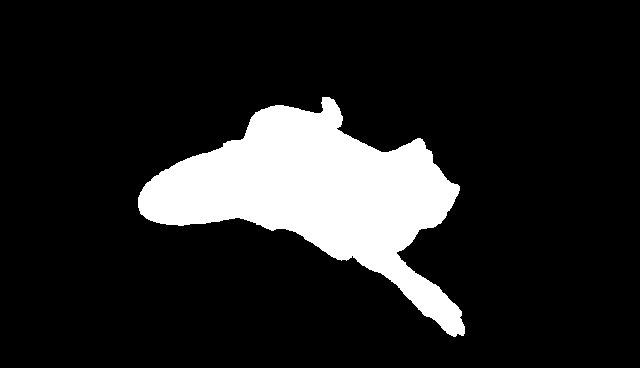}
	\end{subfigure}
	\begin{subfigure}{0.08\textwidth}
		\includegraphics[width=\textwidth]{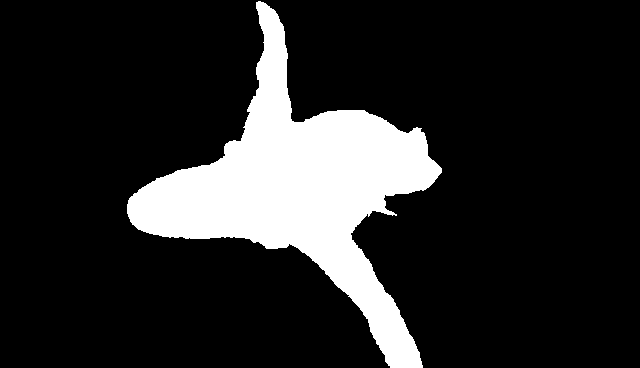}
	\end{subfigure}
	\begin{subfigure}{0.08\textwidth}
		\includegraphics[width=\textwidth]{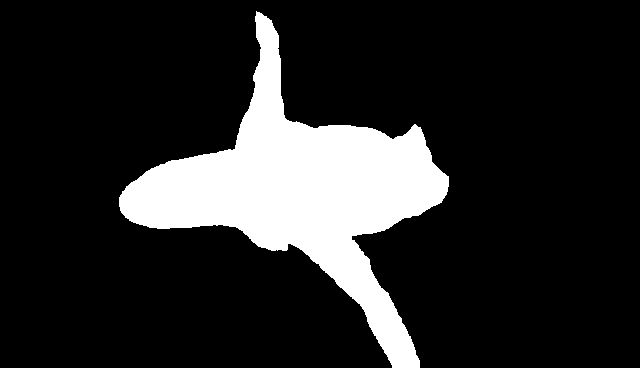}
	\end{subfigure}
	\begin{subfigure}{0.08\textwidth}
		\includegraphics[width=\textwidth]{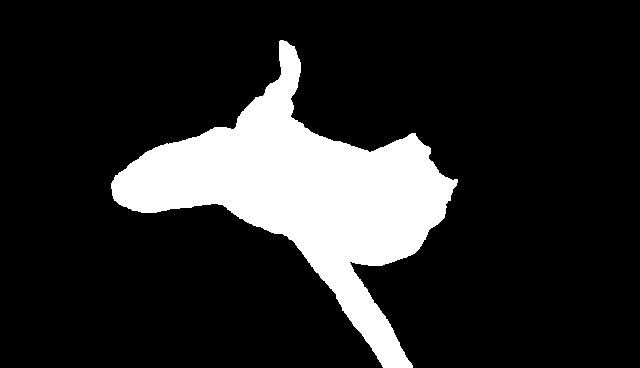}
	\end{subfigure}
	\begin{subfigure}{0.08\textwidth}
		\includegraphics[width=\textwidth]{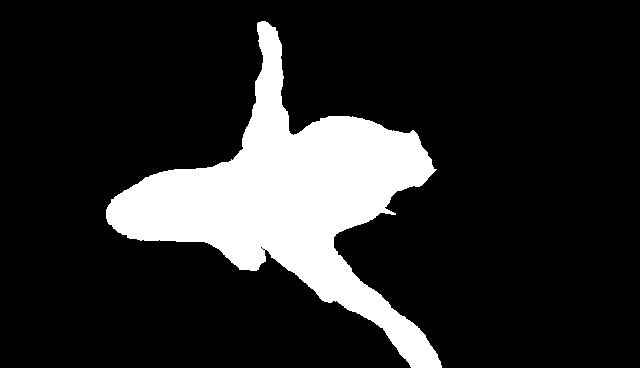}
	\end{subfigure}
	\begin{subfigure}{0.08\textwidth}
		\includegraphics[width=\textwidth]{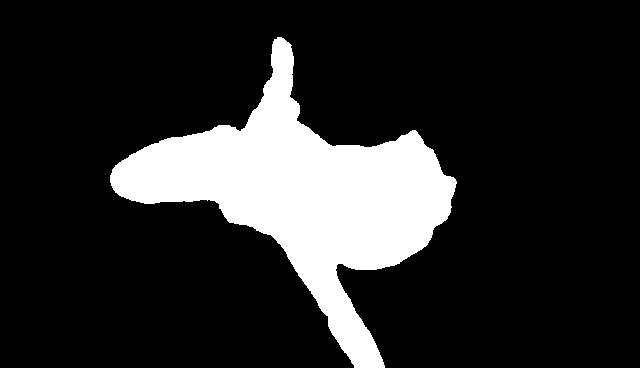}
	\end{subfigure}
	\begin{subfigure}{0.08\textwidth}
		\includegraphics[width=\textwidth]{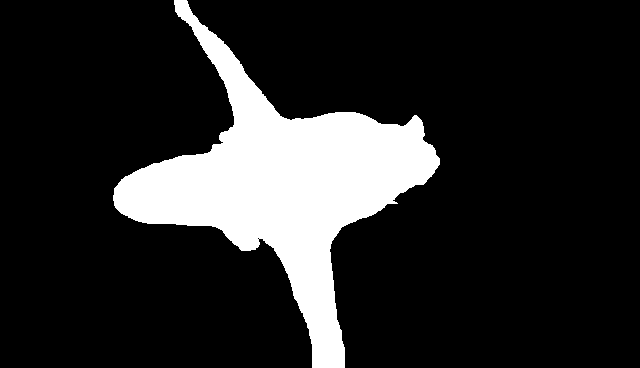}
	\end{subfigure}
	\begin{subfigure}{0.08\textwidth}
		\includegraphics[width=\textwidth]{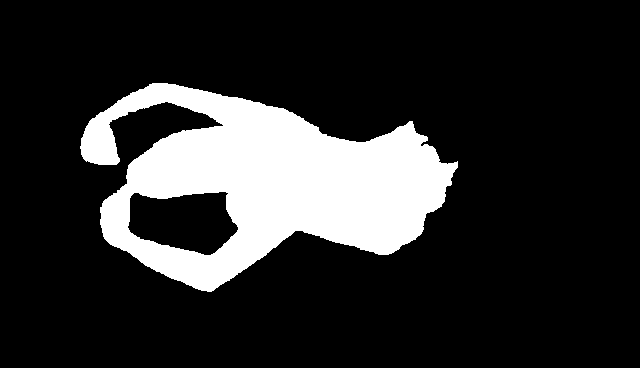}
	\end{subfigure}
	\begin{subfigure}{0.08\textwidth}
		\includegraphics[width=\textwidth]{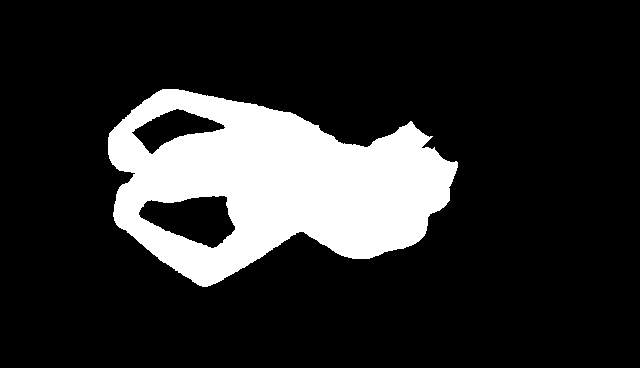}
	\end{subfigure}
	
	\vspace*{1.3mm}
	\begin{subfigure}{0.08\textwidth}
		\includegraphics[width=\textwidth]{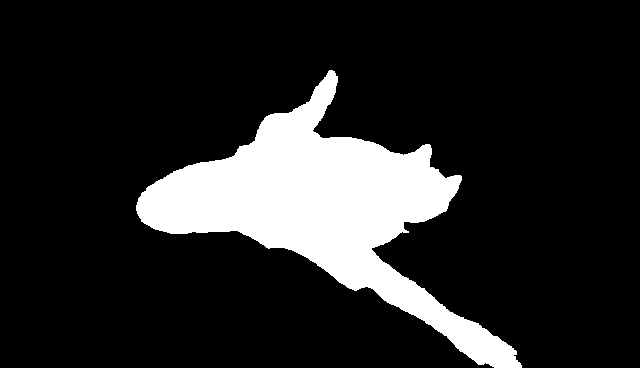}
	\end{subfigure}
	\begin{subfigure}{0.08\textwidth}
		\includegraphics[width=\textwidth]{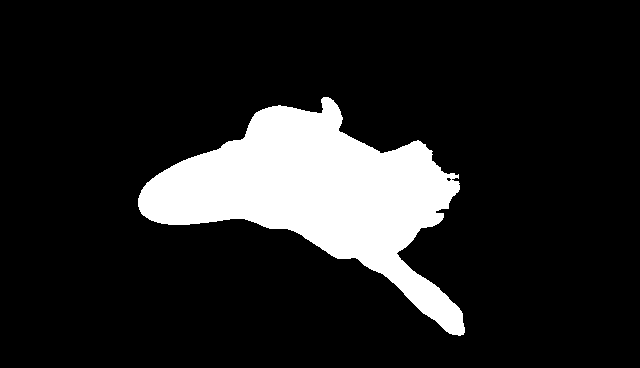}
	\end{subfigure}
	\begin{subfigure}{0.08\textwidth}
		\includegraphics[width=\textwidth]{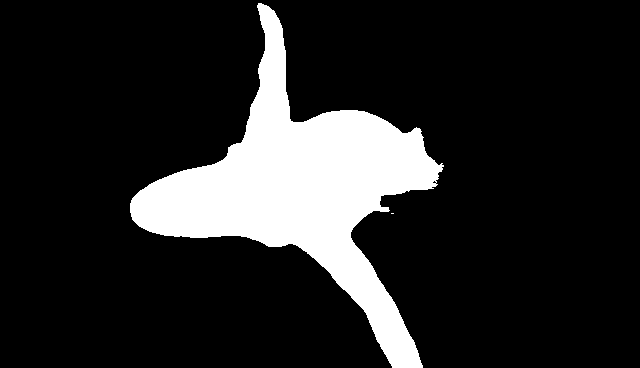}
	\end{subfigure}
	\begin{subfigure}{0.08\textwidth}
		\includegraphics[width=\textwidth]{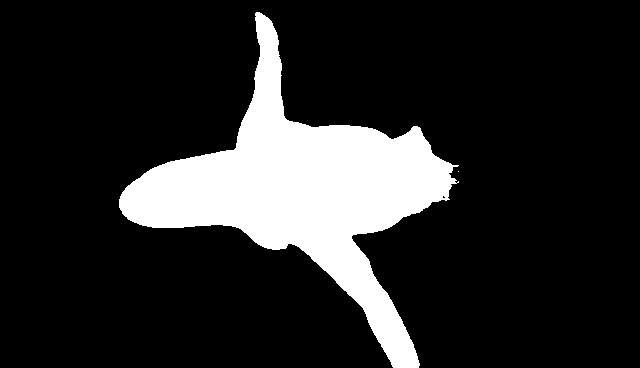}
	\end{subfigure}
	\begin{subfigure}{0.08\textwidth}
		\includegraphics[width=\textwidth]{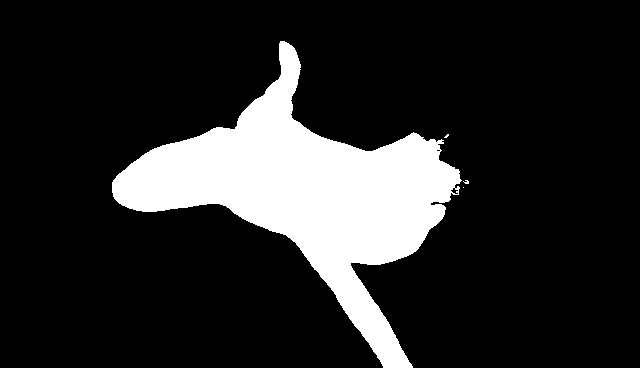}
	\end{subfigure}
	\begin{subfigure}{0.08\textwidth}
		\includegraphics[width=\textwidth]{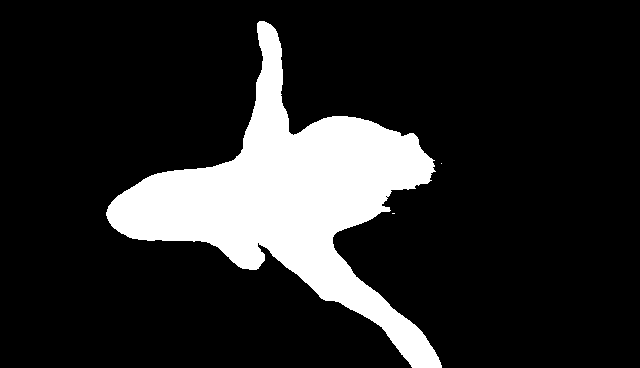}
	\end{subfigure}
	\begin{subfigure}{0.08\textwidth}
		\includegraphics[width=\textwidth]{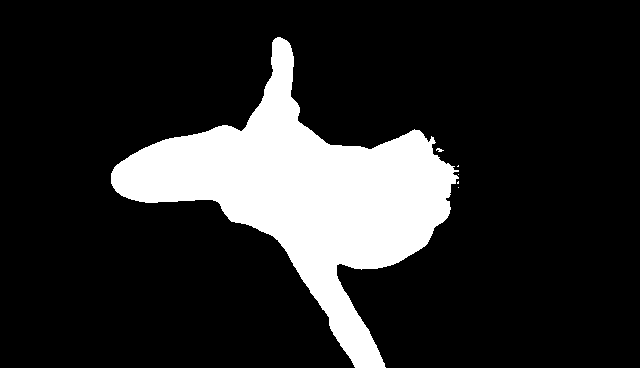}
	\end{subfigure}
	\begin{subfigure}{0.08\textwidth}
		\includegraphics[width=\textwidth]{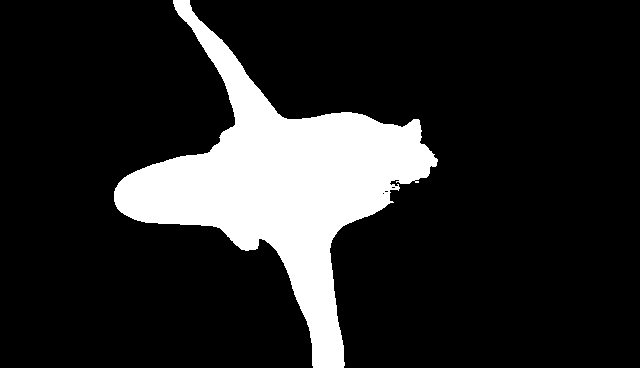}
	\end{subfigure}
	\begin{subfigure}{0.08\textwidth}
		\includegraphics[width=\textwidth]{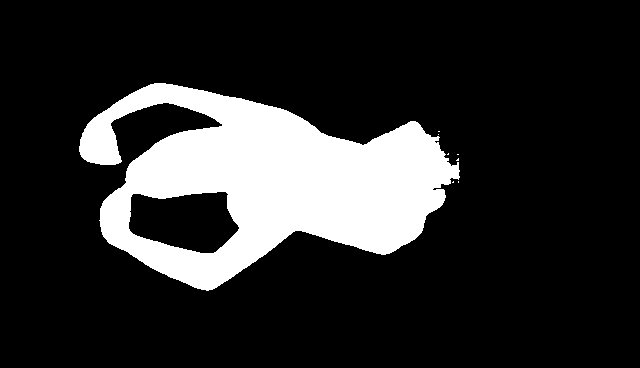}
	\end{subfigure}
	\begin{subfigure}{0.08\textwidth}
		\includegraphics[width=\textwidth]{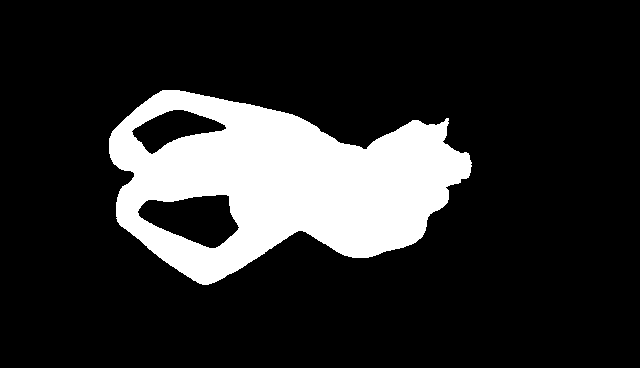}
	\end{subfigure}

	\vspace*{1.3mm}
	\begin{subfigure}{0.08\textwidth}
		\includegraphics[width=\textwidth]{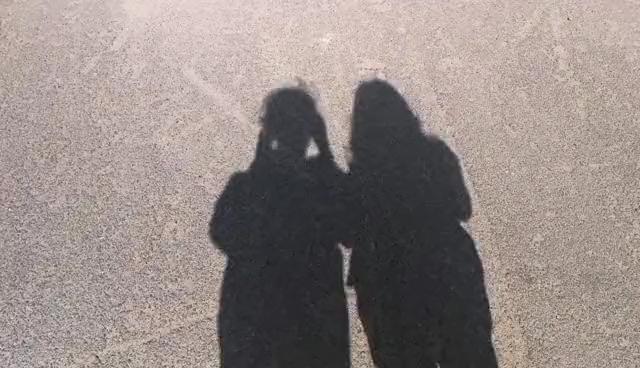}
	\end{subfigure}
	\begin{subfigure}{0.08\textwidth}
		\includegraphics[width=\textwidth]{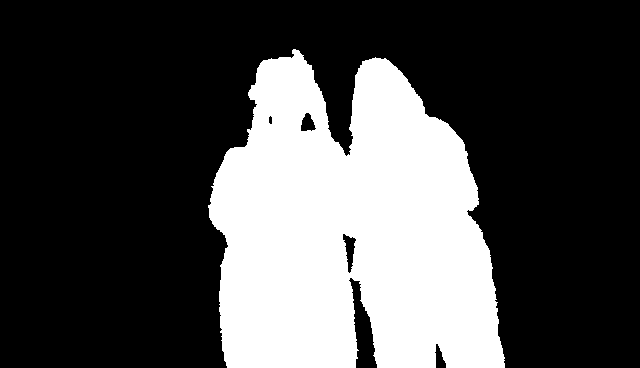}
	\end{subfigure}
	\begin{subfigure}{0.08\textwidth}
		\includegraphics[width=\textwidth]{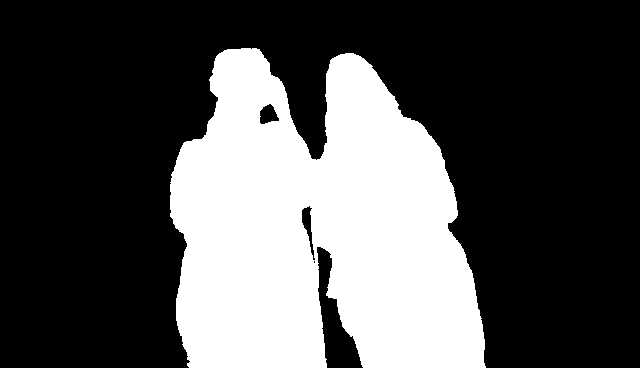}
	\end{subfigure}
	\begin{subfigure}{0.08\textwidth}
		\includegraphics[width=\textwidth]{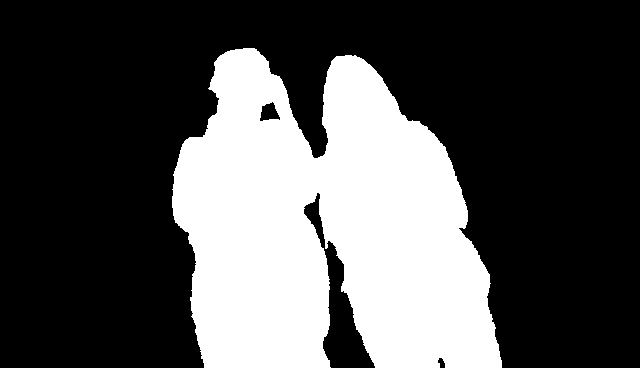}
	\end{subfigure}
	\begin{subfigure}{0.08\textwidth}
		\includegraphics[width=\textwidth]{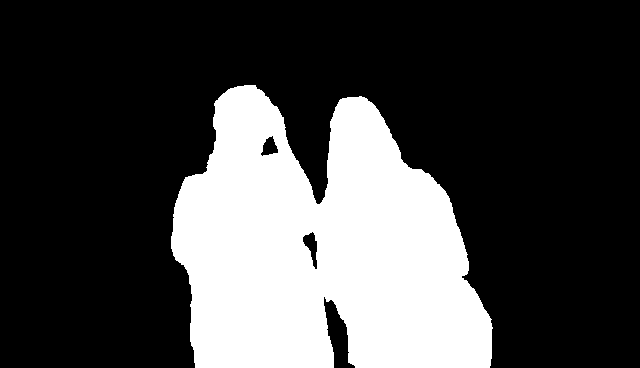}
	\end{subfigure}
	\begin{subfigure}{0.08\textwidth}
		\includegraphics[width=\textwidth]{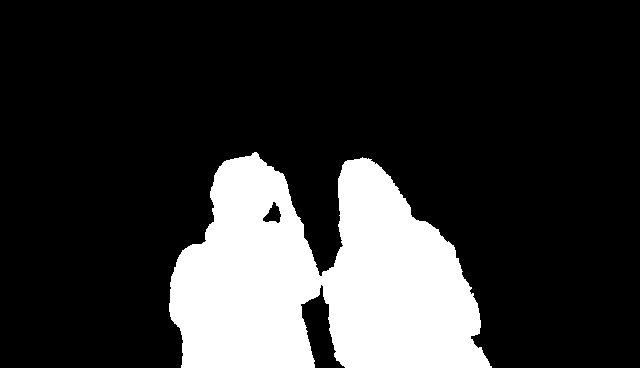}
	\end{subfigure}
	\begin{subfigure}{0.08\textwidth}
		\includegraphics[width=\textwidth]{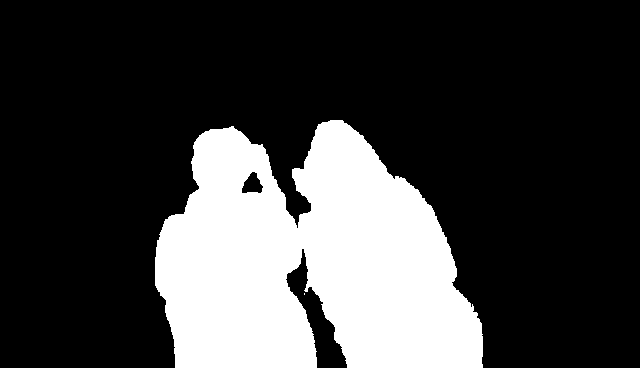}
	\end{subfigure}
	\begin{subfigure}{0.08\textwidth}
		\includegraphics[width=\textwidth]{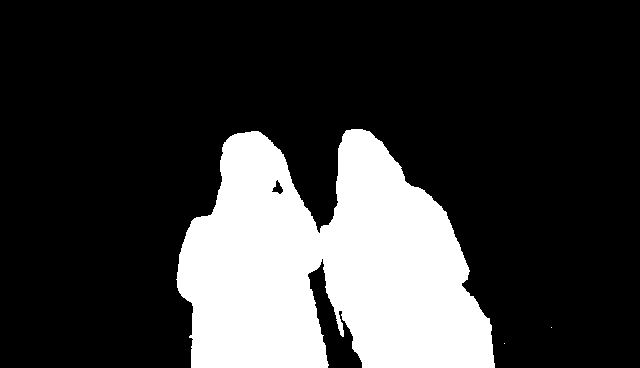}
	\end{subfigure}
	\begin{subfigure}{0.08\textwidth}
		\includegraphics[width=\textwidth]{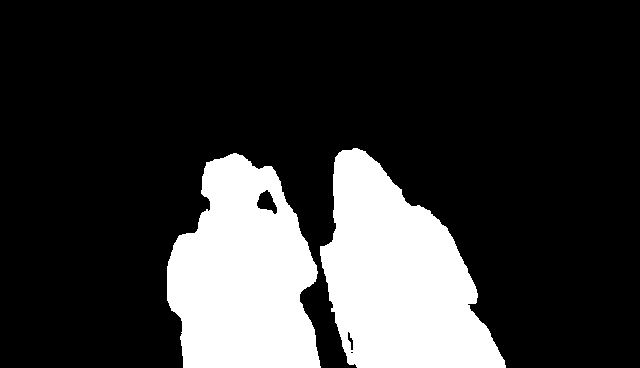}
	\end{subfigure}
	\begin{subfigure}{0.08\textwidth}
		\includegraphics[width=\textwidth]{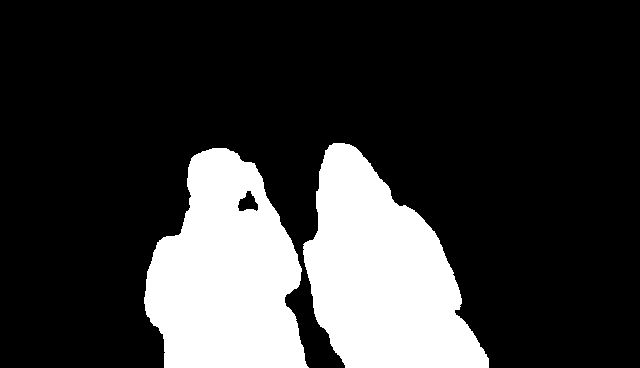}
	\end{subfigure}
	
	\vspace*{1.3mm}
	\begin{subfigure}{0.08\textwidth}
		\includegraphics[width=\textwidth]{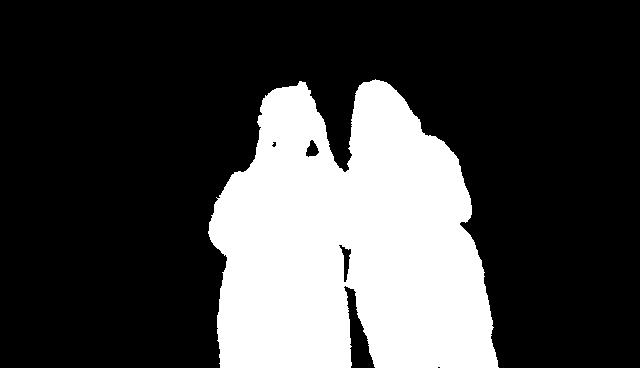}
	\end{subfigure}
	\begin{subfigure}{0.08\textwidth}
		\includegraphics[width=\textwidth]{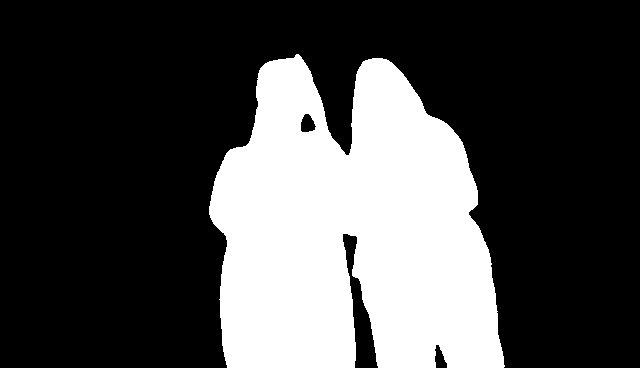}
	\end{subfigure}
	\begin{subfigure}{0.08\textwidth}
		\includegraphics[width=\textwidth]{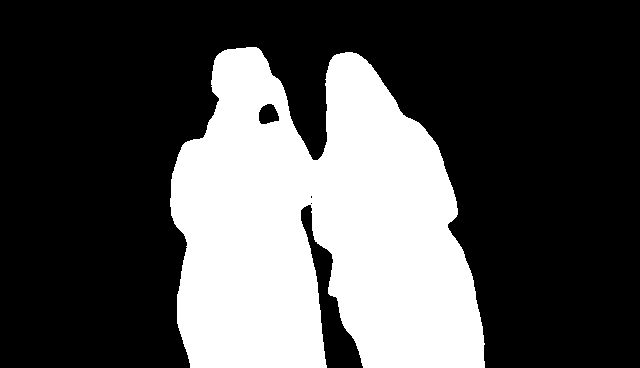}
	\end{subfigure}
	\begin{subfigure}{0.08\textwidth}
		\includegraphics[width=\textwidth]{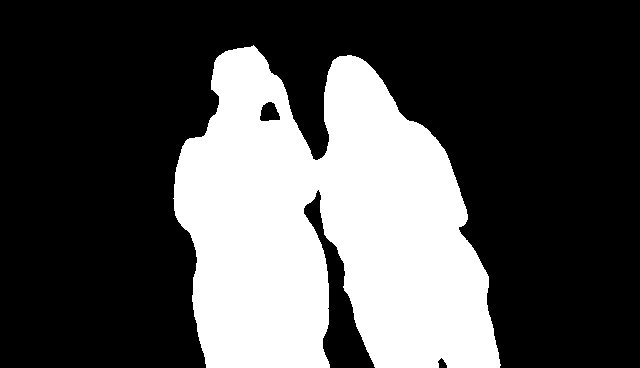}
	\end{subfigure}
	\begin{subfigure}{0.08\textwidth}
		\includegraphics[width=\textwidth]{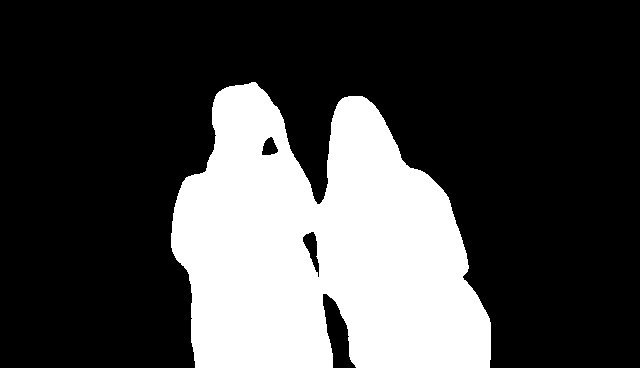}
	\end{subfigure}
	\begin{subfigure}{0.08\textwidth}
		\includegraphics[width=\textwidth]{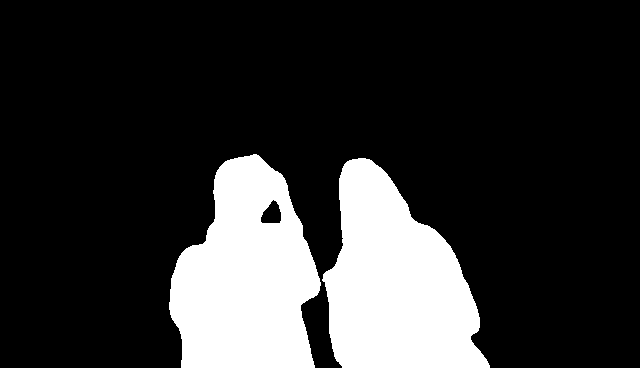}
	\end{subfigure}
	\begin{subfigure}{0.08\textwidth}
		\includegraphics[width=\textwidth]{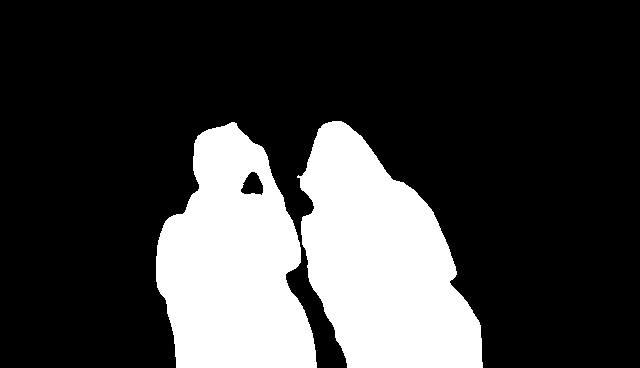}
	\end{subfigure}
	\begin{subfigure}{0.08\textwidth}
		\includegraphics[width=\textwidth]{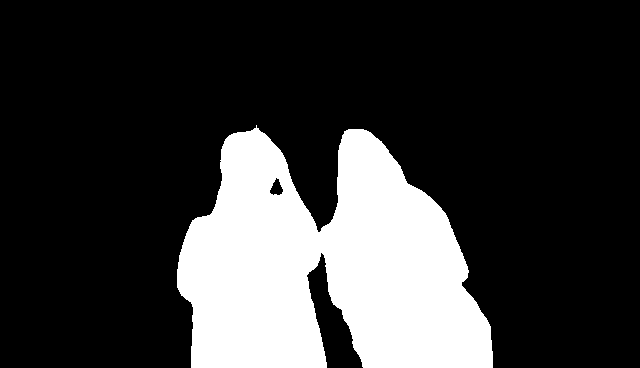}
	\end{subfigure}
	\begin{subfigure}{0.08\textwidth}
		\includegraphics[width=\textwidth]{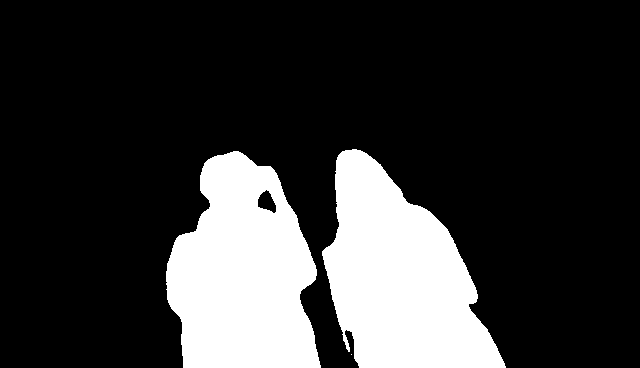}
	\end{subfigure}
	\begin{subfigure}{0.08\textwidth}
		\includegraphics[width=\textwidth]{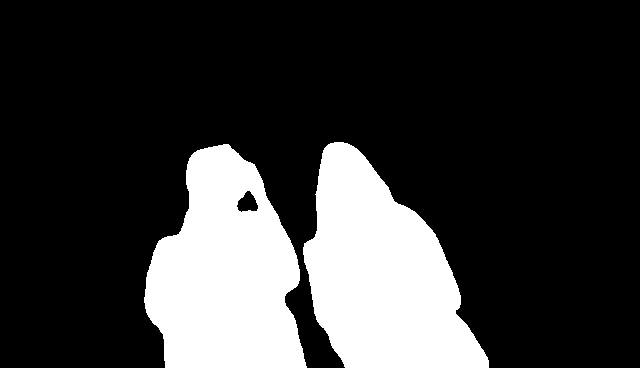}
	\end{subfigure}

	\vspace*{1.3mm}
	\begin{subfigure}{0.08\textwidth}
		\includegraphics[width=\textwidth]{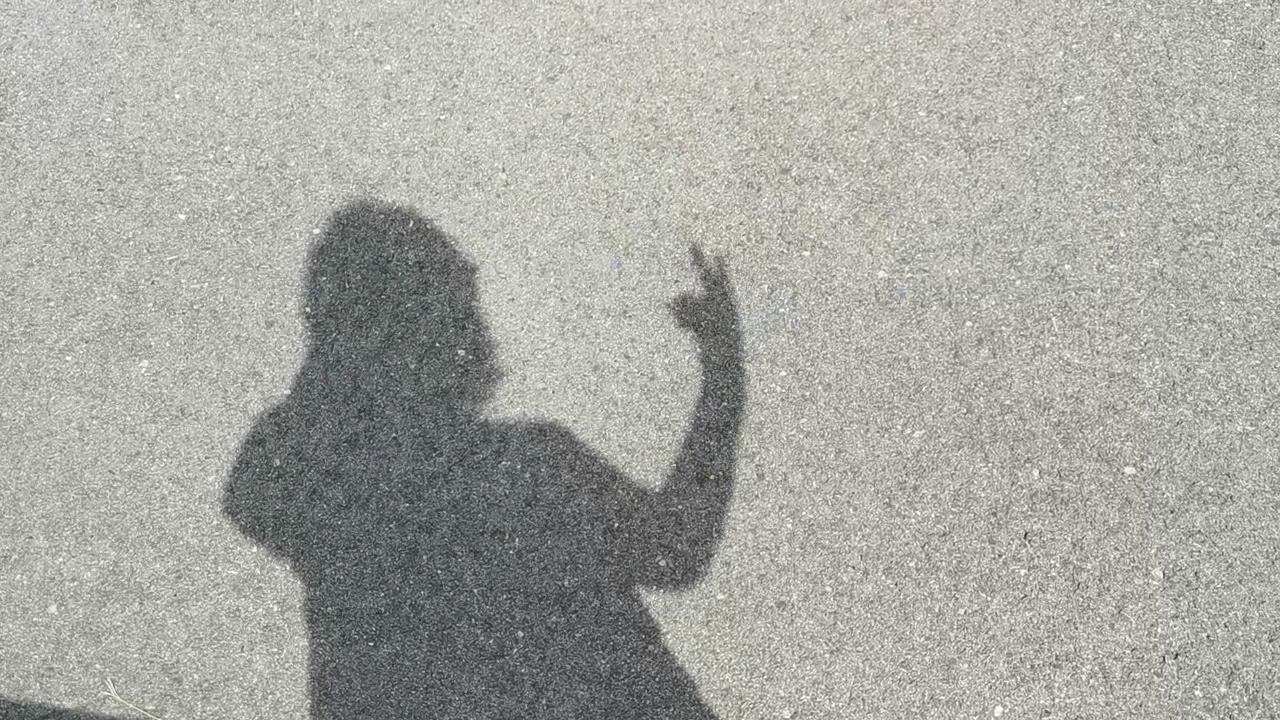}
	\end{subfigure}
	\begin{subfigure}{0.08\textwidth}
		\includegraphics[width=\textwidth]{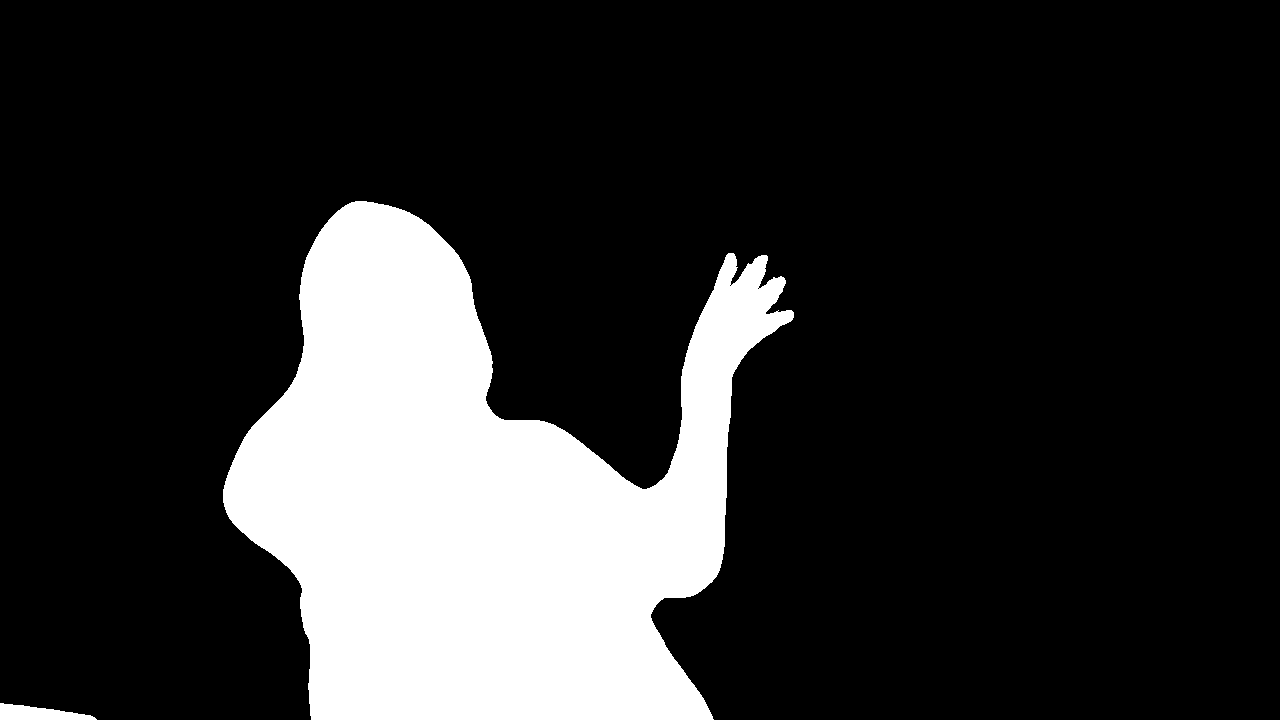}
	\end{subfigure}
	\begin{subfigure}{0.08\textwidth}
		\includegraphics[width=\textwidth]{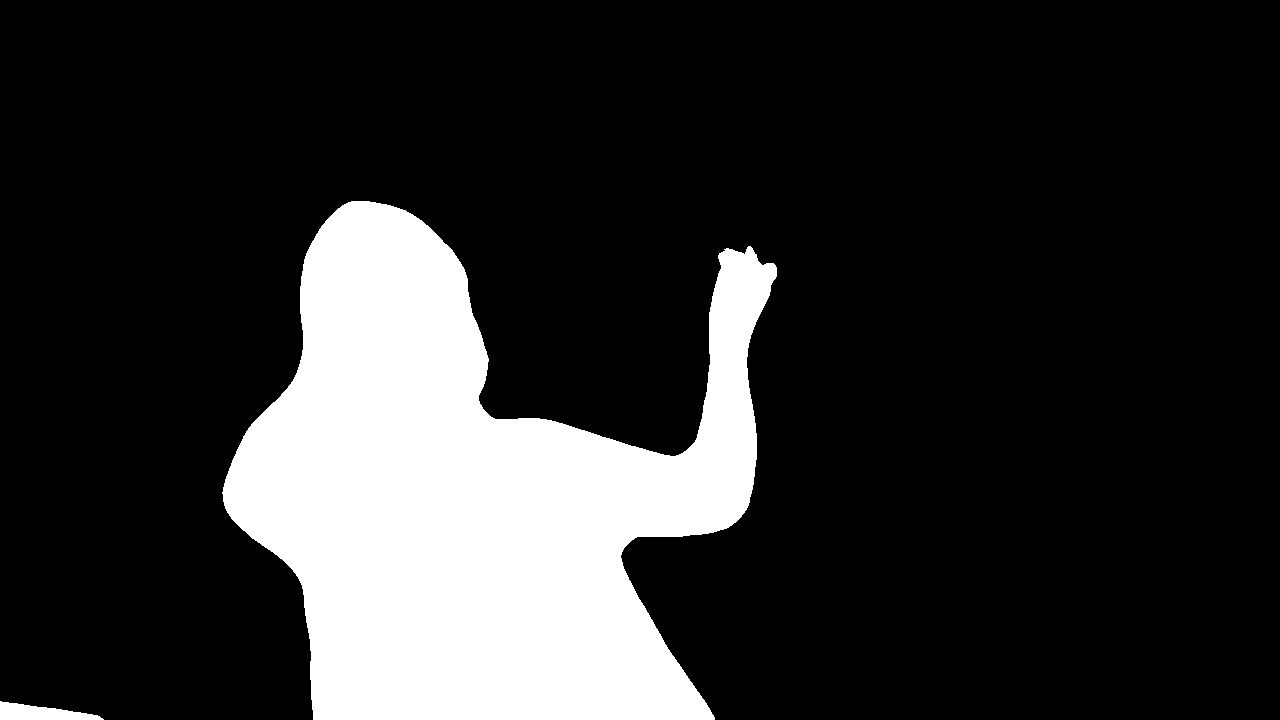}
	\end{subfigure}
	\begin{subfigure}{0.08\textwidth}
		\includegraphics[width=\textwidth]{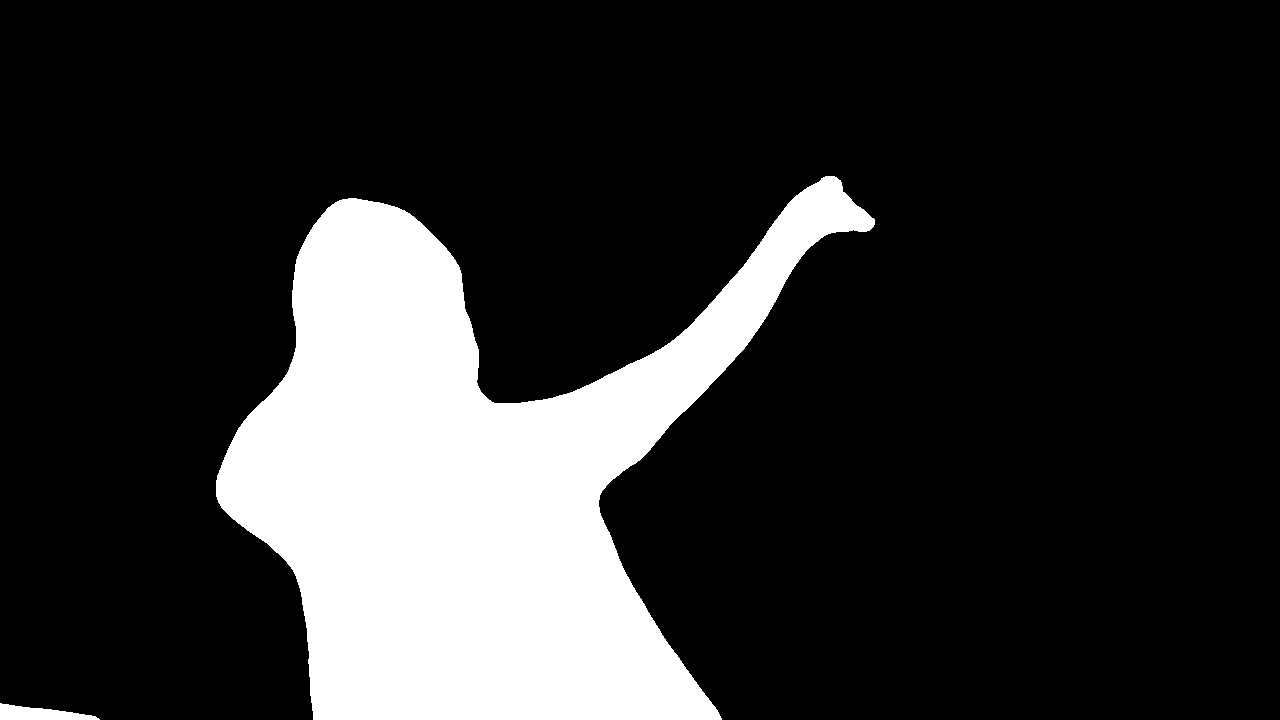}
	\end{subfigure}
	\begin{subfigure}{0.08\textwidth}
		\includegraphics[width=\textwidth]{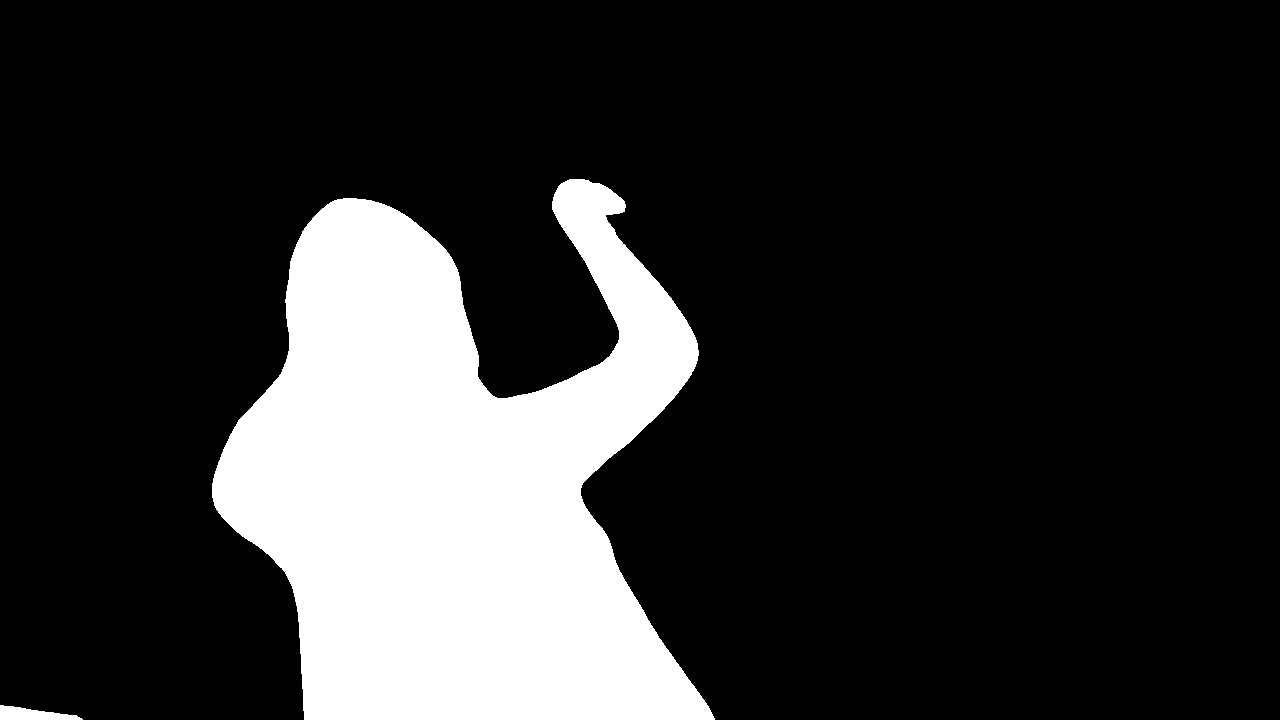}
	\end{subfigure}
	\begin{subfigure}{0.08\textwidth}
		\includegraphics[width=\textwidth]{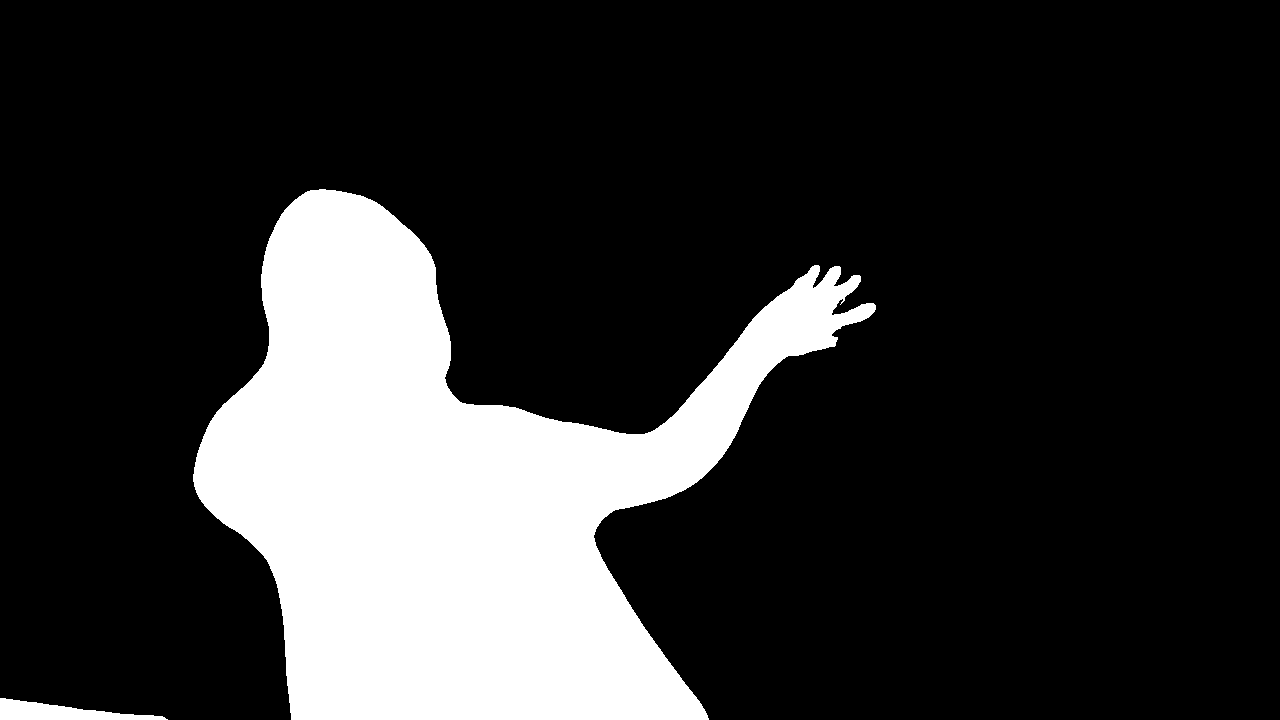}
	\end{subfigure}
	\begin{subfigure}{0.08\textwidth}
		\includegraphics[width=\textwidth]{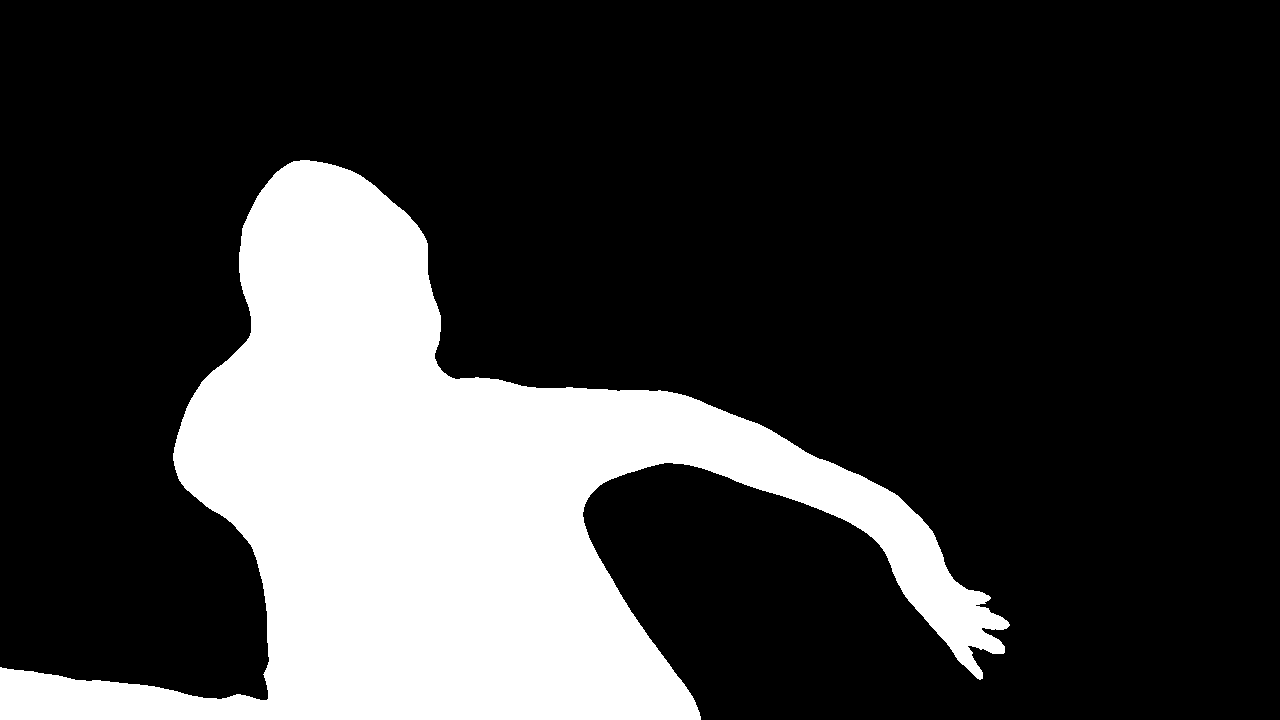}
	\end{subfigure}
	\begin{subfigure}{0.08\textwidth}
		\includegraphics[width=\textwidth]{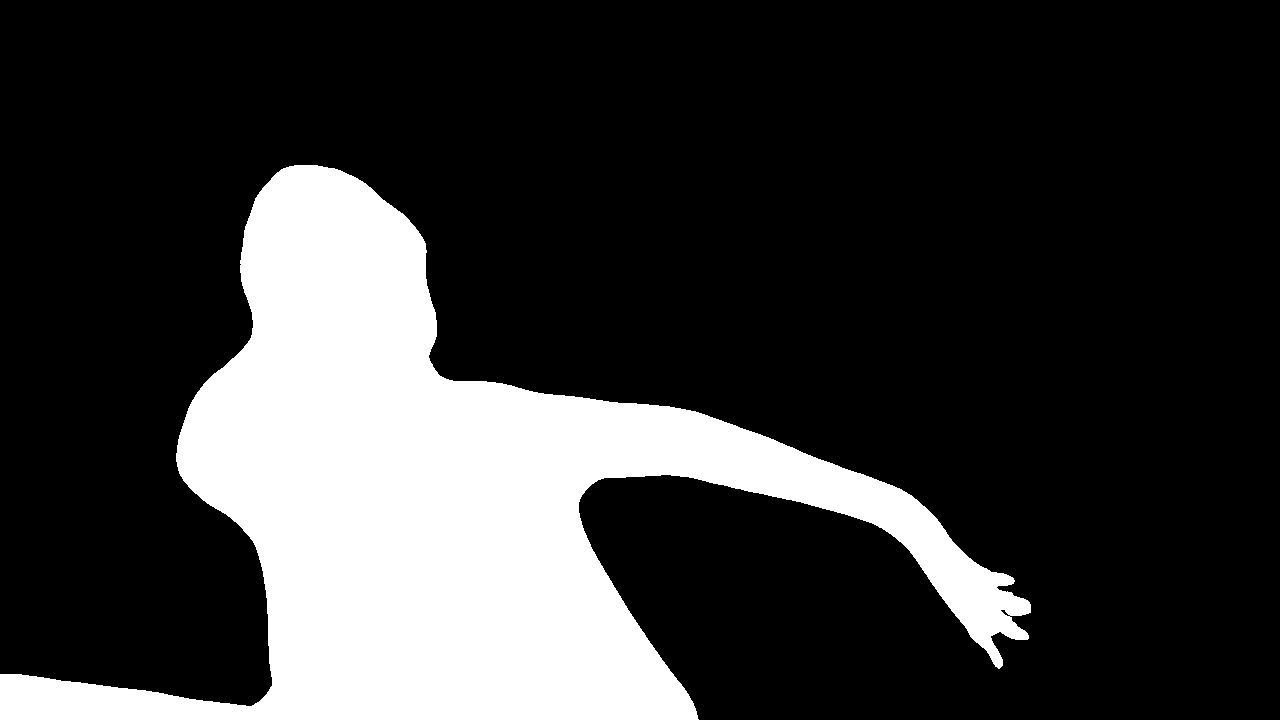}
	\end{subfigure}
	\begin{subfigure}{0.08\textwidth}
		\includegraphics[width=\textwidth]{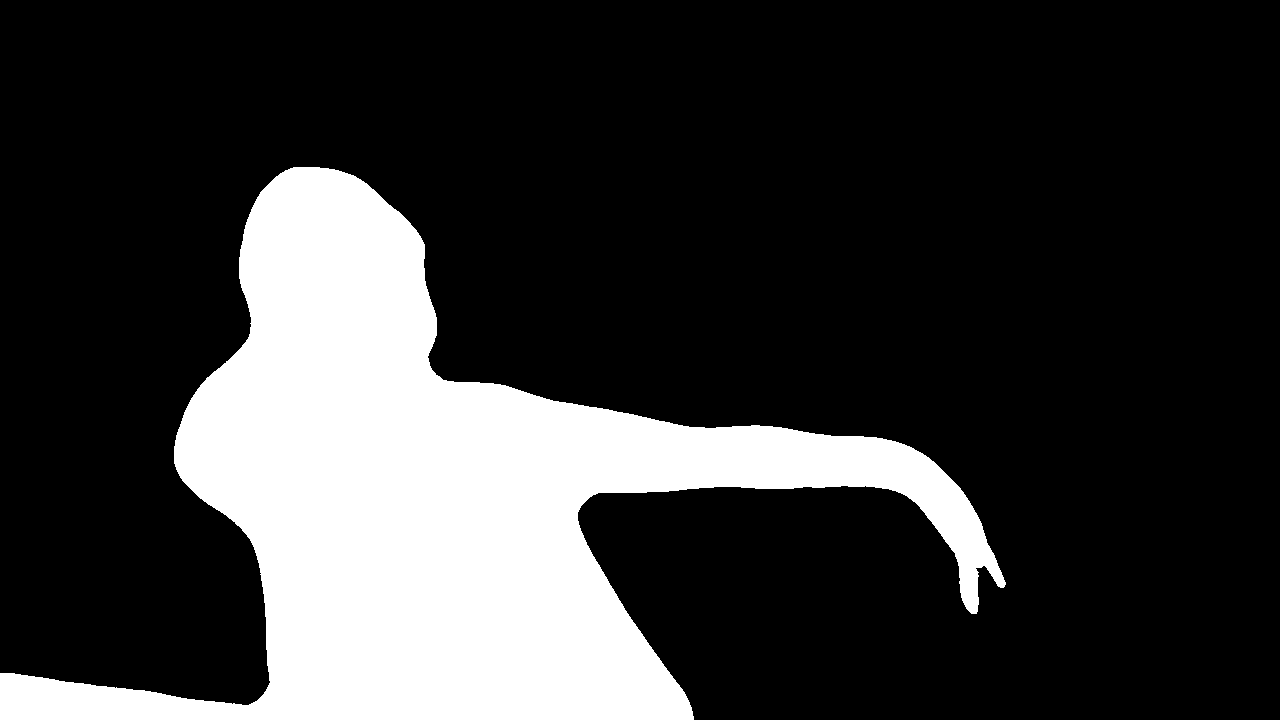}
	\end{subfigure}
	\begin{subfigure}{0.08\textwidth}
		\includegraphics[width=\textwidth]{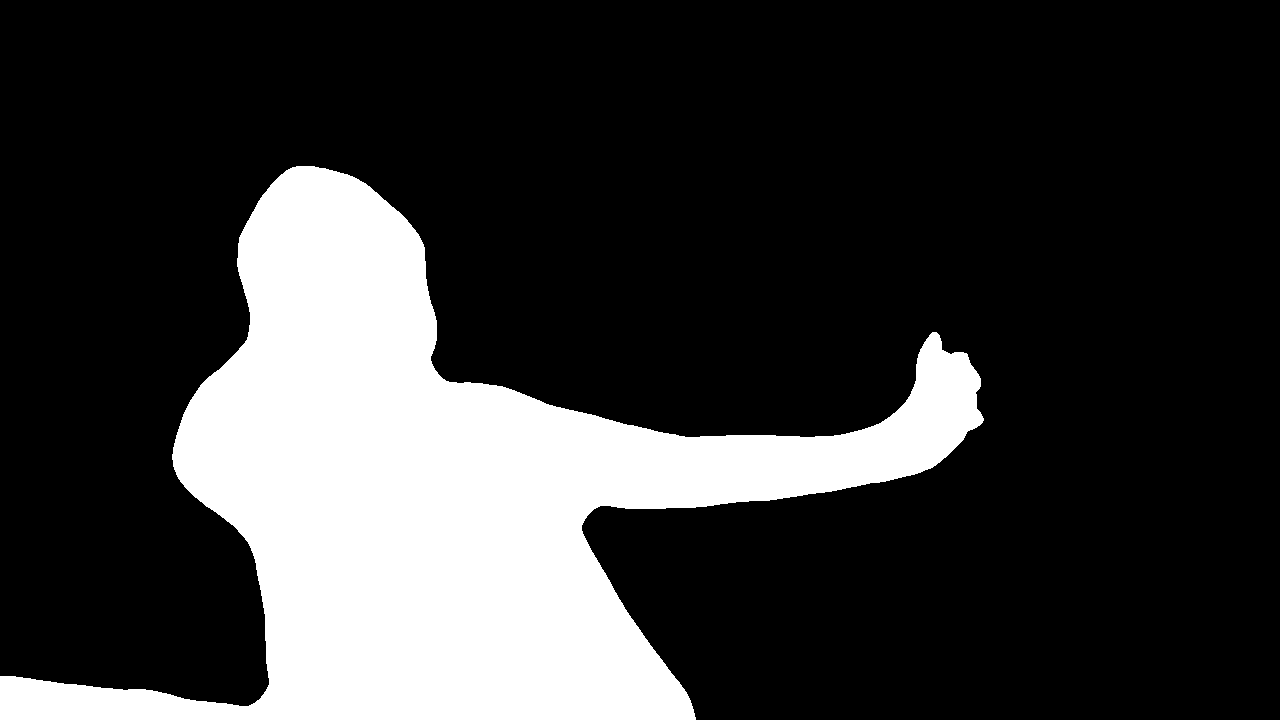}
	\end{subfigure}
	
	\vspace*{1.3mm}
	\begin{subfigure}{0.08\textwidth}
		\includegraphics[width=\textwidth]{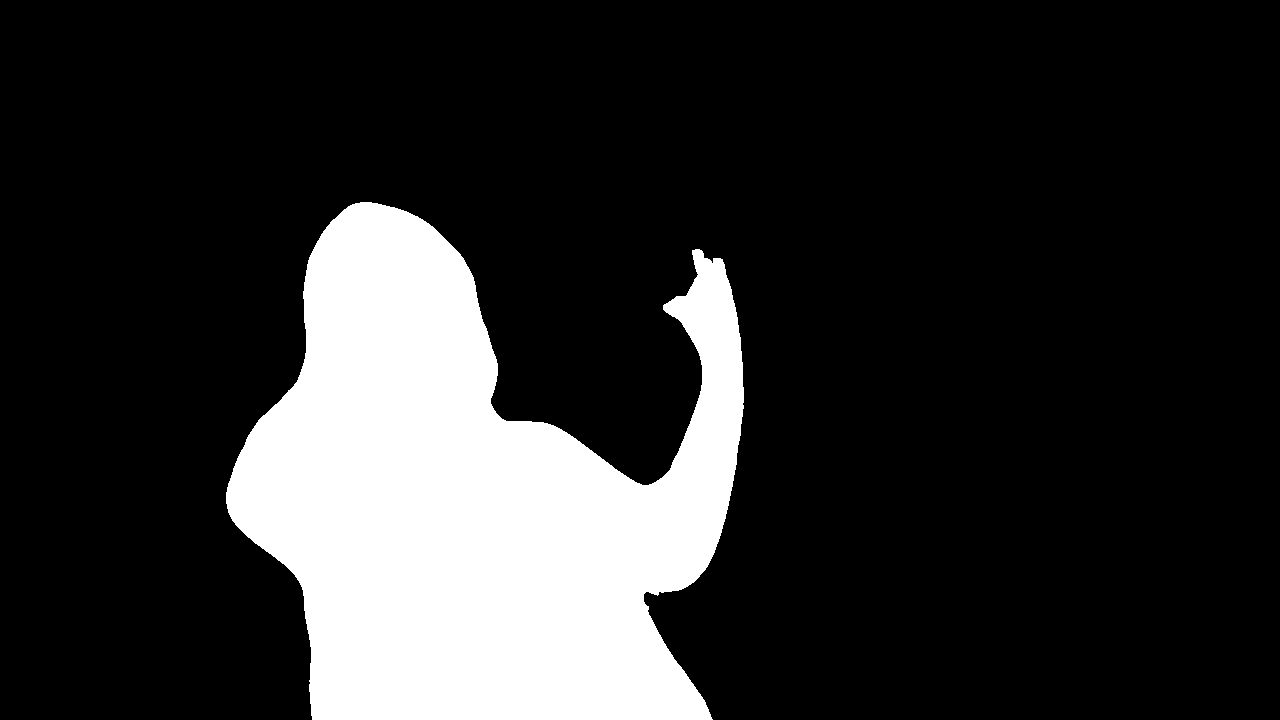}
	\end{subfigure}
	\begin{subfigure}{0.08\textwidth}
		\includegraphics[width=\textwidth]{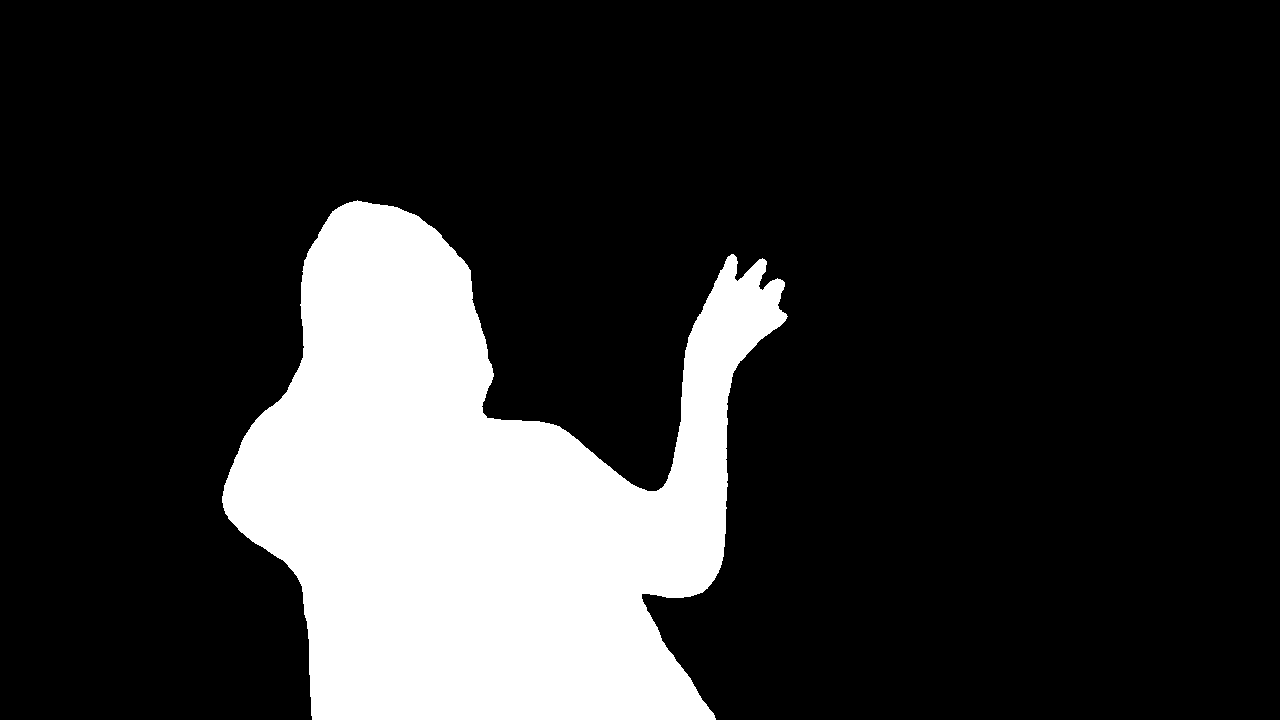}
	\end{subfigure}
	\begin{subfigure}{0.08\textwidth}
		\includegraphics[width=\textwidth]{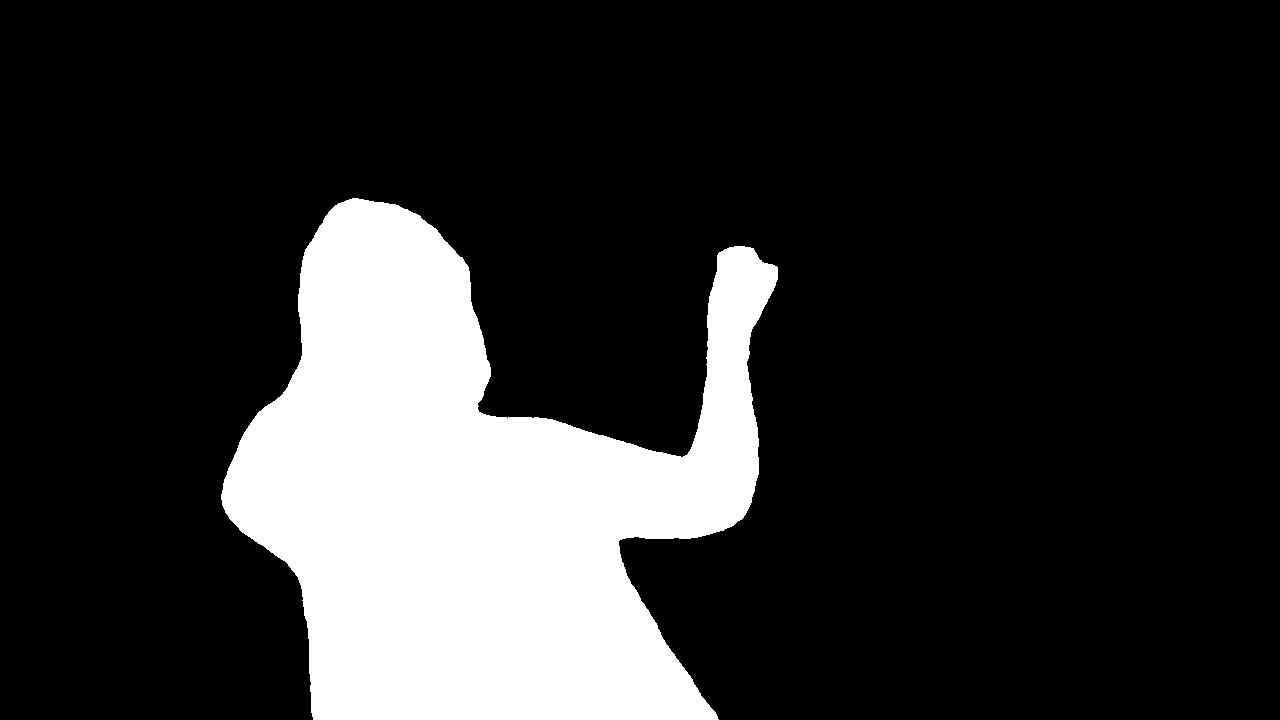}
	\end{subfigure}
	\begin{subfigure}{0.08\textwidth}
		\includegraphics[width=\textwidth]{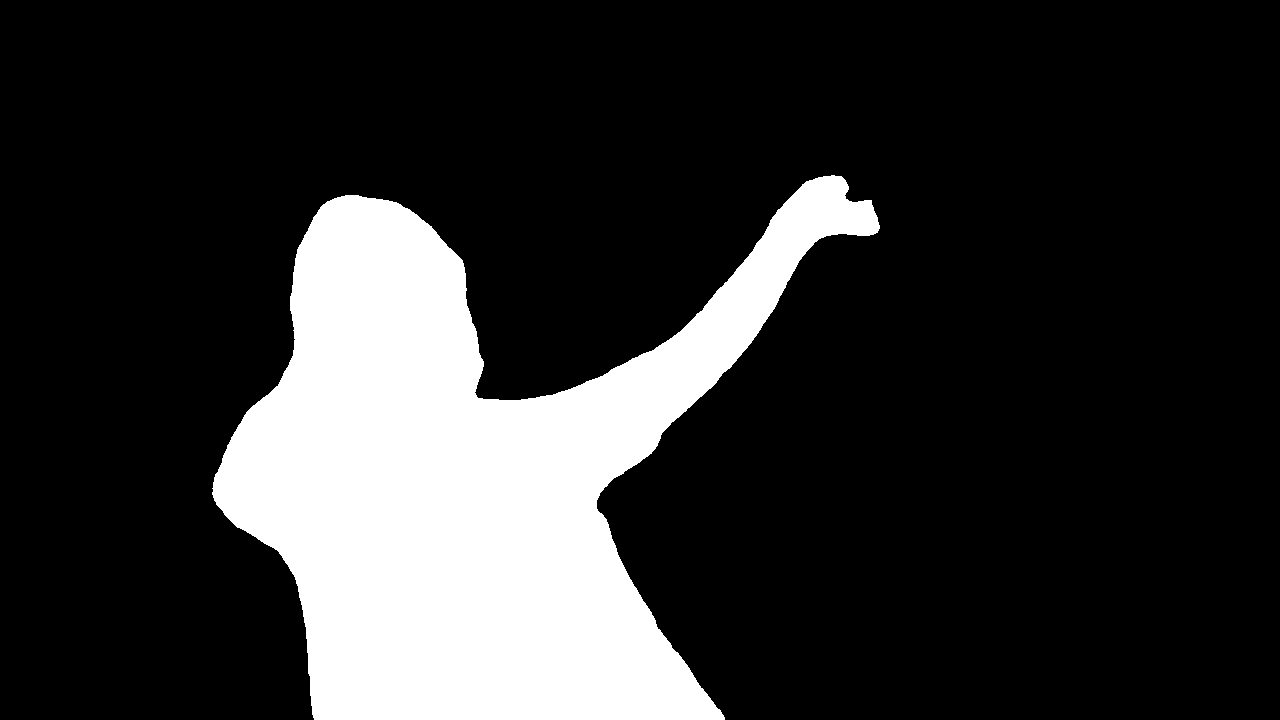}
	\end{subfigure}
	\begin{subfigure}{0.08\textwidth}
		\includegraphics[width=\textwidth]{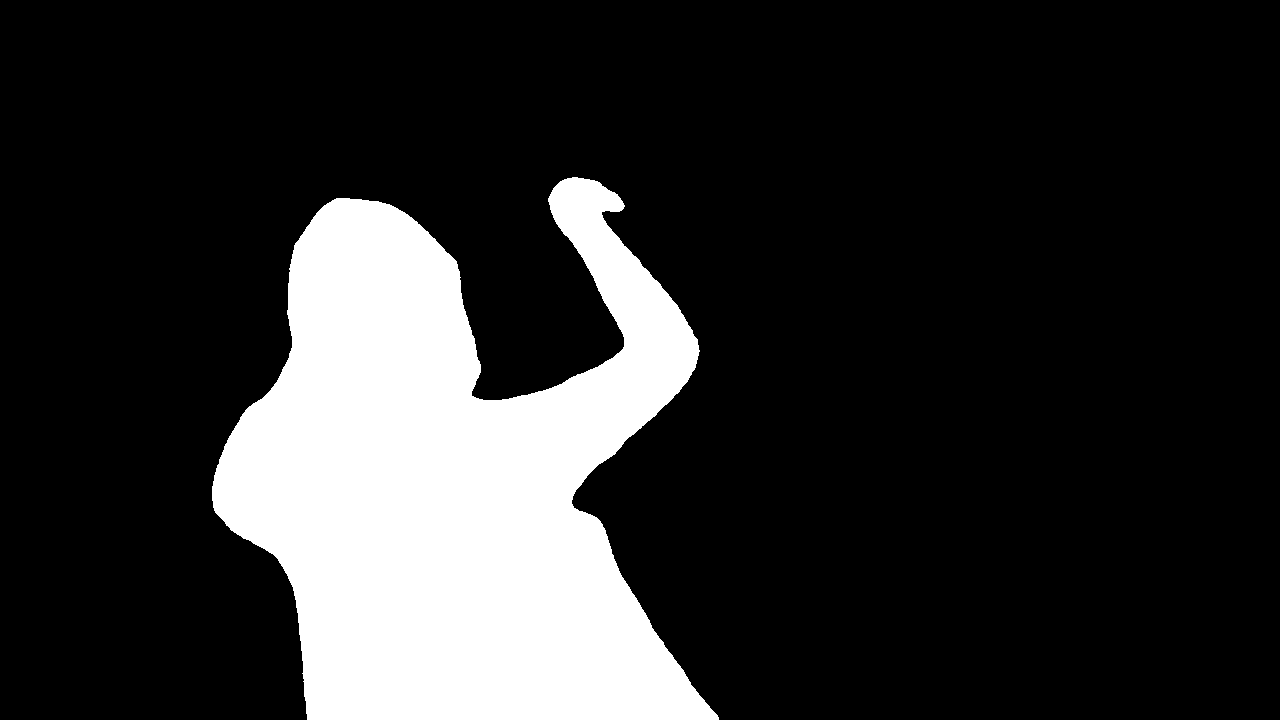}
	\end{subfigure}
	\begin{subfigure}{0.08\textwidth}
		\includegraphics[width=\textwidth]{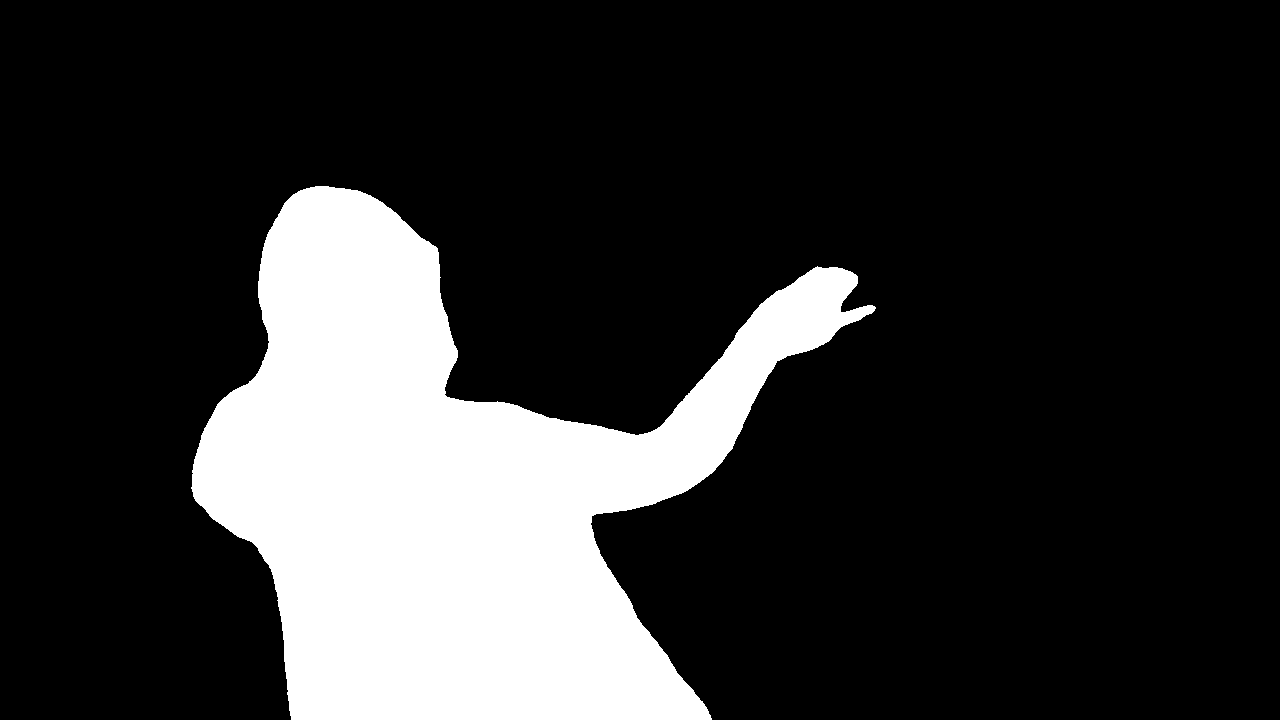}
	\end{subfigure}
	\begin{subfigure}{0.08\textwidth}
		\includegraphics[width=\textwidth]{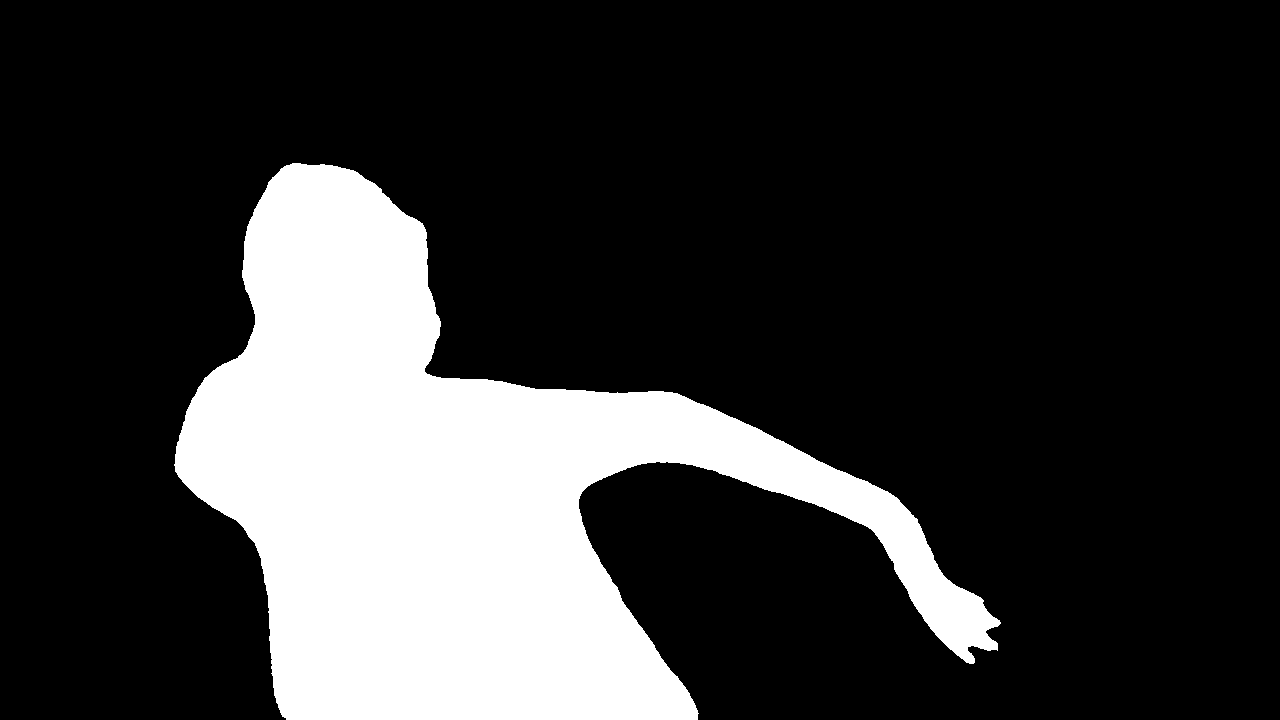}
	\end{subfigure}
	\begin{subfigure}{0.08\textwidth}
		\includegraphics[width=\textwidth]{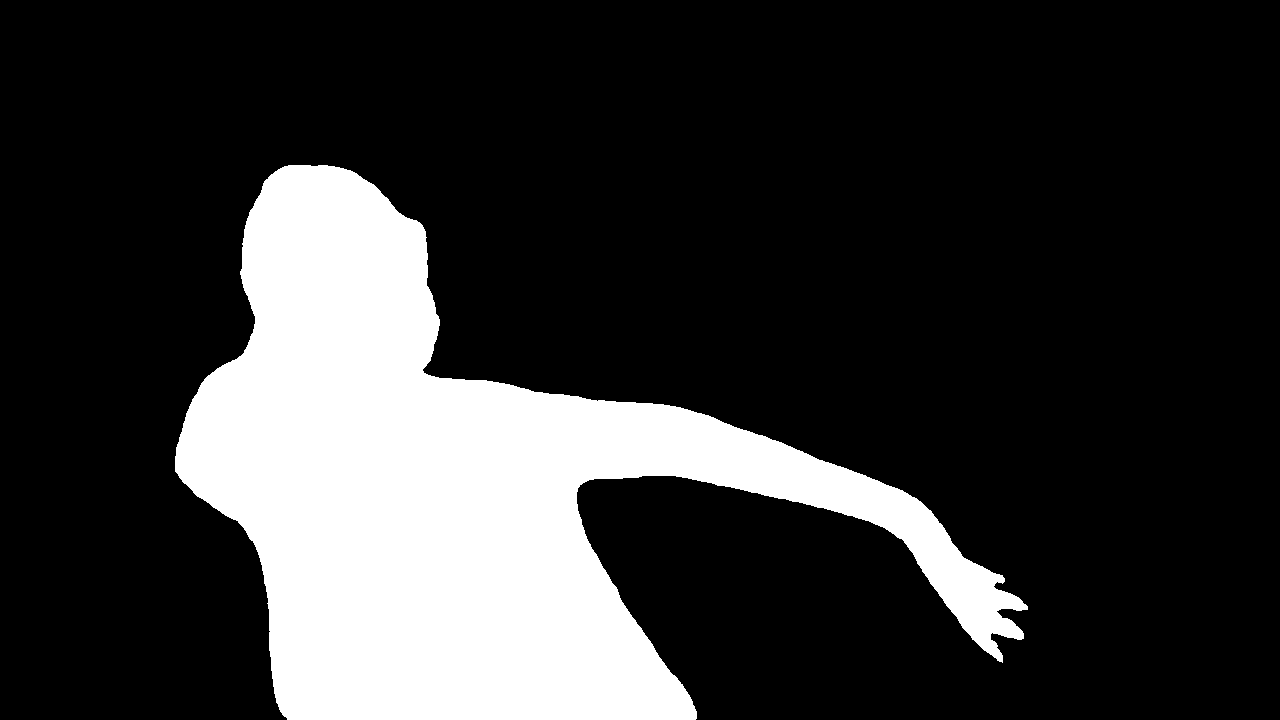}
	\end{subfigure}
	\begin{subfigure}{0.08\textwidth}
		\includegraphics[width=\textwidth]{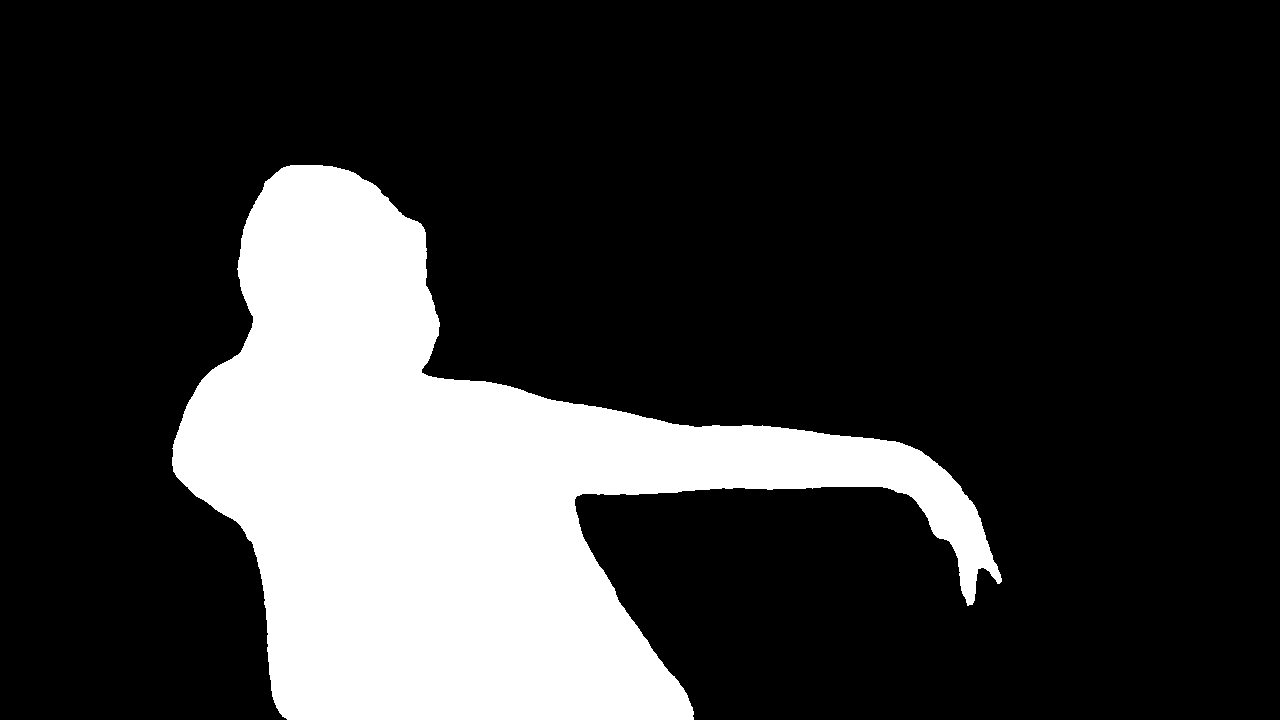}
	\end{subfigure}
	\begin{subfigure}{0.08\textwidth}
		\includegraphics[width=\textwidth]{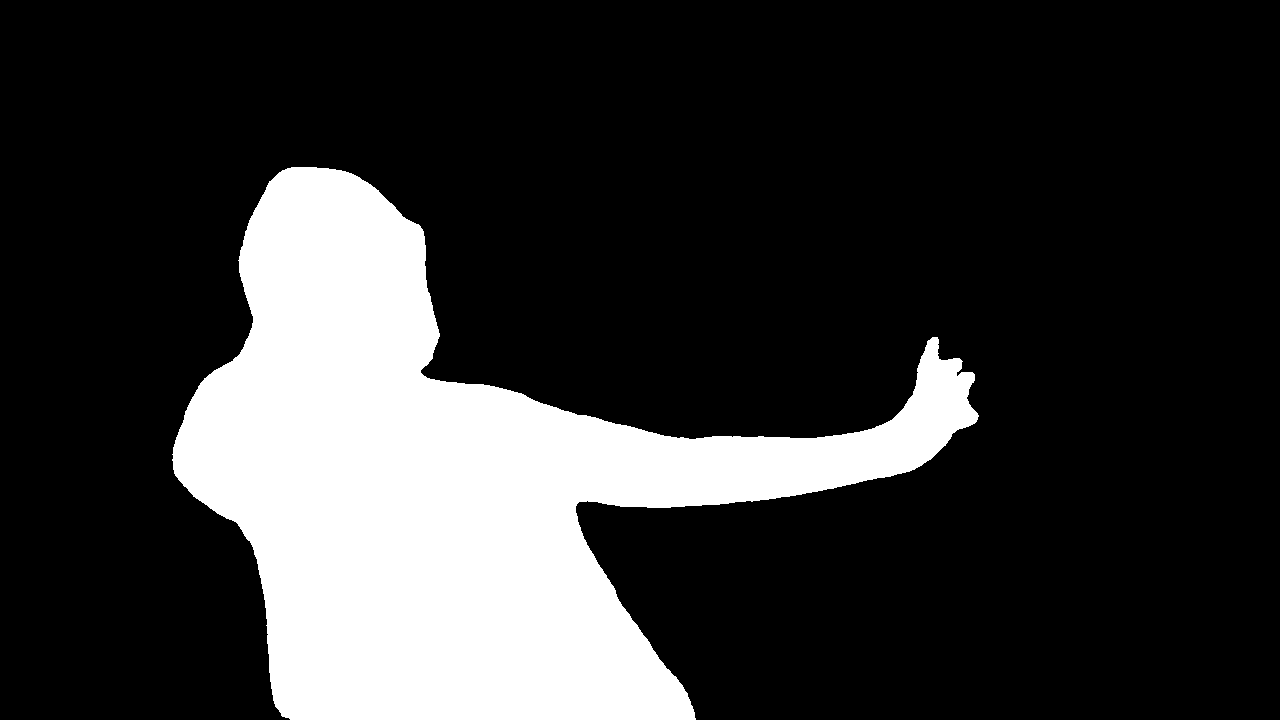}
	\end{subfigure}
	
	\vspace*{1.3mm}
	\begin{subfigure}{0.08\textwidth}
		\includegraphics[width=\textwidth]{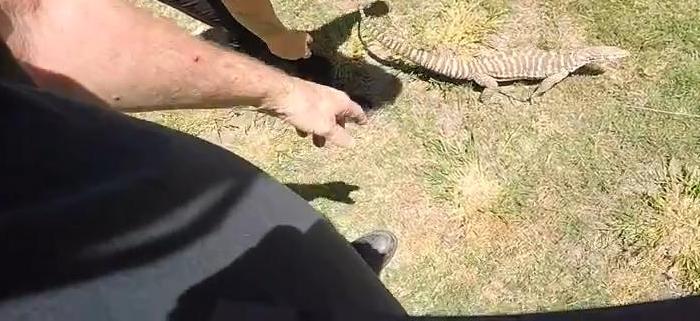}
	\end{subfigure}
	\begin{subfigure}{0.08\textwidth}
		\includegraphics[width=\textwidth]{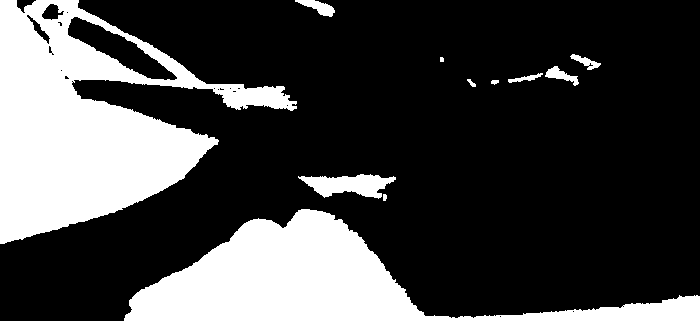}
	\end{subfigure}
	\begin{subfigure}{0.08\textwidth}
		\includegraphics[width=\textwidth]{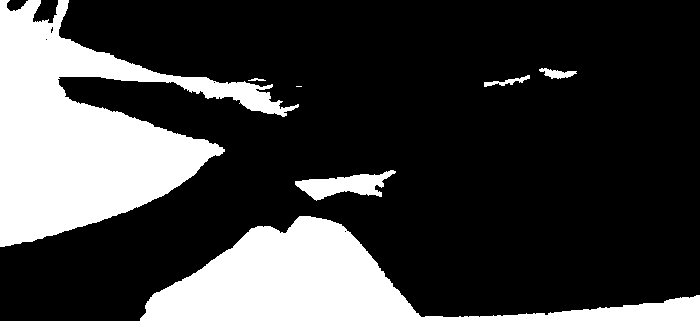}
	\end{subfigure}
	\begin{subfigure}{0.08\textwidth}
		\includegraphics[width=\textwidth]{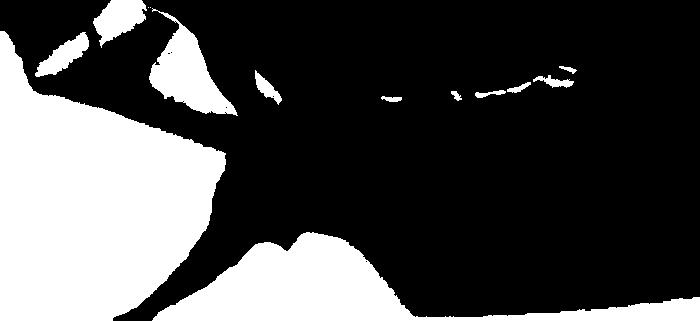}
	\end{subfigure}
	\begin{subfigure}{0.08\textwidth}
		\includegraphics[width=\textwidth]{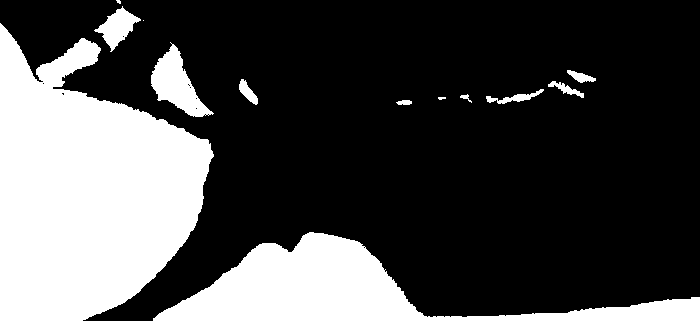}
	\end{subfigure}
	\begin{subfigure}{0.08\textwidth}
		\includegraphics[width=\textwidth]{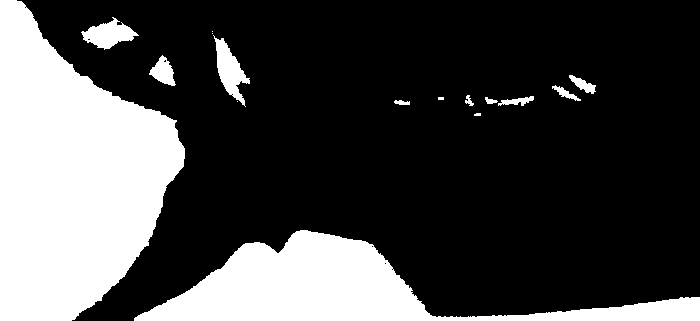}
	\end{subfigure}
	\begin{subfigure}{0.08\textwidth}
		\includegraphics[width=\textwidth]{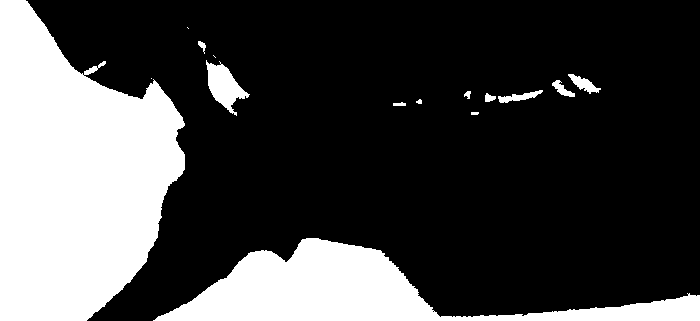}
	\end{subfigure}
	\begin{subfigure}{0.08\textwidth}
		\includegraphics[width=\textwidth]{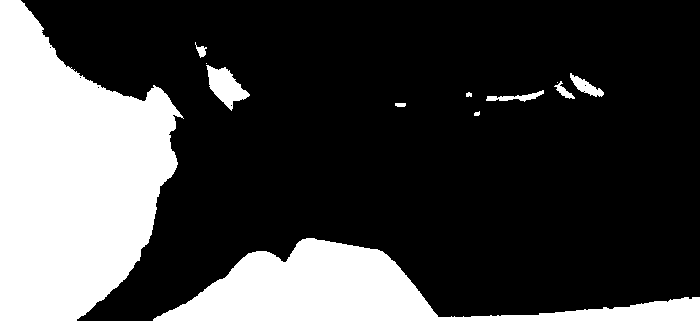}
	\end{subfigure}
	\begin{subfigure}{0.08\textwidth}
		\includegraphics[width=\textwidth]{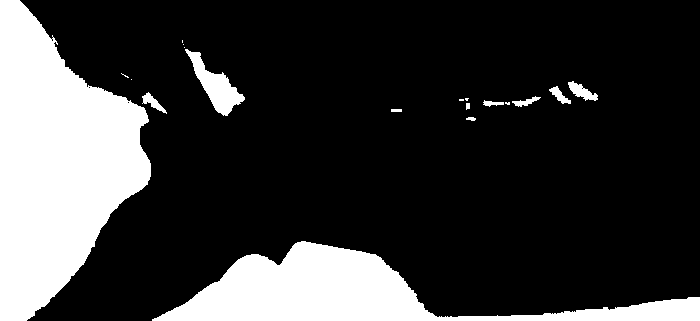}
	\end{subfigure}
	\begin{subfigure}{0.08\textwidth}
		\includegraphics[width=\textwidth]{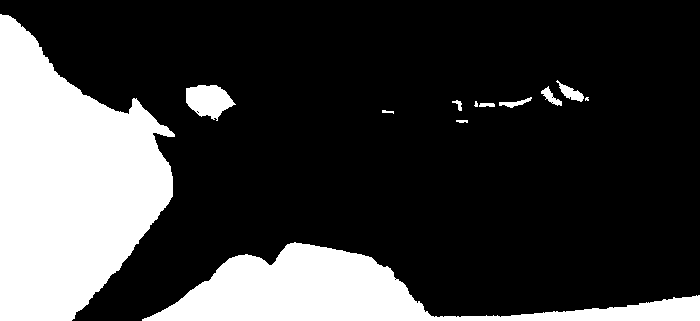}
	\end{subfigure}
	
	\vspace*{1.3mm}
	\begin{subfigure}{0.08\textwidth}
		\includegraphics[width=\textwidth]{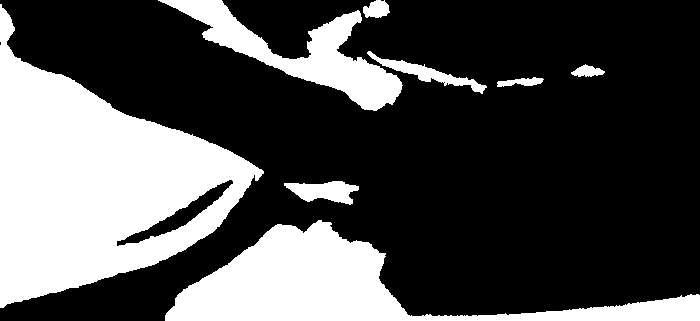}
	\end{subfigure}
	\begin{subfigure}{0.08\textwidth}
		\includegraphics[width=\textwidth]{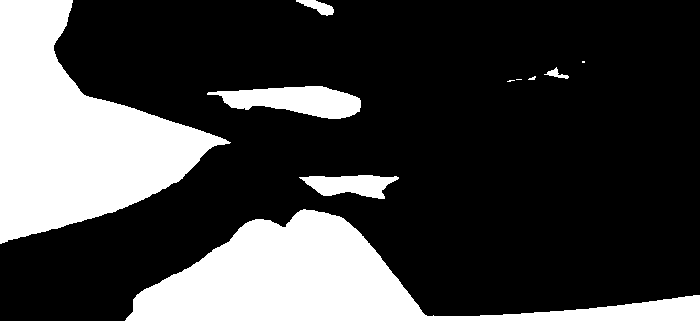}
	\end{subfigure}
	\begin{subfigure}{0.08\textwidth}
		\includegraphics[width=\textwidth]{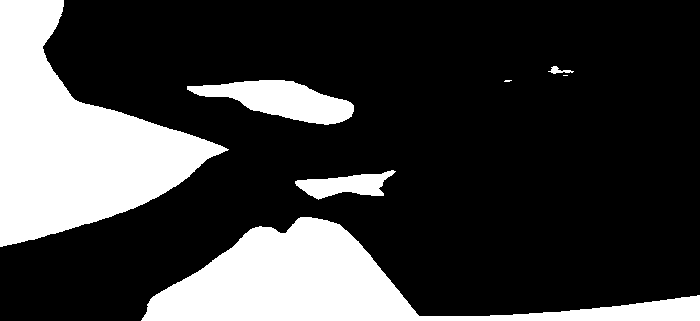}
	\end{subfigure}
	\begin{subfigure}{0.08\textwidth}
		\includegraphics[width=\textwidth]{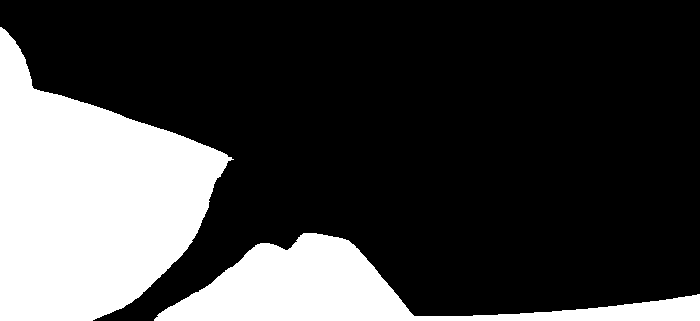}
	\end{subfigure}
	\begin{subfigure}{0.08\textwidth}
		\includegraphics[width=\textwidth]{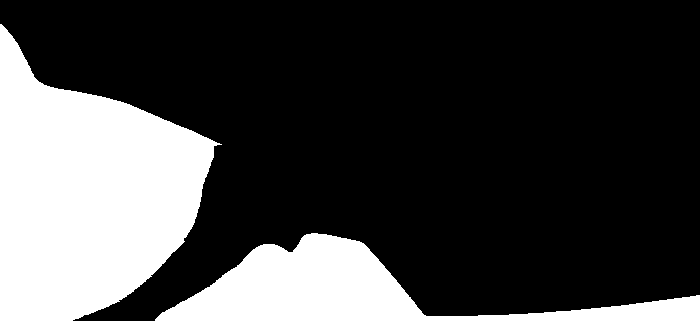}
	\end{subfigure}
	\begin{subfigure}{0.08\textwidth}
		\includegraphics[width=\textwidth]{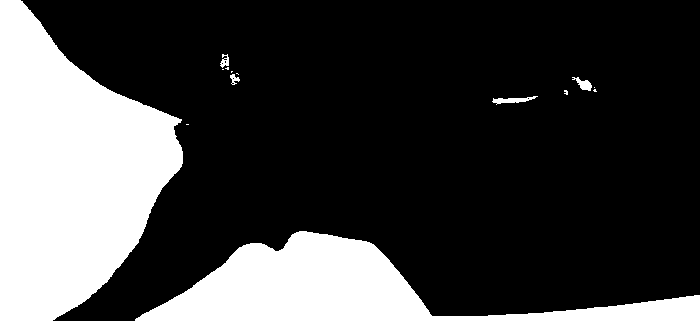}
	\end{subfigure}
	\begin{subfigure}{0.08\textwidth}
		\includegraphics[width=\textwidth]{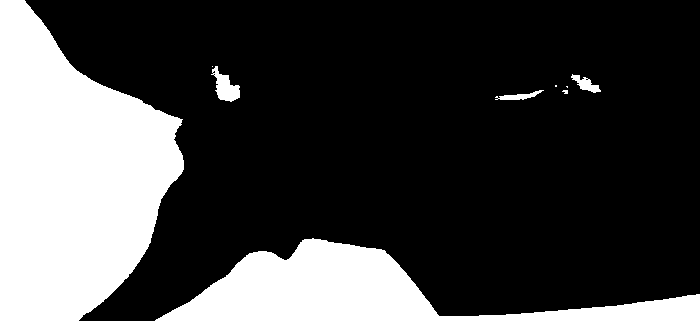}
	\end{subfigure}
	\begin{subfigure}{0.08\textwidth}
		\includegraphics[width=\textwidth]{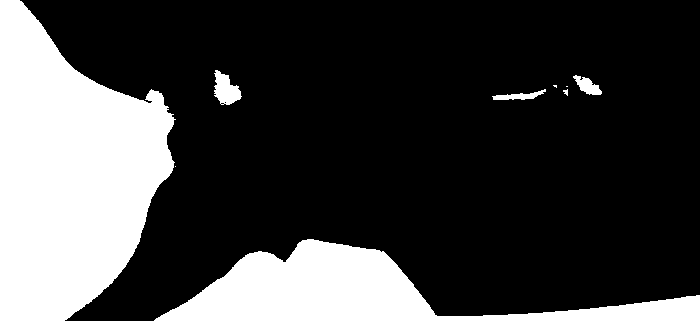}
	\end{subfigure}
	\begin{subfigure}{0.08\textwidth}
		\includegraphics[width=\textwidth]{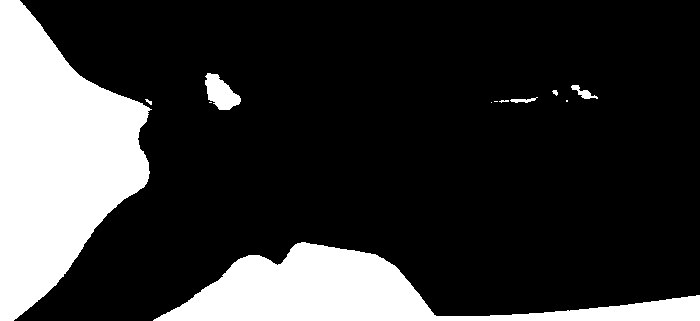}
	\end{subfigure}
	\begin{subfigure}{0.08\textwidth}
		\includegraphics[width=\textwidth]{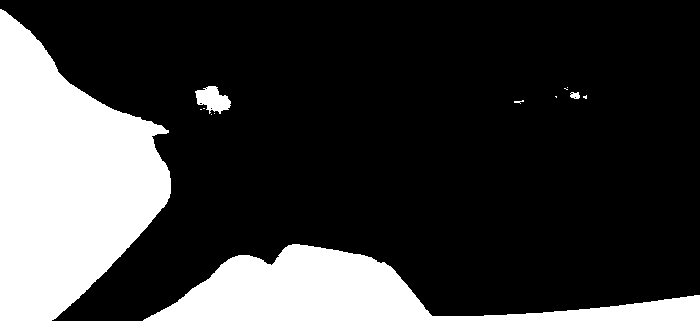}
	\end{subfigure}

	\vspace*{1.3mm}
	\begin{subfigure}{0.08\textwidth}
		\includegraphics[width=\textwidth]{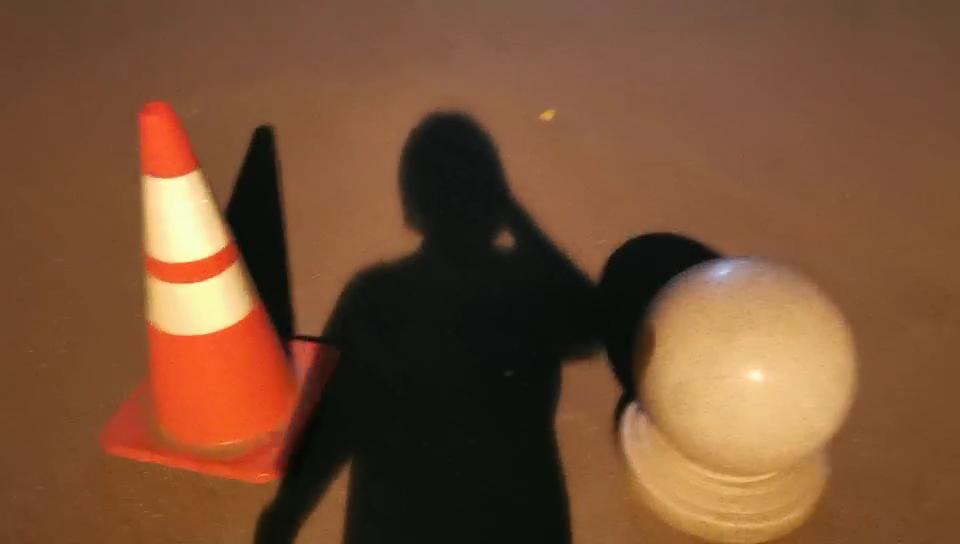}
	\end{subfigure}
	\begin{subfigure}{0.08\textwidth}
		\includegraphics[width=\textwidth]{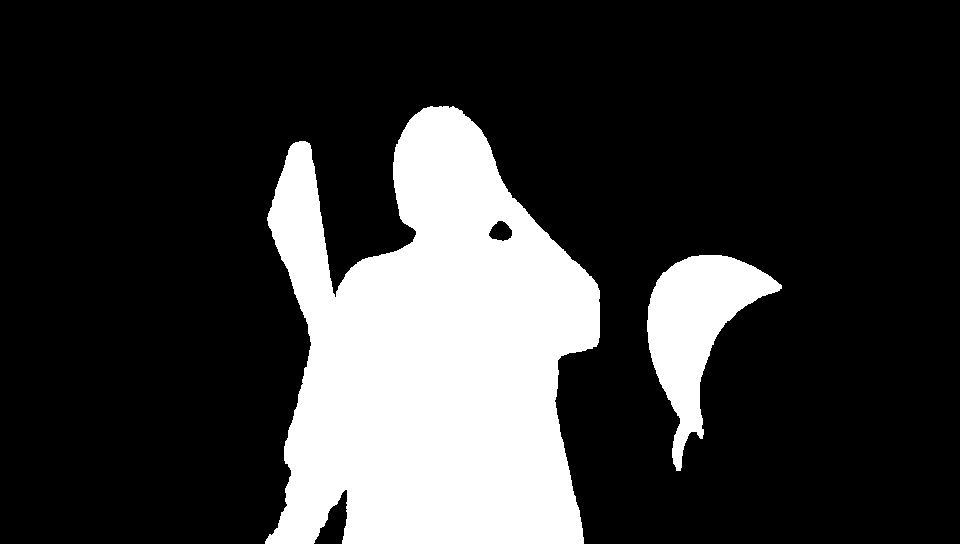}
	\end{subfigure}
	\begin{subfigure}{0.08\textwidth}
		\includegraphics[width=\textwidth]{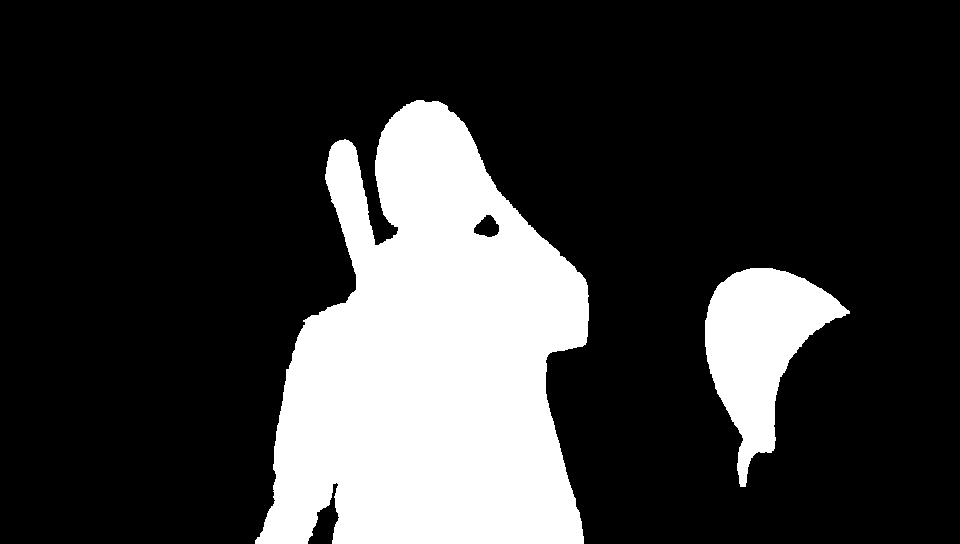}
	\end{subfigure}
	\begin{subfigure}{0.08\textwidth}
		\includegraphics[width=\textwidth]{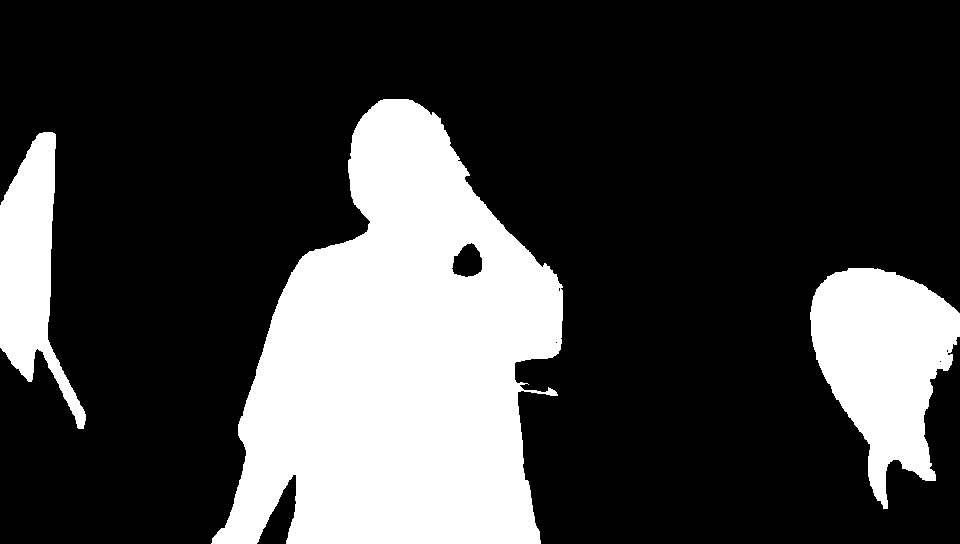}
	\end{subfigure}
	\begin{subfigure}{0.08\textwidth}
		\includegraphics[width=\textwidth]{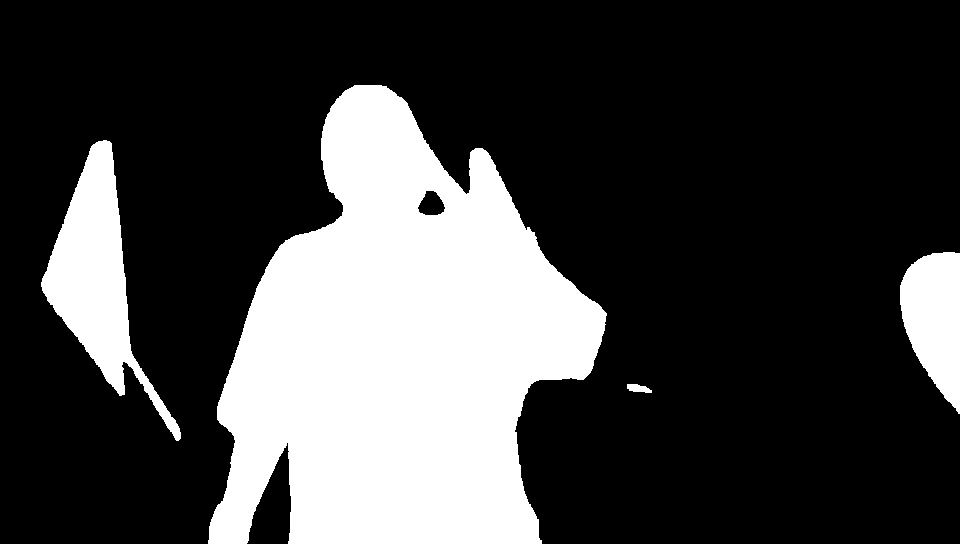}
	\end{subfigure}
	\begin{subfigure}{0.08\textwidth}
		\includegraphics[width=\textwidth]{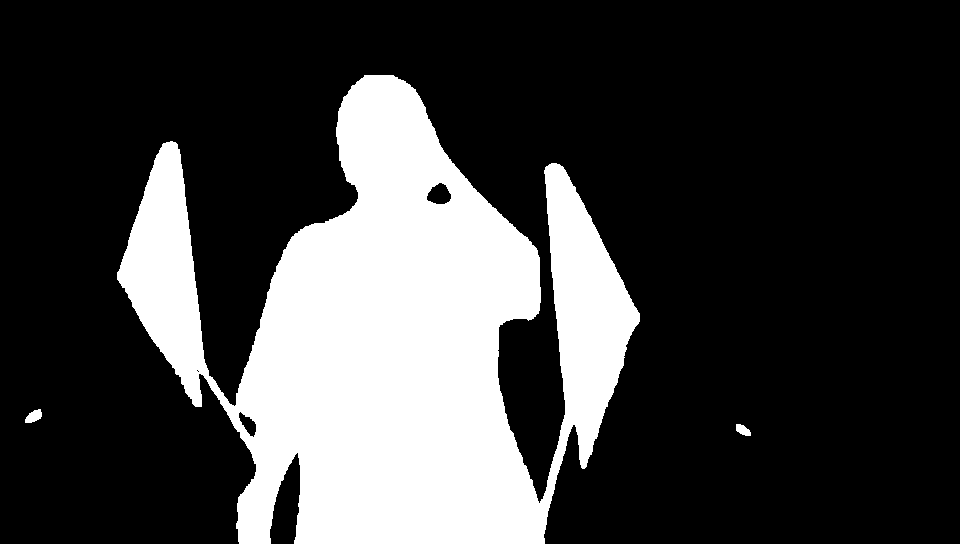}
	\end{subfigure}
	\begin{subfigure}{0.08\textwidth}
		\includegraphics[width=\textwidth]{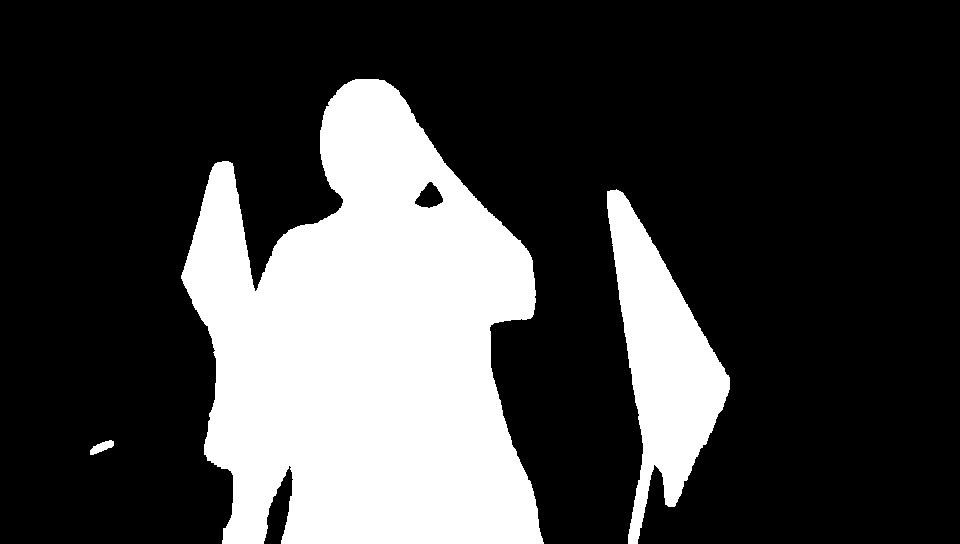}
	\end{subfigure}
	\begin{subfigure}{0.08\textwidth}
		\includegraphics[width=\textwidth]{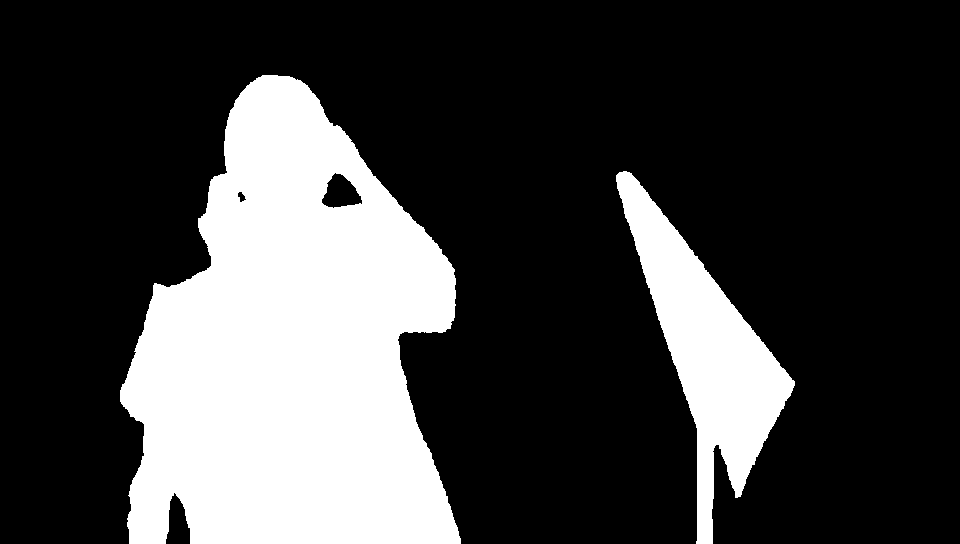}
	\end{subfigure}
	\begin{subfigure}{0.08\textwidth}
		\includegraphics[width=\textwidth]{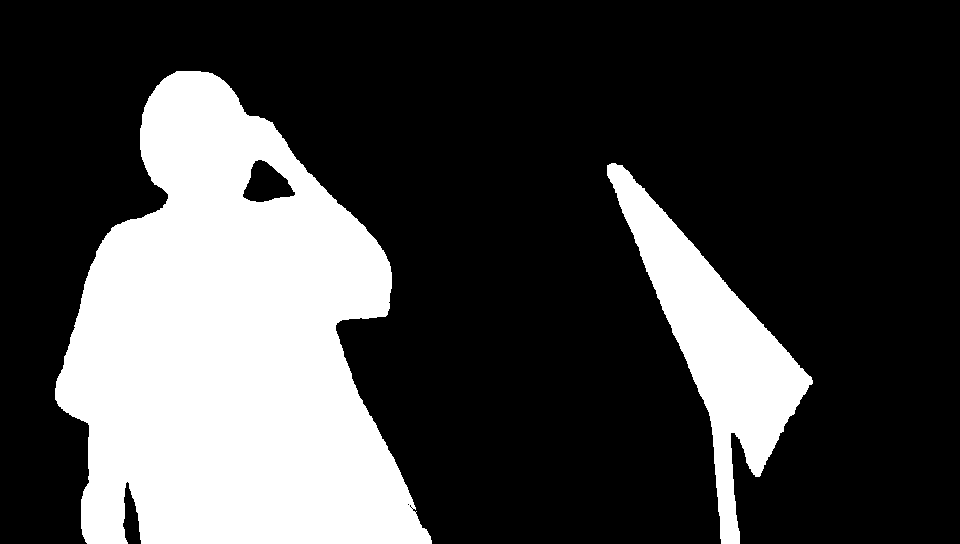}
	\end{subfigure}
	\begin{subfigure}{0.08\textwidth}
		\includegraphics[width=\textwidth]{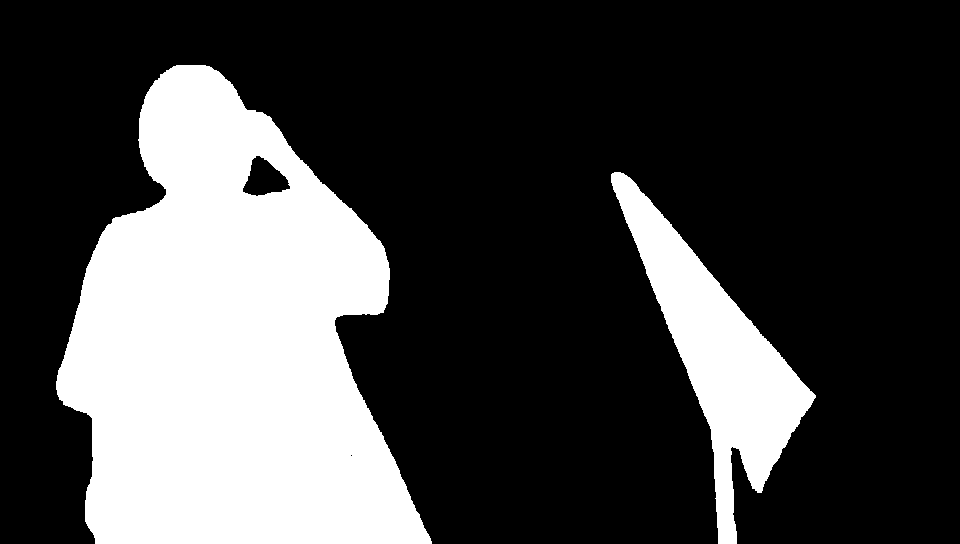}
	\end{subfigure}
	
	\vspace*{1.3mm}
	\begin{subfigure}{0.08\textwidth}
		\includegraphics[width=\textwidth]{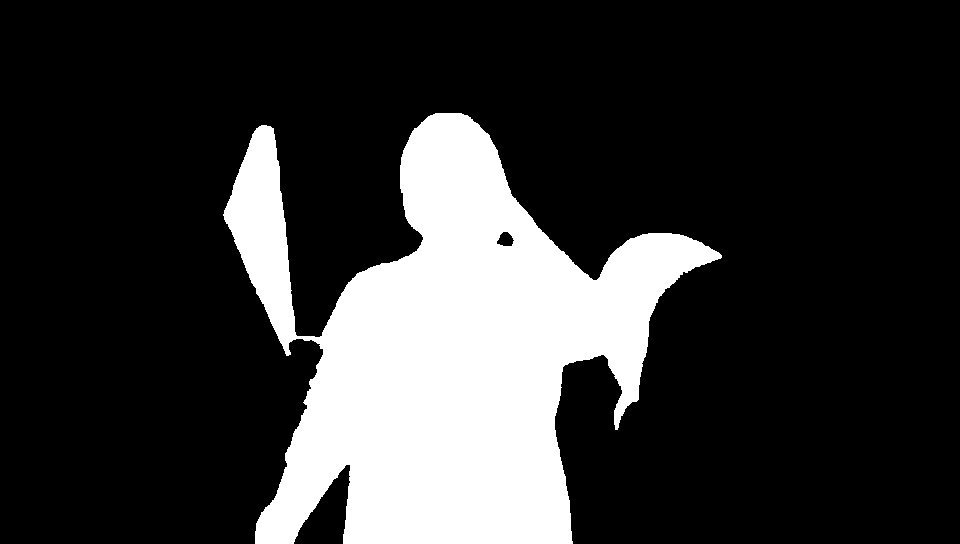}
	\end{subfigure}
	\begin{subfigure}{0.08\textwidth}
		\includegraphics[width=\textwidth]{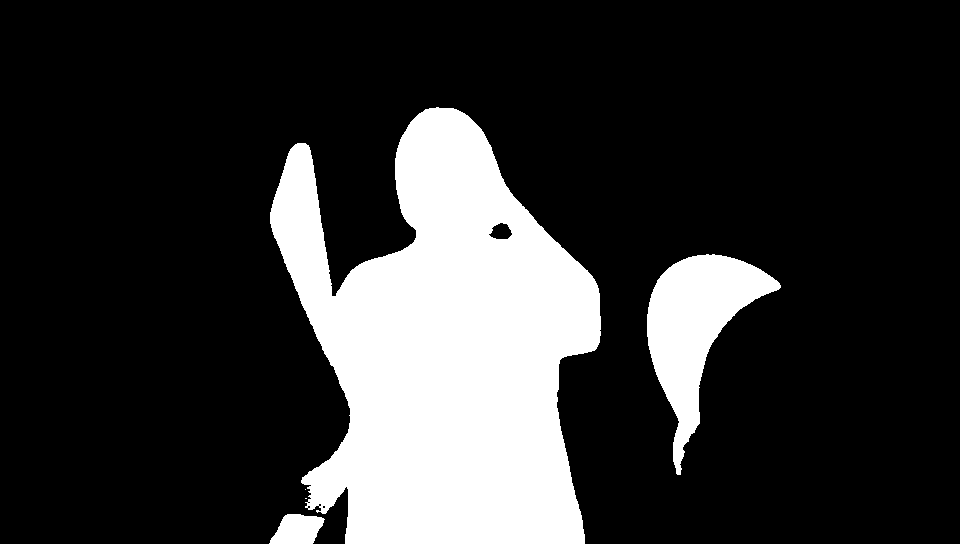}
	\end{subfigure}
	\begin{subfigure}{0.08\textwidth}
		\includegraphics[width=\textwidth]{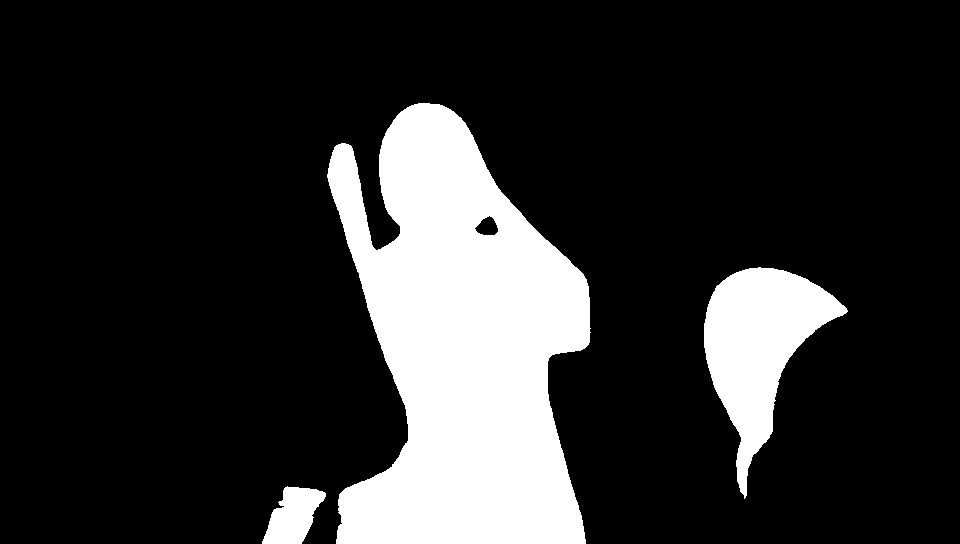}
	\end{subfigure}
	\begin{subfigure}{0.08\textwidth}
		\includegraphics[width=\textwidth]{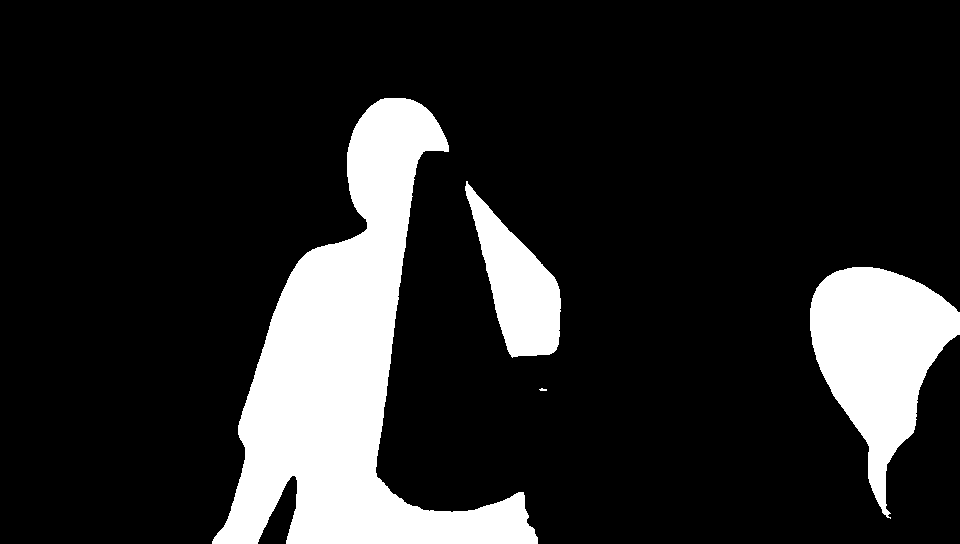}
	\end{subfigure}
	\begin{subfigure}{0.08\textwidth}
		\includegraphics[width=\textwidth]{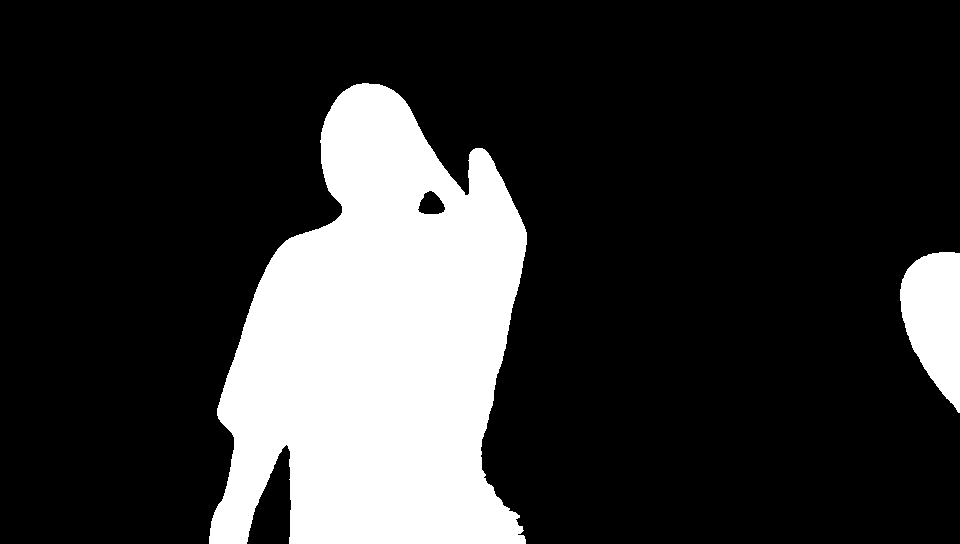}
	\end{subfigure}
	\begin{subfigure}{0.08\textwidth}
		\includegraphics[width=\textwidth]{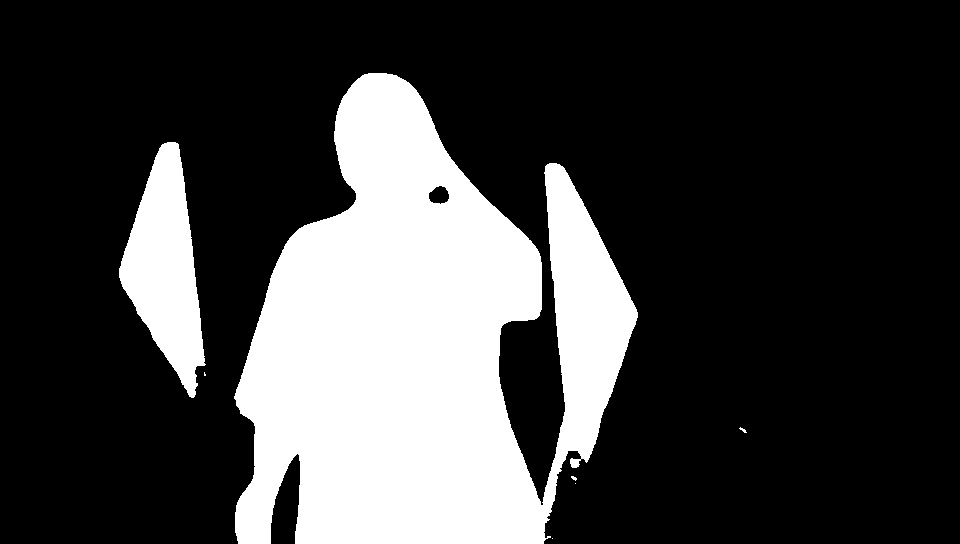}
	\end{subfigure}
	\begin{subfigure}{0.08\textwidth}
		\includegraphics[width=\textwidth]{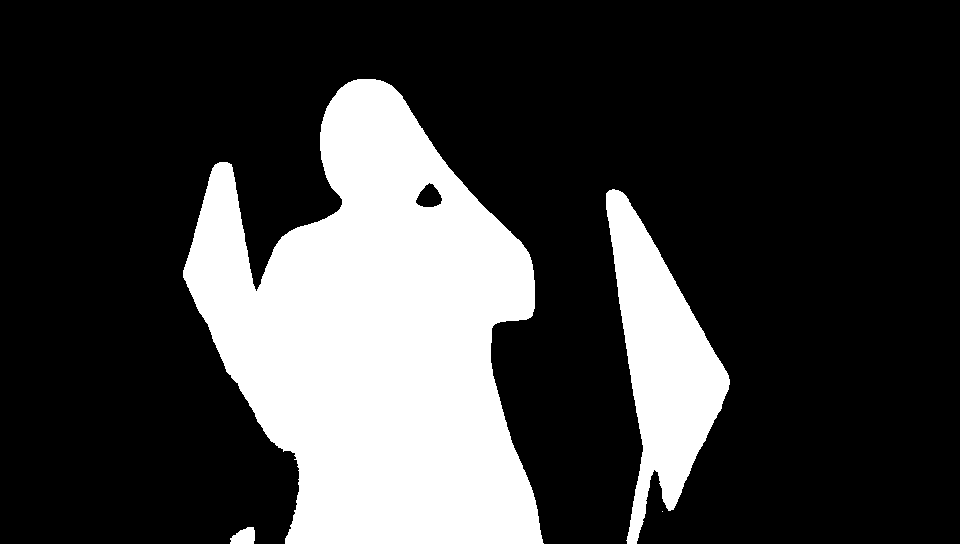}
	\end{subfigure}
	\begin{subfigure}{0.08\textwidth}
		\includegraphics[width=\textwidth]{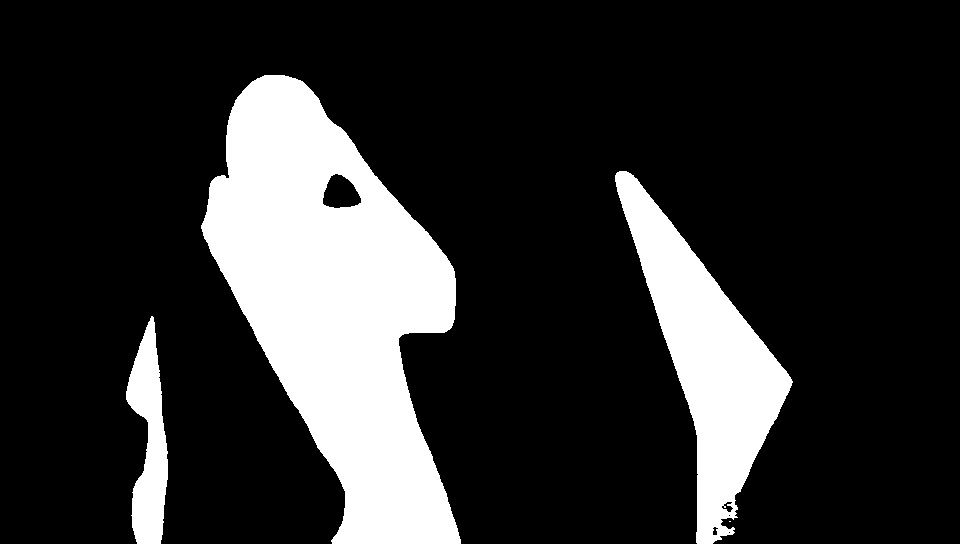}
	\end{subfigure}
	\begin{subfigure}{0.08\textwidth}
		\includegraphics[width=\textwidth]{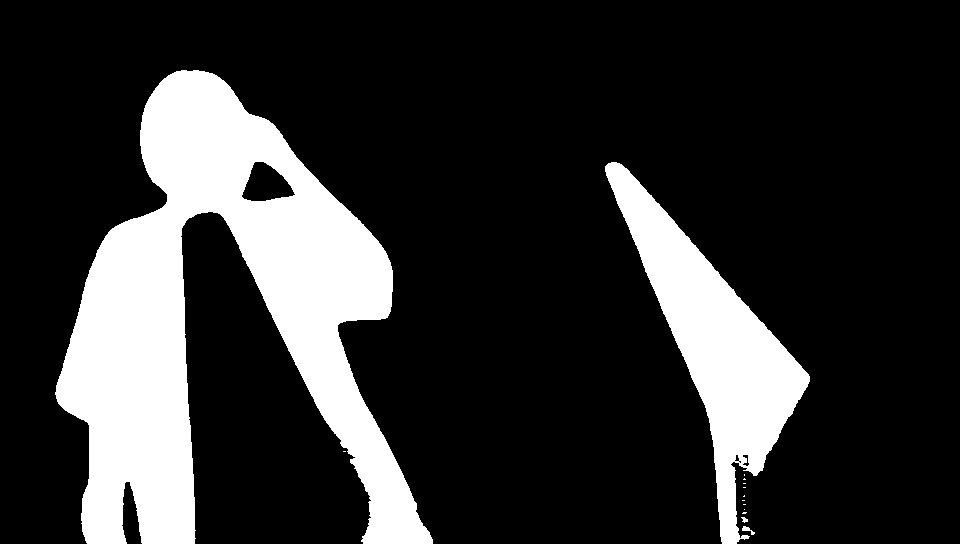}
	\end{subfigure}
	\begin{subfigure}{0.08\textwidth}
		\includegraphics[width=\textwidth]{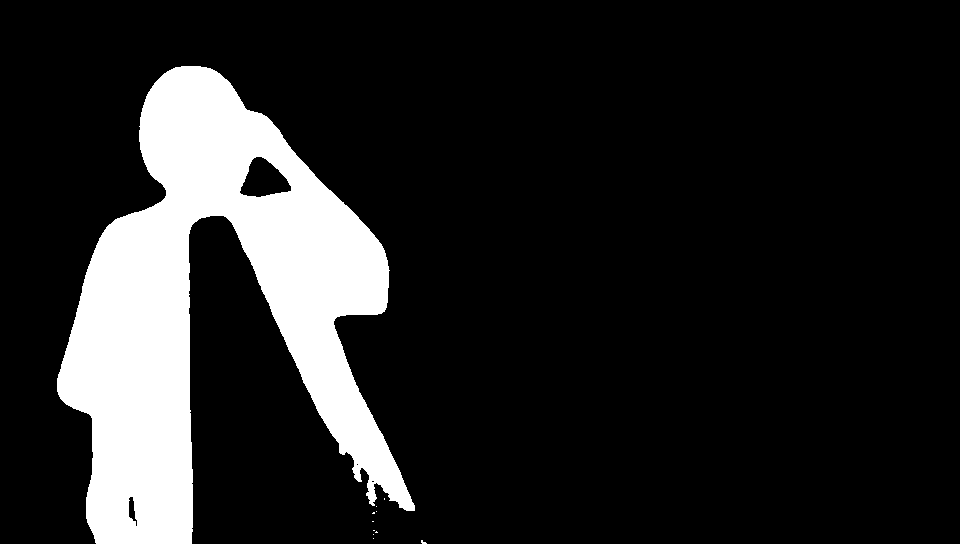}
	\end{subfigure}

	\vspace*{1.3mm}
	\begin{subfigure}{0.08\textwidth}
		\includegraphics[width=\textwidth]{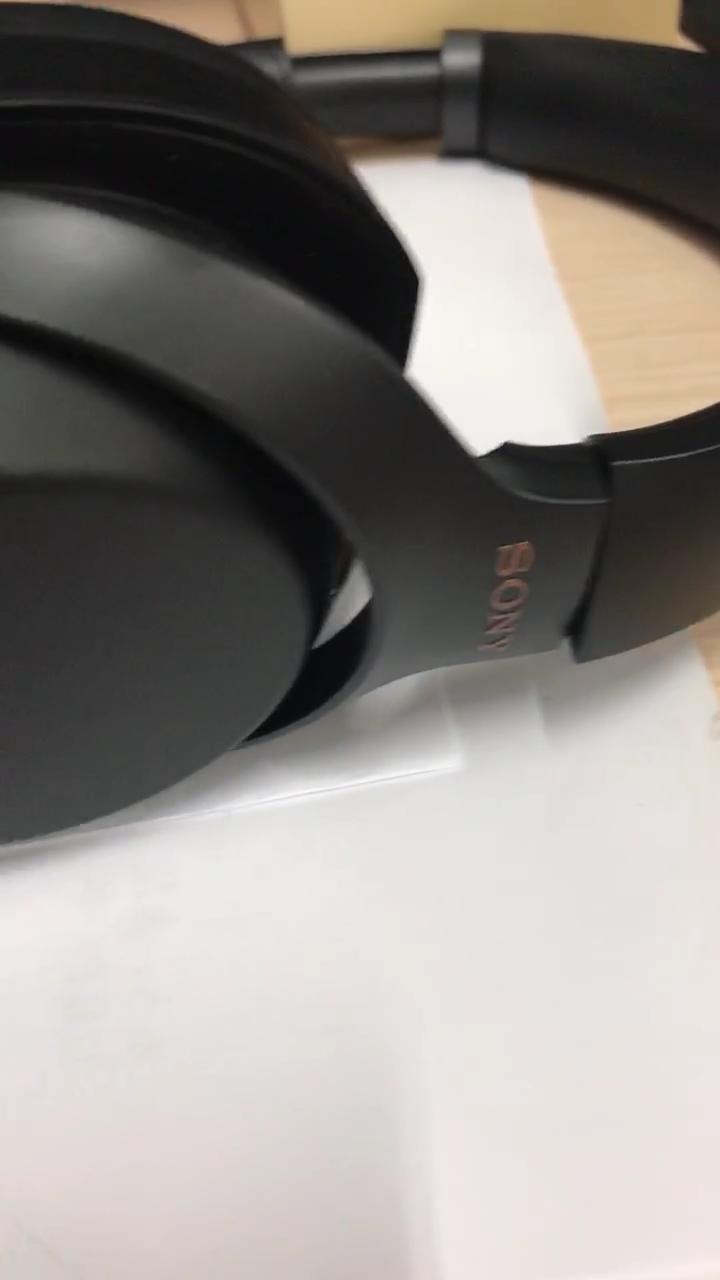}
	\end{subfigure}
	\begin{subfigure}{0.08\textwidth}
		\includegraphics[width=\textwidth]{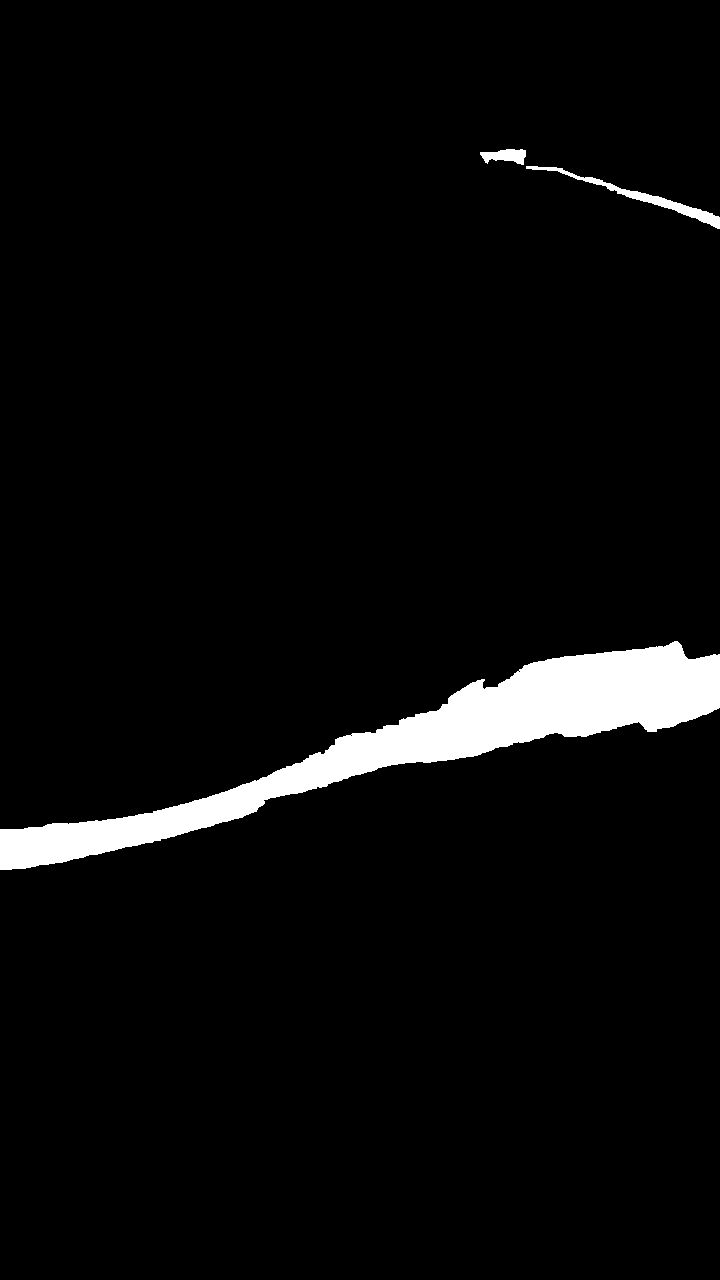}
	\end{subfigure}
	\begin{subfigure}{0.08\textwidth}
		\includegraphics[width=\textwidth]{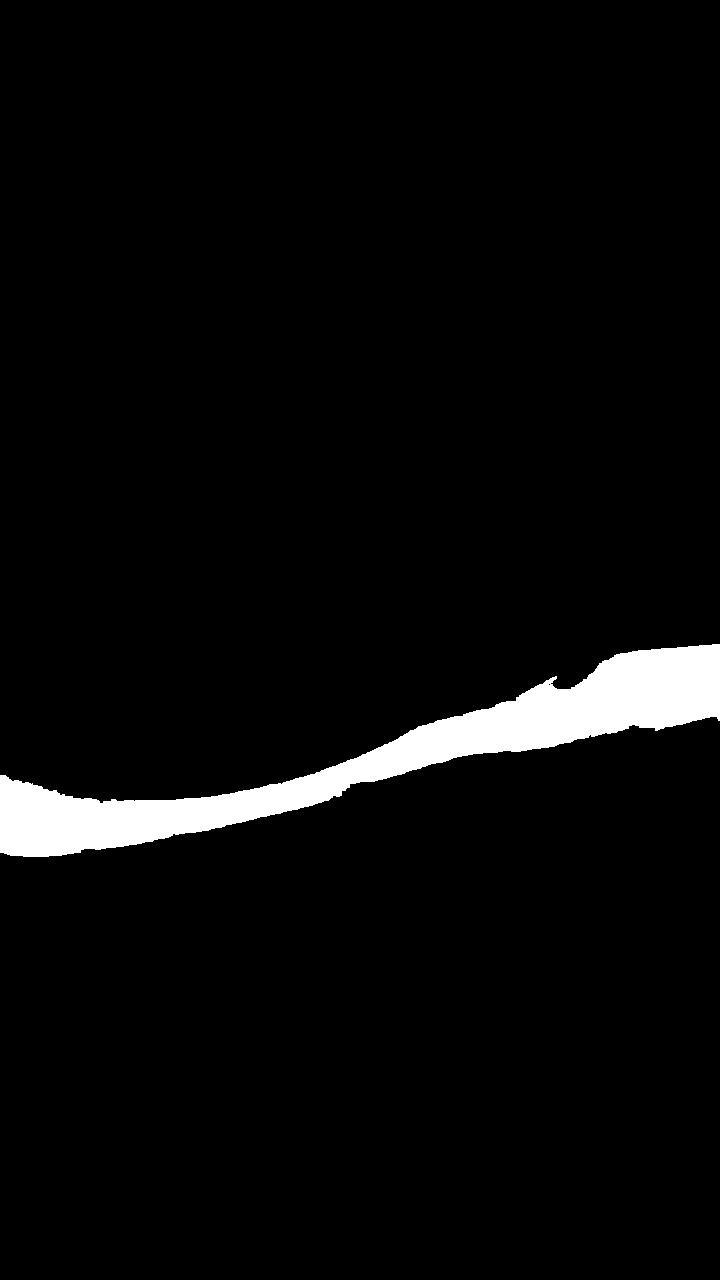}
	\end{subfigure}
	\begin{subfigure}{0.08\textwidth}
		\includegraphics[width=\textwidth]{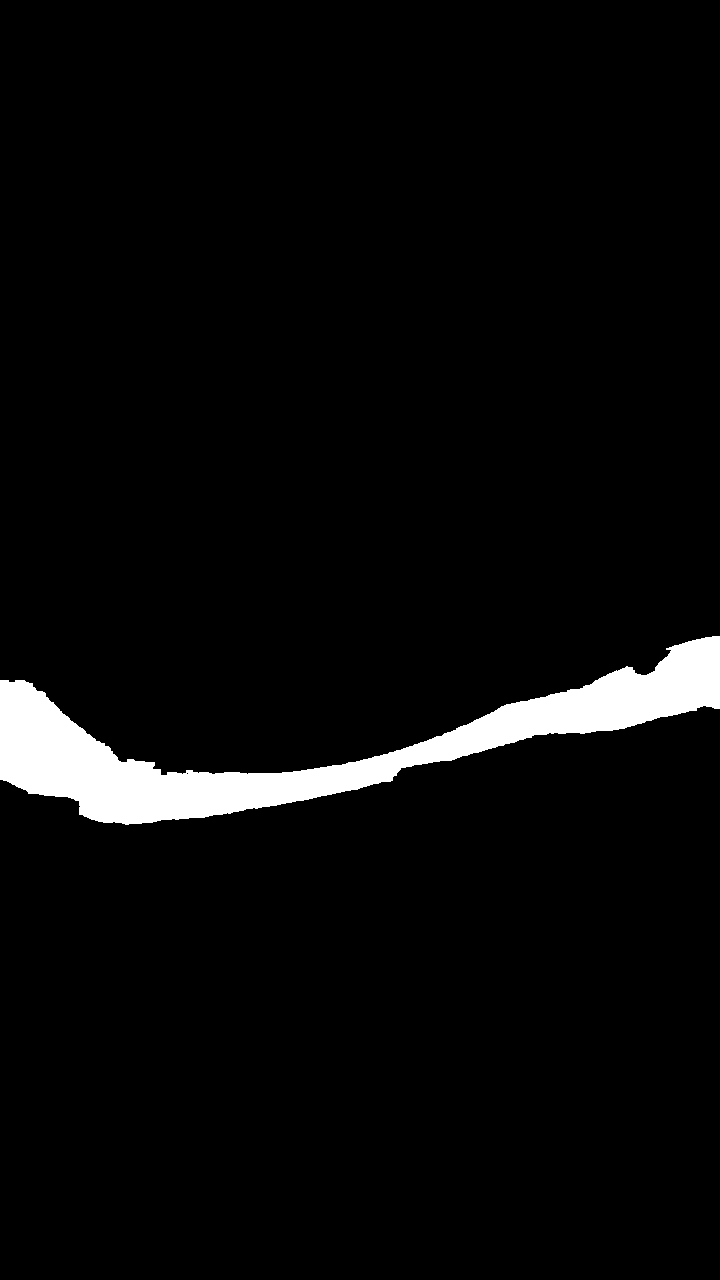}
	\end{subfigure}
	\begin{subfigure}{0.08\textwidth}
		\includegraphics[width=\textwidth]{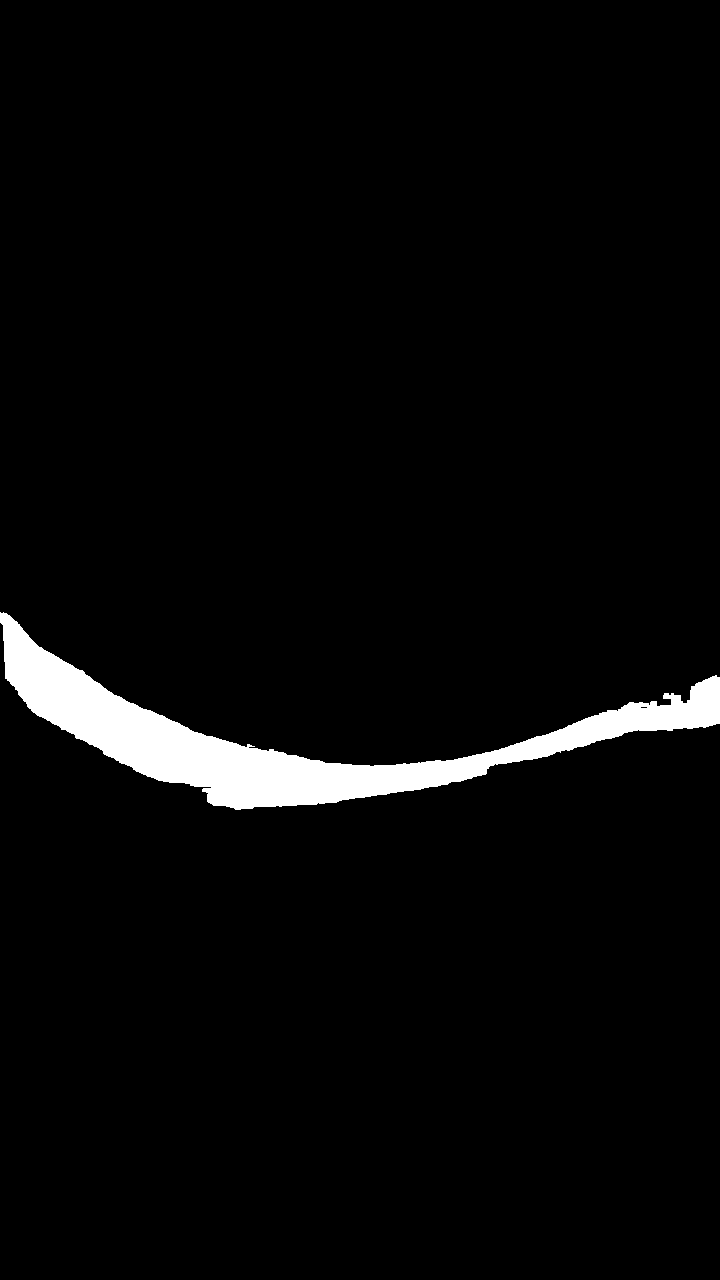}
	\end{subfigure}
	\begin{subfigure}{0.08\textwidth}
		\includegraphics[width=\textwidth]{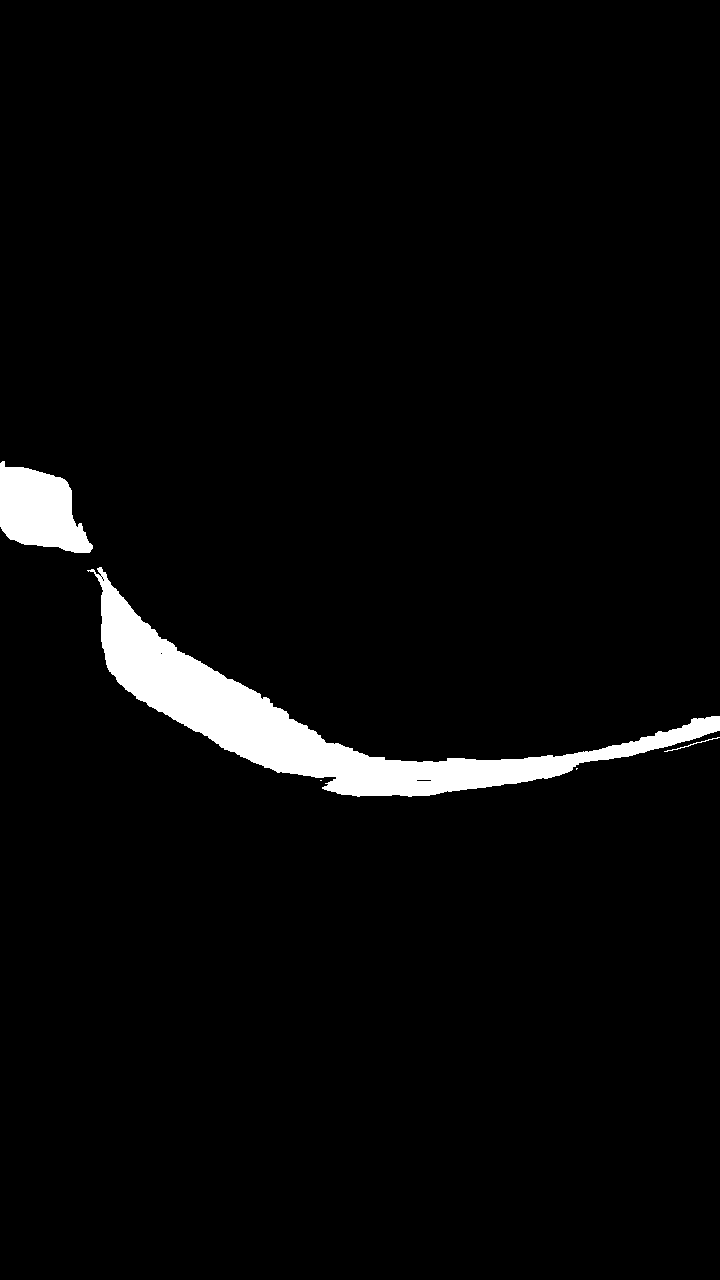}
	\end{subfigure}
	\begin{subfigure}{0.08\textwidth}
		\includegraphics[width=\textwidth]{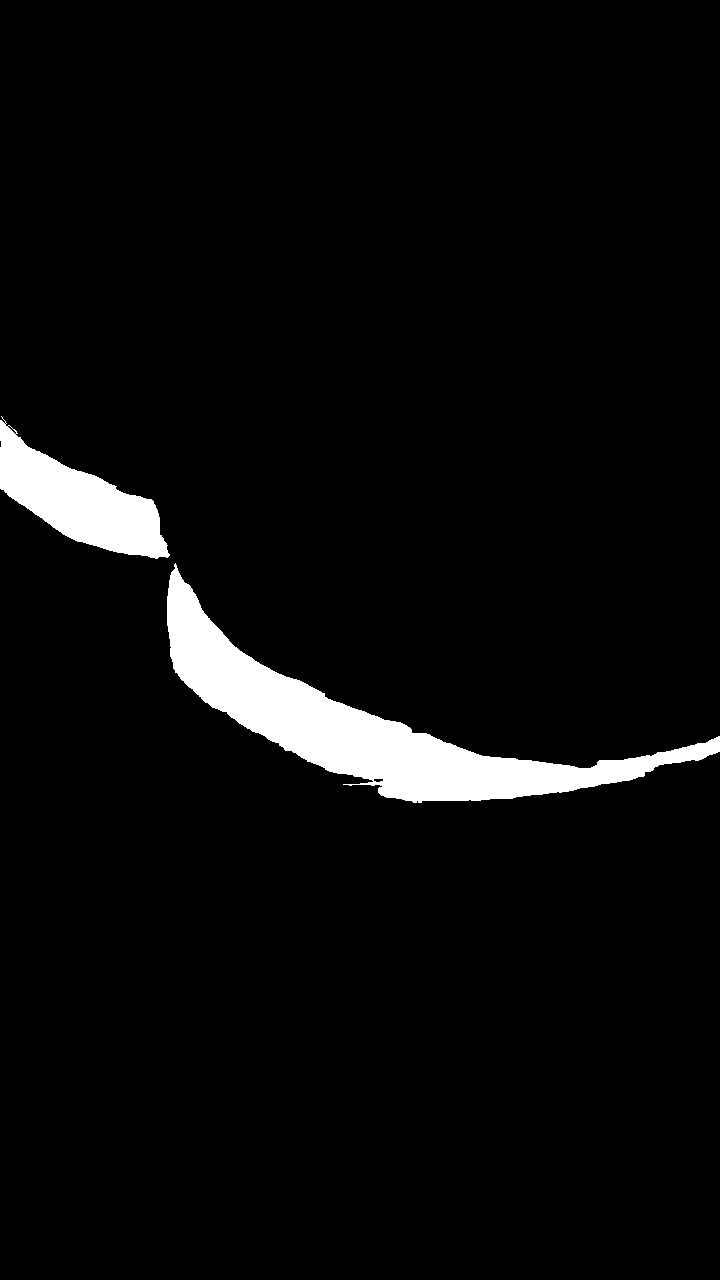}
	\end{subfigure}
	\begin{subfigure}{0.08\textwidth}
		\includegraphics[width=\textwidth]{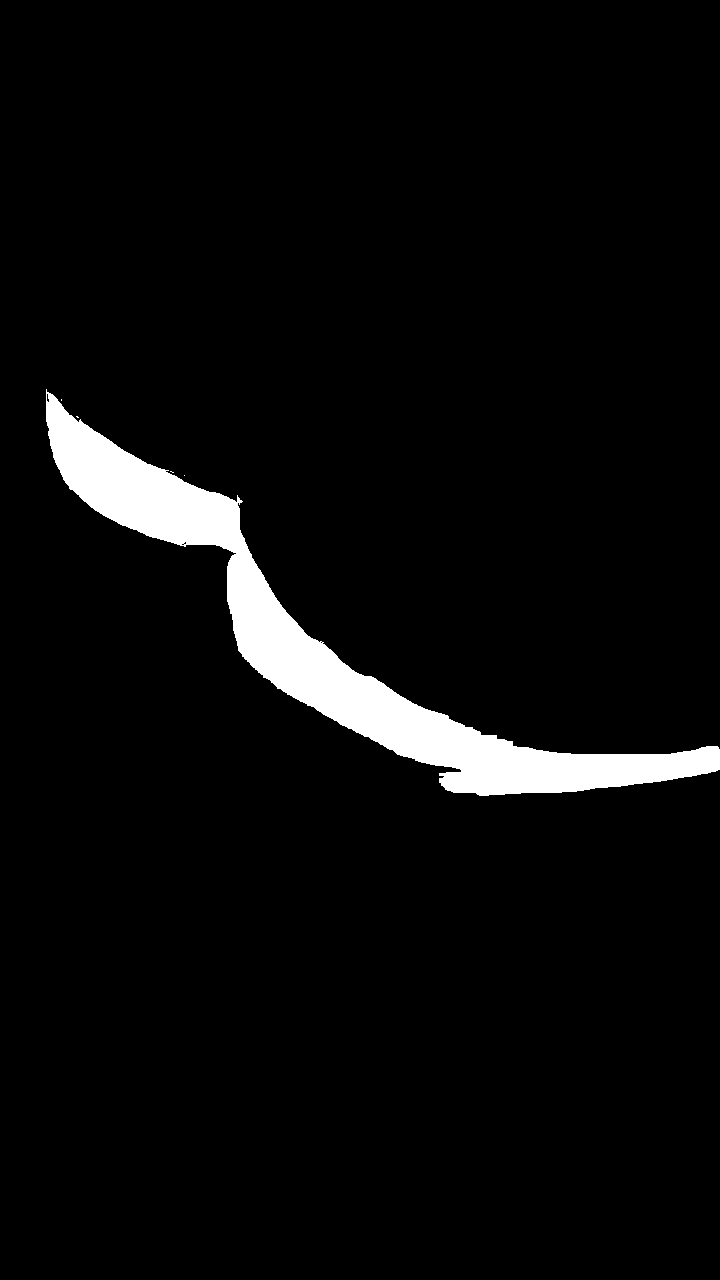}
	\end{subfigure}
	\begin{subfigure}{0.08\textwidth}
		\includegraphics[width=\textwidth]{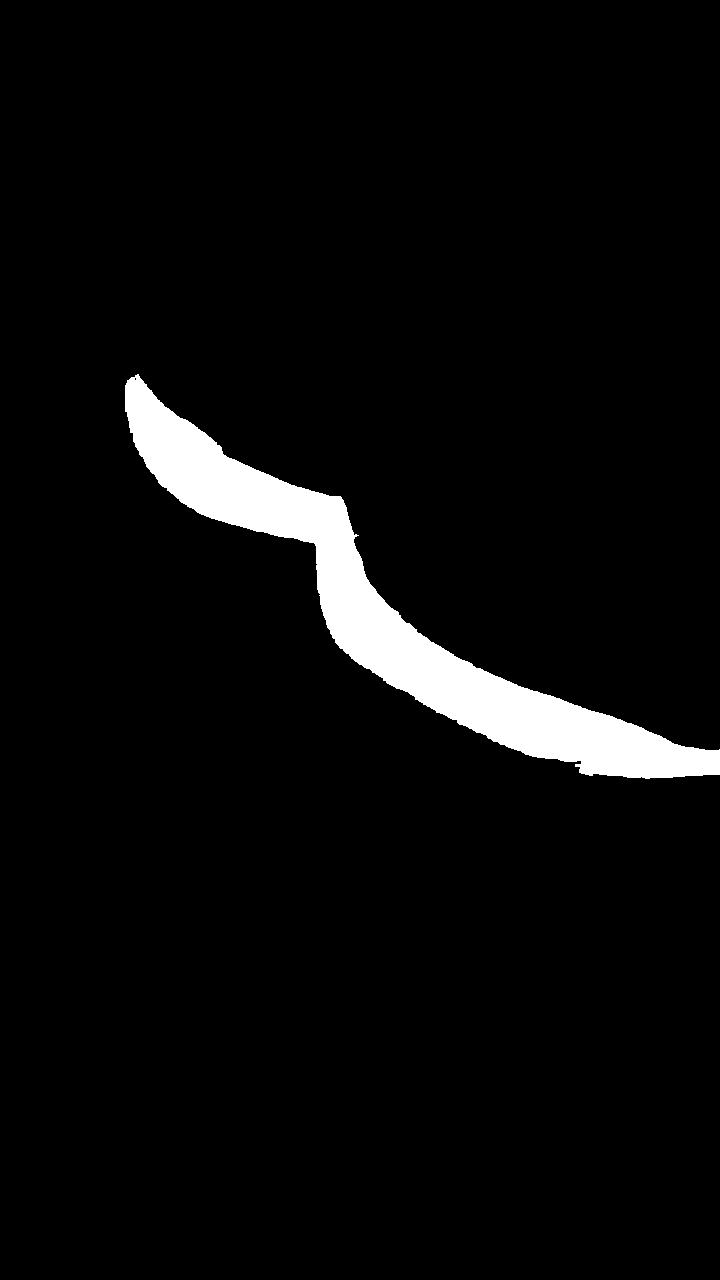}
	\end{subfigure}
	\begin{subfigure}{0.08\textwidth}
		\includegraphics[width=\textwidth]{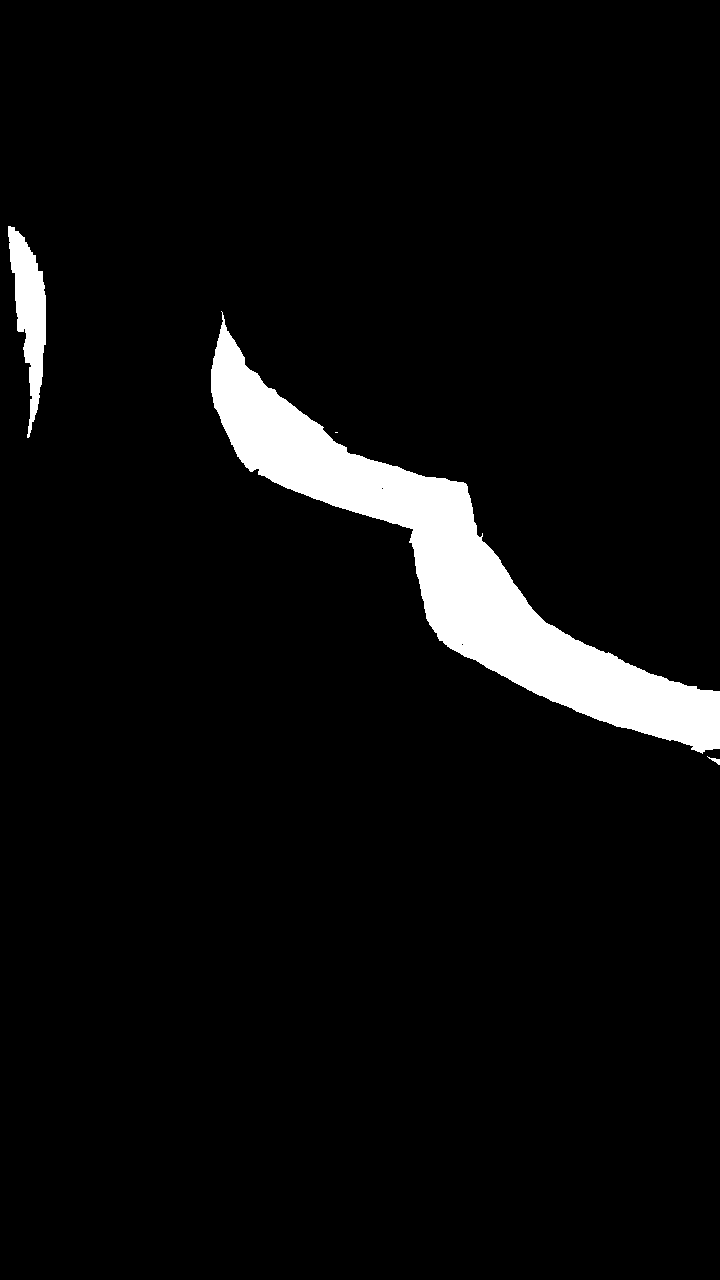}
	\end{subfigure}
	
	\vspace*{1.3mm}
	\begin{subfigure}{0.08\textwidth}
		\includegraphics[width=\textwidth]{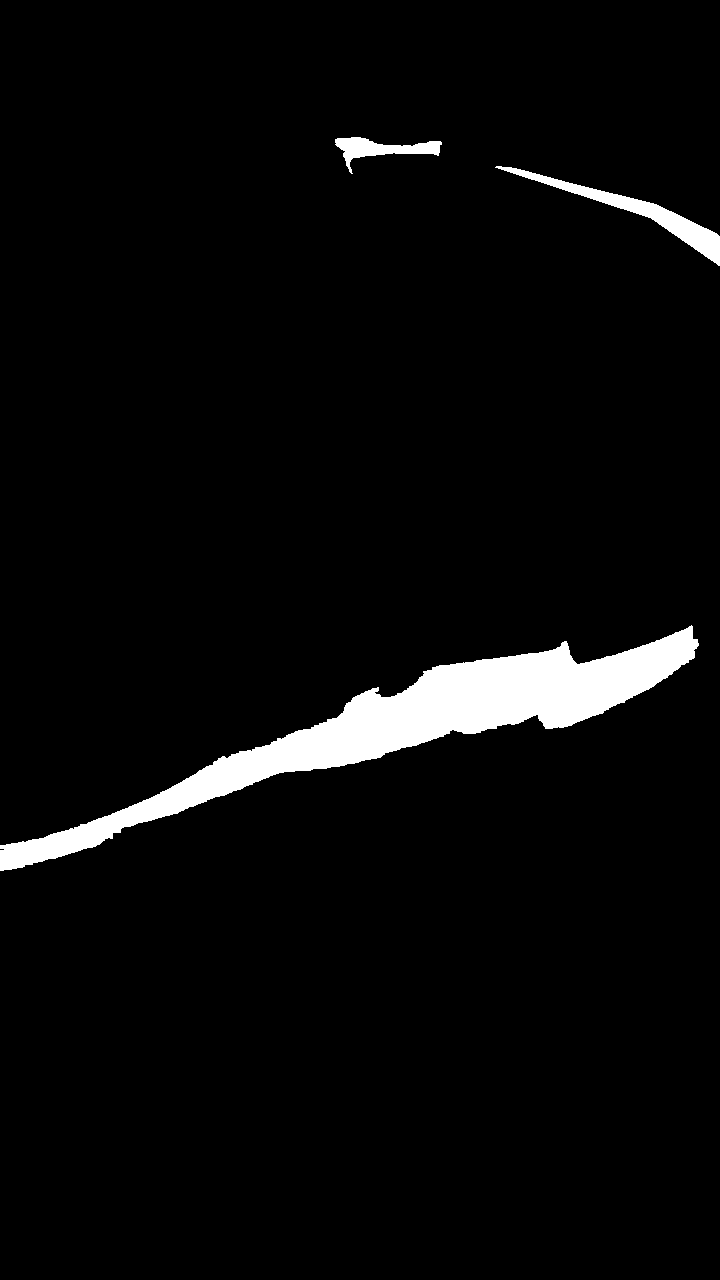}
		\captionsetup{justification=centering}
        \vspace{-5.5mm} \caption{\footnotesize{\\RGB/GT}}
	\end{subfigure}
	\begin{subfigure}{0.08\textwidth}
		\includegraphics[width=\textwidth]{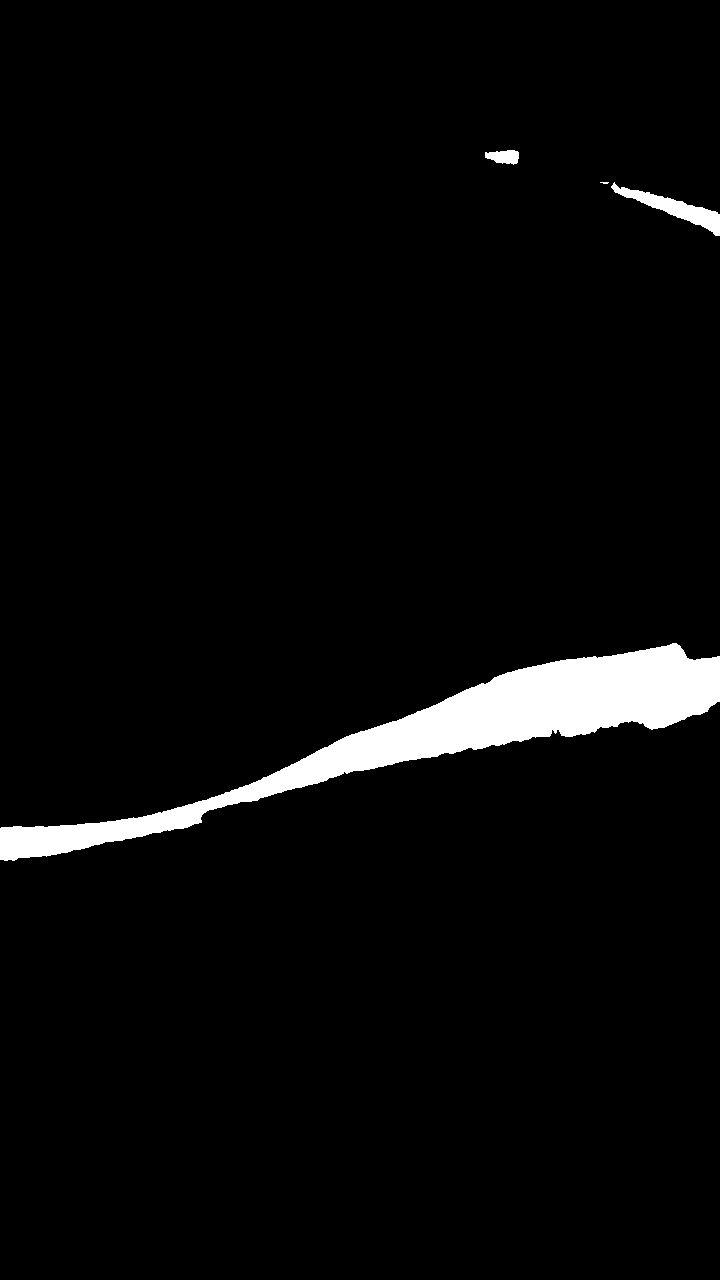}
		\captionsetup{justification=centering}
        \vspace{-5.5mm} \caption{\footnotesize{\\11th}}
	\end{subfigure}
	\begin{subfigure}{0.08\textwidth}
		\includegraphics[width=\textwidth]{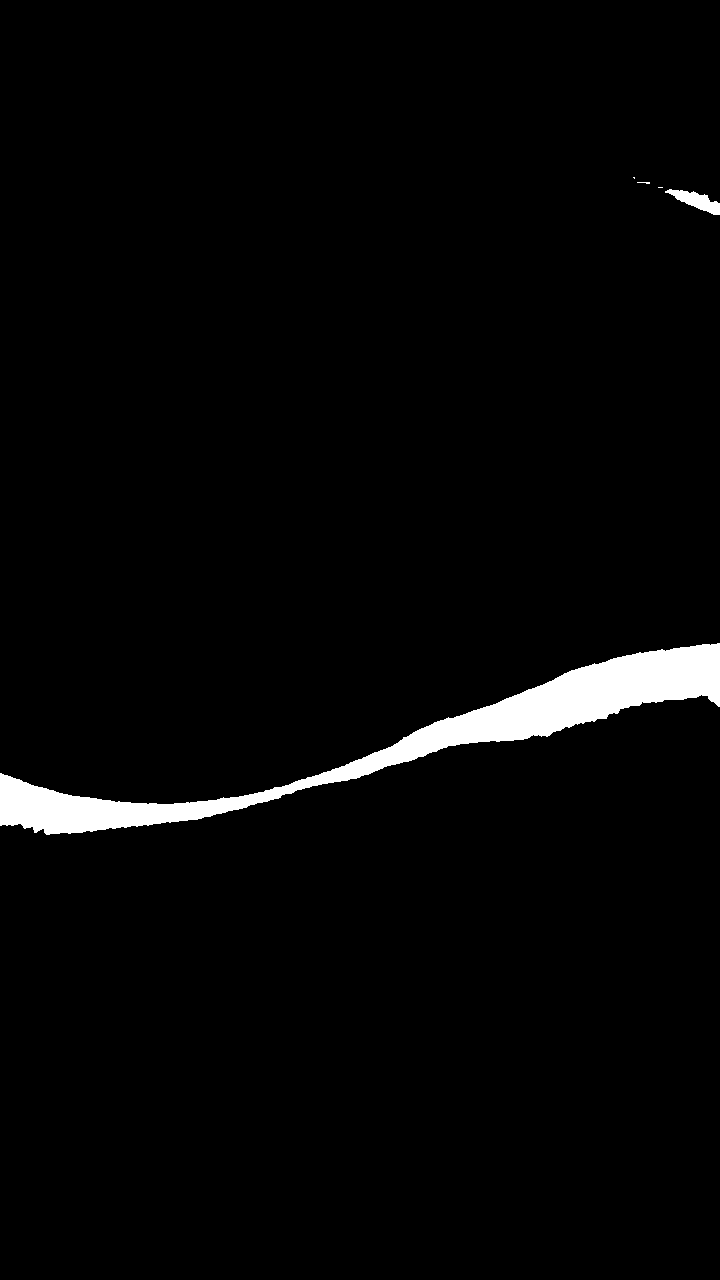}
		\captionsetup{justification=centering}
        \vspace{-5.5mm} \caption{\footnotesize{\\21th}}
	\end{subfigure}
	\begin{subfigure}{0.08\textwidth}
		\includegraphics[width=\textwidth]{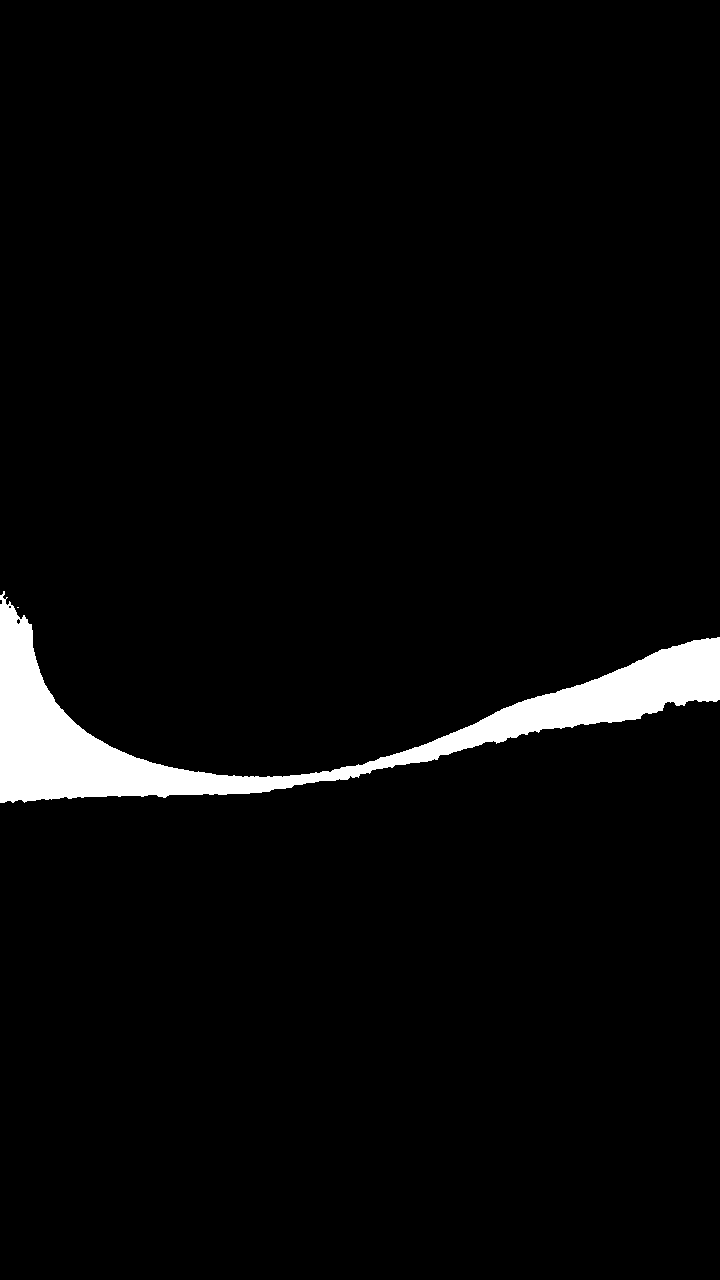}
		\captionsetup{justification=centering}
        \vspace{-5.5mm} \caption{\footnotesize{\\31th}}
	\end{subfigure}
	\begin{subfigure}{0.08\textwidth}
		\includegraphics[width=\textwidth]{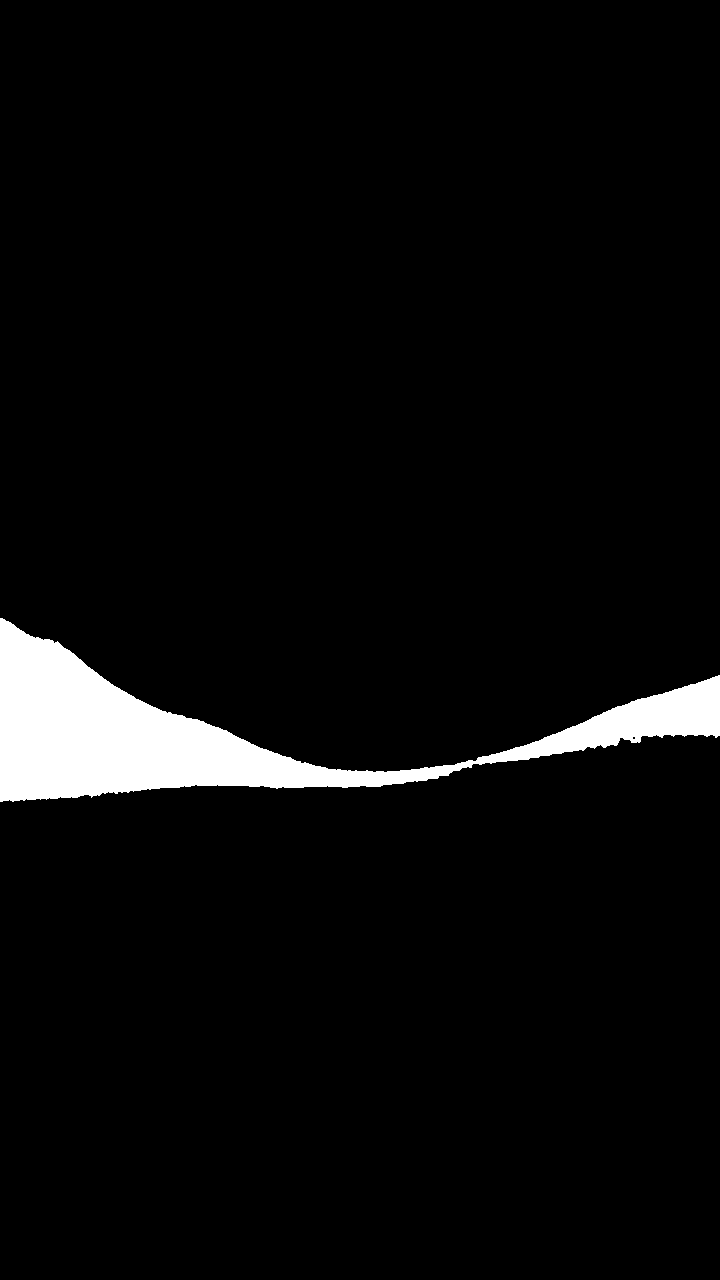}
		\captionsetup{justification=centering}
        \vspace{-5.5mm} \caption{\footnotesize{\\41th}}
	\end{subfigure}
	\begin{subfigure}{0.08\textwidth}
		\includegraphics[width=\textwidth]{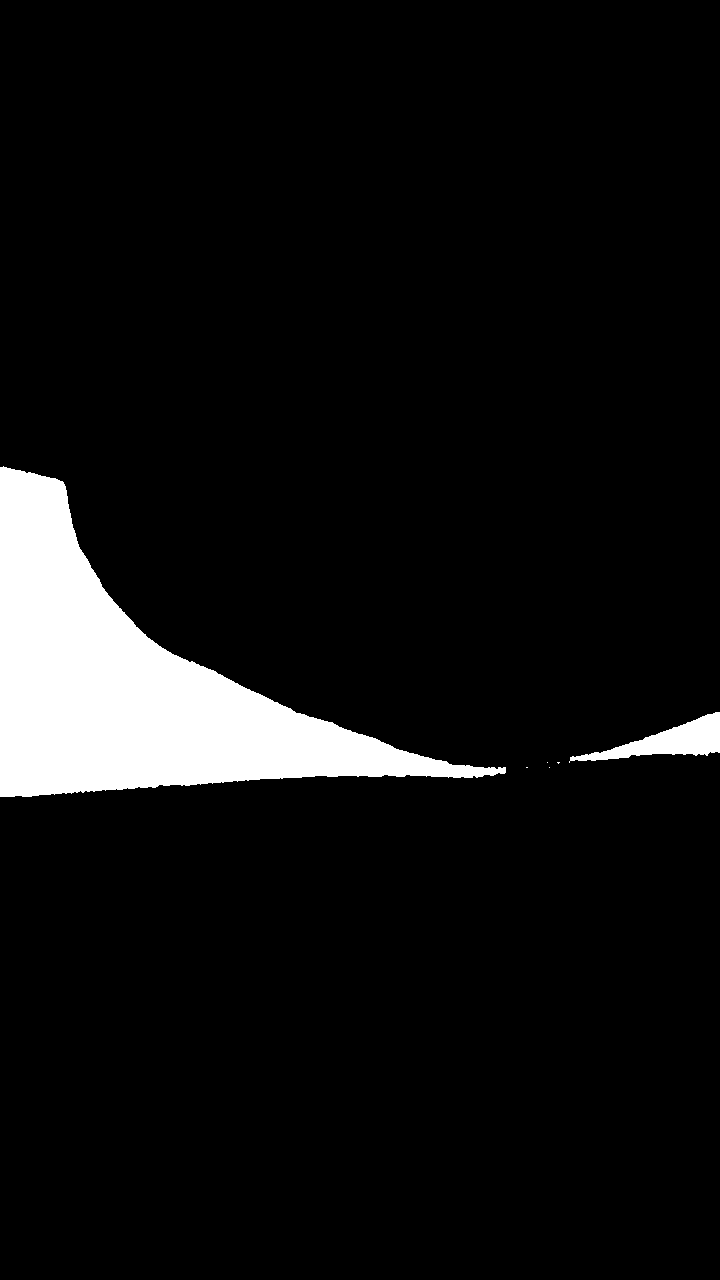}
		\captionsetup{justification=centering}
        \vspace{-5.5mm} \caption{\footnotesize{\\51th}}
	\end{subfigure}
	\begin{subfigure}{0.08\textwidth}
		\includegraphics[width=\textwidth]{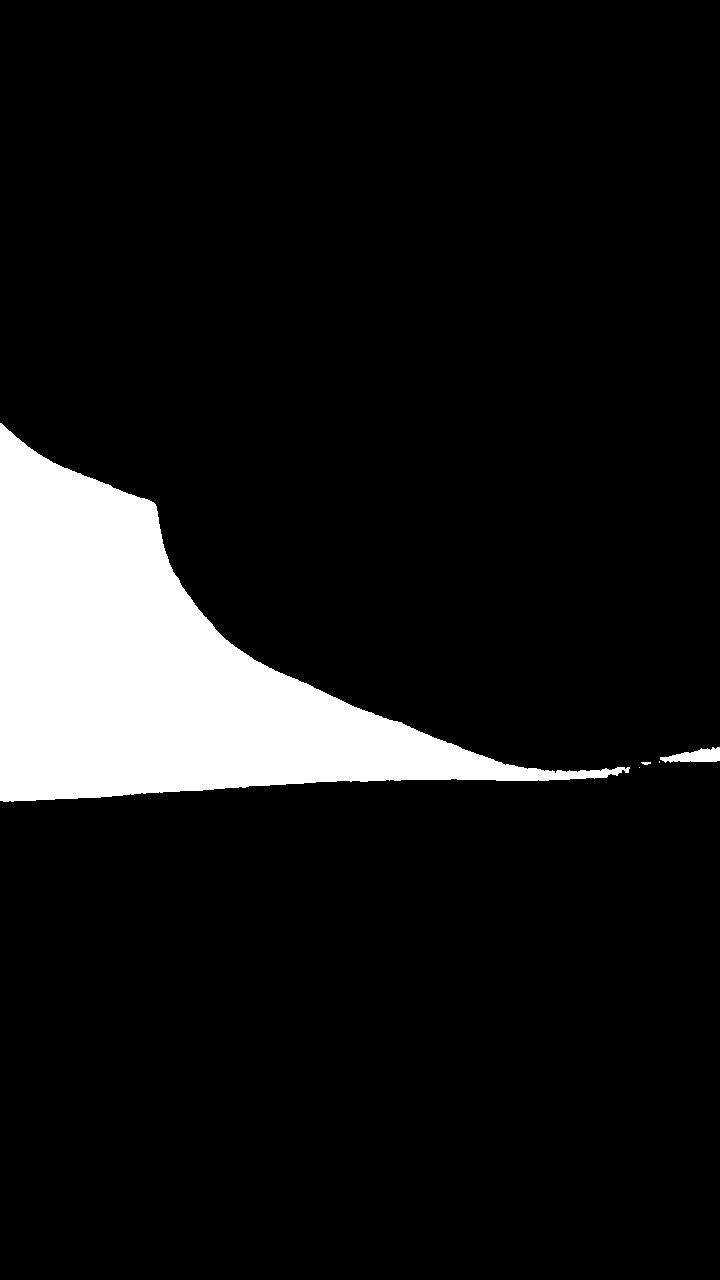}
		\captionsetup{justification=centering}
        \vspace{-5.5mm} \caption{\footnotesize{\\61th}}
	\end{subfigure}
	\begin{subfigure}{0.08\textwidth}
		\includegraphics[width=\textwidth]{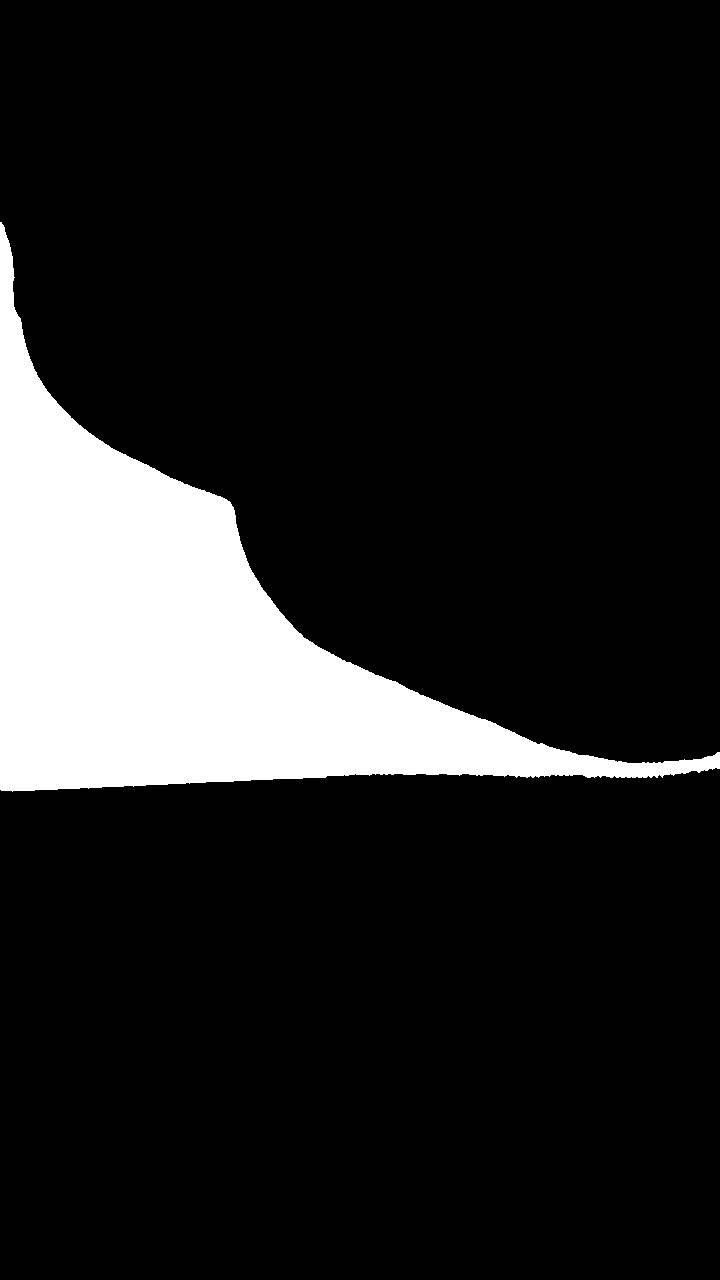}
		\captionsetup{justification=centering}
        \vspace{-5.5mm} \caption{\footnotesize{\\71th}}
	\end{subfigure}
	\begin{subfigure}{0.08\textwidth}
		\includegraphics[width=\textwidth]{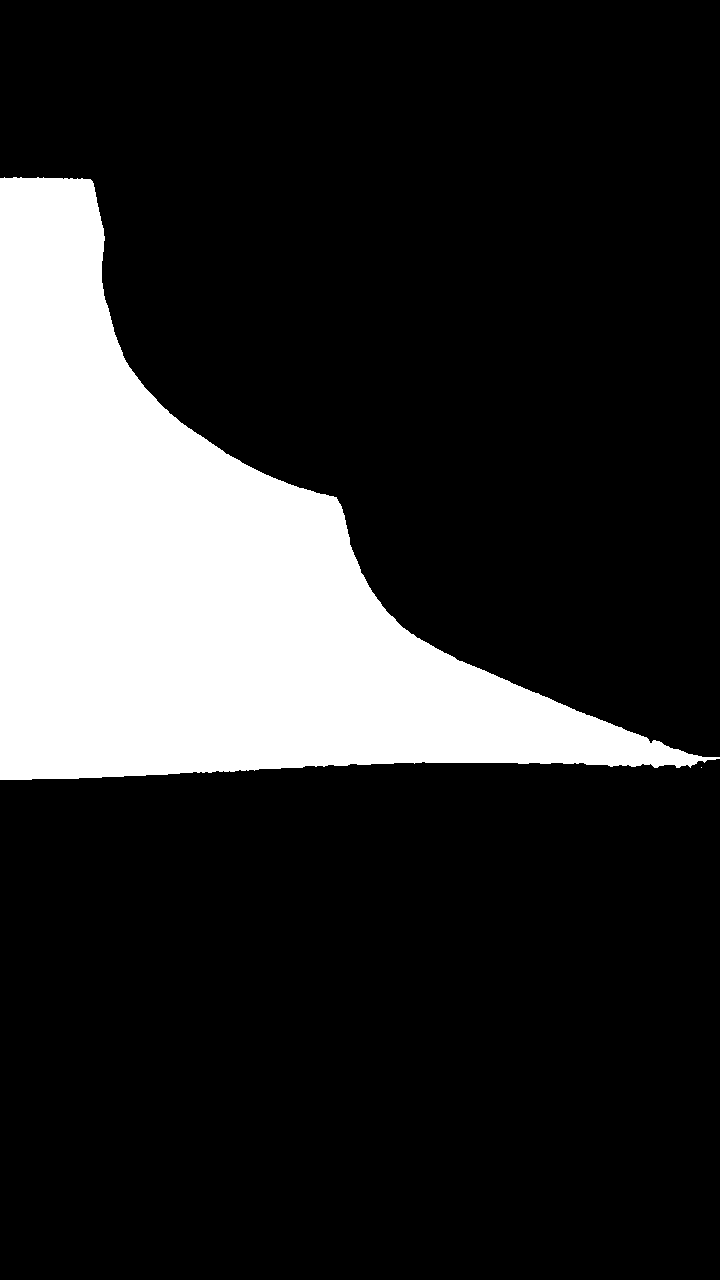}
		\captionsetup{justification=centering}
        \vspace{-5.5mm} \caption{\footnotesize{\\81th}}
	\end{subfigure}
	\begin{subfigure}{0.08\textwidth}
		\includegraphics[width=\textwidth]{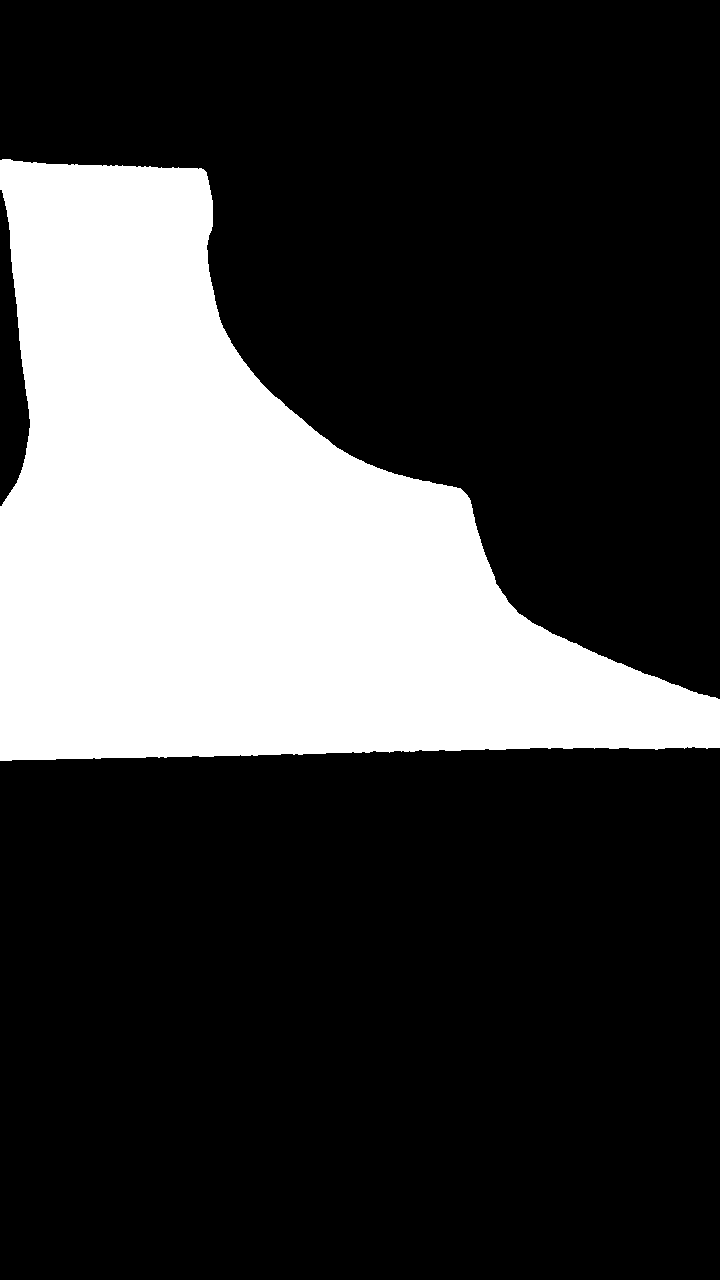}
		\captionsetup{justification=centering}
        \vspace{-5.5mm} \caption{\footnotesize{\\91th}}
	\end{subfigure}

	\caption{Qualitative comparison of the predicted segmentations using mask prompts on the Visha dataset. The first images of every two rows represents the rgb and groud truth image of the 1st frame, while the other images shown in the even row and in the odd row are the ground truth and predicted shadow masks for the 11th, 21th, 31th, 41th, 51th, 61th, 71th, 81th and 91th, respectively. Best viewed on screen.}
	\label{fig:fig_visha_mask}
	
\end{figure*}

\begin{figure*}[!ht]
	\centering
	\vspace*{1.3mm}
	\begin{subfigure}{0.11\textwidth}
		\includegraphics[width=\textwidth]{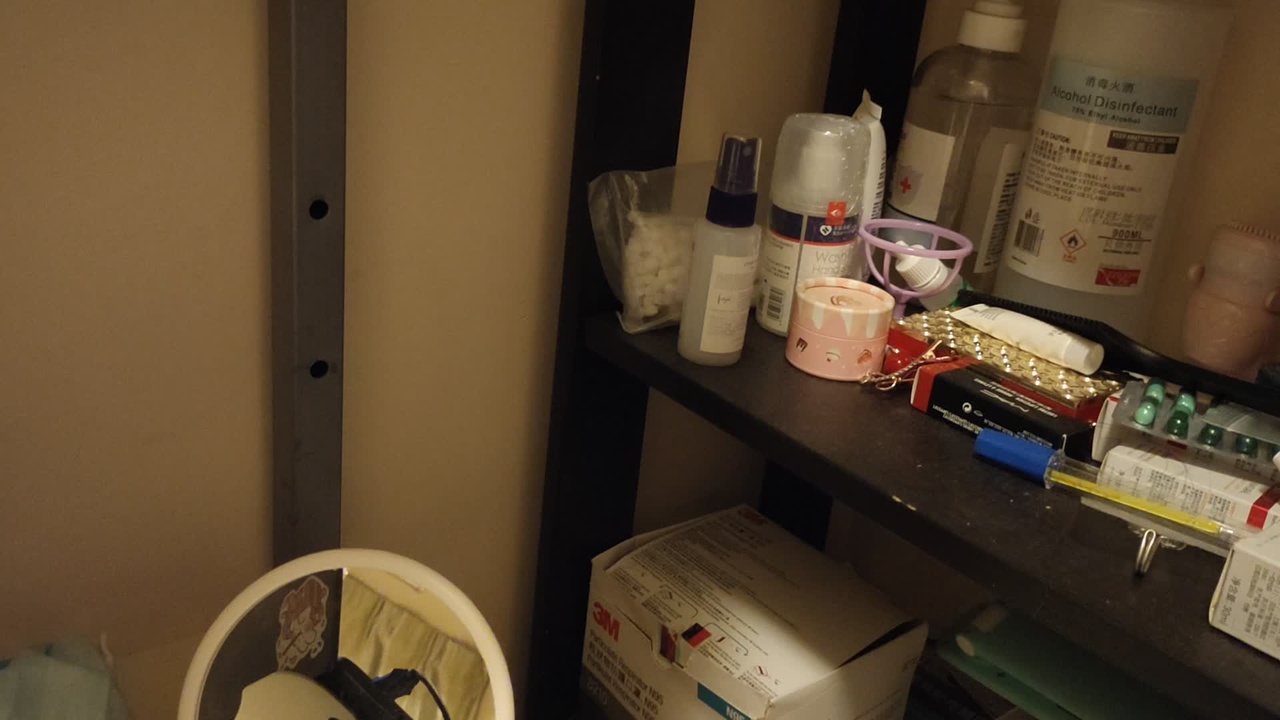}
	\end{subfigure}
	\begin{subfigure}{0.11\textwidth}
		\includegraphics[width=\textwidth]{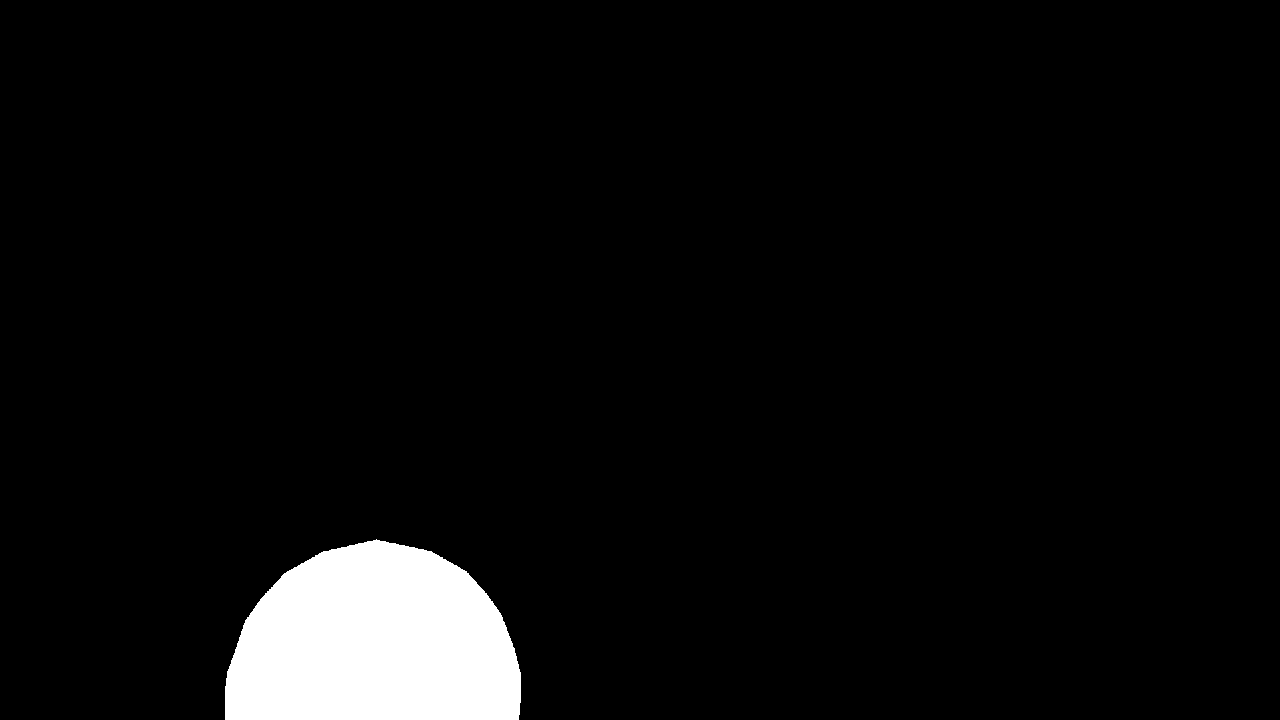}
	\end{subfigure}
	\begin{subfigure}{0.11\textwidth}
		\includegraphics[width=\textwidth]{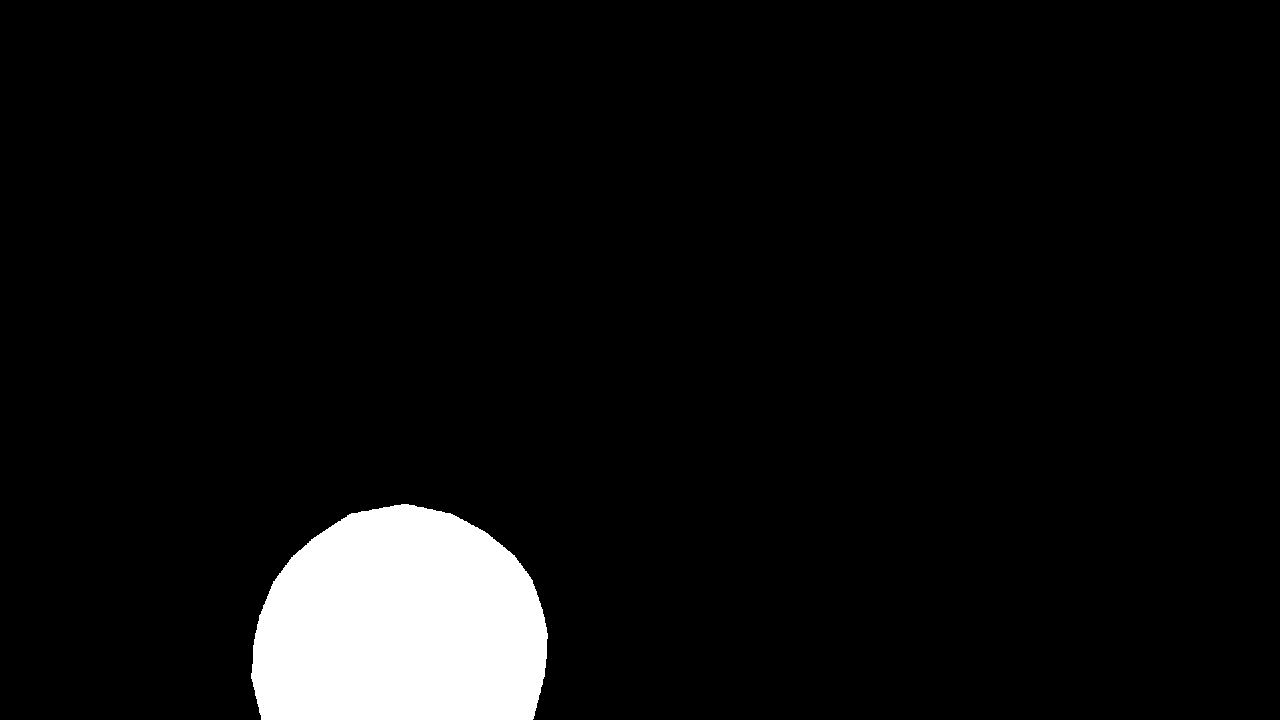}
	\end{subfigure}
	\begin{subfigure}{0.11\textwidth}
		\includegraphics[width=\textwidth]{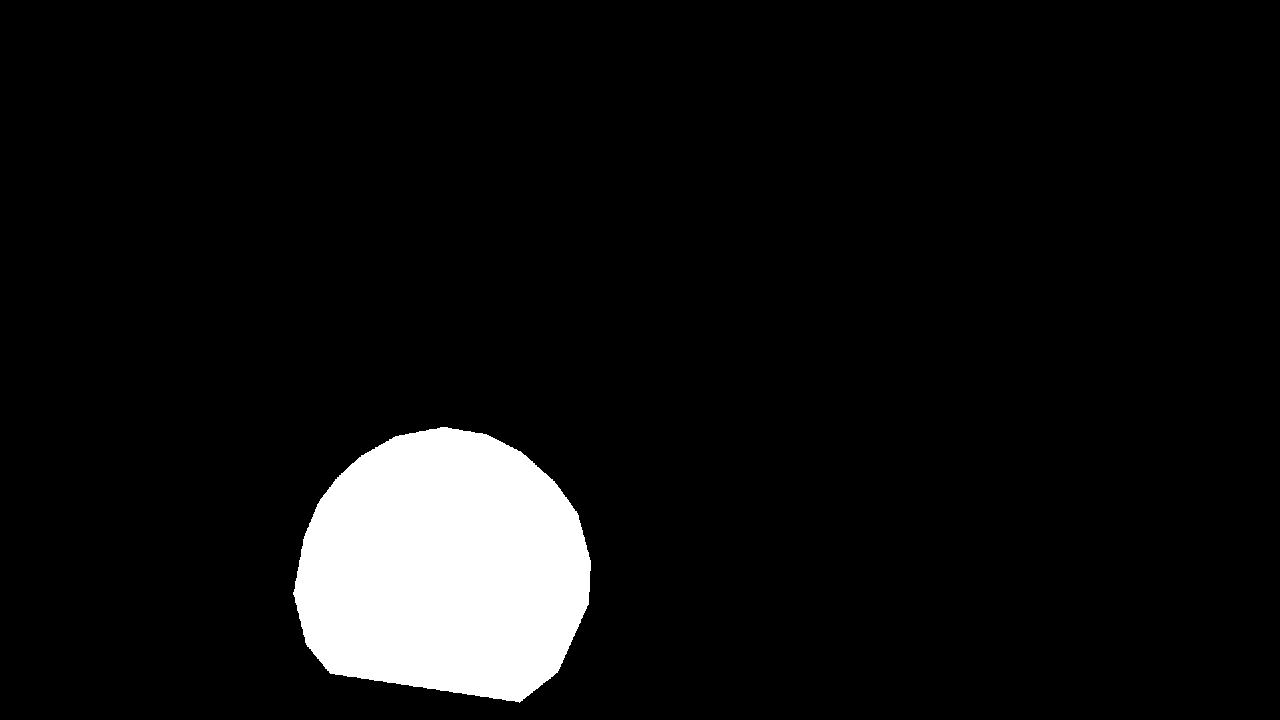}
	\end{subfigure}
	\begin{subfigure}{0.11\textwidth}
		\includegraphics[width=\textwidth]{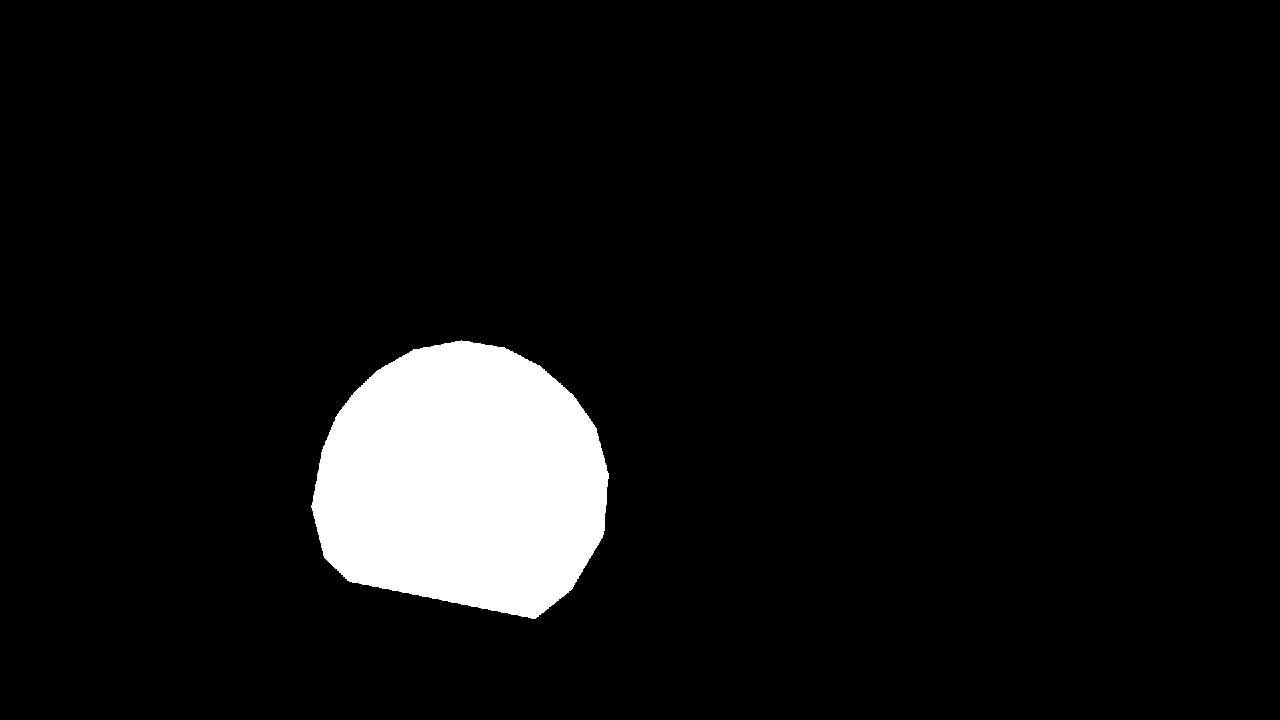}
	\end{subfigure}
	\begin{subfigure}{0.11\textwidth}
		\includegraphics[width=\textwidth]{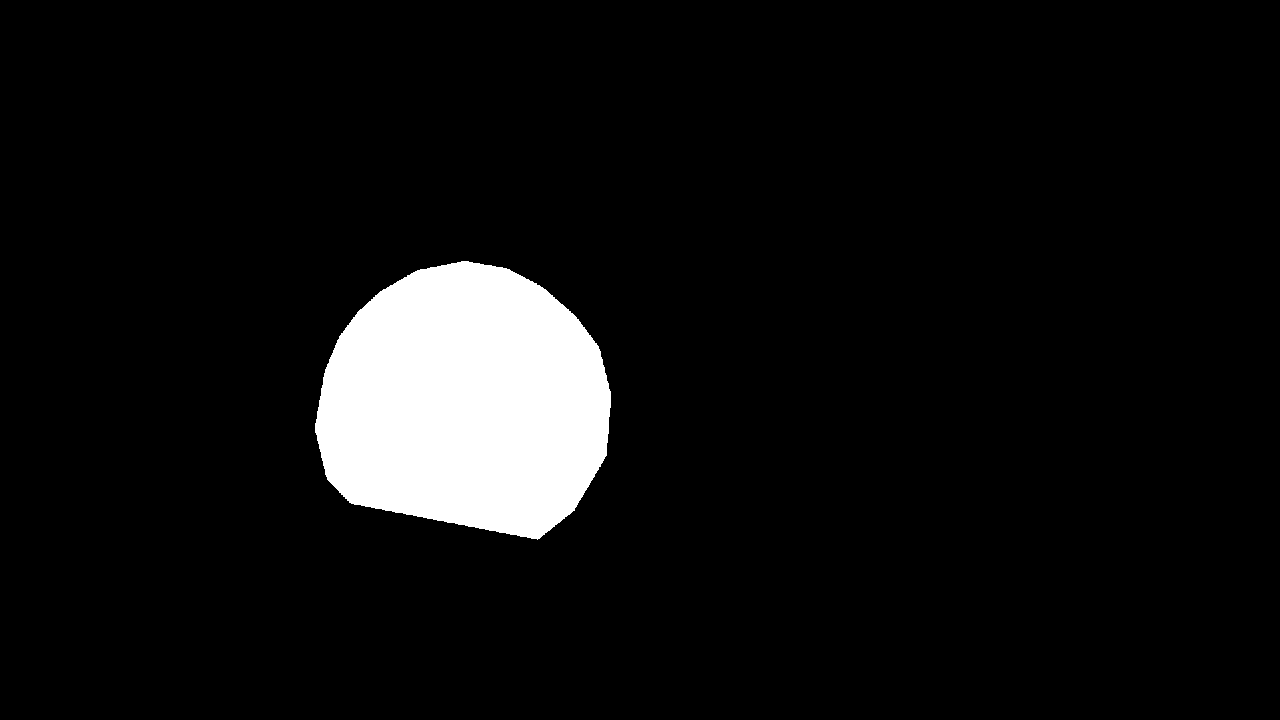}
	\end{subfigure}

	\vspace*{1.3mm}
	\begin{subfigure}{0.11\textwidth}
		\includegraphics[width=\textwidth]{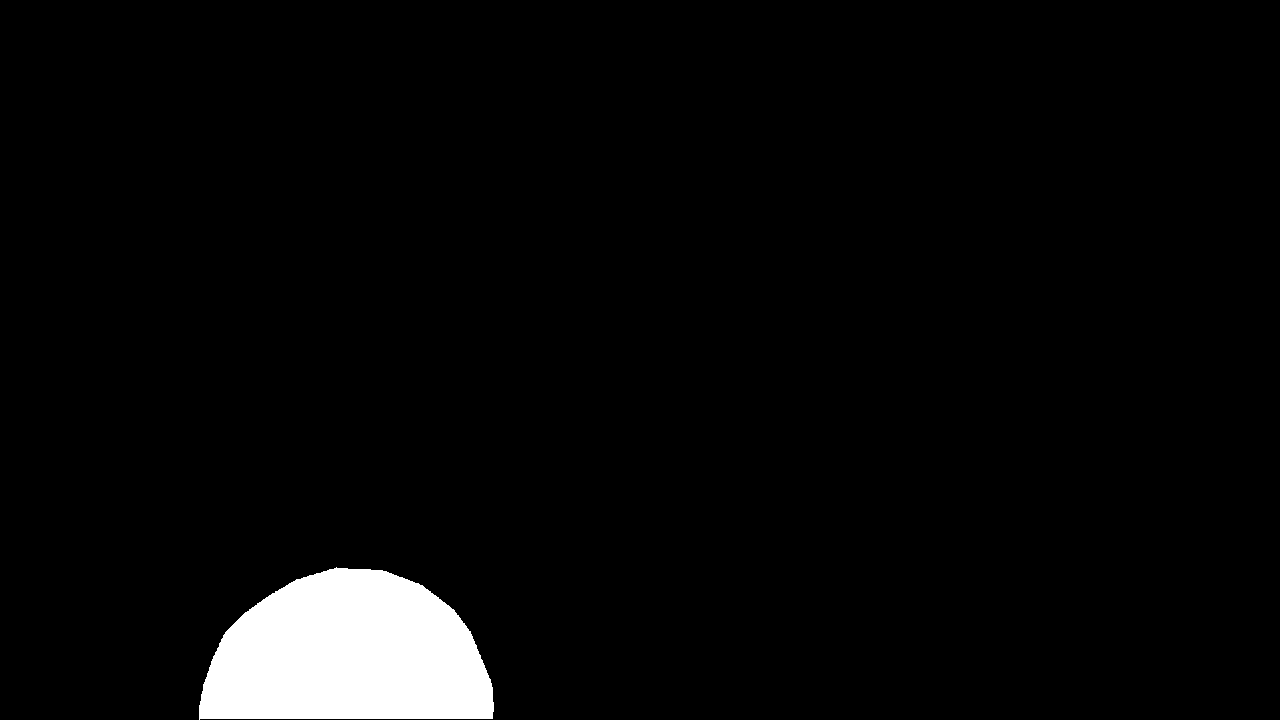}
	\end{subfigure}
	\begin{subfigure}{0.11\textwidth}
		\includegraphics[width=\textwidth]{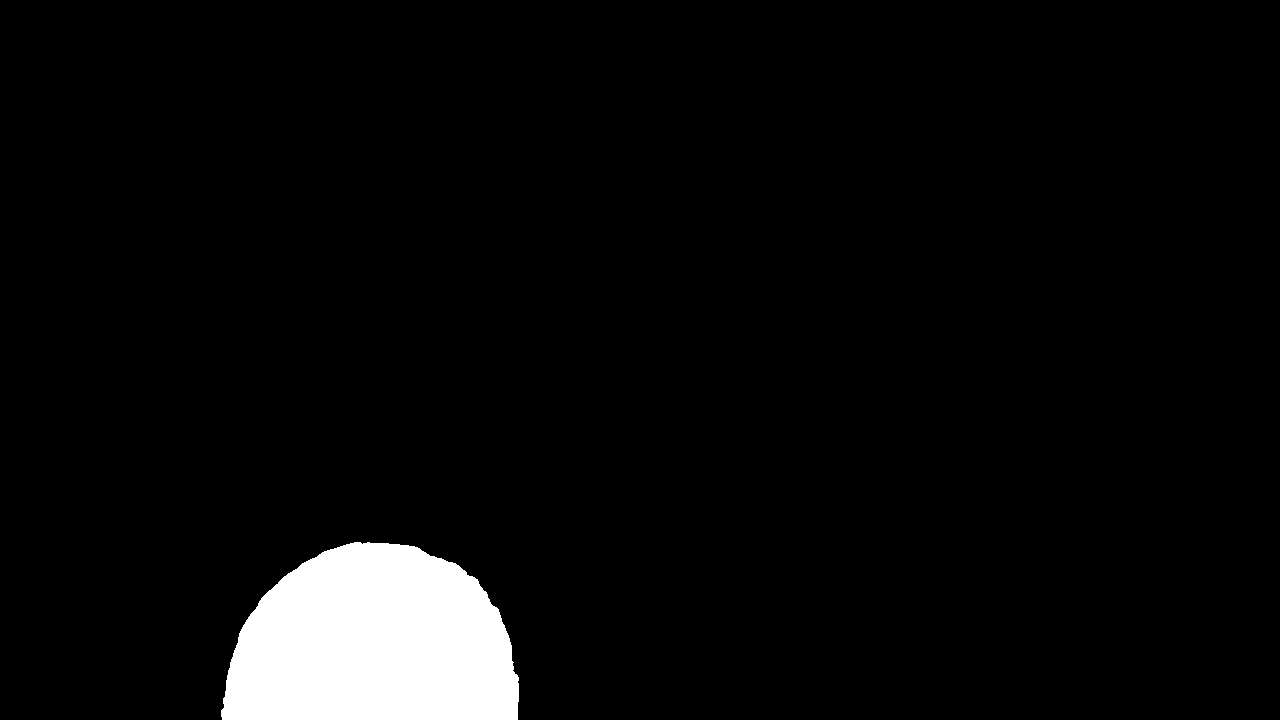}
	\end{subfigure}
	\begin{subfigure}{0.11\textwidth}
		\includegraphics[width=\textwidth]{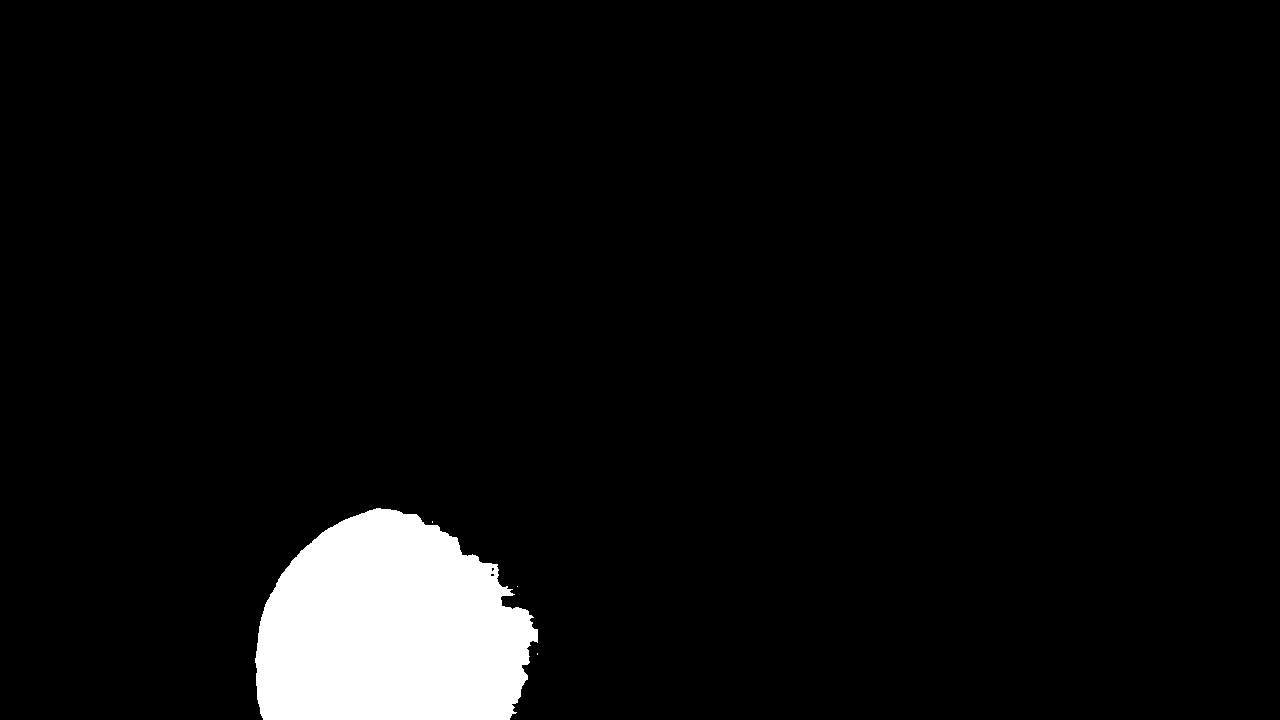}
	\end{subfigure}
	\begin{subfigure}{0.11\textwidth}
		\includegraphics[width=\textwidth]{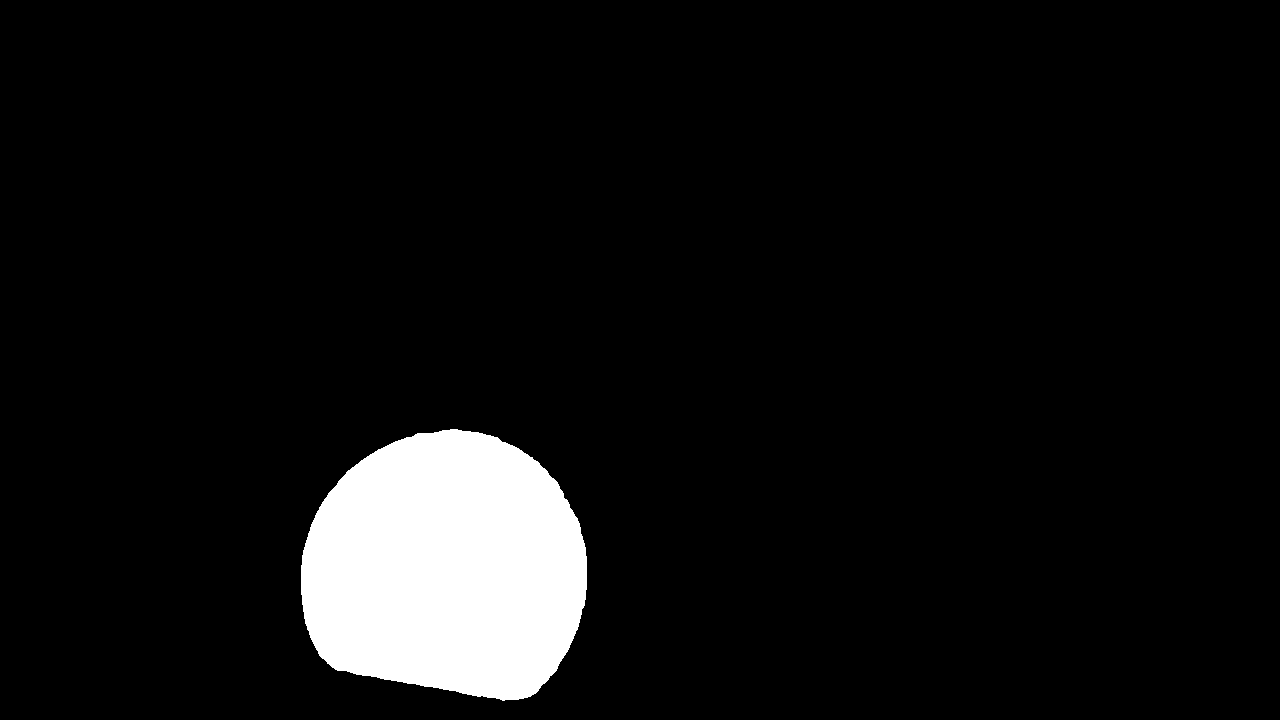}
	\end{subfigure}
	\begin{subfigure}{0.11\textwidth}
		\includegraphics[width=\textwidth]{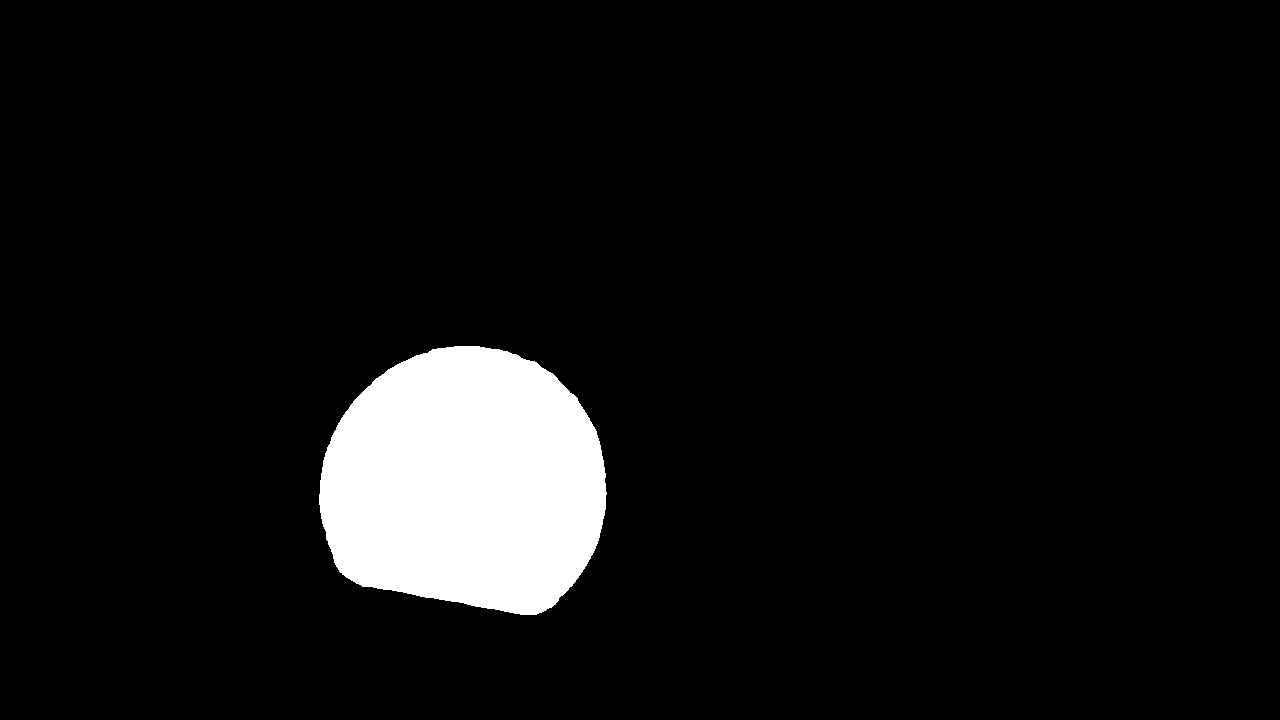}
	\end{subfigure}
	\begin{subfigure}{0.11\textwidth}
		\includegraphics[width=\textwidth]{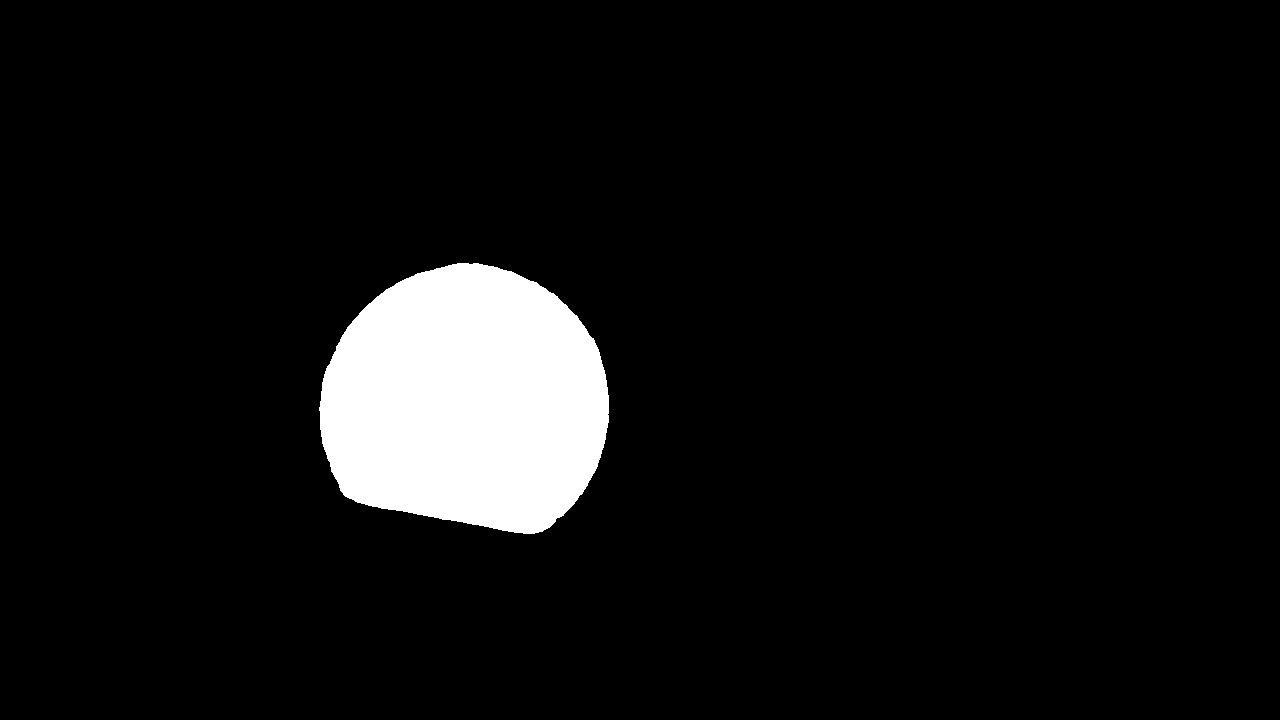}
	\end{subfigure}

	\vspace*{1.3mm}
	\begin{subfigure}{0.11\textwidth}
		\includegraphics[width=\textwidth]{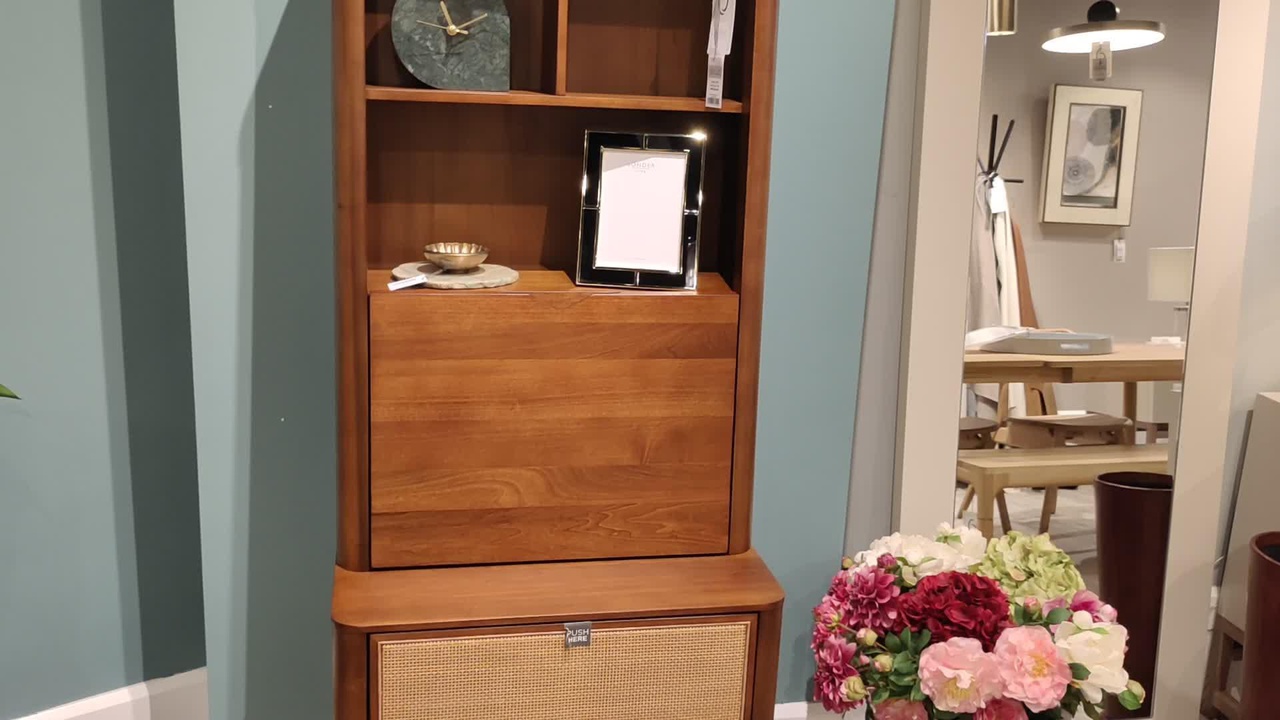}
	\end{subfigure}
	\begin{subfigure}{0.11\textwidth}
		\includegraphics[width=\textwidth]{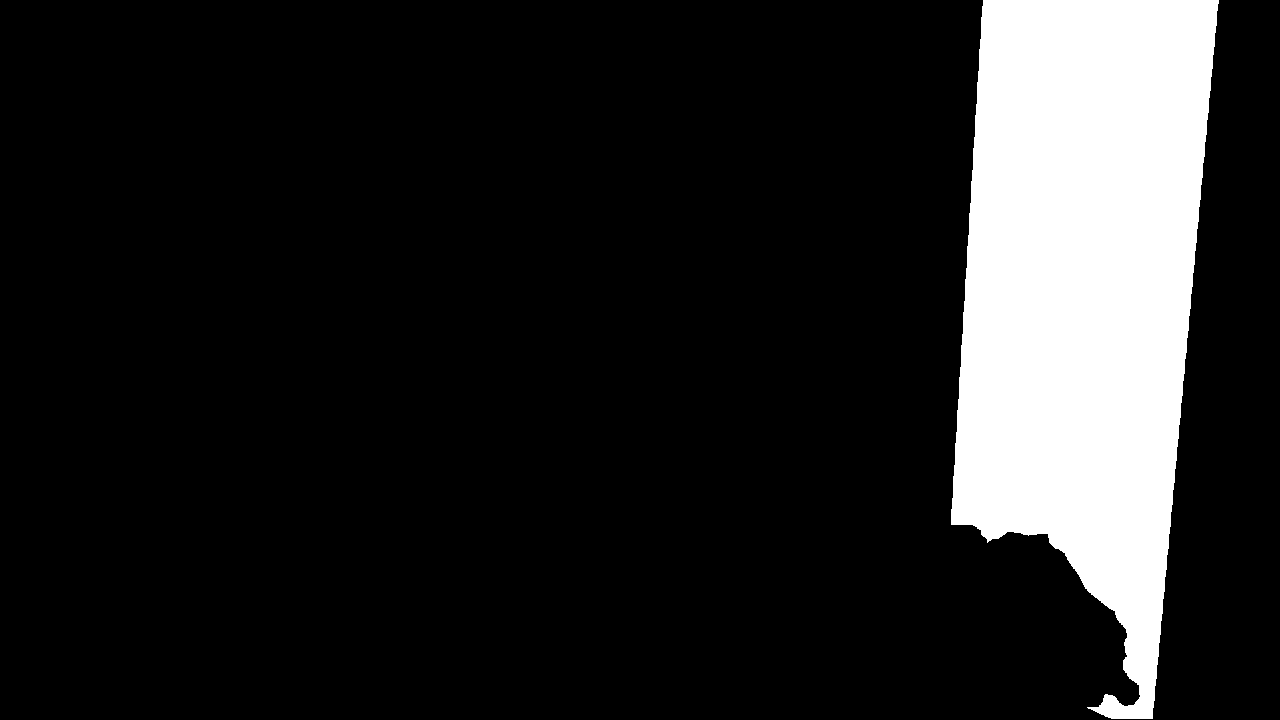}
	\end{subfigure}
	\begin{subfigure}{0.11\textwidth}
		\includegraphics[width=\textwidth]{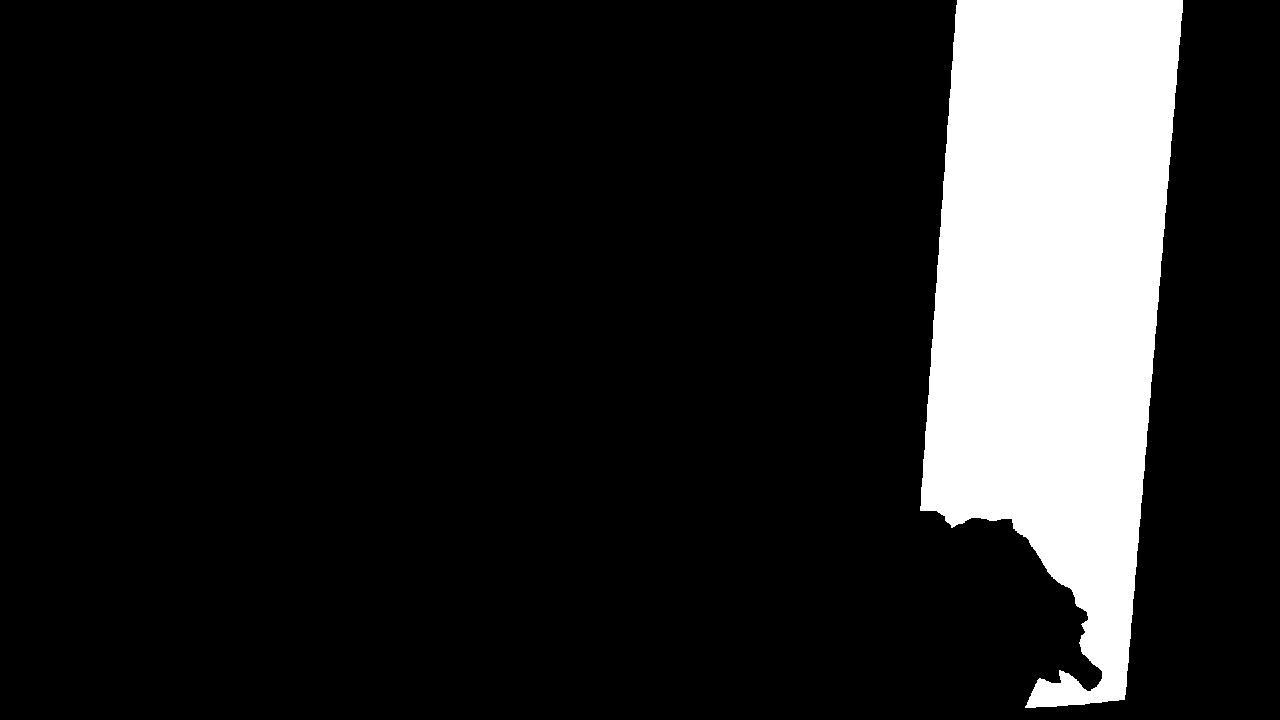}
	\end{subfigure}
	\begin{subfigure}{0.11\textwidth}
		\includegraphics[width=\textwidth]{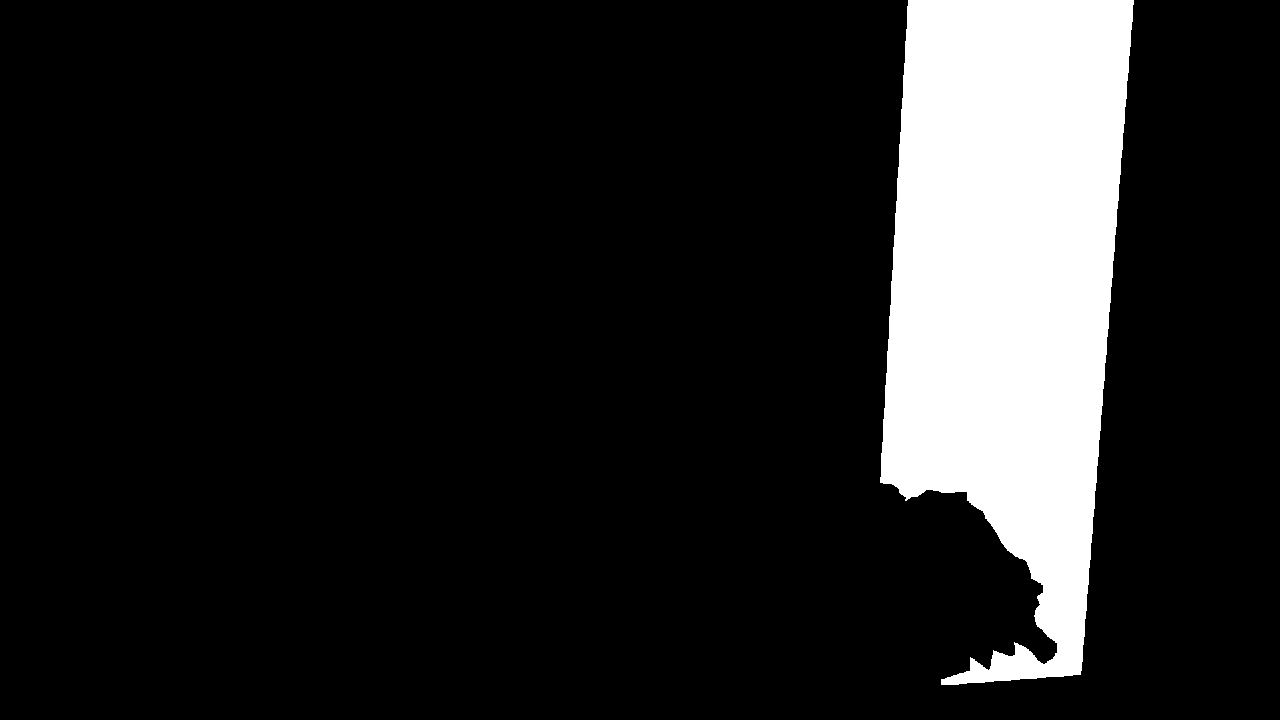}
	\end{subfigure}
	\begin{subfigure}{0.11\textwidth}
		\includegraphics[width=\textwidth]{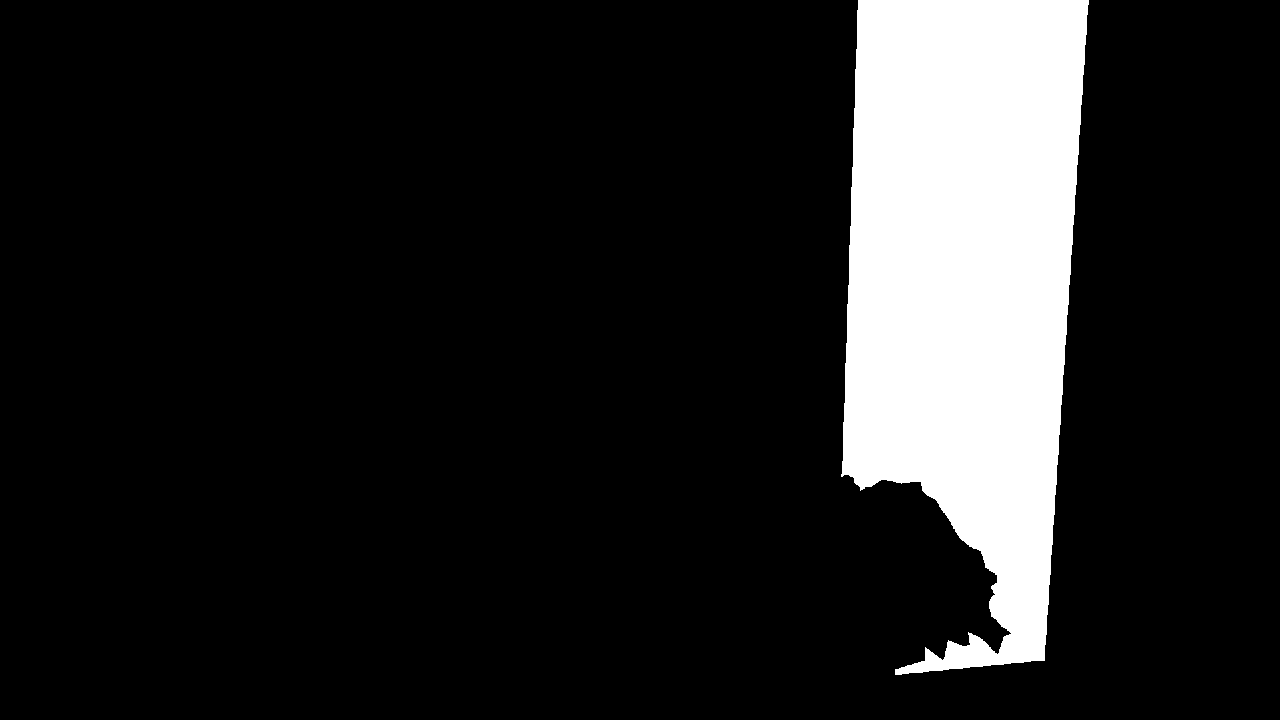}
	\end{subfigure}
	\begin{subfigure}{0.11\textwidth}
		\includegraphics[width=\textwidth]{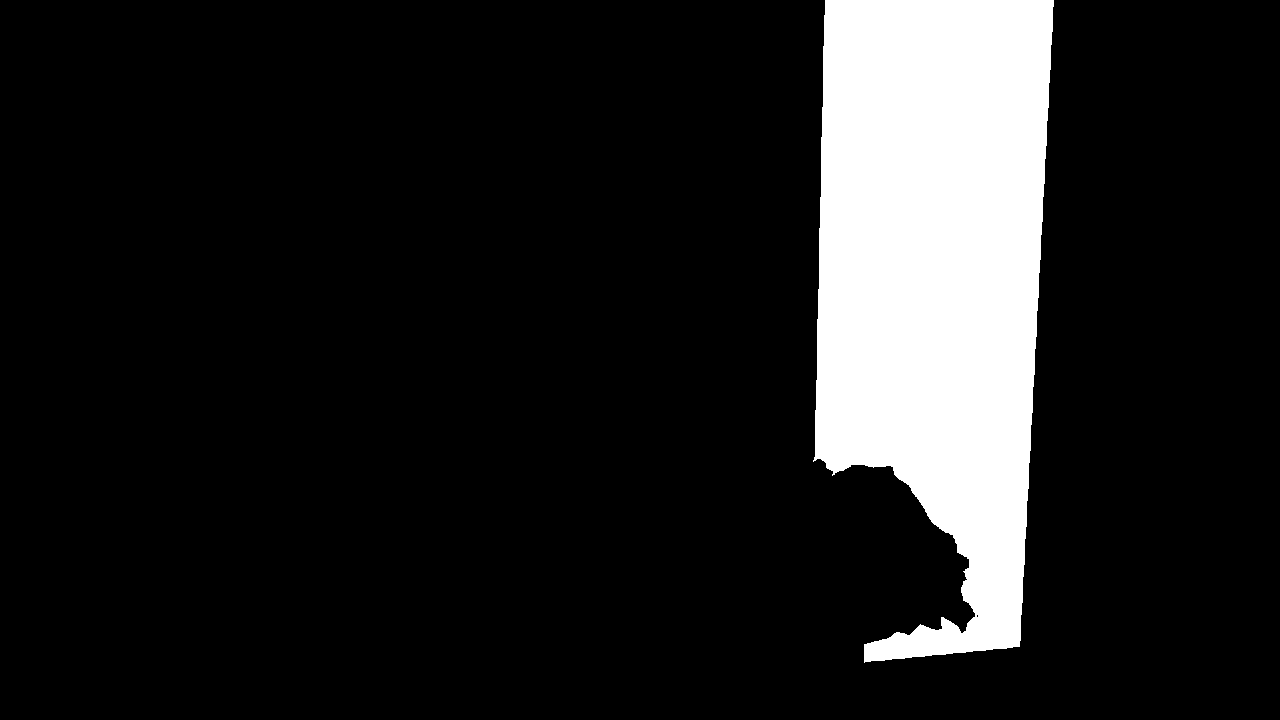}
	\end{subfigure}
	
	\vspace*{1.3mm}
	\begin{subfigure}{0.11\textwidth}
		\includegraphics[width=\textwidth]{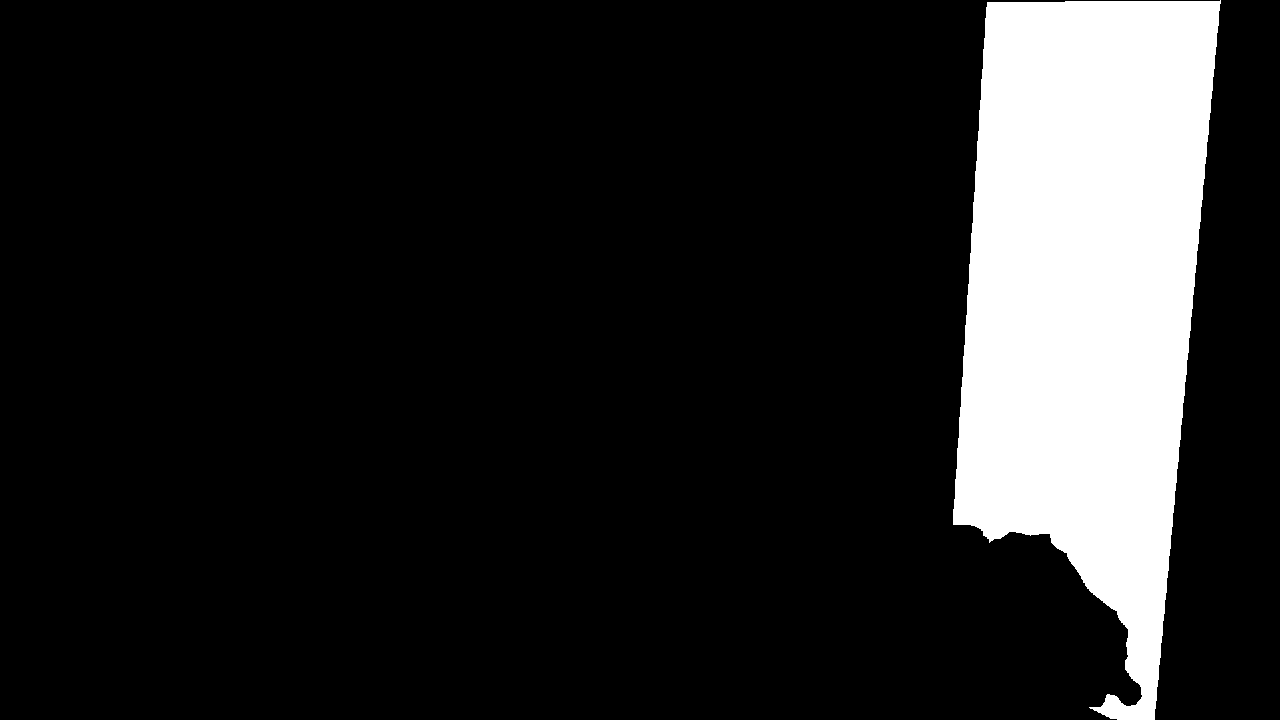}
	\end{subfigure}
	\begin{subfigure}{0.11\textwidth}
		\includegraphics[width=\textwidth]{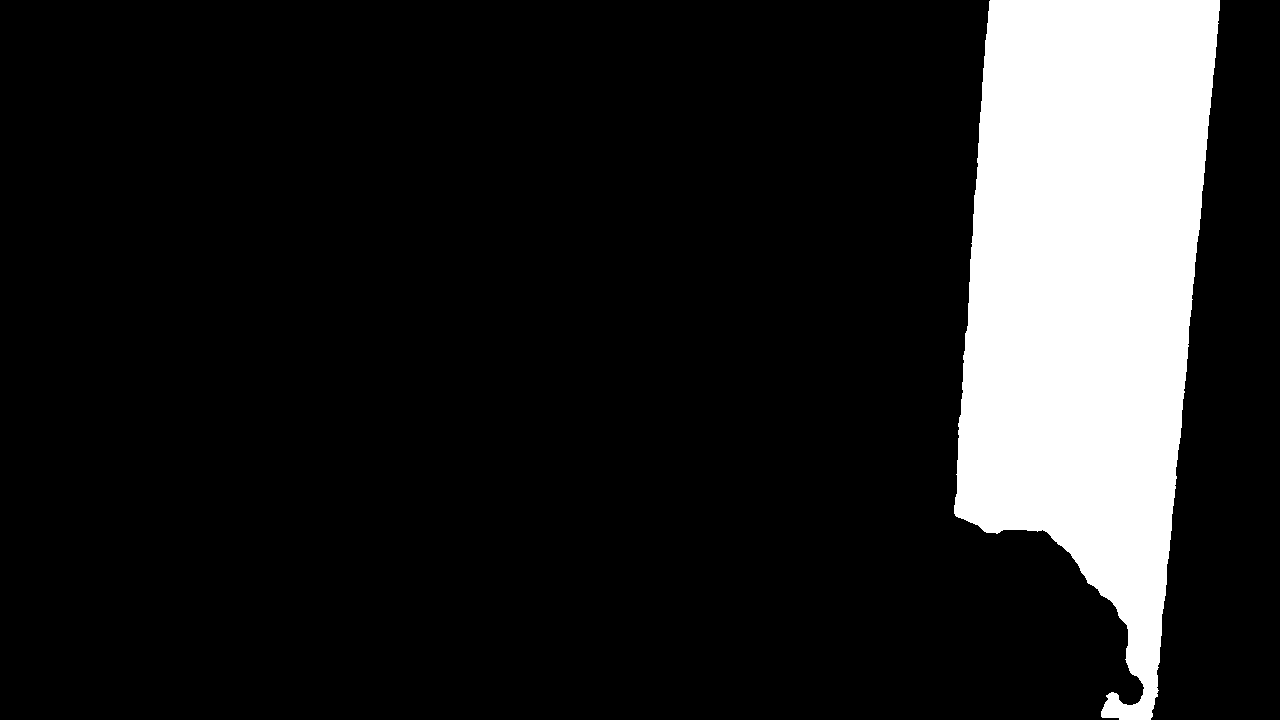}
	\end{subfigure}
	\begin{subfigure}{0.11\textwidth}
		\includegraphics[width=\textwidth]{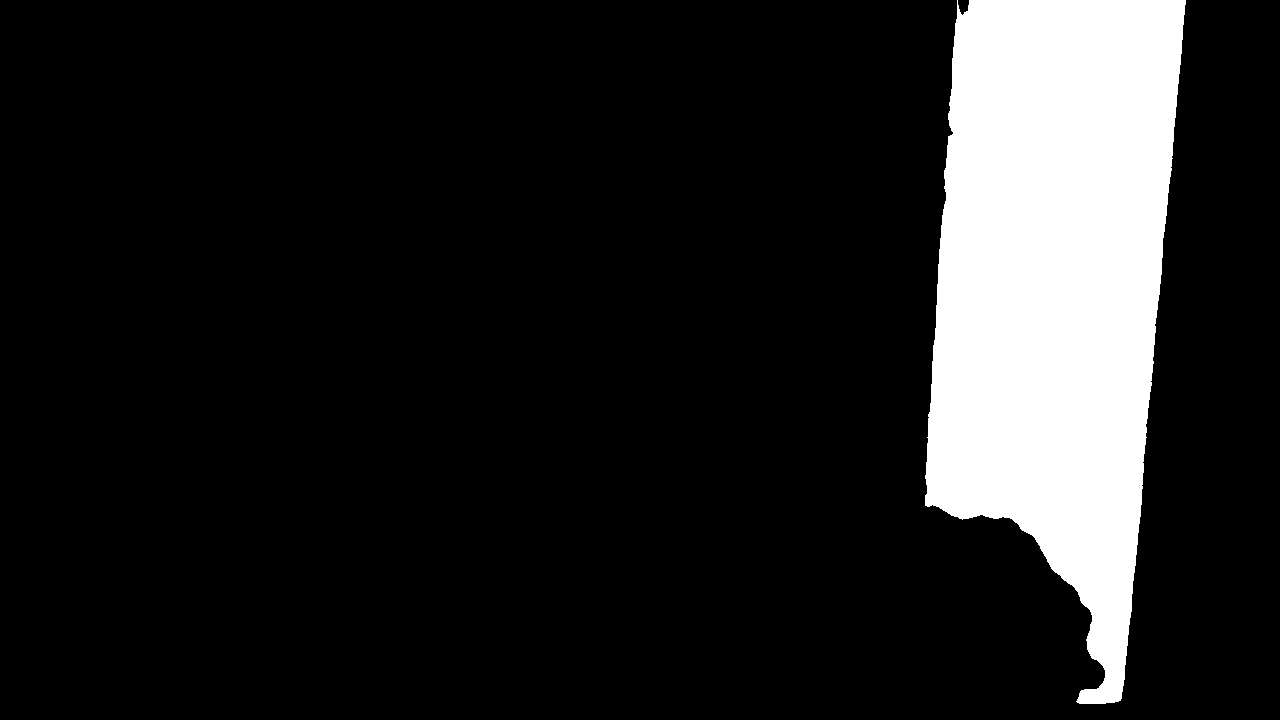}
	\end{subfigure}
	\begin{subfigure}{0.11\textwidth}
		\includegraphics[width=\textwidth]{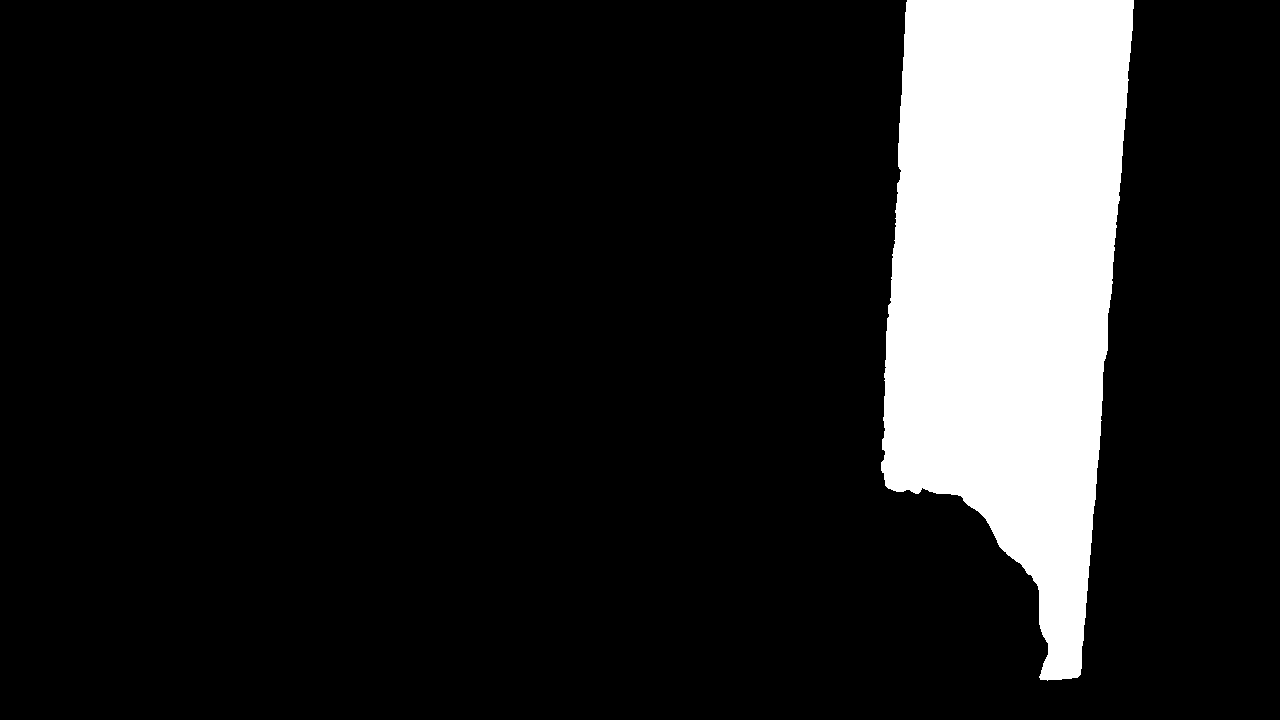}
	\end{subfigure}
	\begin{subfigure}{0.11\textwidth}
		\includegraphics[width=\textwidth]{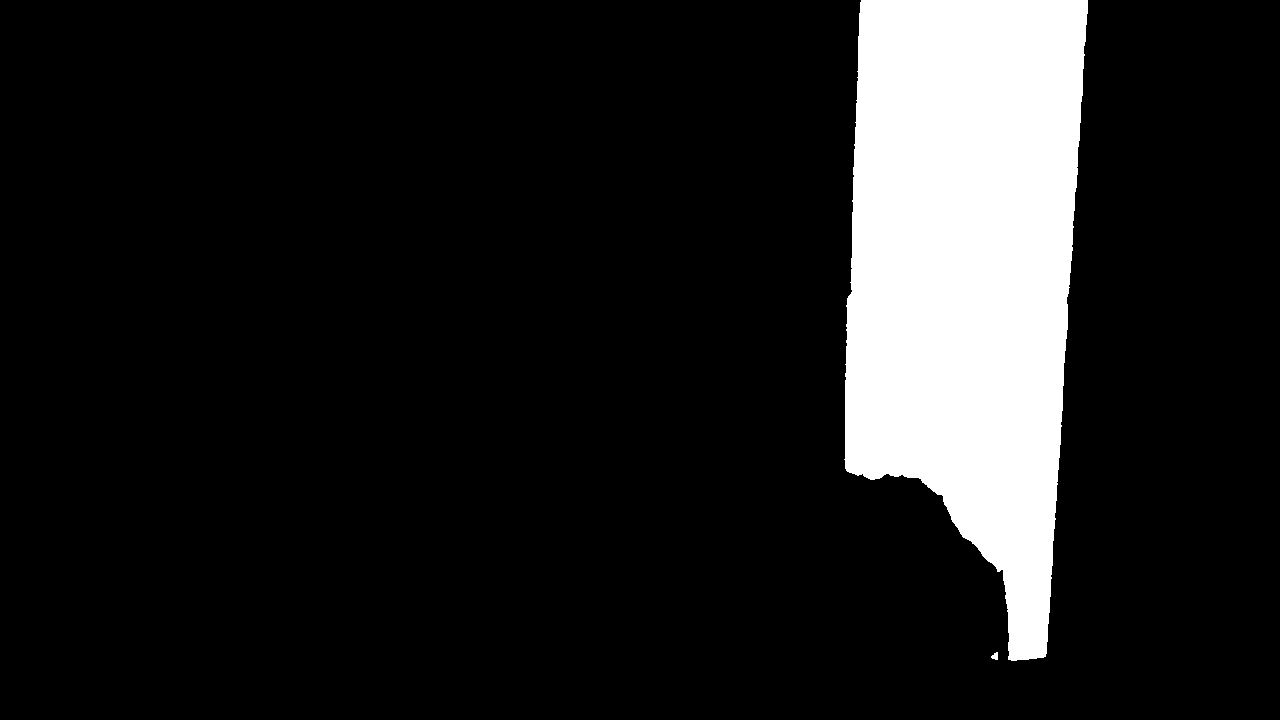}
	\end{subfigure}
	\begin{subfigure}{0.11\textwidth}
		\includegraphics[width=\textwidth]{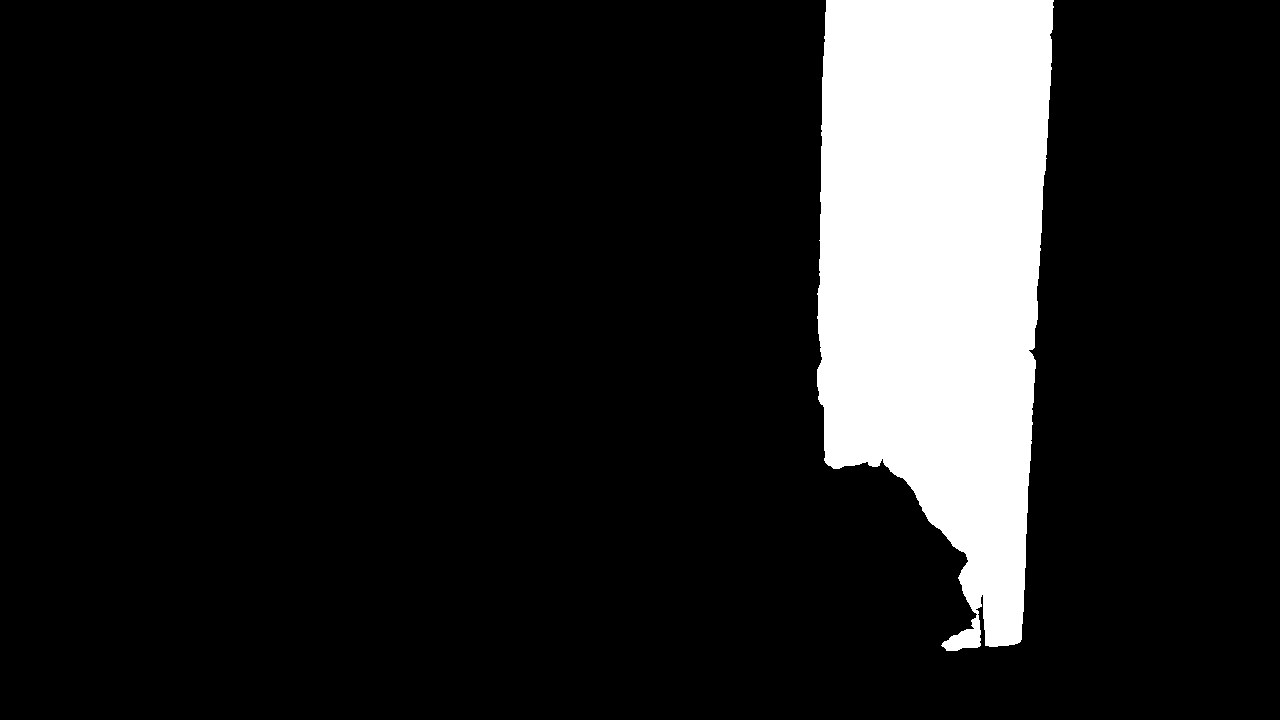}
	\end{subfigure}

	\vspace*{1.3mm}
	\begin{subfigure}{0.11\textwidth}
		\includegraphics[width=\textwidth]{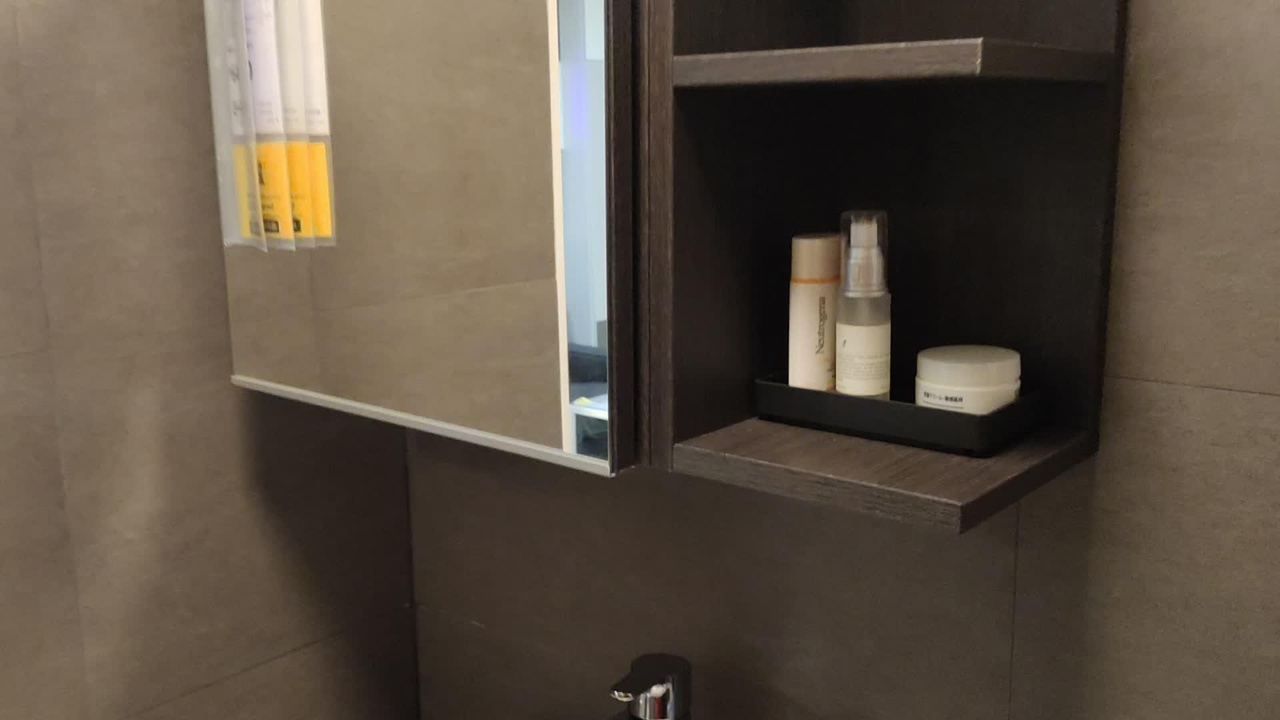}
	\end{subfigure}
	\begin{subfigure}{0.11\textwidth}
		\includegraphics[width=\textwidth]{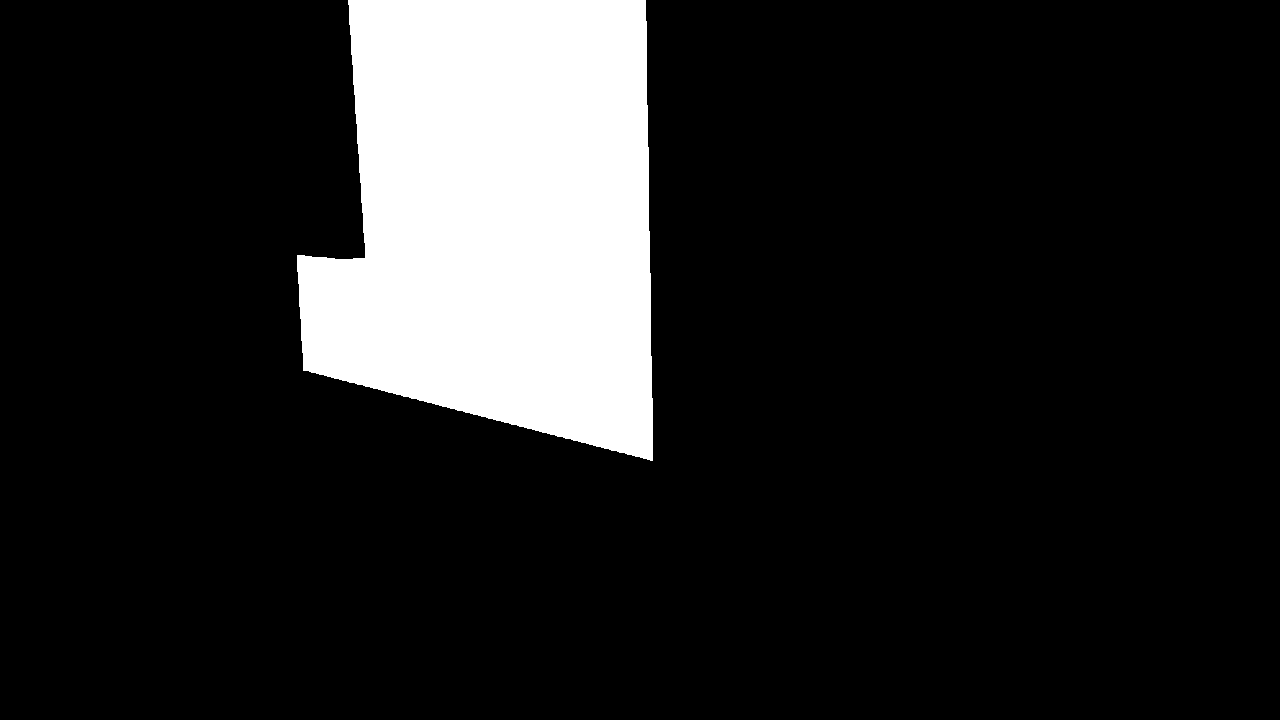}
	\end{subfigure}
	\begin{subfigure}{0.11\textwidth}
		\includegraphics[width=\textwidth]{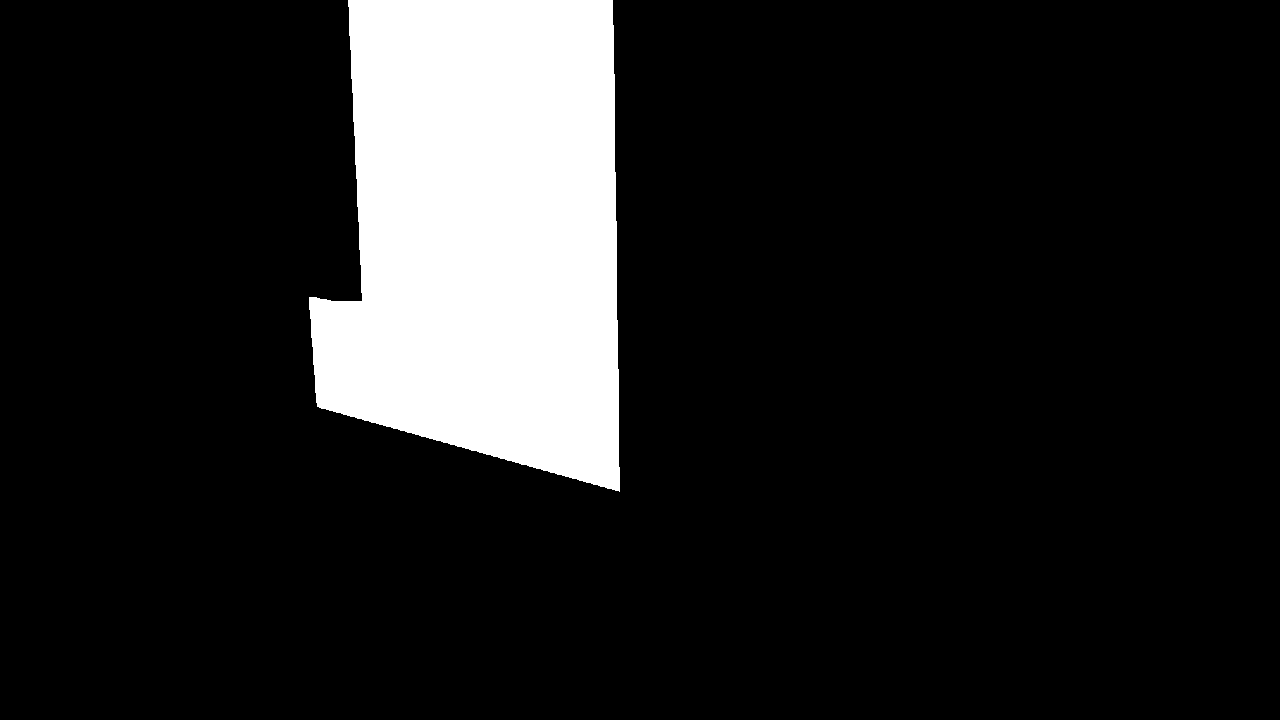}
	\end{subfigure}
	\begin{subfigure}{0.11\textwidth}
		\includegraphics[width=\textwidth]{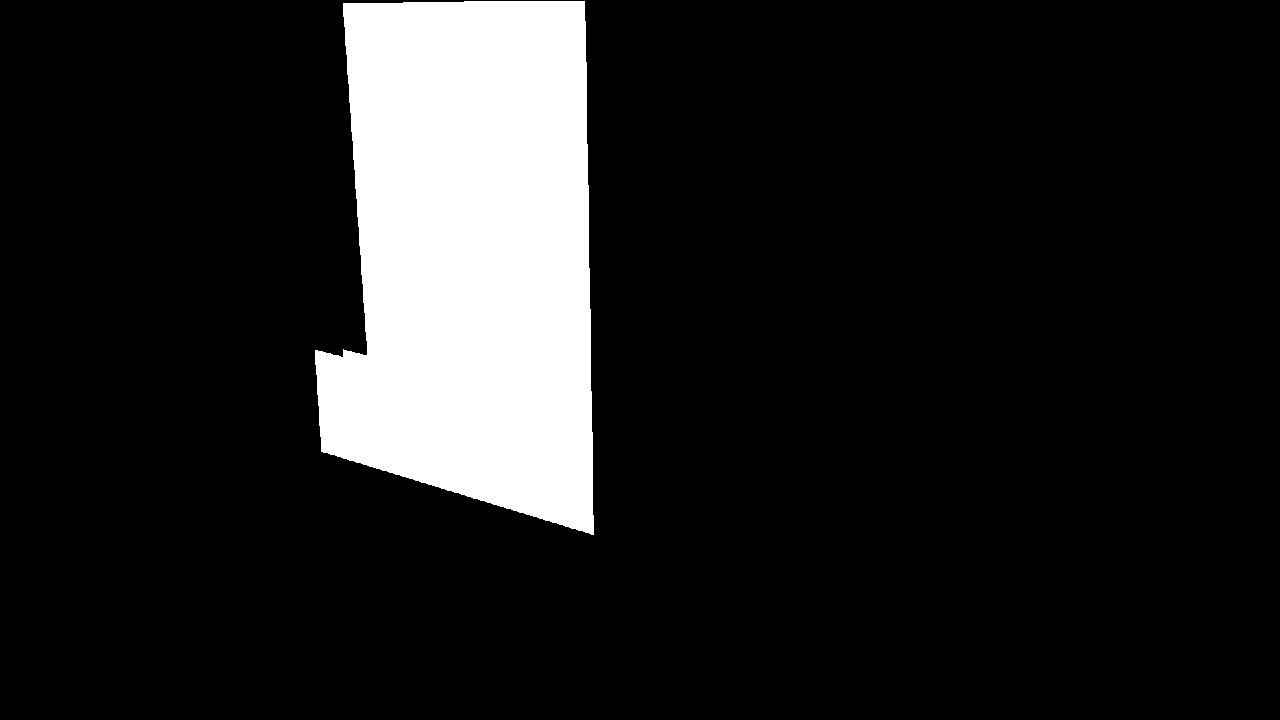}
	\end{subfigure}
	\begin{subfigure}{0.11\textwidth}
		\includegraphics[width=\textwidth]{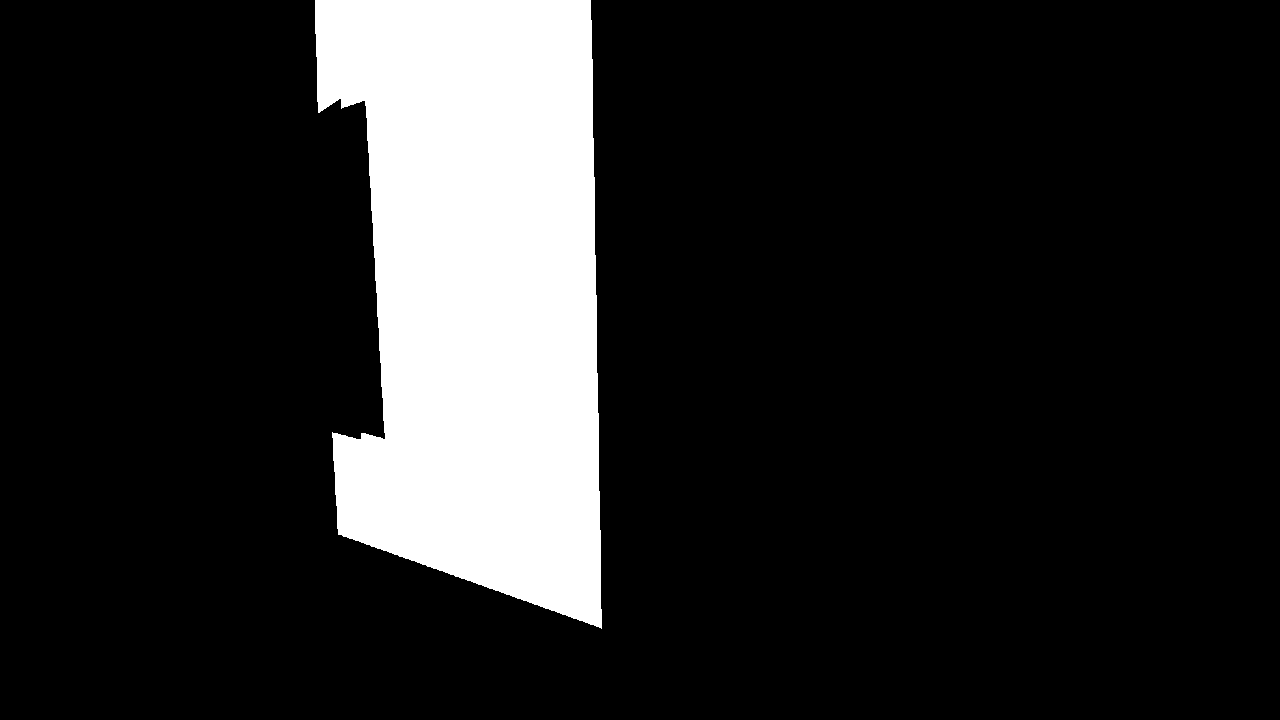}
	\end{subfigure}
	\begin{subfigure}{0.11\textwidth}
		\includegraphics[width=\textwidth]{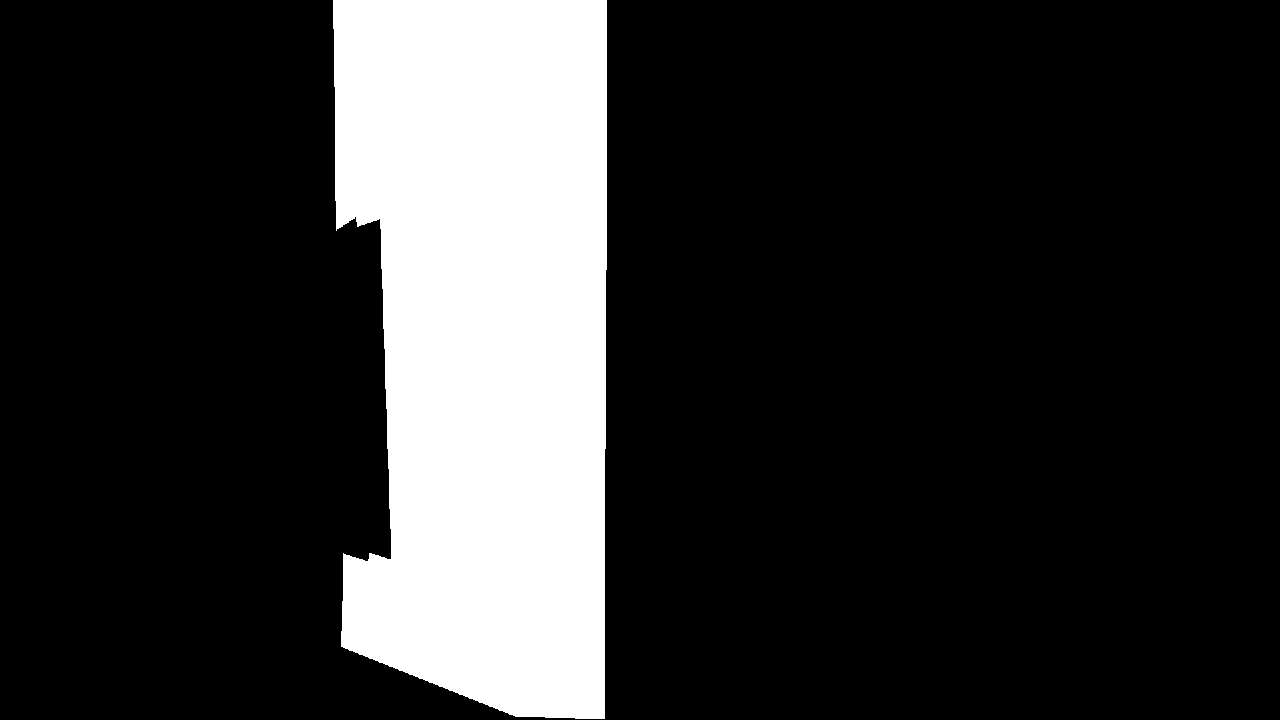}
	\end{subfigure}
	
	\vspace*{1.3mm}
	\begin{subfigure}{0.11\textwidth}
		\includegraphics[width=\textwidth]{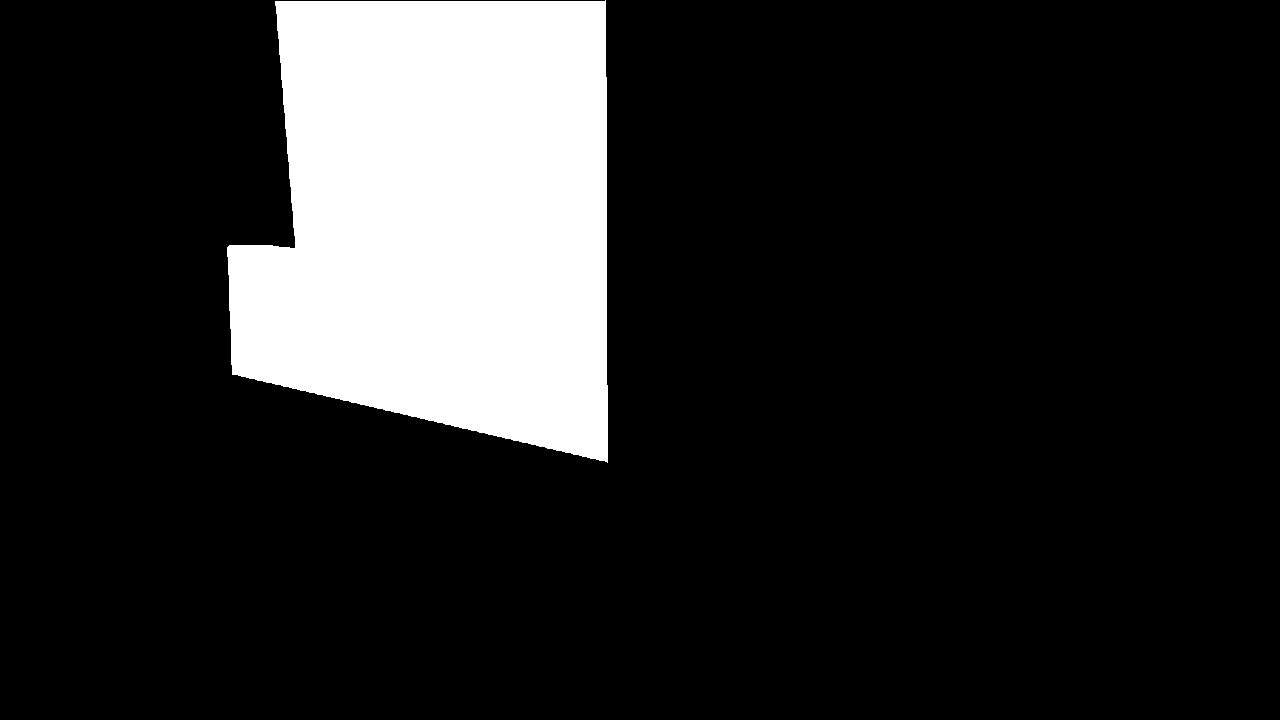}
	\end{subfigure}
	\begin{subfigure}{0.11\textwidth}
		\includegraphics[width=\textwidth]{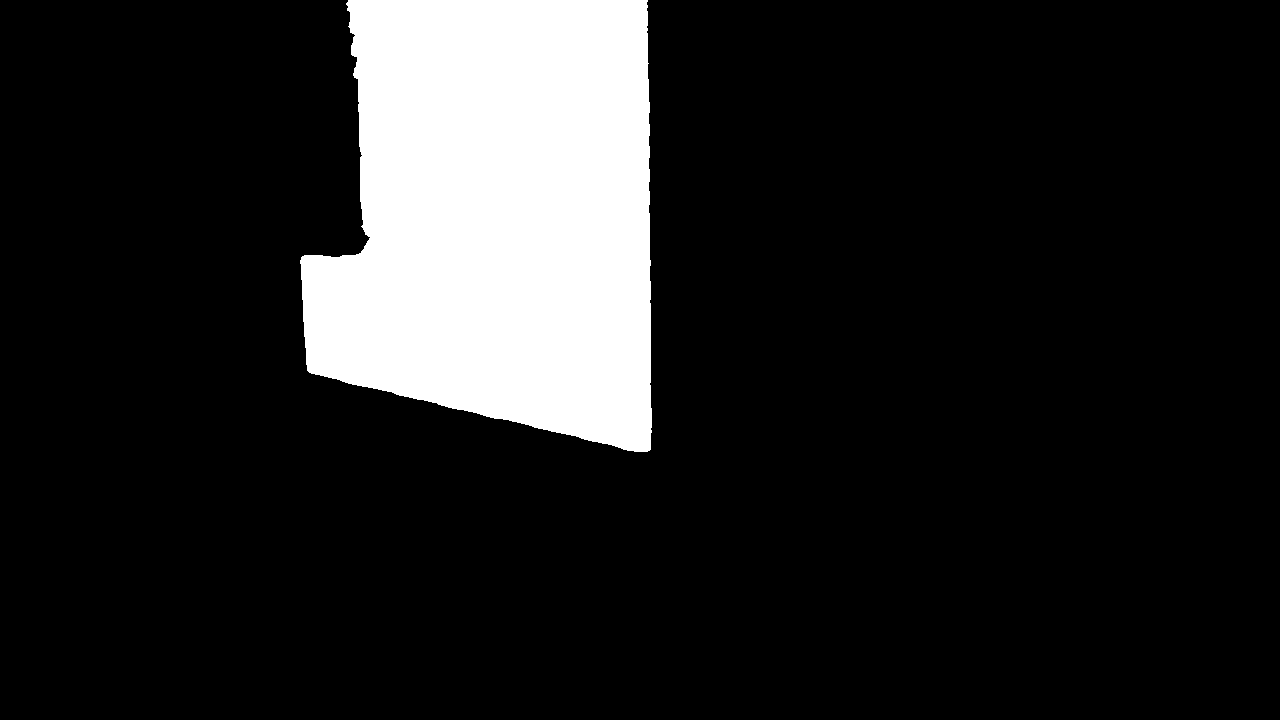}
	\end{subfigure}
	\begin{subfigure}{0.11\textwidth}
		\includegraphics[width=\textwidth]{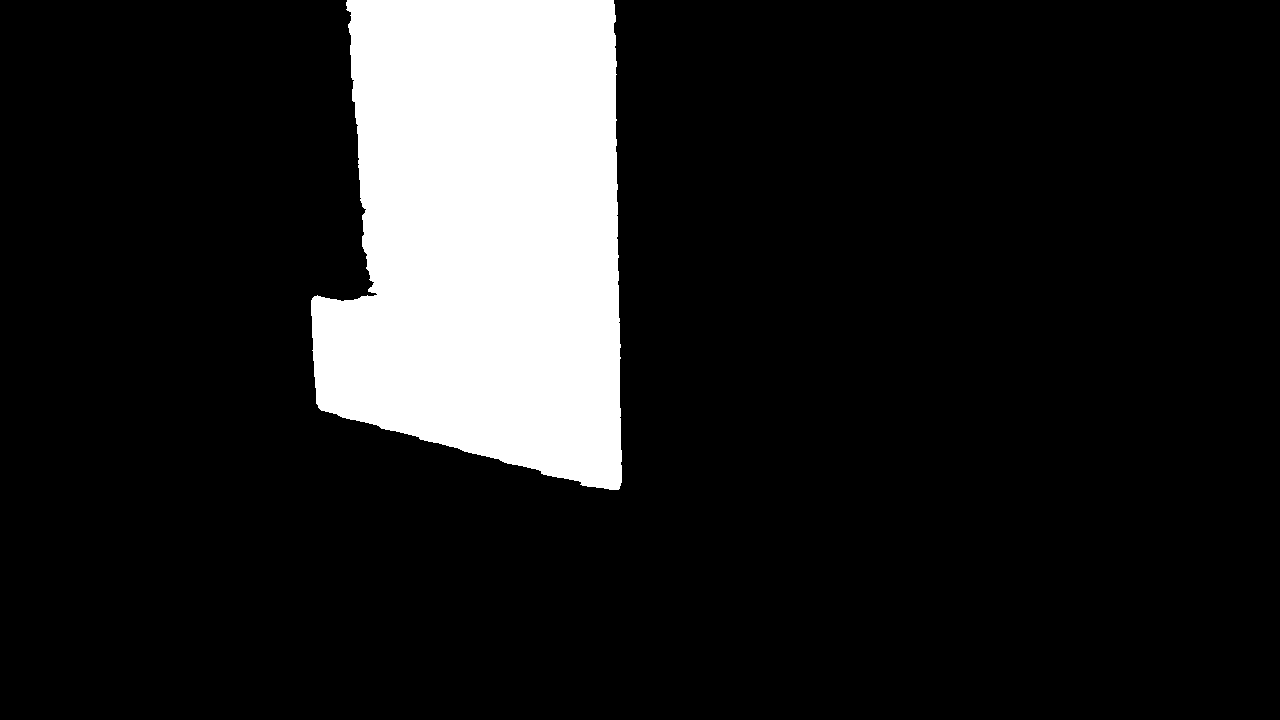}
	\end{subfigure}
	\begin{subfigure}{0.11\textwidth}
		\includegraphics[width=\textwidth]{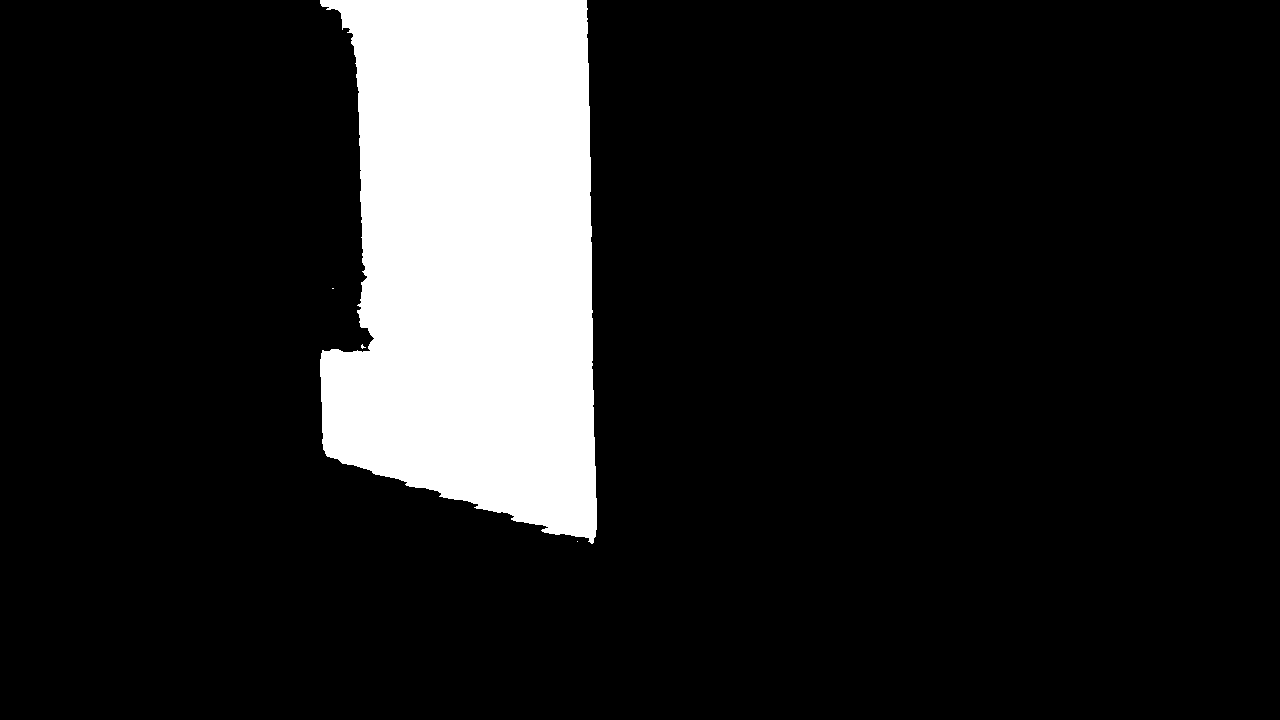}
	\end{subfigure}
	\begin{subfigure}{0.11\textwidth}
		\includegraphics[width=\textwidth]{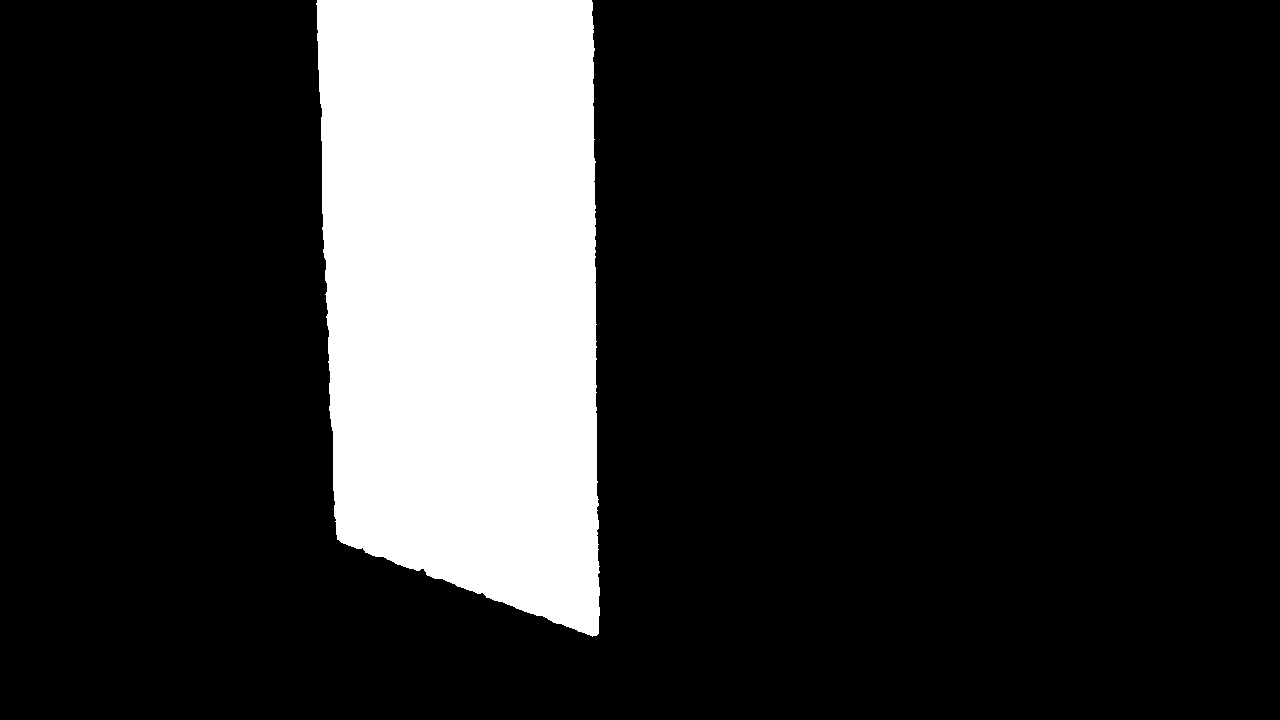}
	\end{subfigure}
	\begin{subfigure}{0.11\textwidth}
		\includegraphics[width=\textwidth]{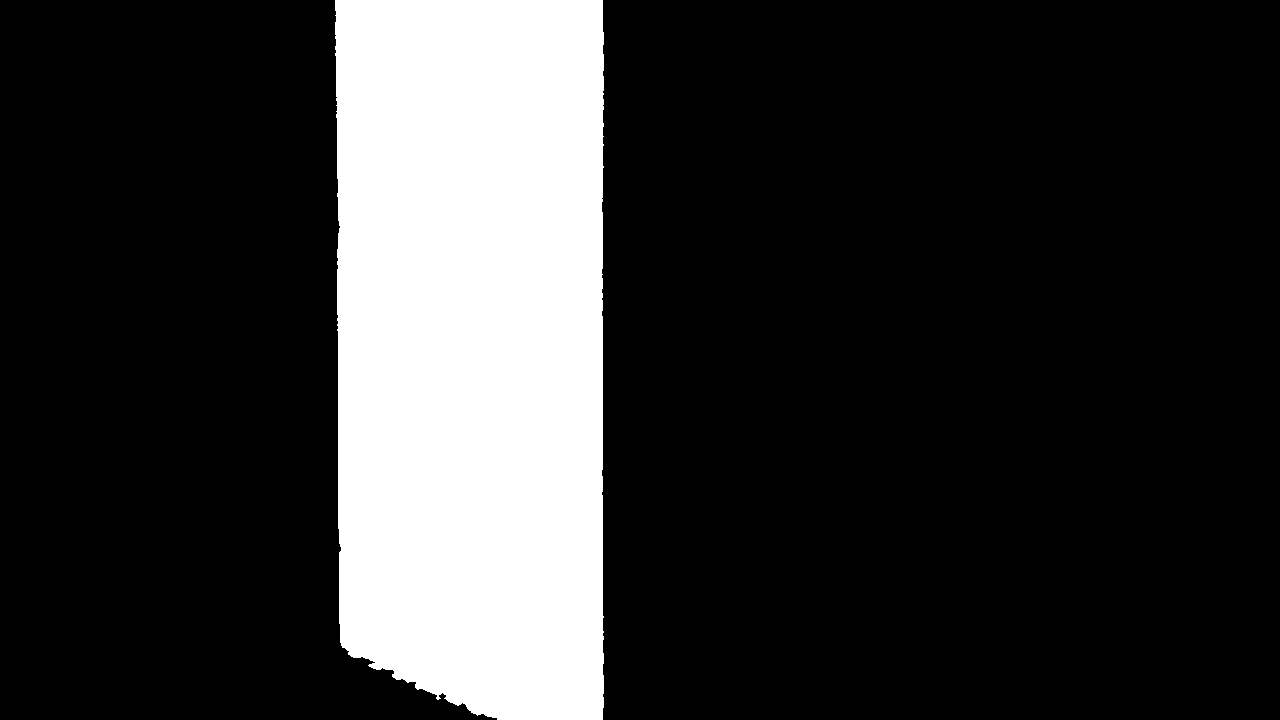}
	\end{subfigure}

	\vspace*{1.3mm}
	\begin{subfigure}{0.11\textwidth}
		\includegraphics[width=\textwidth]{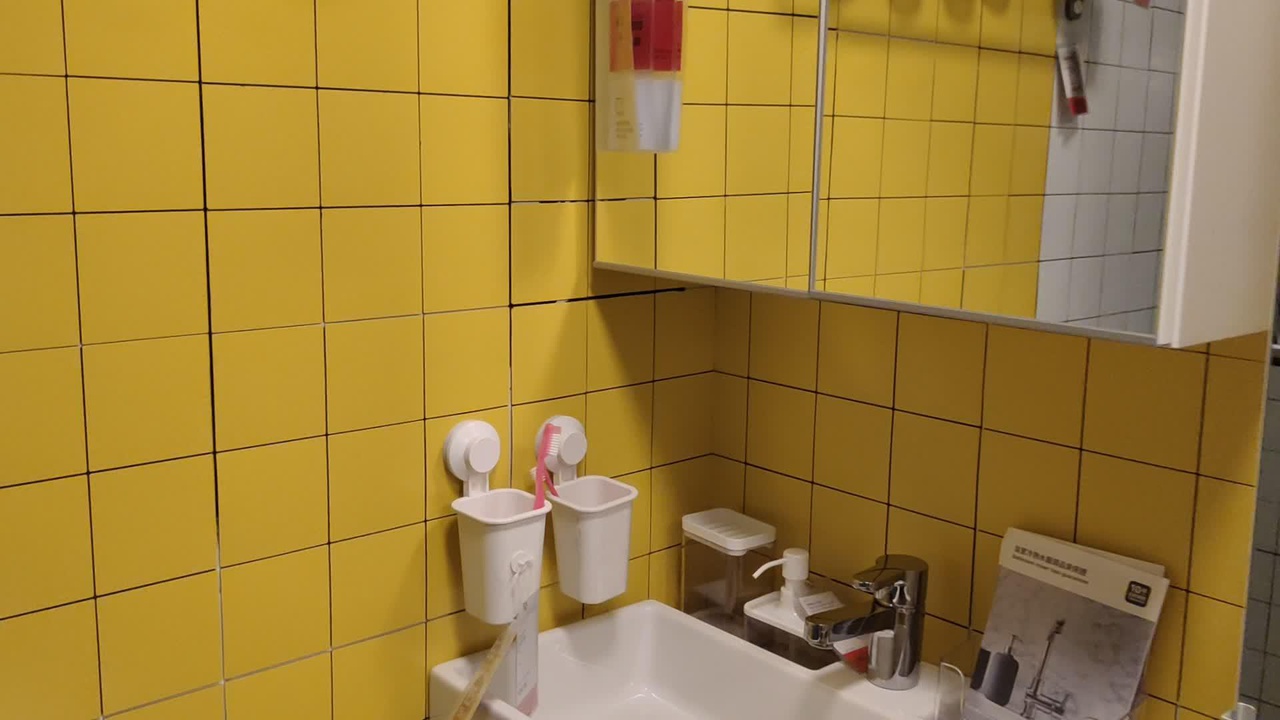}
	\end{subfigure}
	\begin{subfigure}{0.11\textwidth}
		\includegraphics[width=\textwidth]{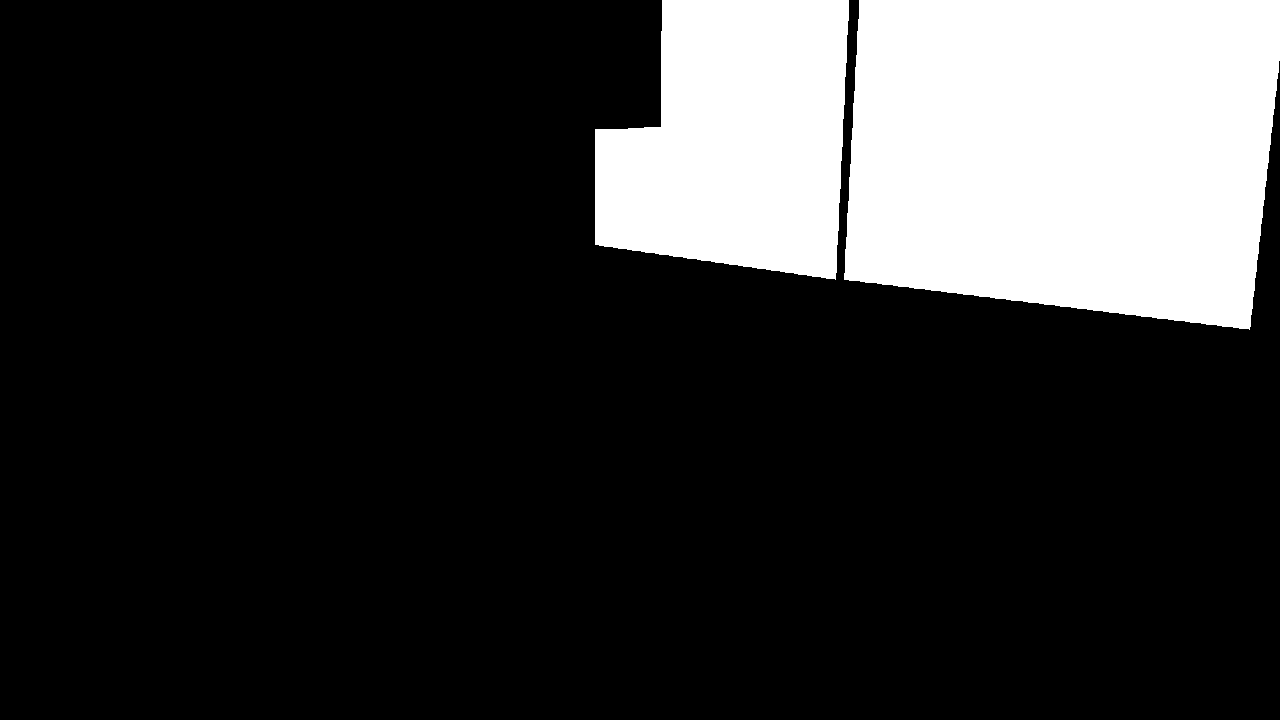}
	\end{subfigure}
	\begin{subfigure}{0.11\textwidth}
		\includegraphics[width=\textwidth]{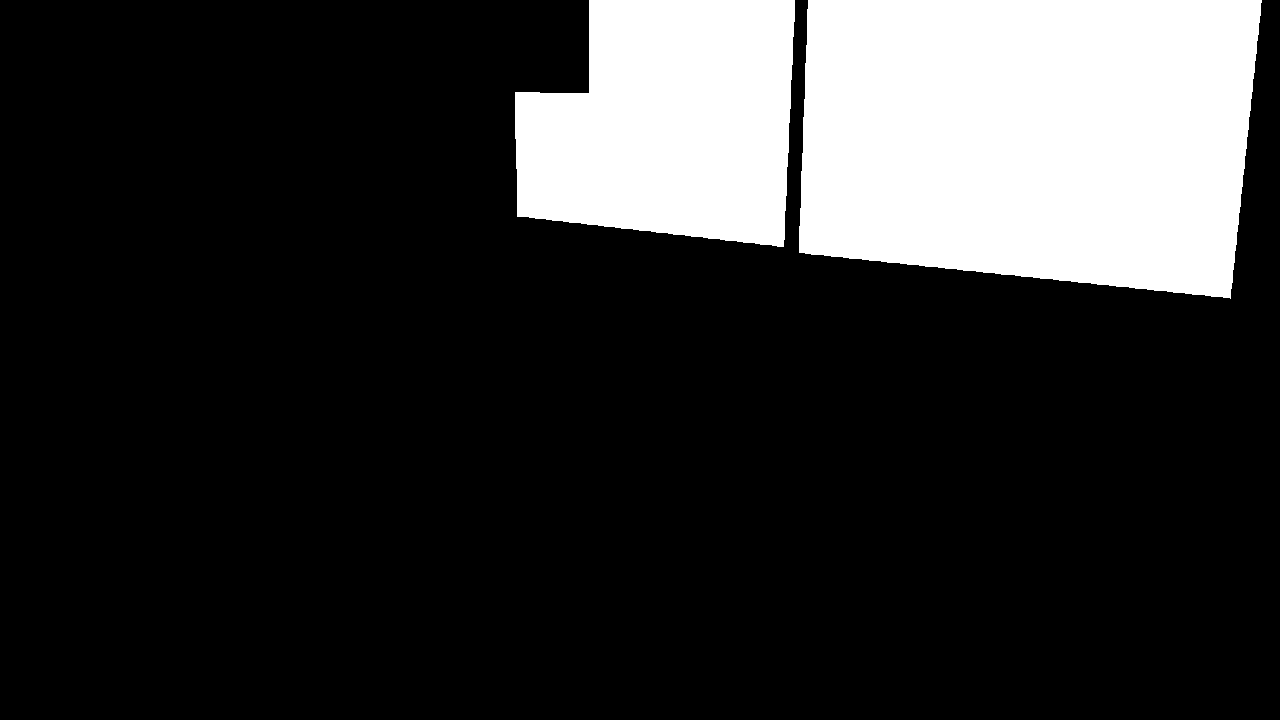}
	\end{subfigure}
	\begin{subfigure}{0.11\textwidth}
		\includegraphics[width=\textwidth]{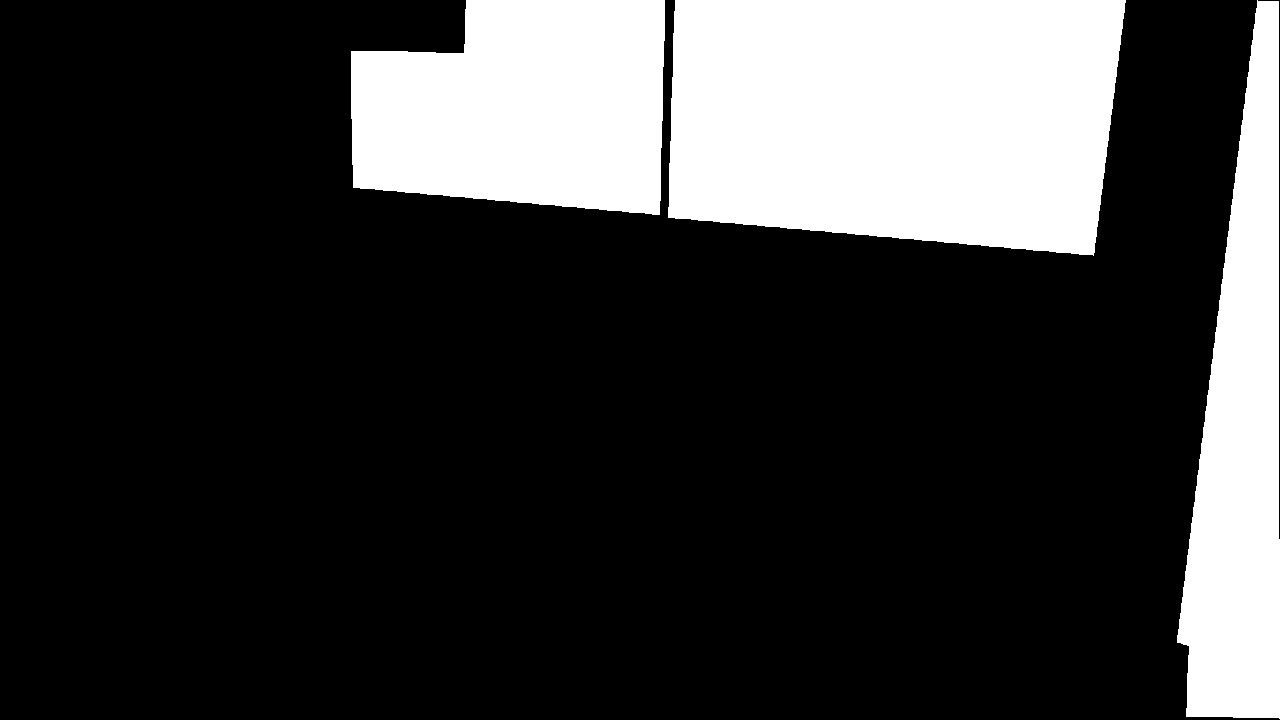}
	\end{subfigure}
	\begin{subfigure}{0.11\textwidth}
		\includegraphics[width=\textwidth]{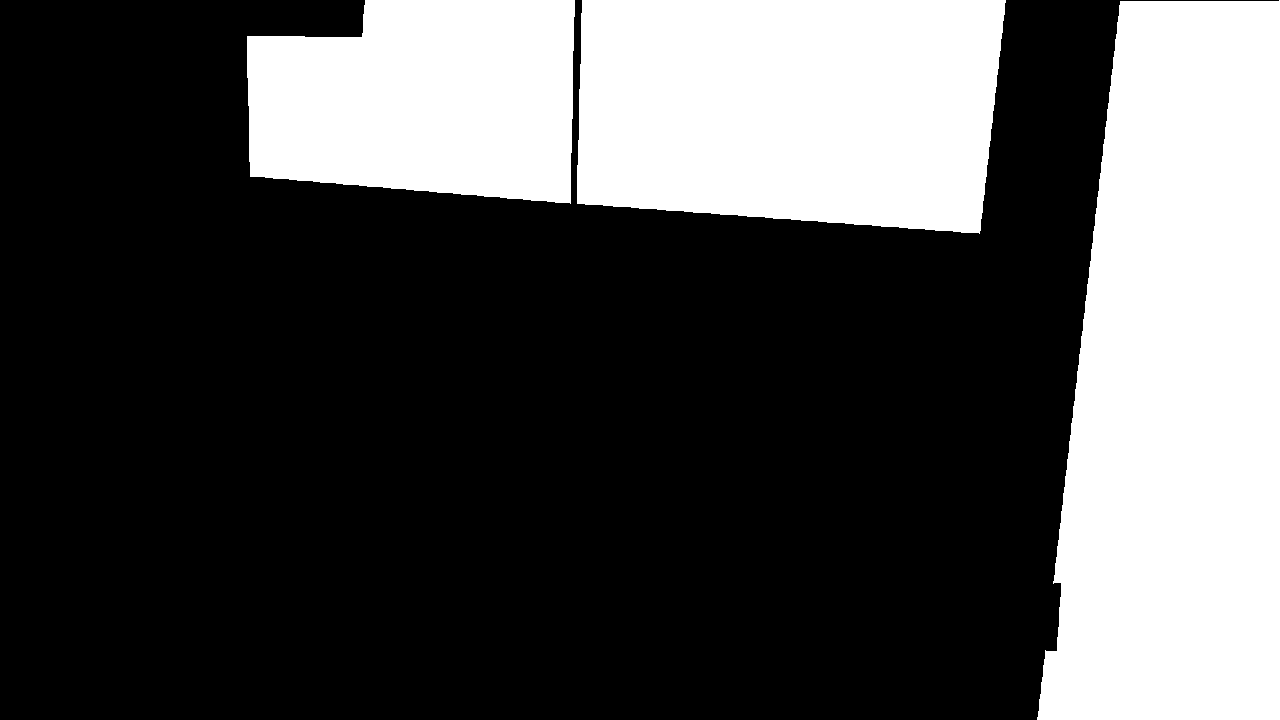}
	\end{subfigure}
	\begin{subfigure}{0.11\textwidth}
		\includegraphics[width=\textwidth]{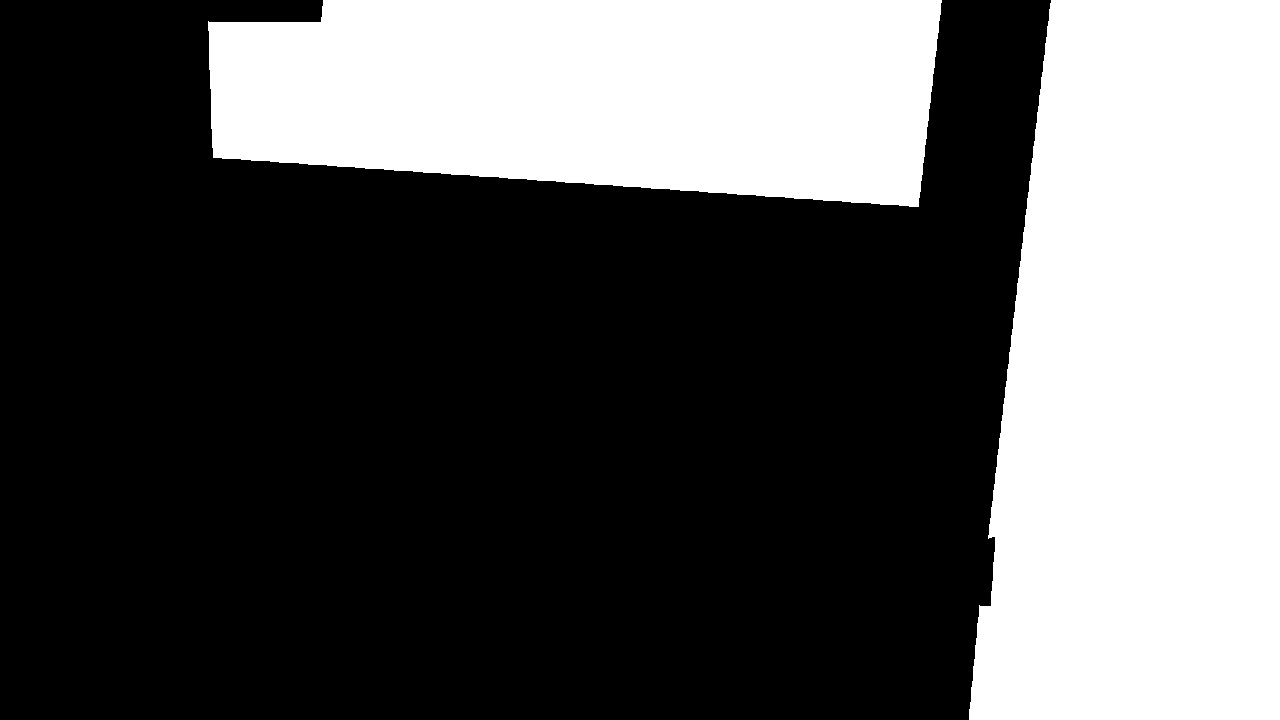}
	\end{subfigure}
	
	\vspace*{1.3mm}
	\begin{subfigure}{0.11\textwidth}
		\includegraphics[width=\textwidth]{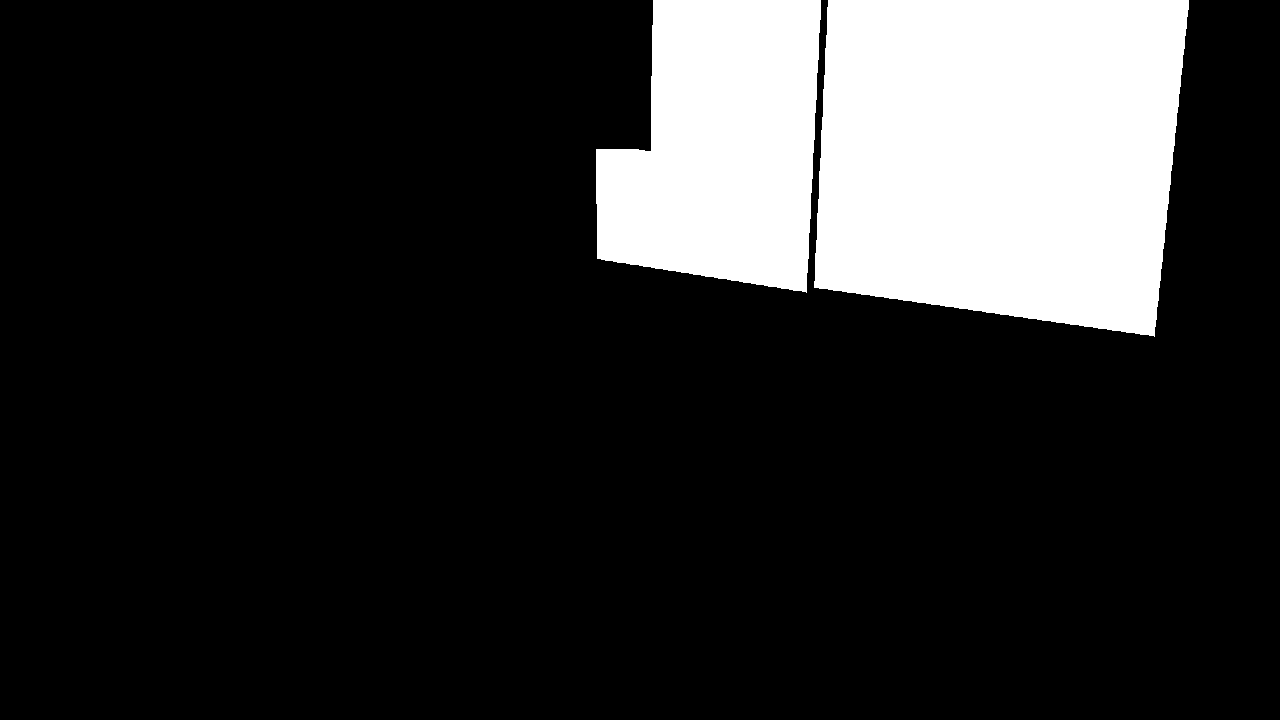}
	\end{subfigure}
	\begin{subfigure}{0.11\textwidth}
		\includegraphics[width=\textwidth]{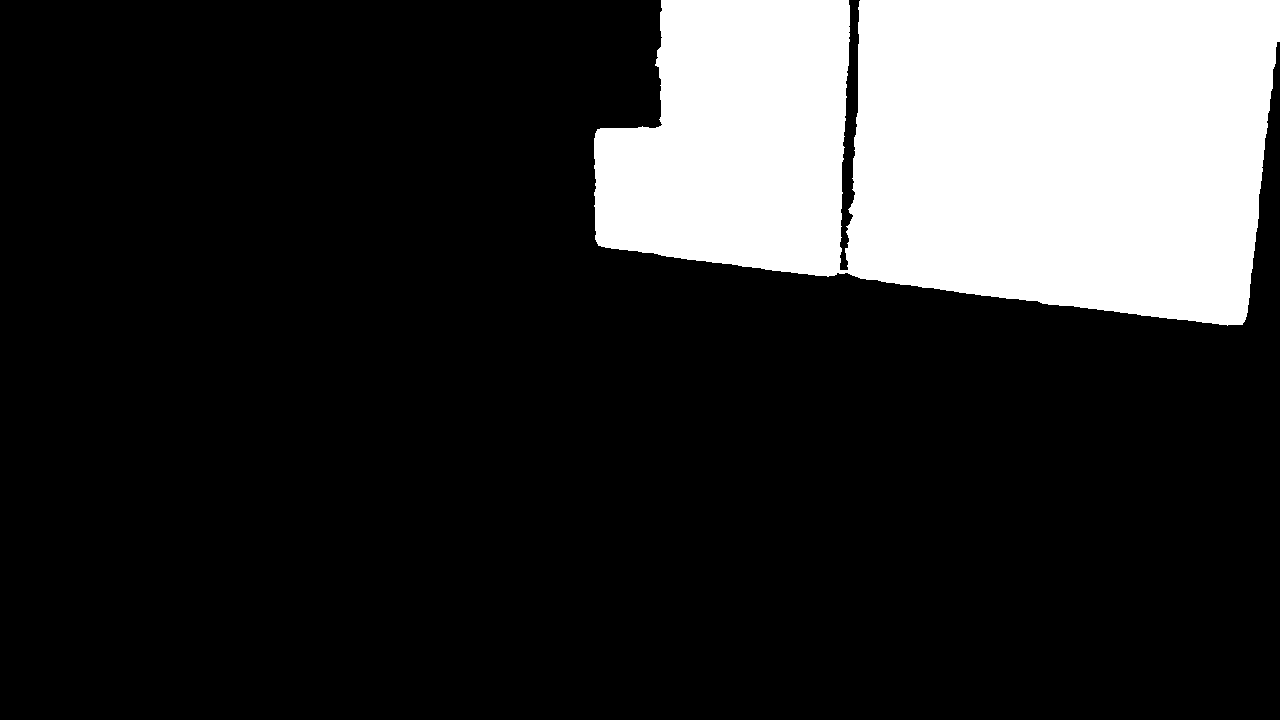}
	\end{subfigure}
	\begin{subfigure}{0.11\textwidth}
		\includegraphics[width=\textwidth]{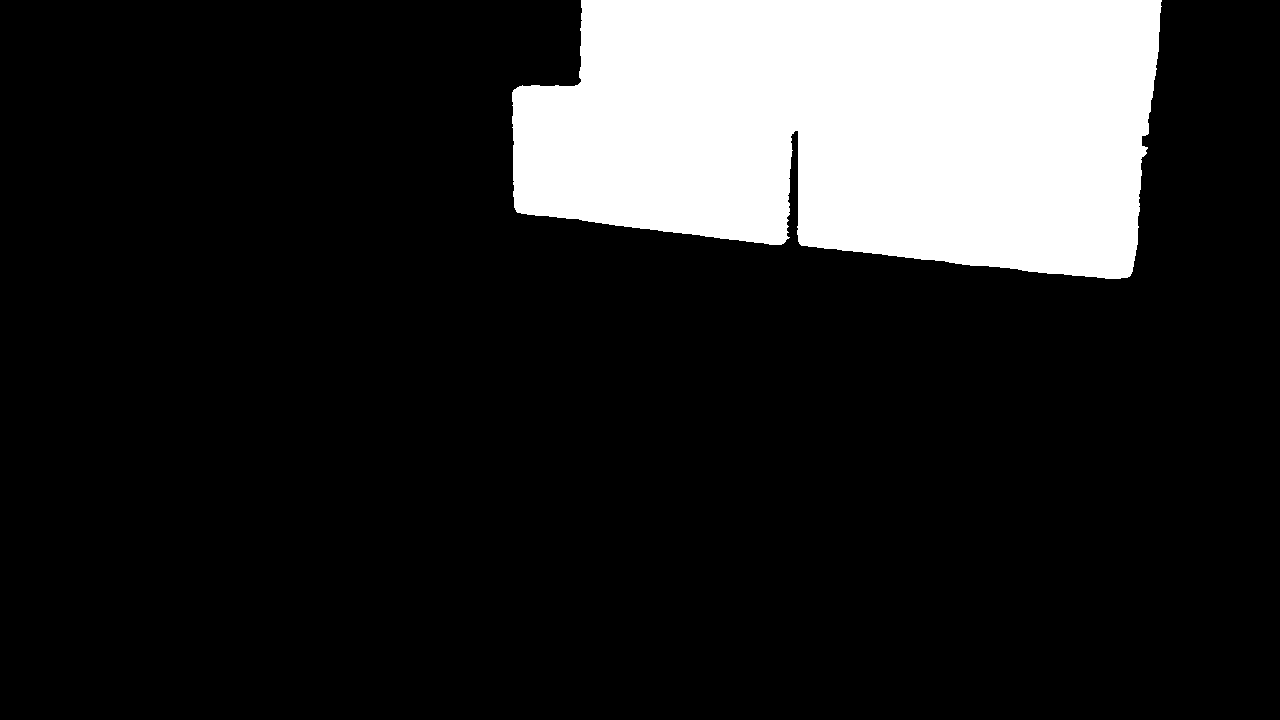}
	\end{subfigure}
	\begin{subfigure}{0.11\textwidth}
		\includegraphics[width=\textwidth]{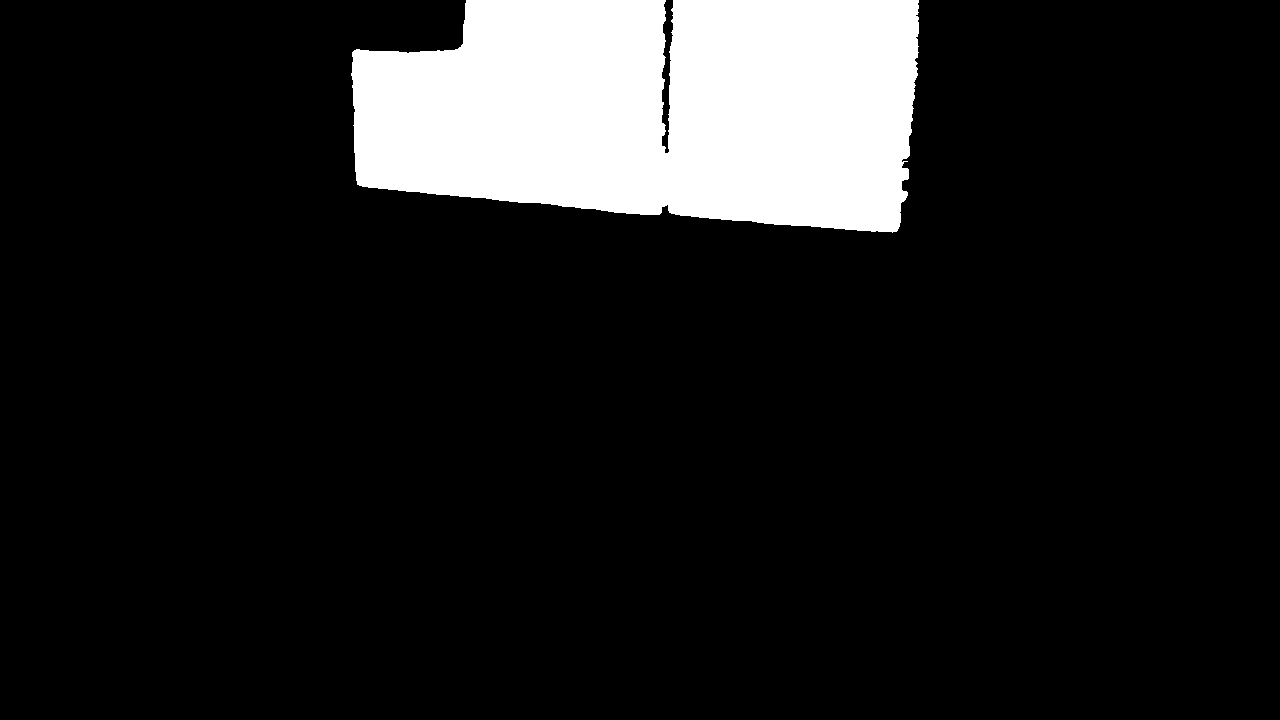}
	\end{subfigure}
	\begin{subfigure}{0.11\textwidth}
		\includegraphics[width=\textwidth]{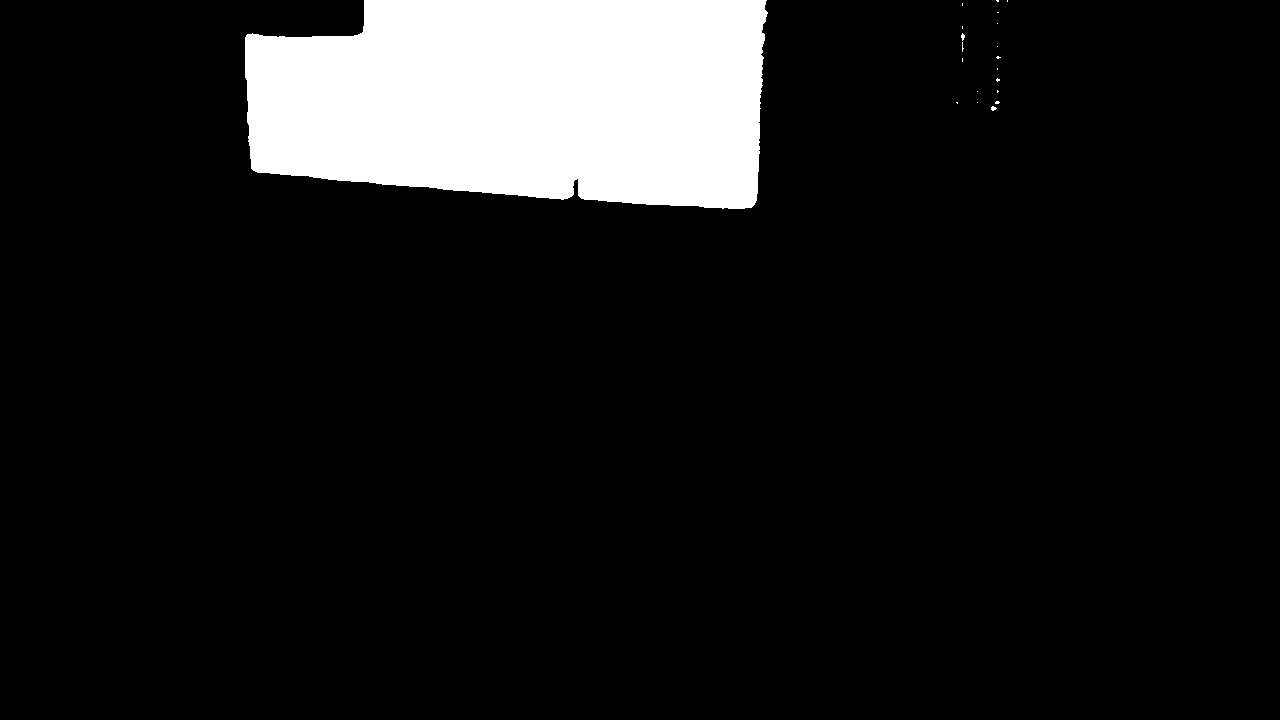}
	\end{subfigure}
	\begin{subfigure}{0.11\textwidth}
		\includegraphics[width=\textwidth]{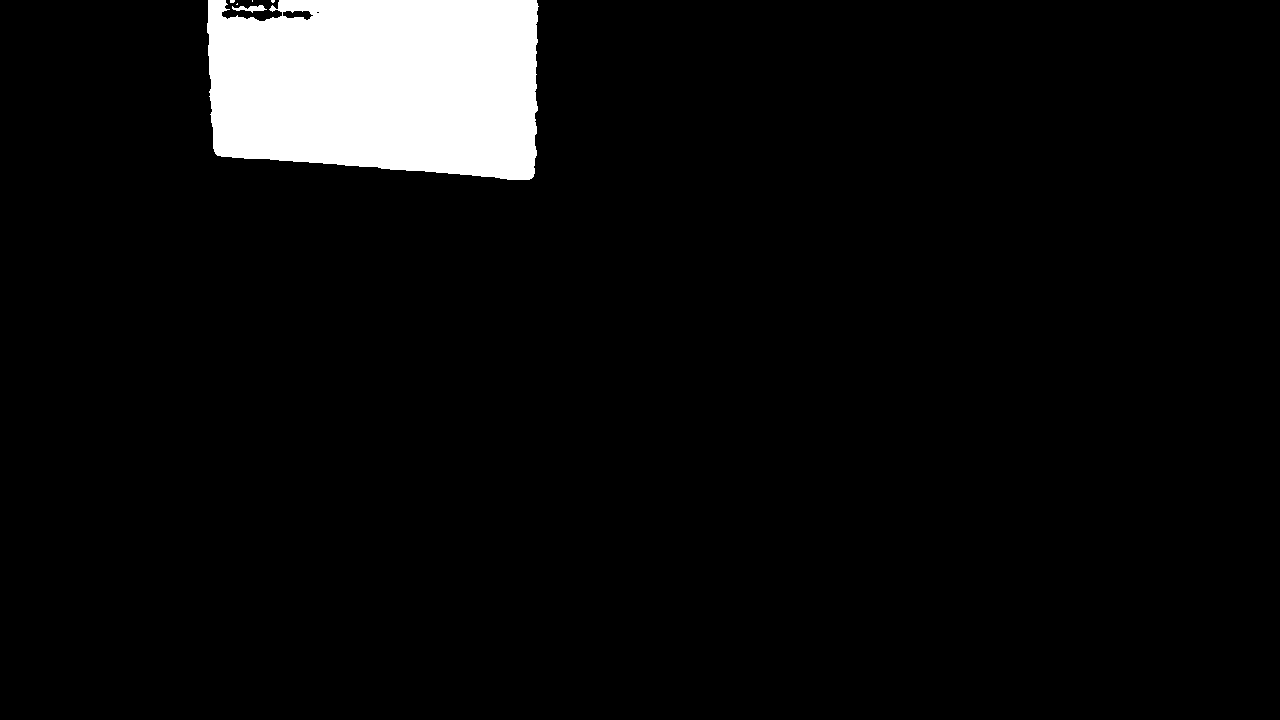}
	\end{subfigure}

	\vspace*{1.3mm}
	\begin{subfigure}{0.11\textwidth}
		\includegraphics[width=\textwidth]{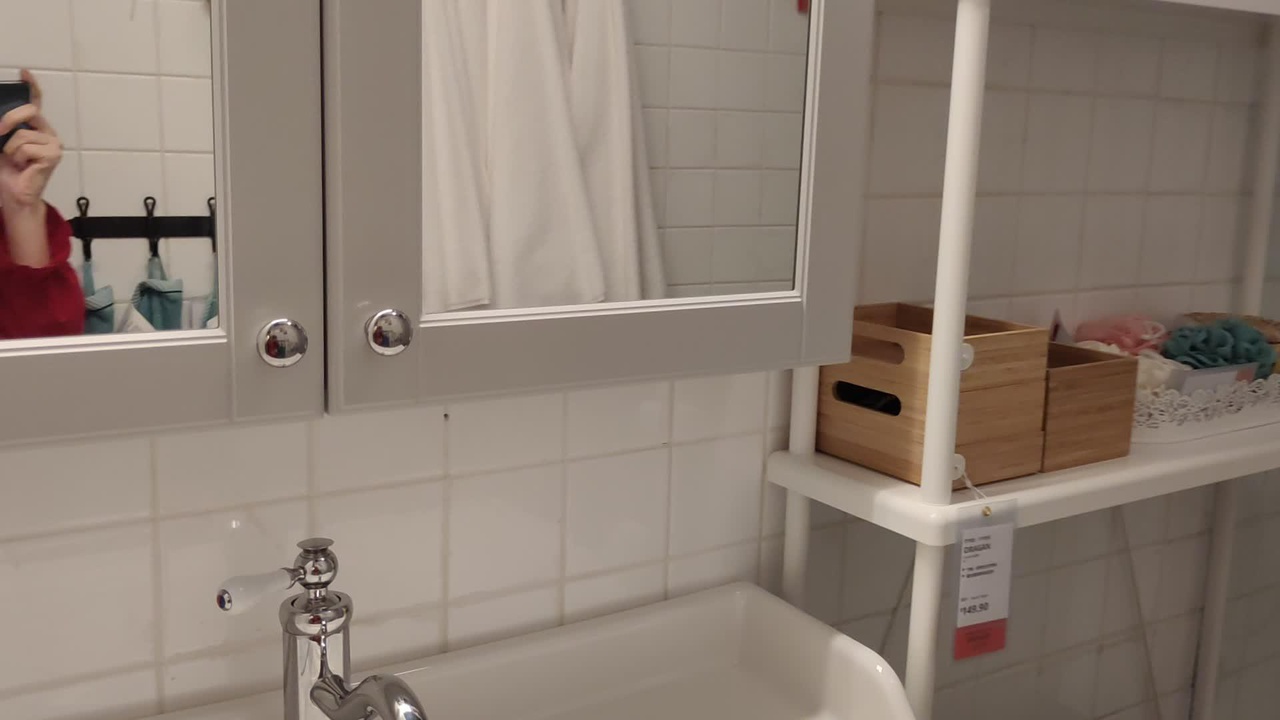}
	\end{subfigure}
	\begin{subfigure}{0.11\textwidth}
		\includegraphics[width=\textwidth]{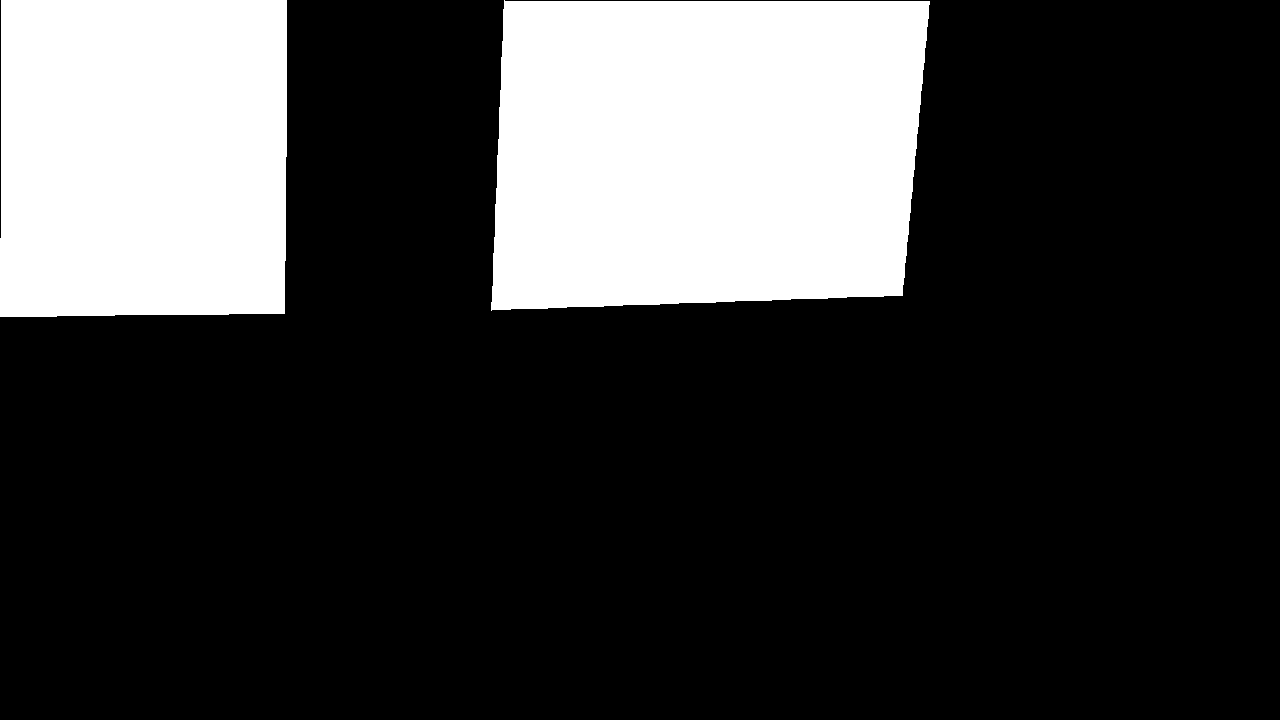}
	\end{subfigure}
	\begin{subfigure}{0.11\textwidth}
		\includegraphics[width=\textwidth]{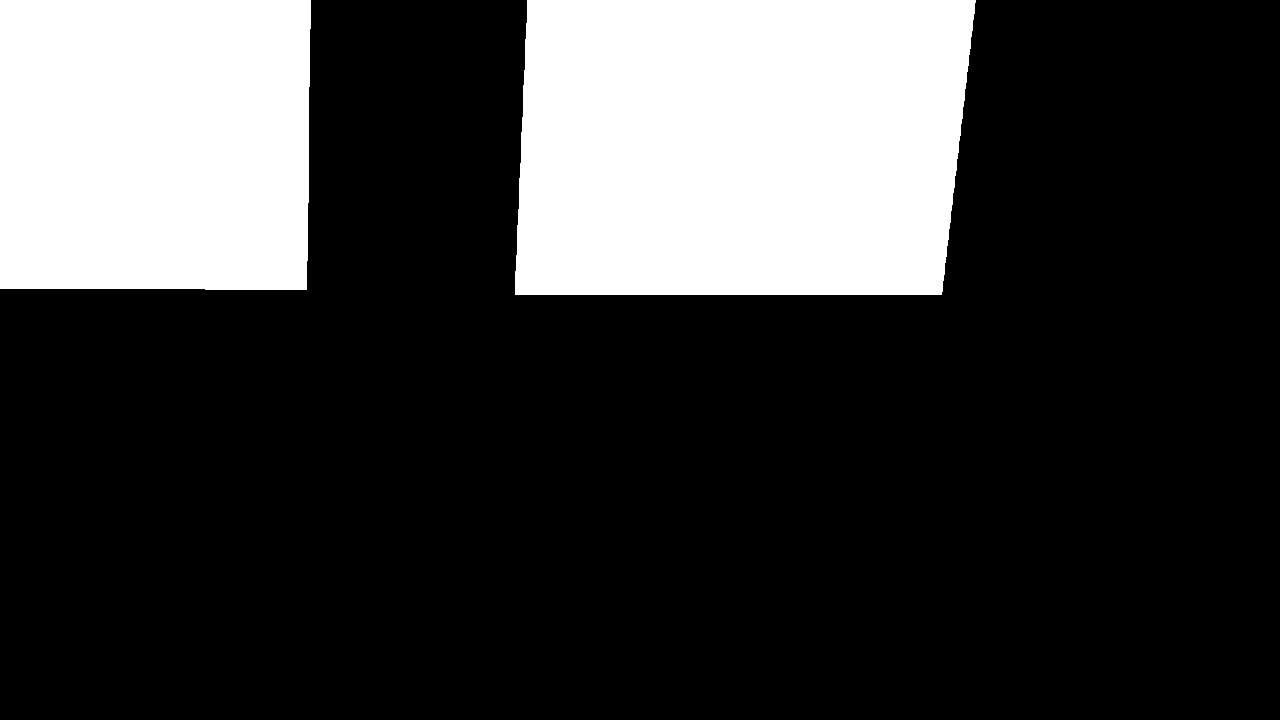}
	\end{subfigure}
	\begin{subfigure}{0.11\textwidth}
		\includegraphics[width=\textwidth]{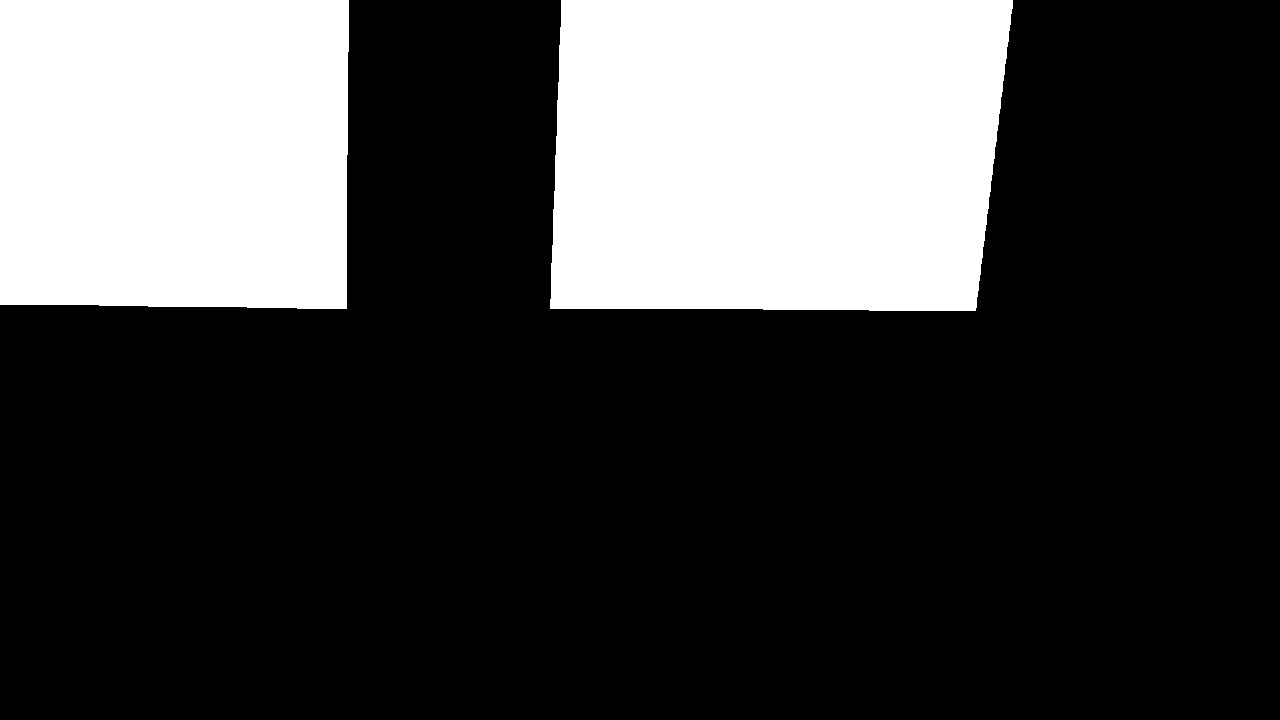}
	\end{subfigure}
	\begin{subfigure}{0.11\textwidth}
		\includegraphics[width=\textwidth]{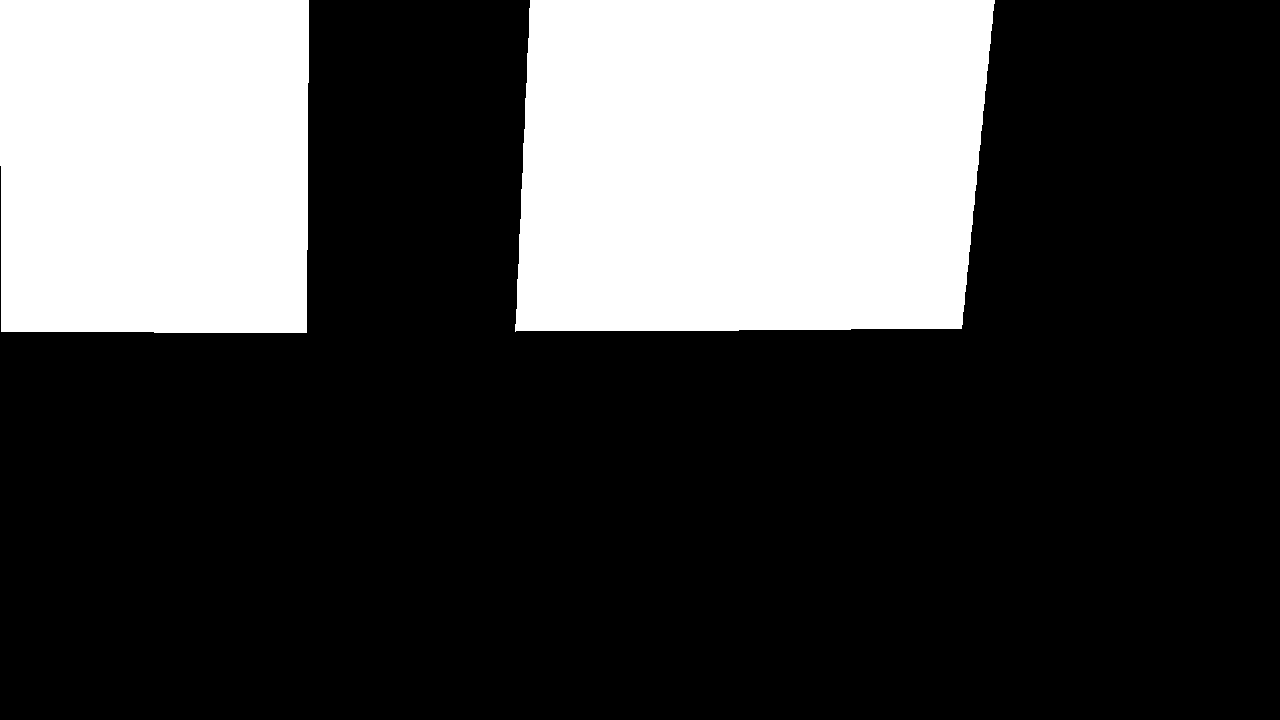}
	\end{subfigure}
	\begin{subfigure}{0.11\textwidth}
		\includegraphics[width=\textwidth]{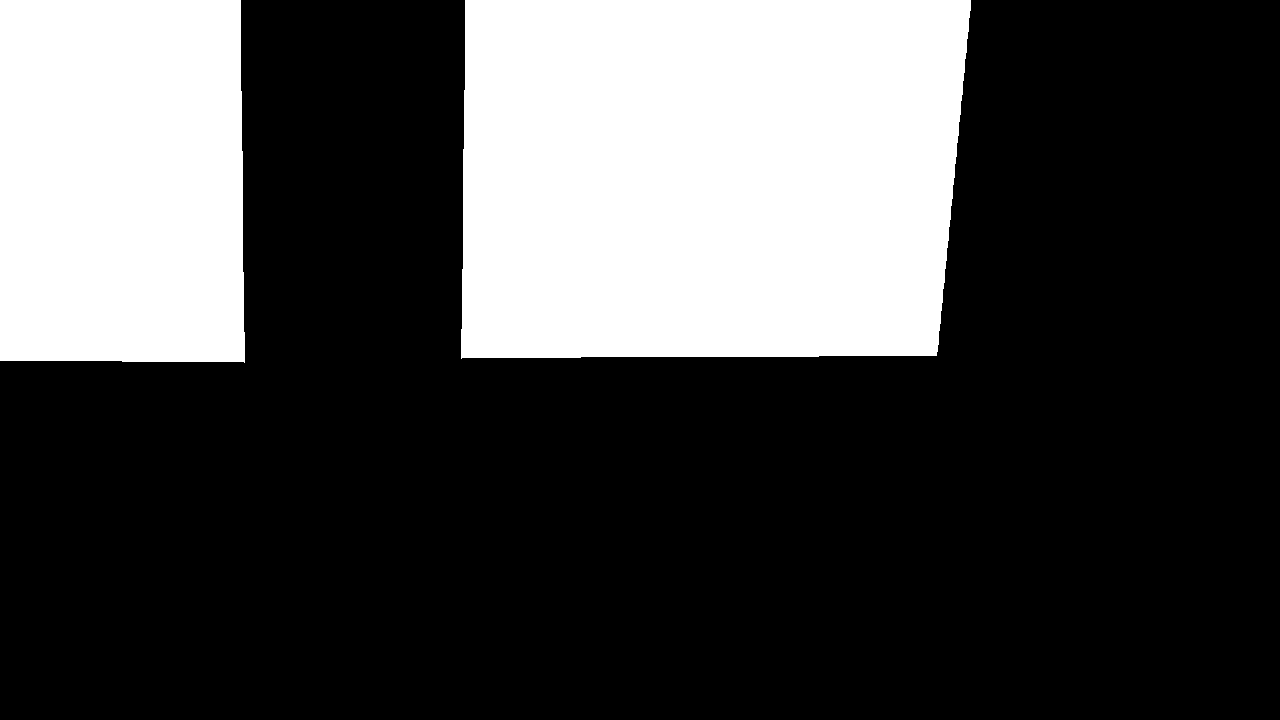}
	\end{subfigure}
	
	\vspace*{1.3mm}
	\begin{subfigure}{0.11\textwidth}
		\includegraphics[width=\textwidth]{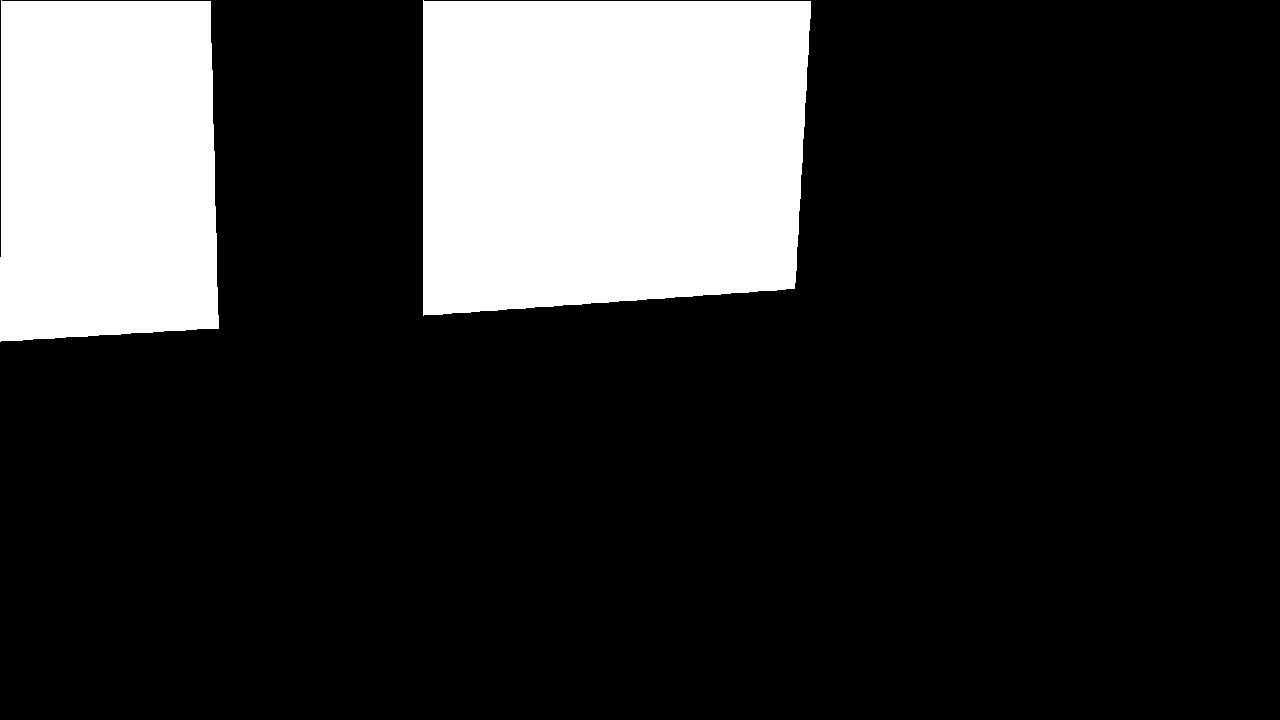}
	\end{subfigure}
	\begin{subfigure}{0.11\textwidth}
		\includegraphics[width=\textwidth]{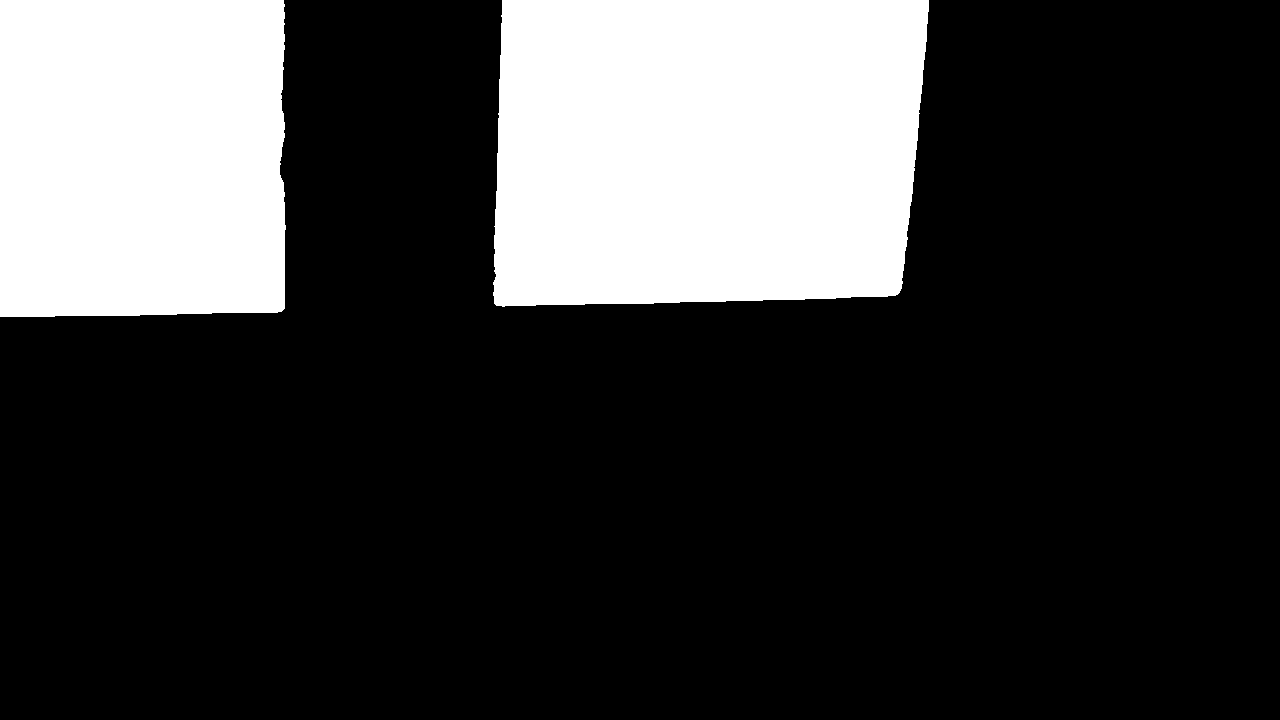}
	\end{subfigure}
	\begin{subfigure}{0.11\textwidth}
		\includegraphics[width=\textwidth]{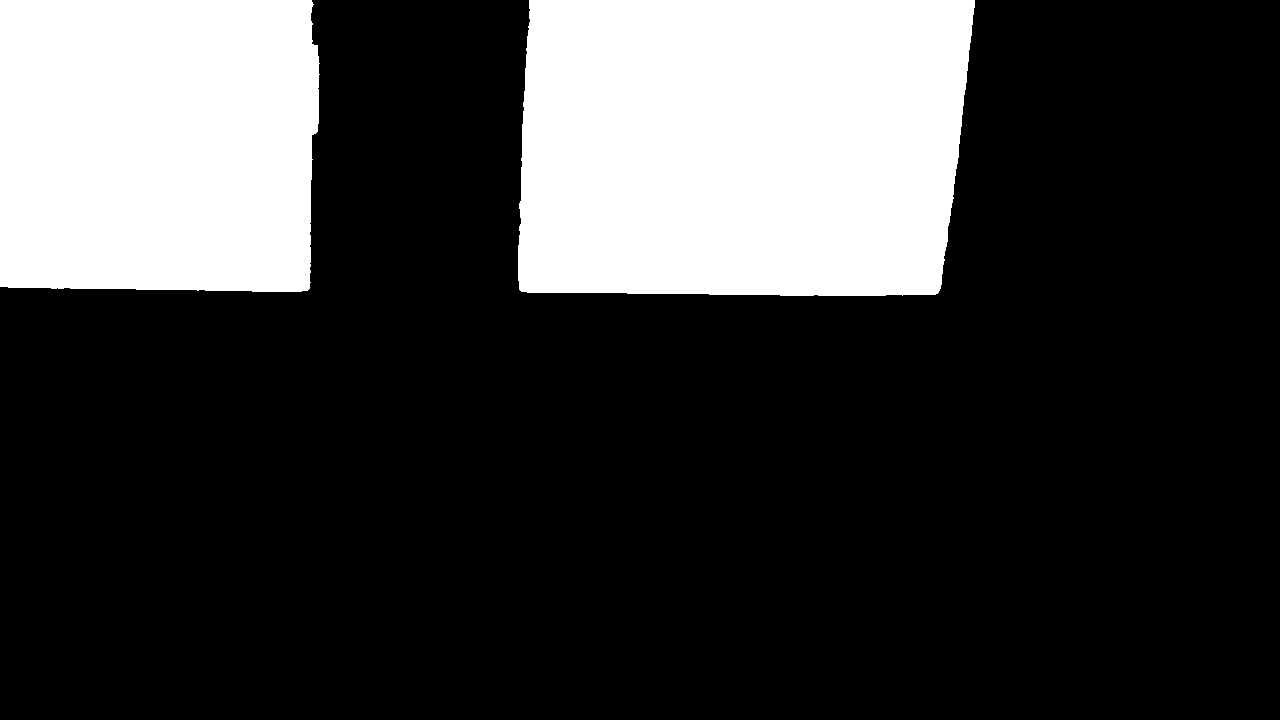}
	\end{subfigure}
	\begin{subfigure}{0.11\textwidth}
		\includegraphics[width=\textwidth]{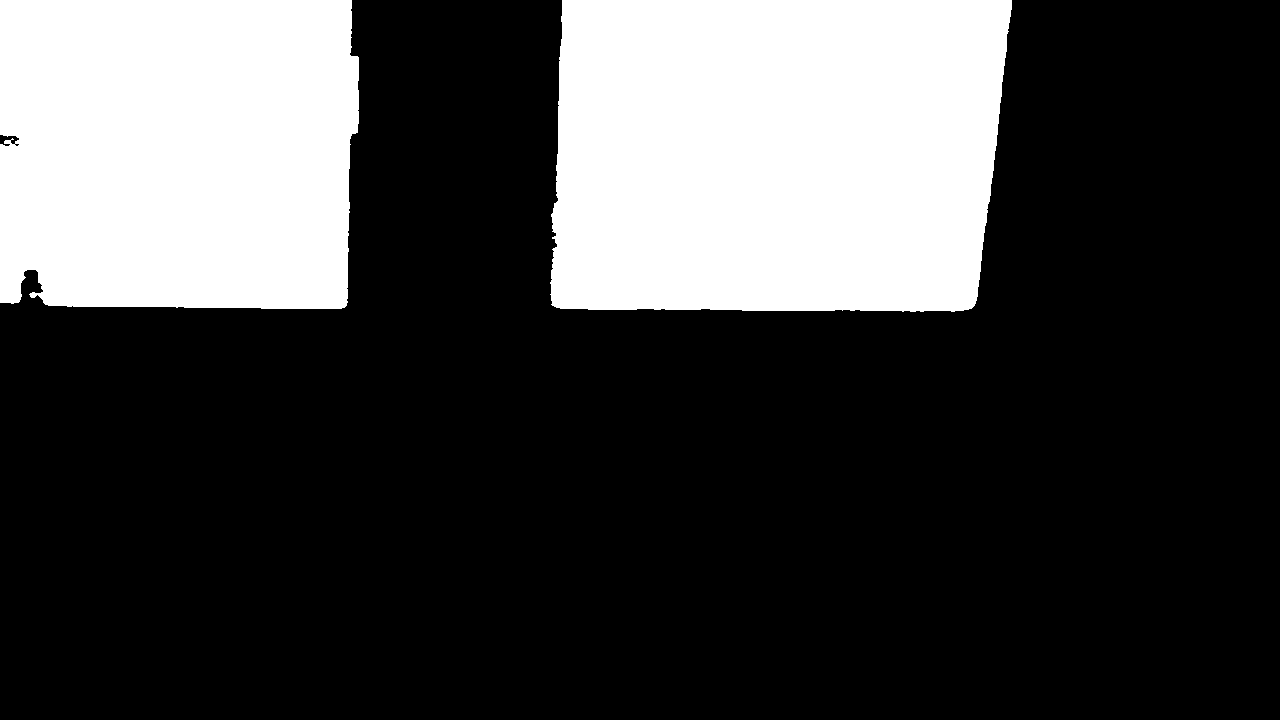}
	\end{subfigure}
	\begin{subfigure}{0.11\textwidth}
		\includegraphics[width=\textwidth]{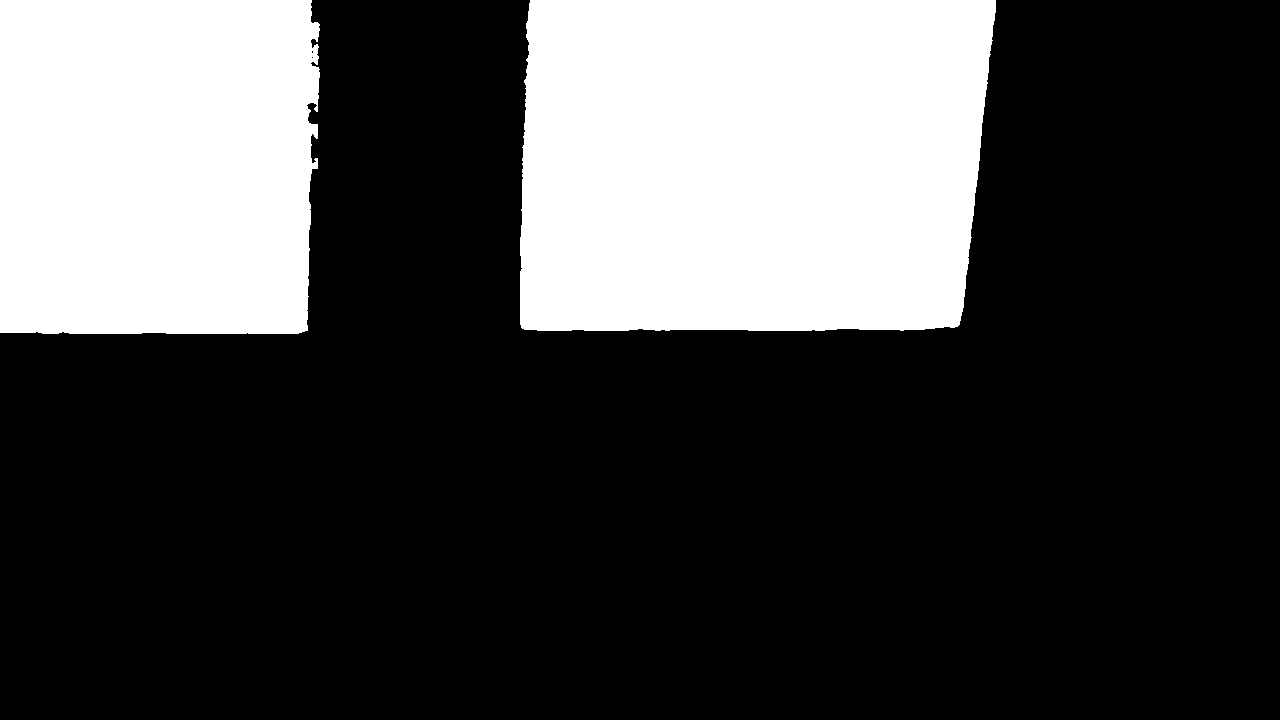}
	\end{subfigure}
	\begin{subfigure}{0.11\textwidth}
		\includegraphics[width=\textwidth]{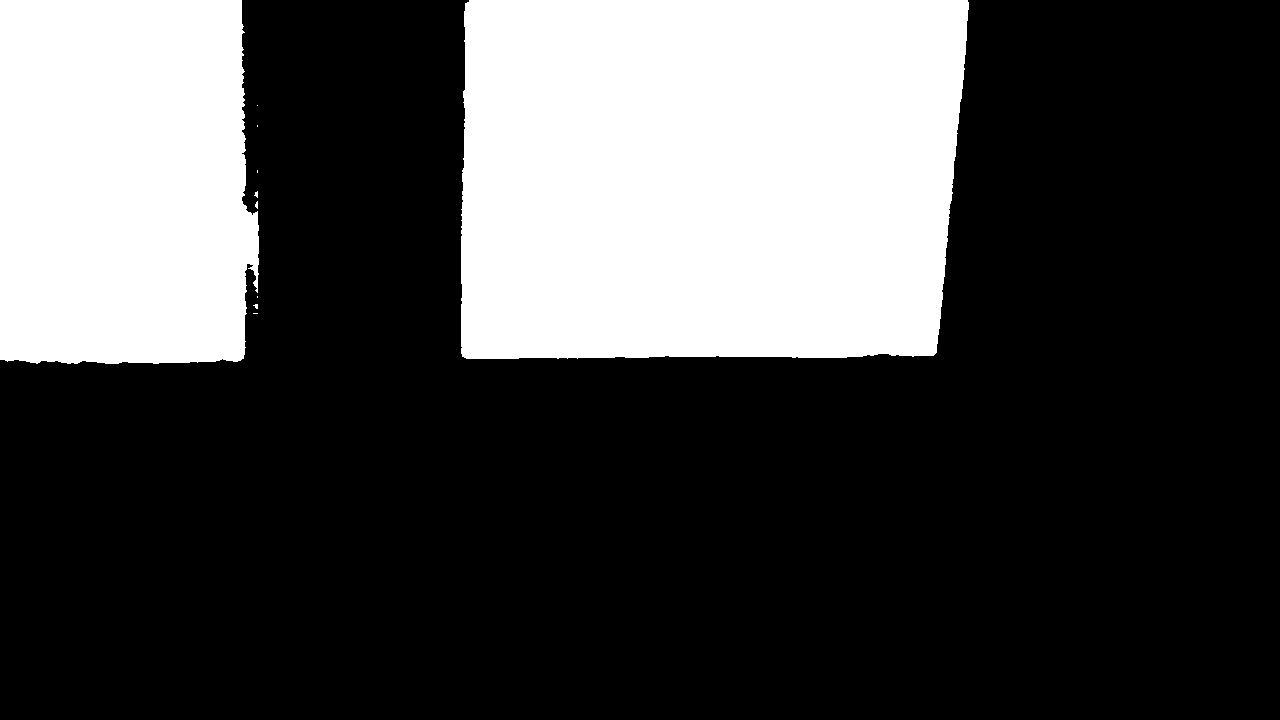}
	\end{subfigure}
	
	\vspace*{1.3mm}
	\begin{subfigure}{0.11\textwidth}
		\includegraphics[width=\textwidth]{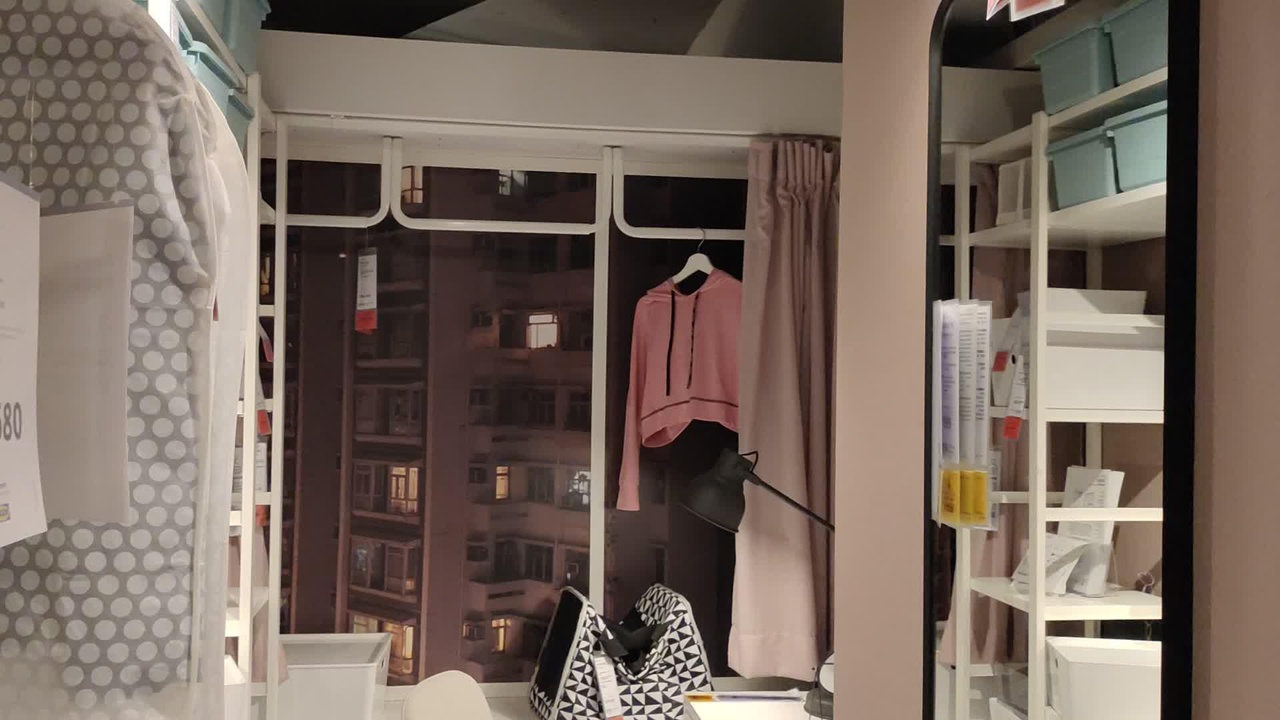}
	\end{subfigure}
	\begin{subfigure}{0.11\textwidth}
		\includegraphics[width=\textwidth]{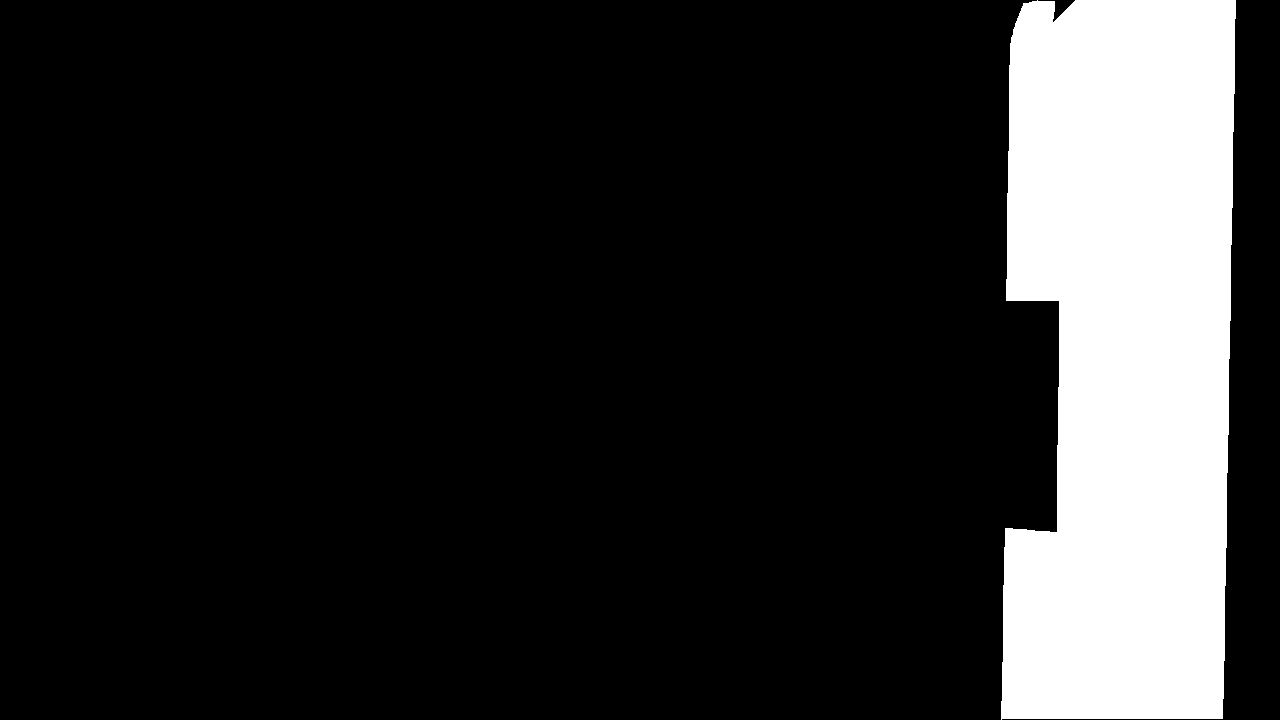}
	\end{subfigure}
	\begin{subfigure}{0.11\textwidth}
		\includegraphics[width=\textwidth]{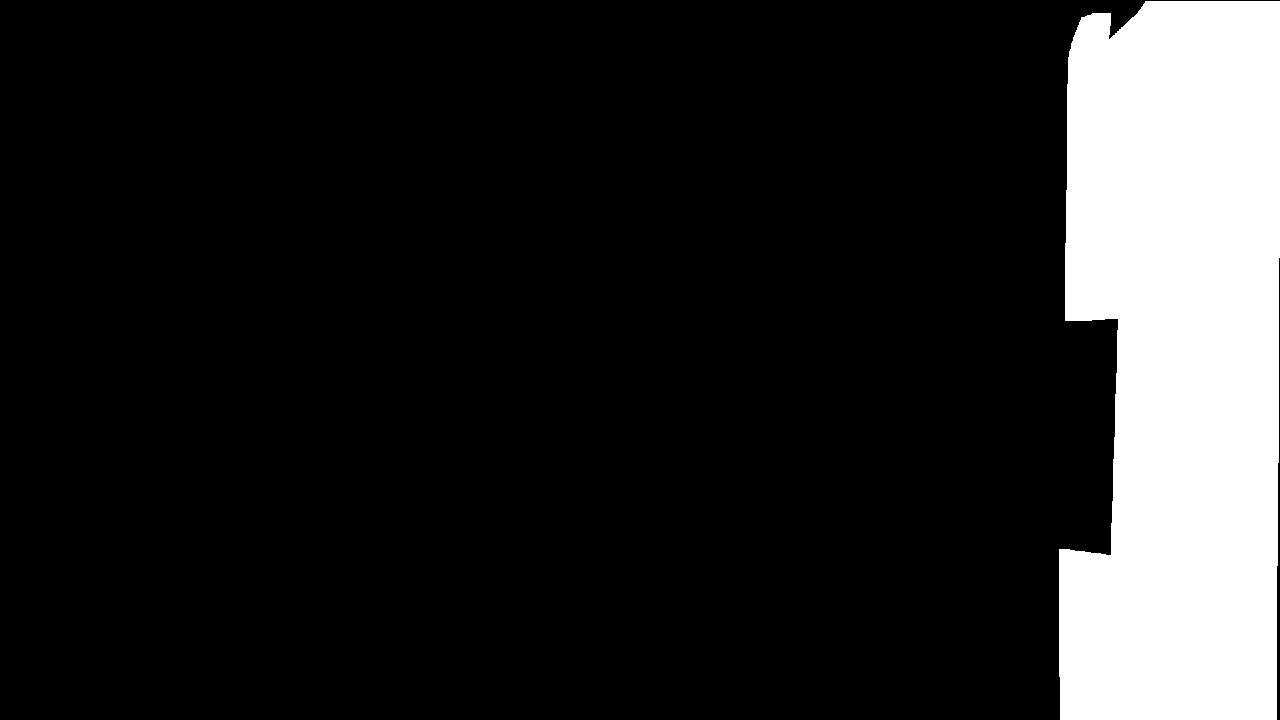}
	\end{subfigure}
	\begin{subfigure}{0.11\textwidth}
		\includegraphics[width=\textwidth]{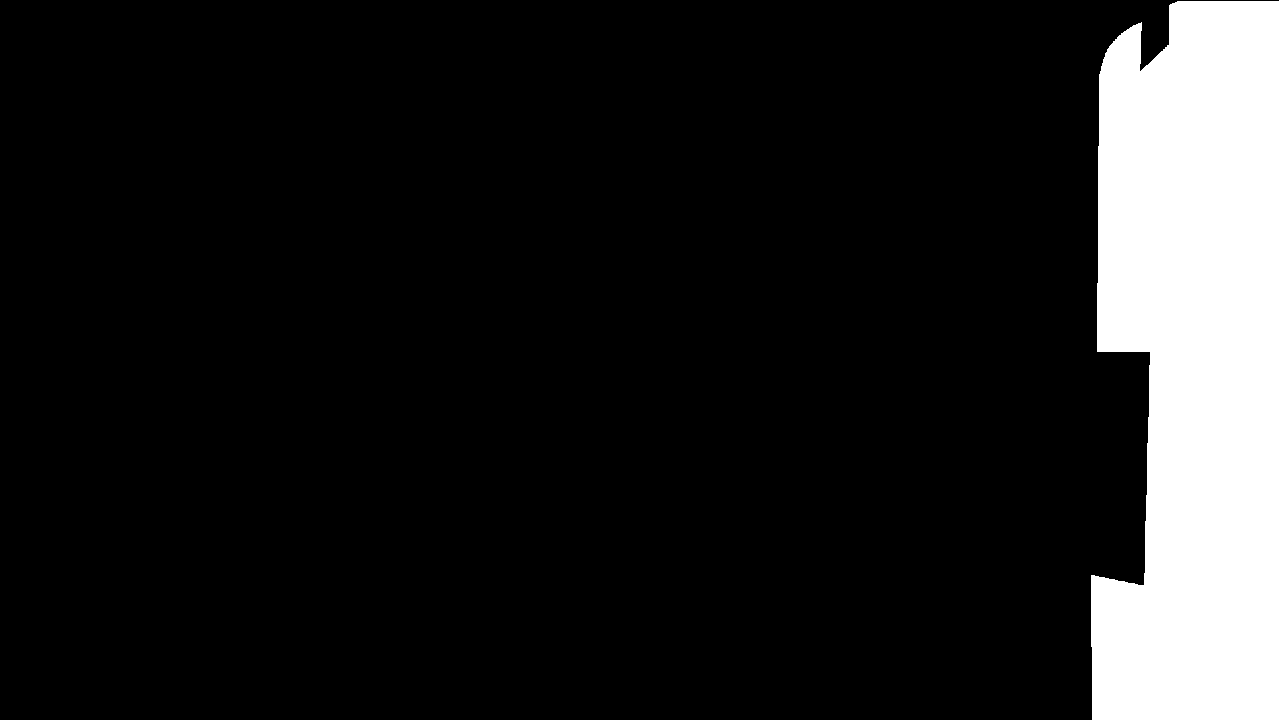}
	\end{subfigure}
	\begin{subfigure}{0.11\textwidth}
		\includegraphics[width=\textwidth]{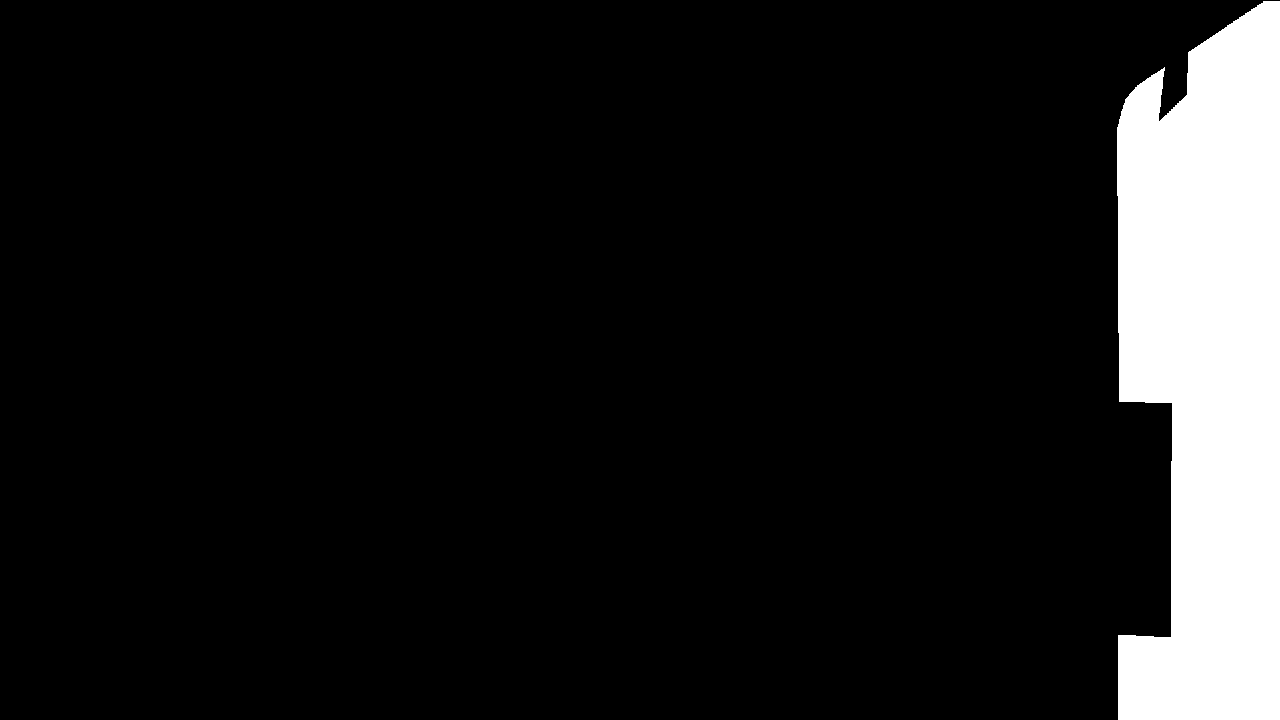}
	\end{subfigure}
	\begin{subfigure}{0.11\textwidth}
		\includegraphics[width=\textwidth]{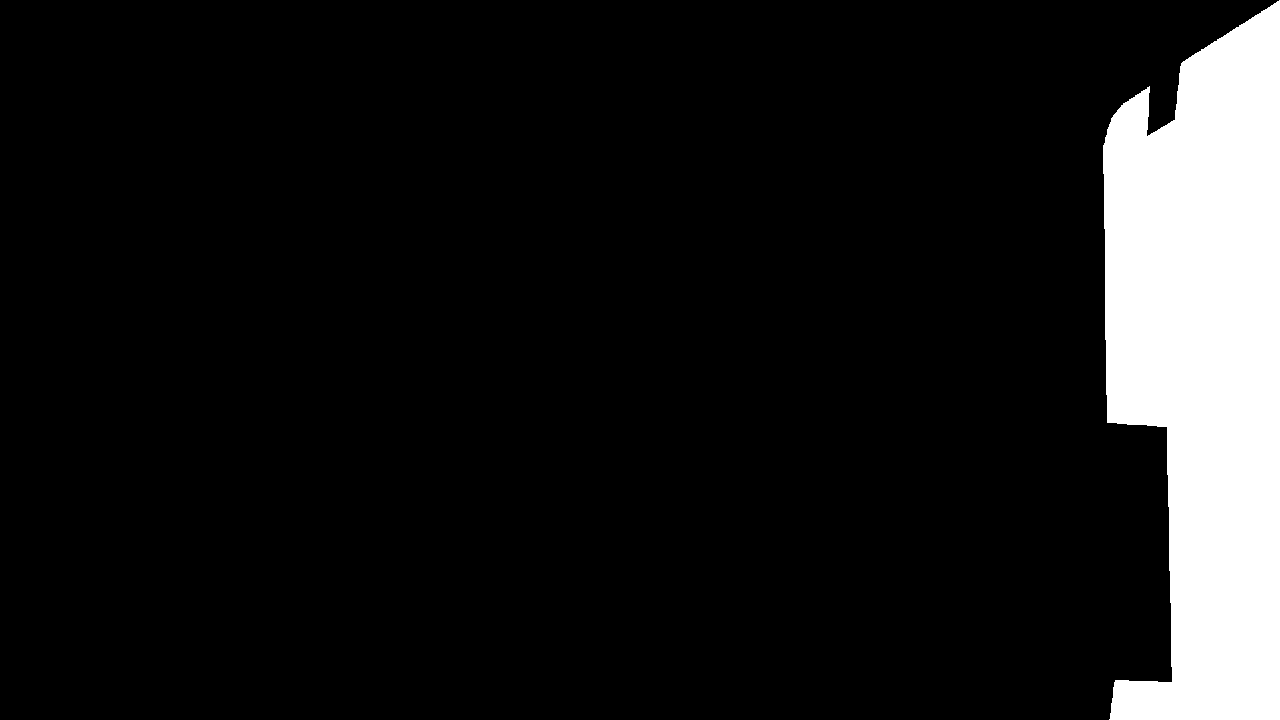}
	\end{subfigure}
	
	\vspace*{1.3mm}
	\begin{subfigure}{0.11\textwidth}
		\includegraphics[width=\textwidth]{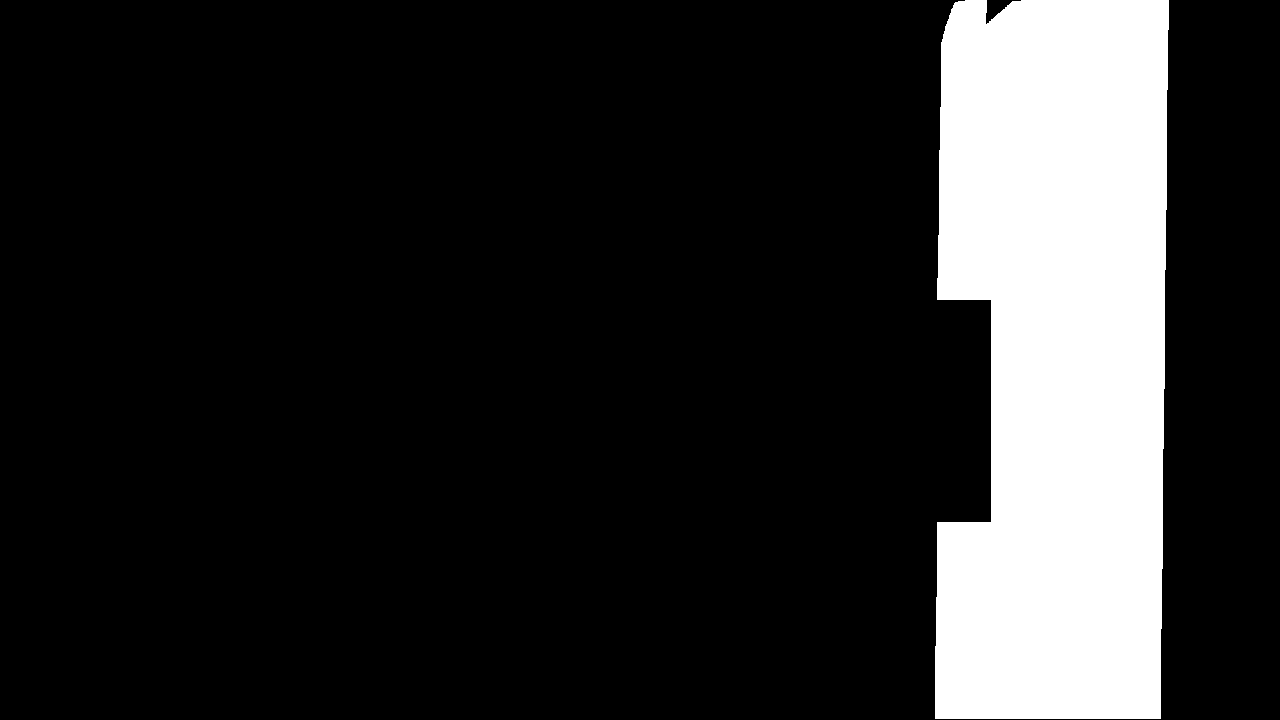}
	\end{subfigure}
	\begin{subfigure}{0.11\textwidth}
		\includegraphics[width=\textwidth]{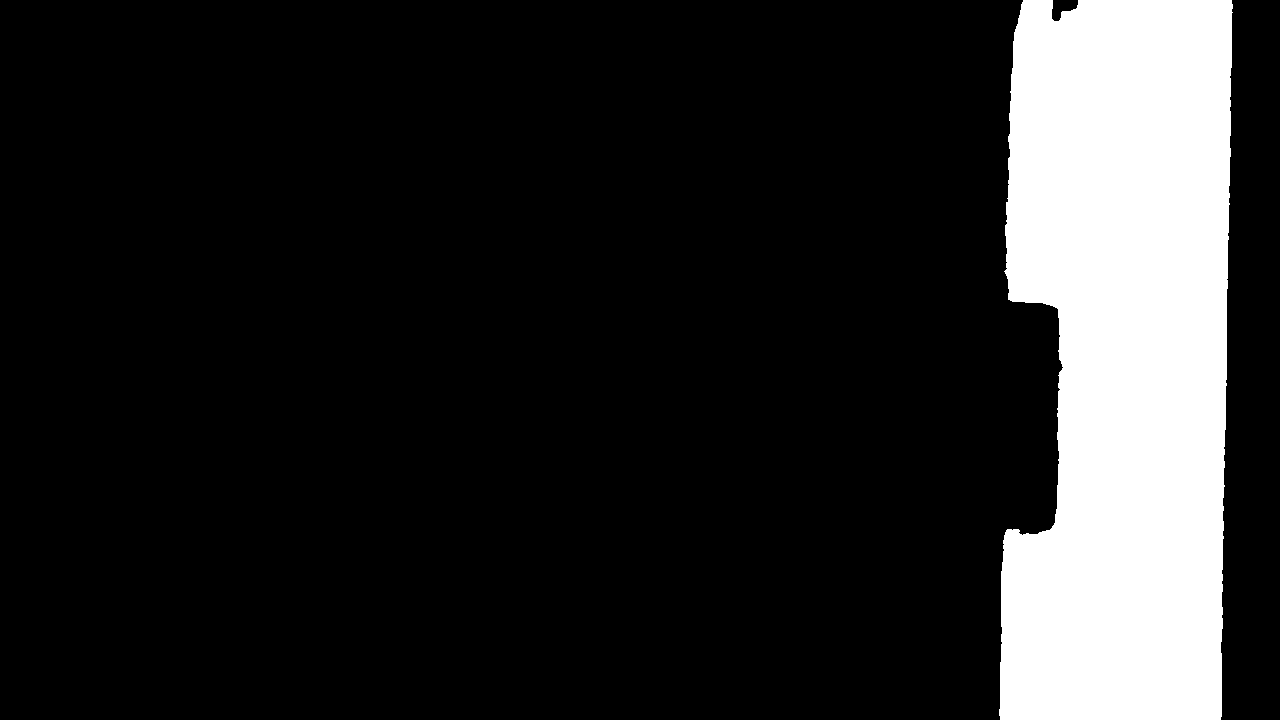}
	\end{subfigure}
	\begin{subfigure}{0.11\textwidth}
		\includegraphics[width=\textwidth]{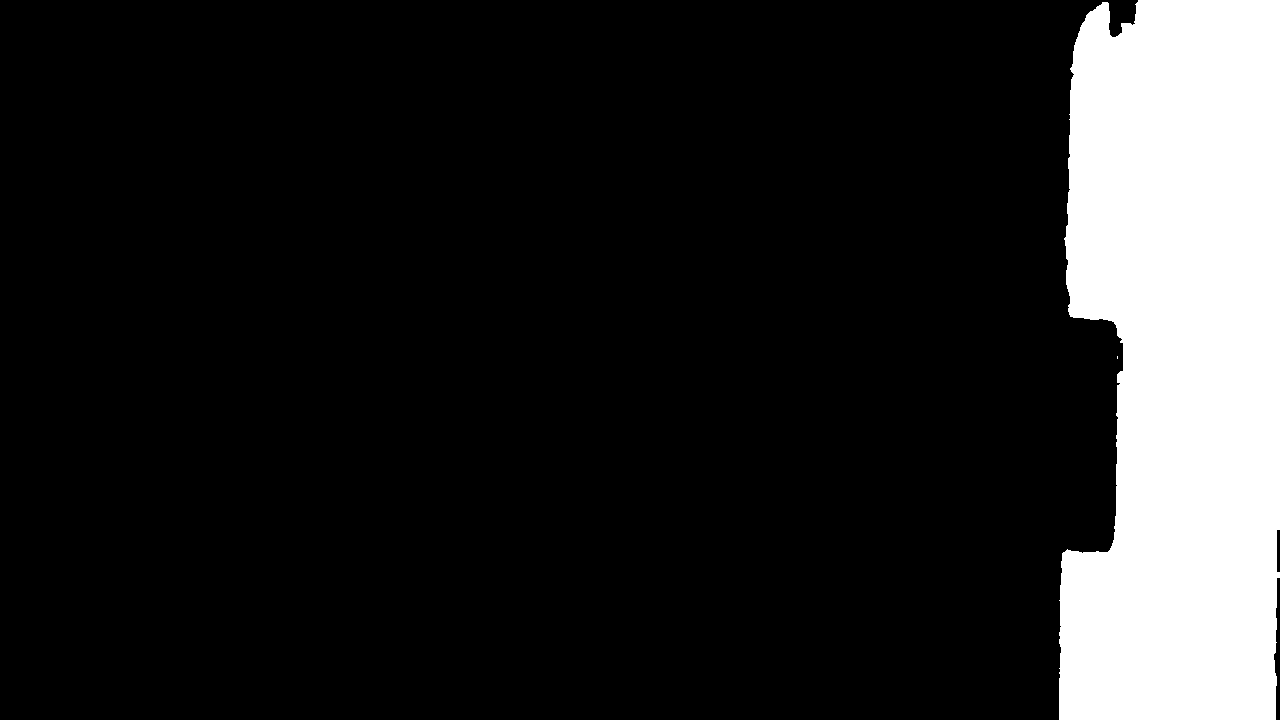}
	\end{subfigure}
	\begin{subfigure}{0.11\textwidth}
		\includegraphics[width=\textwidth]{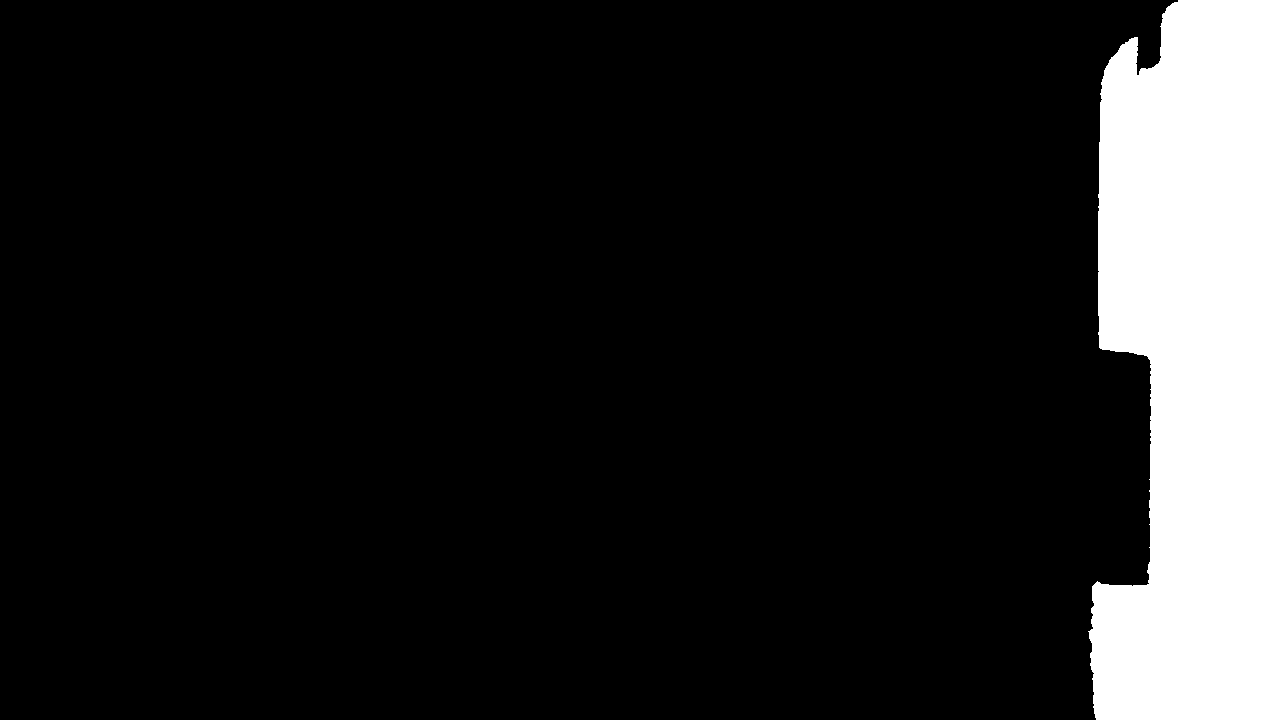}
	\end{subfigure}
	\begin{subfigure}{0.11\textwidth}
		\includegraphics[width=\textwidth]{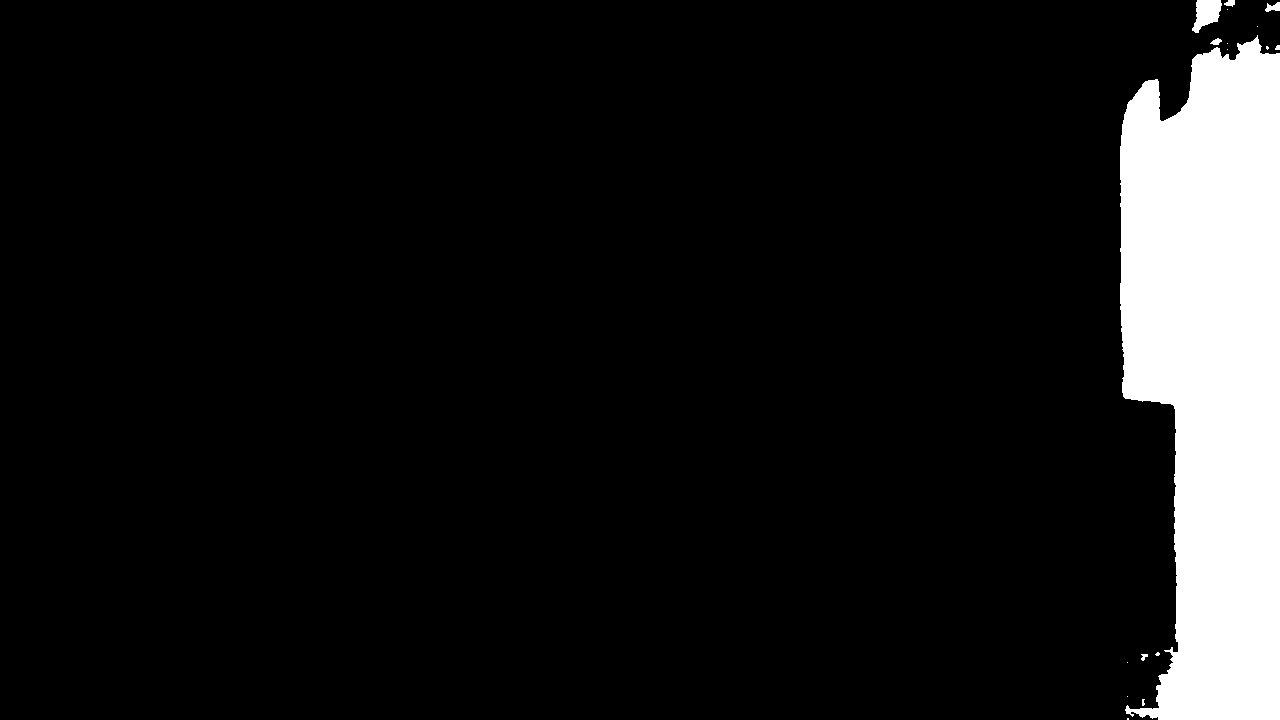}
	\end{subfigure}
	\begin{subfigure}{0.11\textwidth}
		\includegraphics[width=\textwidth]{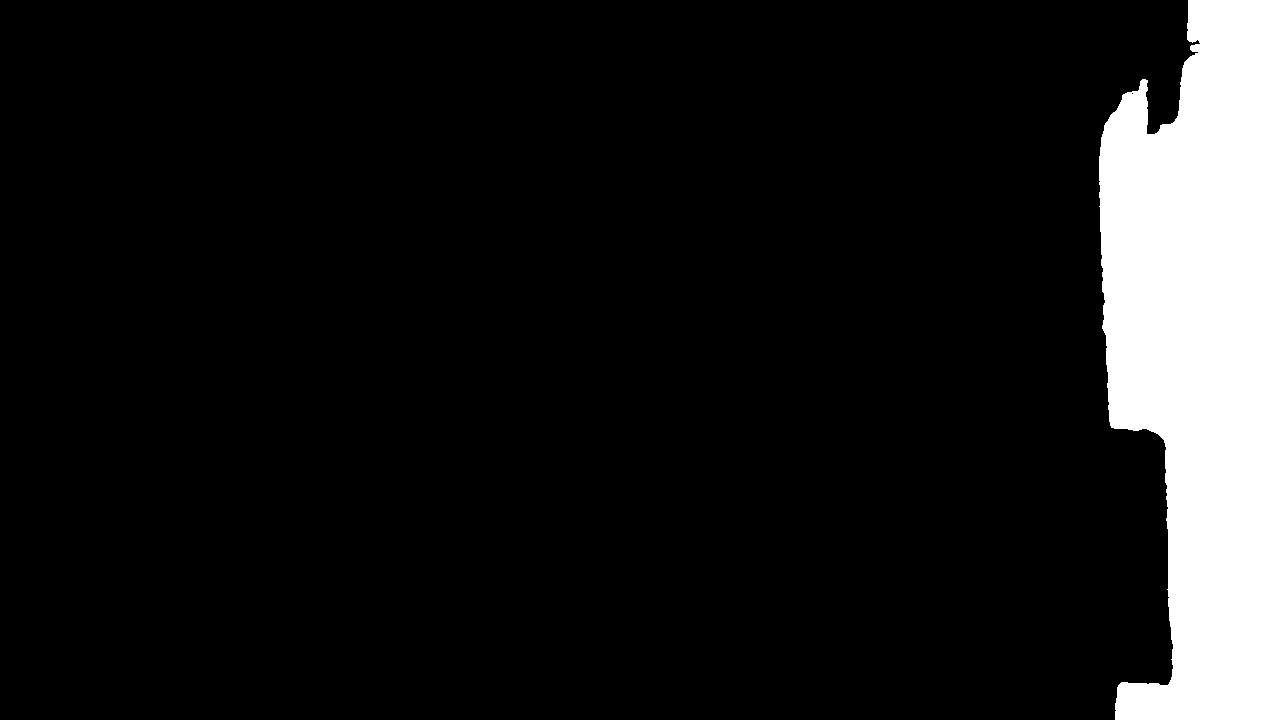}
	\end{subfigure}

	\vspace*{1.3mm}
	\begin{subfigure}{0.11\textwidth}
		\includegraphics[width=\textwidth]{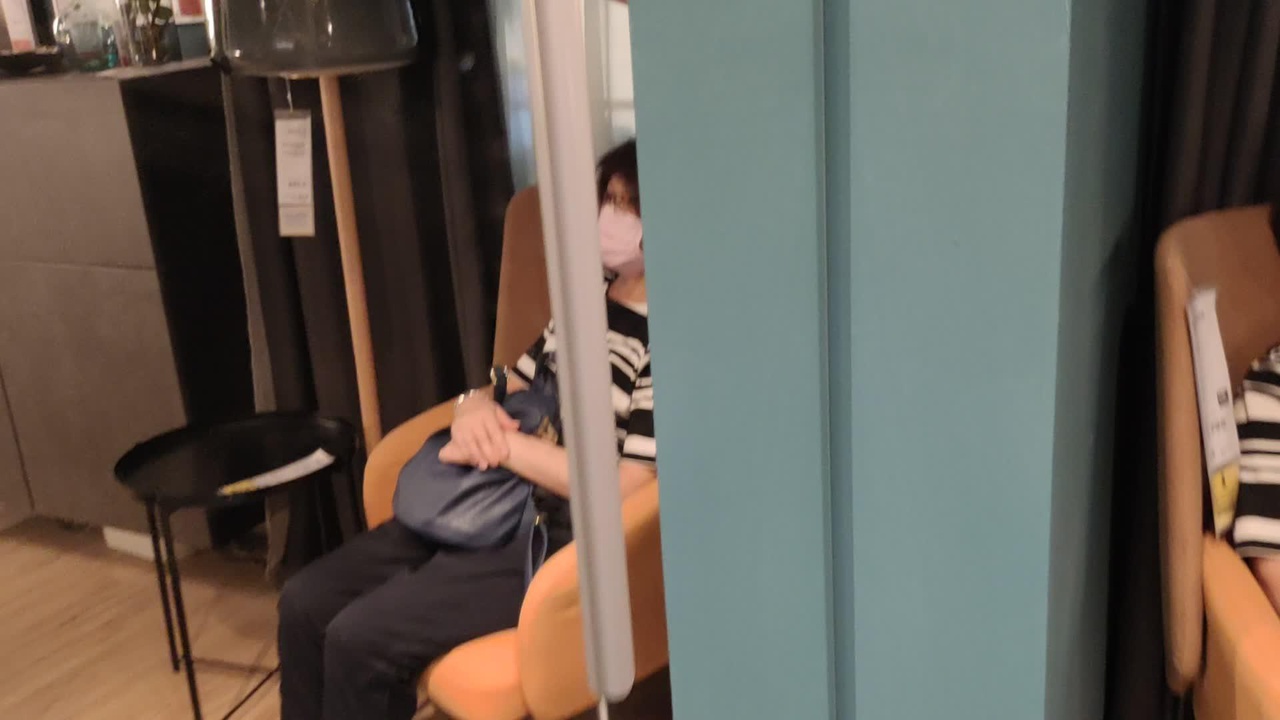}
	\end{subfigure}
	\begin{subfigure}{0.11\textwidth}
		\includegraphics[width=\textwidth]{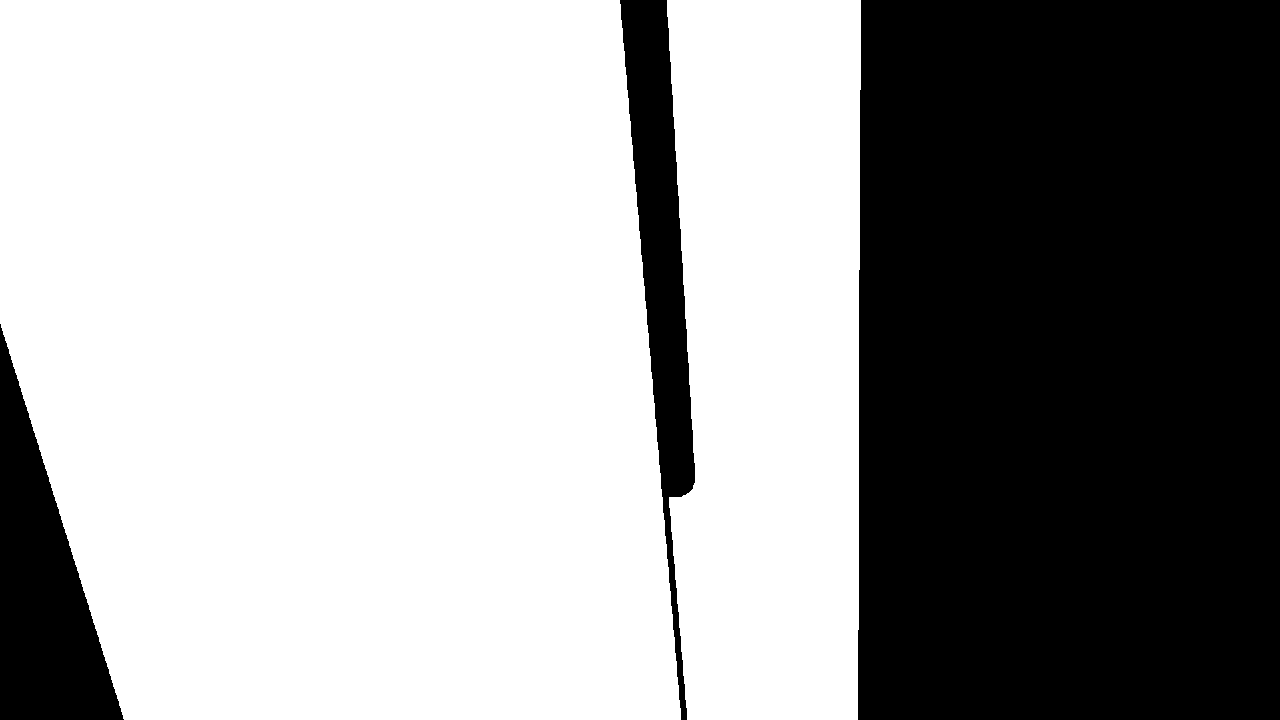}
	\end{subfigure}
	\begin{subfigure}{0.11\textwidth}
		\includegraphics[width=\textwidth]{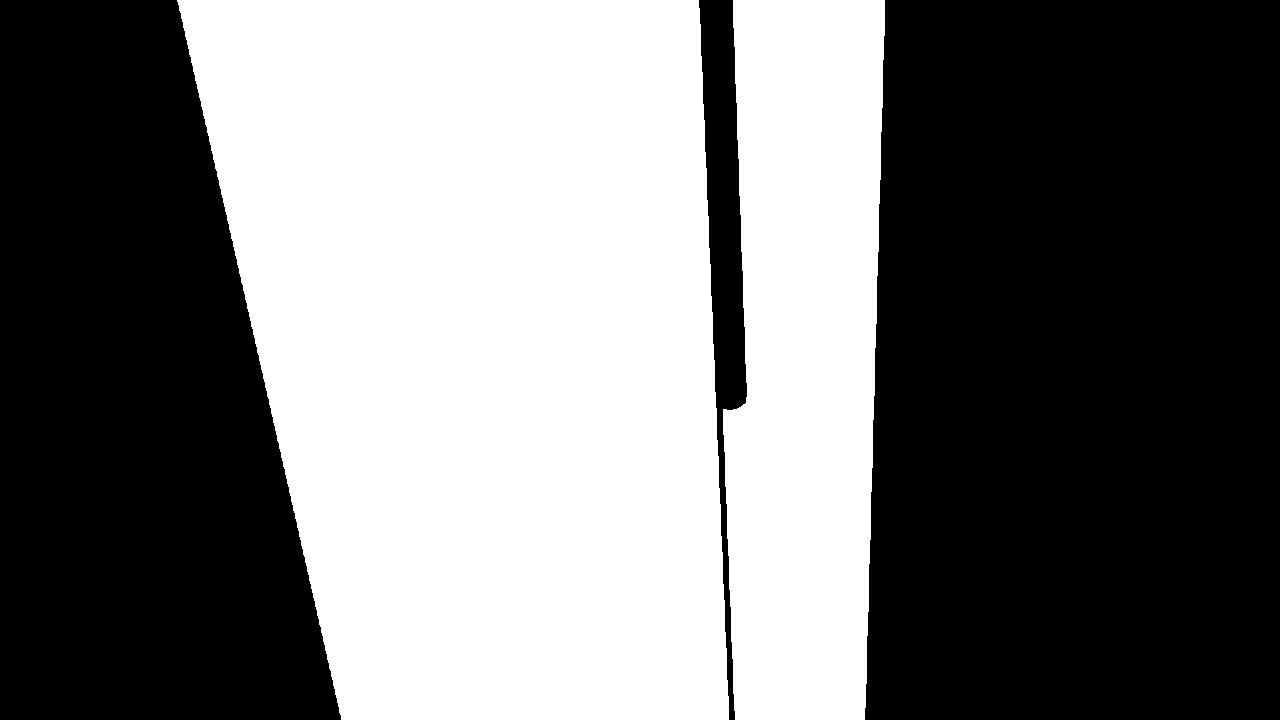}
	\end{subfigure}
	\begin{subfigure}{0.11\textwidth}
		\includegraphics[width=\textwidth]{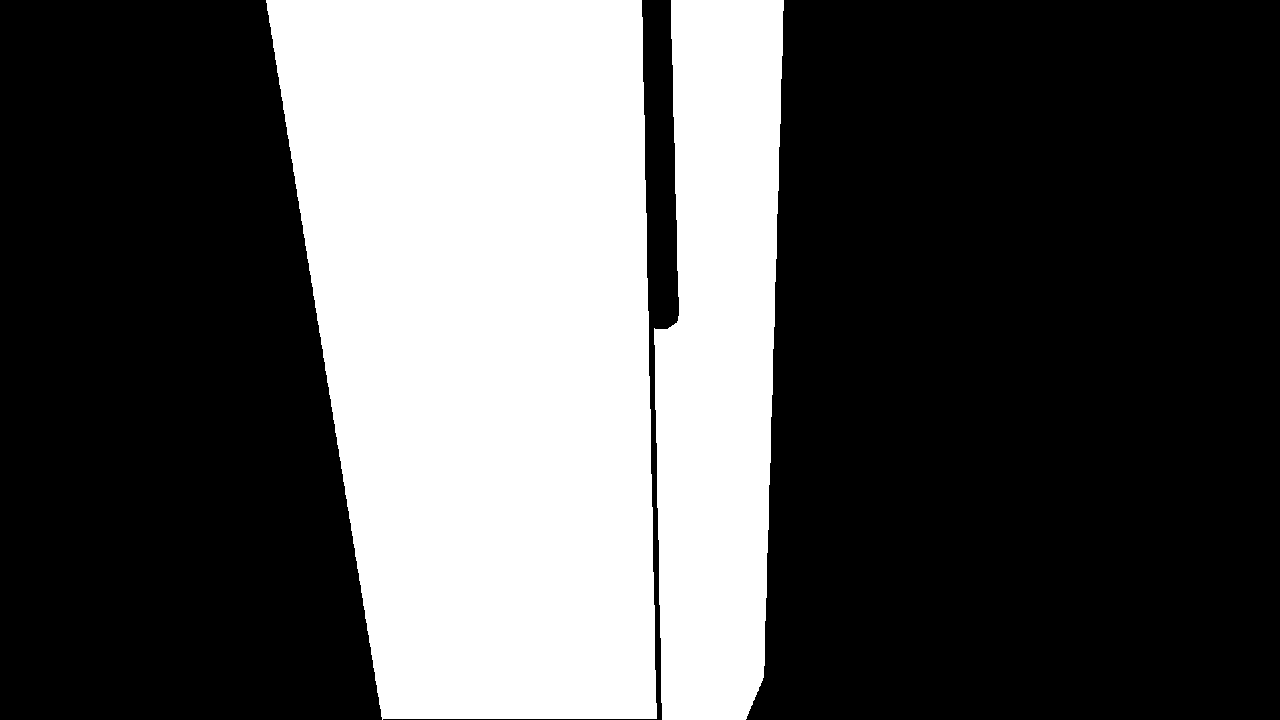}
	\end{subfigure}
	\begin{subfigure}{0.11\textwidth}
		\includegraphics[width=\textwidth]{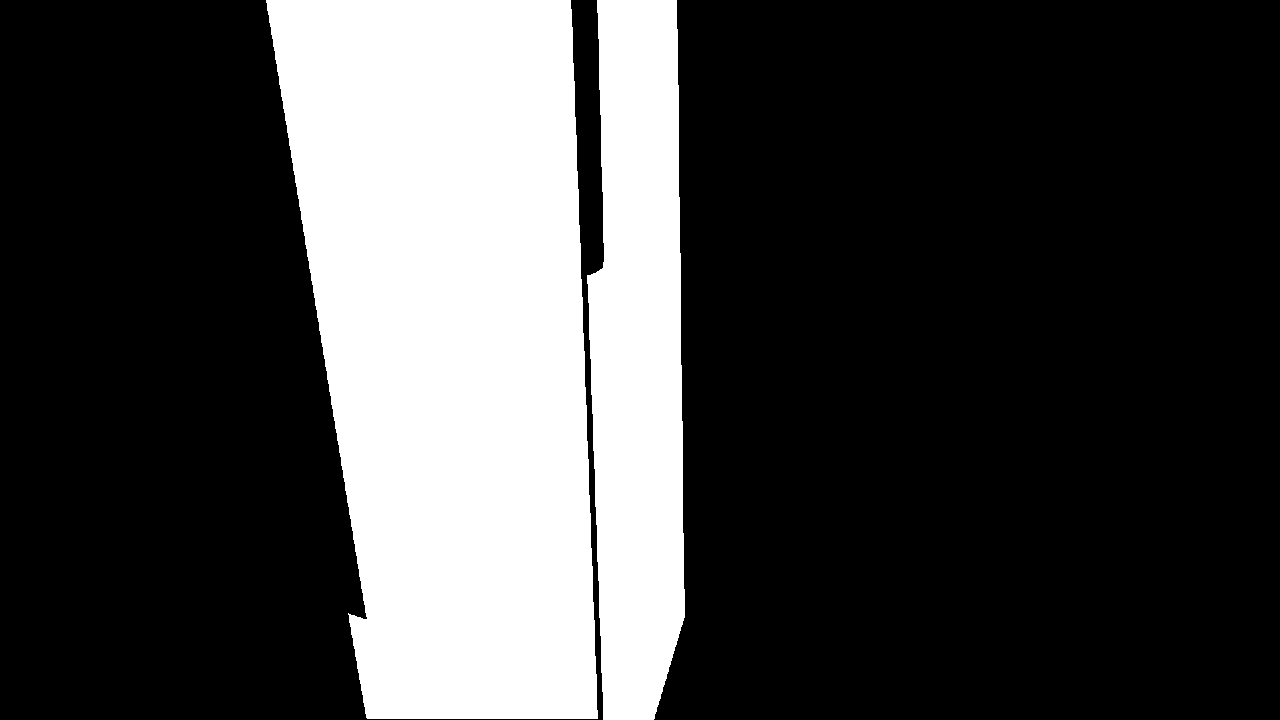}
	\end{subfigure}
	\begin{subfigure}{0.11\textwidth}
		\includegraphics[width=\textwidth]{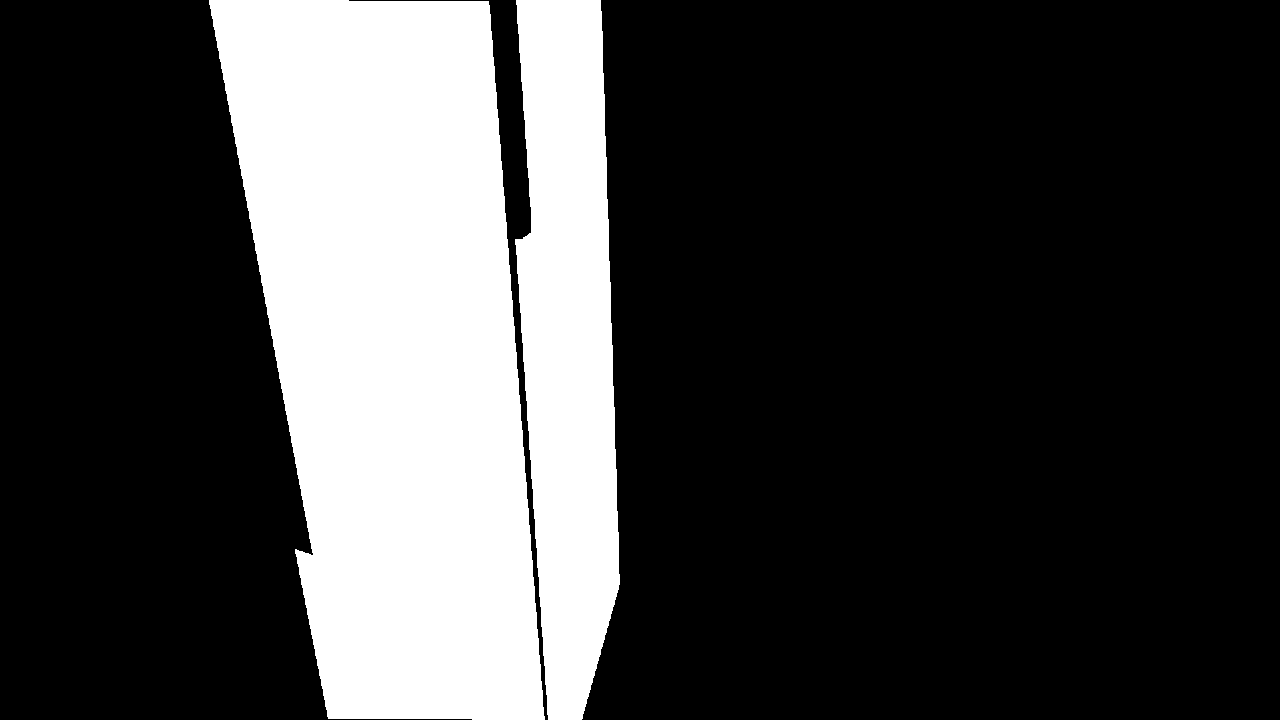}
	\end{subfigure}
	
	\vspace*{1.3mm}
	\begin{subfigure}{0.11\textwidth}
		\includegraphics[width=\textwidth]{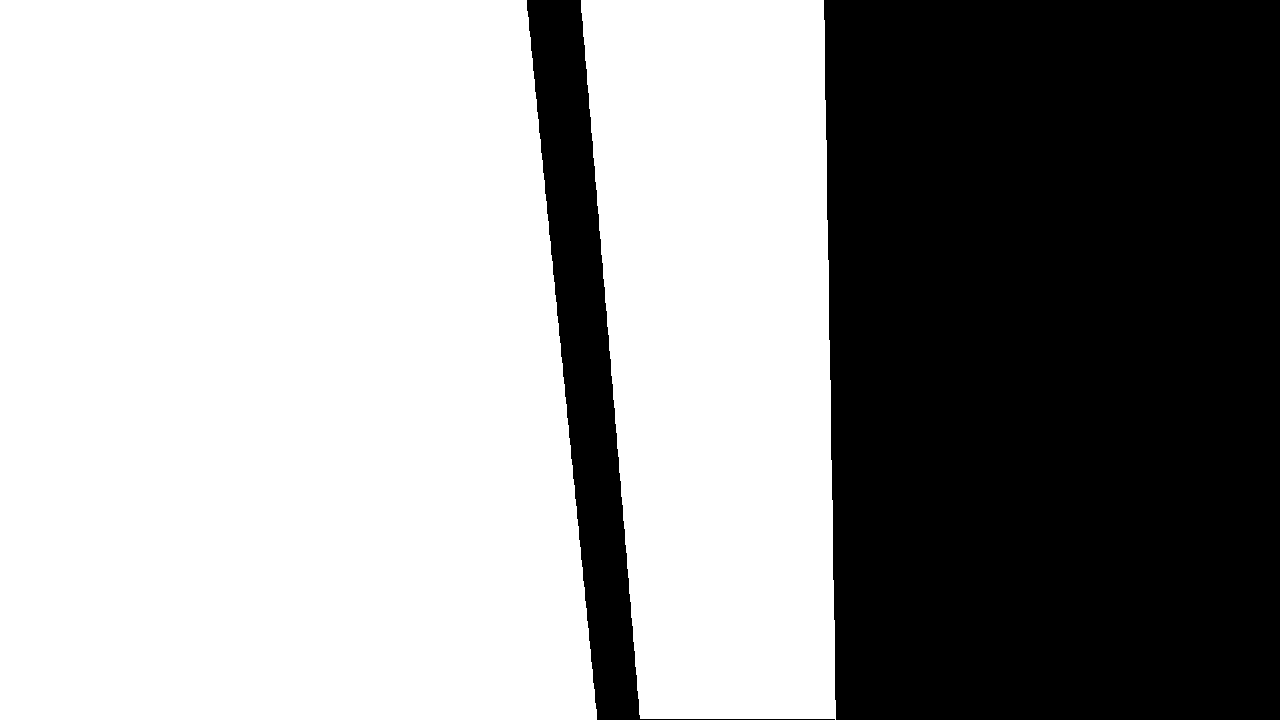}
	\end{subfigure}
	\begin{subfigure}{0.11\textwidth}
		\includegraphics[width=\textwidth]{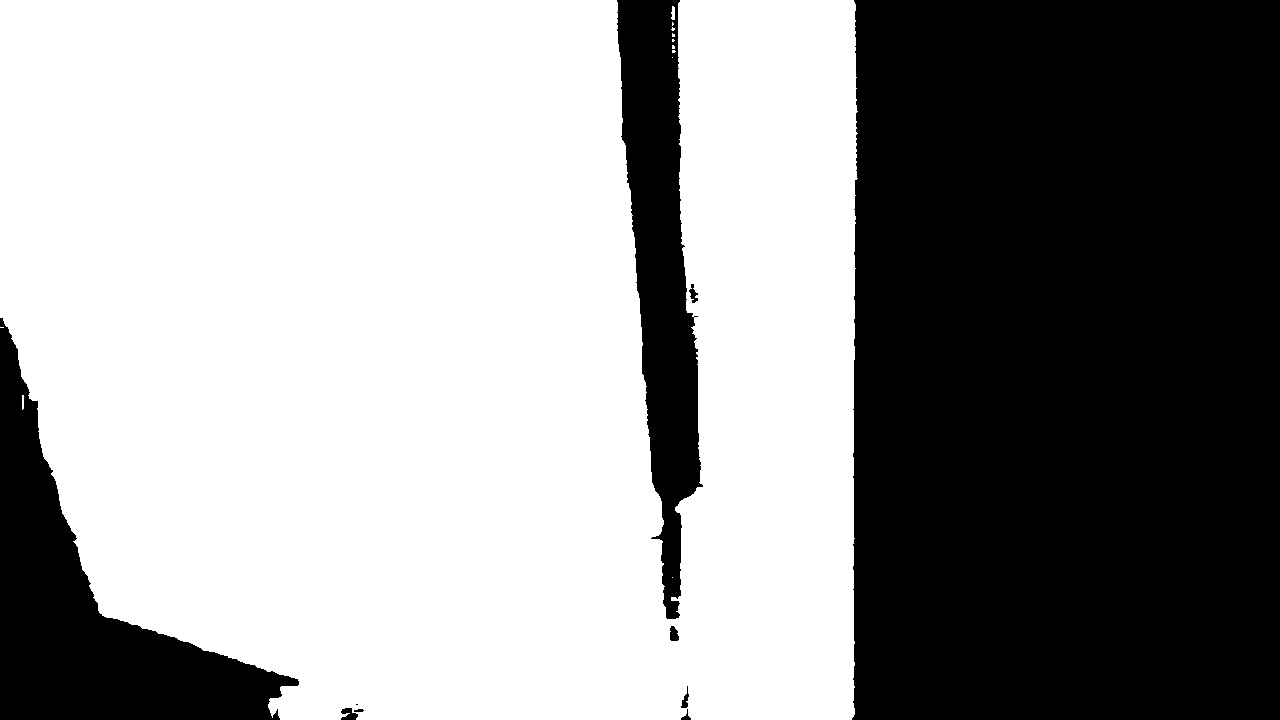}
	\end{subfigure}
	\begin{subfigure}{0.11\textwidth}
		\includegraphics[width=\textwidth]{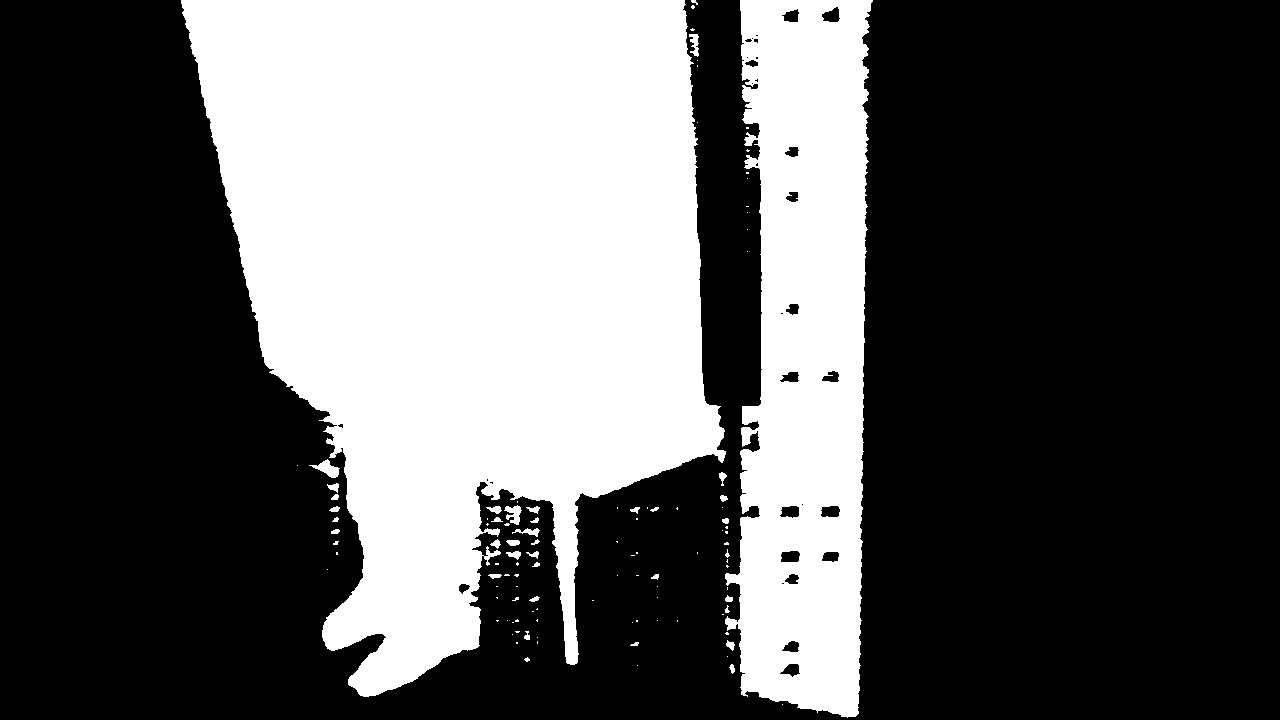}
	\end{subfigure}
	\begin{subfigure}{0.11\textwidth}
		\includegraphics[width=\textwidth]{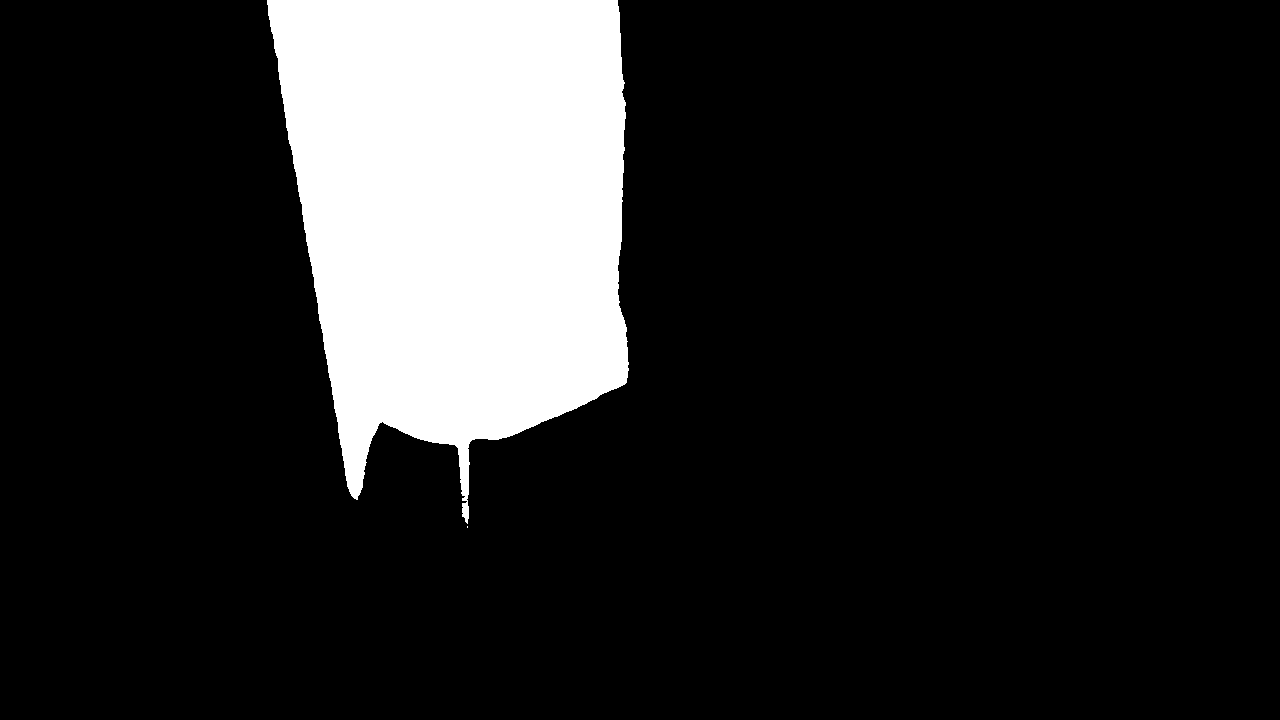}
	\end{subfigure}
	\begin{subfigure}{0.11\textwidth}
		\includegraphics[width=\textwidth]{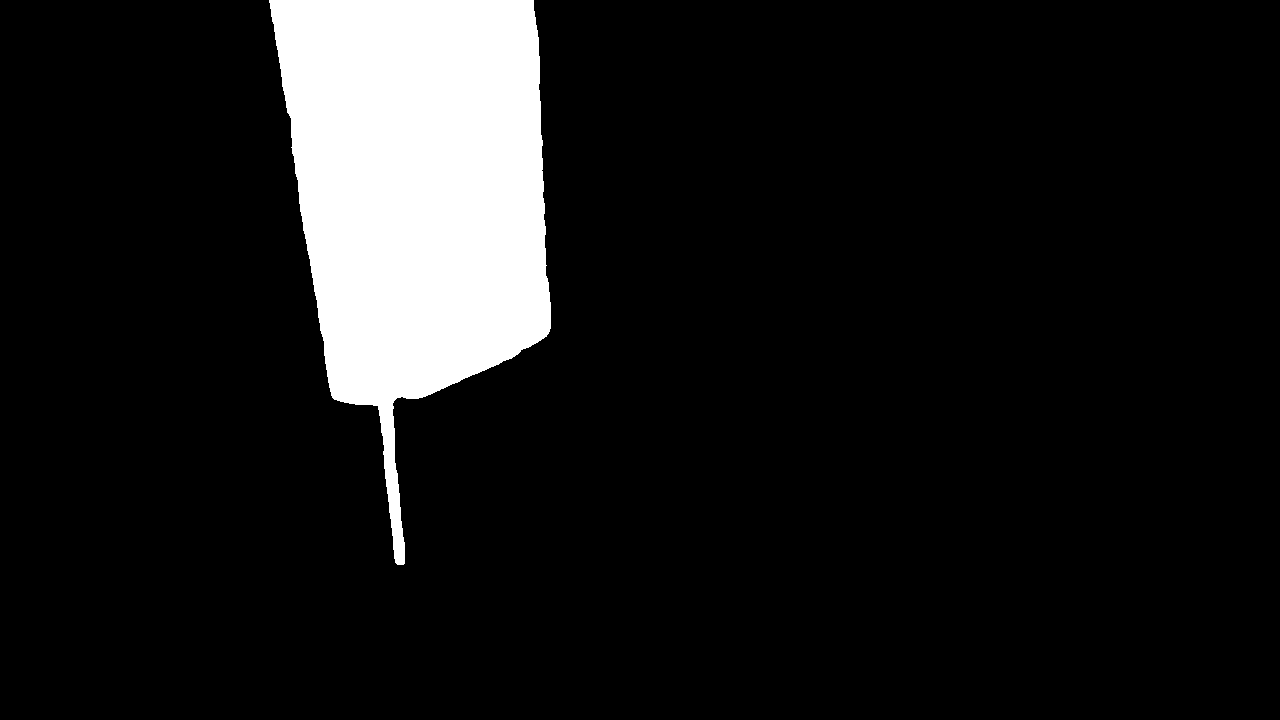}
	\end{subfigure}
	\begin{subfigure}{0.11\textwidth}
		\includegraphics[width=\textwidth]{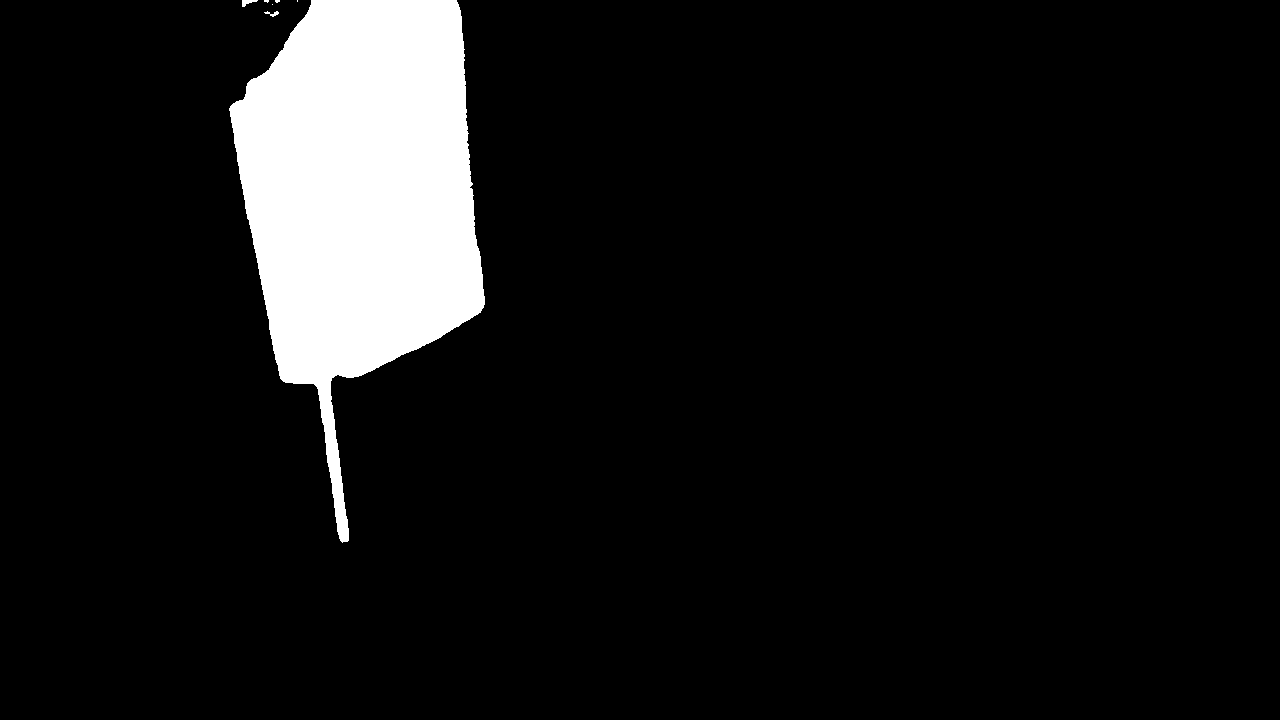}
	\end{subfigure}

	\caption{Qualitative comparison of the predicted segmentations using mask prompts on the VMD dataset. The first images of every two rows represents the rgb and groud truth image of the 1st frame, while the other images shown in the even row and in the odd row are the ground truth and predicted shadow masks for the 11th, 21th, 31th, 41th, 51th, , respectively. Best viewed on screen.}
	\label{fig:fig_vmd_mask}
	
\end{figure*}

\begin{figure*}[!ht]
	\centering
	\vspace*{1.3mm}
	\begin{subfigure}{0.12\textwidth}
		\includegraphics[width=\textwidth]{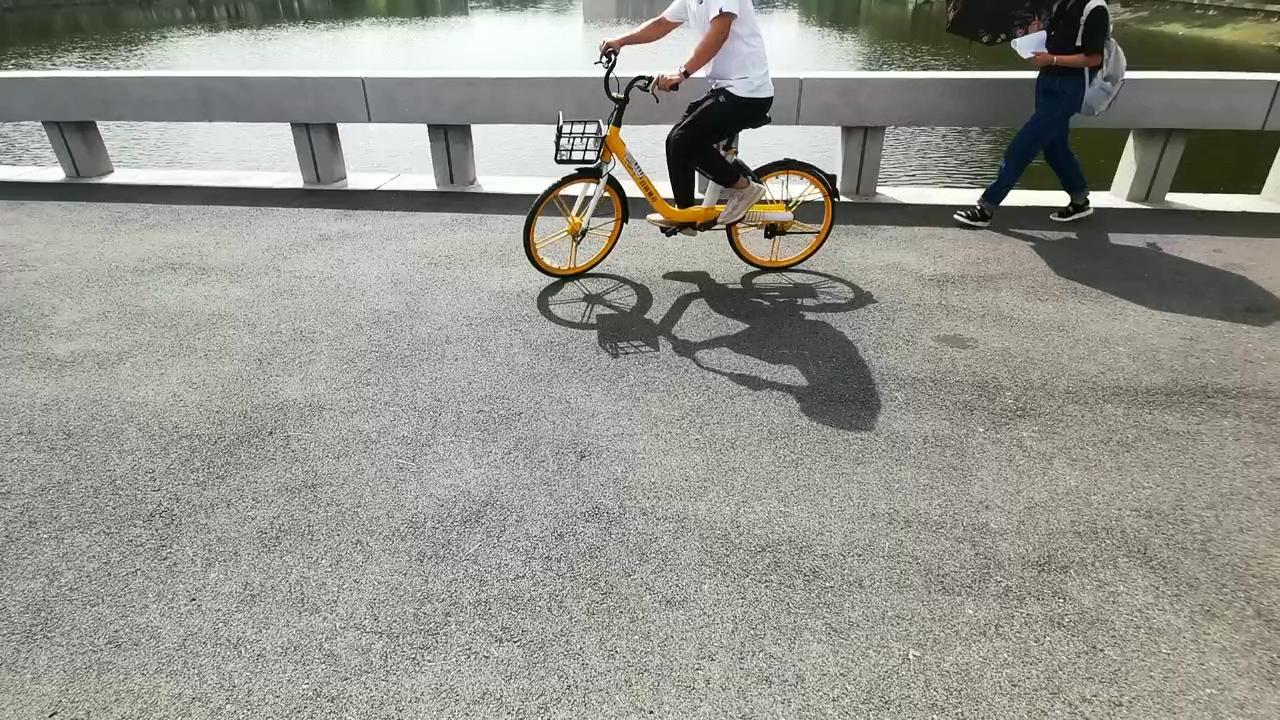}
	\end{subfigure}
	\begin{subfigure}{0.12\textwidth}
		\includegraphics[width=\textwidth]{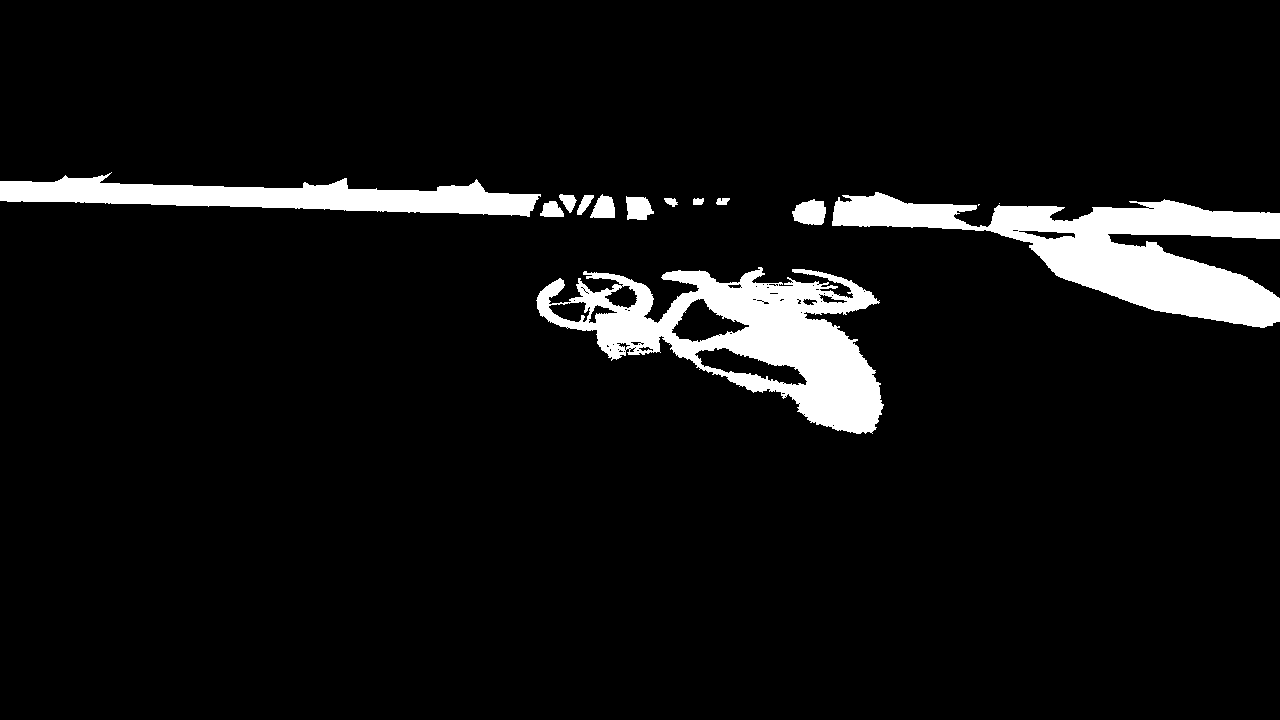}
	\end{subfigure}
	\begin{subfigure}{0.12\textwidth}
		\includegraphics[width=\textwidth]{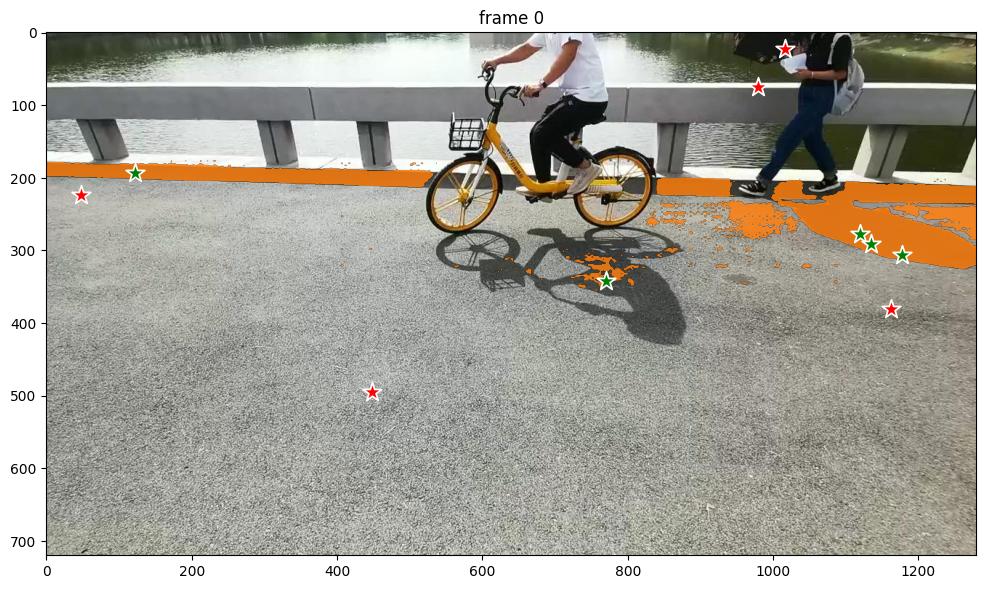}
	\end{subfigure}
	\begin{subfigure}{0.12\textwidth}
		\includegraphics[width=\textwidth]{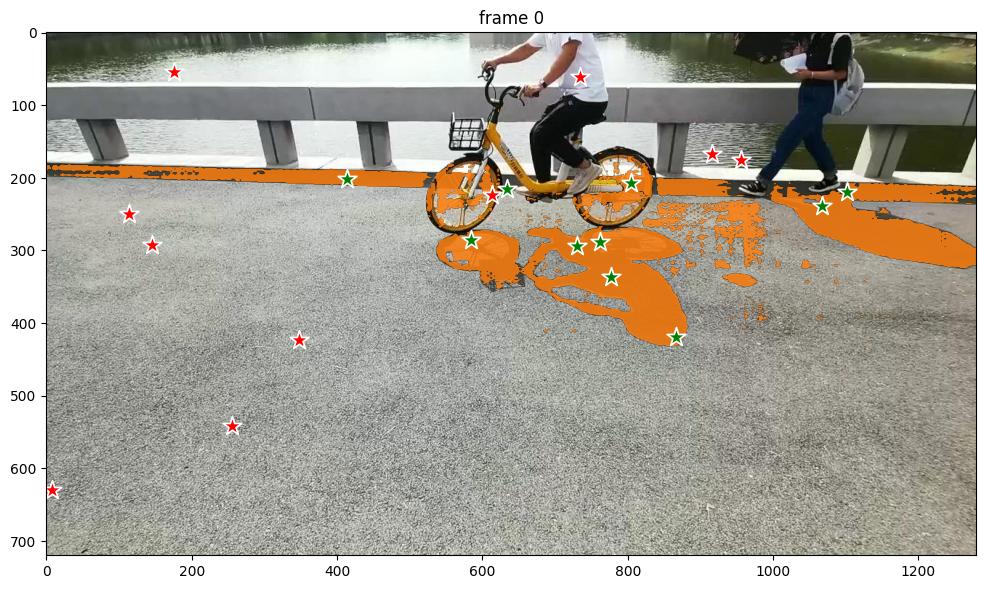}
	\end{subfigure}
	\begin{subfigure}{0.12\textwidth}
		\includegraphics[width=\textwidth]{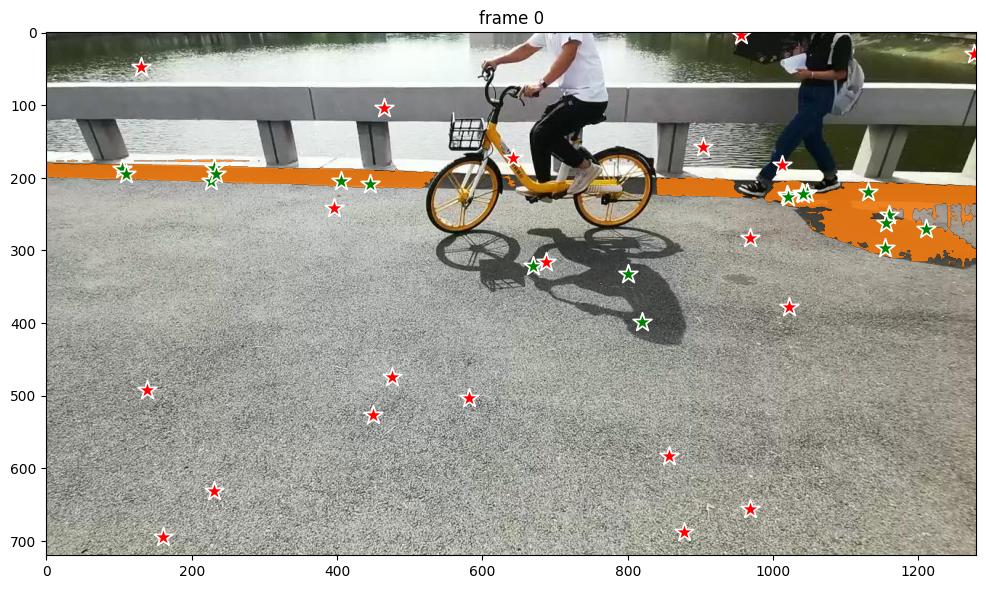}
	\end{subfigure}
	\begin{subfigure}{0.12\textwidth}
		\includegraphics[width=\textwidth]{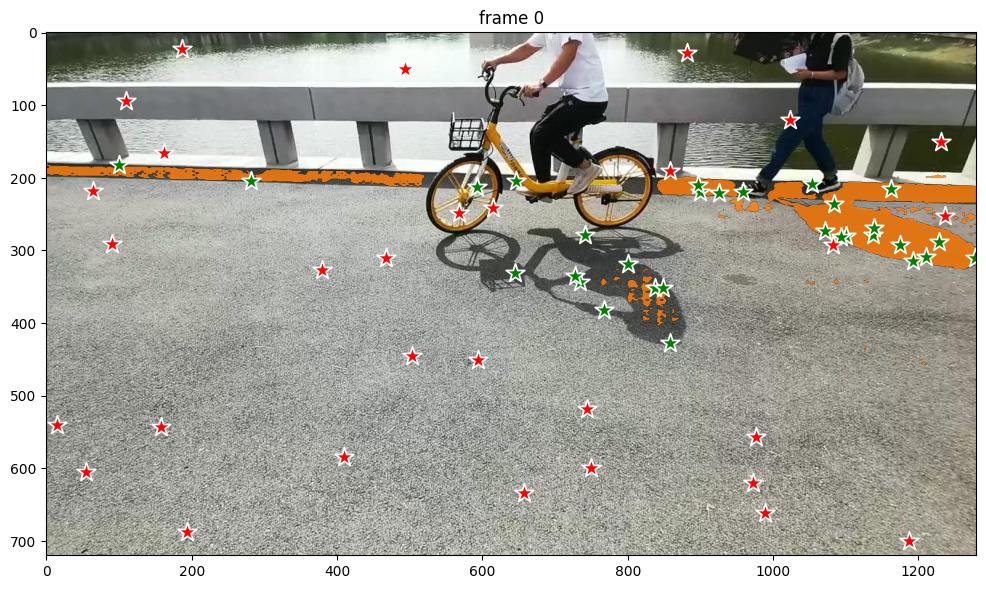}
	\end{subfigure}
	\begin{subfigure}{0.12\textwidth}
		\includegraphics[width=\textwidth]{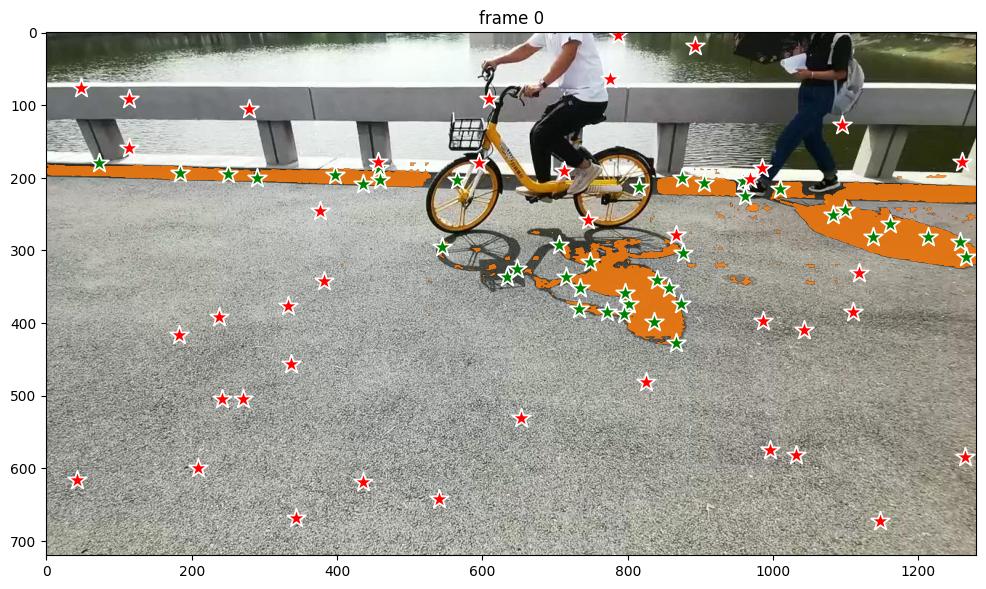}
	\end{subfigure}
	\begin{subfigure}{0.12\textwidth}
		\includegraphics[width=\textwidth]{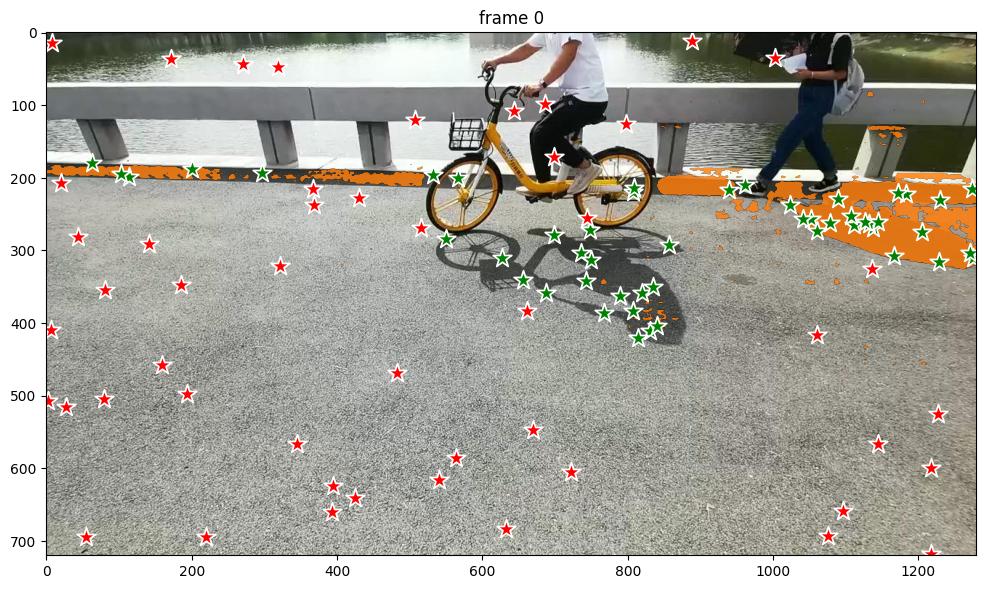}
	\end{subfigure}

	\vspace*{1.3mm}
	\begin{subfigure}{0.12\textwidth}
		\includegraphics[width=\textwidth]{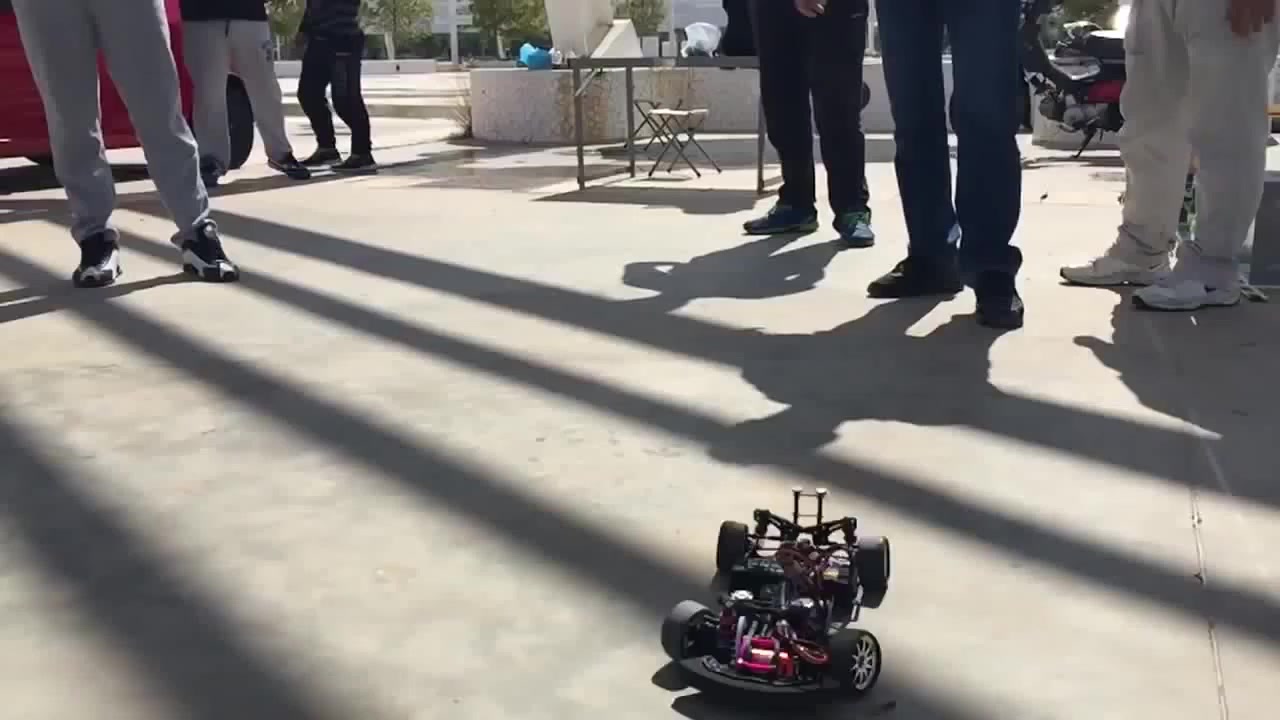}
	\end{subfigure}
	\begin{subfigure}{0.12\textwidth}
		\includegraphics[width=\textwidth]{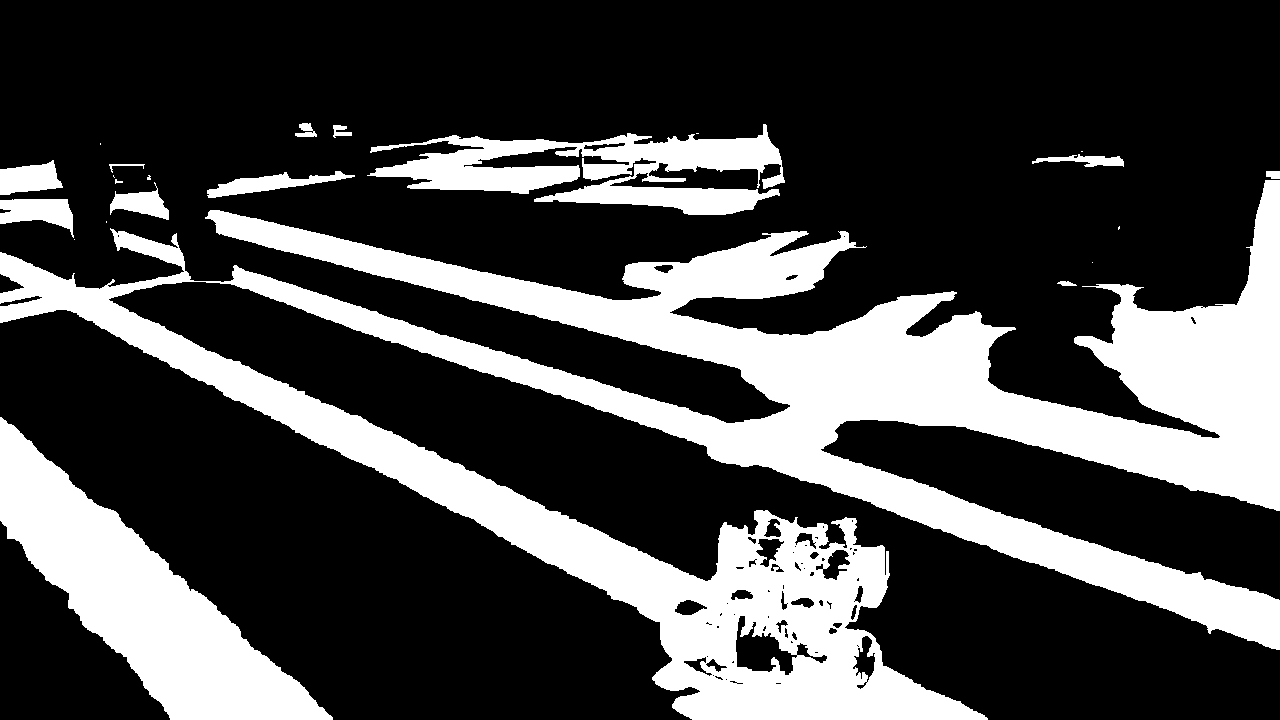}
	\end{subfigure}
	\begin{subfigure}{0.12\textwidth}
		\includegraphics[width=\textwidth]{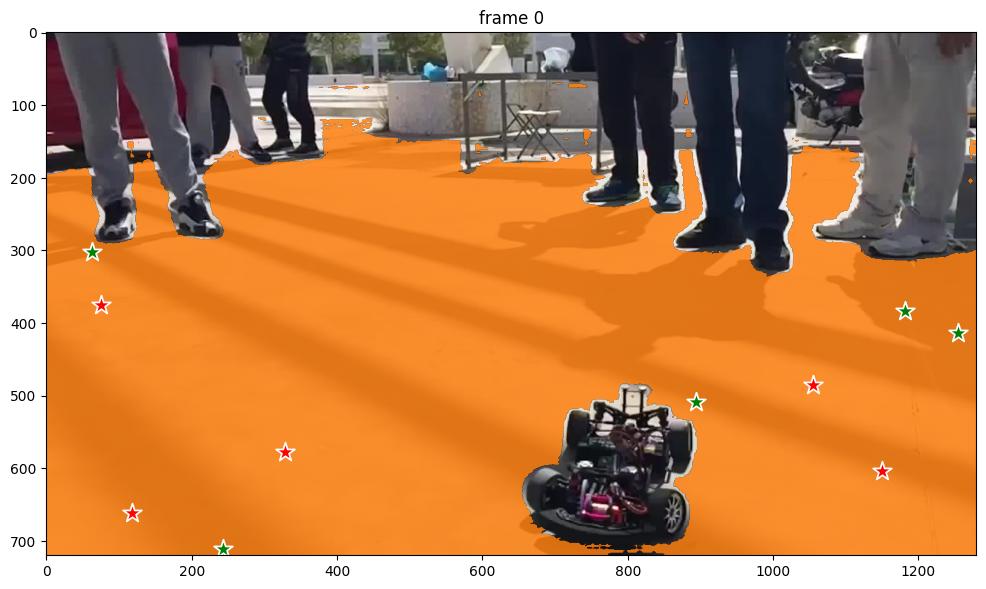}
	\end{subfigure}
	\begin{subfigure}{0.12\textwidth}
		\includegraphics[width=\textwidth]{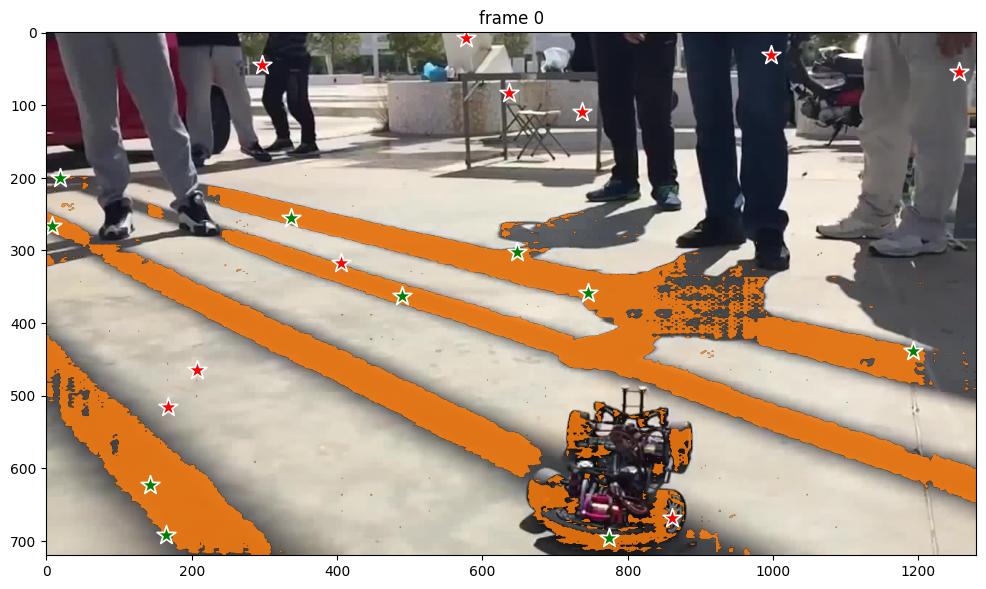}
	\end{subfigure}
	\begin{subfigure}{0.12\textwidth}
		\includegraphics[width=\textwidth]{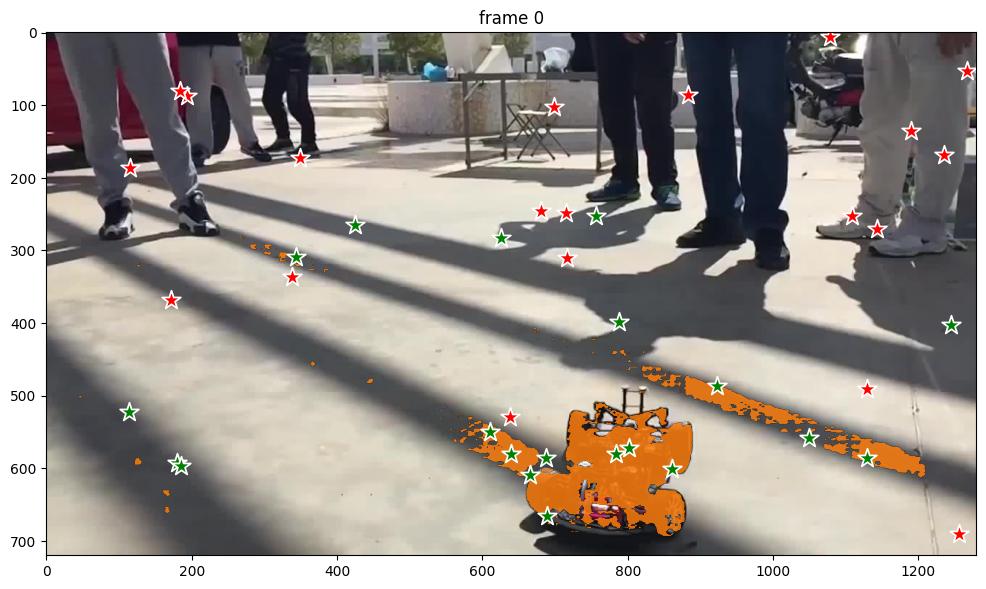}
	\end{subfigure}
	\begin{subfigure}{0.12\textwidth}
		\includegraphics[width=\textwidth]{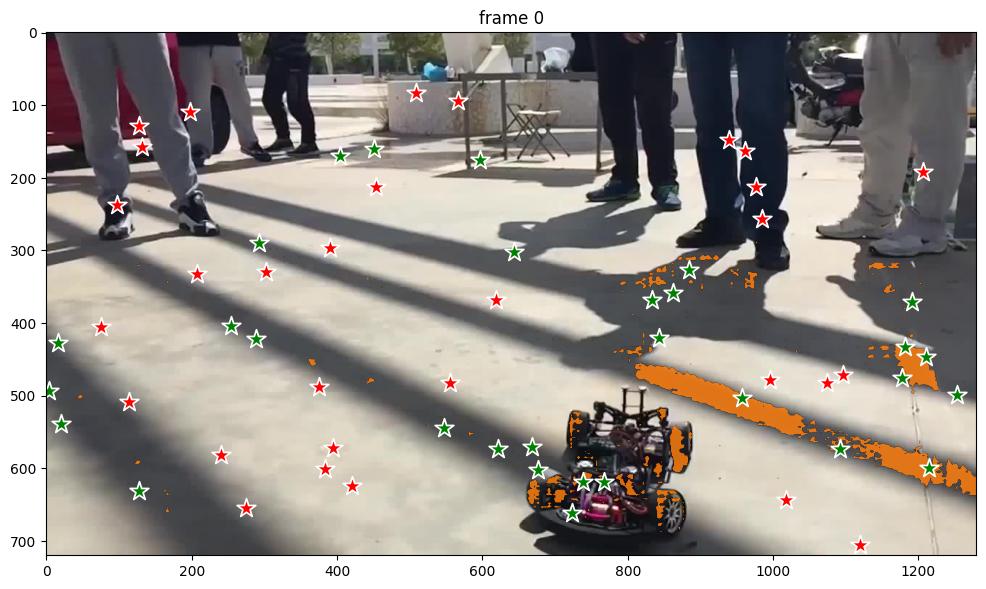}
	\end{subfigure}
	\begin{subfigure}{0.12\textwidth}
		\includegraphics[width=\textwidth]{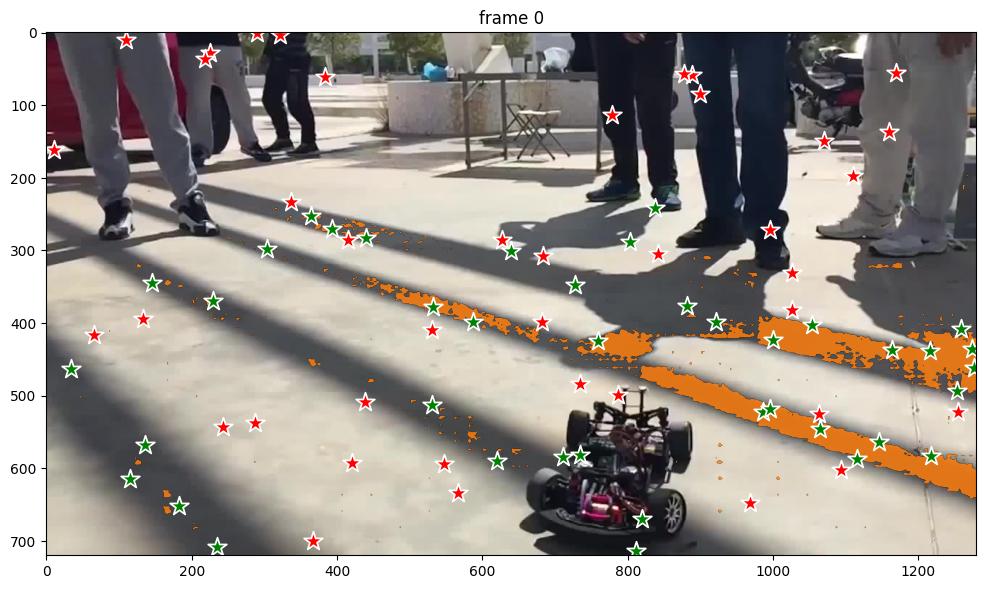}
	\end{subfigure}
	\begin{subfigure}{0.12\textwidth}
		\includegraphics[width=\textwidth]{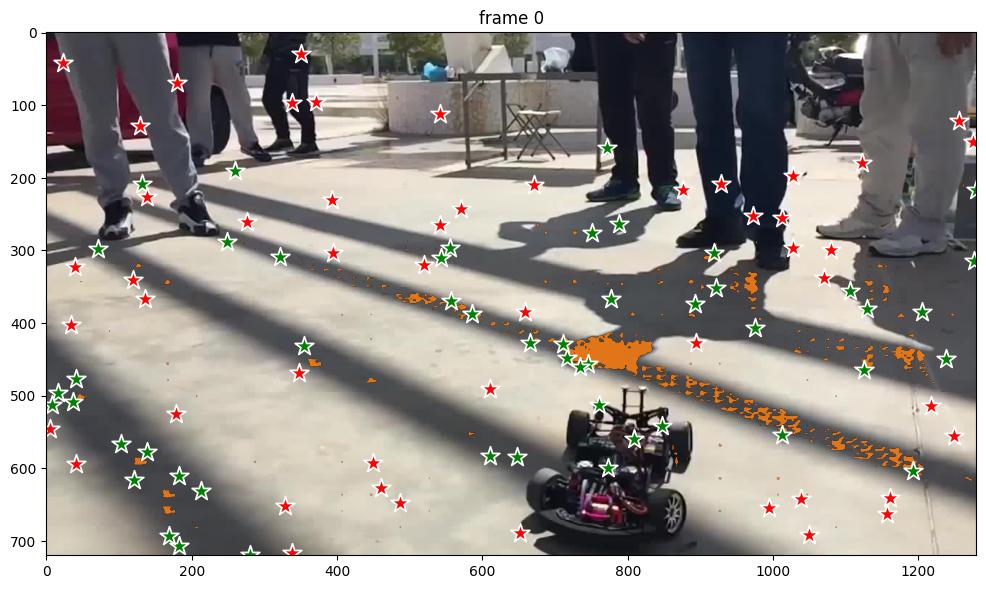}
	\end{subfigure}

	\vspace*{1.3mm}
	\begin{subfigure}{0.12\textwidth}
		\includegraphics[width=\textwidth]{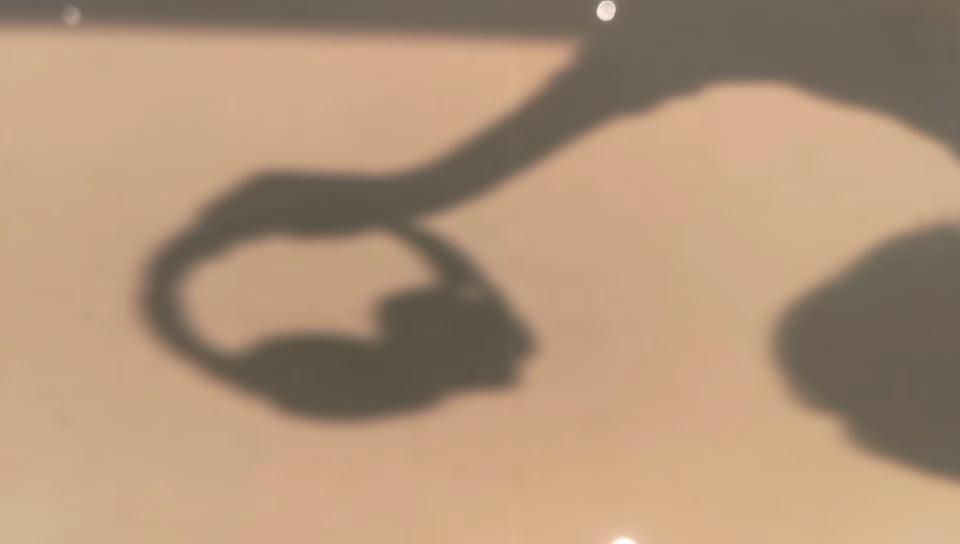}
	\end{subfigure}
	\begin{subfigure}{0.12\textwidth}
		\includegraphics[width=\textwidth]{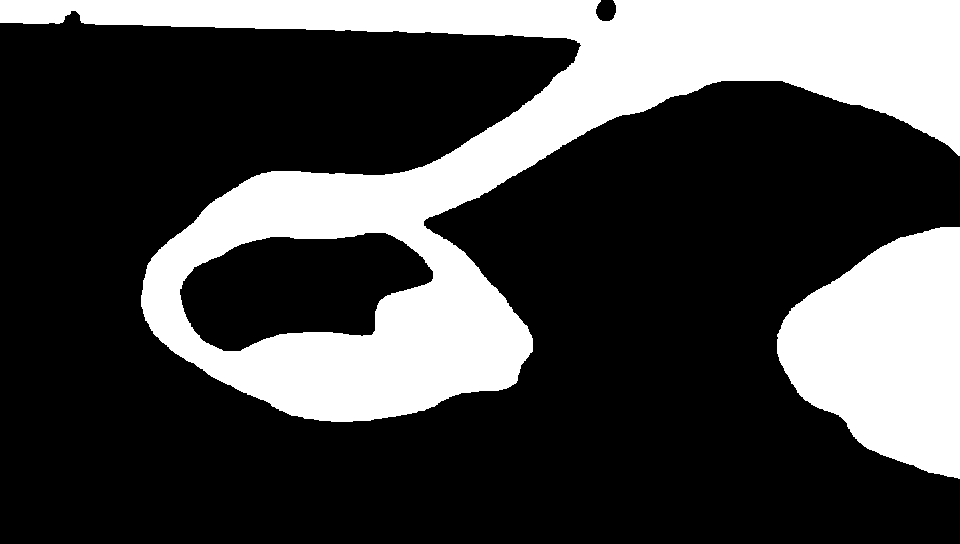}
	\end{subfigure}
	\begin{subfigure}{0.12\textwidth}
		\includegraphics[width=\textwidth]{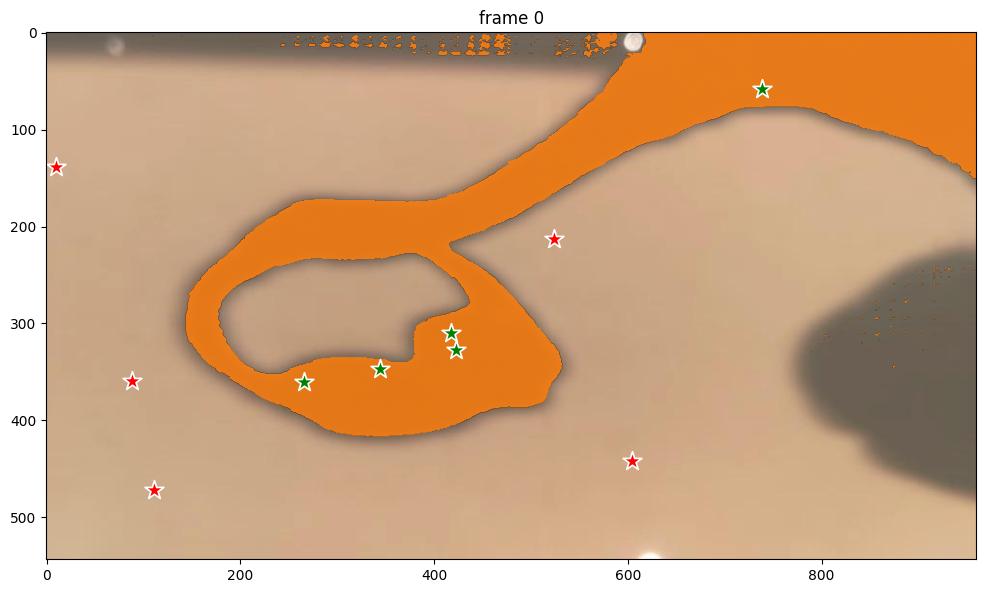}
	\end{subfigure}
	\begin{subfigure}{0.12\textwidth}
		\includegraphics[width=\textwidth]{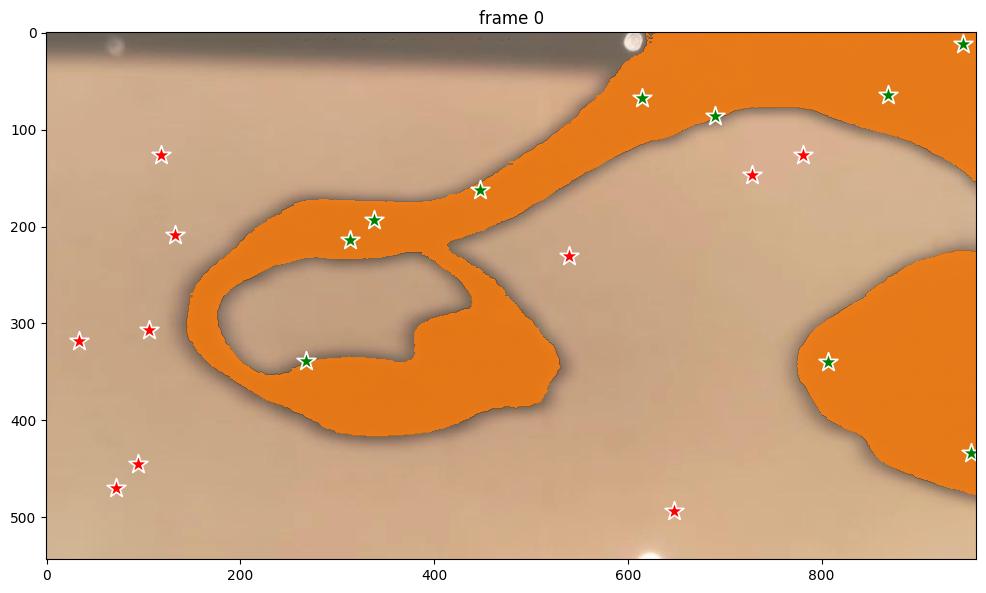}
	\end{subfigure}
	\begin{subfigure}{0.12\textwidth}
		\includegraphics[width=\textwidth]{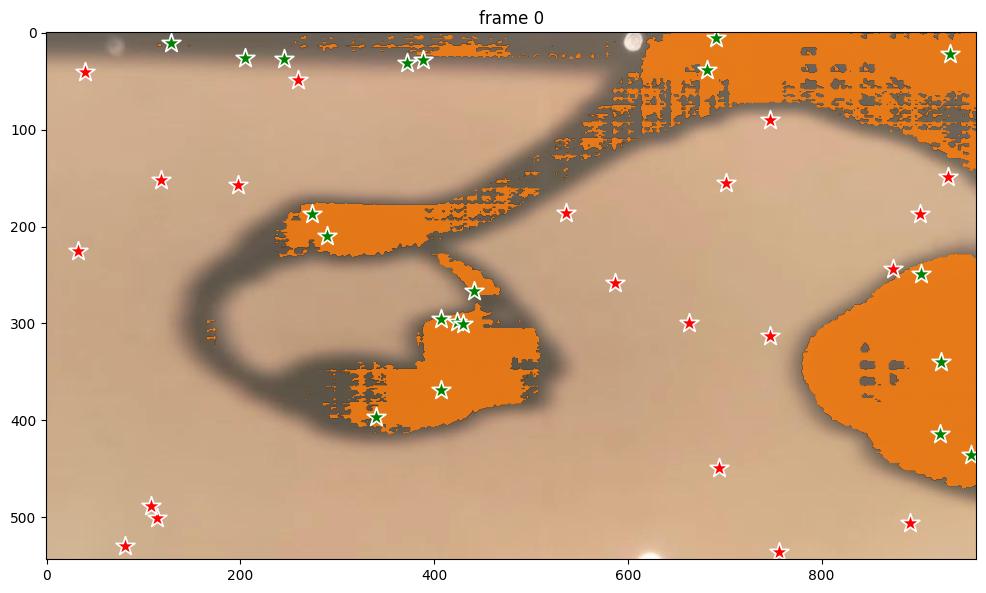}
	\end{subfigure}
	\begin{subfigure}{0.12\textwidth}
		\includegraphics[width=\textwidth]{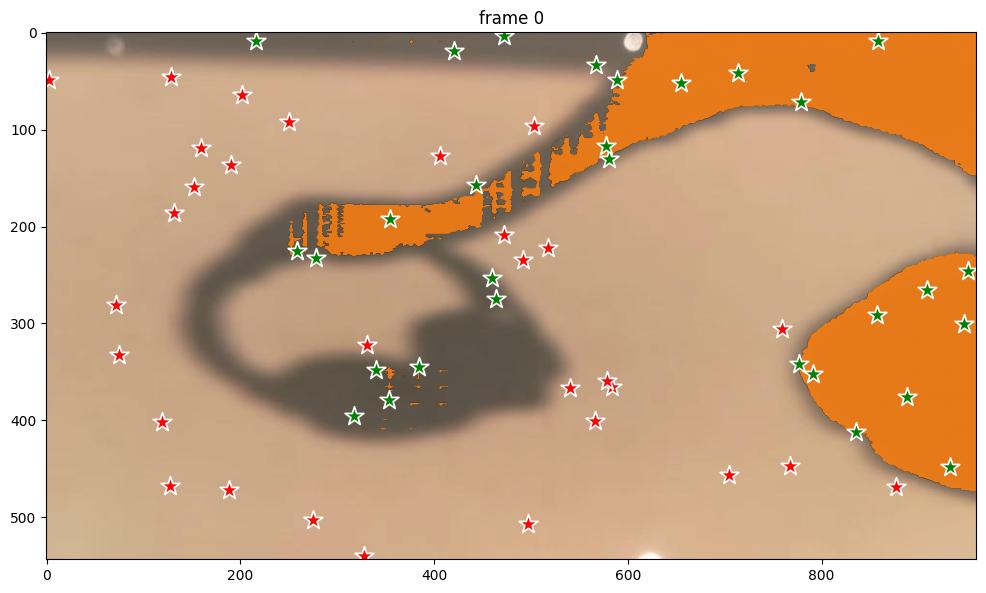}
	\end{subfigure}
	\begin{subfigure}{0.12\textwidth}
		\includegraphics[width=\textwidth]{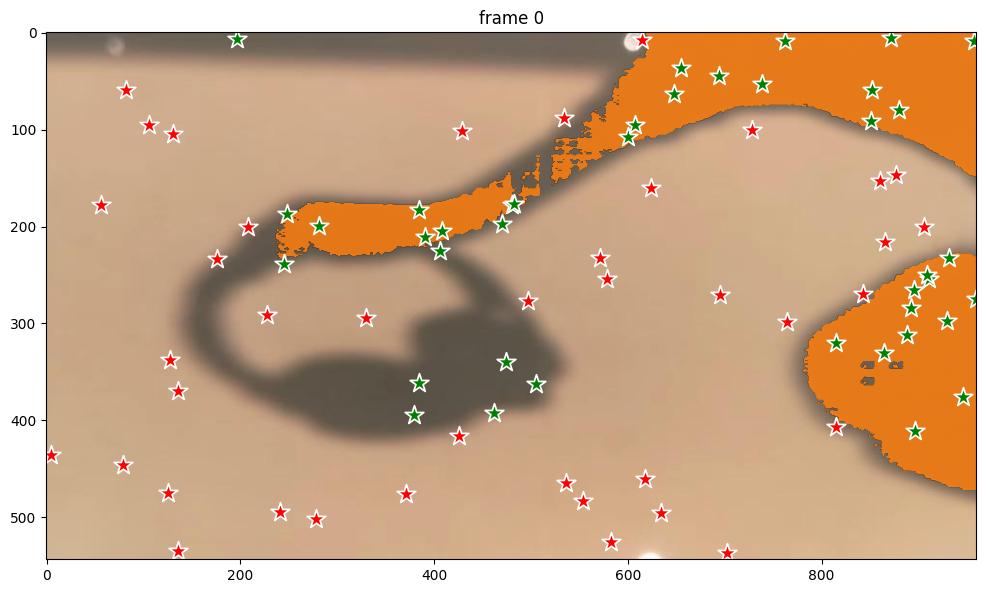}
	\end{subfigure}
	\begin{subfigure}{0.12\textwidth}
		\includegraphics[width=\textwidth]{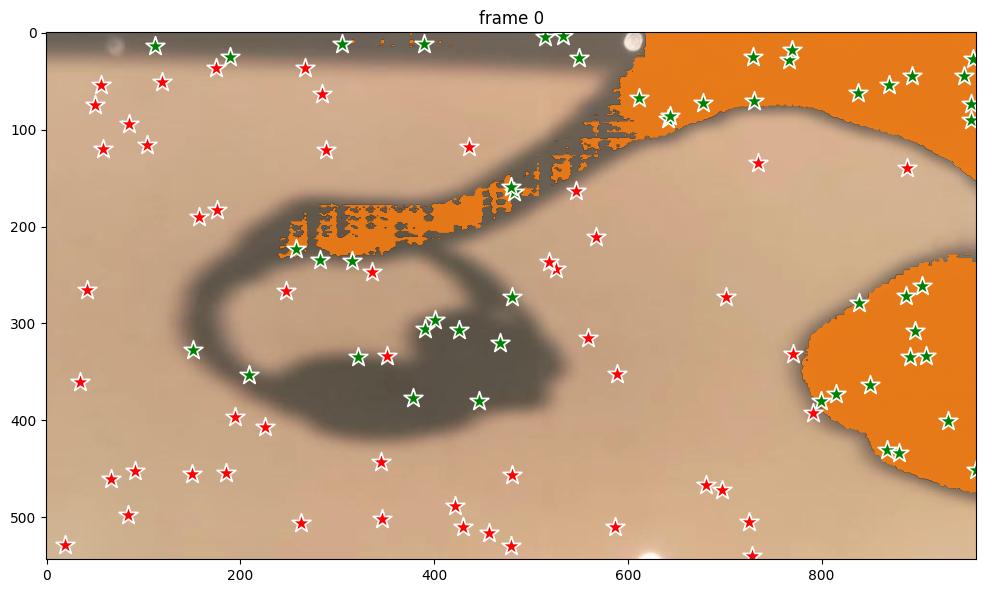}
	\end{subfigure}

	\vspace*{1.3mm}
	\begin{subfigure}{0.12\textwidth}
		\includegraphics[width=\textwidth]{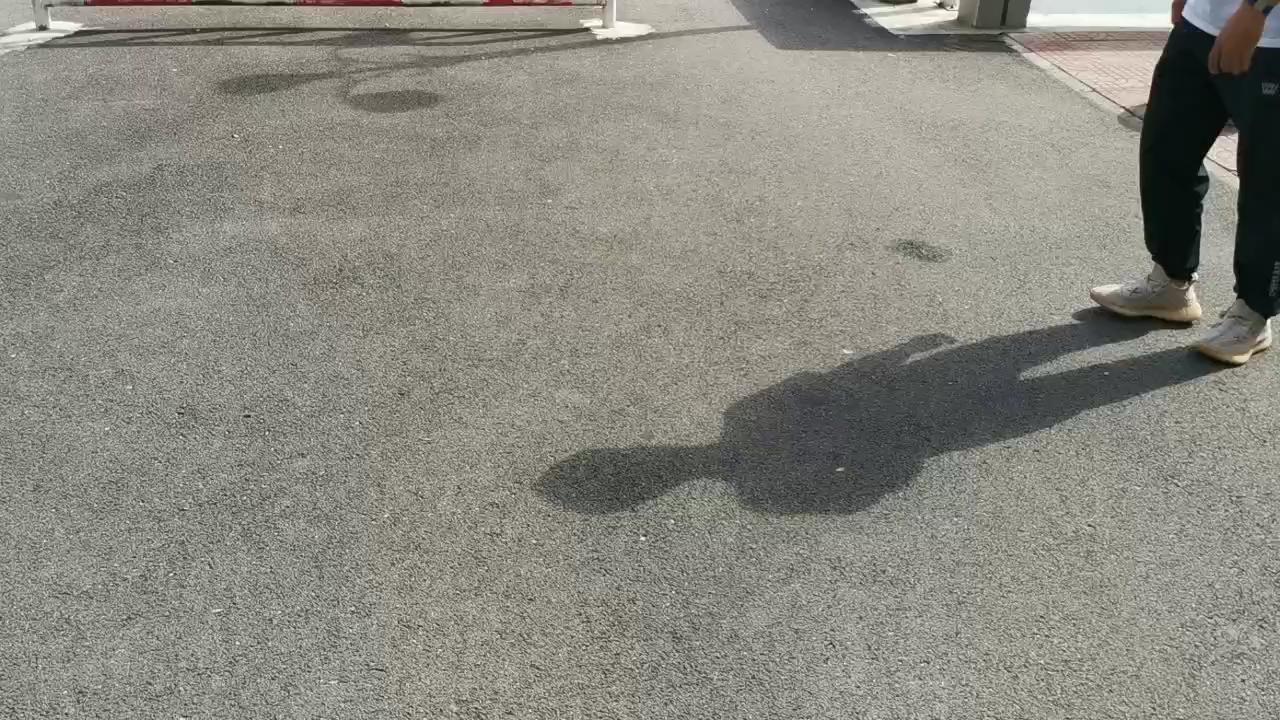}
	\end{subfigure}
	\begin{subfigure}{0.12\textwidth}
		\includegraphics[width=\textwidth]{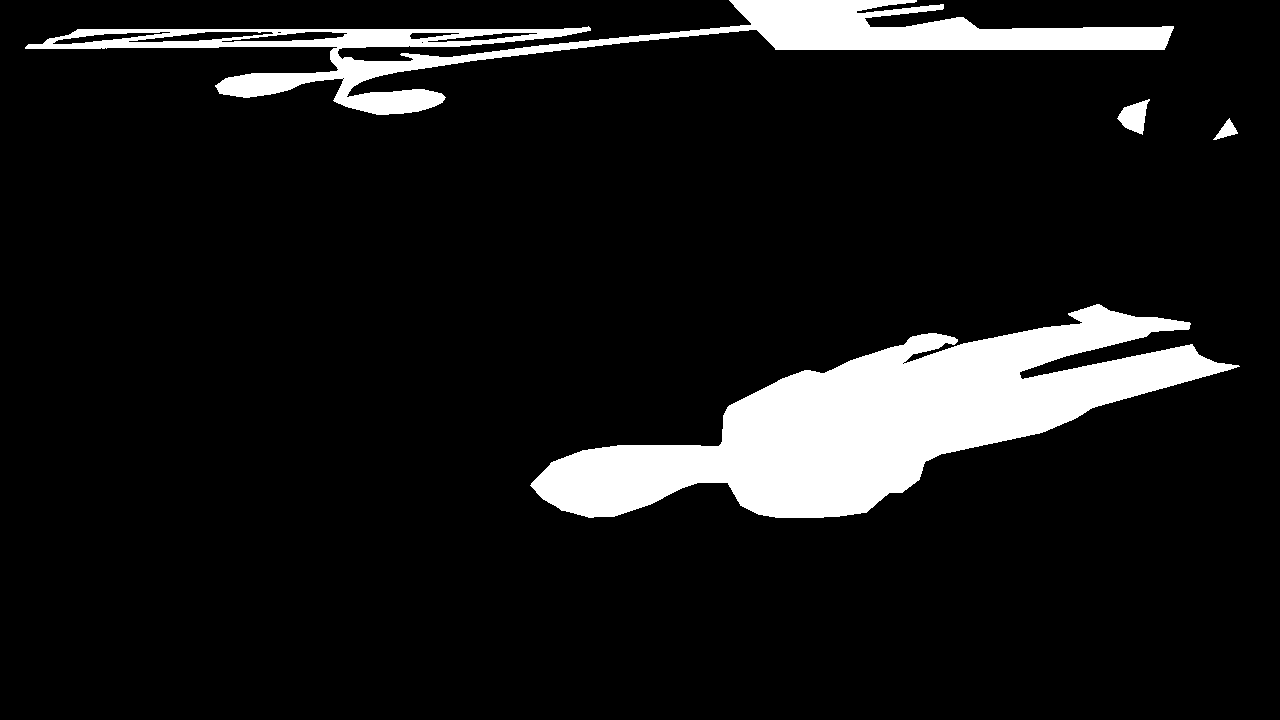}
	\end{subfigure}
	\begin{subfigure}{0.12\textwidth}
		\includegraphics[width=\textwidth]{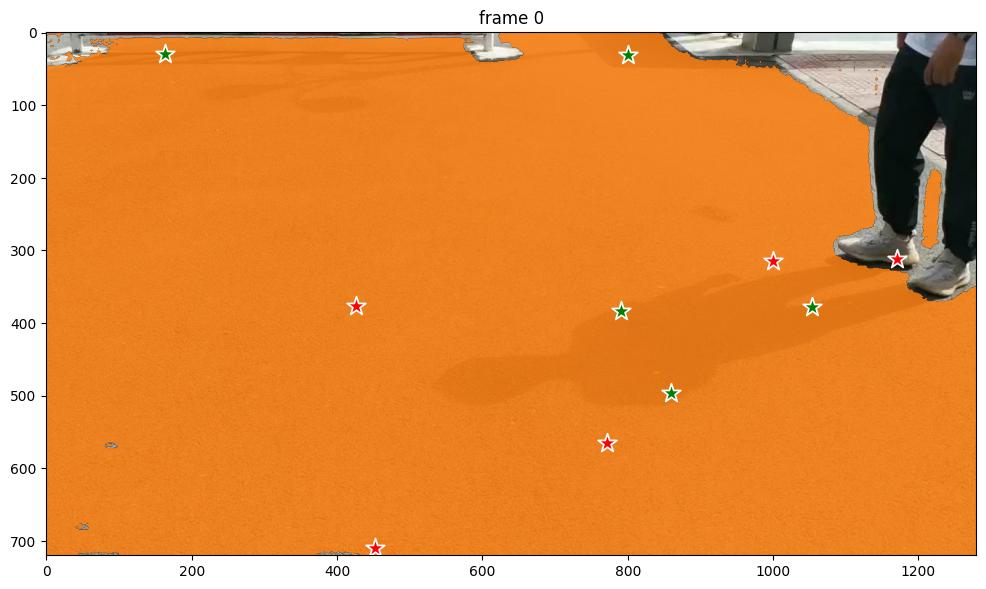}
	\end{subfigure}
	\begin{subfigure}{0.12\textwidth}
		\includegraphics[width=\textwidth]{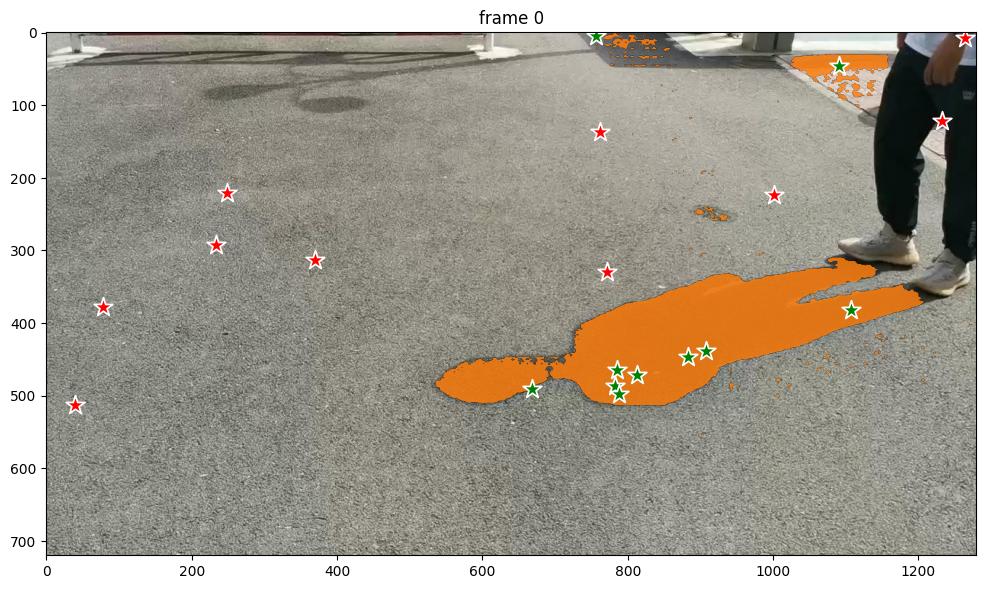}
	\end{subfigure}
	\begin{subfigure}{0.12\textwidth}
		\includegraphics[width=\textwidth]{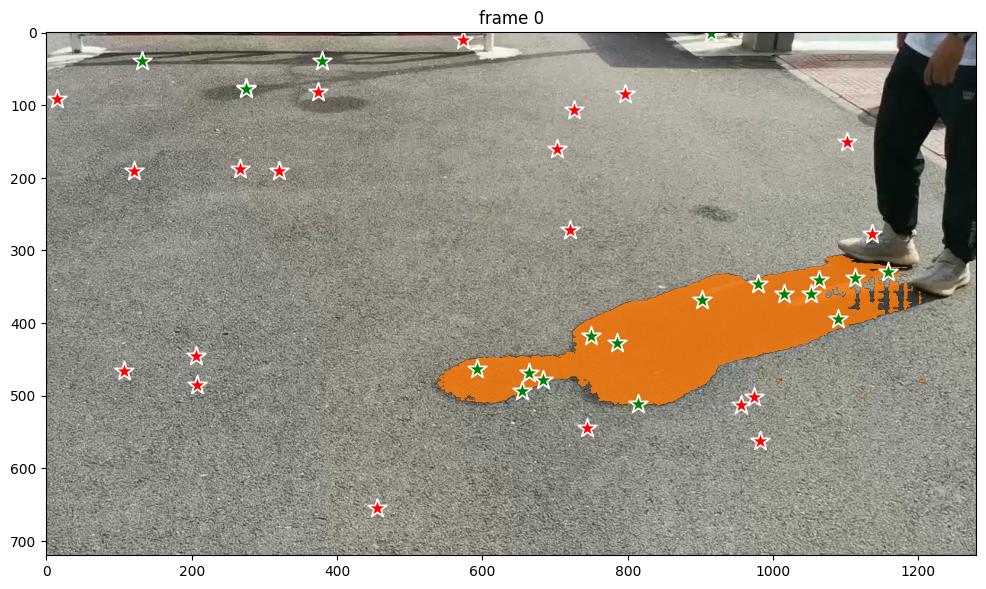}
	\end{subfigure}
	\begin{subfigure}{0.12\textwidth}
		\includegraphics[width=\textwidth]{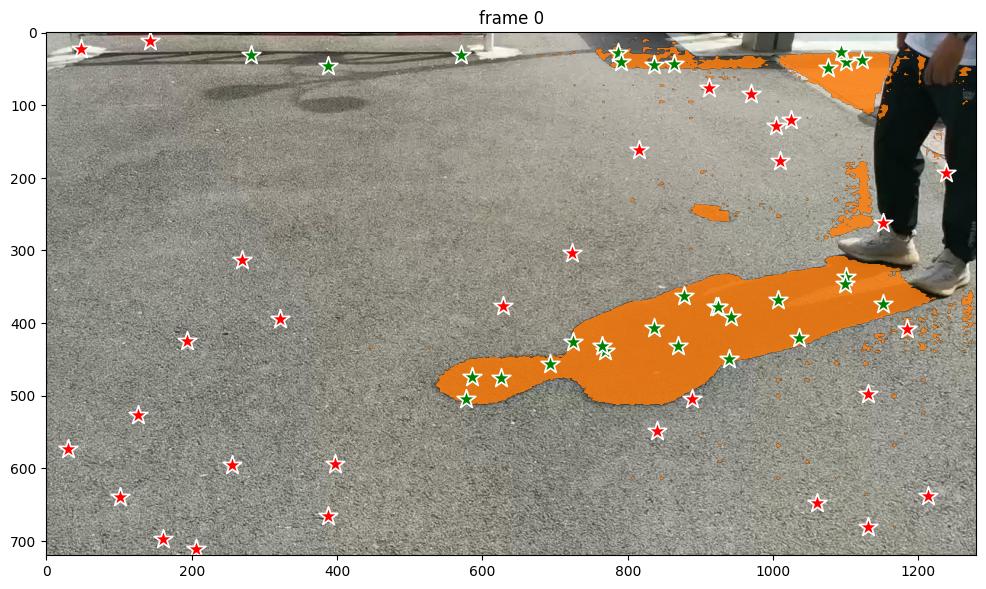}
	\end{subfigure}
	\begin{subfigure}{0.12\textwidth}
		\includegraphics[width=\textwidth]{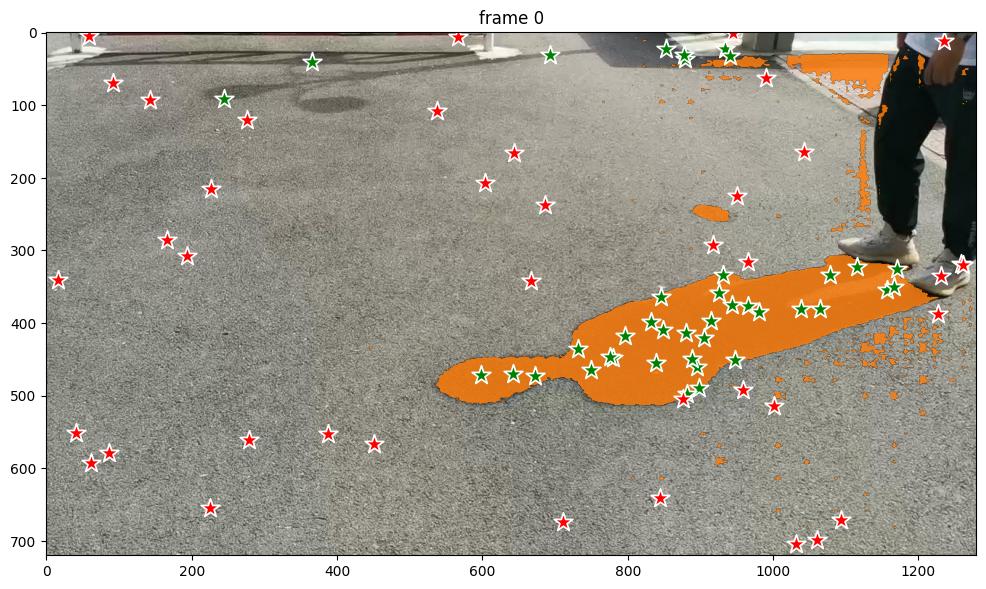}
	\end{subfigure}
	\begin{subfigure}{0.12\textwidth}
		\includegraphics[width=\textwidth]{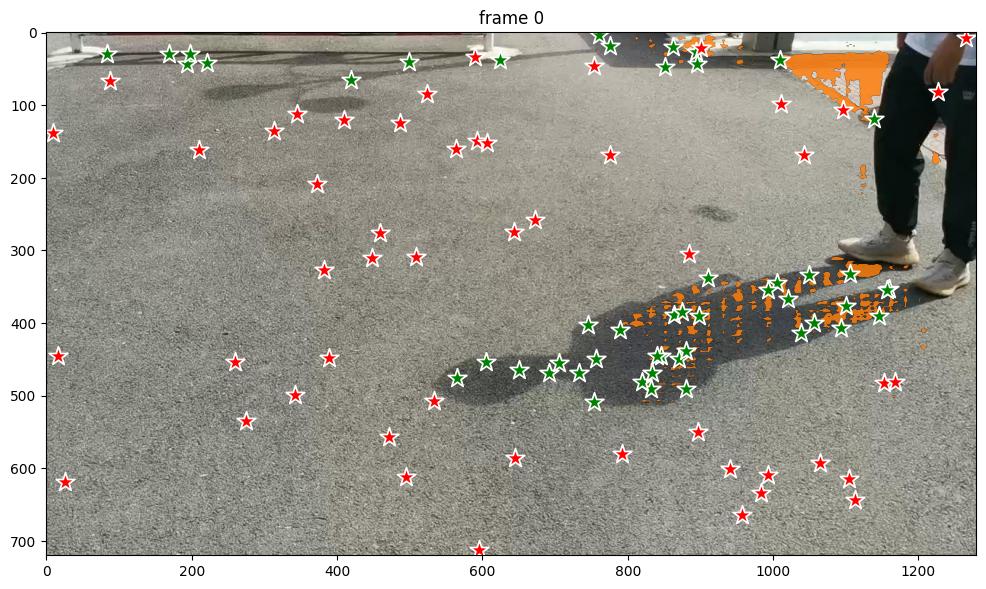}
	\end{subfigure}

	\vspace*{1.3mm}
	\begin{subfigure}{0.12\textwidth}
		\includegraphics[width=\textwidth]{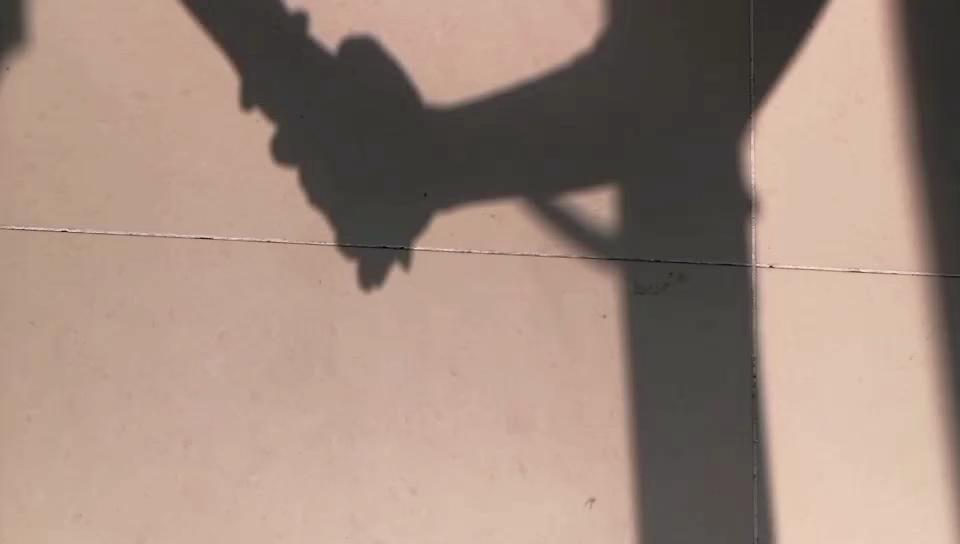}
	\end{subfigure}
	\begin{subfigure}{0.12\textwidth}
		\includegraphics[width=\textwidth]{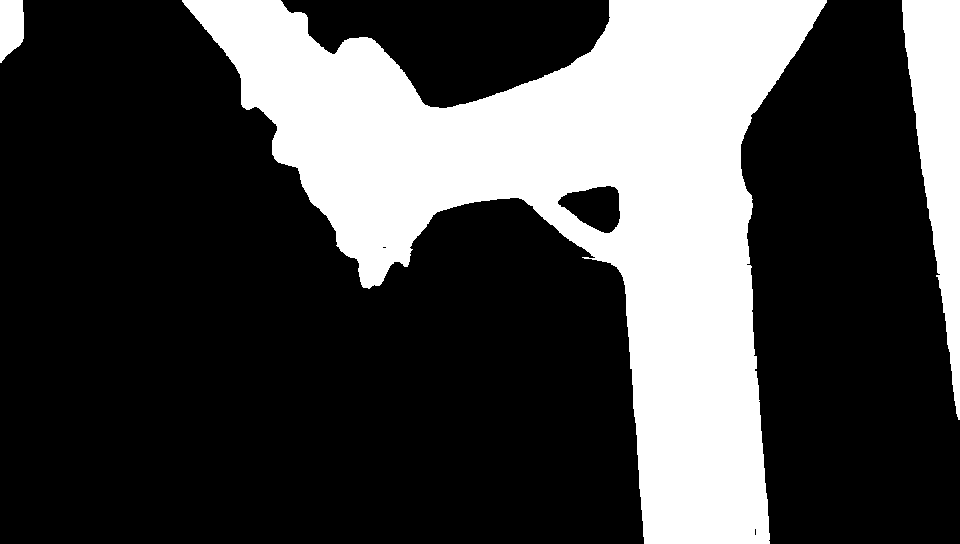}
	\end{subfigure}
	\begin{subfigure}{0.12\textwidth}
		\includegraphics[width=\textwidth]{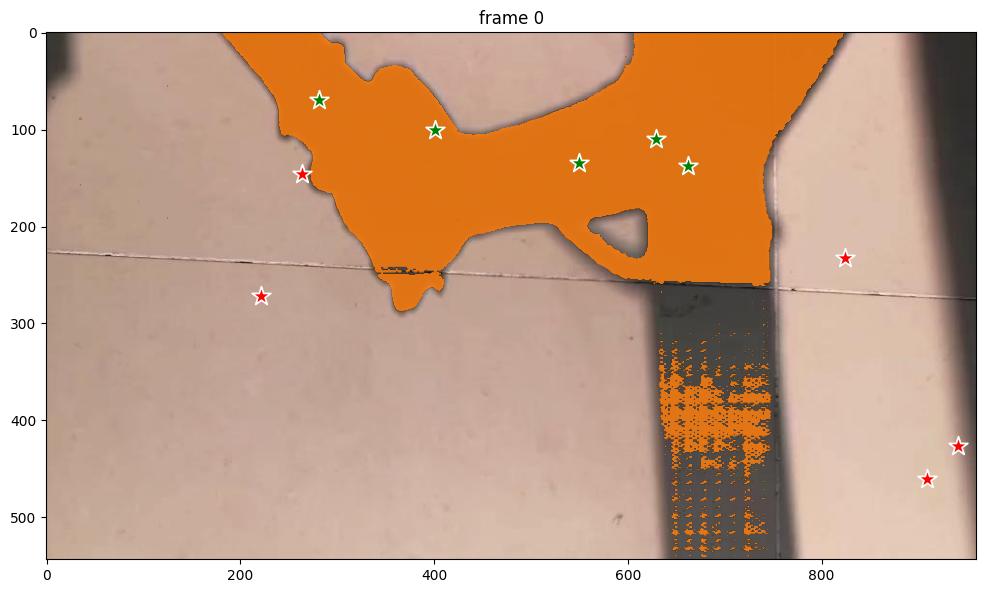}
	\end{subfigure}
	\begin{subfigure}{0.12\textwidth}
		\includegraphics[width=\textwidth]{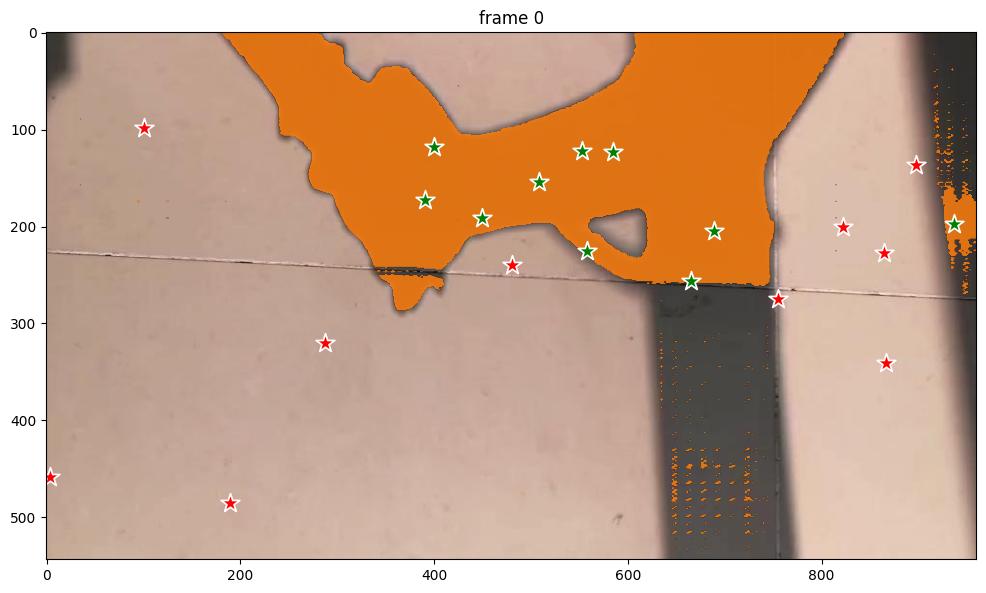}
	\end{subfigure}
	\begin{subfigure}{0.12\textwidth}
		\includegraphics[width=\textwidth]{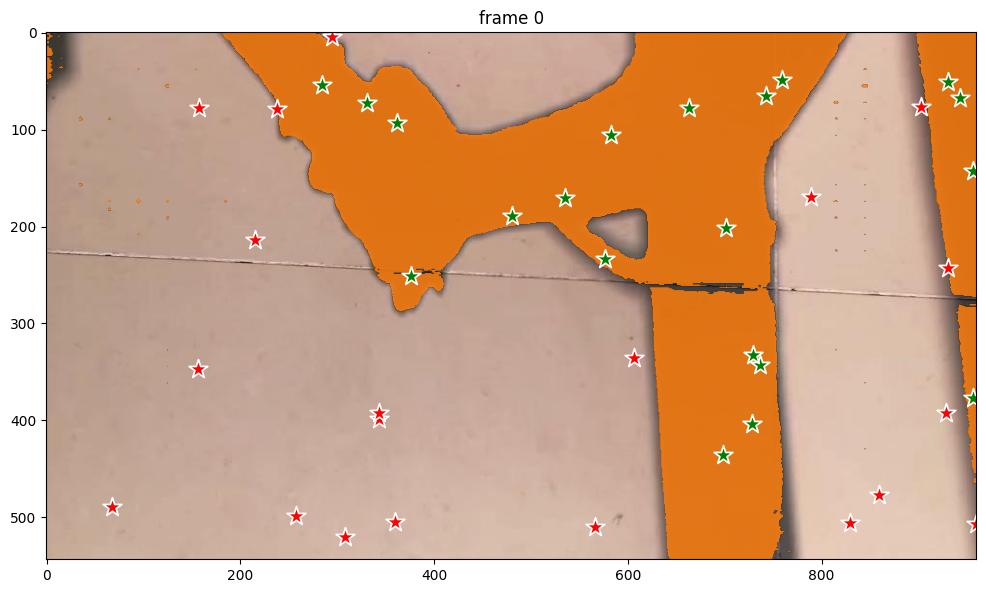}
	\end{subfigure}
	\begin{subfigure}{0.12\textwidth}
		\includegraphics[width=\textwidth]{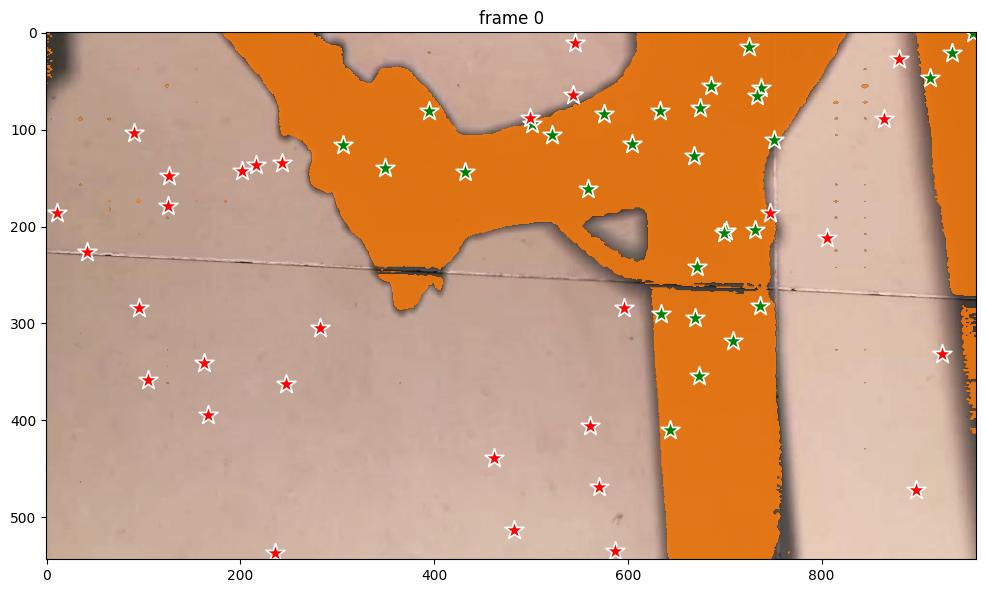}
	\end{subfigure}
	\begin{subfigure}{0.12\textwidth}
		\includegraphics[width=\textwidth]{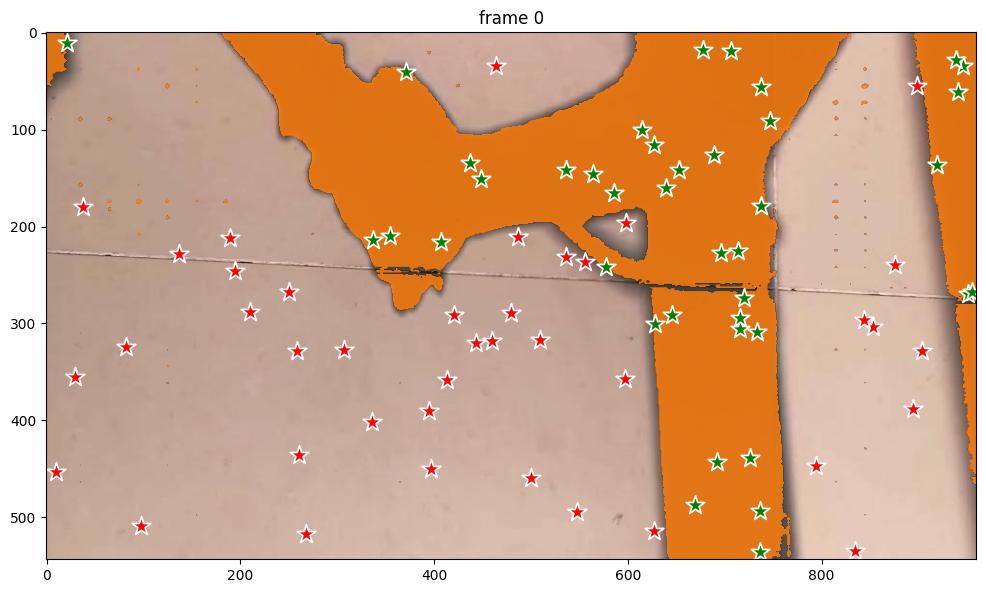}
	\end{subfigure}
	\begin{subfigure}{0.12\textwidth}
		\includegraphics[width=\textwidth]{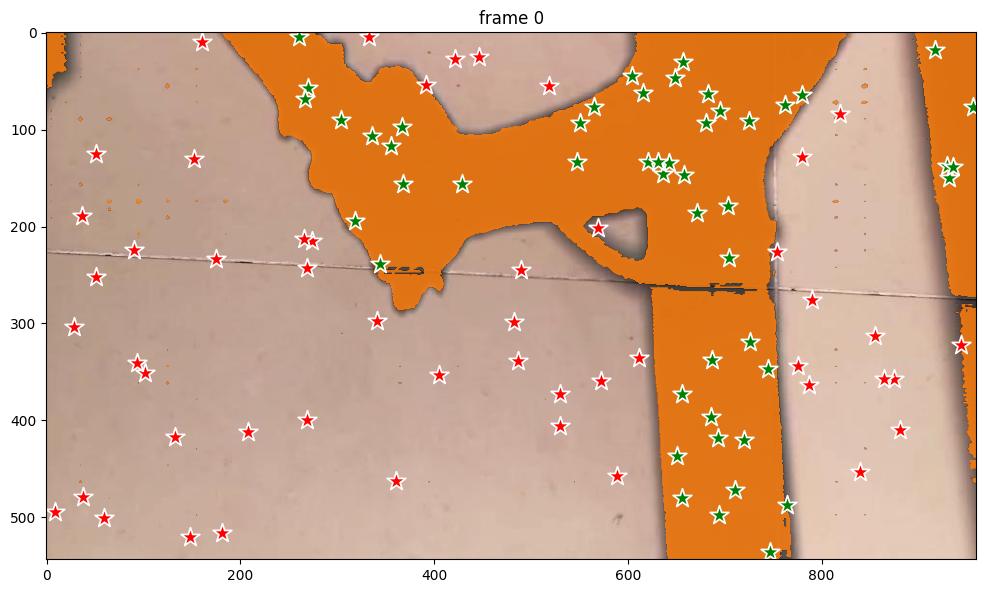}
	\end{subfigure}

	\vspace*{1.3mm}
	\begin{subfigure}{0.12\textwidth}
		\includegraphics[width=\textwidth]{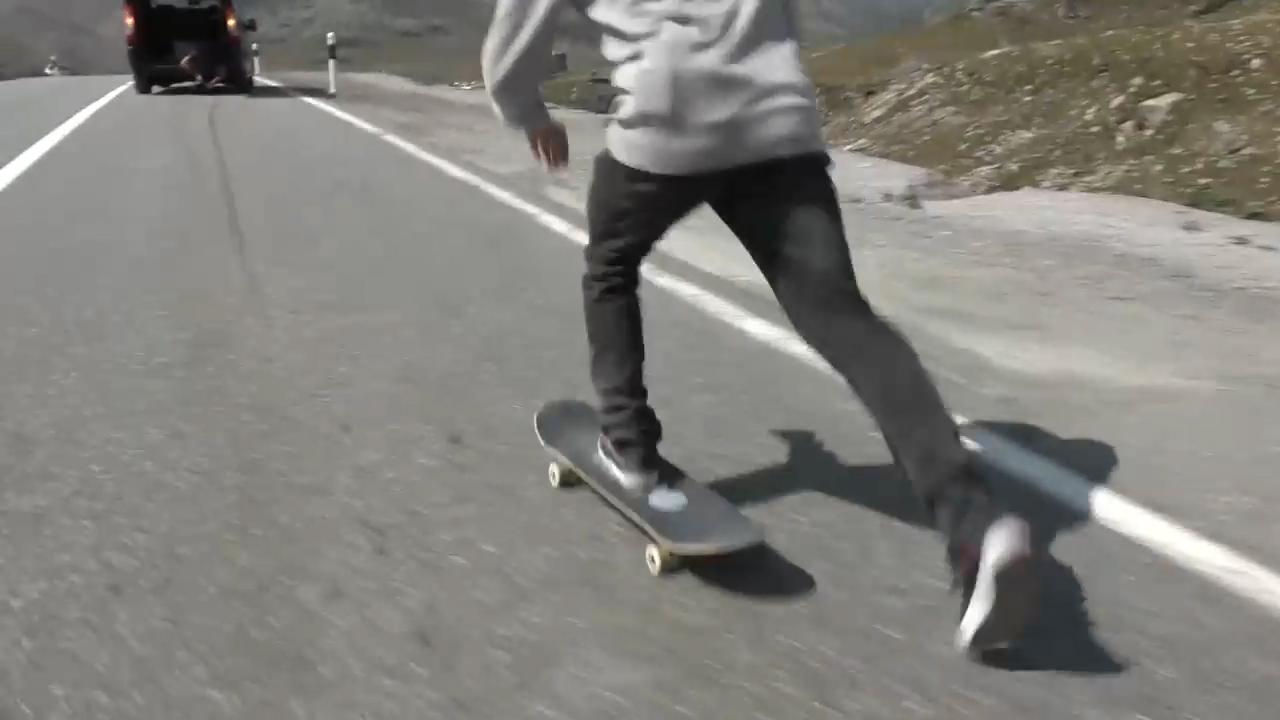}
	\end{subfigure}
	\begin{subfigure}{0.12\textwidth}
		\includegraphics[width=\textwidth]{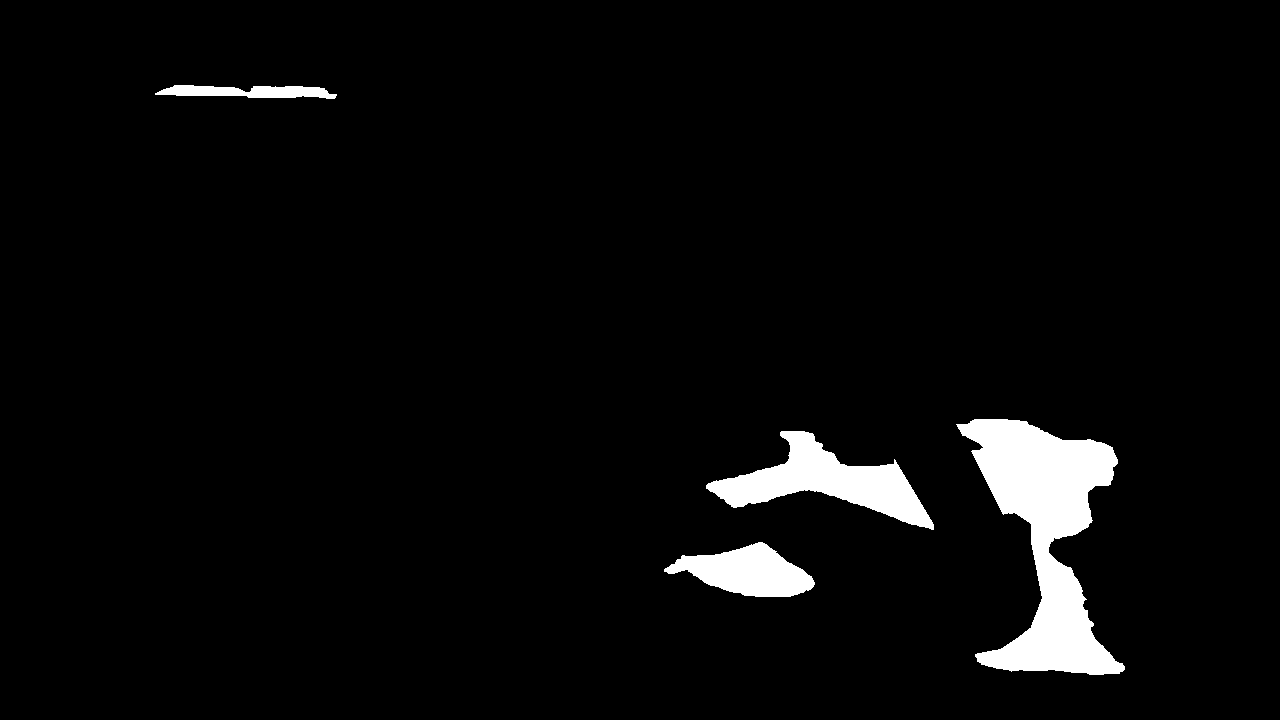}
	\end{subfigure}
	\begin{subfigure}{0.12\textwidth}
		\includegraphics[width=\textwidth]{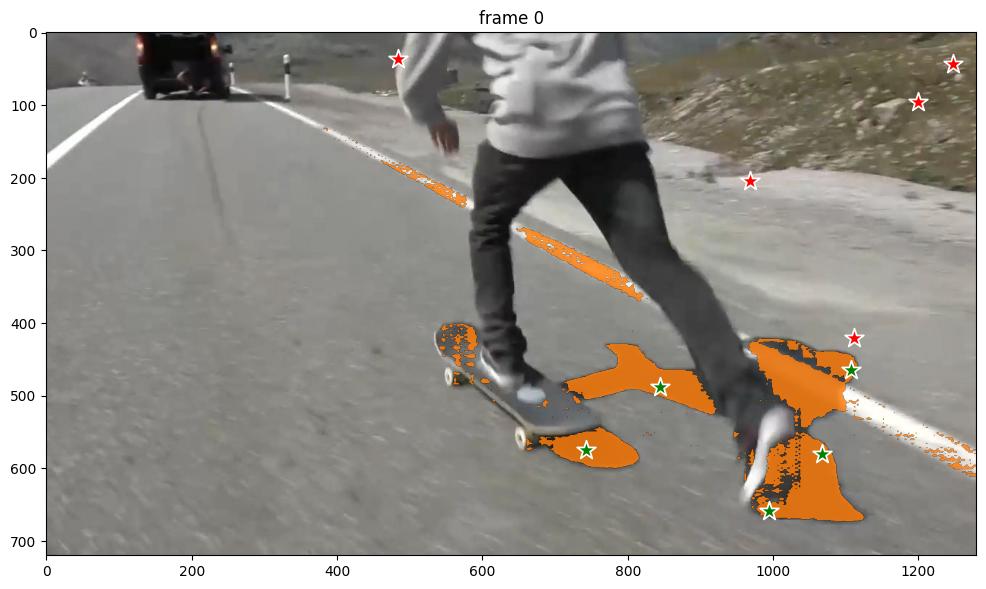}
	\end{subfigure}
	\begin{subfigure}{0.12\textwidth}
		\includegraphics[width=\textwidth]{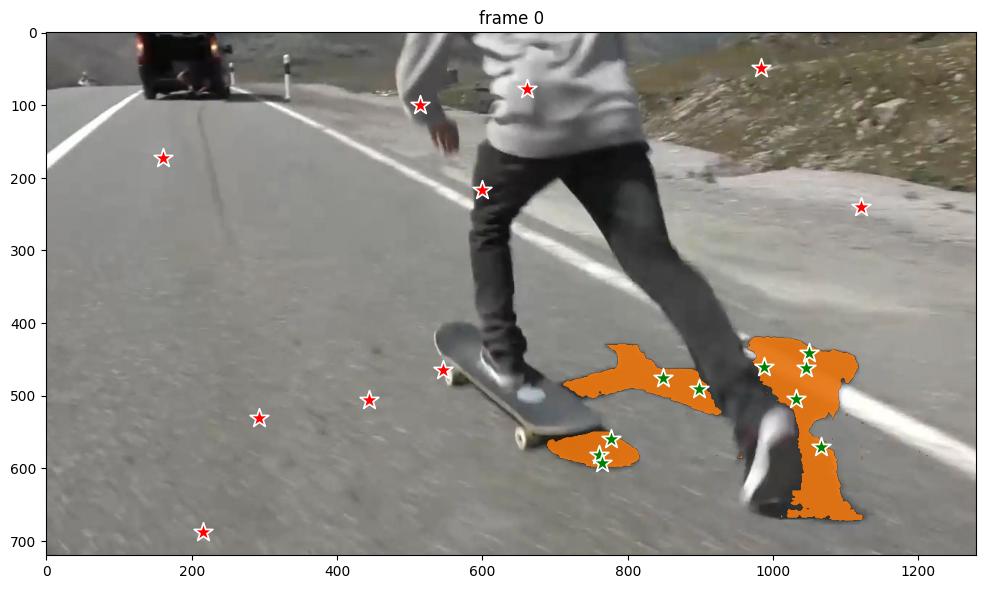}
	\end{subfigure}
	\begin{subfigure}{0.12\textwidth}
		\includegraphics[width=\textwidth]{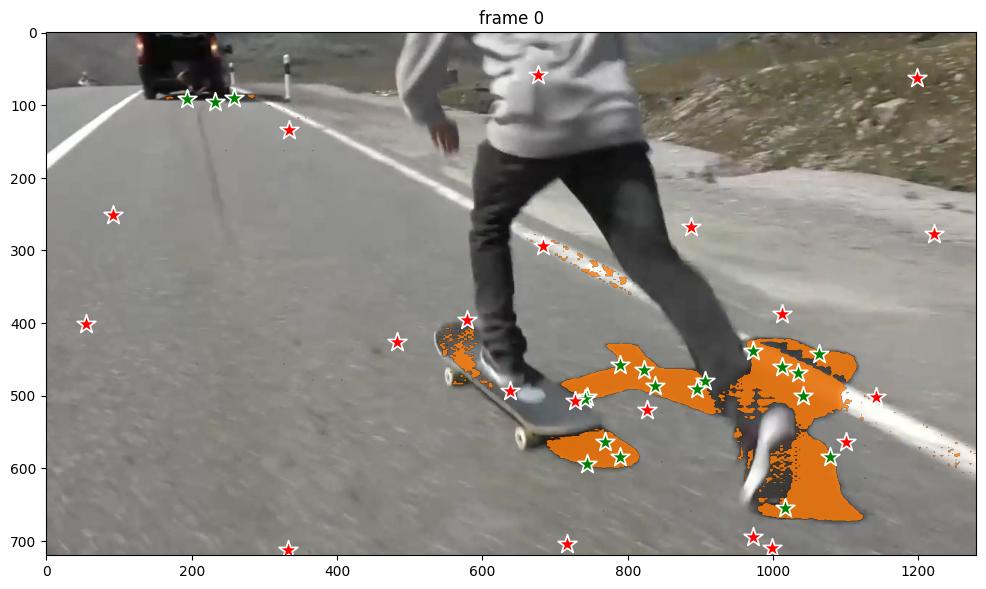}
	\end{subfigure}
	\begin{subfigure}{0.12\textwidth}
		\includegraphics[width=\textwidth]{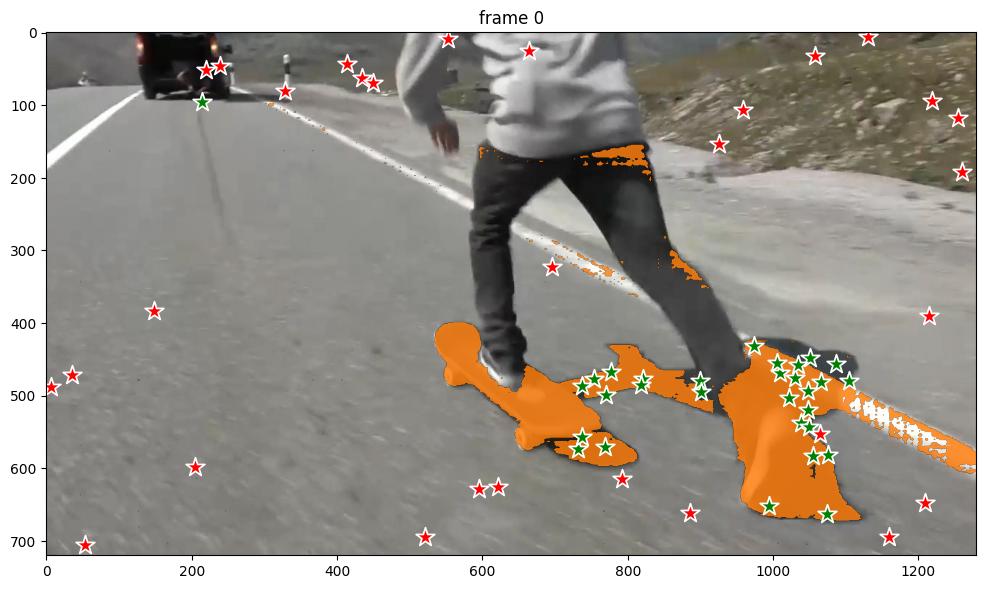}
	\end{subfigure}
	\begin{subfigure}{0.12\textwidth}
		\includegraphics[width=\textwidth]{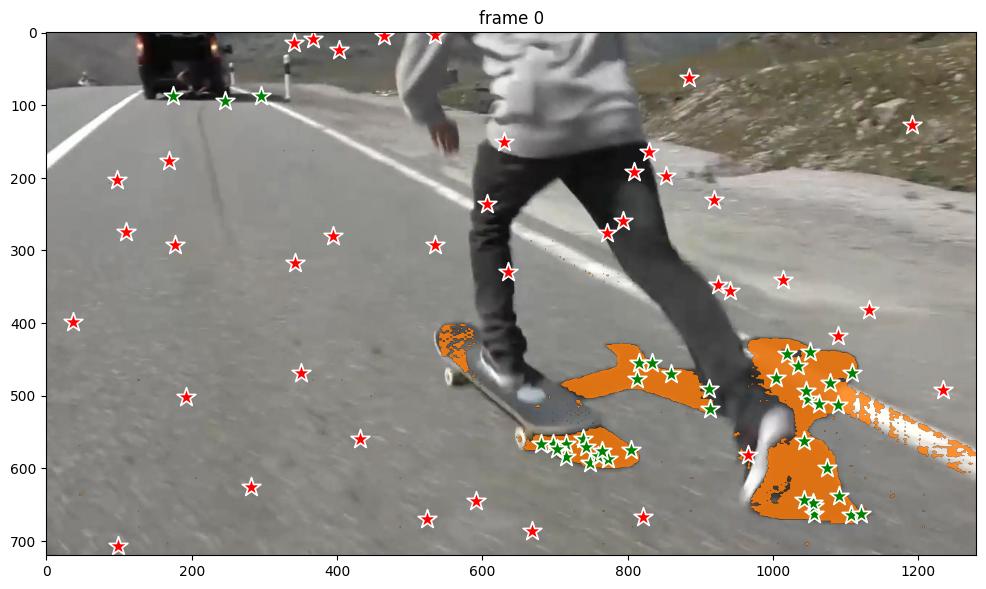}
	\end{subfigure}
	\begin{subfigure}{0.12\textwidth}
		\includegraphics[width=\textwidth]{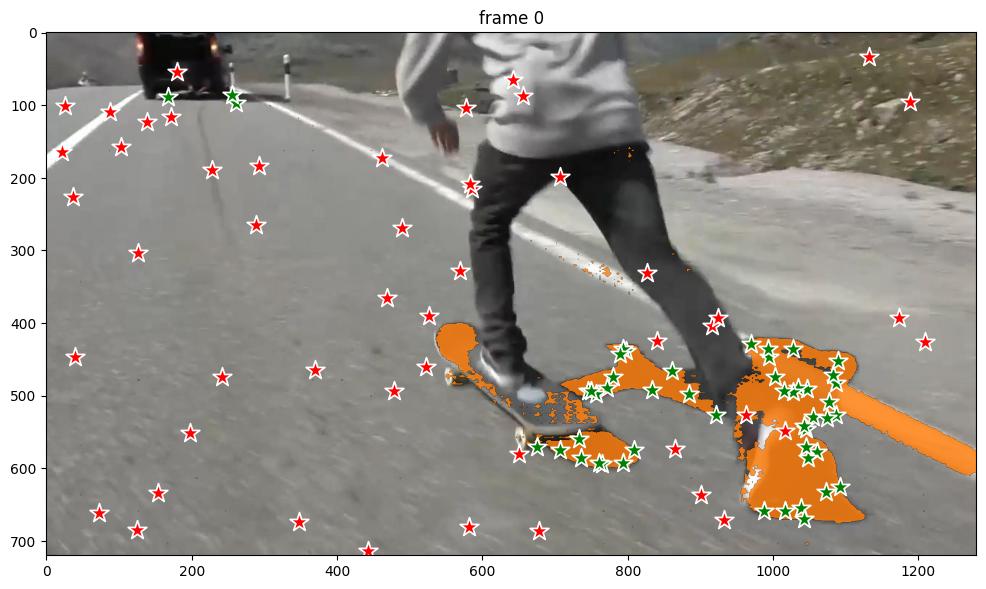}
	\end{subfigure}

	\vspace*{1.3mm}
	\begin{subfigure}{0.12\textwidth}
		\includegraphics[width=\textwidth]{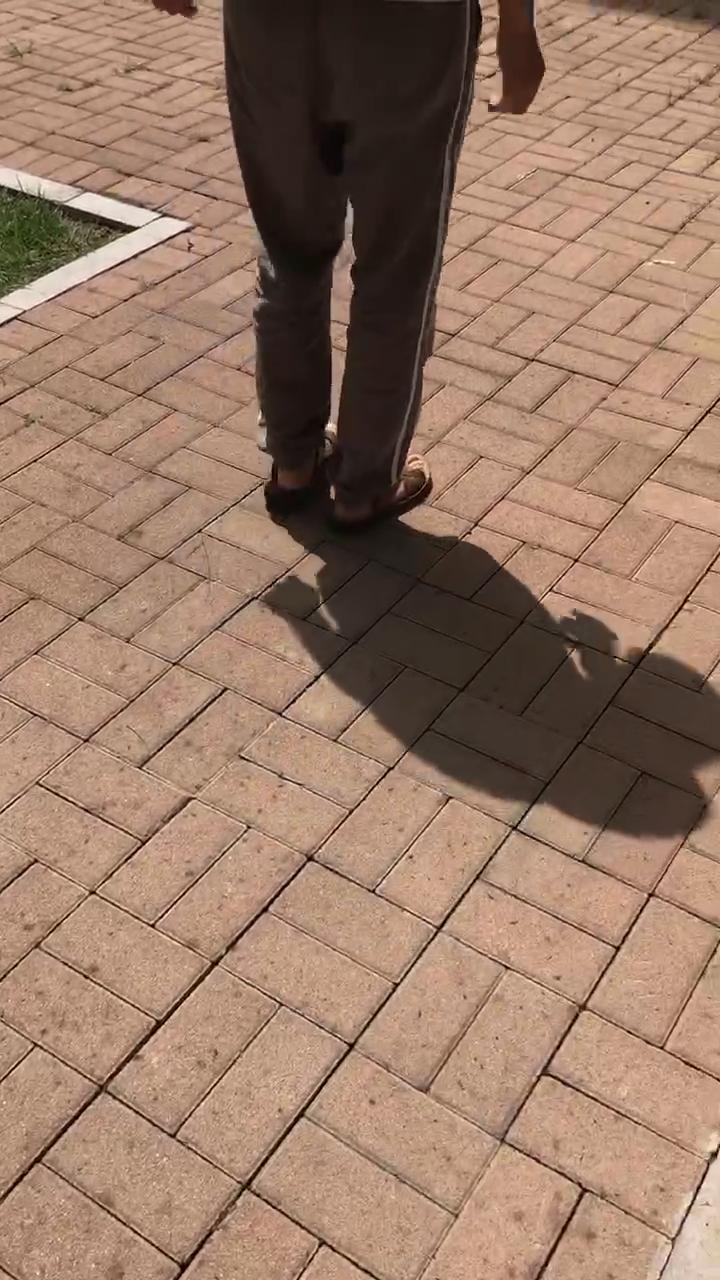}
	\end{subfigure}
	\begin{subfigure}{0.12\textwidth}
		\includegraphics[width=\textwidth]{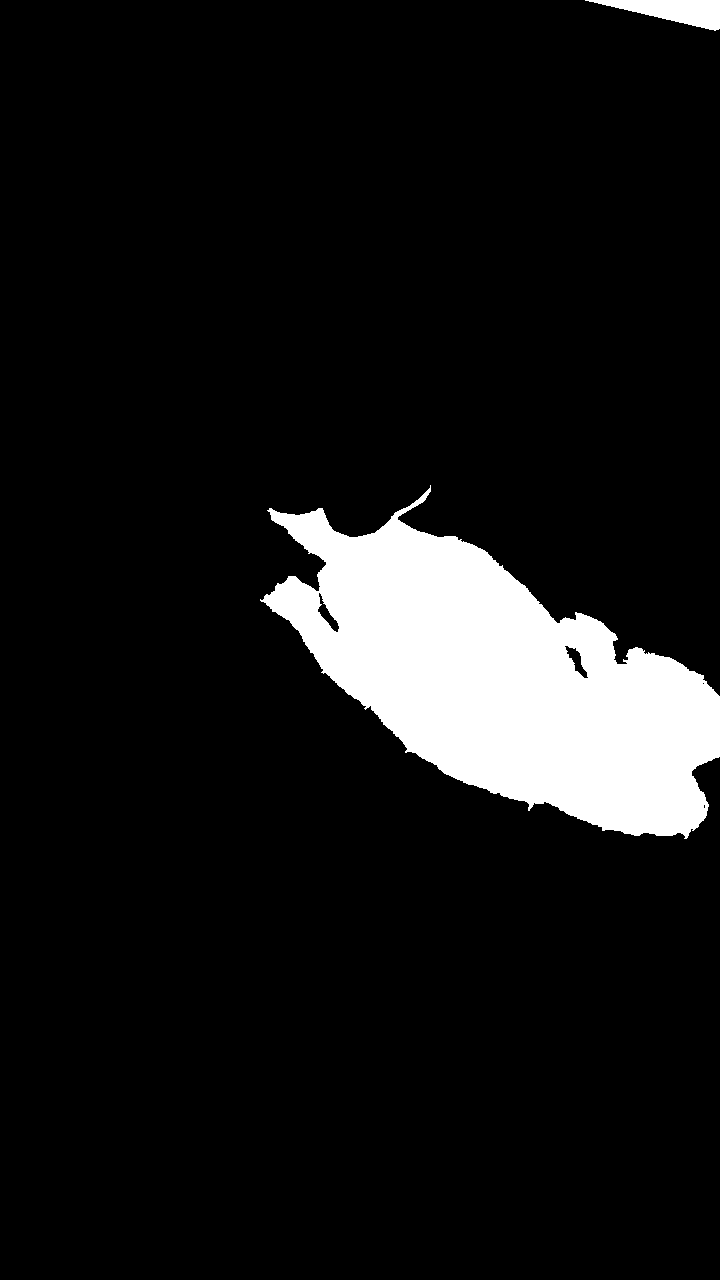}
	\end{subfigure}
	\begin{subfigure}{0.12\textwidth}
		\includegraphics[width=\textwidth]{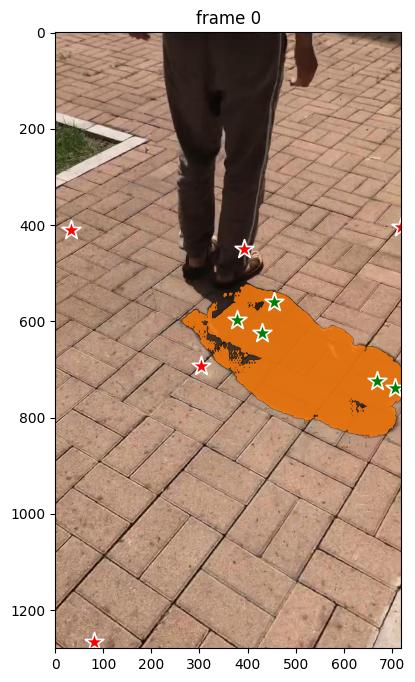}
	\end{subfigure}
	\begin{subfigure}{0.12\textwidth}
		\includegraphics[width=\textwidth]{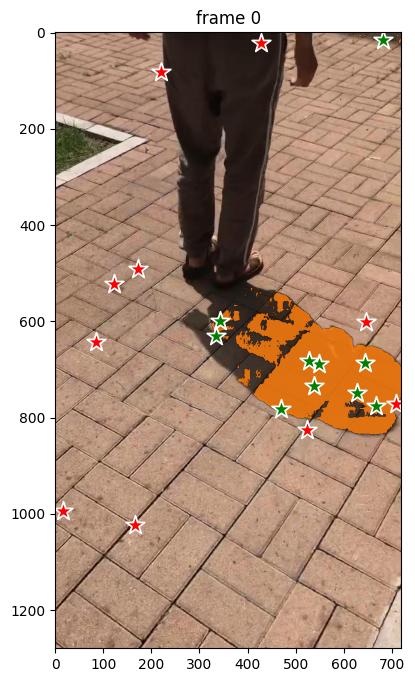}
	\end{subfigure}
	\begin{subfigure}{0.12\textwidth}
		\includegraphics[width=\textwidth]{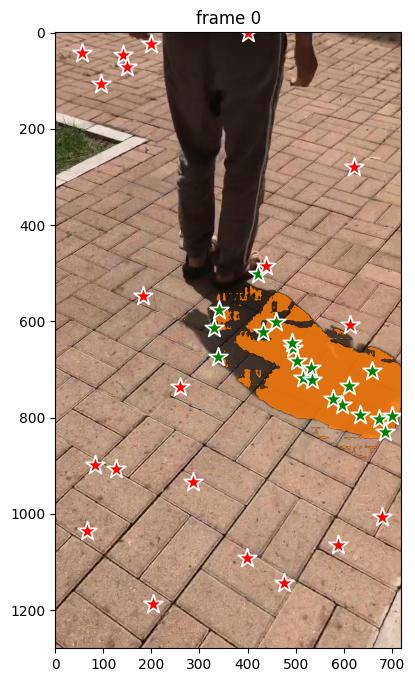}
	\end{subfigure}
	\begin{subfigure}{0.12\textwidth}
		\includegraphics[width=\textwidth]{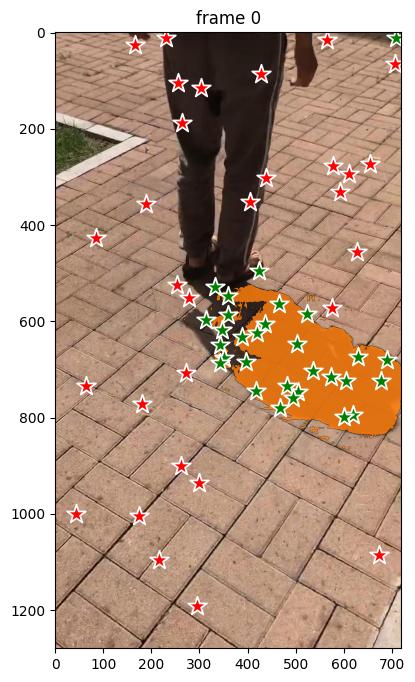}
	\end{subfigure}
	\begin{subfigure}{0.12\textwidth}
		\includegraphics[width=\textwidth]{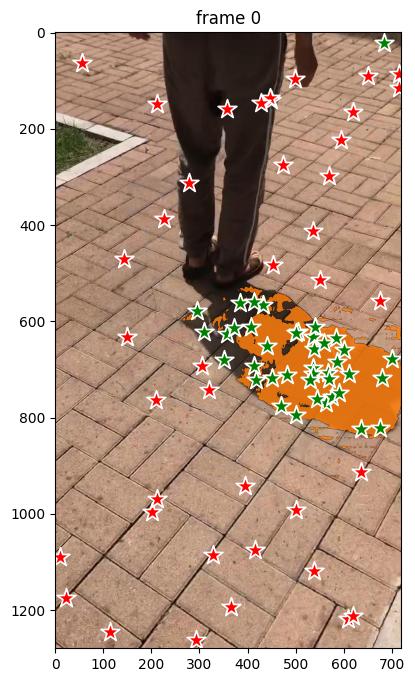}
	\end{subfigure}
	\begin{subfigure}{0.12\textwidth}
		\includegraphics[width=\textwidth]{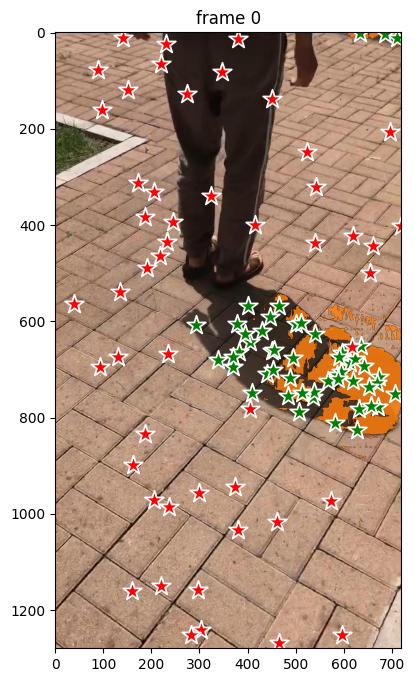}
	\end{subfigure}

	\vspace*{1.3mm}
	\begin{subfigure}{0.12\textwidth}
		\includegraphics[width=\textwidth]{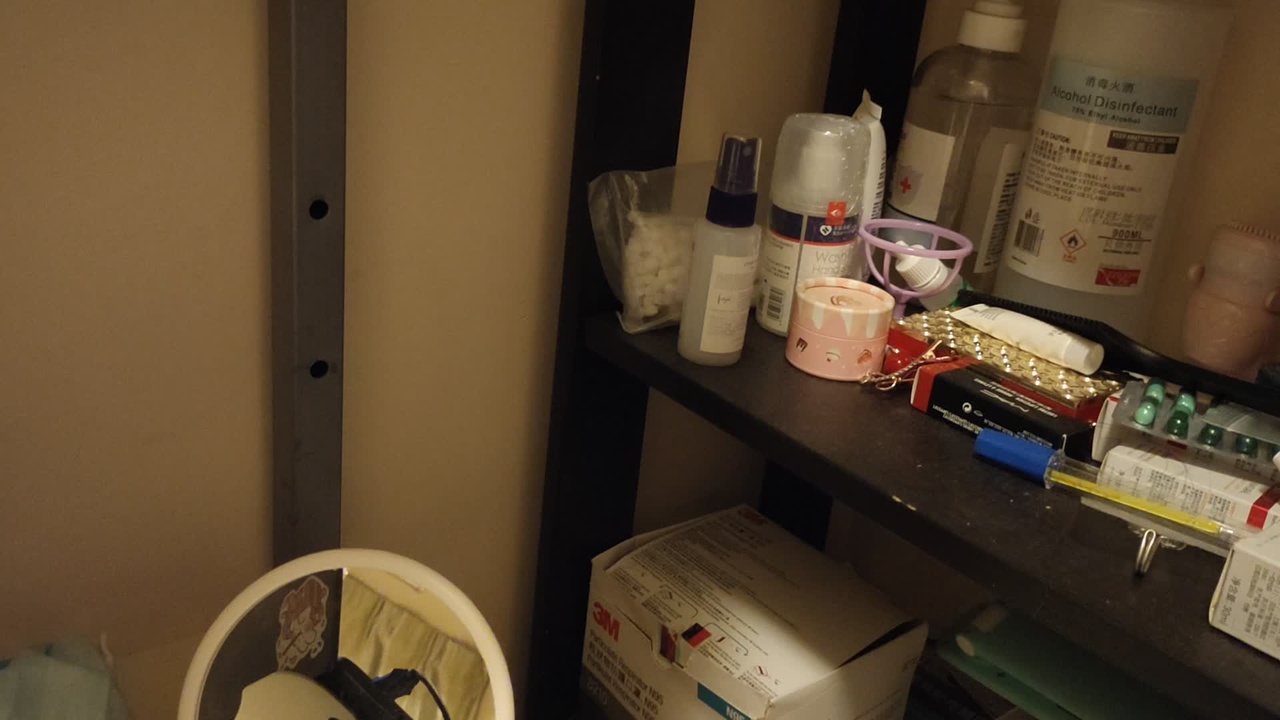}
	\end{subfigure}
	\begin{subfigure}{0.12\textwidth}
		\includegraphics[width=\textwidth]{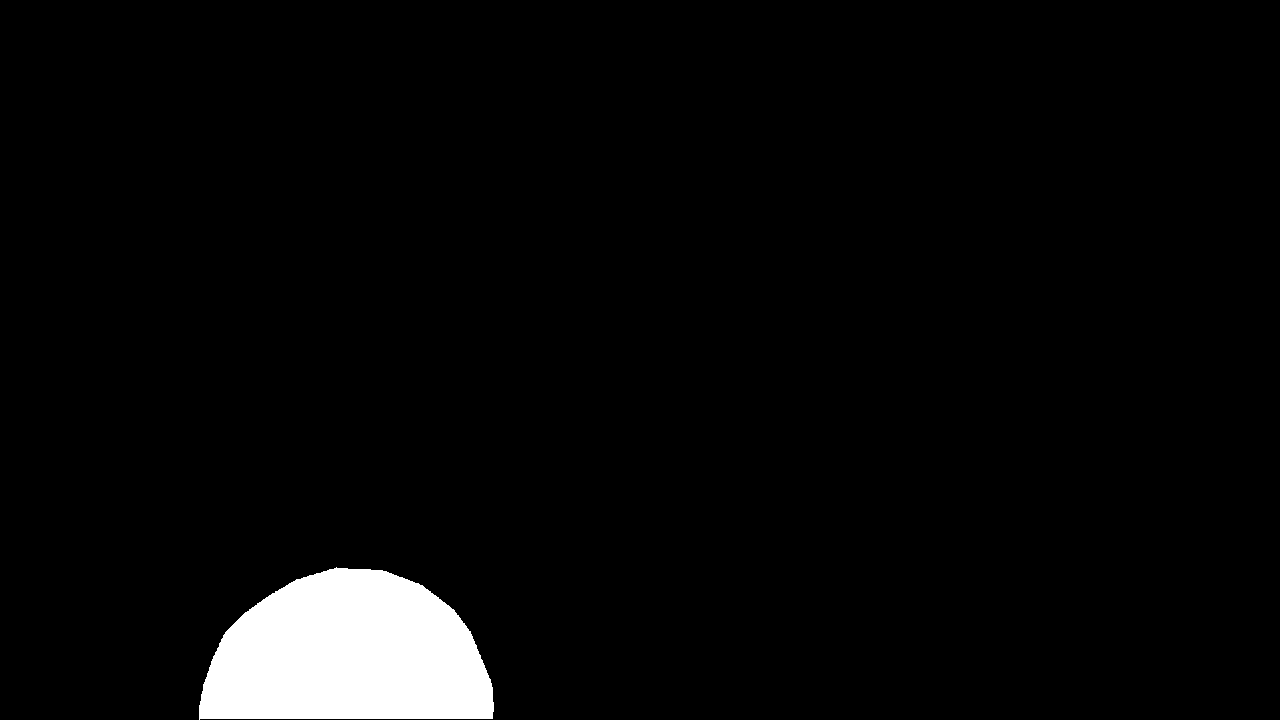}
	\end{subfigure}
	\begin{subfigure}{0.12\textwidth}
		\includegraphics[width=\textwidth]{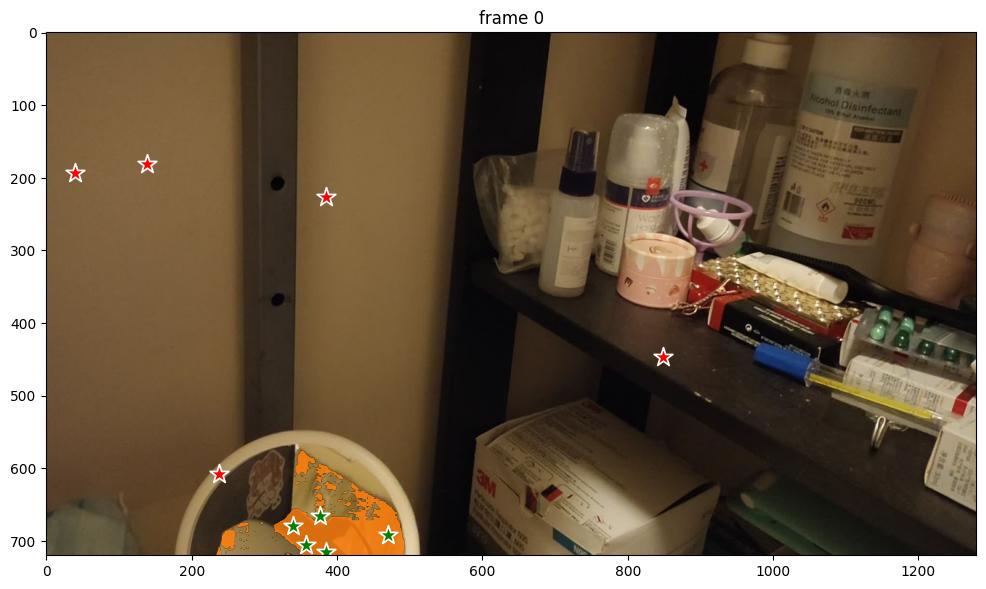}
	\end{subfigure}
	\begin{subfigure}{0.12\textwidth}
		\includegraphics[width=\textwidth]{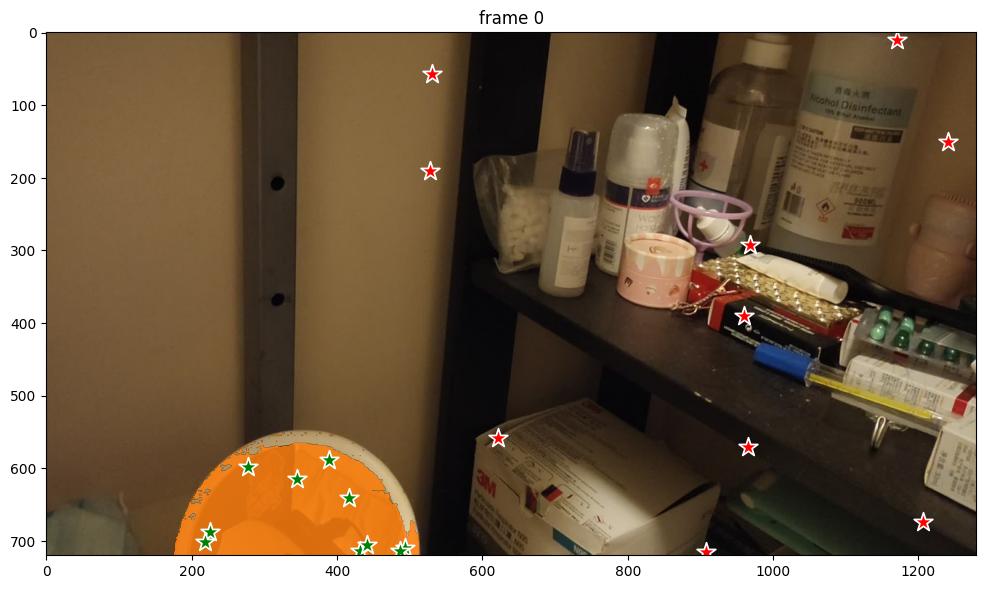}
	\end{subfigure}
	\begin{subfigure}{0.12\textwidth}
		\includegraphics[width=\textwidth]{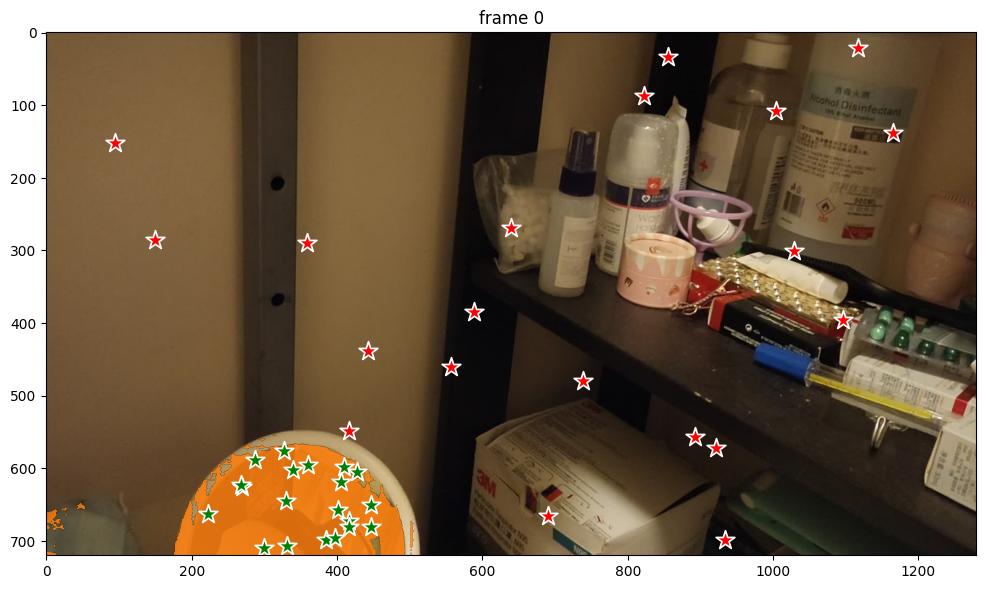}
	\end{subfigure}
	\begin{subfigure}{0.12\textwidth}
		\includegraphics[width=\textwidth]{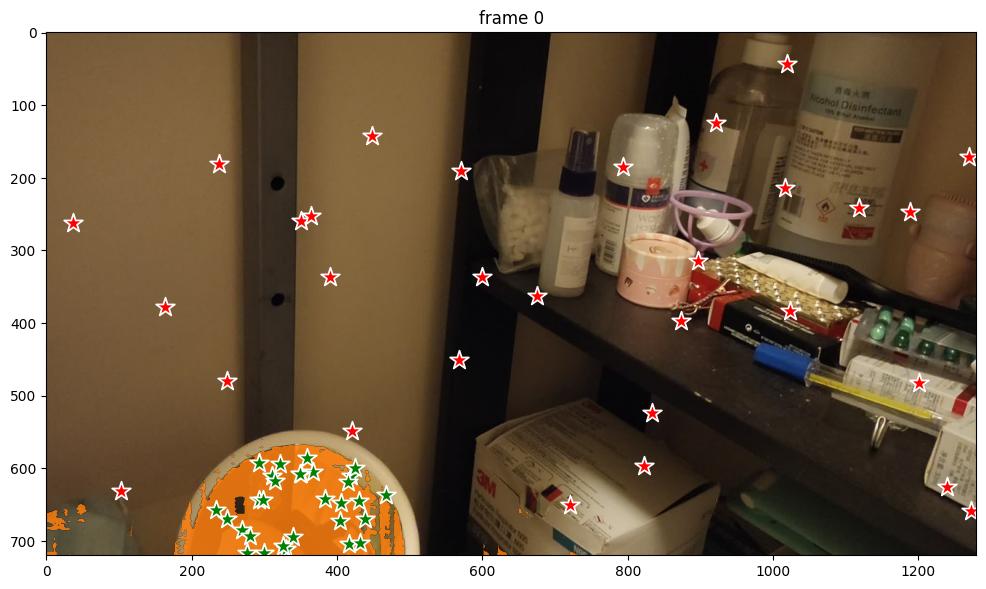}
	\end{subfigure}
	\begin{subfigure}{0.12\textwidth}
		\includegraphics[width=\textwidth]{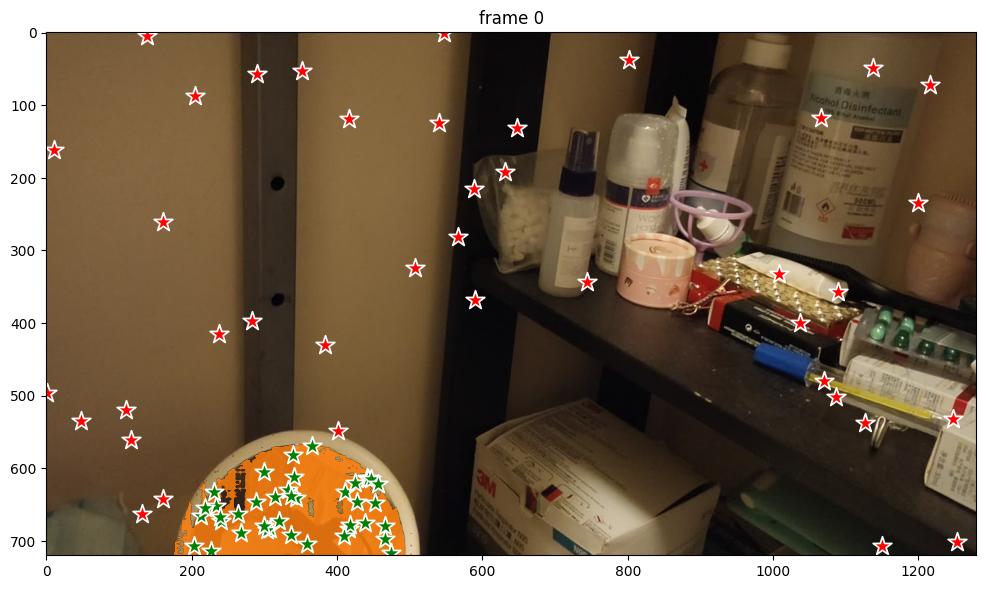}
	\end{subfigure}
	\begin{subfigure}{0.12\textwidth}
		\includegraphics[width=\textwidth]{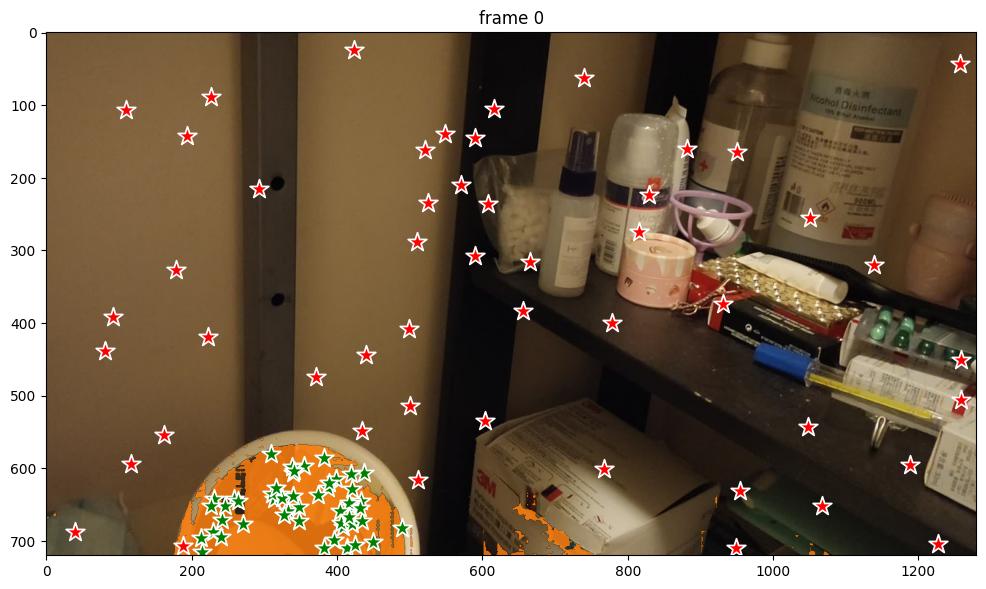}
	\end{subfigure}

	\vspace*{1.3mm}
	\begin{subfigure}{0.12\textwidth}
		\includegraphics[width=\textwidth]{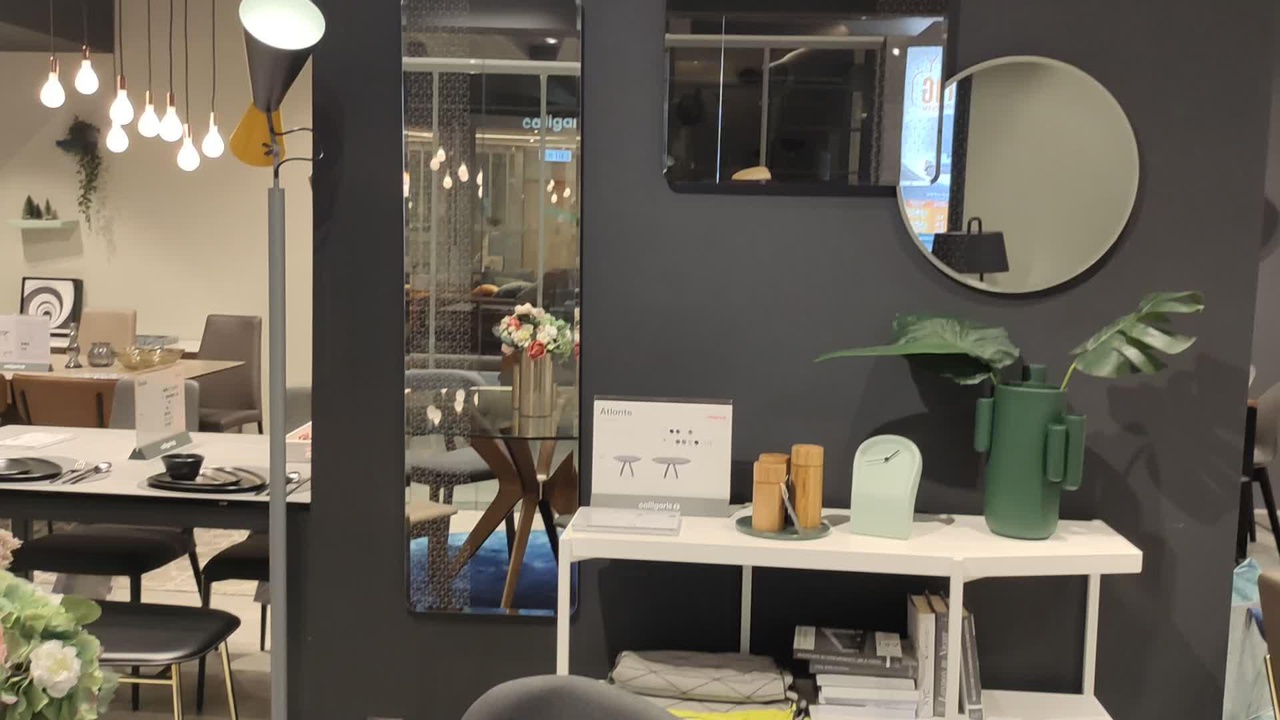}
	\end{subfigure}
	\begin{subfigure}{0.12\textwidth}
		\includegraphics[width=\textwidth]{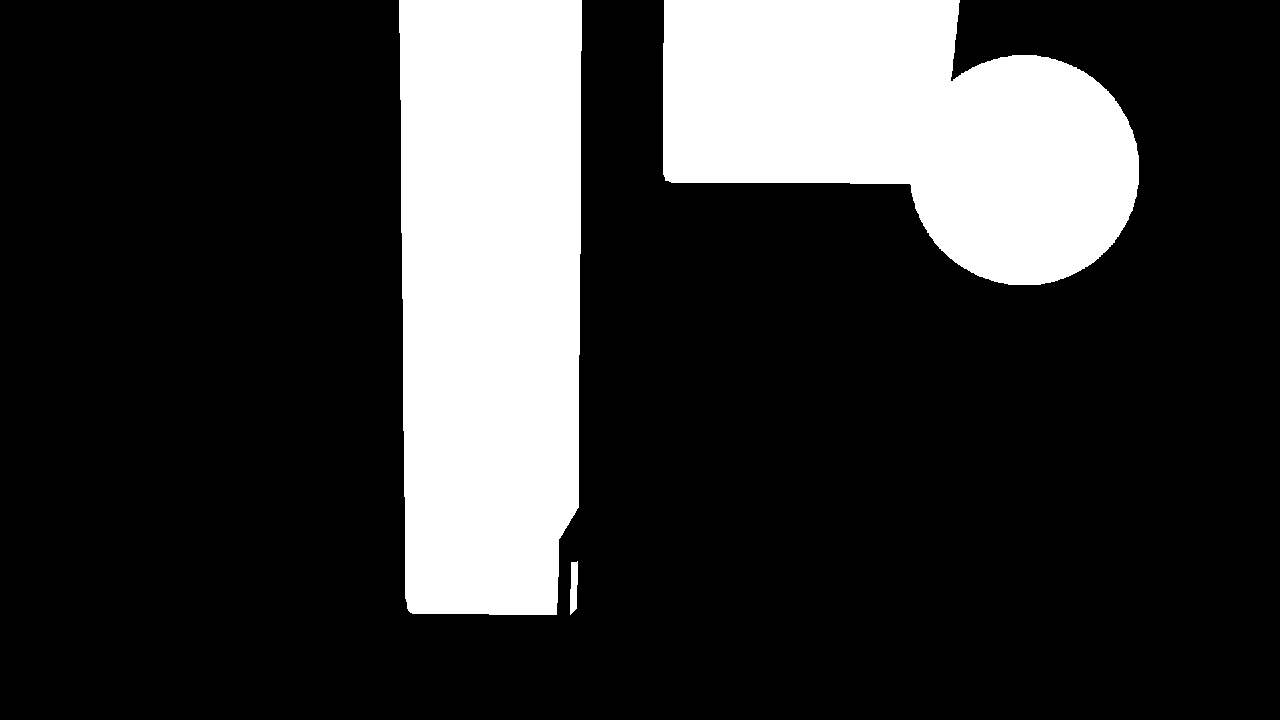}
	\end{subfigure}
	\begin{subfigure}{0.12\textwidth}
		\includegraphics[width=\textwidth]{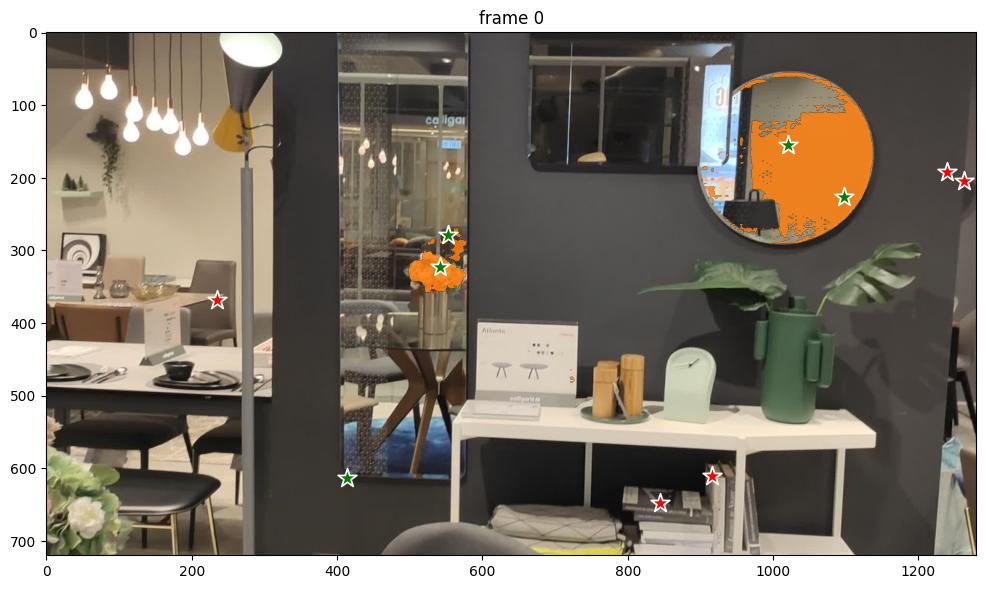}
	\end{subfigure}
	\begin{subfigure}{0.12\textwidth}
		\includegraphics[width=\textwidth]{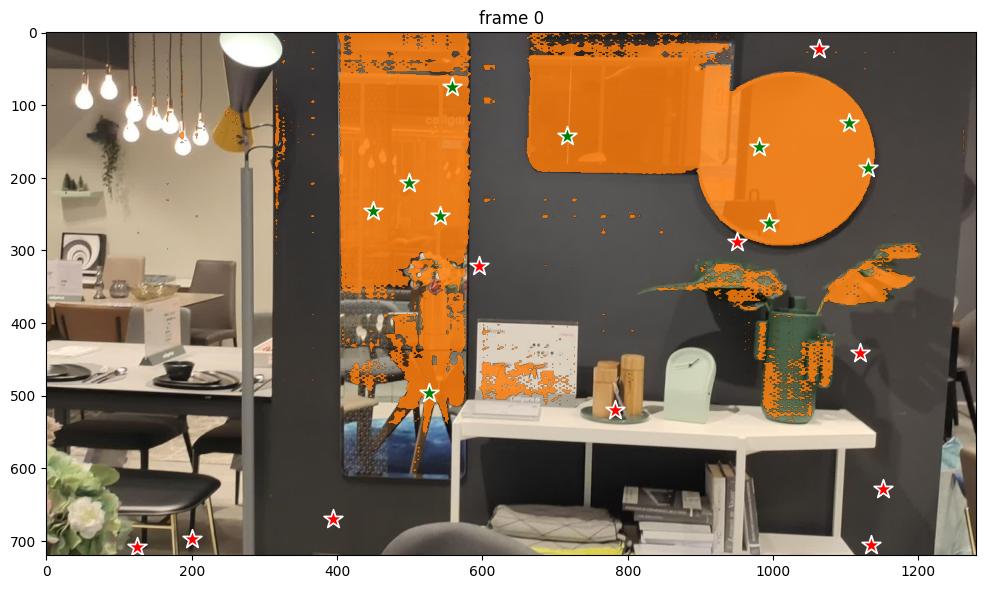}
	\end{subfigure}
	\begin{subfigure}{0.12\textwidth}
		\includegraphics[width=\textwidth]{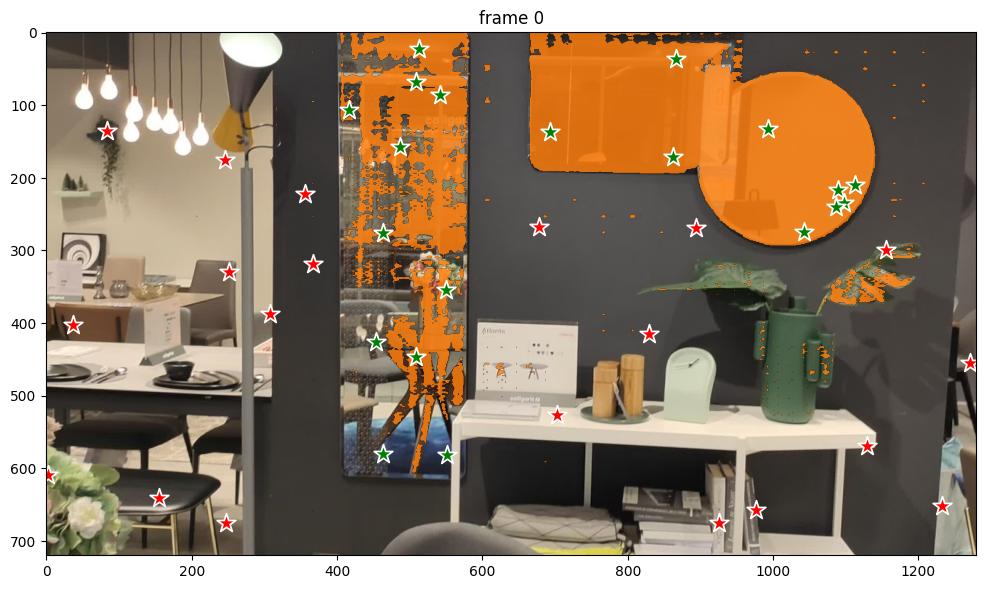}
	\end{subfigure}
	\begin{subfigure}{0.12\textwidth}
		\includegraphics[width=\textwidth]{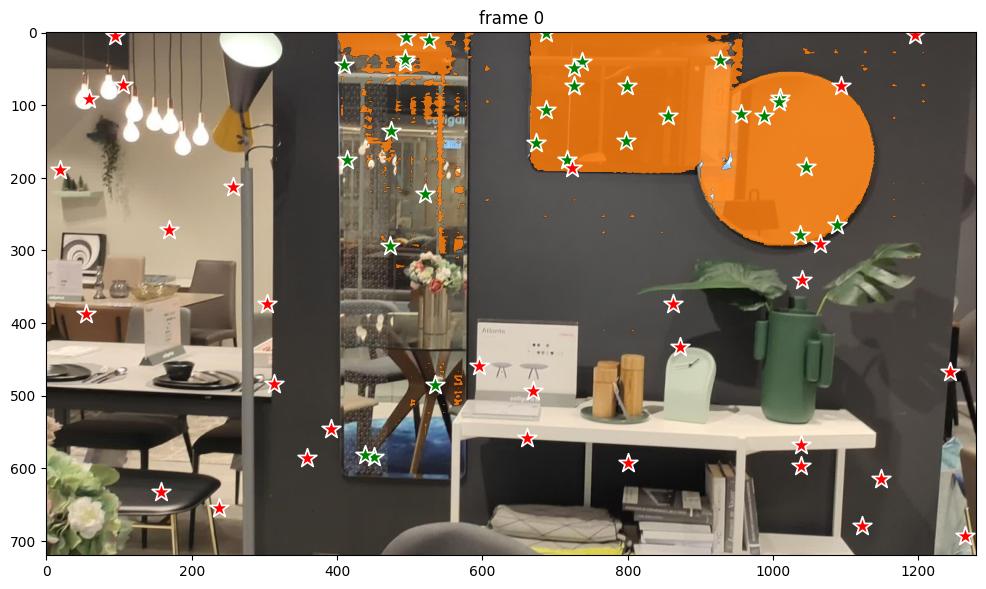}
	\end{subfigure}
	\begin{subfigure}{0.12\textwidth}
		\includegraphics[width=\textwidth]{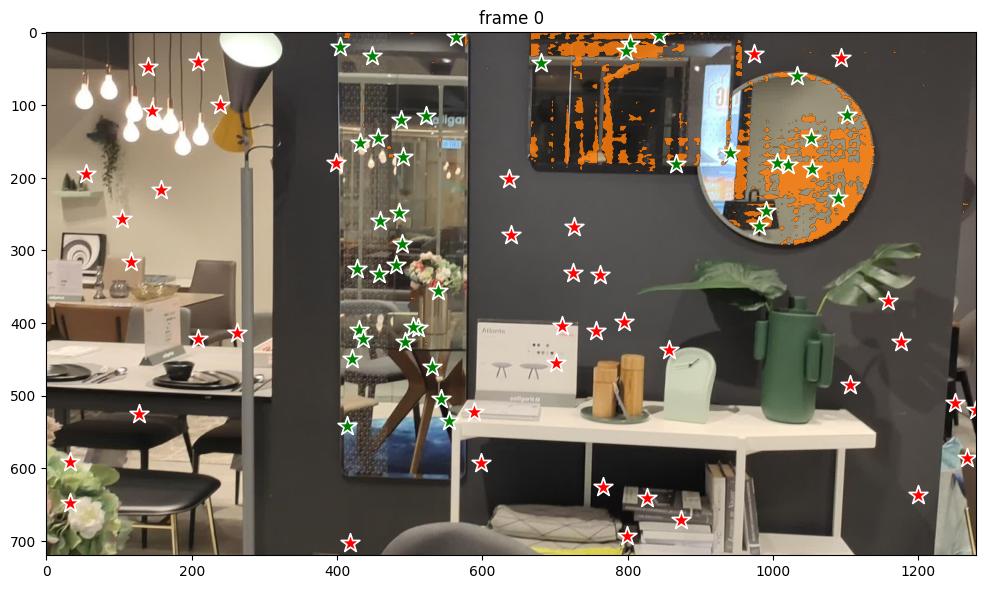}
	\end{subfigure}
	\begin{subfigure}{0.12\textwidth}
		\includegraphics[width=\textwidth]{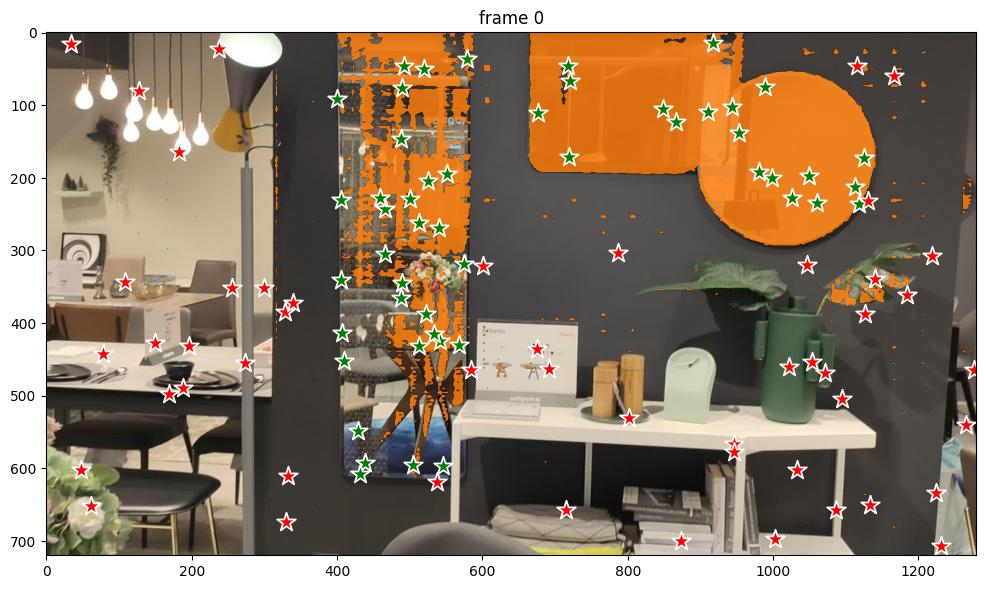}
	\end{subfigure}

	\vspace*{1.3mm}
	\begin{subfigure}{0.12\textwidth}
		\includegraphics[width=\textwidth]{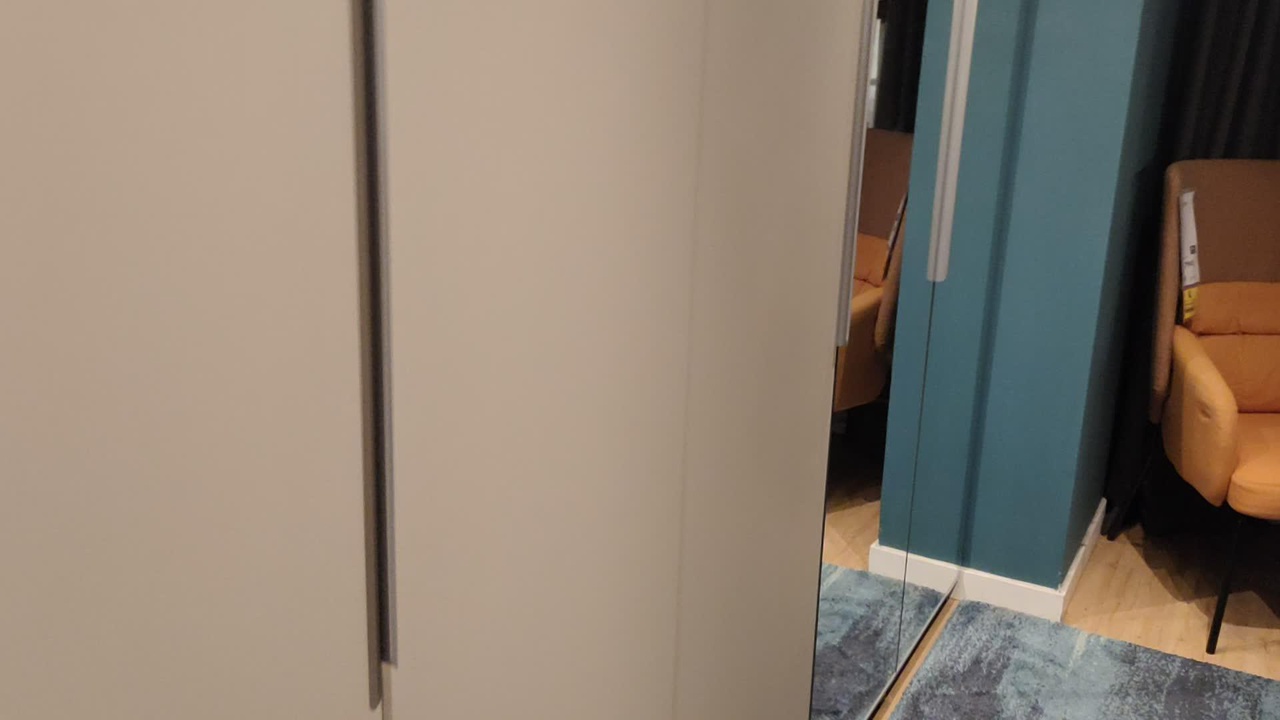}
	\end{subfigure}
	\begin{subfigure}{0.12\textwidth}
		\includegraphics[width=\textwidth]{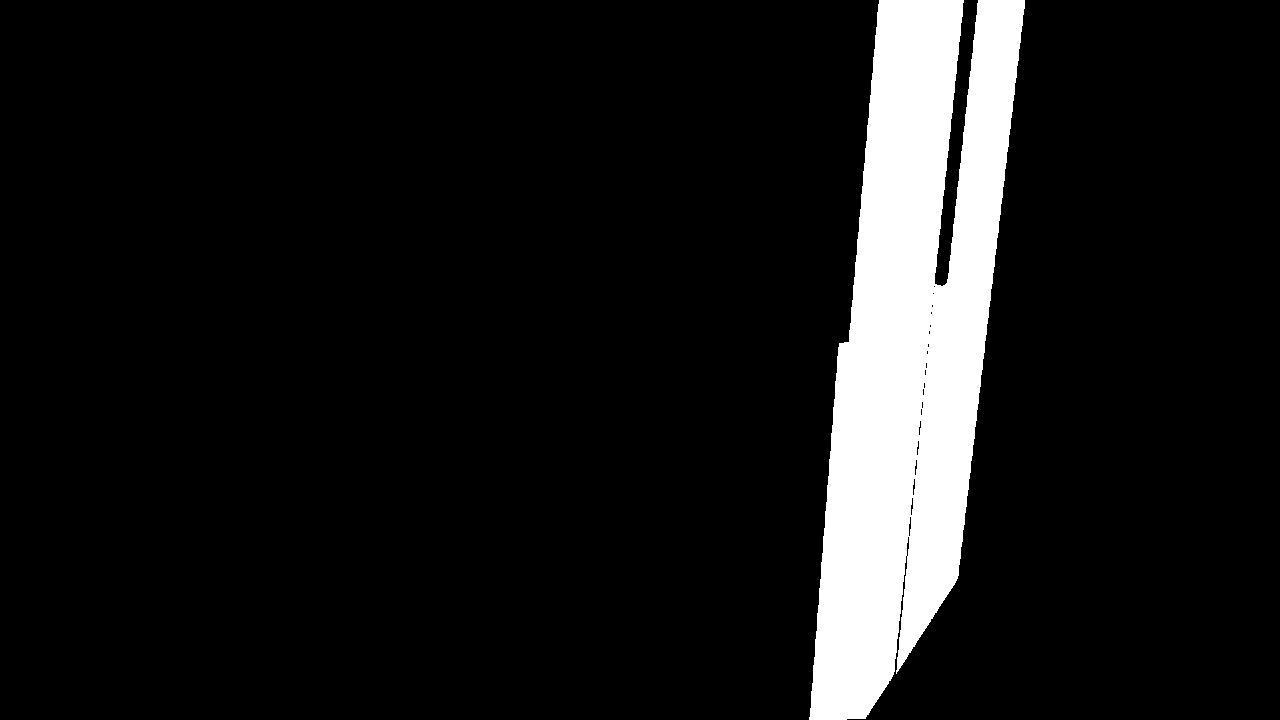}
	\end{subfigure}
	\begin{subfigure}{0.12\textwidth}
		\includegraphics[width=\textwidth]{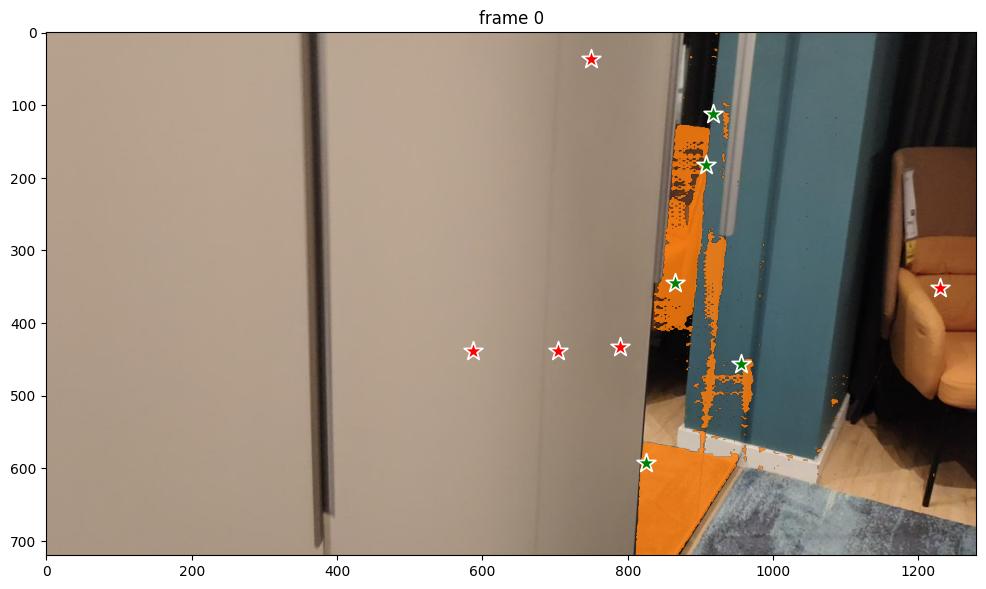}
	\end{subfigure}
	\begin{subfigure}{0.12\textwidth}
		\includegraphics[width=\textwidth]{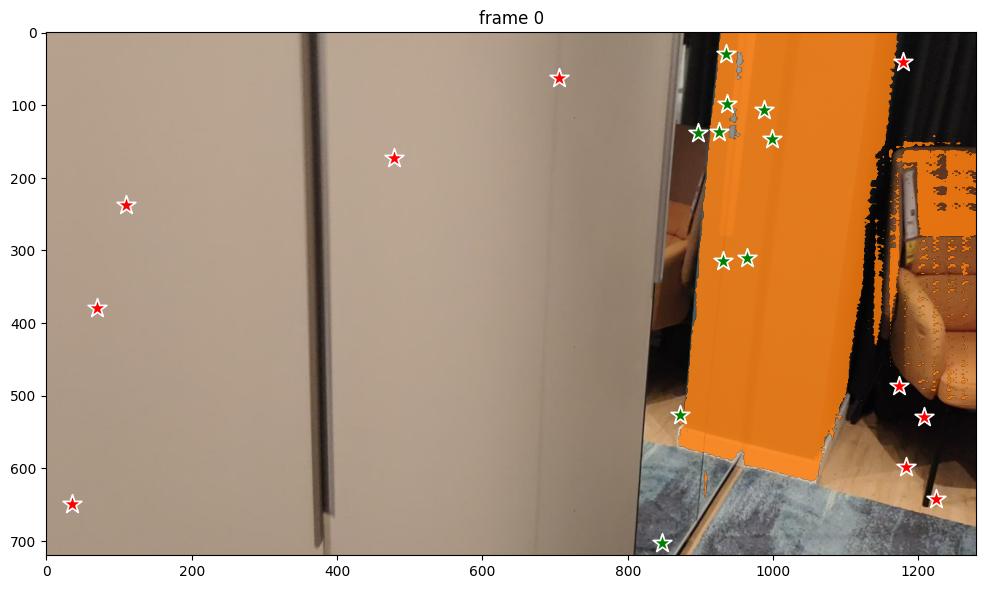}
	\end{subfigure}
	\begin{subfigure}{0.12\textwidth}
		\includegraphics[width=\textwidth]{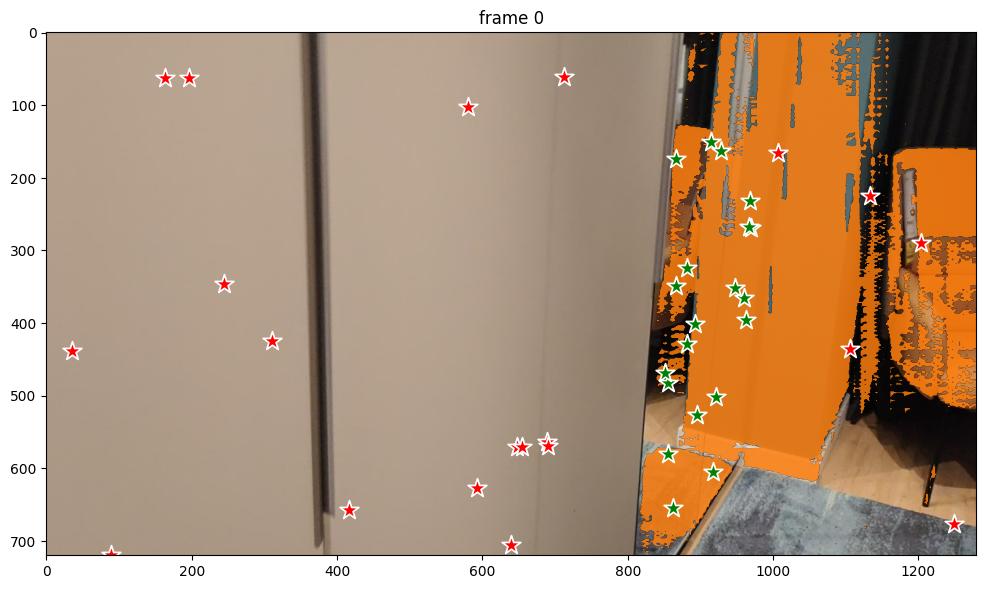}
	\end{subfigure}
	\begin{subfigure}{0.12\textwidth}
		\includegraphics[width=\textwidth]{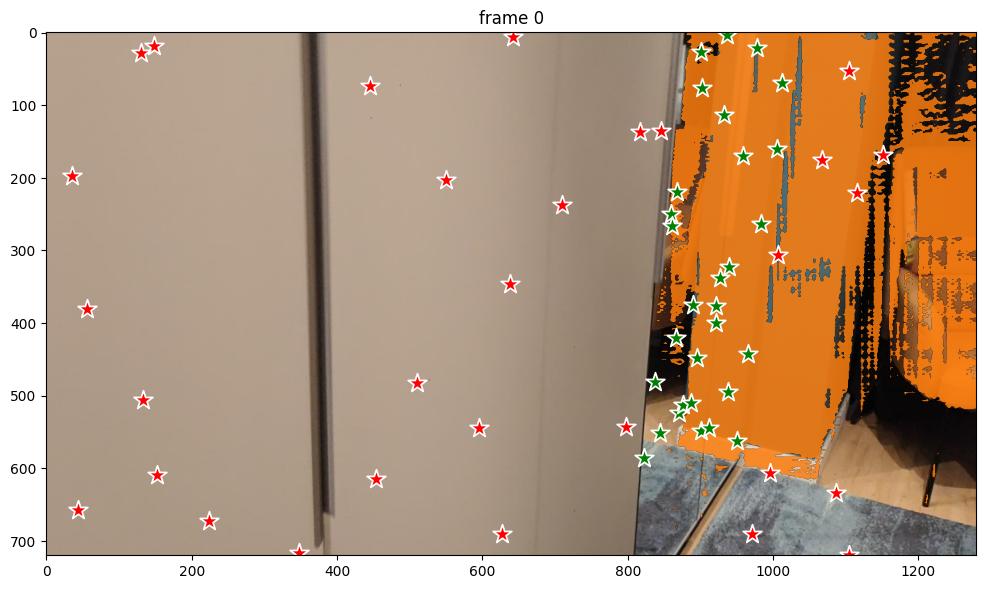}
	\end{subfigure}
	\begin{subfigure}{0.12\textwidth}
		\includegraphics[width=\textwidth]{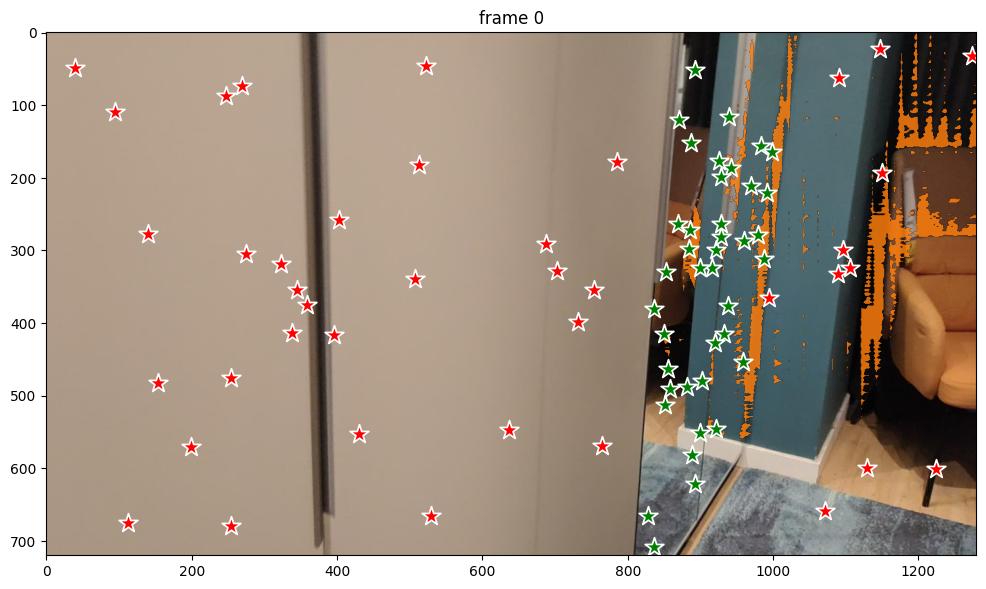}
	\end{subfigure}
	\begin{subfigure}{0.12\textwidth}
		\includegraphics[width=\textwidth]{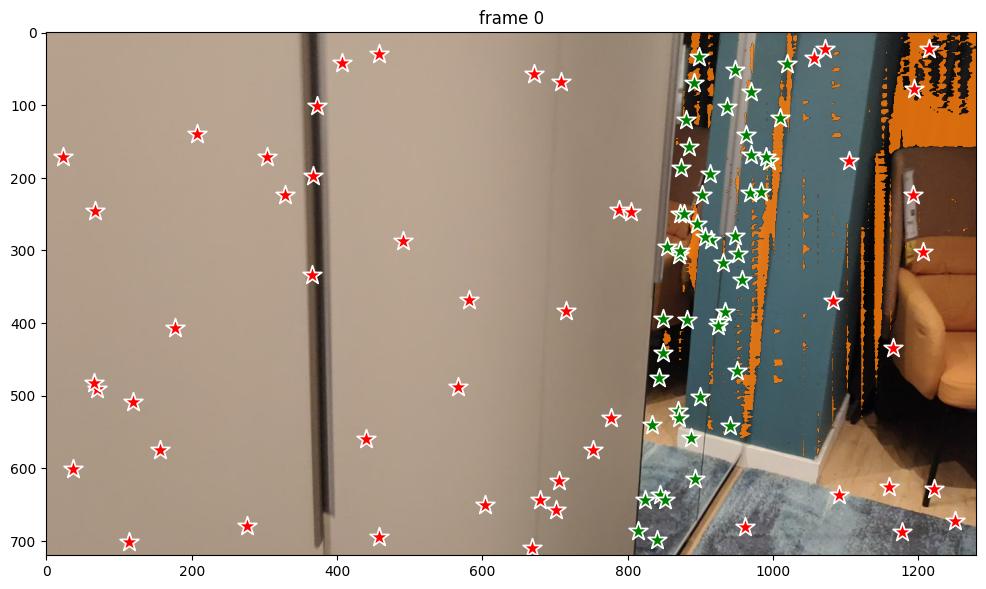}
	\end{subfigure}

	\vspace*{1.3mm}
	\begin{subfigure}{0.12\textwidth}
		\includegraphics[width=\textwidth]{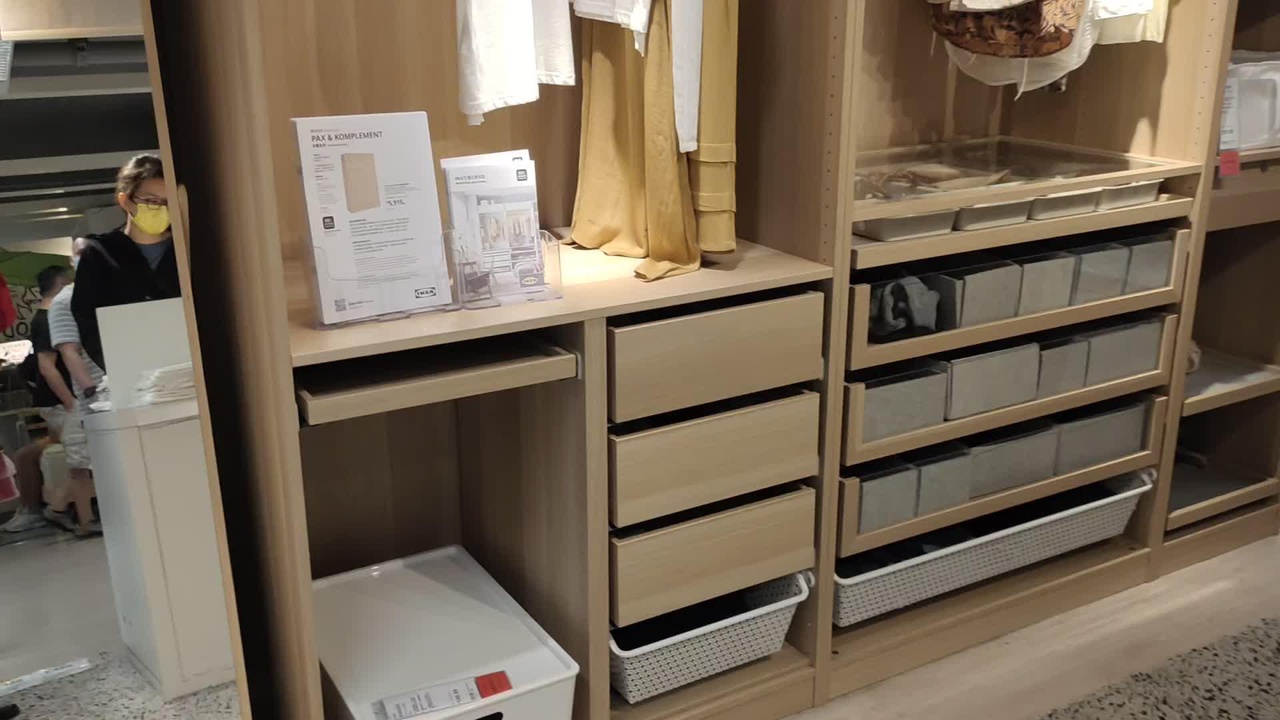}
	\end{subfigure}
	\begin{subfigure}{0.12\textwidth}
		\includegraphics[width=\textwidth]{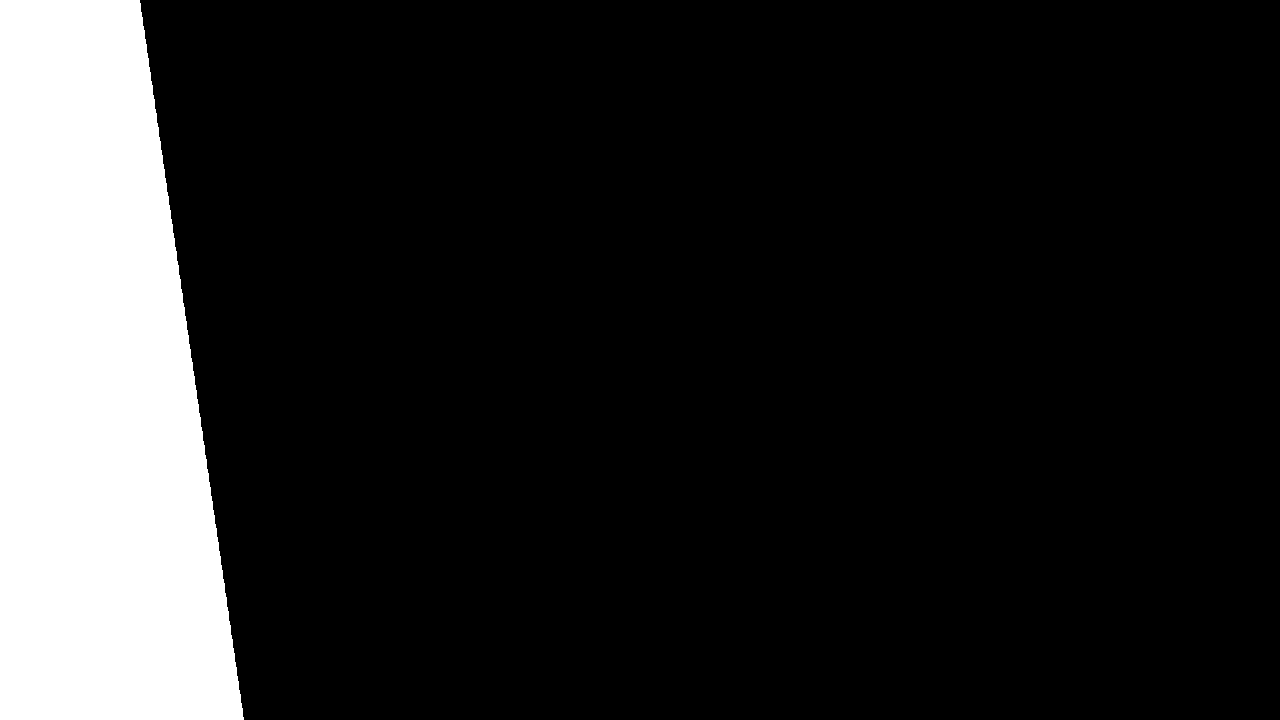}
	\end{subfigure}
	\begin{subfigure}{0.12\textwidth}
		\includegraphics[width=\textwidth]{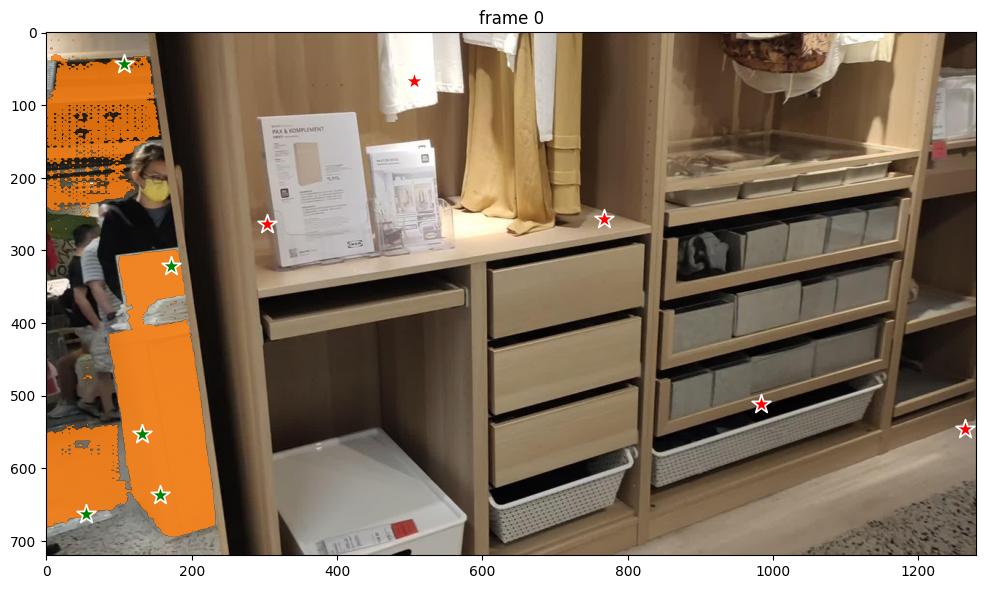}
	\end{subfigure}
	\begin{subfigure}{0.12\textwidth}
		\includegraphics[width=\textwidth]{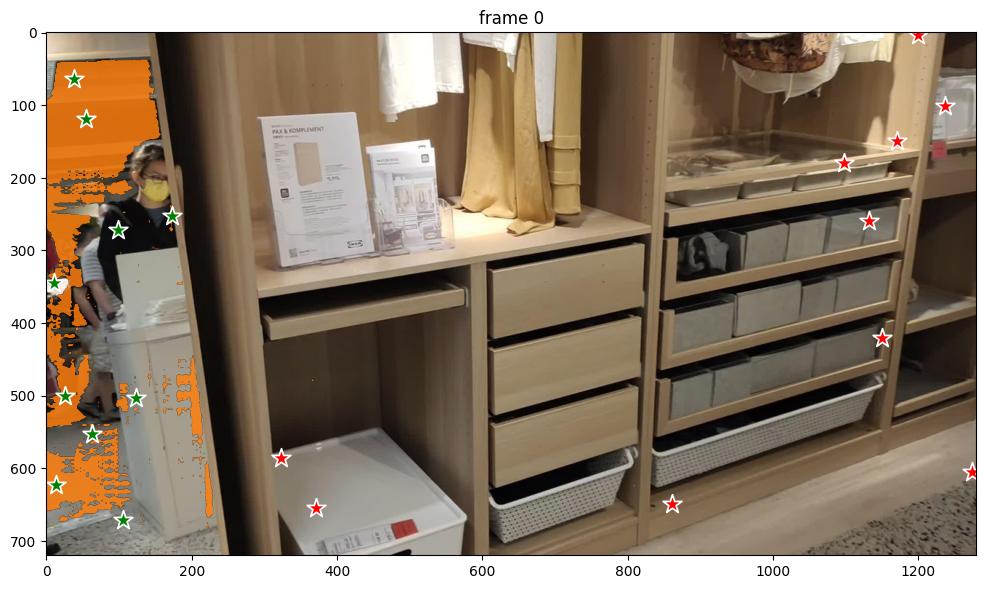}
	\end{subfigure}
	\begin{subfigure}{0.12\textwidth}
		\includegraphics[width=\textwidth]{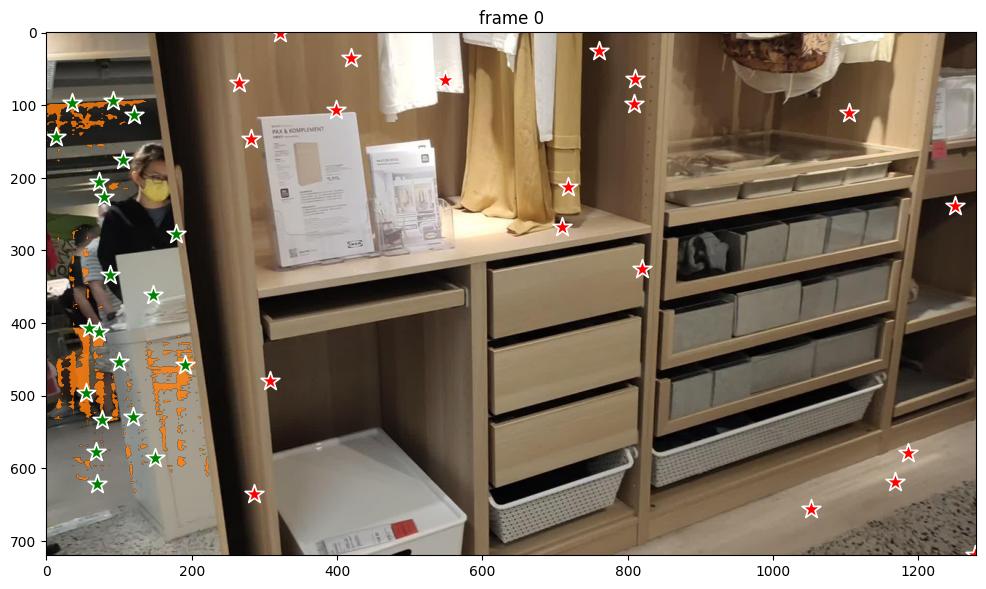}
	\end{subfigure}
	\begin{subfigure}{0.12\textwidth}
		\includegraphics[width=\textwidth]{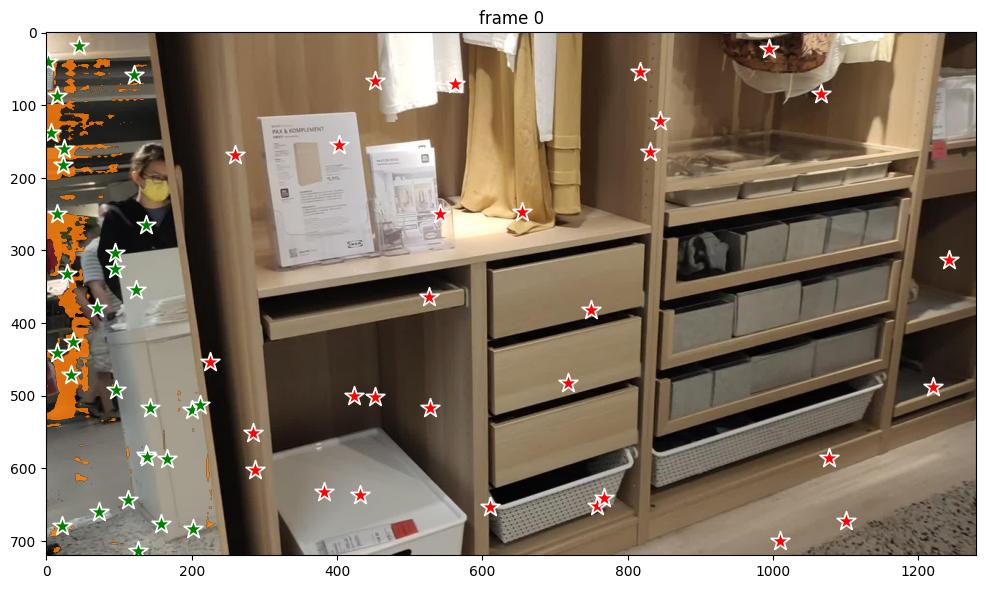}
	\end{subfigure}
	\begin{subfigure}{0.12\textwidth}
		\includegraphics[width=\textwidth]{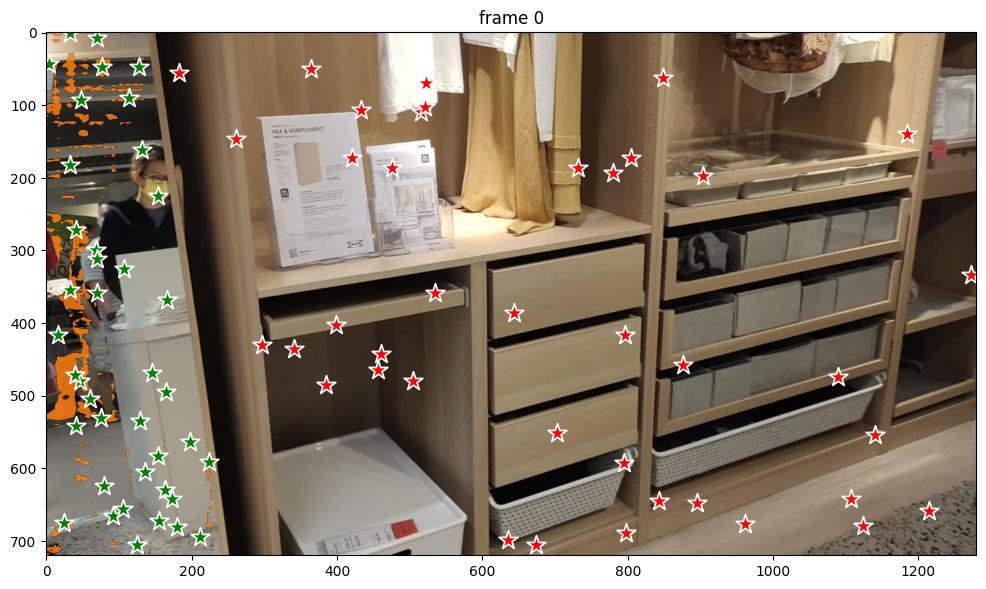}
	\end{subfigure}
	\begin{subfigure}{0.12\textwidth}
		\includegraphics[width=\textwidth]{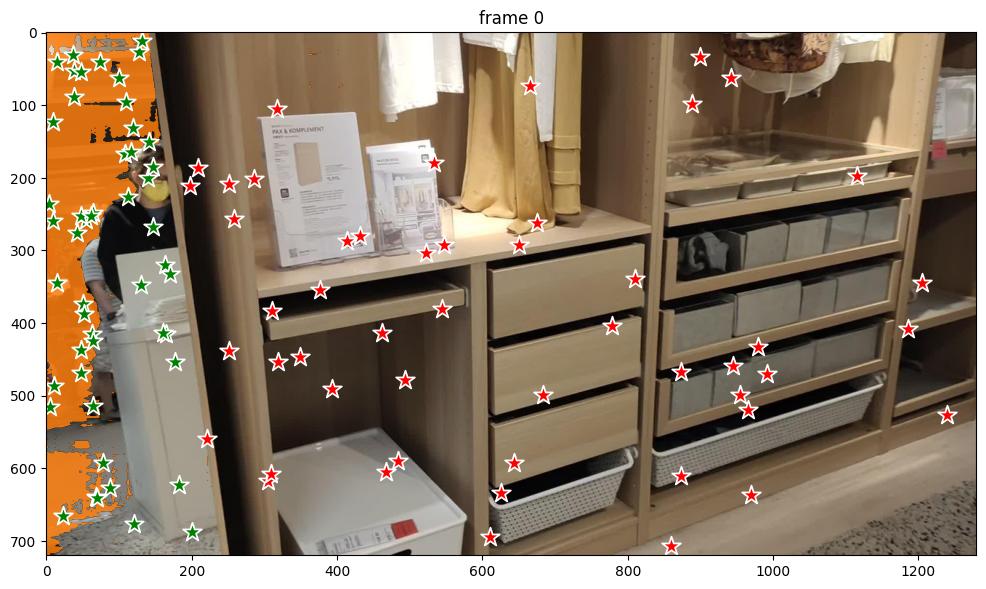}
	\end{subfigure}

	\vspace*{1.3mm}
	\begin{subfigure}{0.12\textwidth}
		\includegraphics[width=\textwidth]{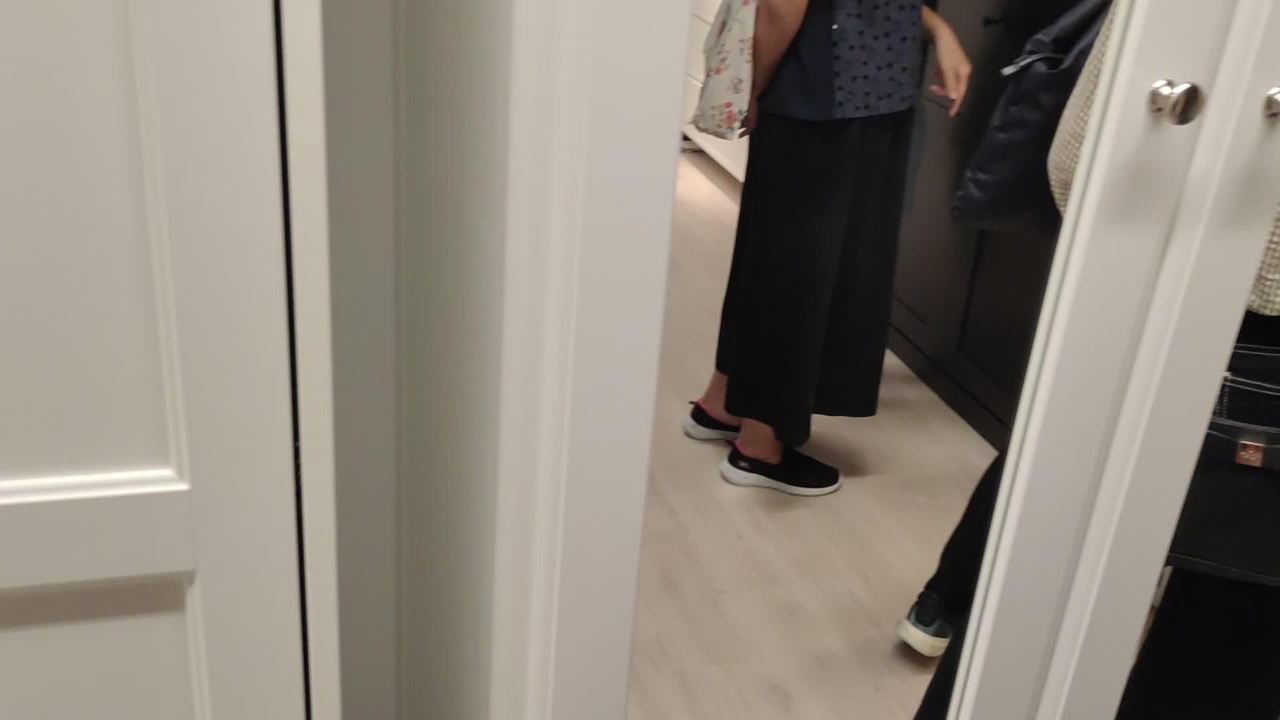}
		\captionsetup{justification=centering}
        \vspace{-5.5mm} \caption{\footnotesize{\\input \\images}}
	\end{subfigure}
	\begin{subfigure}{0.12\textwidth}
		\includegraphics[width=\textwidth]{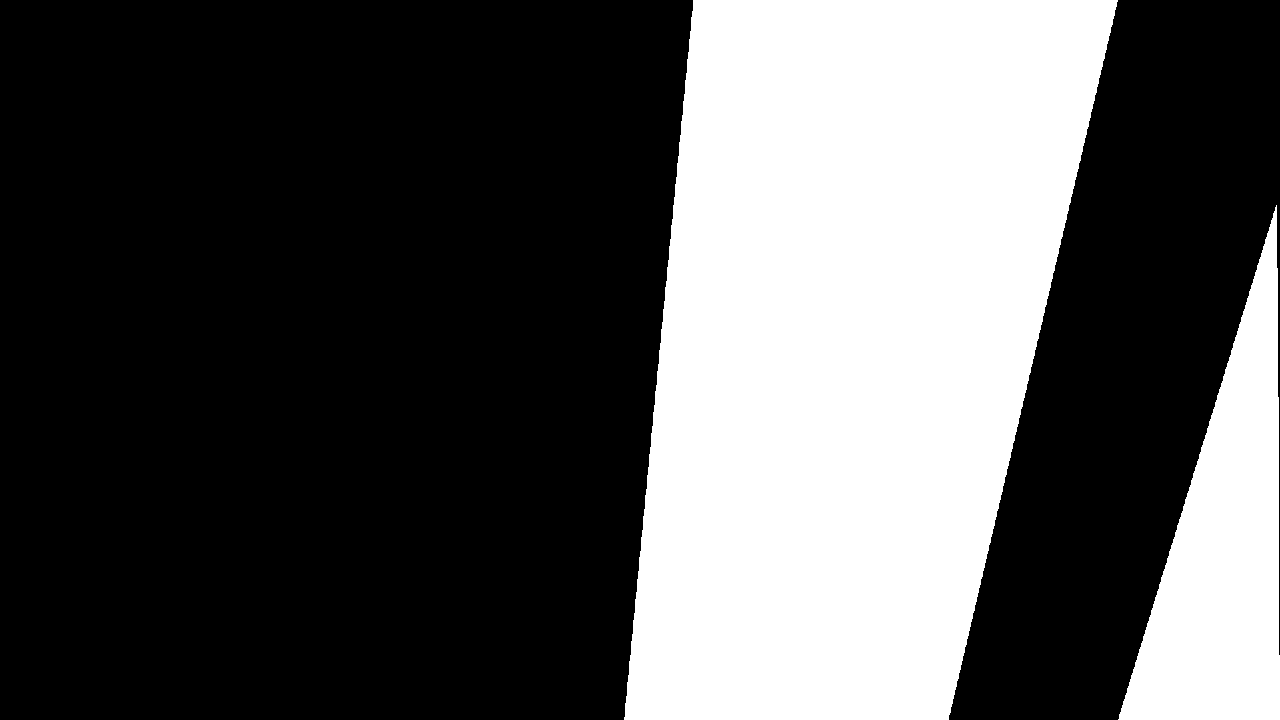}
		\captionsetup{justification=centering}
        \vspace{-5.5mm} \caption{\footnotesize{\\ground \\truths}}
	\end{subfigure}
	\begin{subfigure}{0.12\textwidth}
		\includegraphics[width=\textwidth]{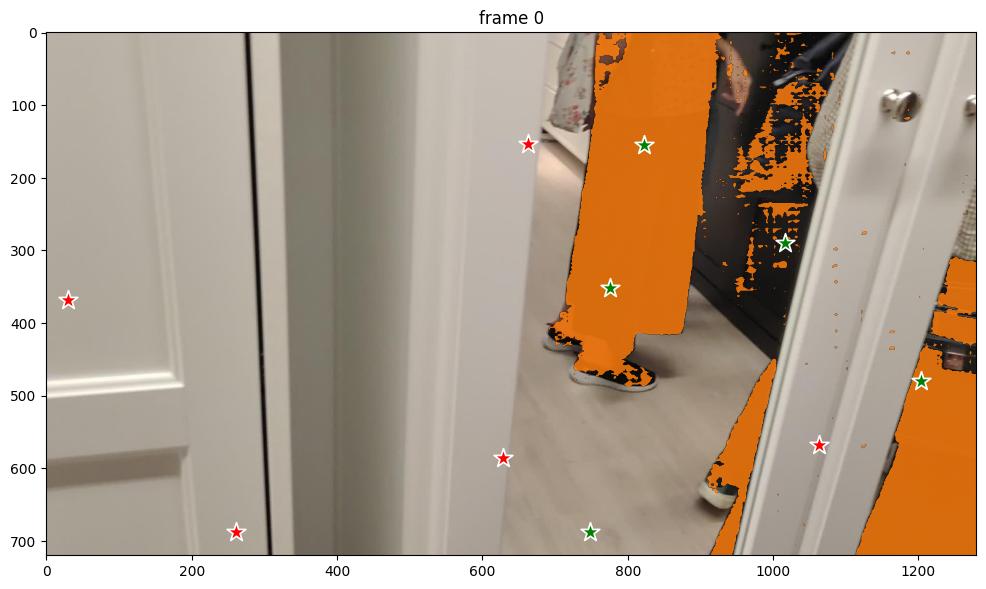}
		\captionsetup{justification=centering}
        \vspace{-5.5mm} \caption{\footnotesize{\\5 \\points}}
	\end{subfigure}
	\begin{subfigure}{0.12\textwidth}
		\includegraphics[width=\textwidth]{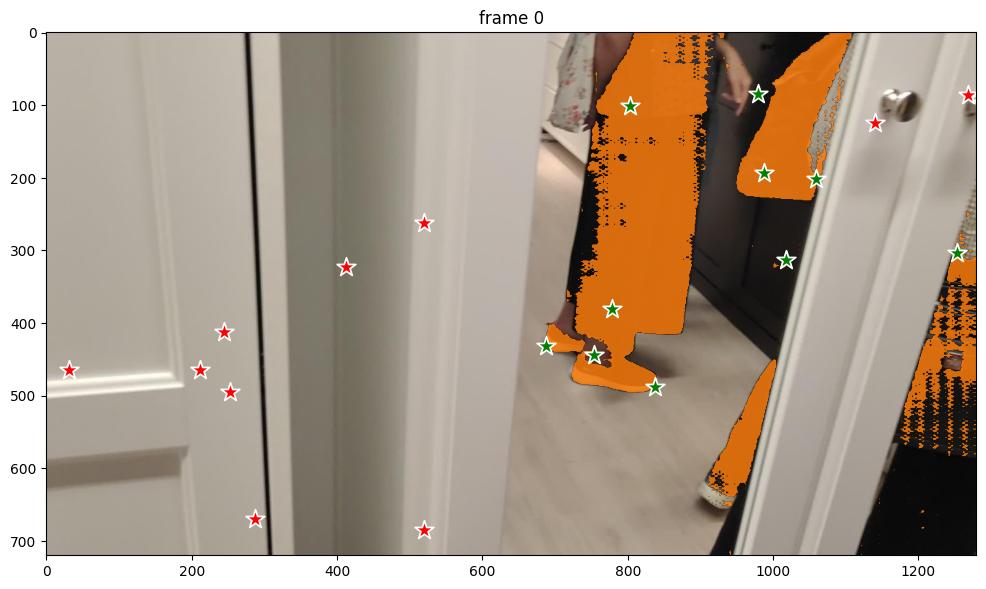}
		\captionsetup{justification=centering}
        \vspace{-5.5mm} \caption{\footnotesize{\\10 \\points}}
	\end{subfigure}
	\begin{subfigure}{0.12\textwidth}
		\includegraphics[width=\textwidth]{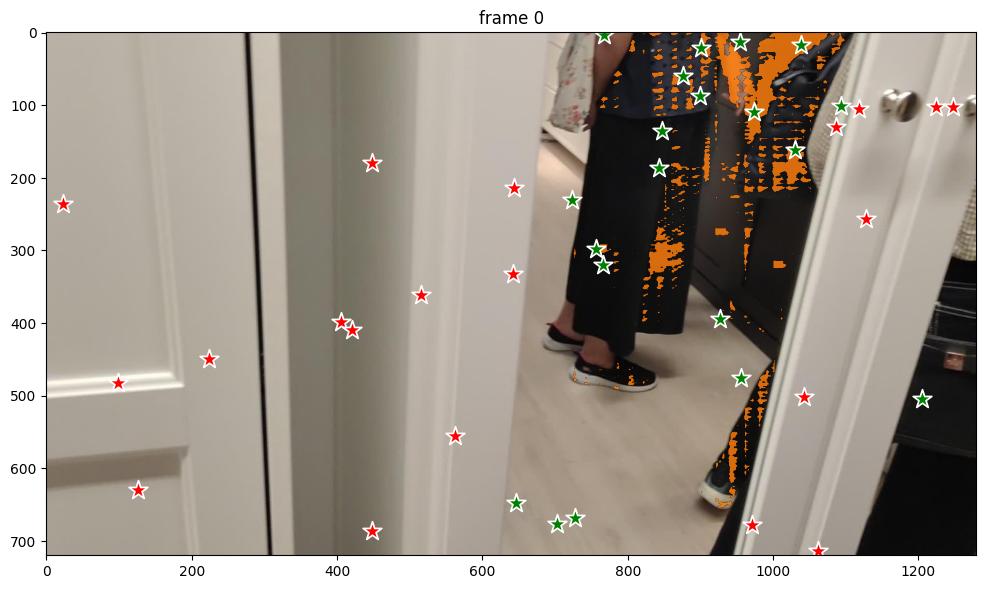}
		\captionsetup{justification=centering}
        \vspace{-5.5mm} \caption{\footnotesize{\\20 \\points}}
	\end{subfigure}
	\begin{subfigure}{0.12\textwidth}
		\includegraphics[width=\textwidth]{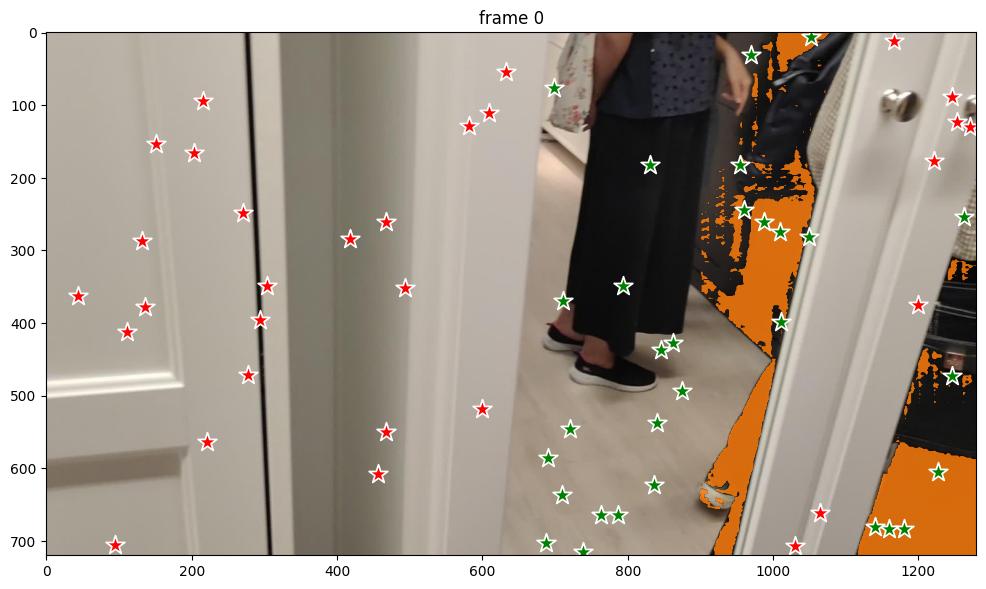}
		\captionsetup{justification=centering}
        \vspace{-5.5mm} \caption{\footnotesize{\\30 \\points}}
	\end{subfigure}
	\begin{subfigure}{0.12\textwidth}
		\includegraphics[width=\textwidth]{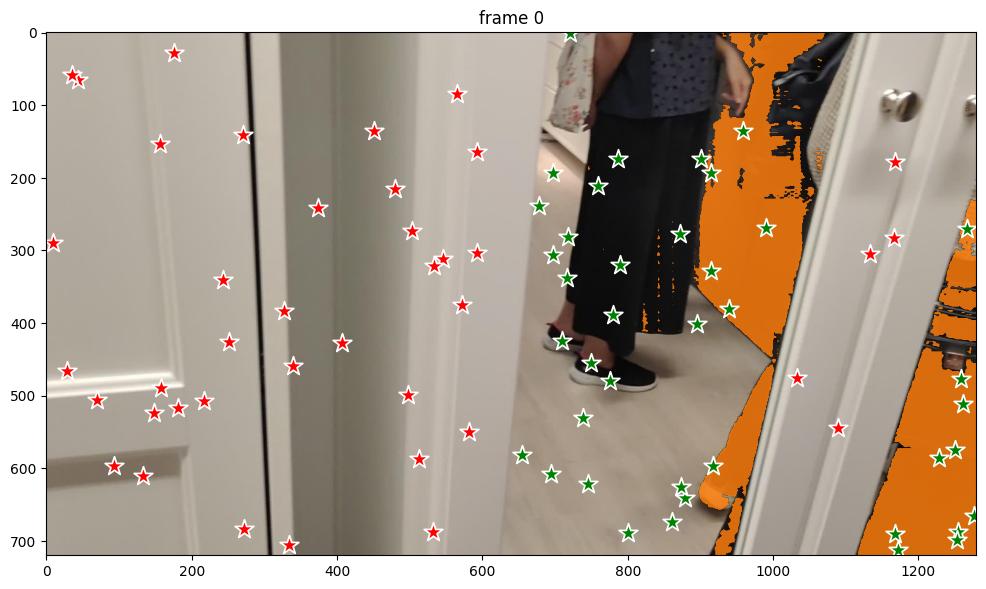}
		\captionsetup{justification=centering}
        \vspace{-5.5mm} \caption{\footnotesize{\\40 \\points}}
	\end{subfigure}
	\begin{subfigure}{0.12\textwidth}
		\includegraphics[width=\textwidth]{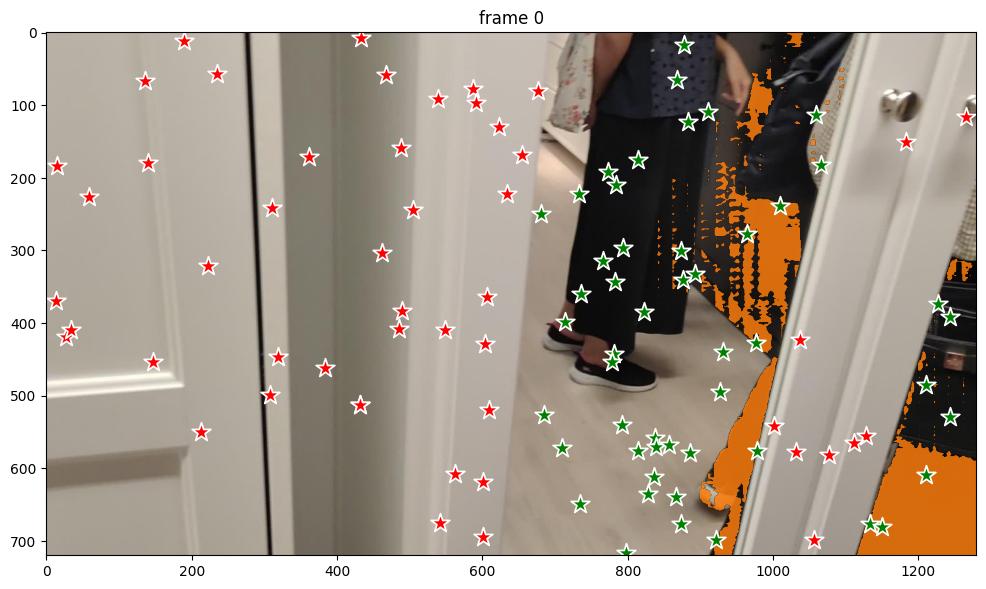}
		\captionsetup{justification=centering}
        \vspace{-5.5mm} \caption{\footnotesize{\\50 \\points}}
	\end{subfigure}

	\caption{Visual effects of masks (organe areas) generated by varying number of point prompts using SAM2. The first two columns show the input images and the ground truths. The remaining columns show the initial segmentation results (the 1st frame in a video) with different numbers of point prompts. The number of point prompts is indicated at the bottom of each column. We use green and red five-pointed stars to represent the positive and negative predictions, respectively. Best viewed on screen.}
	\label{fig:fig_pts}
	
\end{figure*}

\begin{figure*}[!ht]
	\centering
	\vspace*{1.3mm}
	\begin{subfigure}{0.087\textwidth}
		\includegraphics[width=\textwidth]{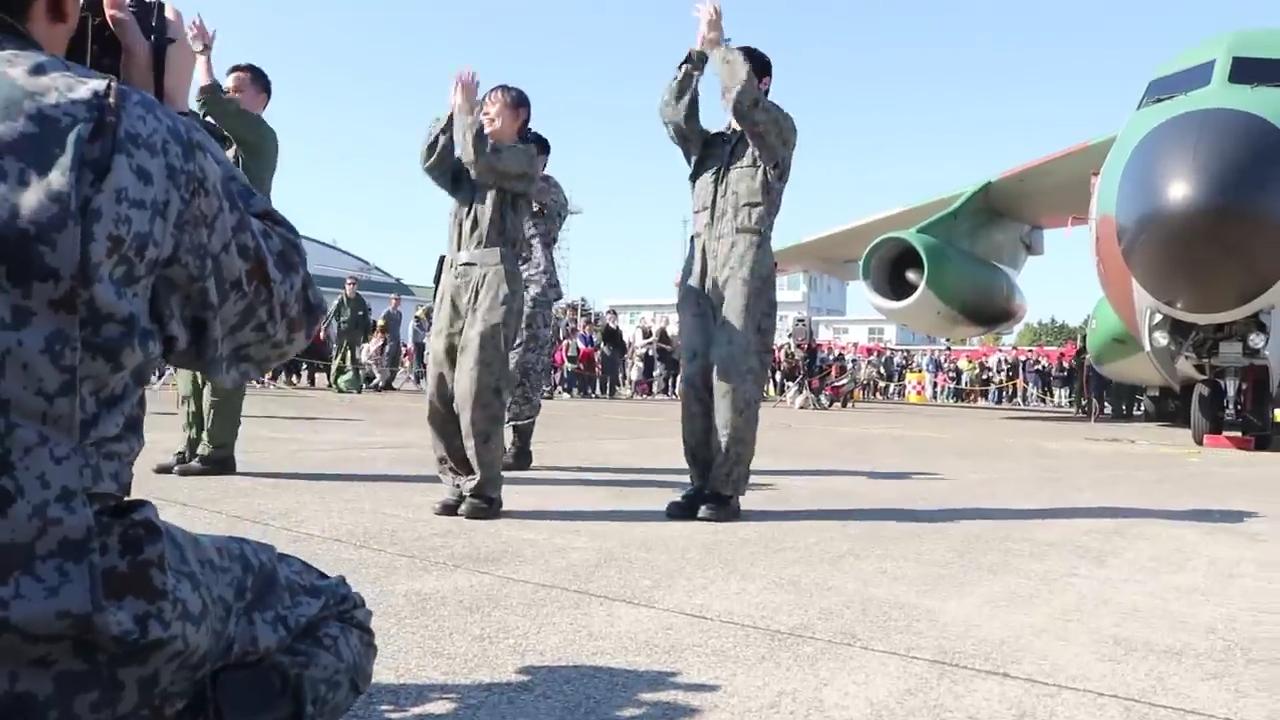}
	\end{subfigure}
	\begin{subfigure}{0.087\textwidth}
		\includegraphics[width=\textwidth]{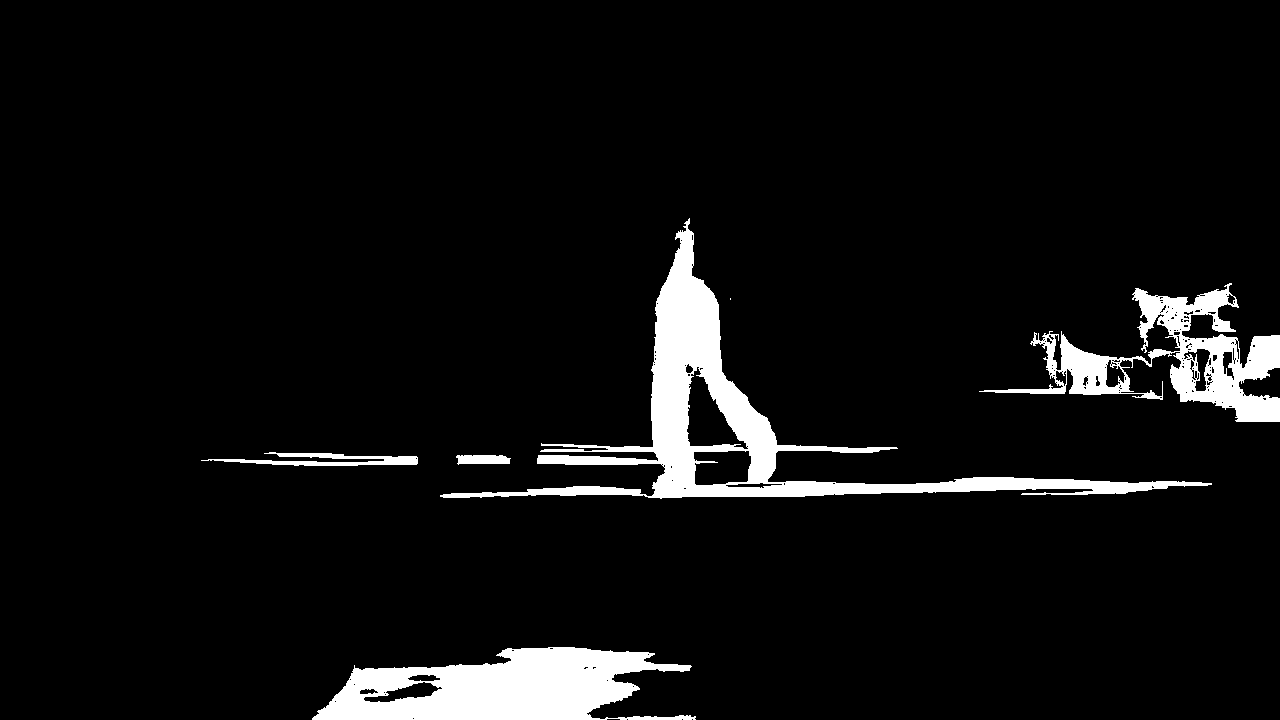}
	\end{subfigure}
	\begin{subfigure}{0.087\textwidth}
		\includegraphics[width=\textwidth]{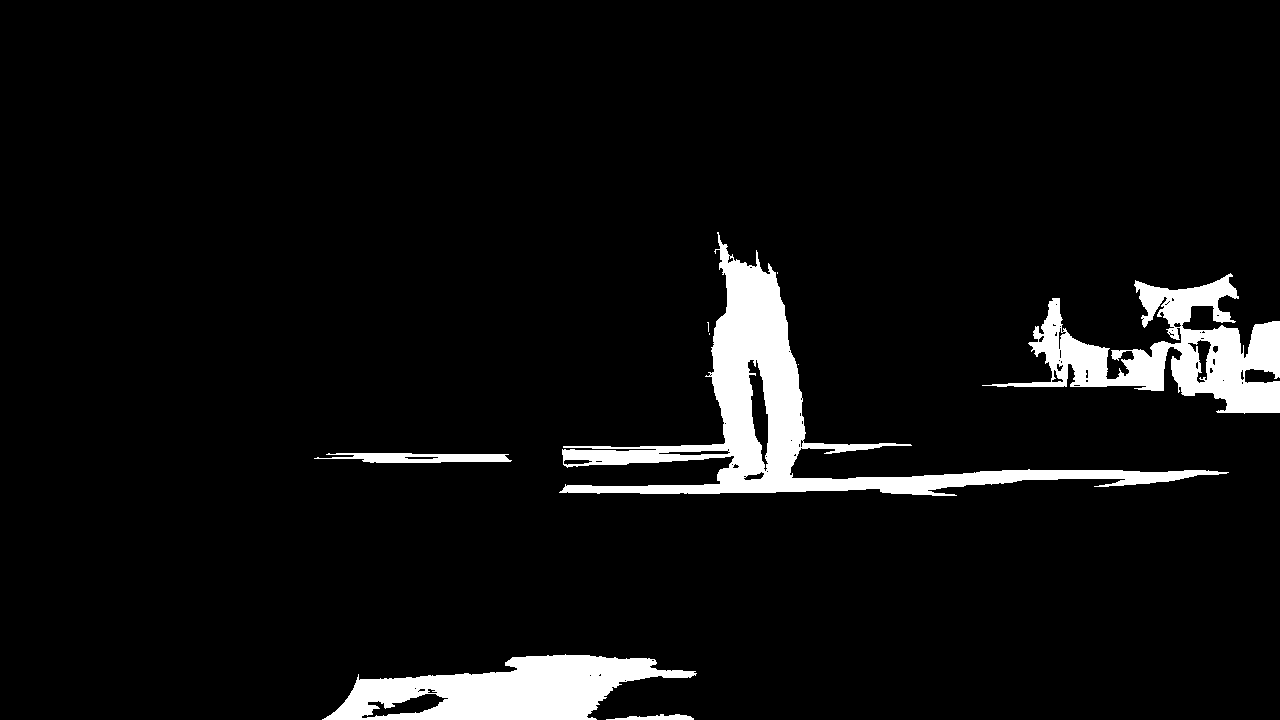}
	\end{subfigure}
	\begin{subfigure}{0.087\textwidth}
		\includegraphics[width=\textwidth]{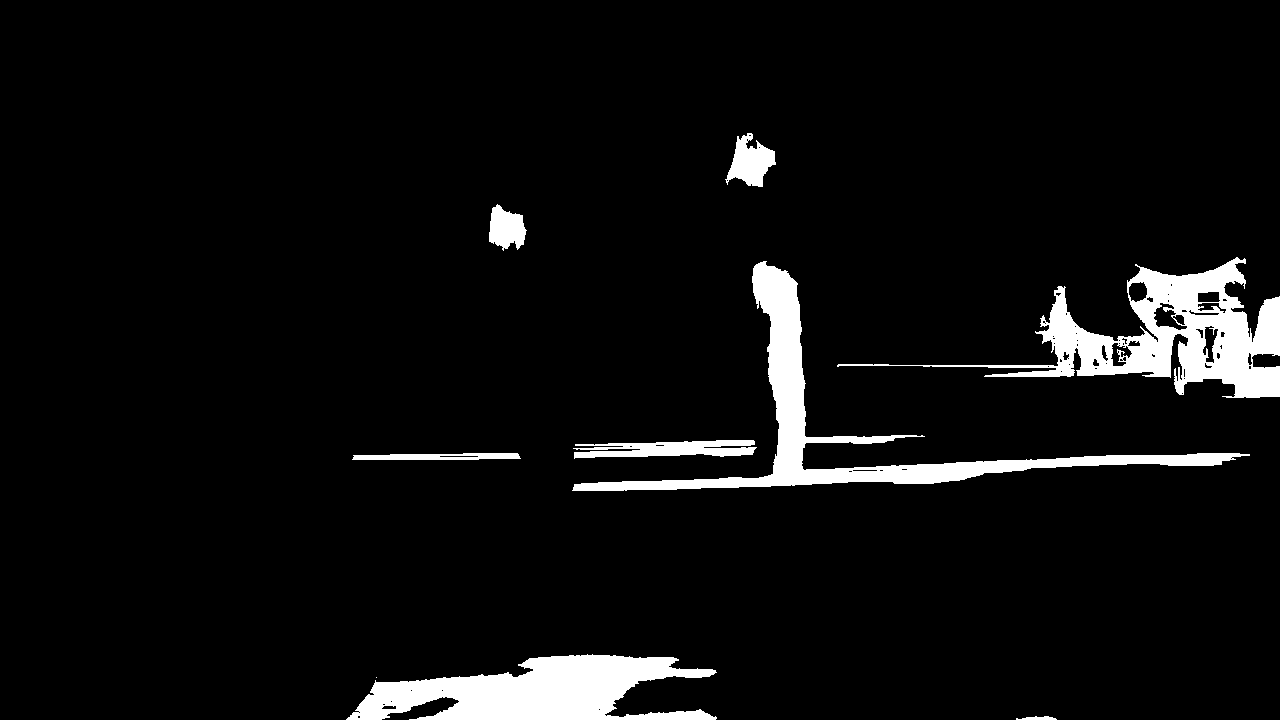}
	\end{subfigure}
	\begin{subfigure}{0.087\textwidth}
		\includegraphics[width=\textwidth]{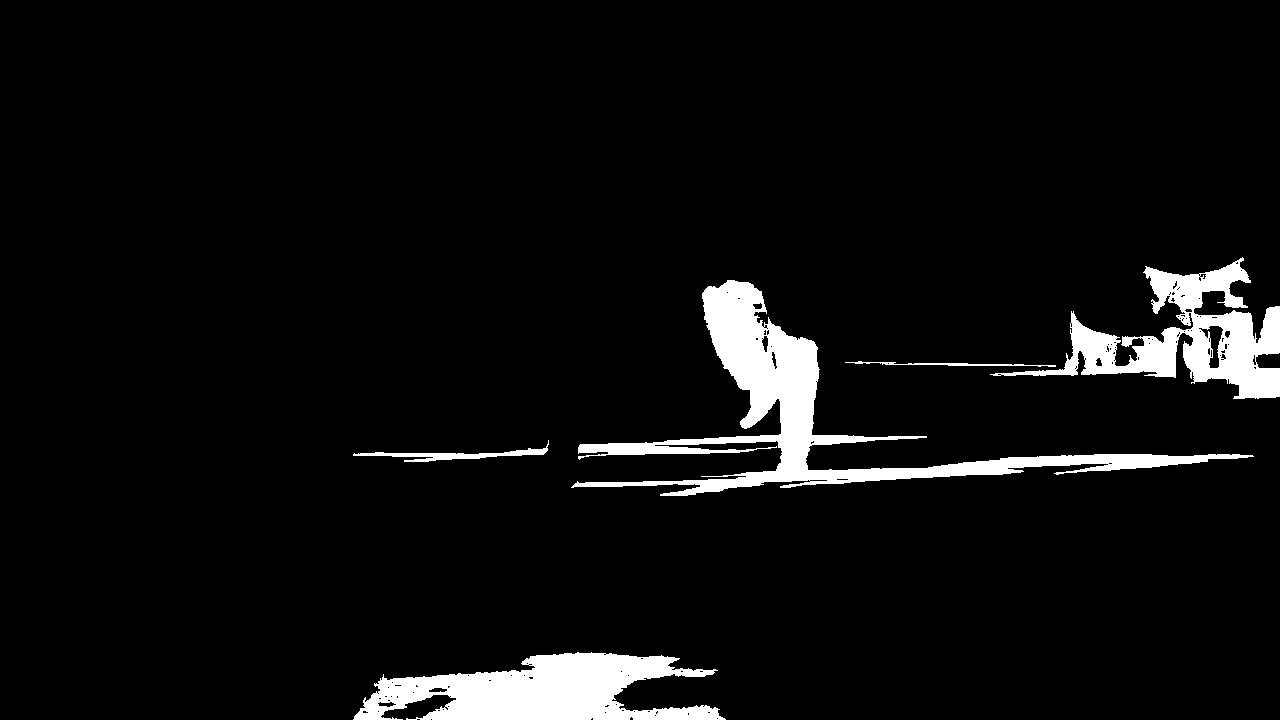}
	\end{subfigure}
	\begin{subfigure}{0.087\textwidth}
		\includegraphics[width=\textwidth]{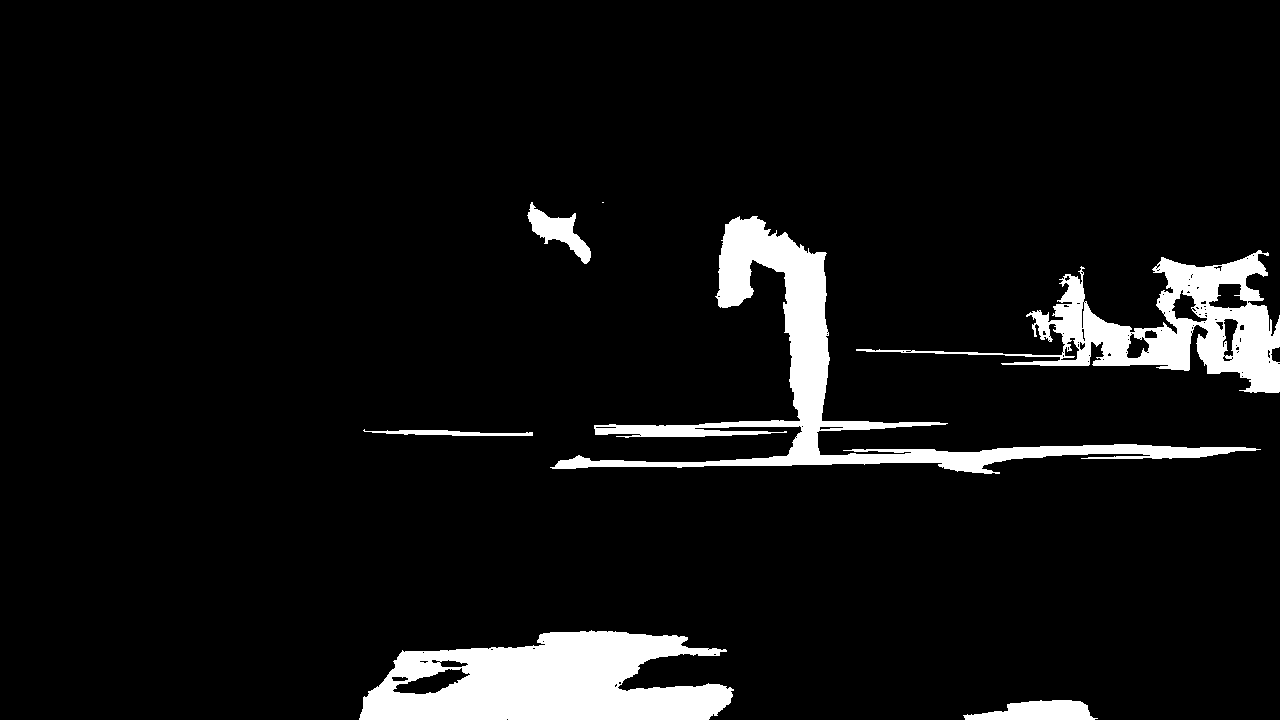}
	\end{subfigure}
	\begin{subfigure}{0.087\textwidth}
		\includegraphics[width=\textwidth]{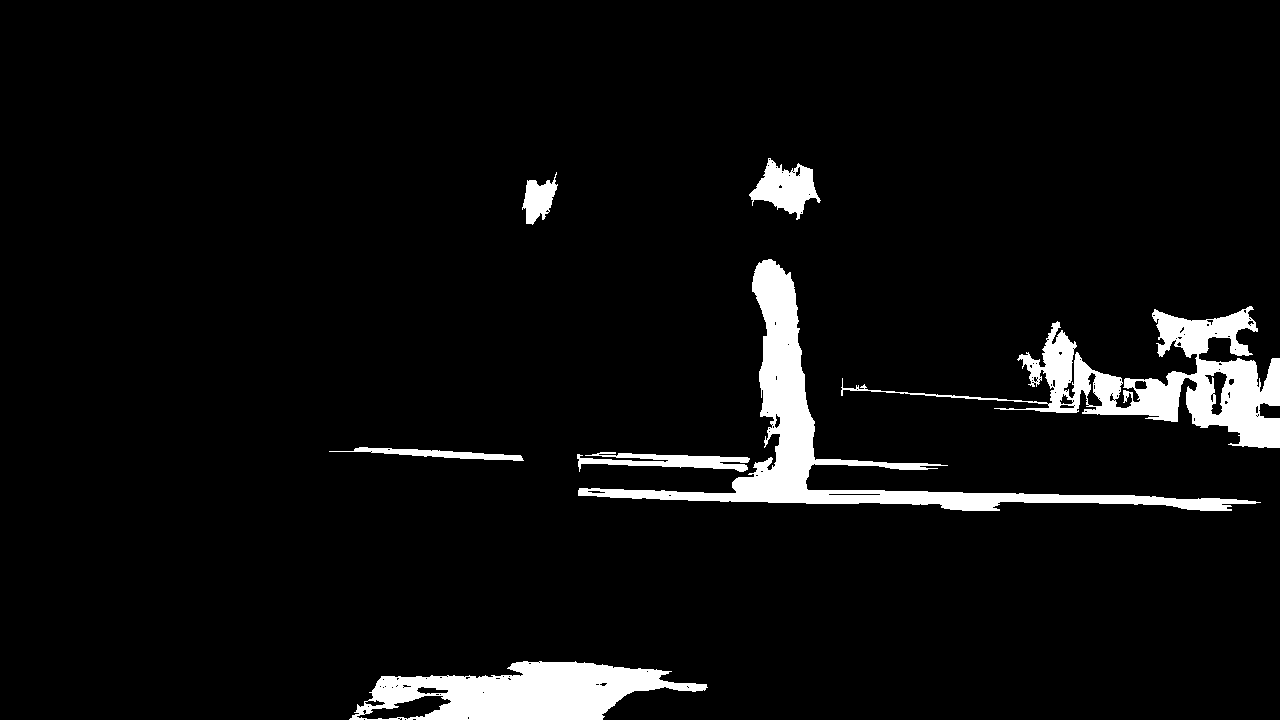}
	\end{subfigure}
	\begin{subfigure}{0.087\textwidth}
		\includegraphics[width=\textwidth]{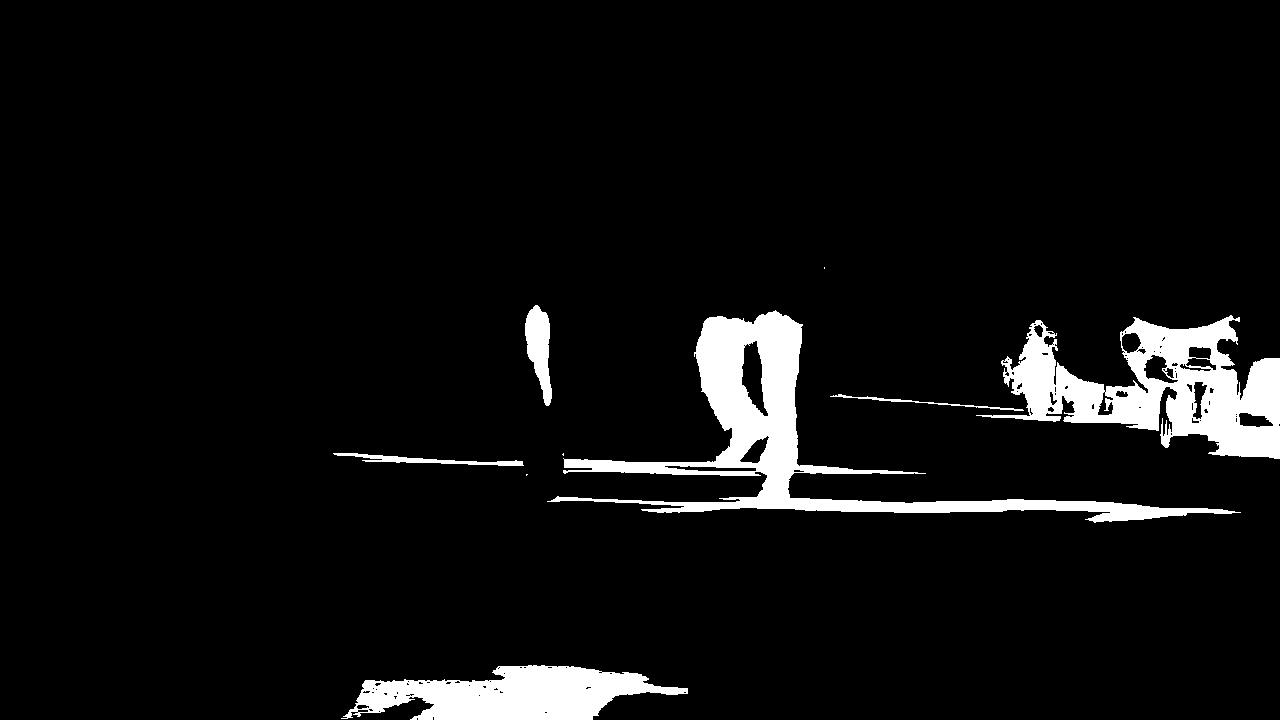}
	\end{subfigure}
	\begin{subfigure}{0.087\textwidth}
		\includegraphics[width=\textwidth]{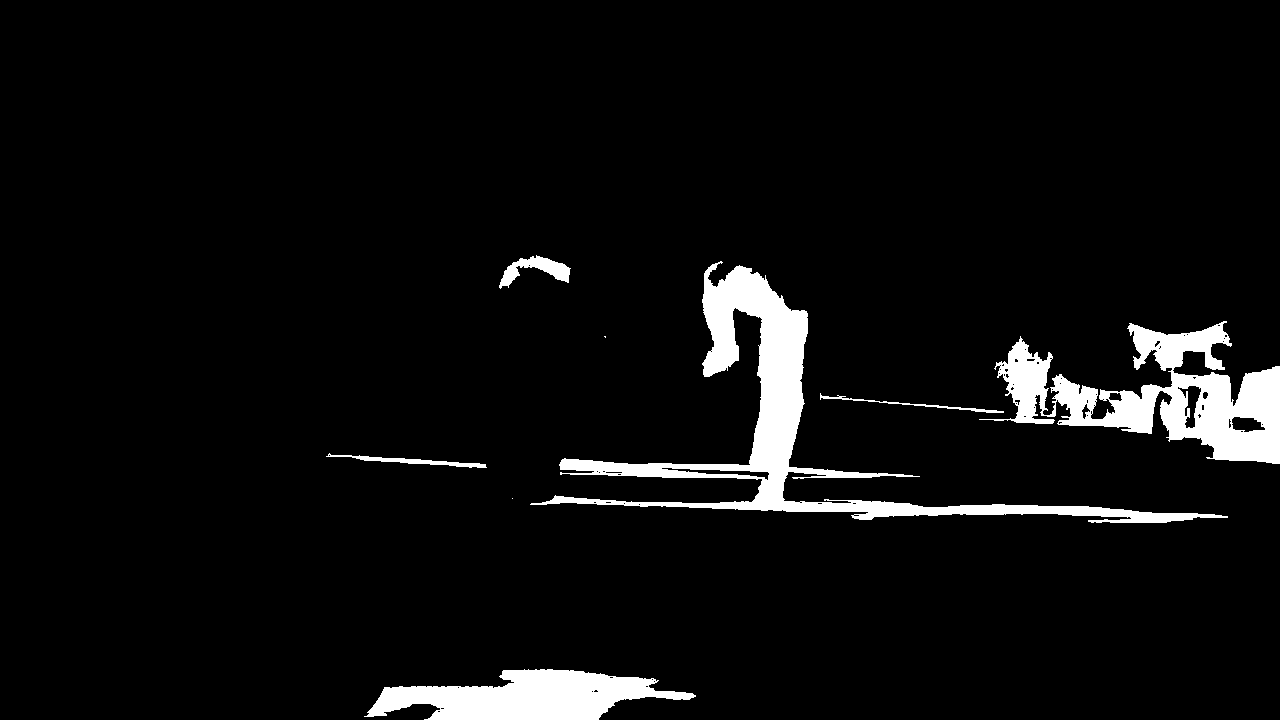}
	\end{subfigure}
	\begin{subfigure}{0.087\textwidth}
		\includegraphics[width=\textwidth]{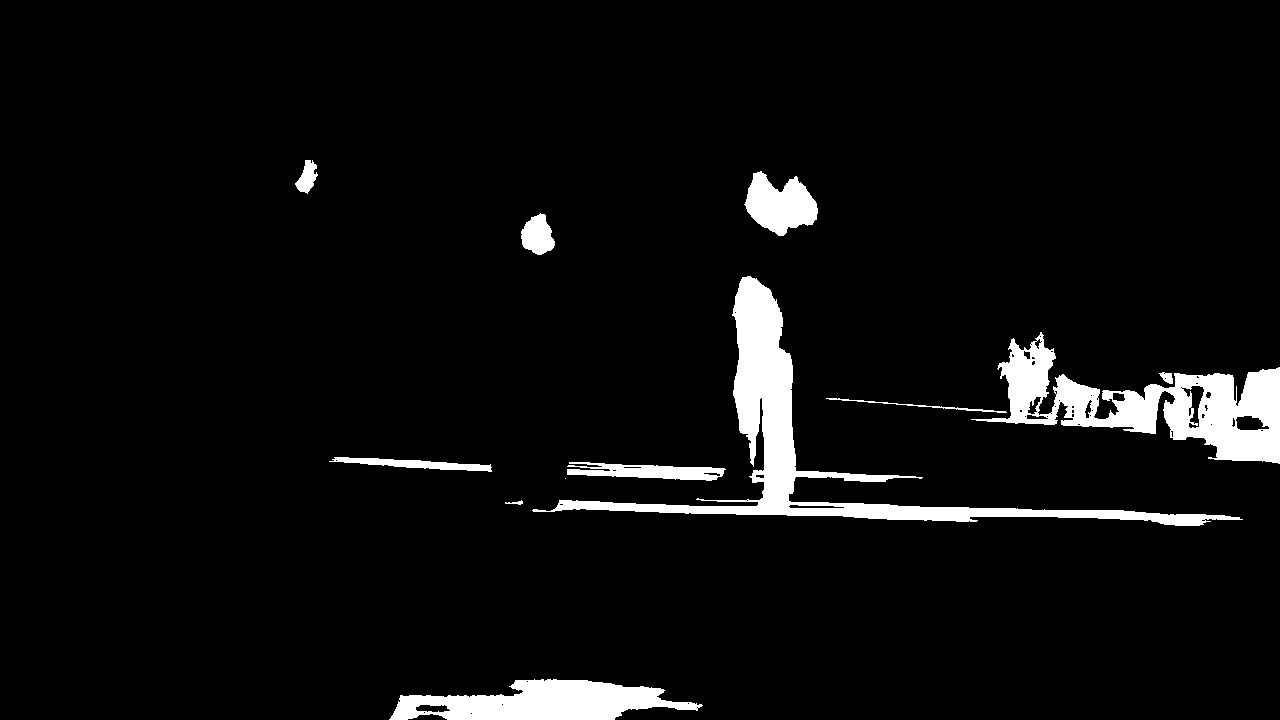}
	\end{subfigure}

	\vspace*{1.3mm}
	\begin{subfigure}{0.087\textwidth}
		\includegraphics[width=\textwidth]{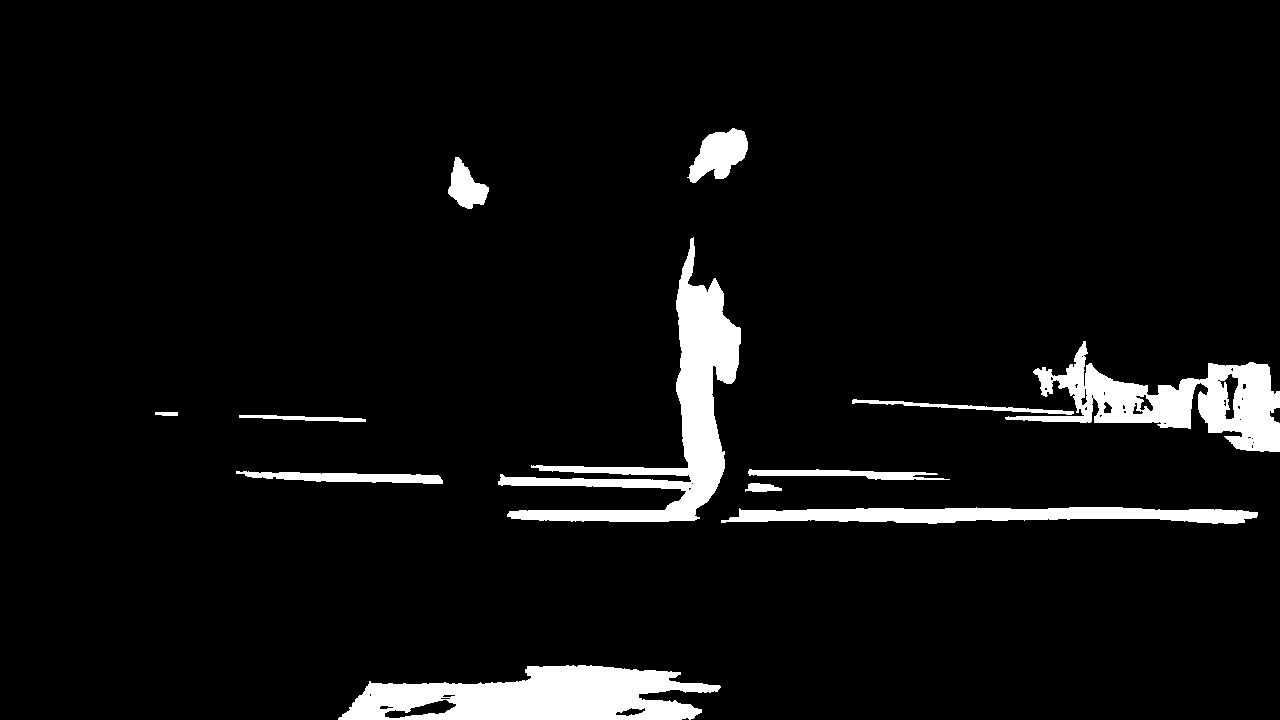}
	\end{subfigure}
	\begin{subfigure}{0.087\textwidth}
		\includegraphics[width=\textwidth]{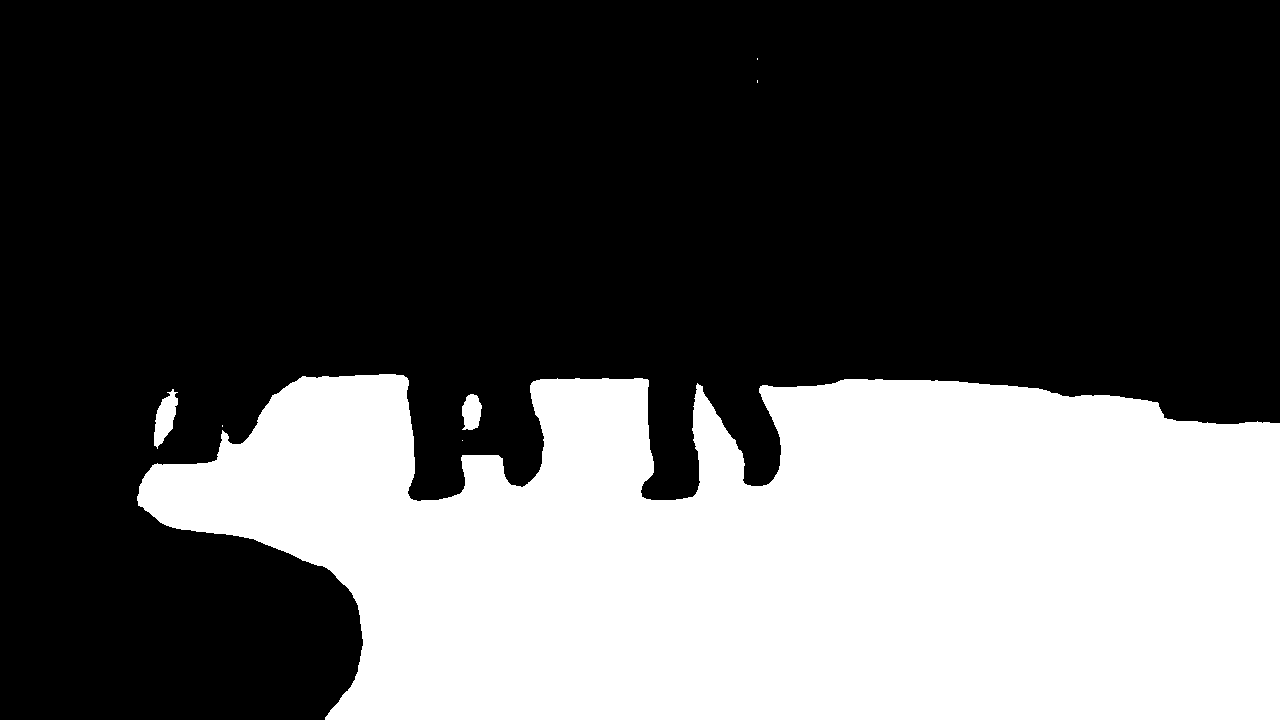}
	\end{subfigure}
	\begin{subfigure}{0.087\textwidth}
		\includegraphics[width=\textwidth]{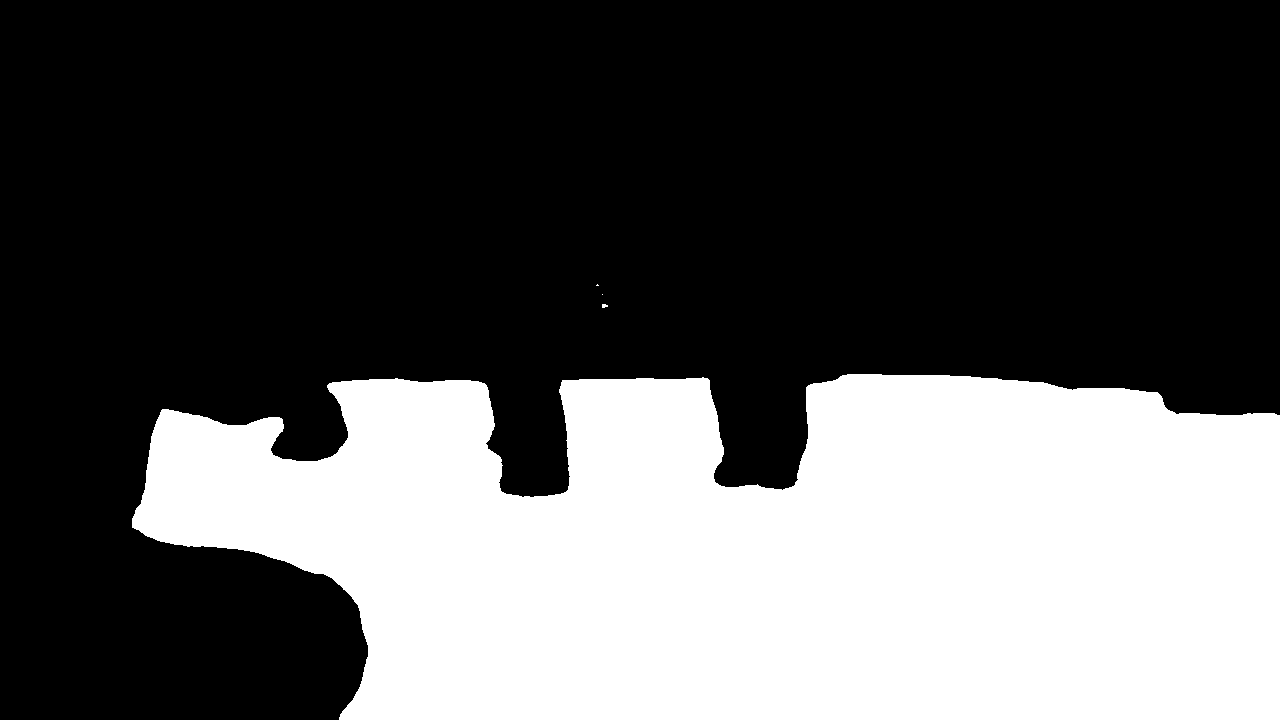}
	\end{subfigure}
	\begin{subfigure}{0.087\textwidth}
		\includegraphics[width=\textwidth]{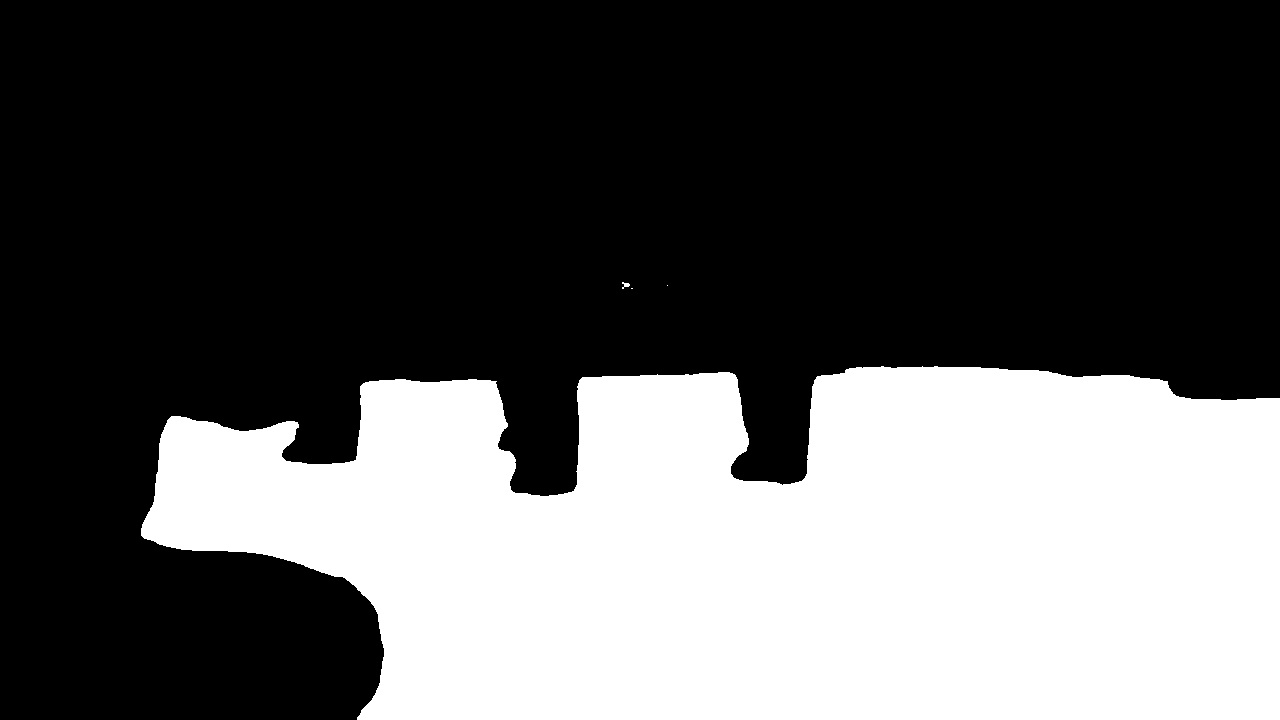}
	\end{subfigure}
	\begin{subfigure}{0.087\textwidth}
		\includegraphics[width=\textwidth]{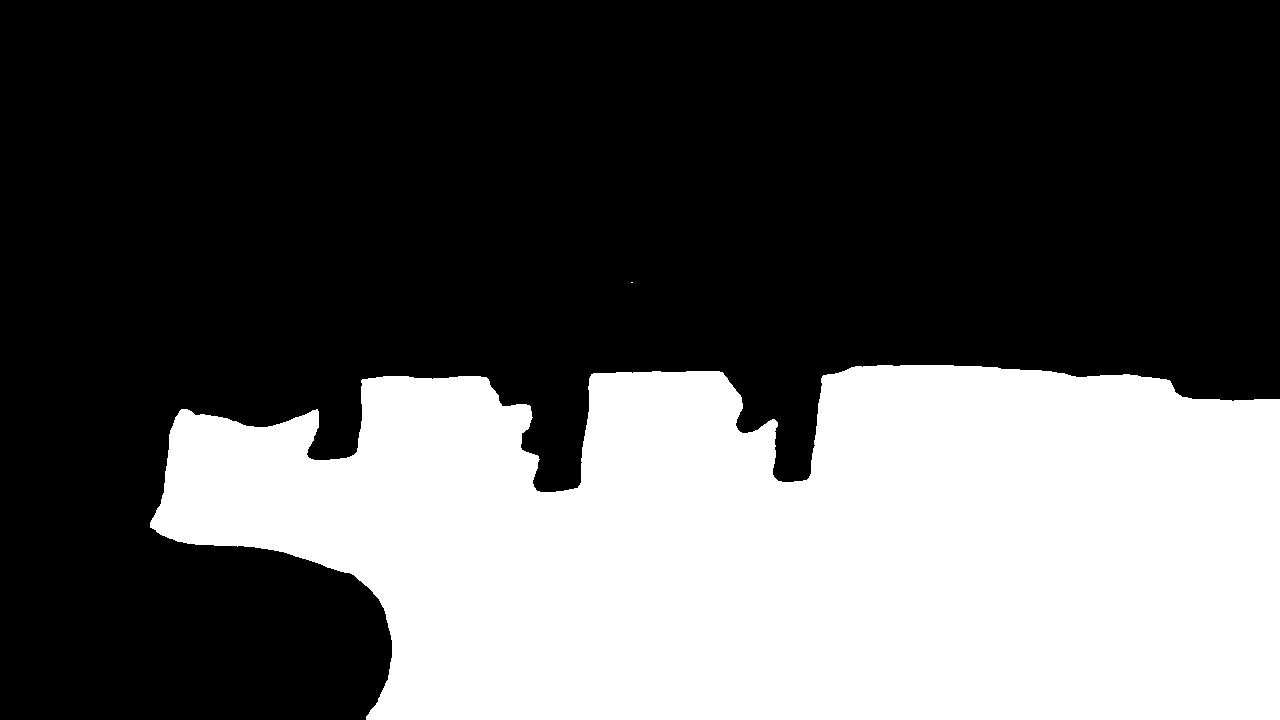}
	\end{subfigure}
	\begin{subfigure}{0.087\textwidth}
		\includegraphics[width=\textwidth]{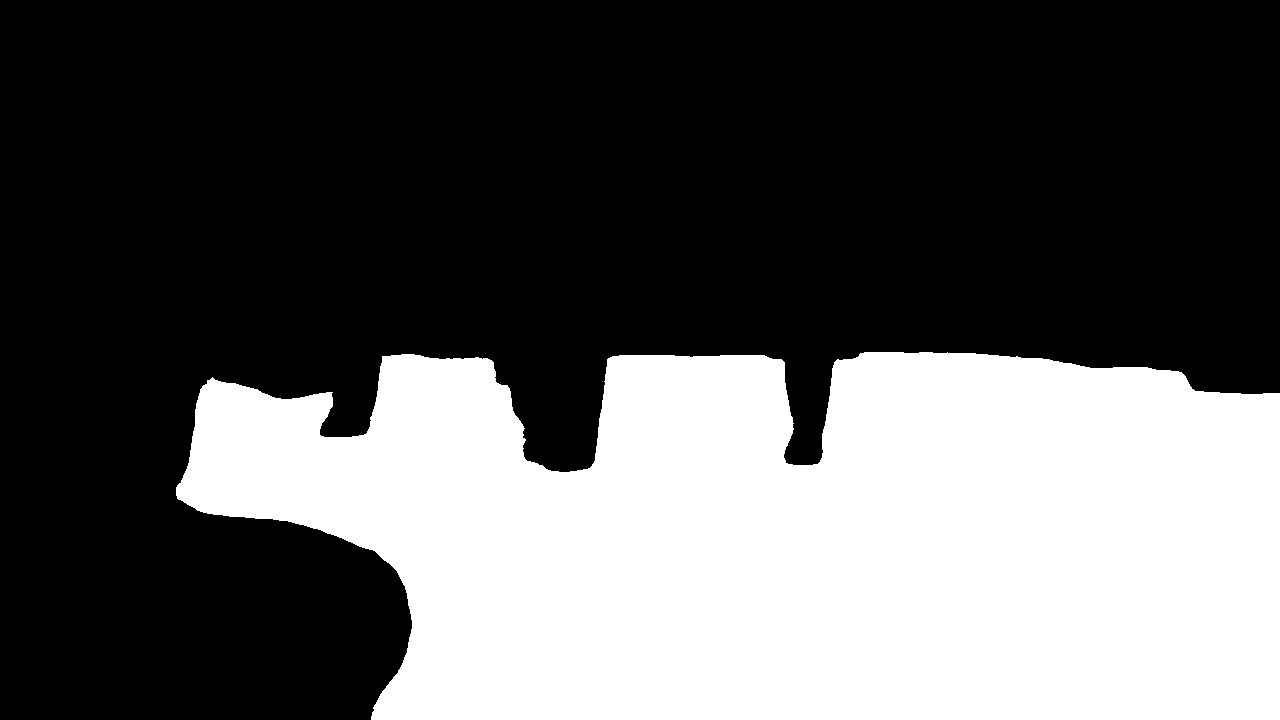}
	\end{subfigure}
	\begin{subfigure}{0.087\textwidth}
		\includegraphics[width=\textwidth]{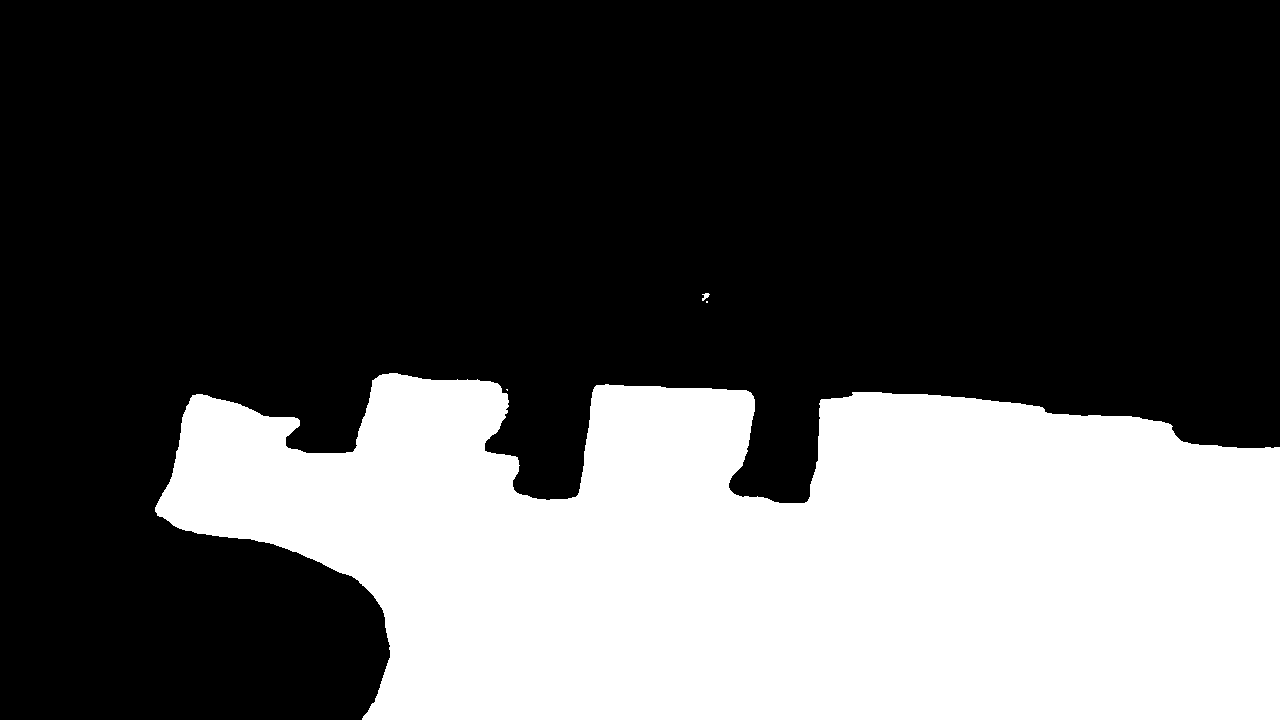}
	\end{subfigure}
	\begin{subfigure}{0.087\textwidth}
		\includegraphics[width=\textwidth]{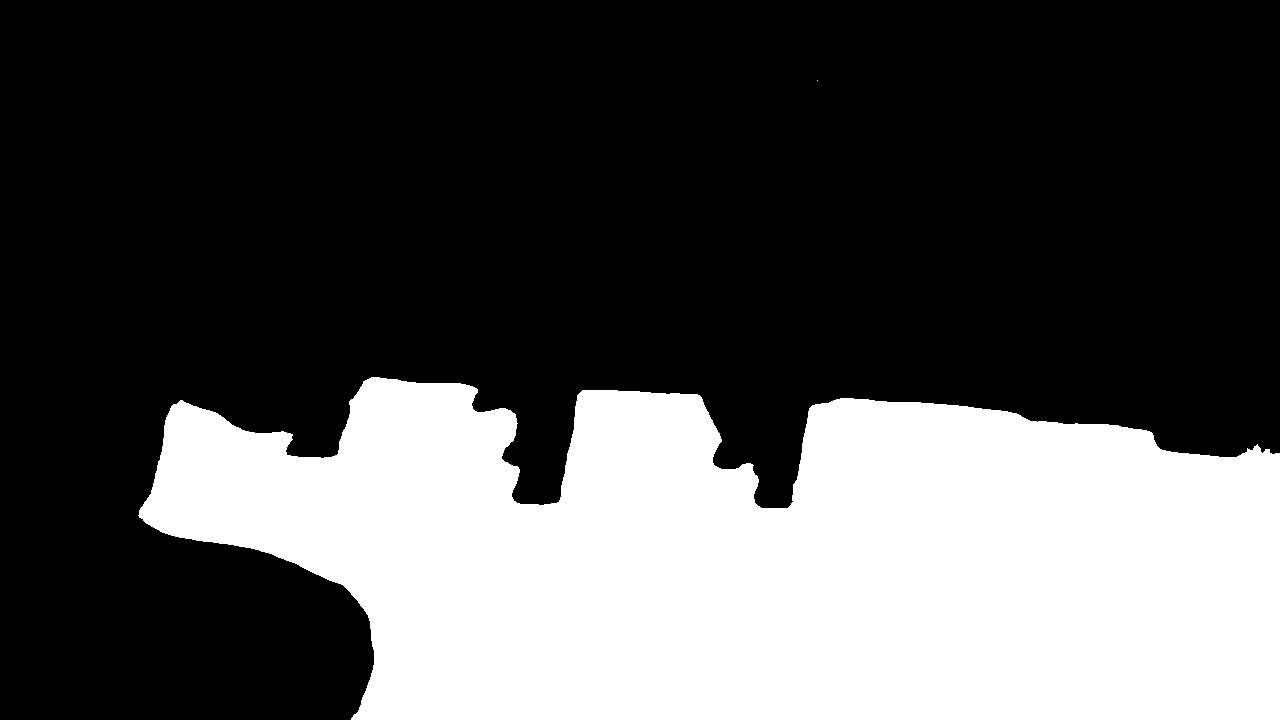}
	\end{subfigure}
	\begin{subfigure}{0.087\textwidth}
		\includegraphics[width=\textwidth]{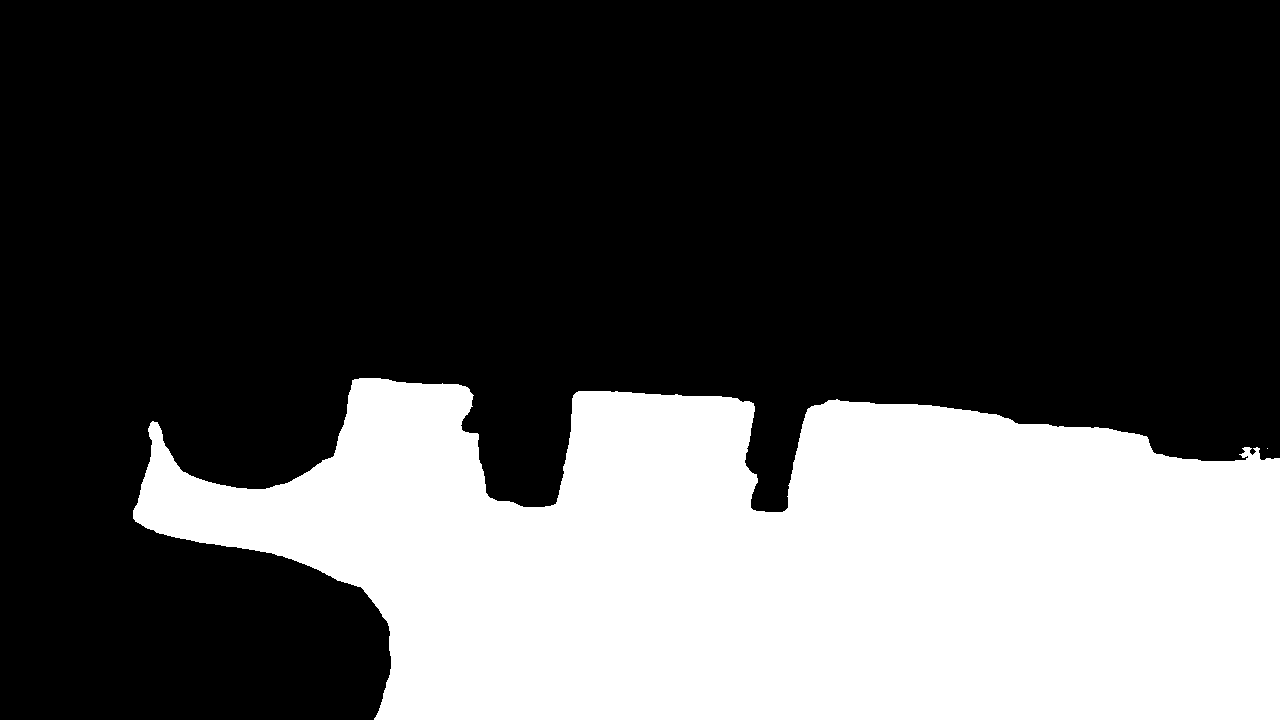}
	\end{subfigure}
	\begin{subfigure}{0.087\textwidth}
		\includegraphics[width=\textwidth]{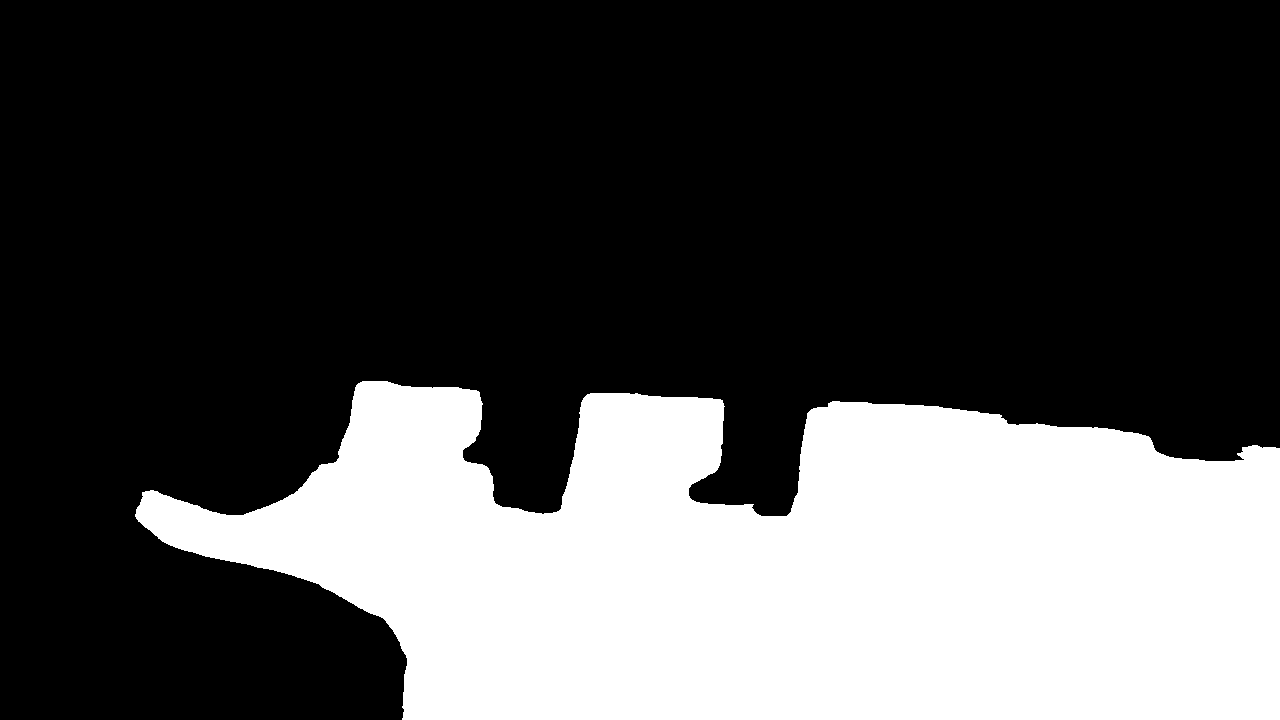}
	\end{subfigure}

	\vspace*{1.3mm}
	\begin{subfigure}{0.087\textwidth}
		\includegraphics[width=\textwidth]{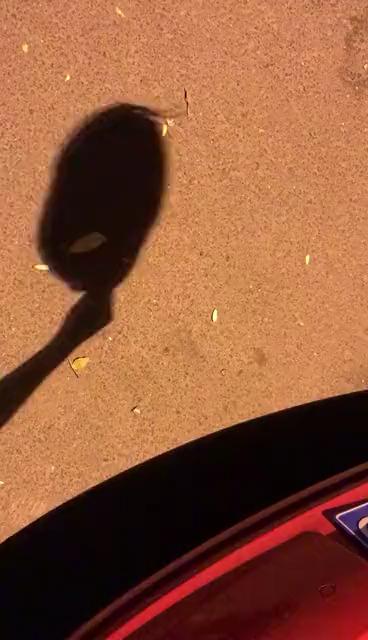}
	\end{subfigure}
	\begin{subfigure}{0.087\textwidth}
		\includegraphics[width=\textwidth]{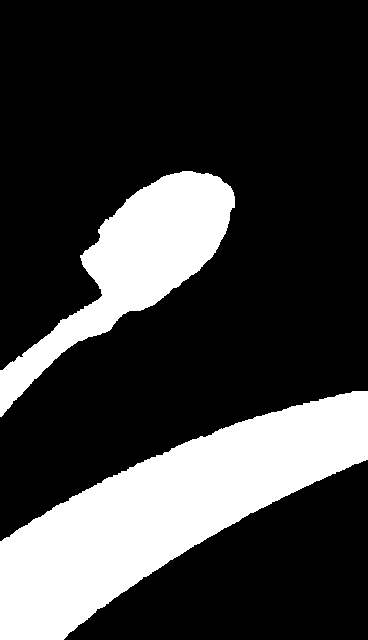}
	\end{subfigure}
	\begin{subfigure}{0.087\textwidth}
		\includegraphics[width=\textwidth]{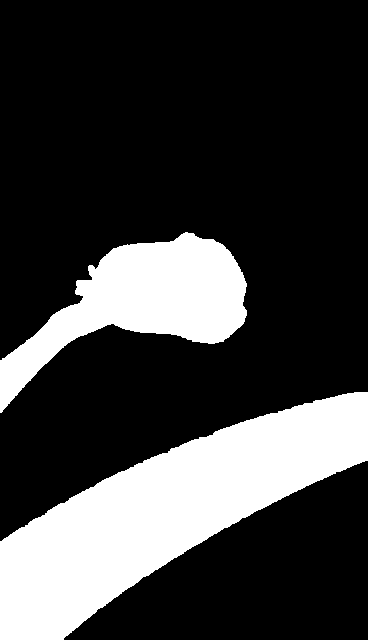}
	\end{subfigure}
	\begin{subfigure}{0.087\textwidth}
		\includegraphics[width=\textwidth]{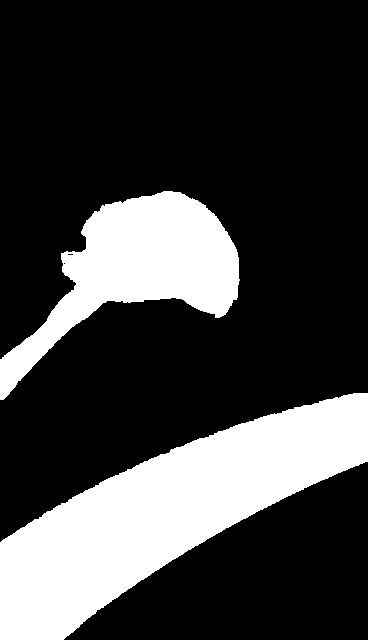}
	\end{subfigure}
	\begin{subfigure}{0.087\textwidth}
		\includegraphics[width=\textwidth]{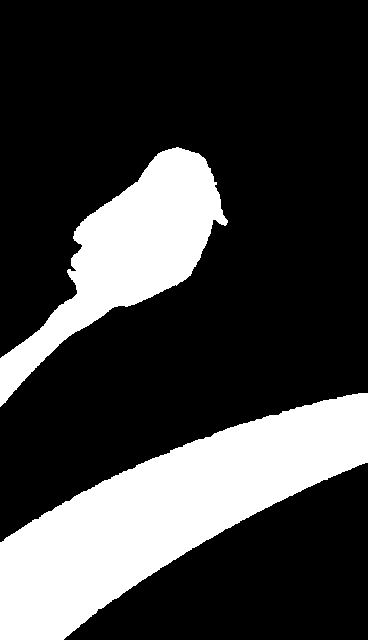}
	\end{subfigure}
	\begin{subfigure}{0.087\textwidth}
		\includegraphics[width=\textwidth]{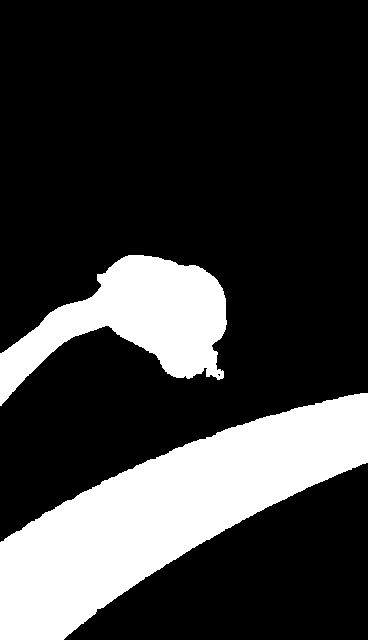}
	\end{subfigure}
	\begin{subfigure}{0.087\textwidth}
		\includegraphics[width=\textwidth]{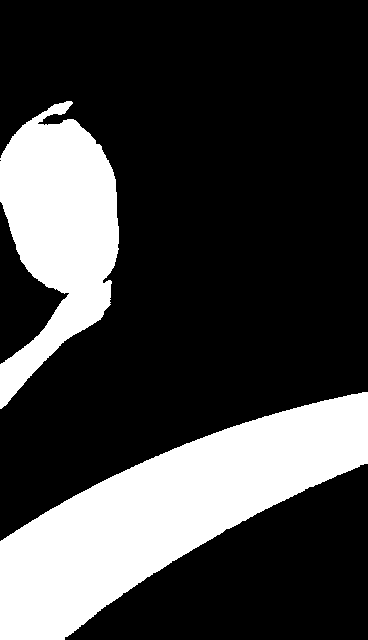}
	\end{subfigure}
	\begin{subfigure}{0.087\textwidth}
		\includegraphics[width=\textwidth]{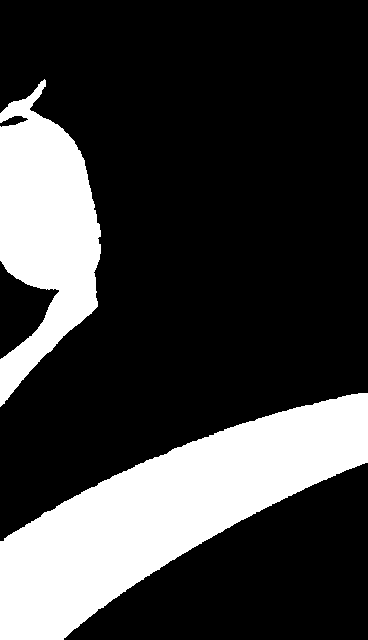}
	\end{subfigure}
	\begin{subfigure}{0.087\textwidth}
		\includegraphics[width=\textwidth]{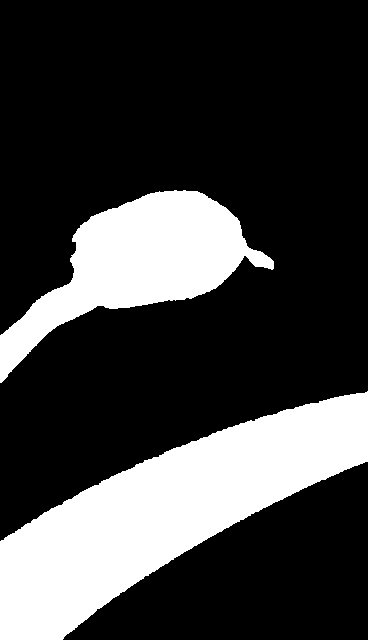}
	\end{subfigure}
	\begin{subfigure}{0.087\textwidth}
		\includegraphics[width=\textwidth]{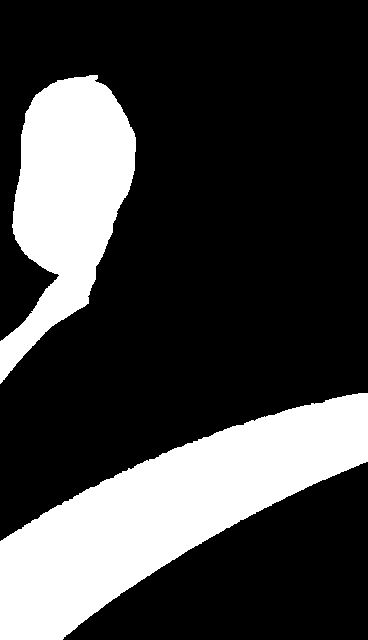}
	\end{subfigure}
	
	\vspace*{1.3mm}
	\begin{subfigure}{0.087\textwidth}
		\includegraphics[width=\textwidth]{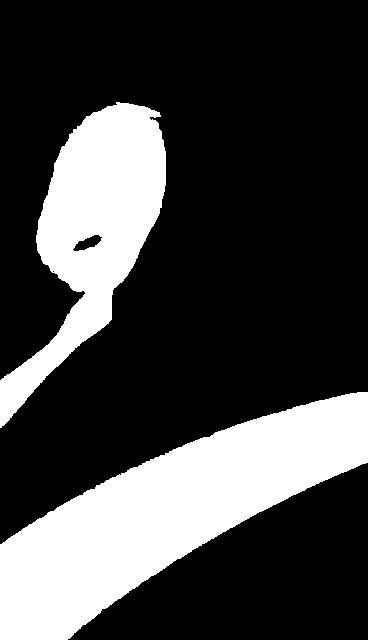}
	\end{subfigure}
	\begin{subfigure}{0.087\textwidth}
		\includegraphics[width=\textwidth]{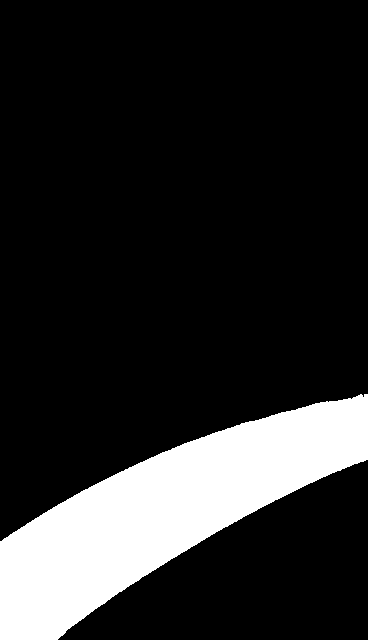}
	\end{subfigure}
	\begin{subfigure}{0.087\textwidth}
		\includegraphics[width=\textwidth]{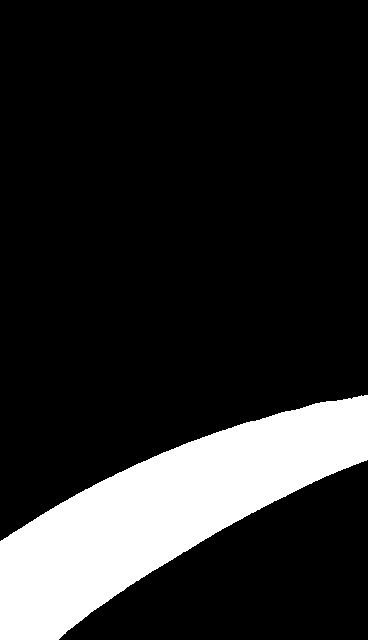}
	\end{subfigure}
	\begin{subfigure}{0.087\textwidth}
		\includegraphics[width=\textwidth]{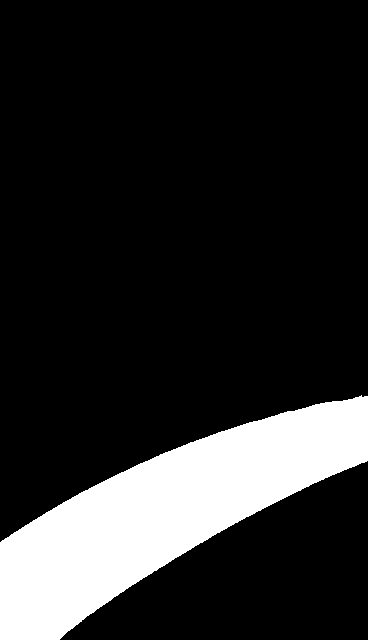}
	\end{subfigure}
	\begin{subfigure}{0.087\textwidth}
		\includegraphics[width=\textwidth]{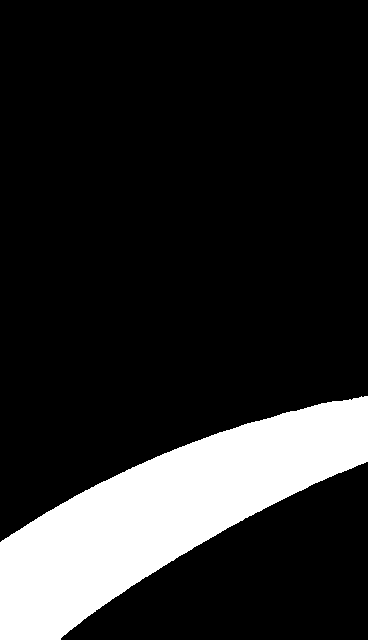}
	\end{subfigure}
	\begin{subfigure}{0.087\textwidth}
		\includegraphics[width=\textwidth]{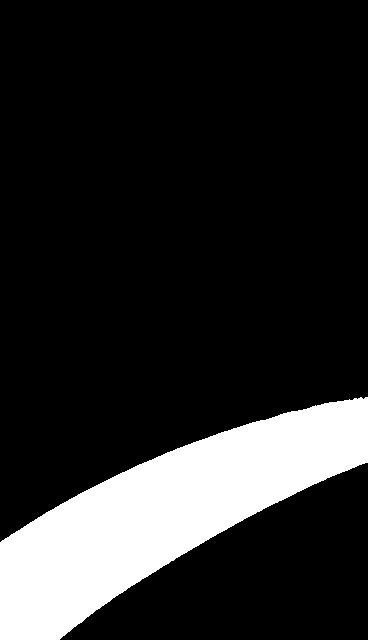}
	\end{subfigure}
	\begin{subfigure}{0.087\textwidth}
		\includegraphics[width=\textwidth]{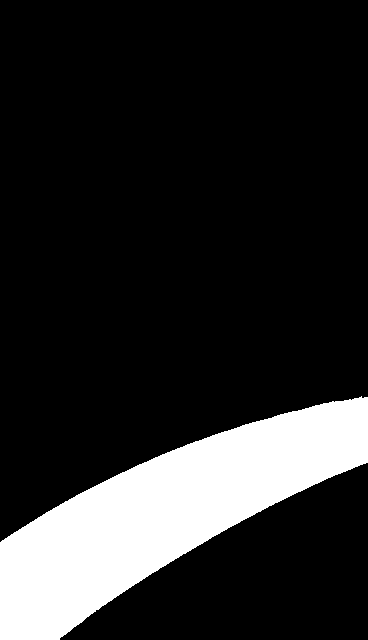}
	\end{subfigure}
	\begin{subfigure}{0.087\textwidth}
		\includegraphics[width=\textwidth]{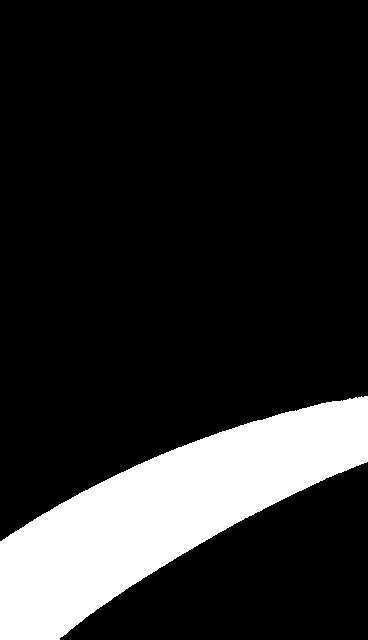}
	\end{subfigure}
	\begin{subfigure}{0.087\textwidth}
		\includegraphics[width=\textwidth]{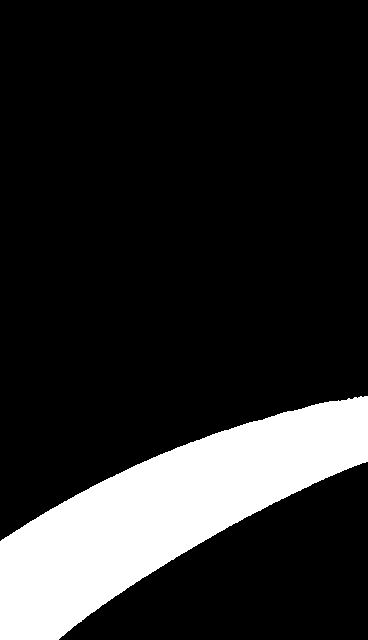}
	\end{subfigure}
	\begin{subfigure}{0.087\textwidth}
		\includegraphics[width=\textwidth]{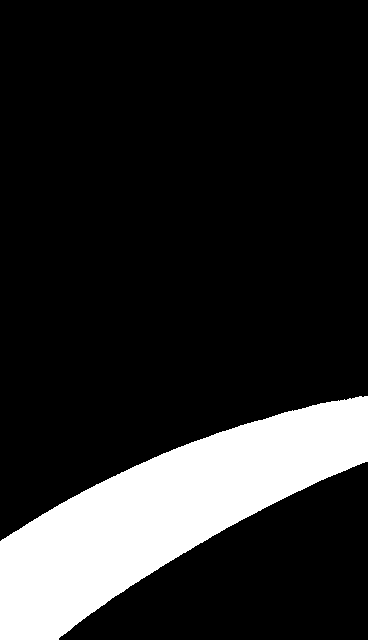}
	\end{subfigure}

	\vspace*{1.3mm}
	\begin{subfigure}{0.087\textwidth}
		\includegraphics[width=\textwidth]{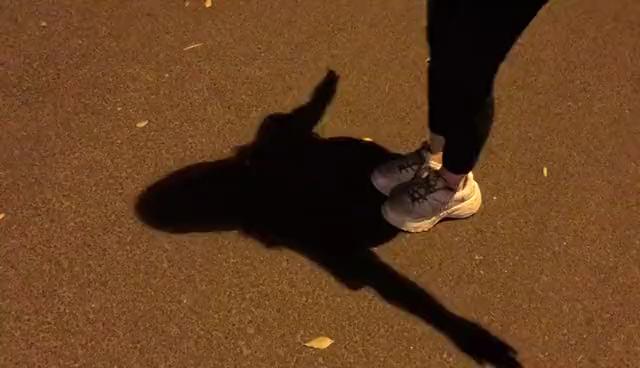}
	\end{subfigure}
	\begin{subfigure}{0.087\textwidth}
		\includegraphics[width=\textwidth]{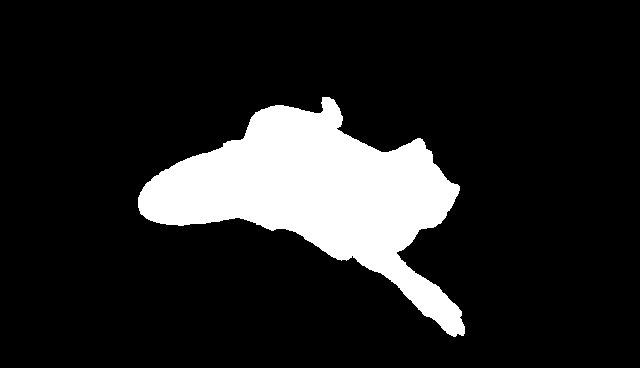}
	\end{subfigure}
	\begin{subfigure}{0.087\textwidth}
		\includegraphics[width=\textwidth]{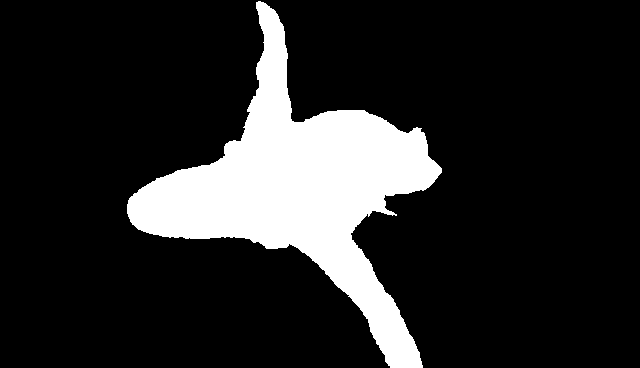}
	\end{subfigure}
	\begin{subfigure}{0.087\textwidth}
		\includegraphics[width=\textwidth]{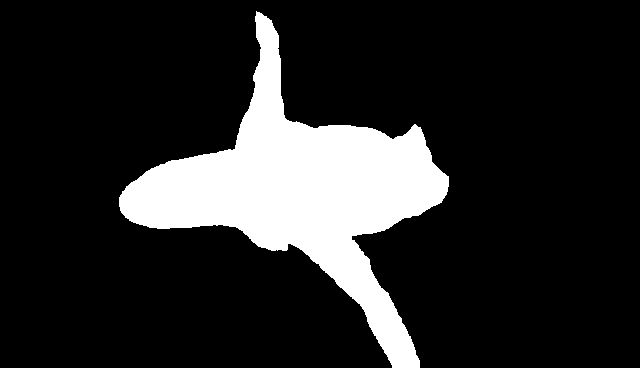}
	\end{subfigure}
	\begin{subfigure}{0.087\textwidth}
		\includegraphics[width=\textwidth]{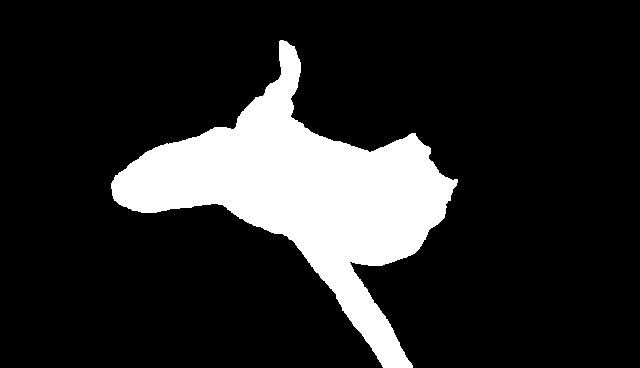}
	\end{subfigure}
	\begin{subfigure}{0.087\textwidth}
		\includegraphics[width=\textwidth]{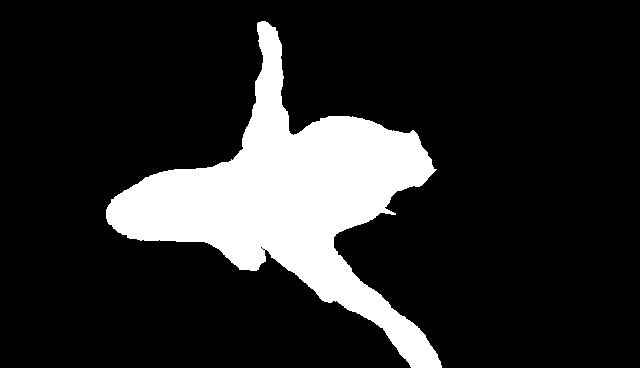}
	\end{subfigure}
	\begin{subfigure}{0.087\textwidth}
		\includegraphics[width=\textwidth]{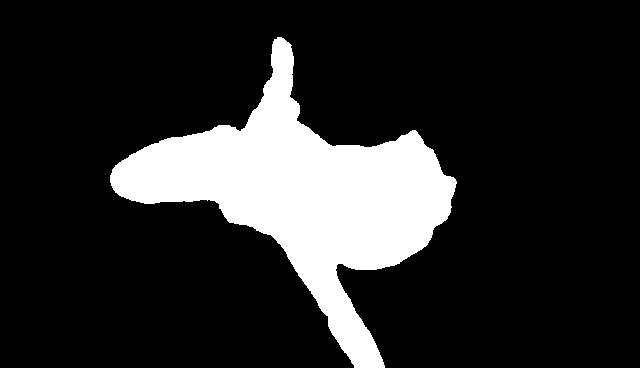}
	\end{subfigure}
	\begin{subfigure}{0.087\textwidth}
		\includegraphics[width=\textwidth]{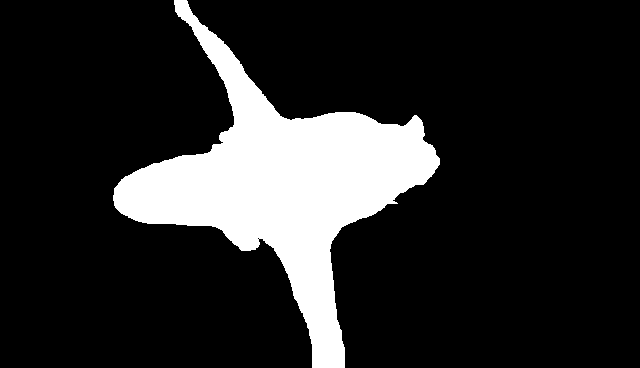}
	\end{subfigure}
	\begin{subfigure}{0.087\textwidth}
		\includegraphics[width=\textwidth]{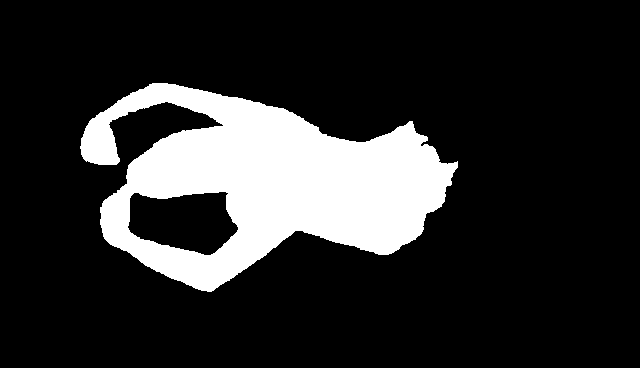}
	\end{subfigure}
	\begin{subfigure}{0.087\textwidth}
		\includegraphics[width=\textwidth]{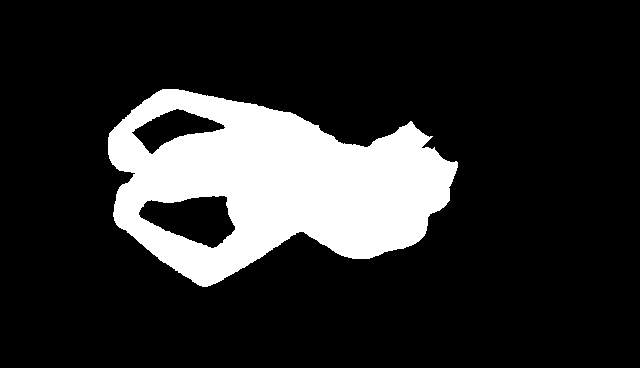}
	\end{subfigure}
	
	\vspace*{1.3mm}
	\begin{subfigure}{0.087\textwidth}
		\includegraphics[width=\textwidth]{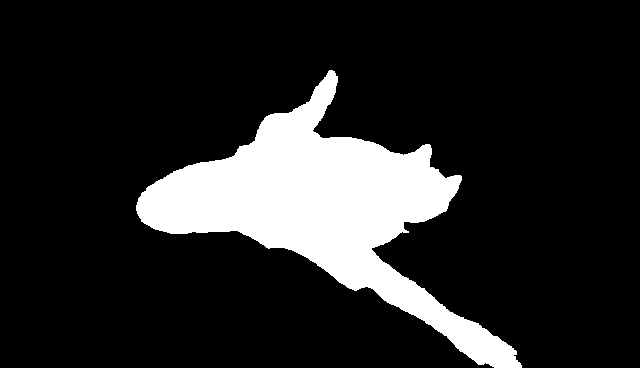}
	\end{subfigure}
	\begin{subfigure}{0.087\textwidth}
		\includegraphics[width=\textwidth]{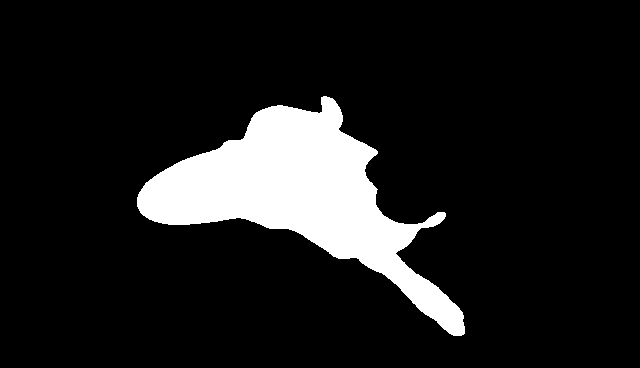}
	\end{subfigure}
	\begin{subfigure}{0.087\textwidth}
		\includegraphics[width=\textwidth]{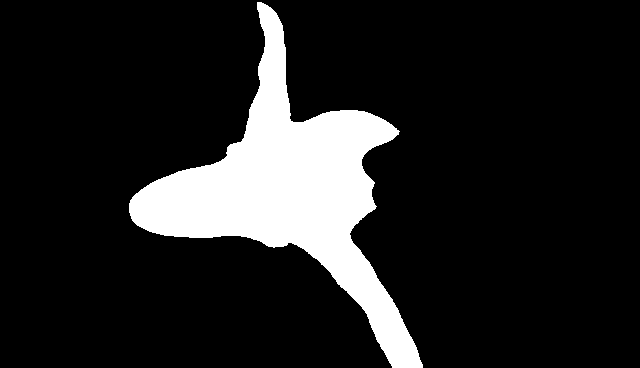}
	\end{subfigure}
	\begin{subfigure}{0.087\textwidth}
		\includegraphics[width=\textwidth]{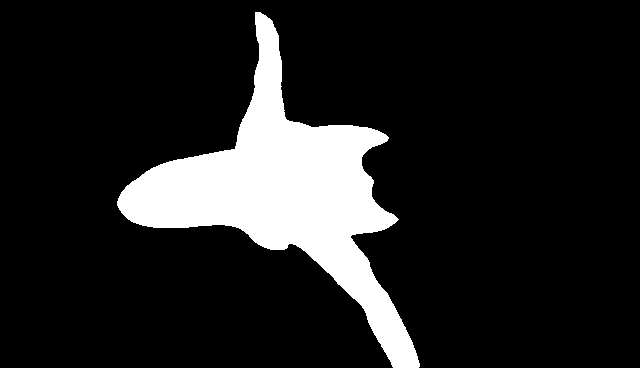}
	\end{subfigure}
	\begin{subfigure}{0.087\textwidth}
		\includegraphics[width=\textwidth]{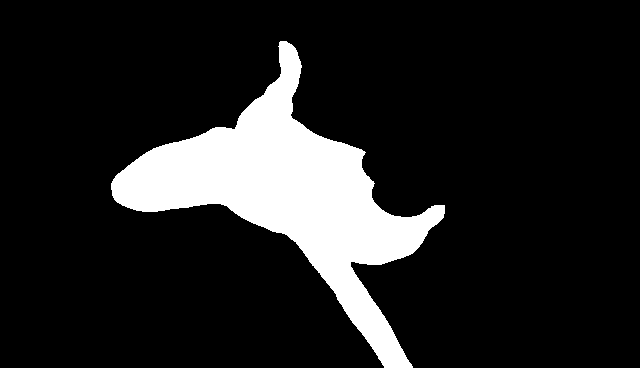}
	\end{subfigure}
	\begin{subfigure}{0.087\textwidth}
		\includegraphics[width=\textwidth]{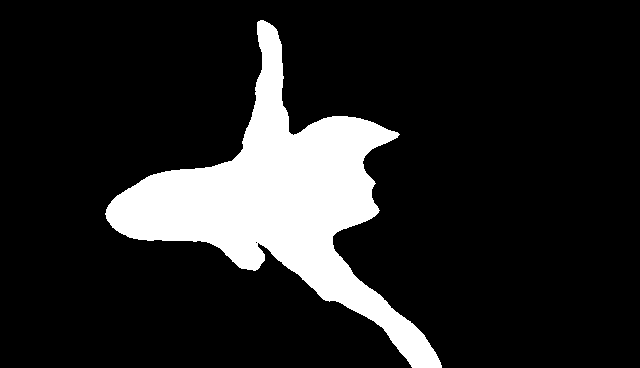}
	\end{subfigure}
	\begin{subfigure}{0.087\textwidth}
		\includegraphics[width=\textwidth]{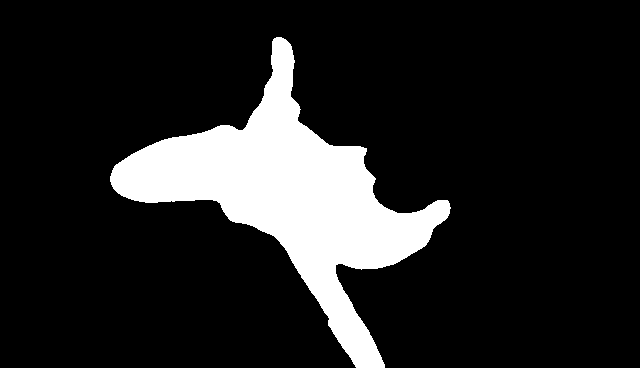}
	\end{subfigure}
	\begin{subfigure}{0.087\textwidth}
		\includegraphics[width=\textwidth]{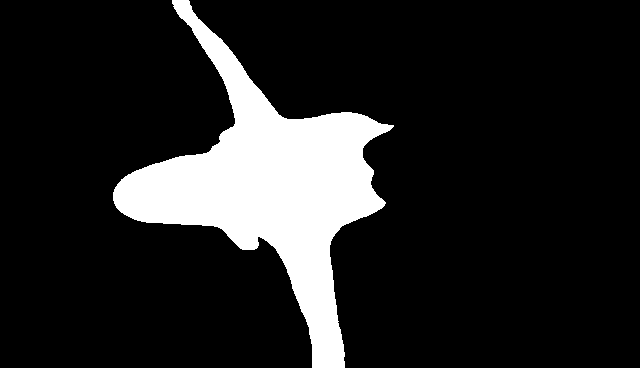}
	\end{subfigure}
	\begin{subfigure}{0.087\textwidth}
		\includegraphics[width=\textwidth]{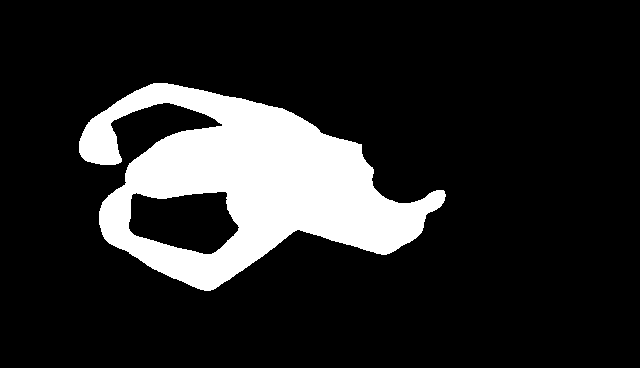}
	\end{subfigure}
	\begin{subfigure}{0.087\textwidth}
		\includegraphics[width=\textwidth]{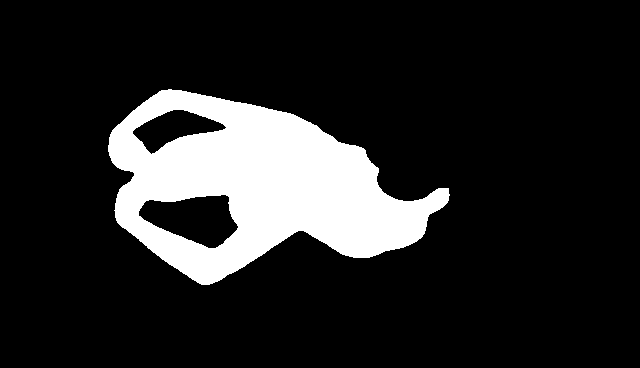}
	\end{subfigure}

	\vspace*{1.3mm}
	\begin{subfigure}{0.087\textwidth}
		\includegraphics[width=\textwidth]{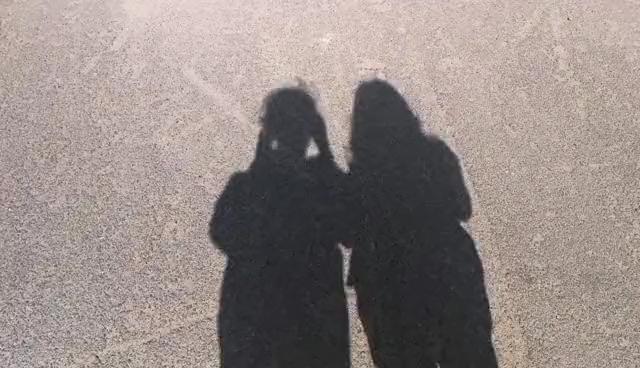}
	\end{subfigure}
	\begin{subfigure}{0.087\textwidth}
		\includegraphics[width=\textwidth]{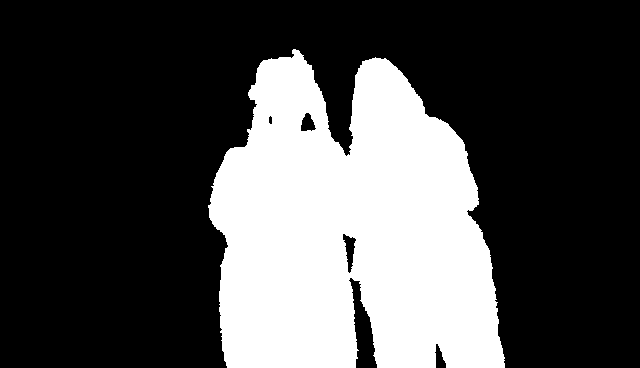}
	\end{subfigure}
	\begin{subfigure}{0.087\textwidth}
		\includegraphics[width=\textwidth]{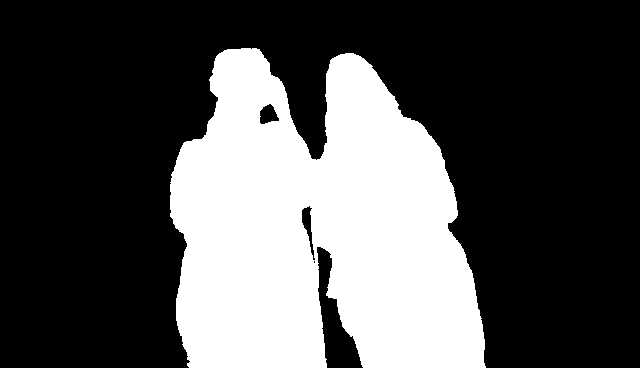}
	\end{subfigure}
	\begin{subfigure}{0.087\textwidth}
		\includegraphics[width=\textwidth]{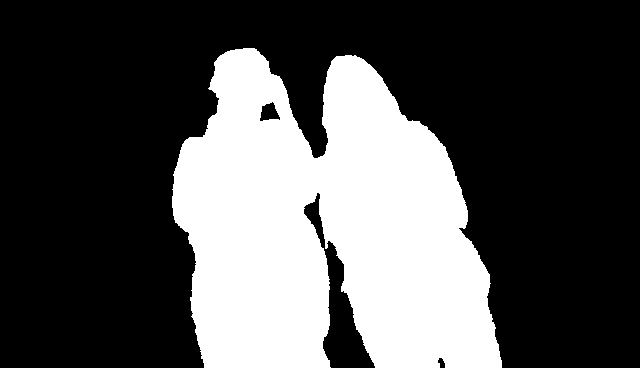}
	\end{subfigure}
	\begin{subfigure}{0.087\textwidth}
		\includegraphics[width=\textwidth]{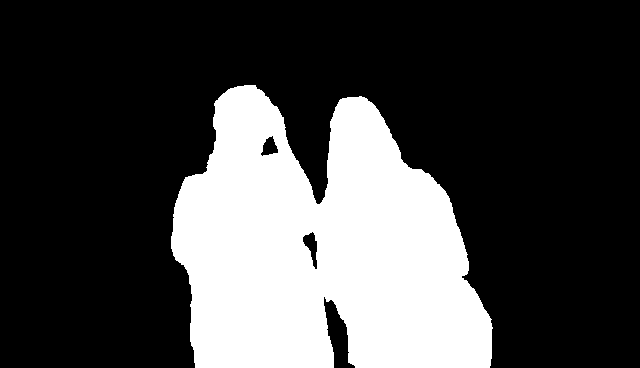}
	\end{subfigure}
	\begin{subfigure}{0.087\textwidth}
		\includegraphics[width=\textwidth]{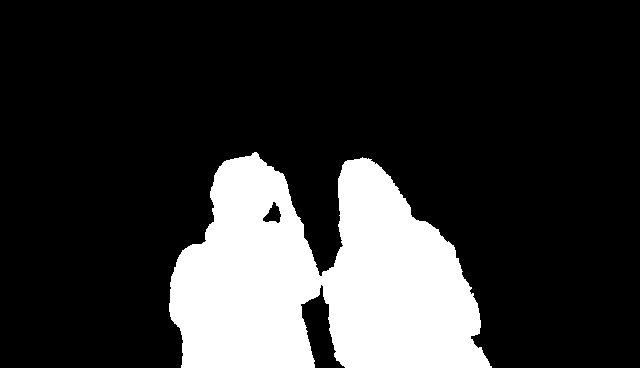}
	\end{subfigure}
	\begin{subfigure}{0.087\textwidth}
		\includegraphics[width=\textwidth]{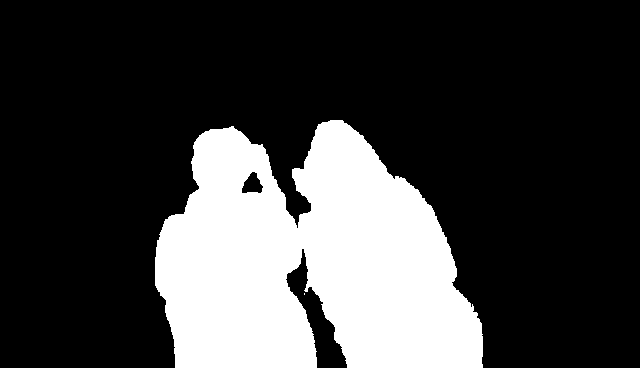}
	\end{subfigure}
	\begin{subfigure}{0.087\textwidth}
		\includegraphics[width=\textwidth]{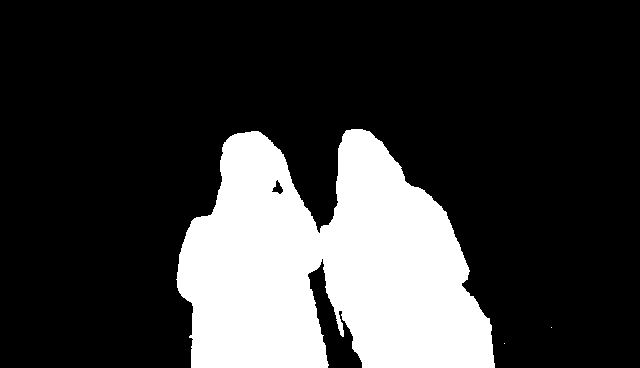}
	\end{subfigure}
	\begin{subfigure}{0.087\textwidth}
		\includegraphics[width=\textwidth]{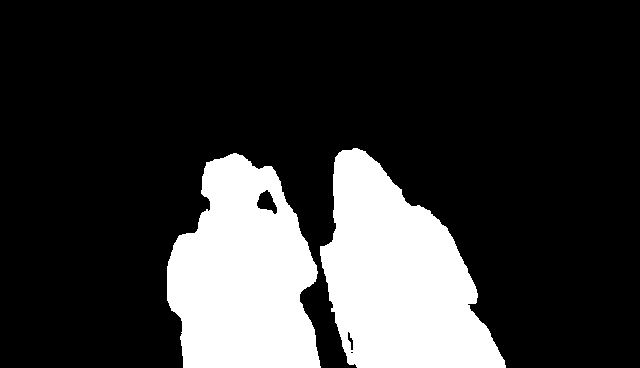}
	\end{subfigure}
	\begin{subfigure}{0.087\textwidth}
		\includegraphics[width=\textwidth]{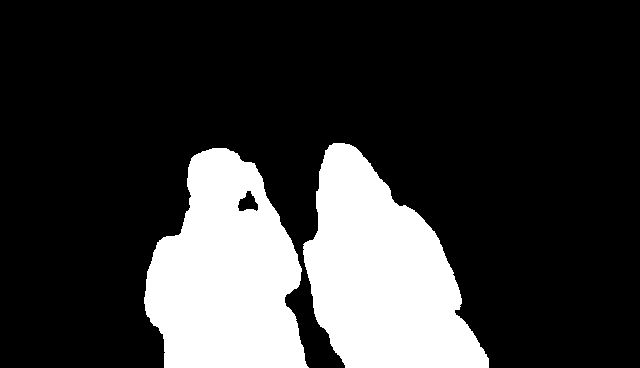}
	\end{subfigure}
	
	\vspace*{1.3mm}
	\begin{subfigure}{0.087\textwidth}
		\includegraphics[width=\textwidth]{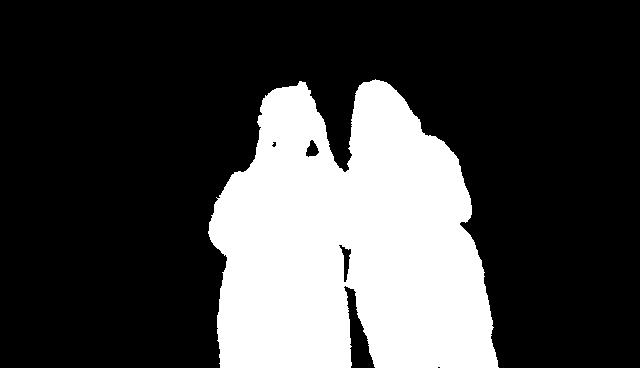}
	\end{subfigure}
	\begin{subfigure}{0.087\textwidth}
		\includegraphics[width=\textwidth]{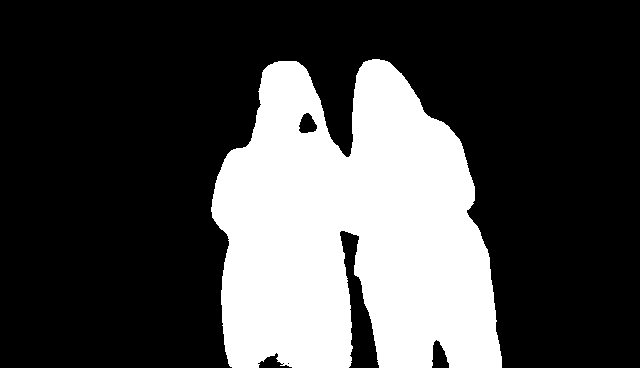}
	\end{subfigure}
	\begin{subfigure}{0.087\textwidth}
		\includegraphics[width=\textwidth]{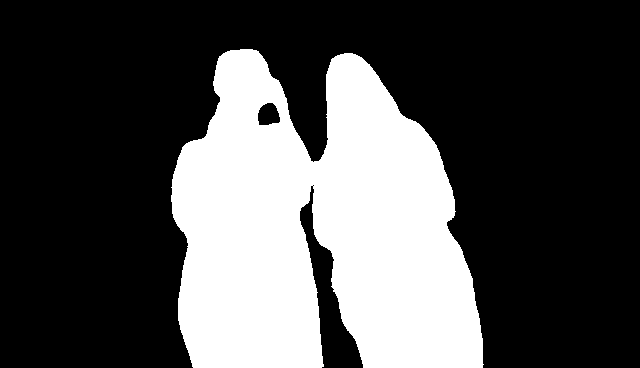}
	\end{subfigure}
	\begin{subfigure}{0.087\textwidth}
		\includegraphics[width=\textwidth]{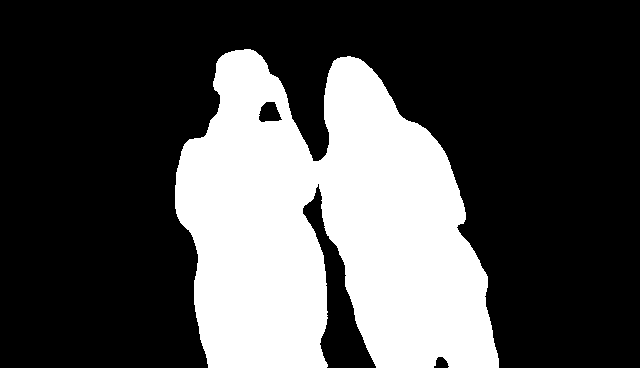}
	\end{subfigure}
	\begin{subfigure}{0.087\textwidth}
		\includegraphics[width=\textwidth]{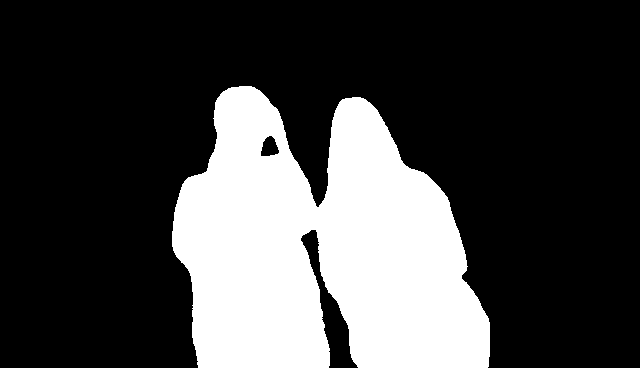}
	\end{subfigure}
	\begin{subfigure}{0.087\textwidth}
		\includegraphics[width=\textwidth]{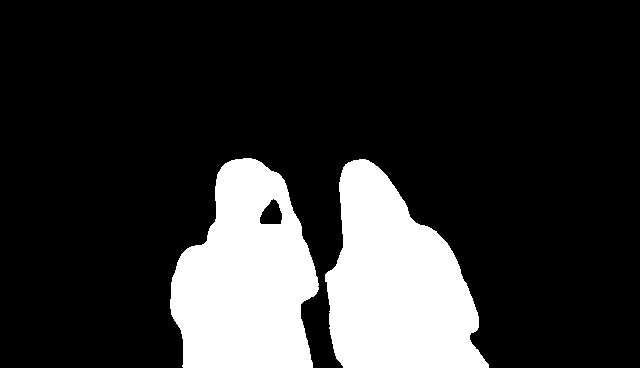}
	\end{subfigure}
	\begin{subfigure}{0.087\textwidth}
		\includegraphics[width=\textwidth]{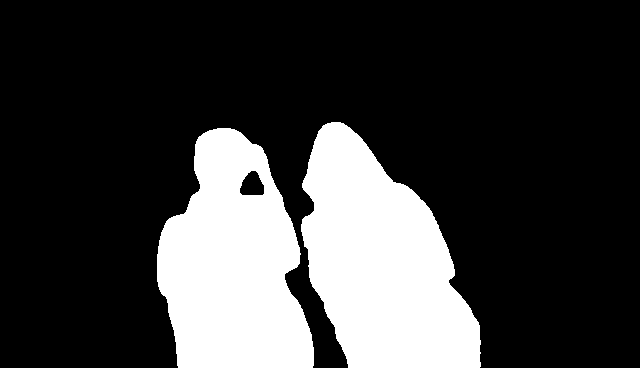}
	\end{subfigure}
	\begin{subfigure}{0.087\textwidth}
		\includegraphics[width=\textwidth]{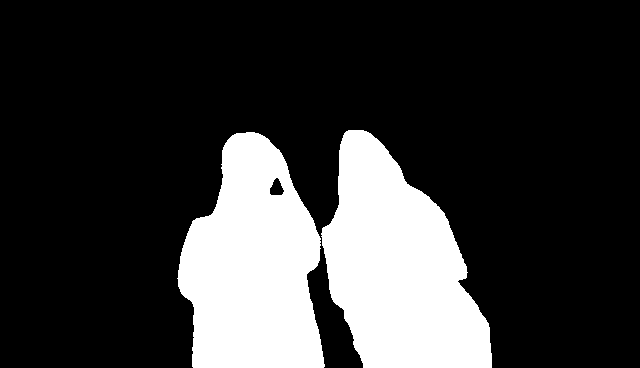}
	\end{subfigure}
	\begin{subfigure}{0.087\textwidth}
		\includegraphics[width=\textwidth]{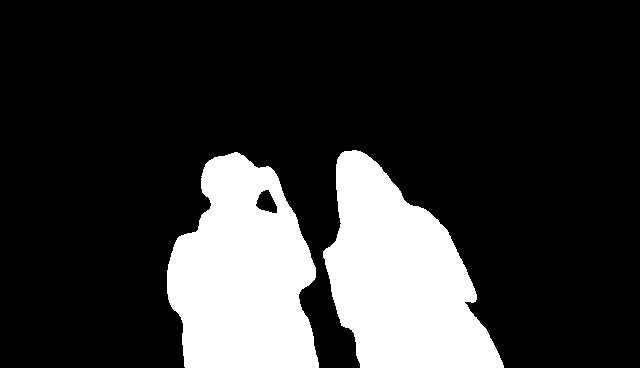}
	\end{subfigure}
	\begin{subfigure}{0.087\textwidth}
		\includegraphics[width=\textwidth]{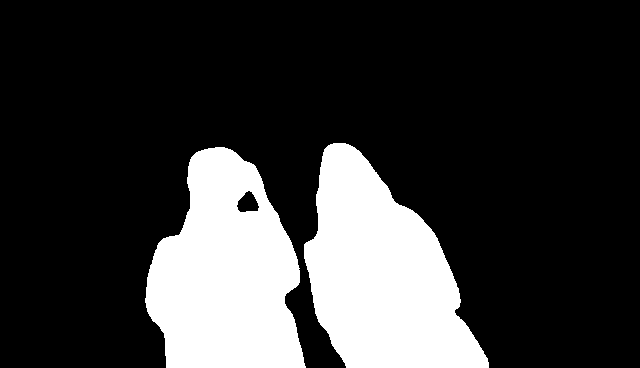}
	\end{subfigure}

	\vspace*{1.3mm}
	\begin{subfigure}{0.087\textwidth}
		\includegraphics[width=\textwidth]{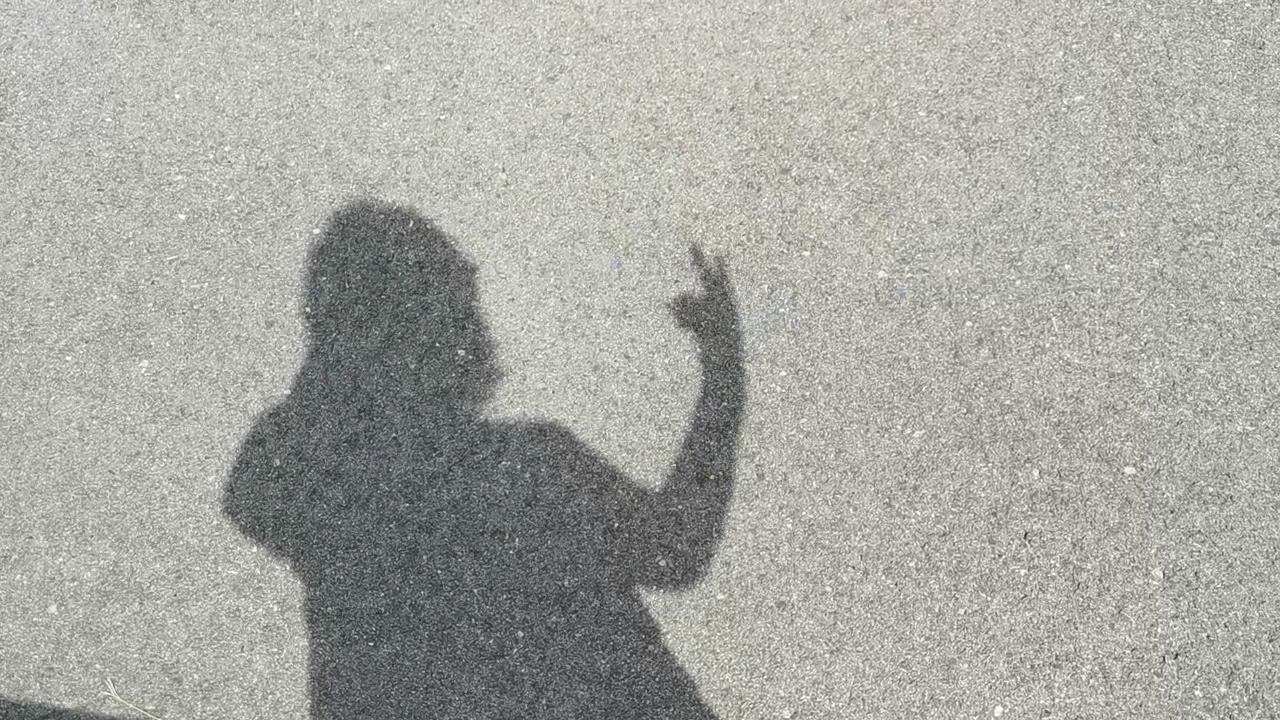}
	\end{subfigure}
	\begin{subfigure}{0.087\textwidth}
		\includegraphics[width=\textwidth]{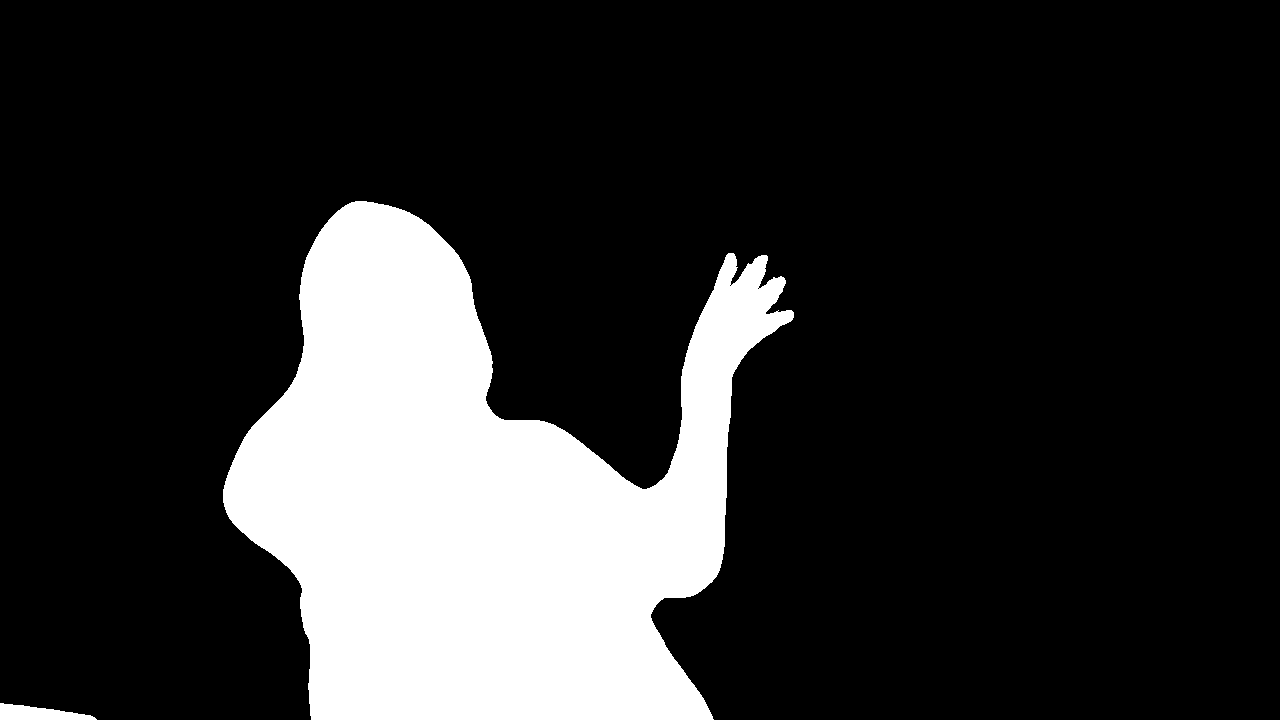}
	\end{subfigure}
	\begin{subfigure}{0.087\textwidth}
		\includegraphics[width=\textwidth]{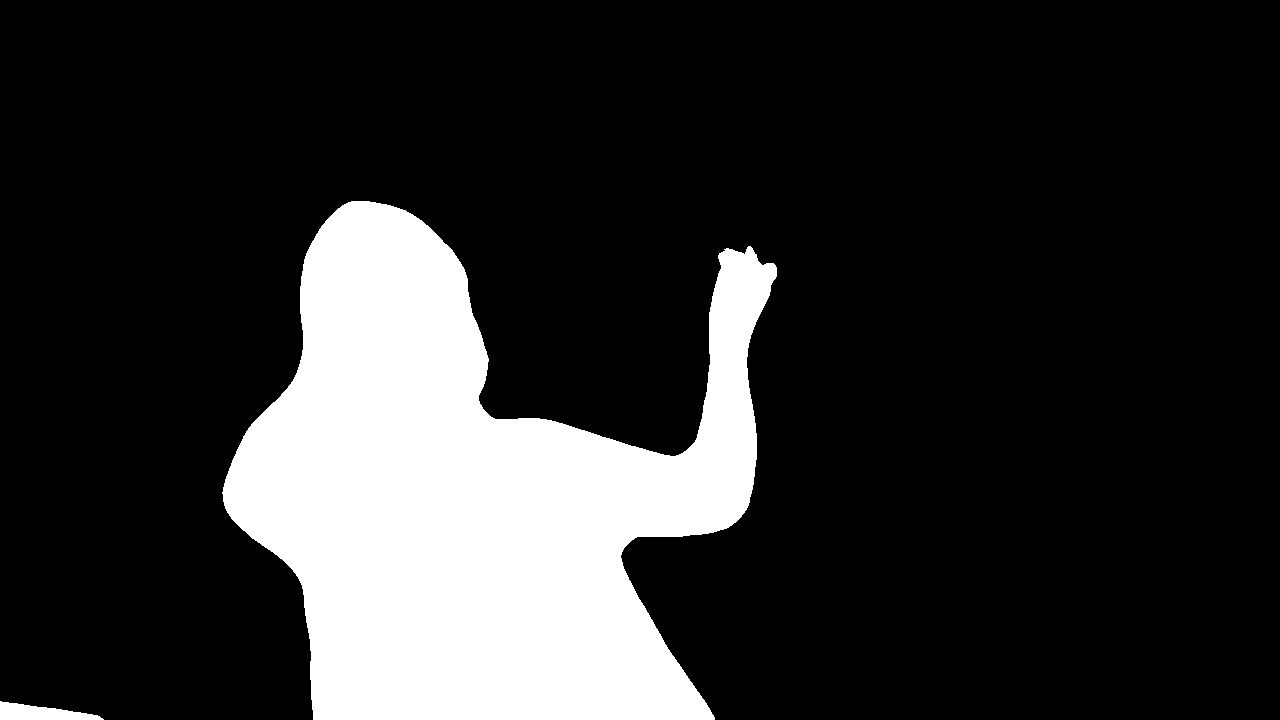}
	\end{subfigure}
	\begin{subfigure}{0.087\textwidth}
		\includegraphics[width=\textwidth]{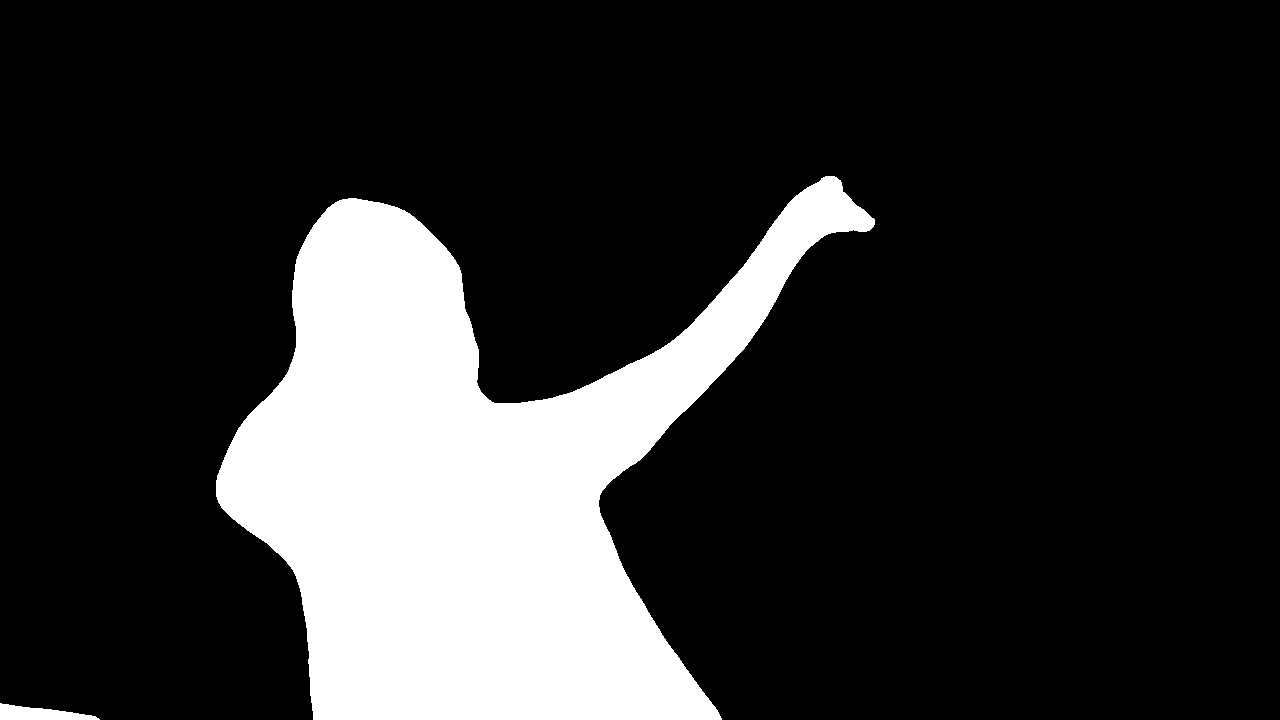}
	\end{subfigure}
	\begin{subfigure}{0.087\textwidth}
		\includegraphics[width=\textwidth]{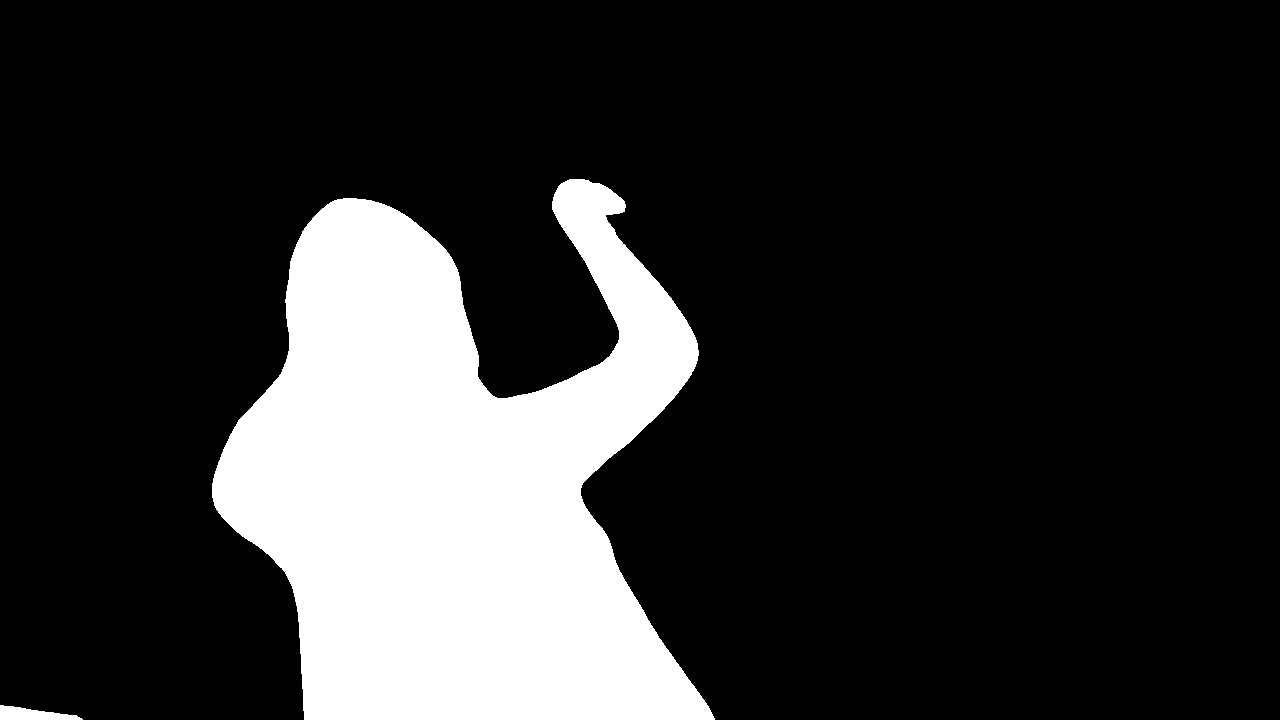}
	\end{subfigure}
	\begin{subfigure}{0.087\textwidth}
		\includegraphics[width=\textwidth]{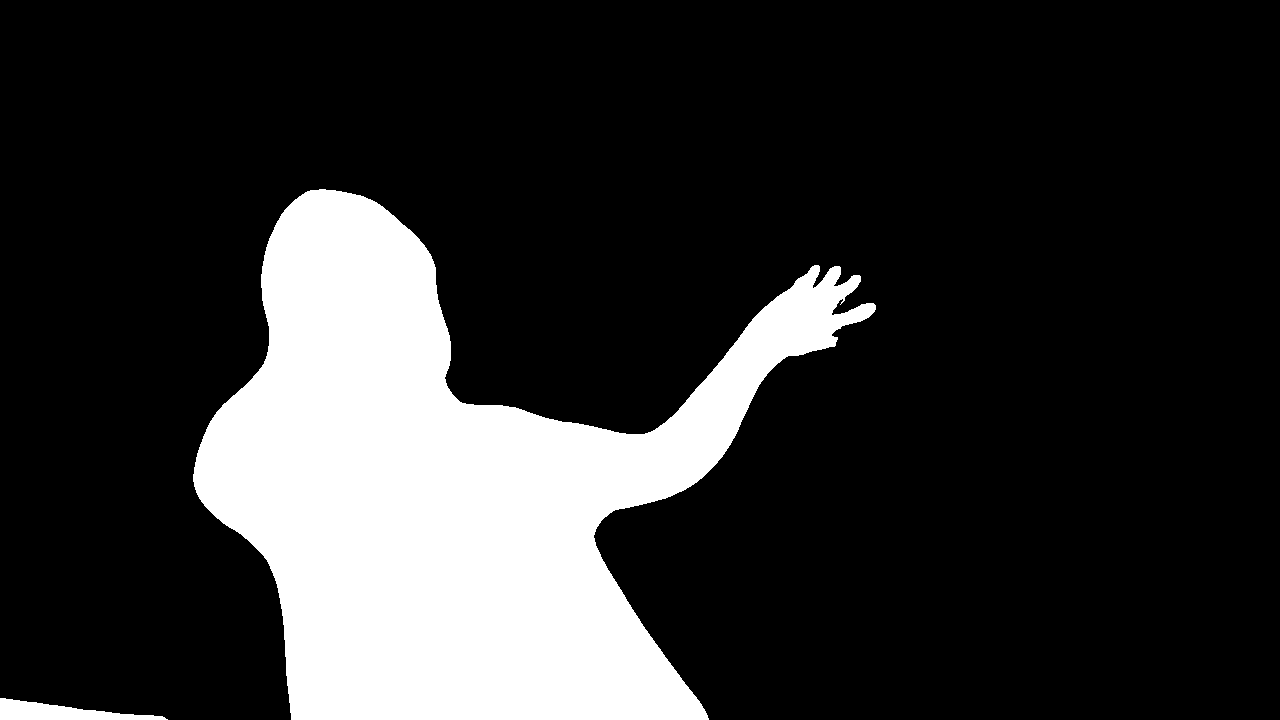}
	\end{subfigure}
	\begin{subfigure}{0.087\textwidth}
		\includegraphics[width=\textwidth]{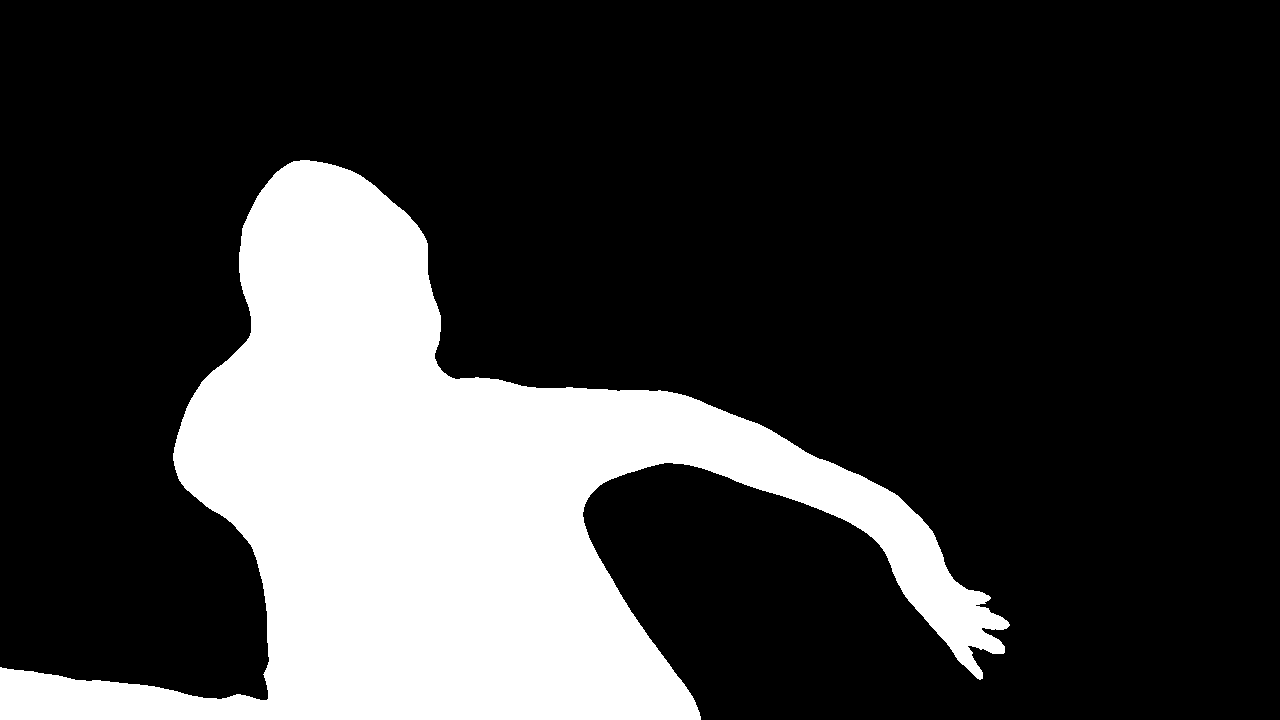}
	\end{subfigure}
	\begin{subfigure}{0.087\textwidth}
		\includegraphics[width=\textwidth]{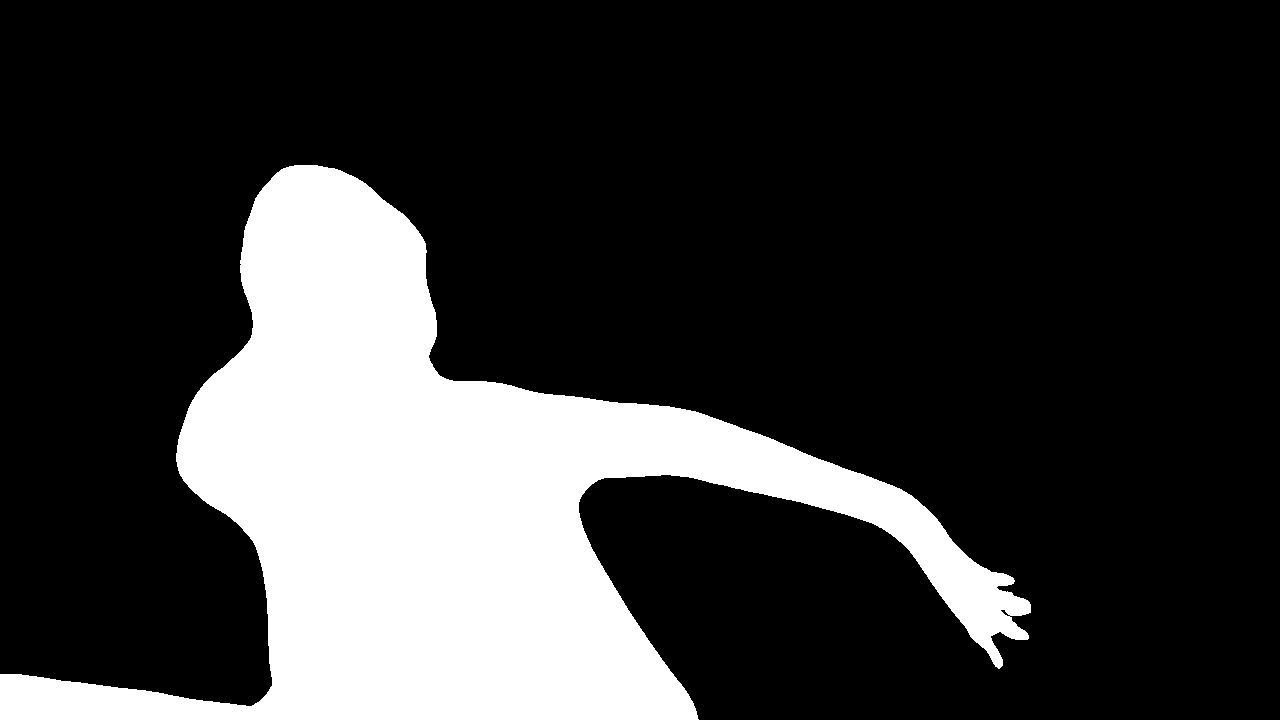}
	\end{subfigure}
	\begin{subfigure}{0.087\textwidth}
		\includegraphics[width=\textwidth]{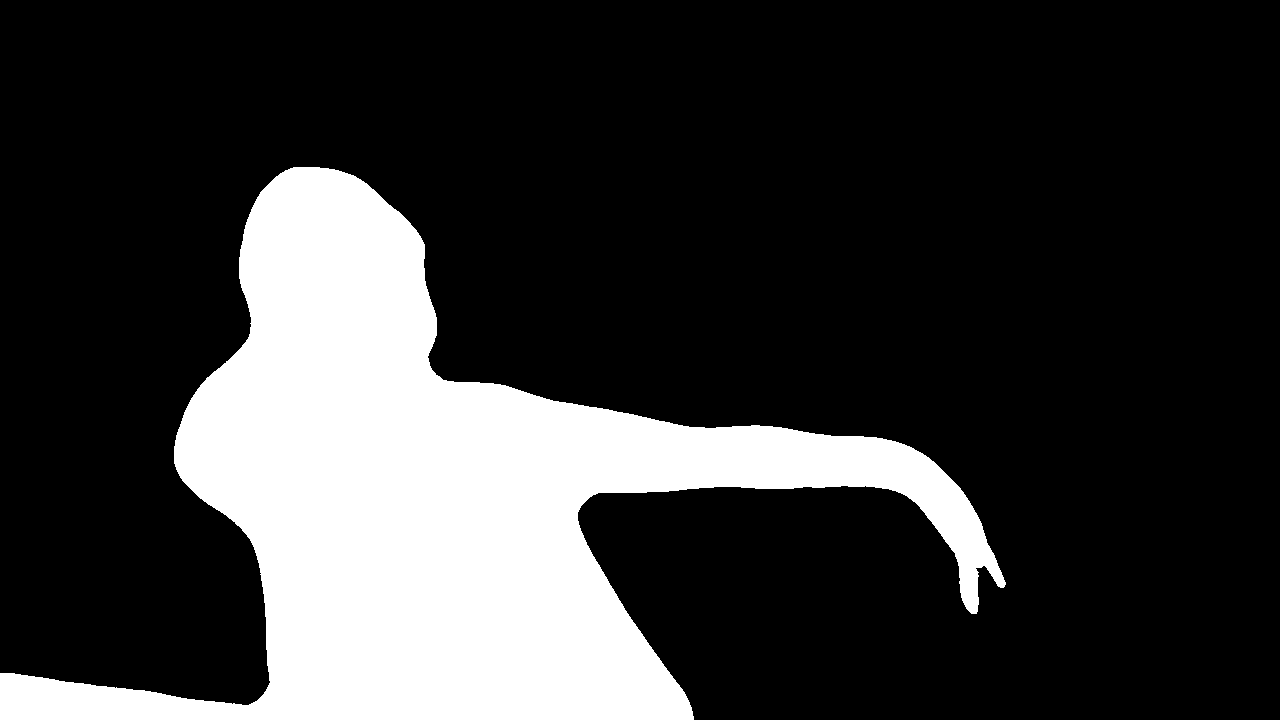}
	\end{subfigure}
	\begin{subfigure}{0.087\textwidth}
		\includegraphics[width=\textwidth]{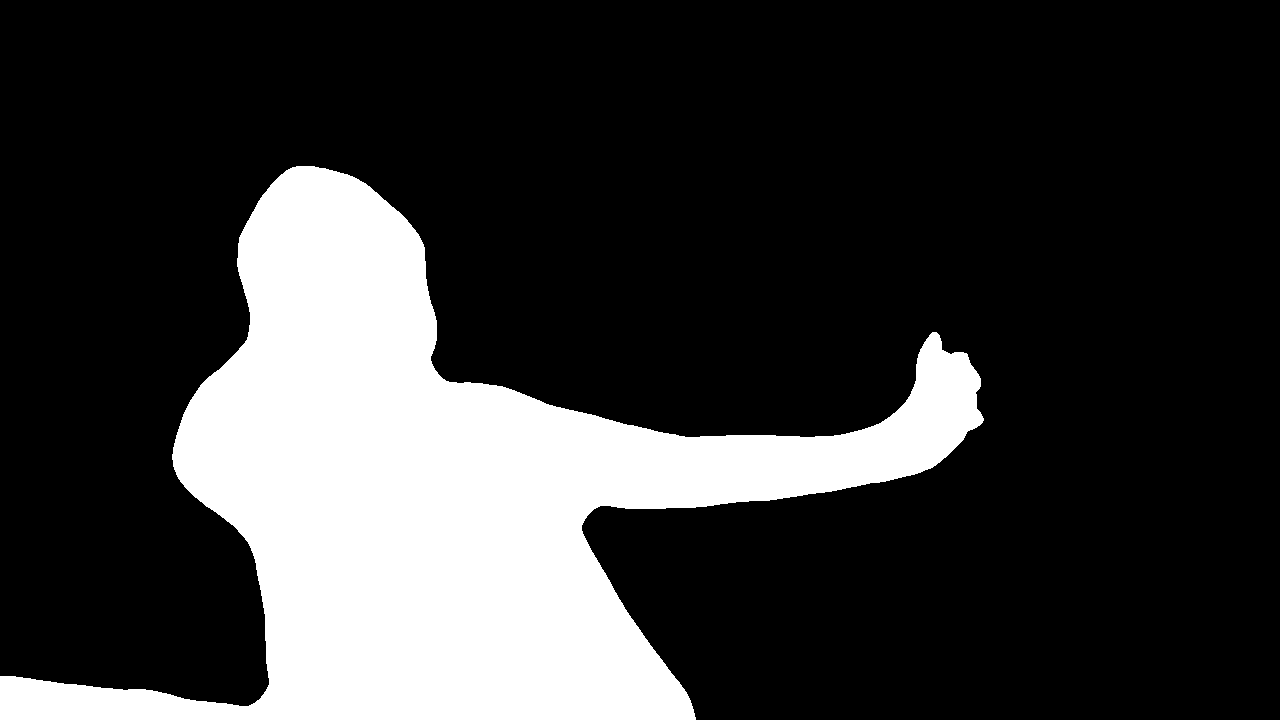}
	\end{subfigure}
	
	\vspace*{1.3mm}
	\begin{subfigure}{0.087\textwidth}
		\includegraphics[width=\textwidth]{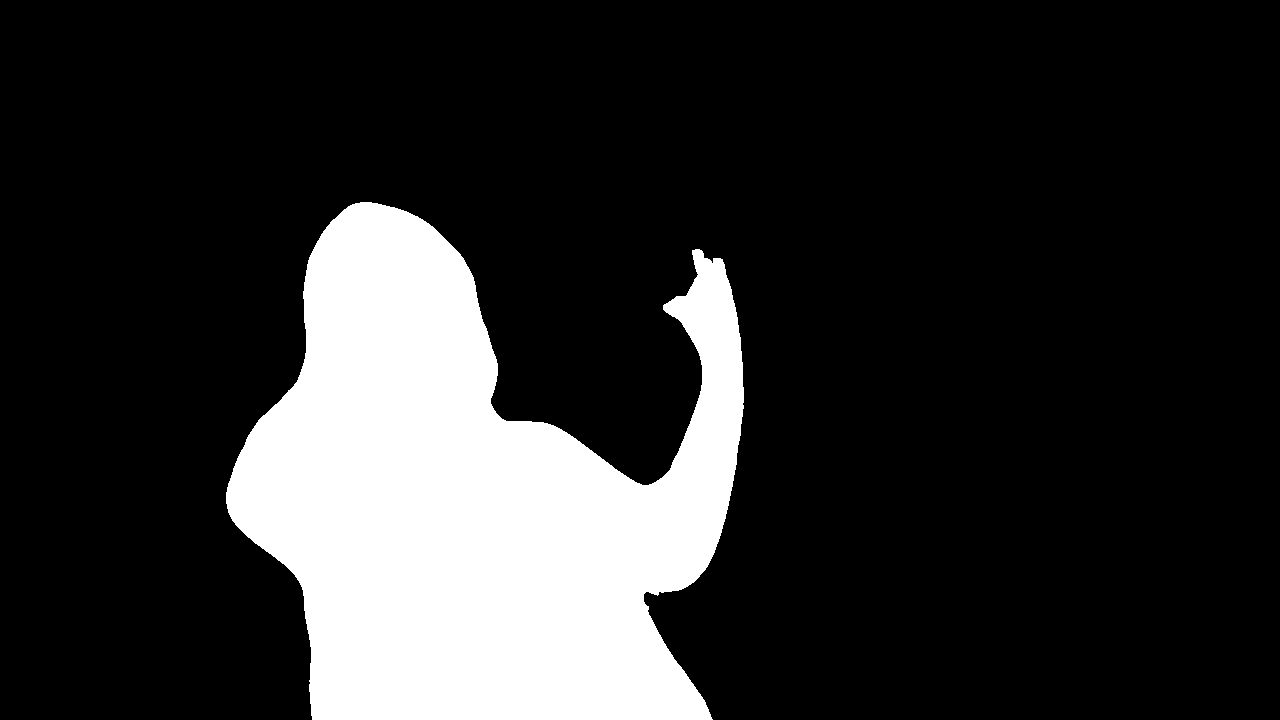}
	\end{subfigure}
	\begin{subfigure}{0.087\textwidth}
		\includegraphics[width=\textwidth]{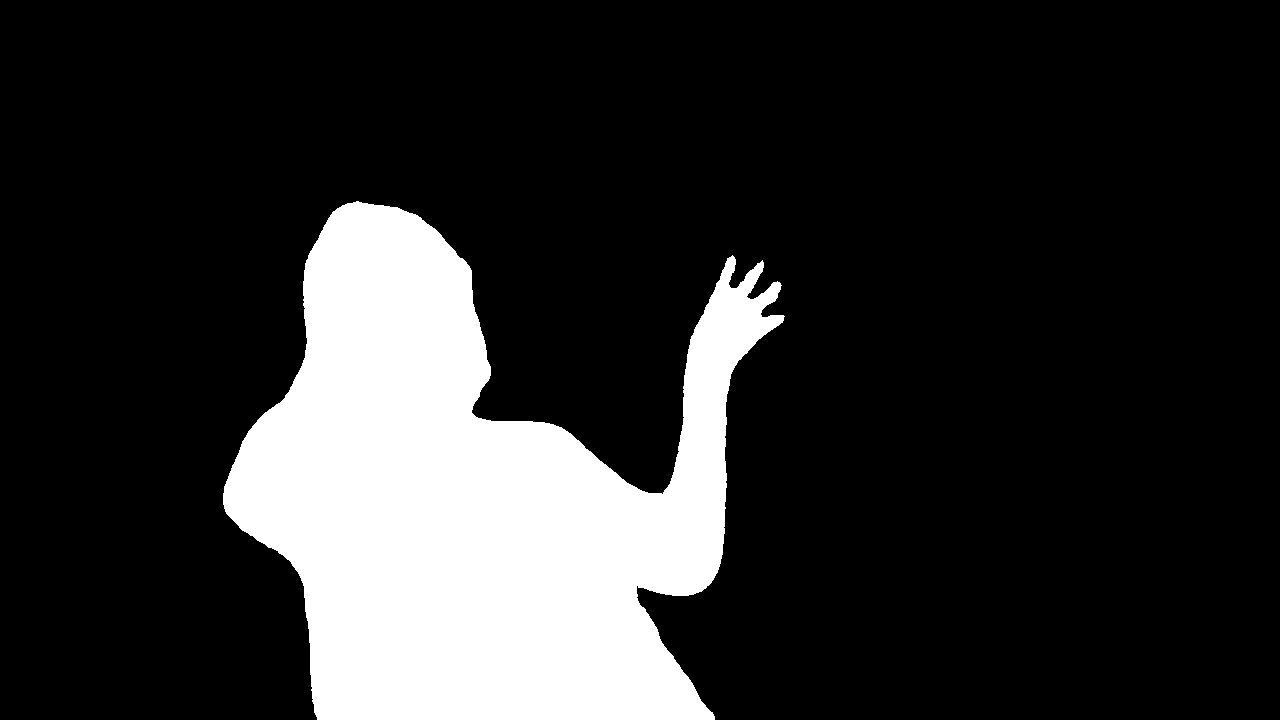}
	\end{subfigure}
	\begin{subfigure}{0.087\textwidth}
		\includegraphics[width=\textwidth]{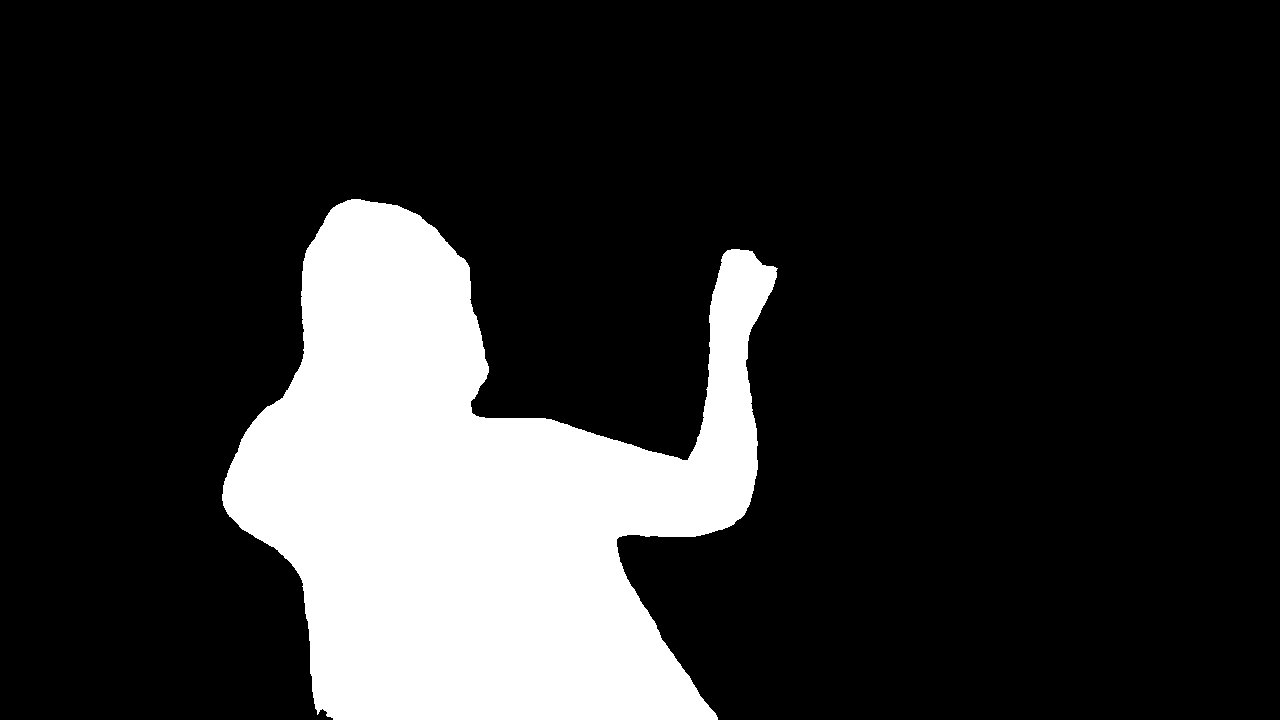}
	\end{subfigure}
	\begin{subfigure}{0.087\textwidth}
		\includegraphics[width=\textwidth]{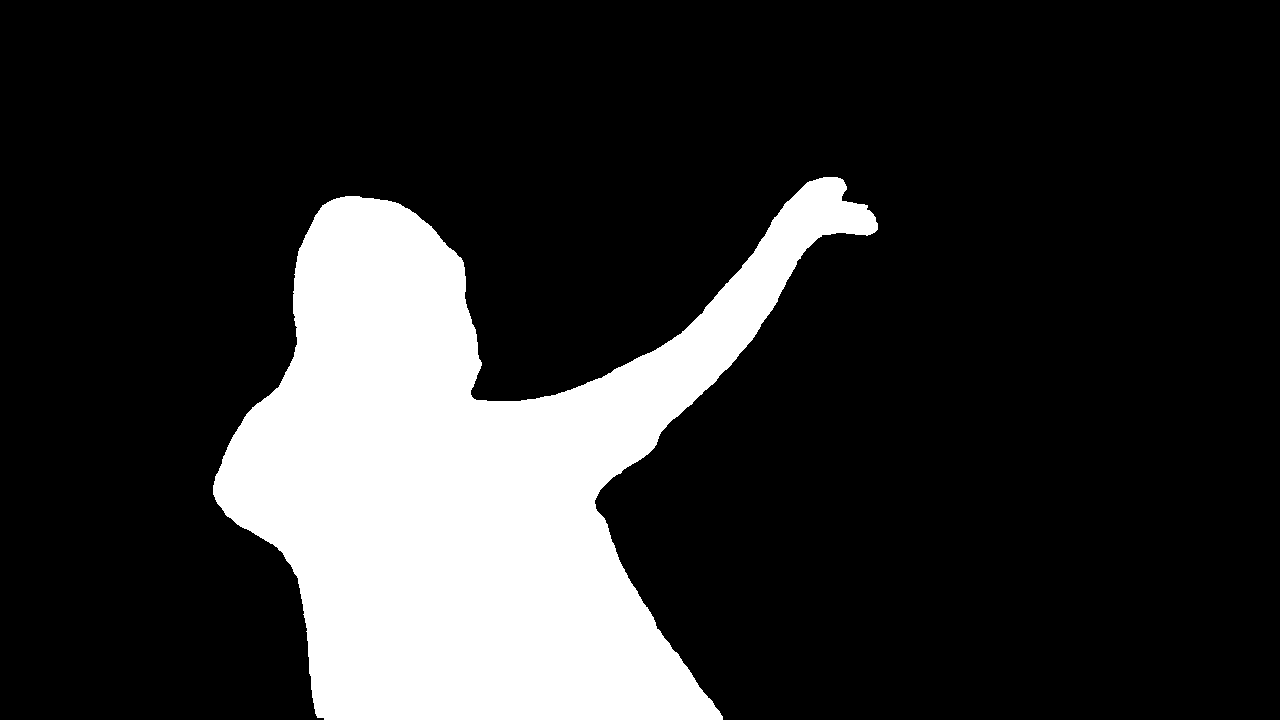}
	\end{subfigure}
	\begin{subfigure}{0.087\textwidth}
		\includegraphics[width=\textwidth]{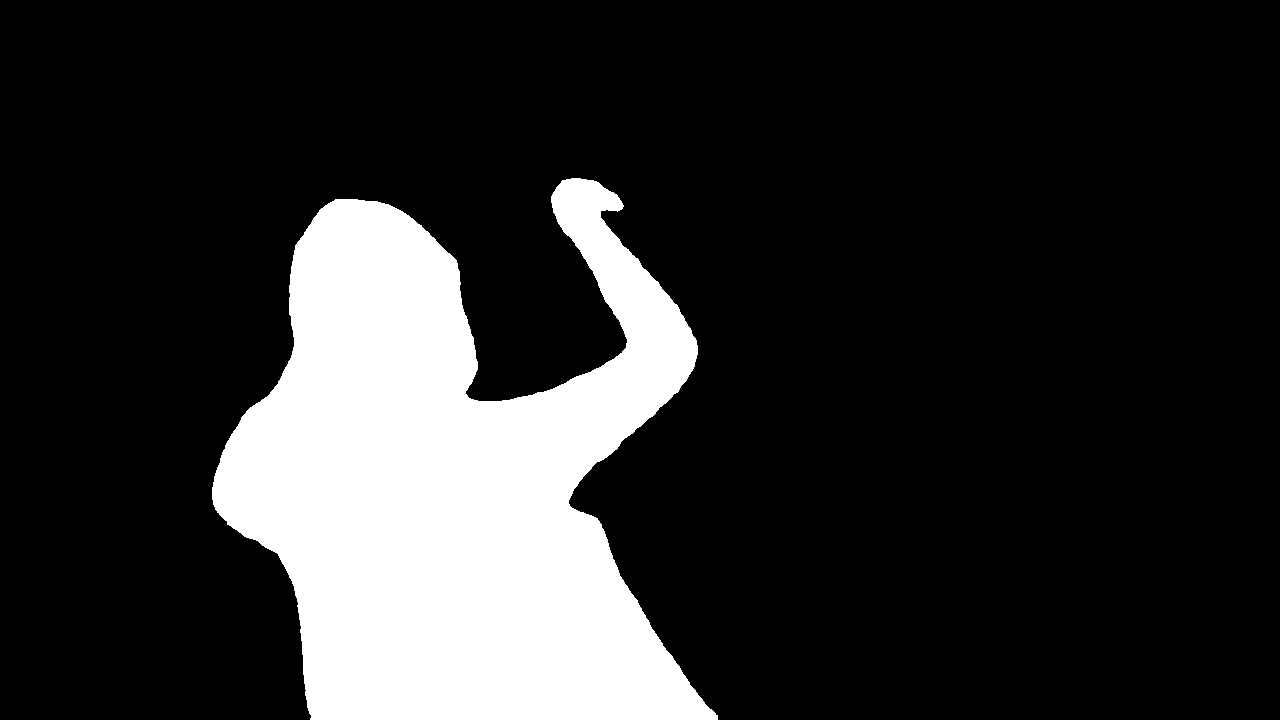}
	\end{subfigure}
	\begin{subfigure}{0.087\textwidth}
		\includegraphics[width=\textwidth]{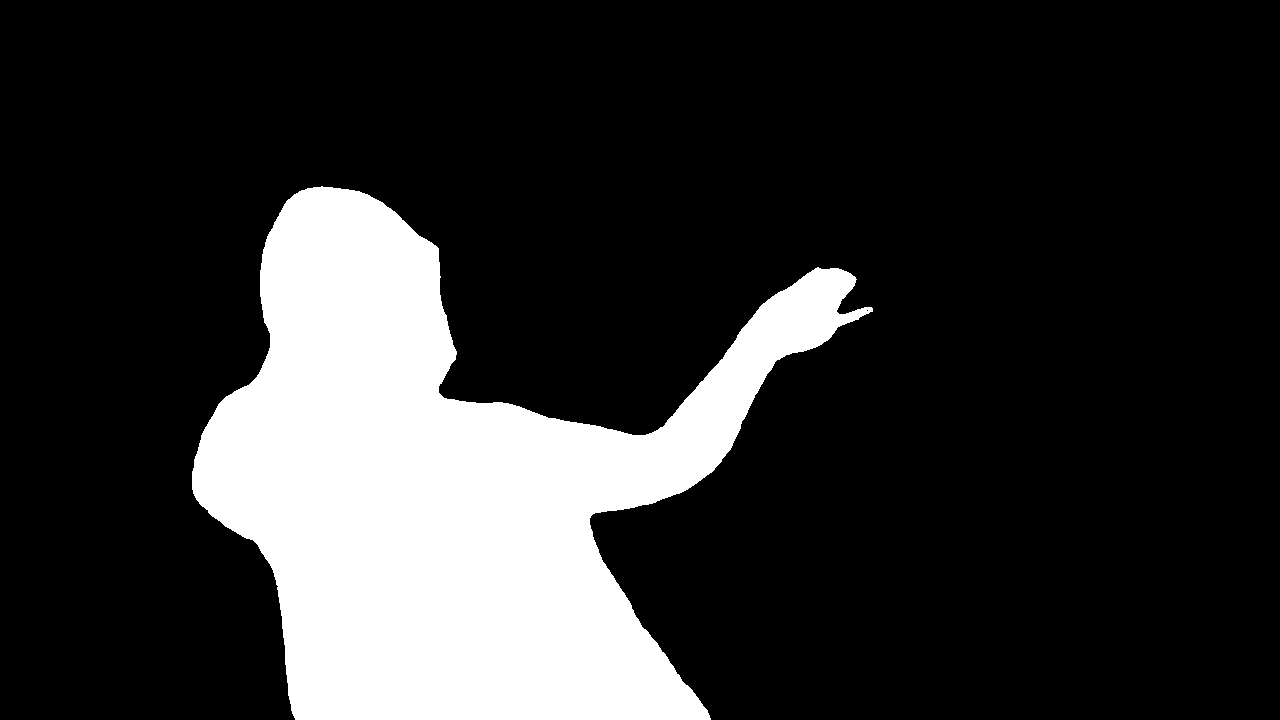}
	\end{subfigure}
	\begin{subfigure}{0.087\textwidth}
		\includegraphics[width=\textwidth]{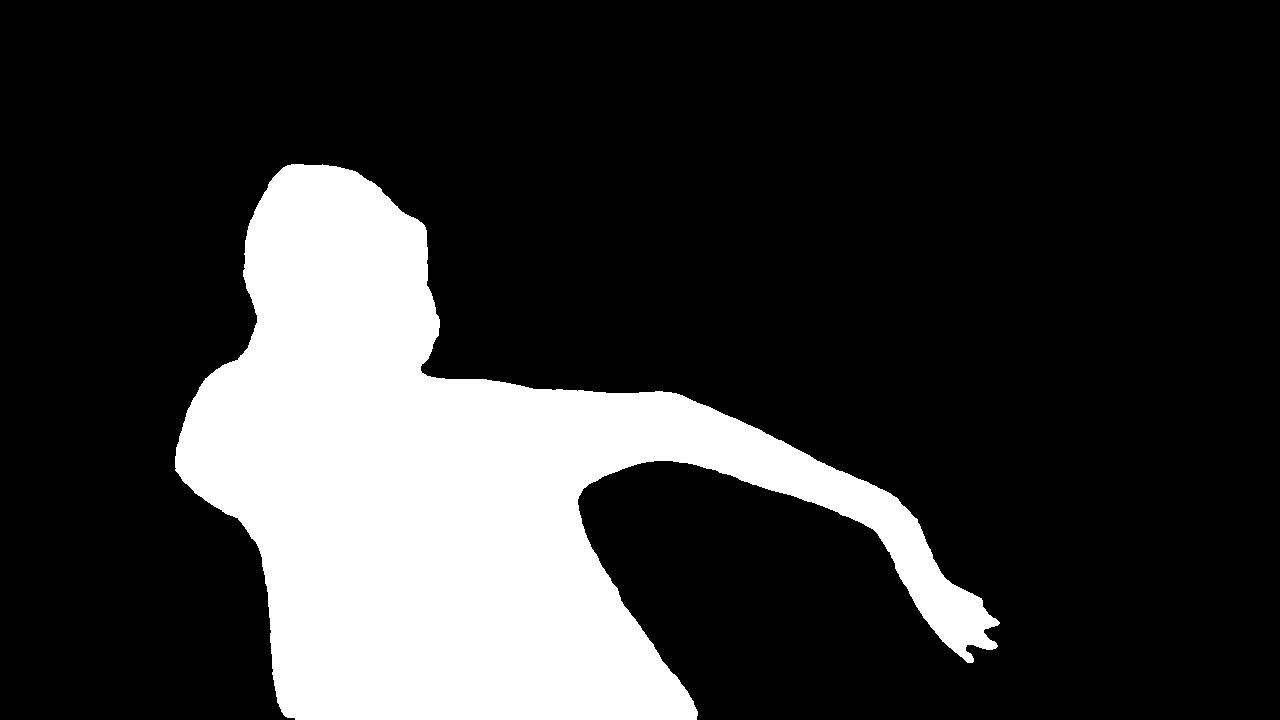}
	\end{subfigure}
	\begin{subfigure}{0.087\textwidth}
		\includegraphics[width=\textwidth]{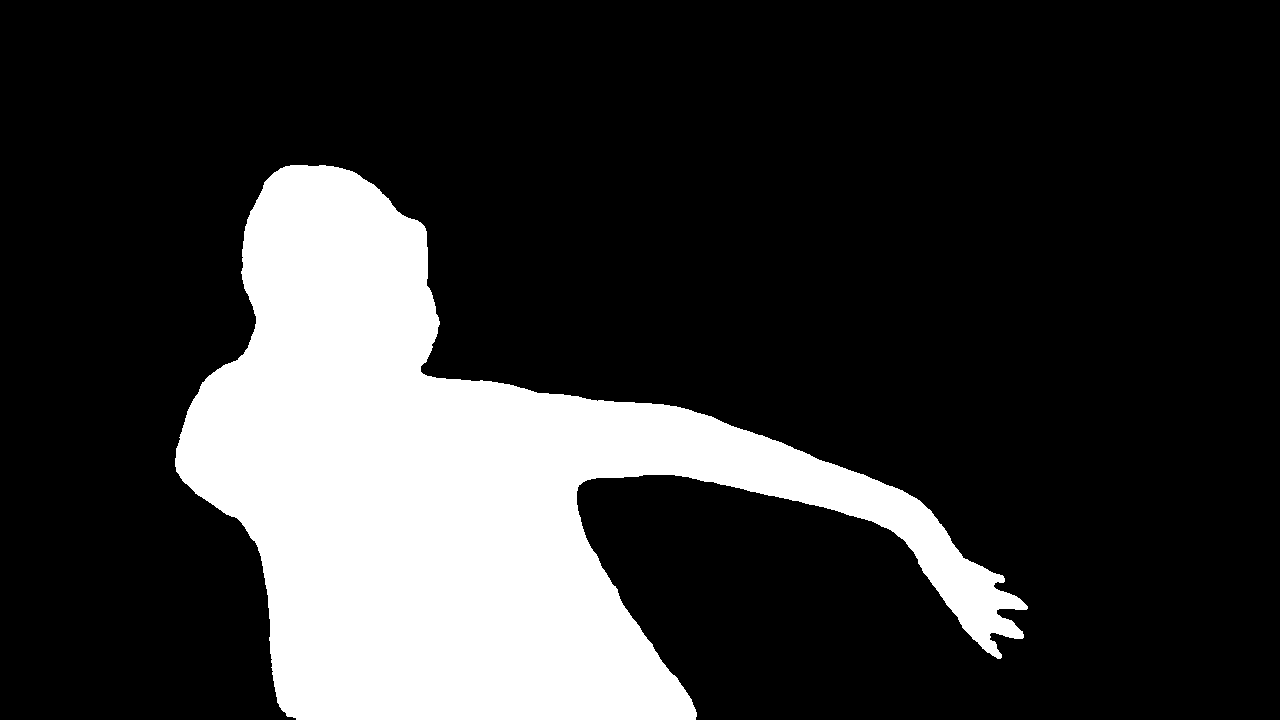}
	\end{subfigure}
	\begin{subfigure}{0.087\textwidth}
		\includegraphics[width=\textwidth]{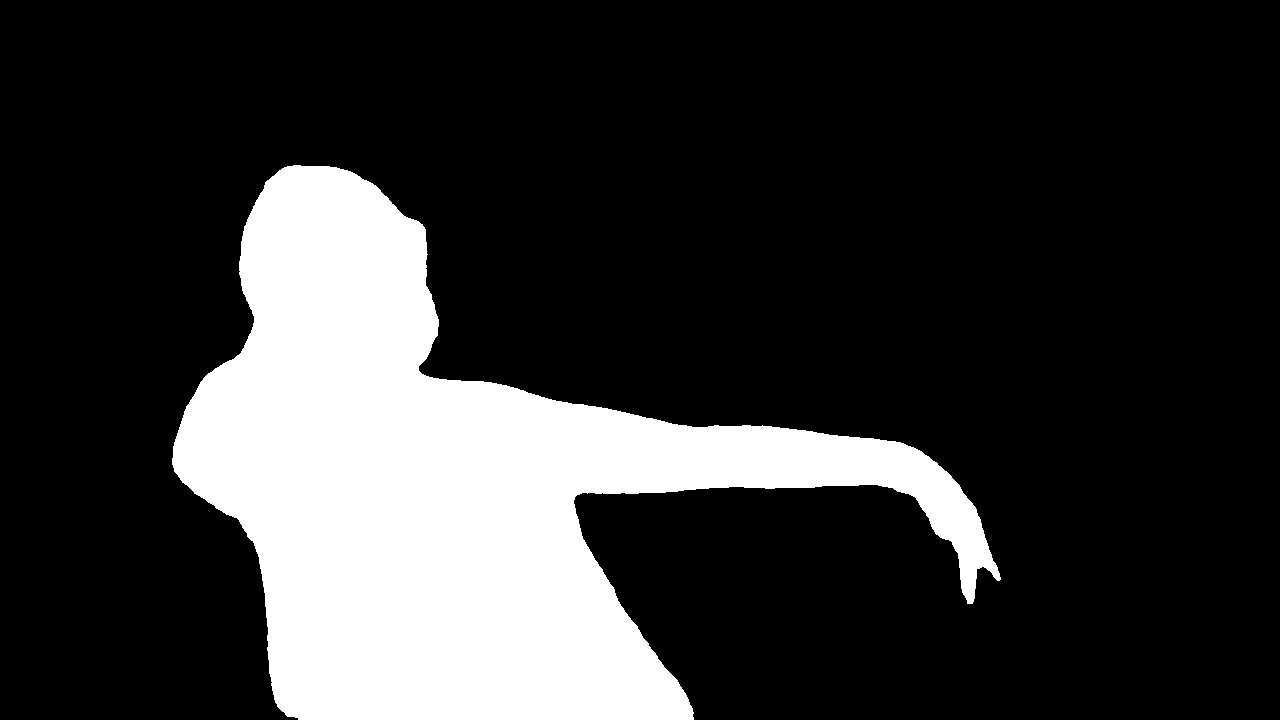}
	\end{subfigure}
	\begin{subfigure}{0.087\textwidth}
		\includegraphics[width=\textwidth]{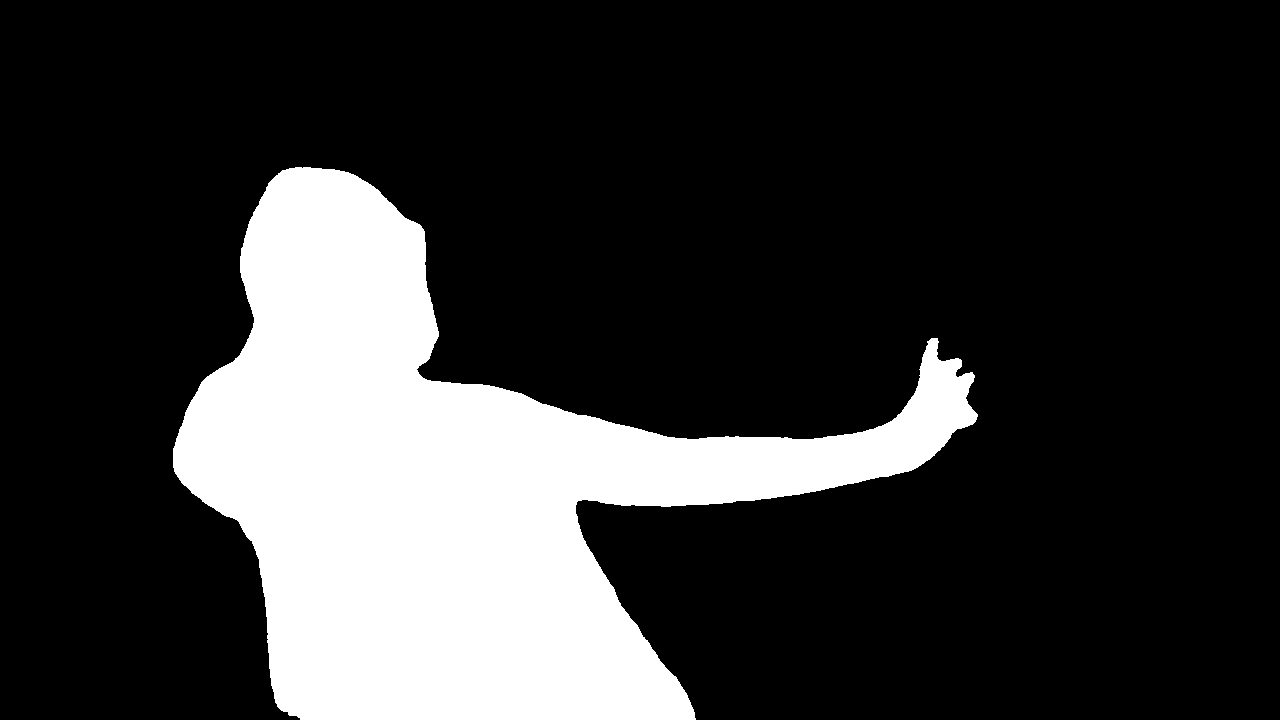}
	\end{subfigure}
	
	\vspace*{1.3mm}
	\begin{subfigure}{0.087\textwidth}
		\includegraphics[width=\textwidth]{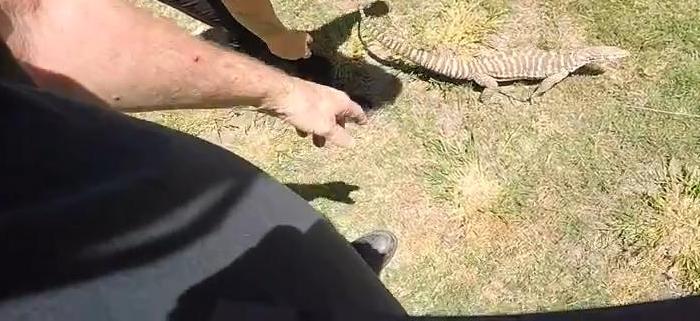}
	\end{subfigure}
	\begin{subfigure}{0.087\textwidth}
		\includegraphics[width=\textwidth]{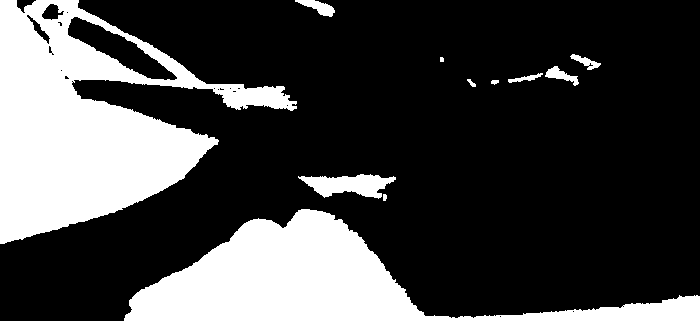}
	\end{subfigure}
	\begin{subfigure}{0.087\textwidth}
		\includegraphics[width=\textwidth]{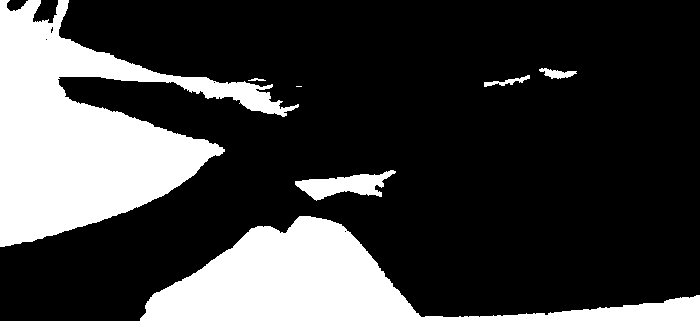}
	\end{subfigure}
	\begin{subfigure}{0.087\textwidth}
		\includegraphics[width=\textwidth]{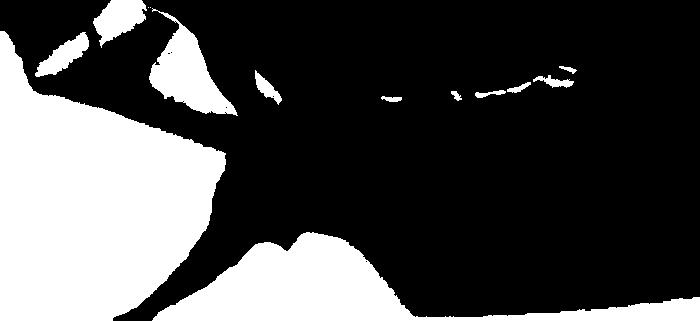}
	\end{subfigure}
	\begin{subfigure}{0.087\textwidth}
		\includegraphics[width=\textwidth]{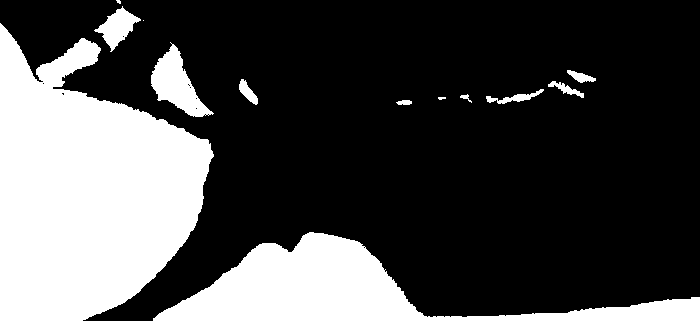}
	\end{subfigure}
	\begin{subfigure}{0.087\textwidth}
		\includegraphics[width=\textwidth]{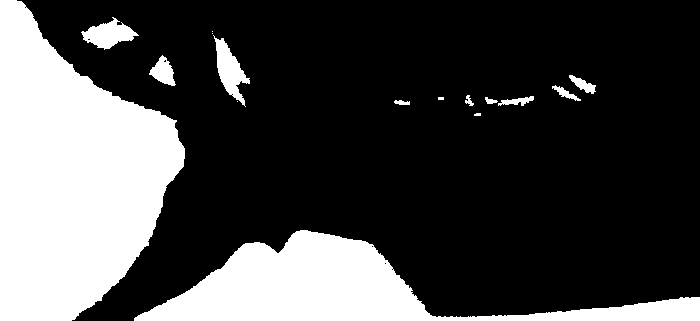}
	\end{subfigure}
	\begin{subfigure}{0.087\textwidth}
		\includegraphics[width=\textwidth]{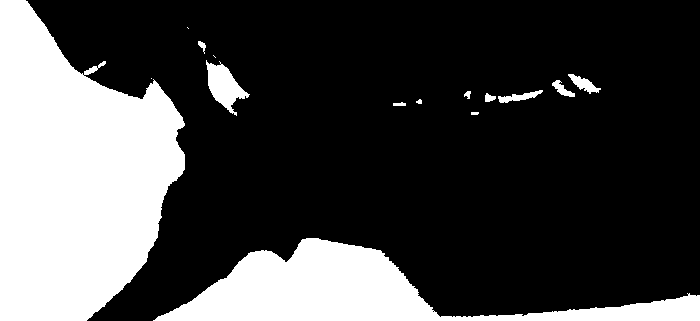}
	\end{subfigure}
	\begin{subfigure}{0.087\textwidth}
		\includegraphics[width=\textwidth]{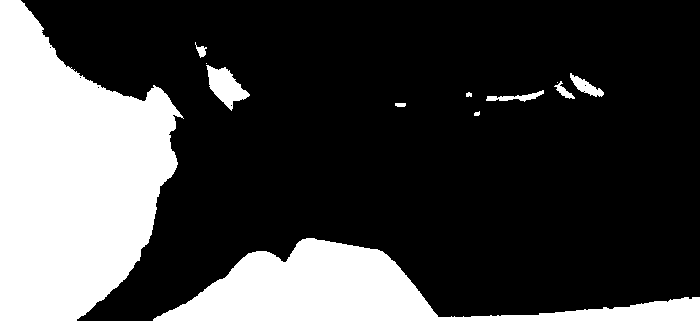}
	\end{subfigure}
	\begin{subfigure}{0.087\textwidth}
		\includegraphics[width=\textwidth]{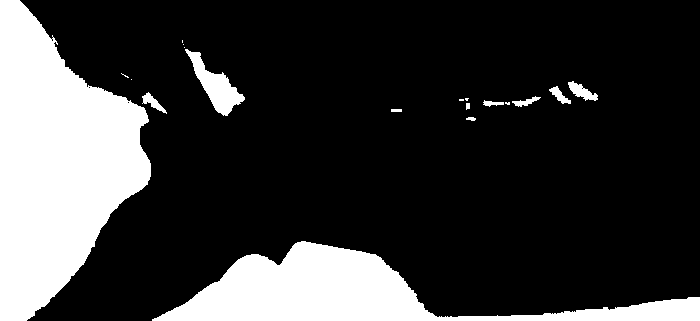}
	\end{subfigure}
	\begin{subfigure}{0.087\textwidth}
		\includegraphics[width=\textwidth]{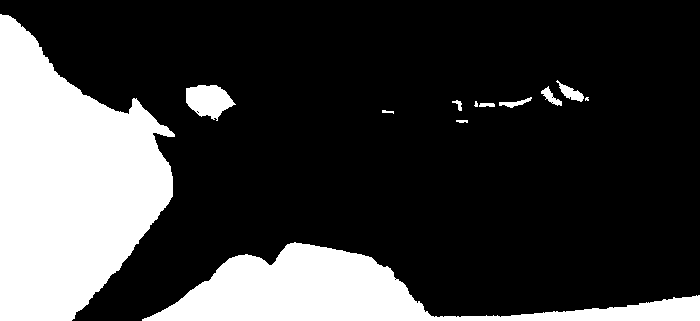}
	\end{subfigure}
	
	\vspace*{1.3mm}
	\begin{subfigure}{0.087\textwidth}
		\includegraphics[width=\textwidth]{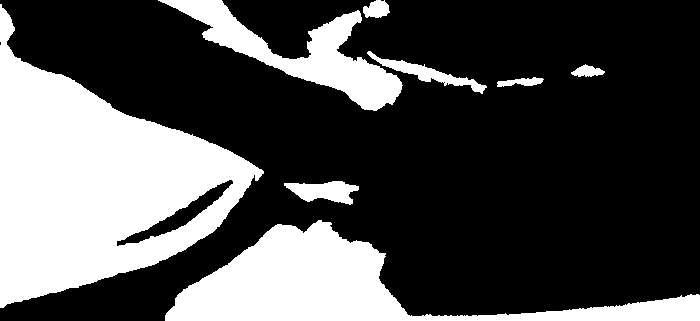}
	\end{subfigure}
	\begin{subfigure}{0.087\textwidth}
		\includegraphics[width=\textwidth]{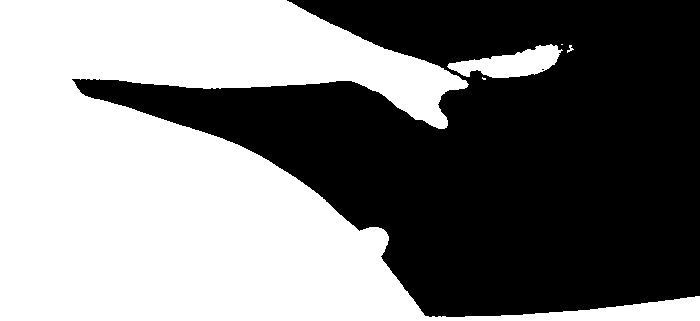}
	\end{subfigure}
	\begin{subfigure}{0.087\textwidth}
		\includegraphics[width=\textwidth]{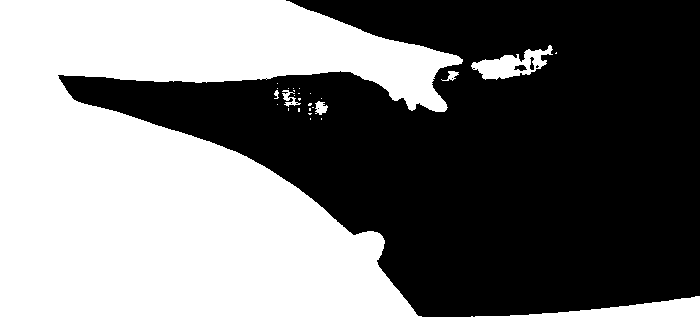}
	\end{subfigure}
	\begin{subfigure}{0.087\textwidth}
		\includegraphics[width=\textwidth]{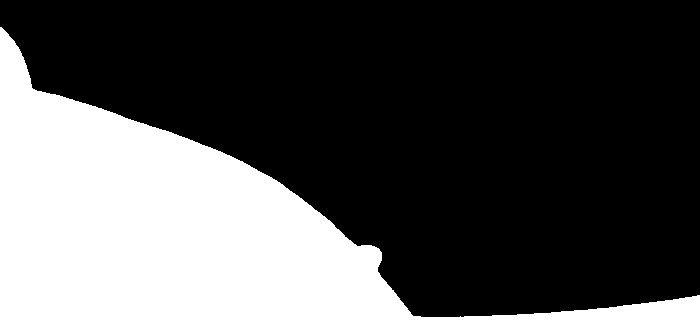}
	\end{subfigure}
	\begin{subfigure}{0.087\textwidth}
		\includegraphics[width=\textwidth]{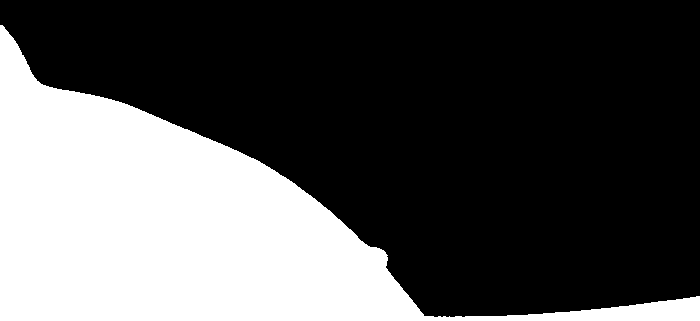}
	\end{subfigure}
	\begin{subfigure}{0.087\textwidth}
		\includegraphics[width=\textwidth]{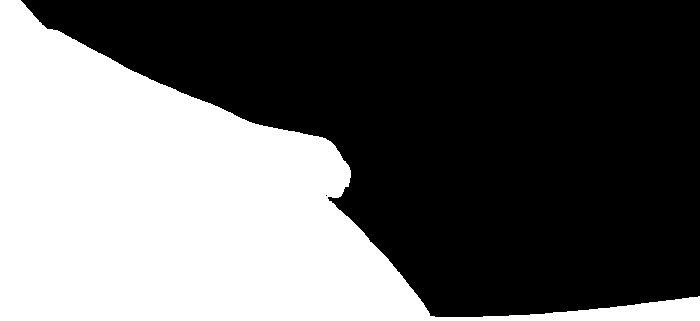}
	\end{subfigure}
	\begin{subfigure}{0.087\textwidth}
		\includegraphics[width=\textwidth]{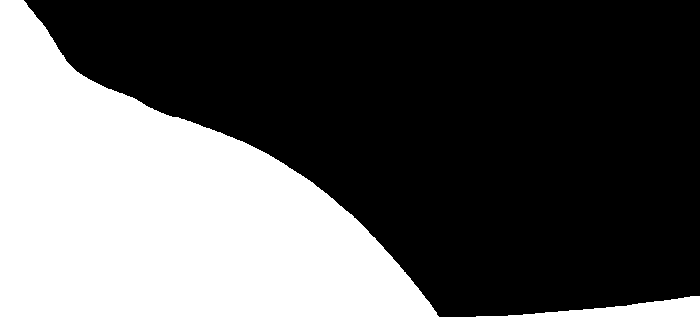}
	\end{subfigure}
	\begin{subfigure}{0.087\textwidth}
		\includegraphics[width=\textwidth]{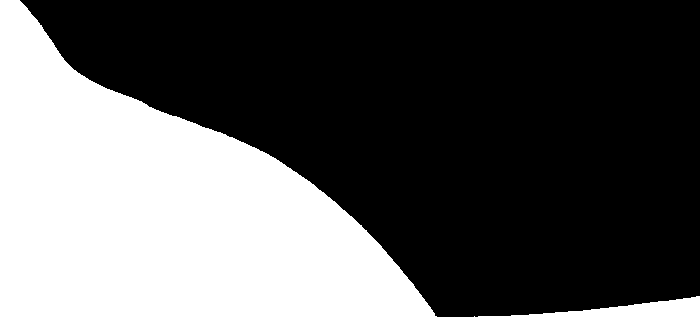}
	\end{subfigure}
	\begin{subfigure}{0.087\textwidth}
		\includegraphics[width=\textwidth]{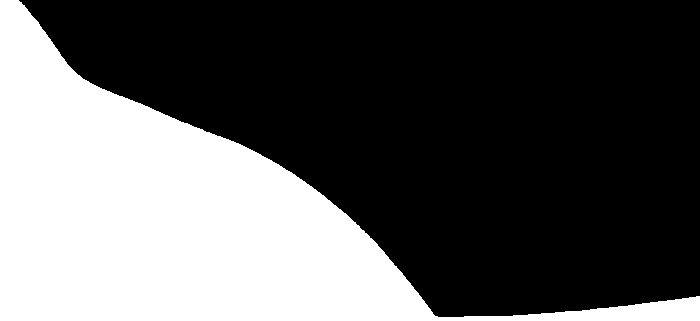}
	\end{subfigure}
	\begin{subfigure}{0.087\textwidth}
		\includegraphics[width=\textwidth]{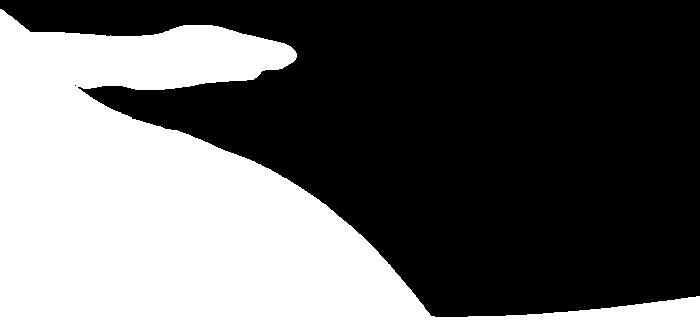}
	\end{subfigure}

	\vspace*{1.3mm}
	\begin{subfigure}{0.087\textwidth}
		\includegraphics[width=\textwidth]{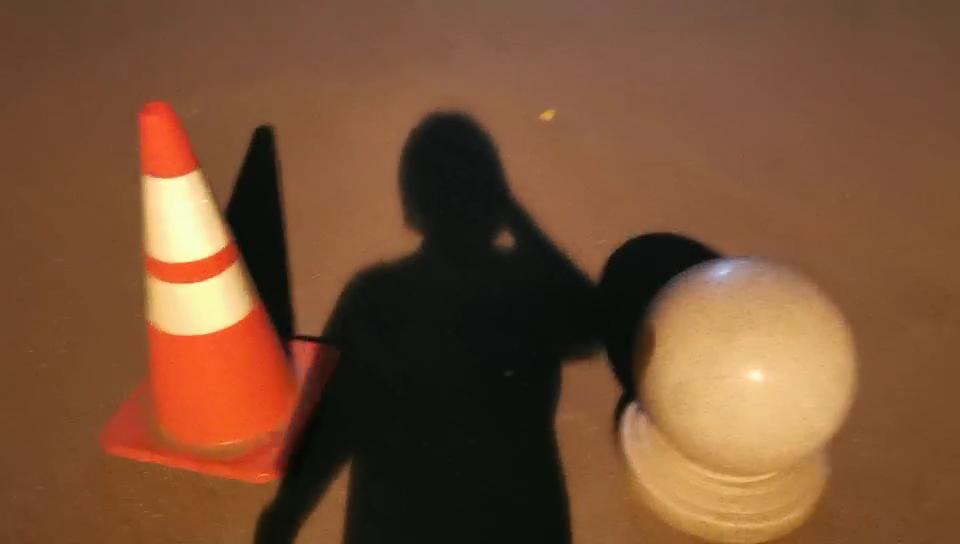}
	\end{subfigure}
	\begin{subfigure}{0.087\textwidth}
		\includegraphics[width=\textwidth]{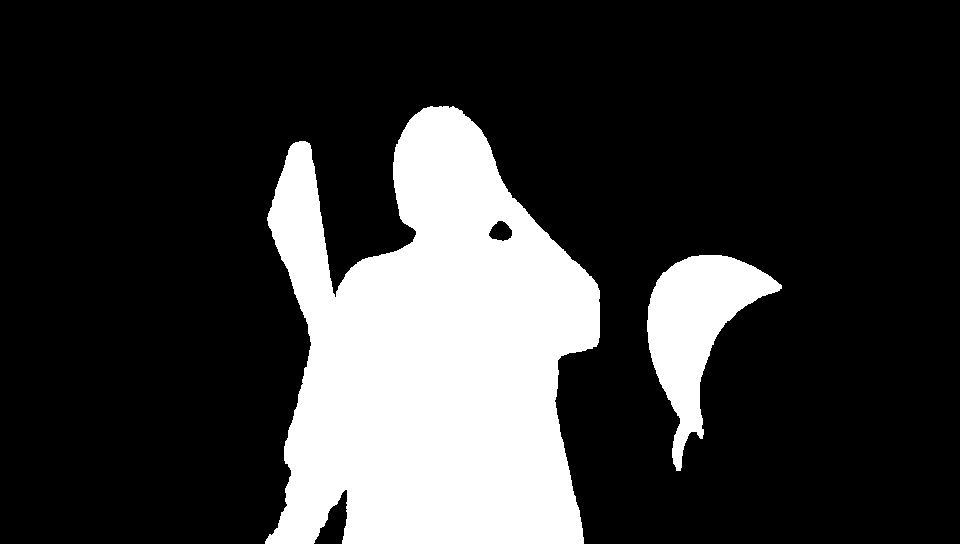}
	\end{subfigure}
	\begin{subfigure}{0.087\textwidth}
		\includegraphics[width=\textwidth]{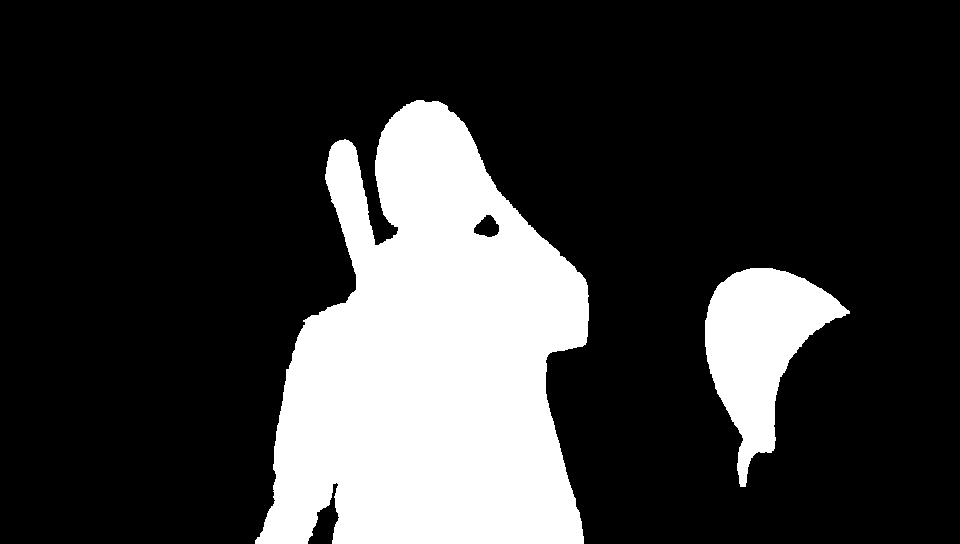}
	\end{subfigure}
	\begin{subfigure}{0.087\textwidth}
		\includegraphics[width=\textwidth]{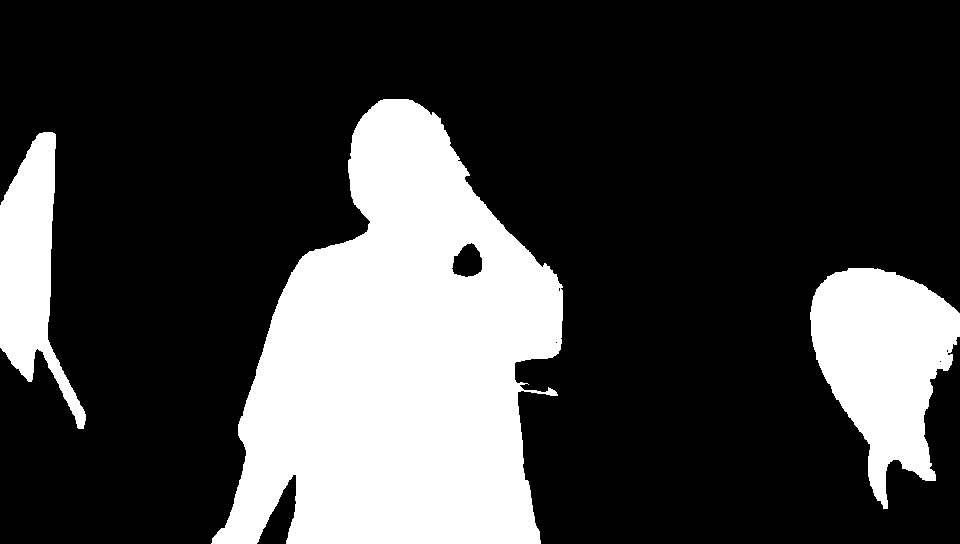}
	\end{subfigure}
	\begin{subfigure}{0.087\textwidth}
		\includegraphics[width=\textwidth]{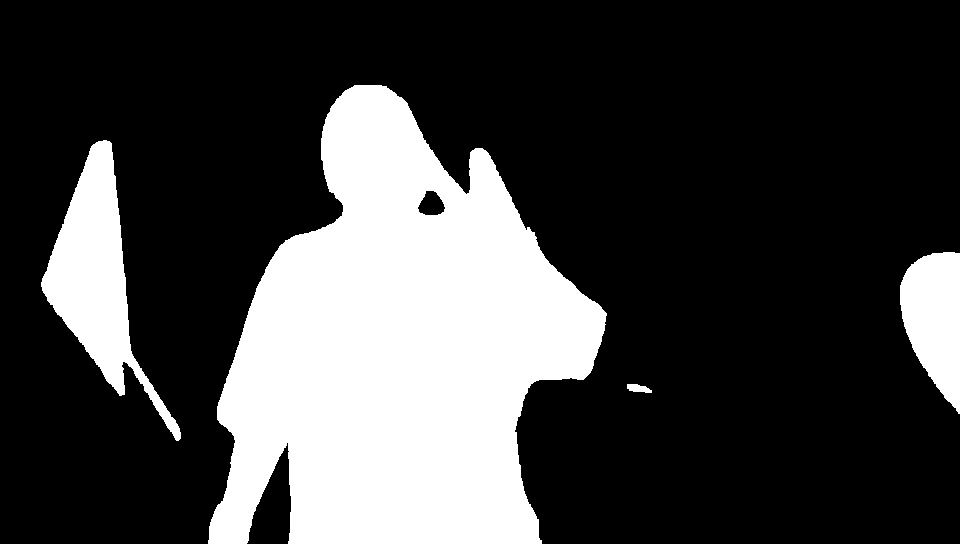}
	\end{subfigure}
	\begin{subfigure}{0.087\textwidth}
		\includegraphics[width=\textwidth]{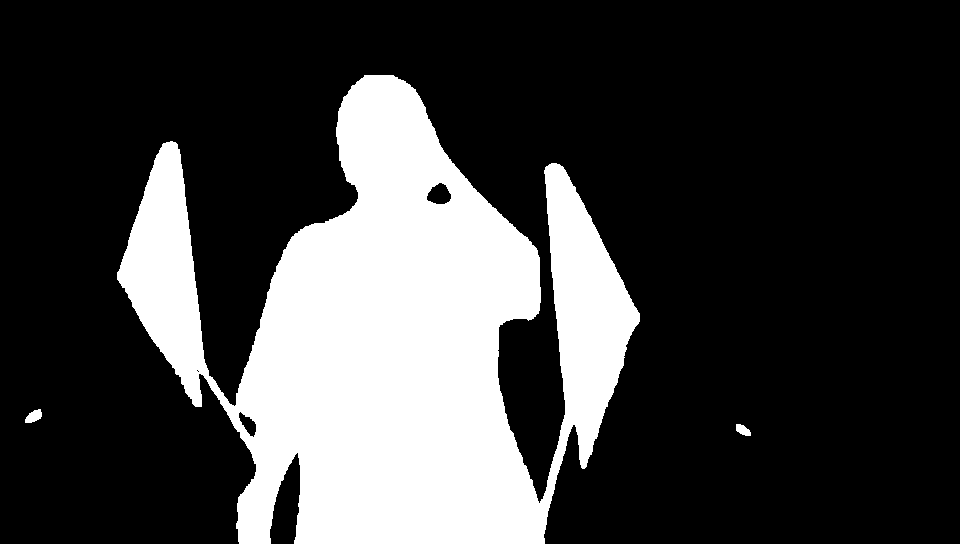}
	\end{subfigure}
	\begin{subfigure}{0.087\textwidth}
		\includegraphics[width=\textwidth]{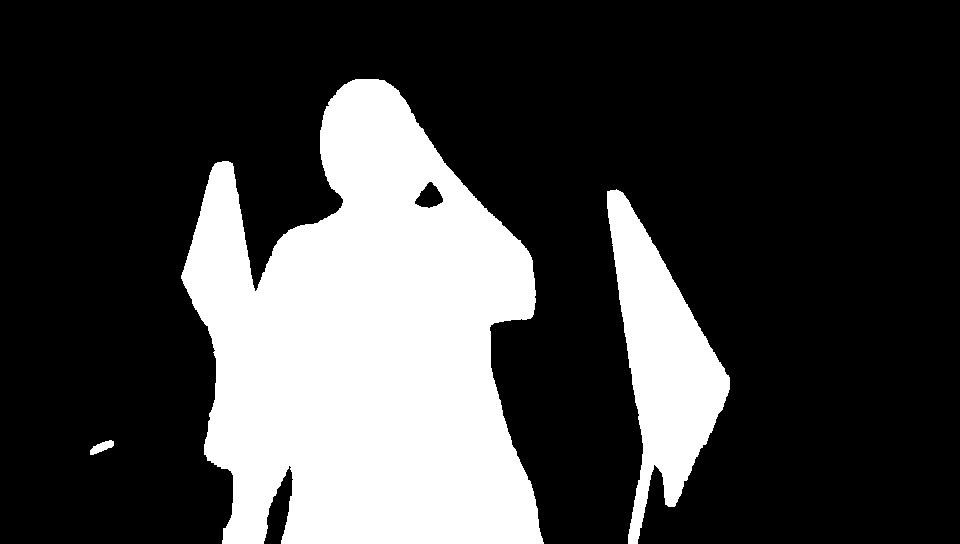}
	\end{subfigure}
	\begin{subfigure}{0.087\textwidth}
		\includegraphics[width=\textwidth]{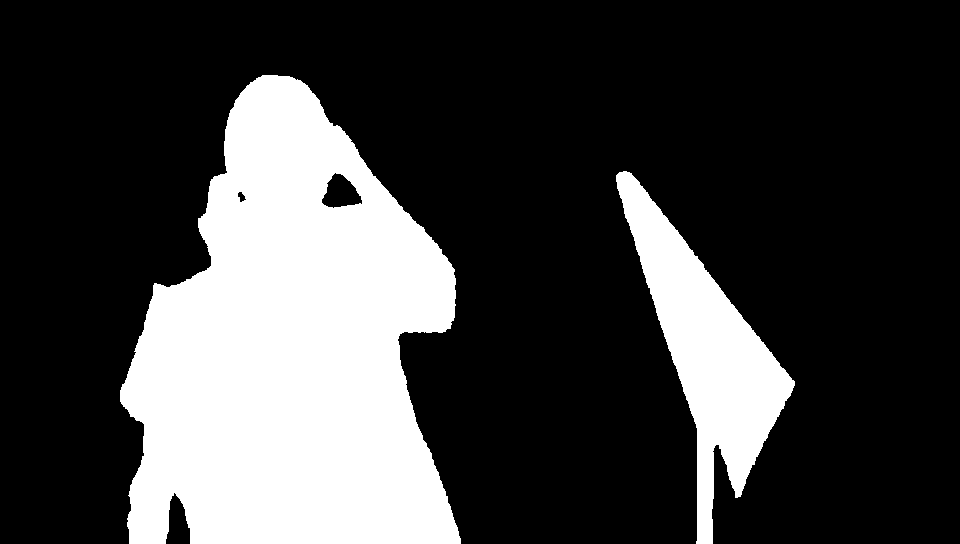}
	\end{subfigure}
	\begin{subfigure}{0.087\textwidth}
		\includegraphics[width=\textwidth]{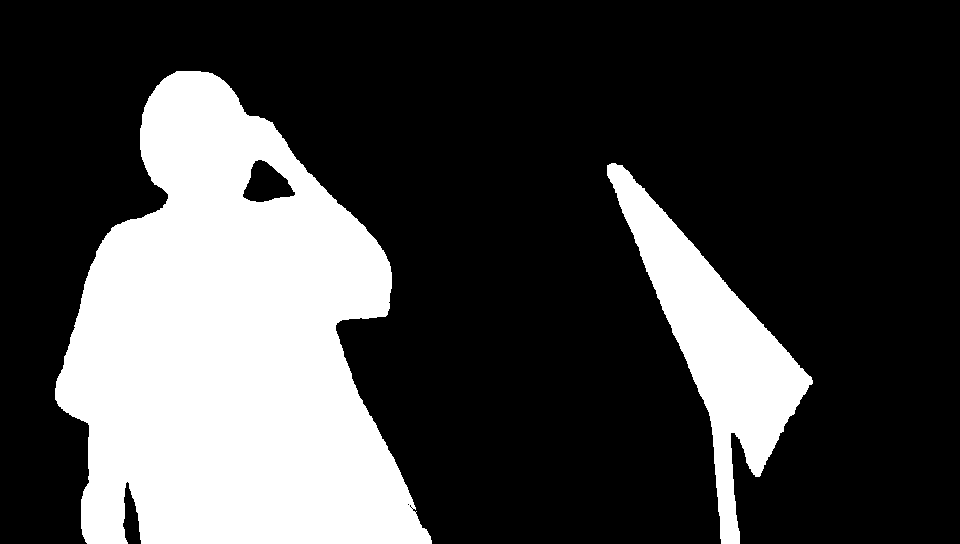}
	\end{subfigure}
	\begin{subfigure}{0.087\textwidth}
		\includegraphics[width=\textwidth]{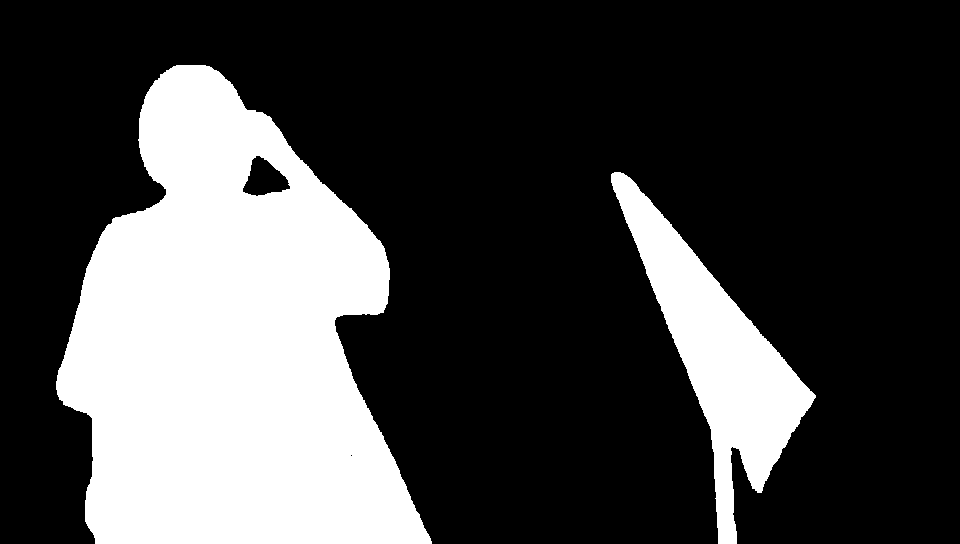}
	\end{subfigure}
	
	\vspace*{1.3mm}
	\begin{subfigure}{0.087\textwidth}
		\includegraphics[width=\textwidth]{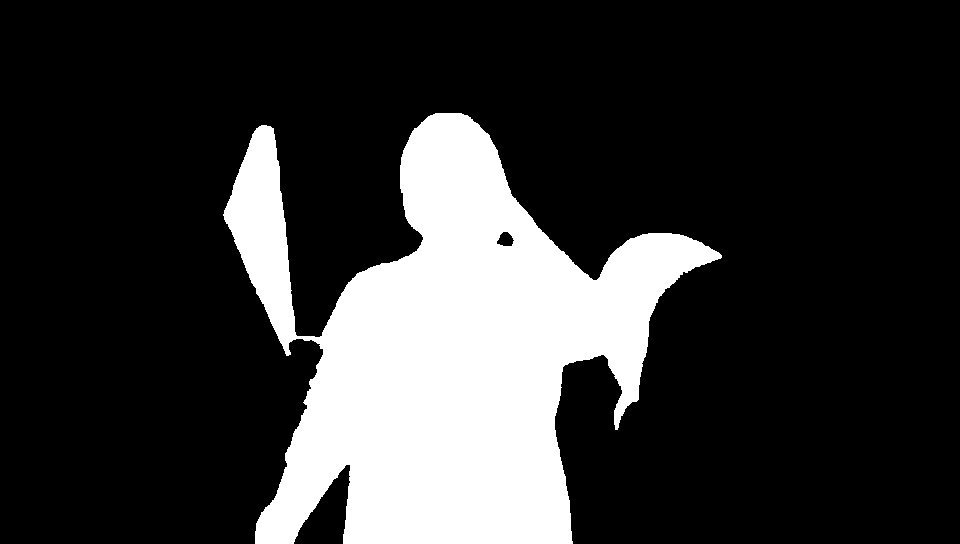}
	\end{subfigure}
	\begin{subfigure}{0.087\textwidth}
		\includegraphics[width=\textwidth]{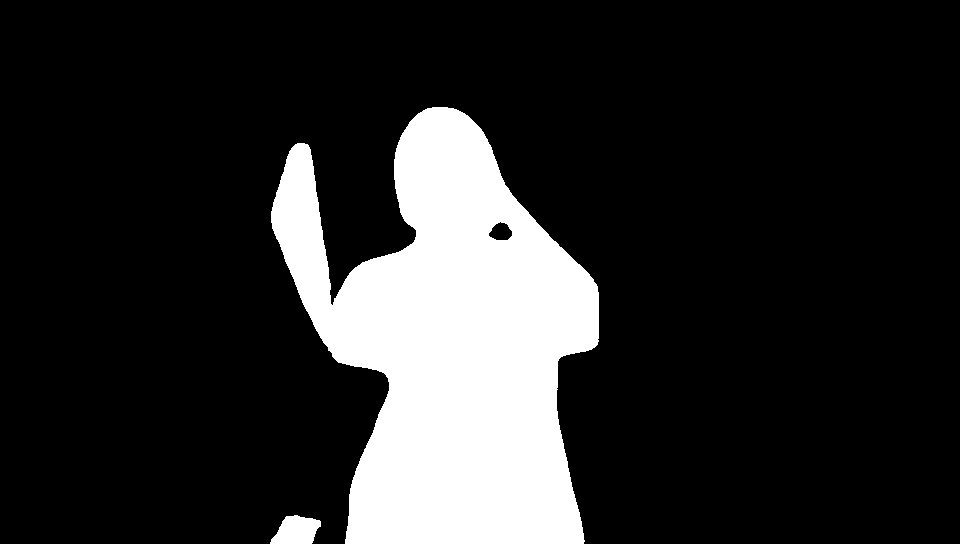}
	\end{subfigure}
	\begin{subfigure}{0.087\textwidth}
		\includegraphics[width=\textwidth]{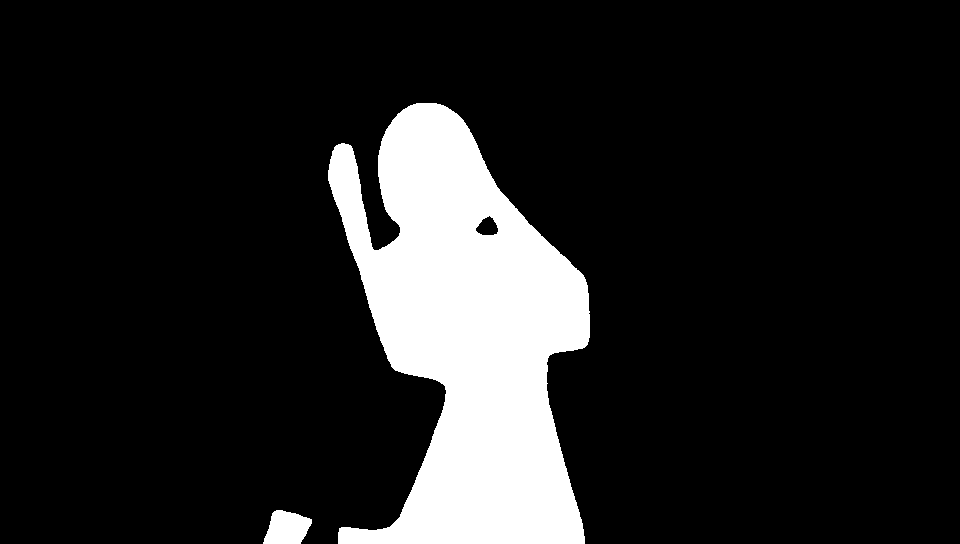}
	\end{subfigure}
	\begin{subfigure}{0.087\textwidth}
		\includegraphics[width=\textwidth]{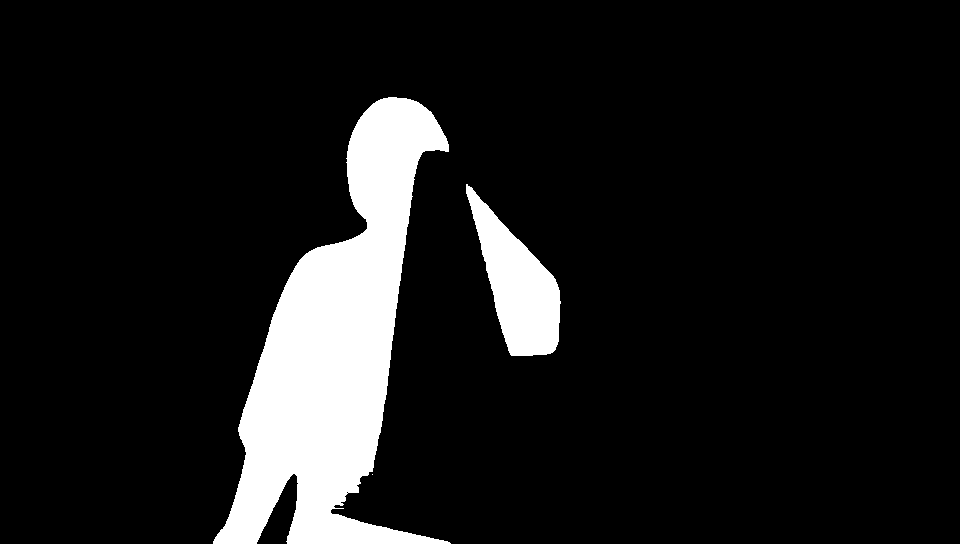}
	\end{subfigure}
	\begin{subfigure}{0.087\textwidth}
		\includegraphics[width=\textwidth]{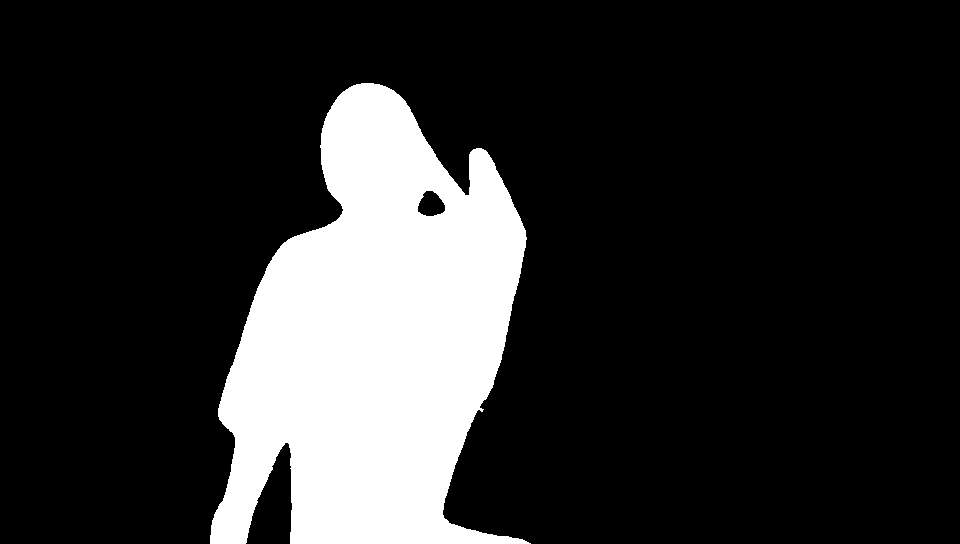}
	\end{subfigure}
	\begin{subfigure}{0.087\textwidth}
		\includegraphics[width=\textwidth]{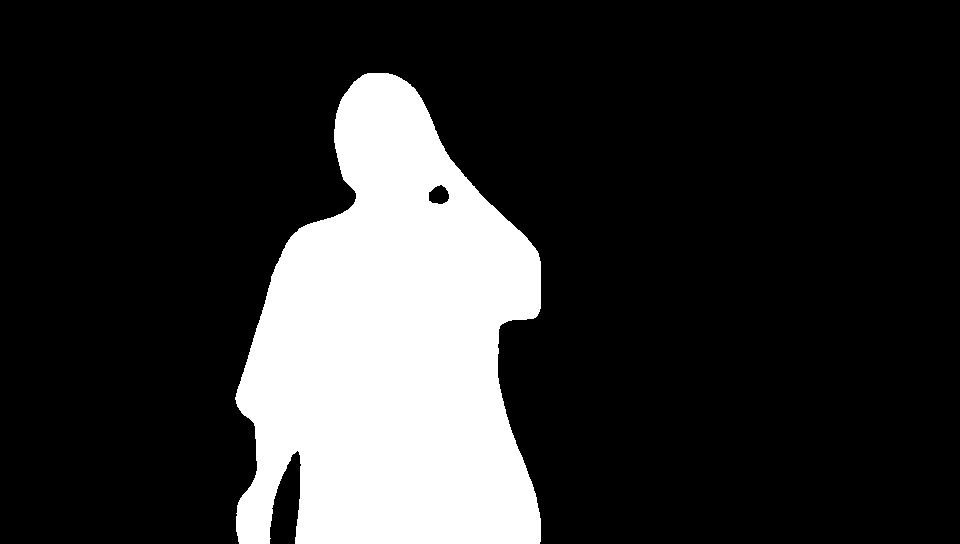}
	\end{subfigure}
	\begin{subfigure}{0.087\textwidth}
		\includegraphics[width=\textwidth]{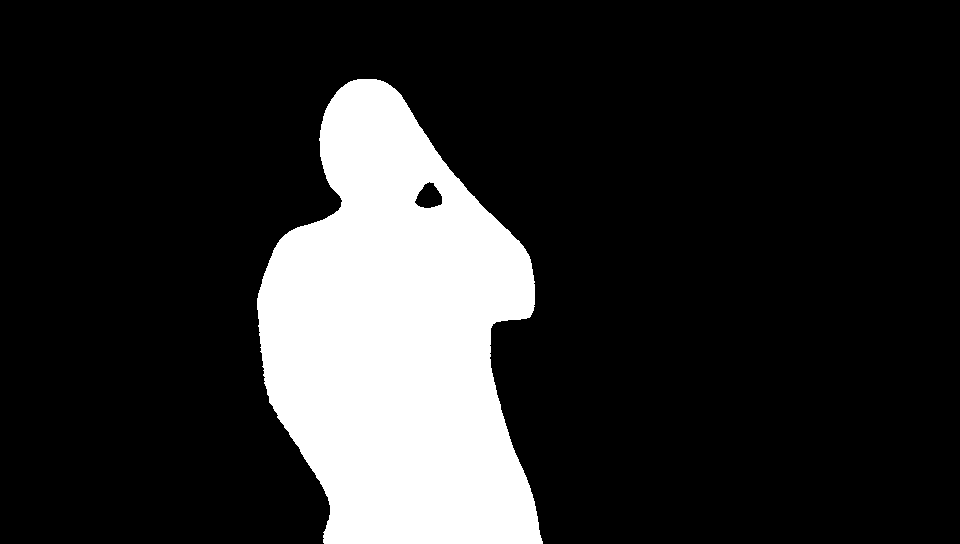}
	\end{subfigure}
	\begin{subfigure}{0.087\textwidth}
		\includegraphics[width=\textwidth]{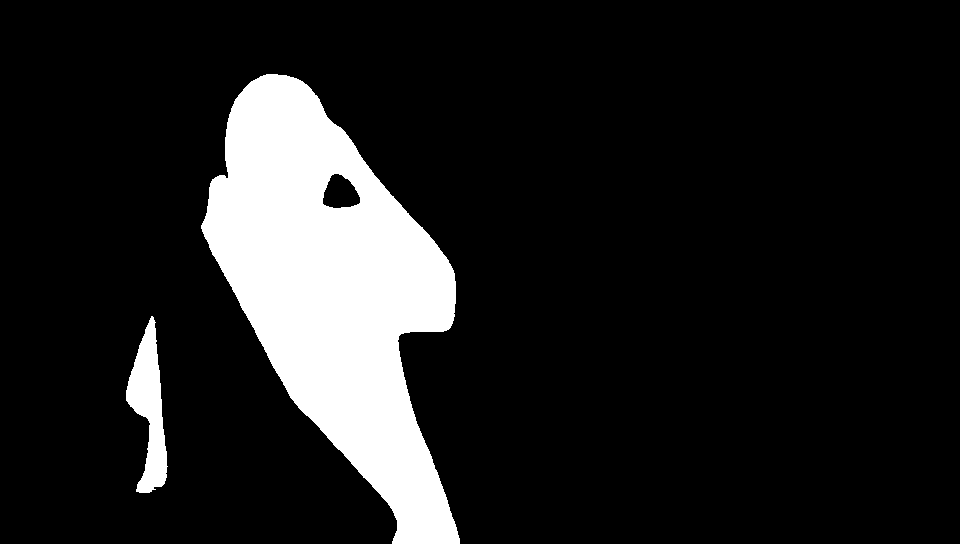}
	\end{subfigure}
	\begin{subfigure}{0.087\textwidth}
		\includegraphics[width=\textwidth]{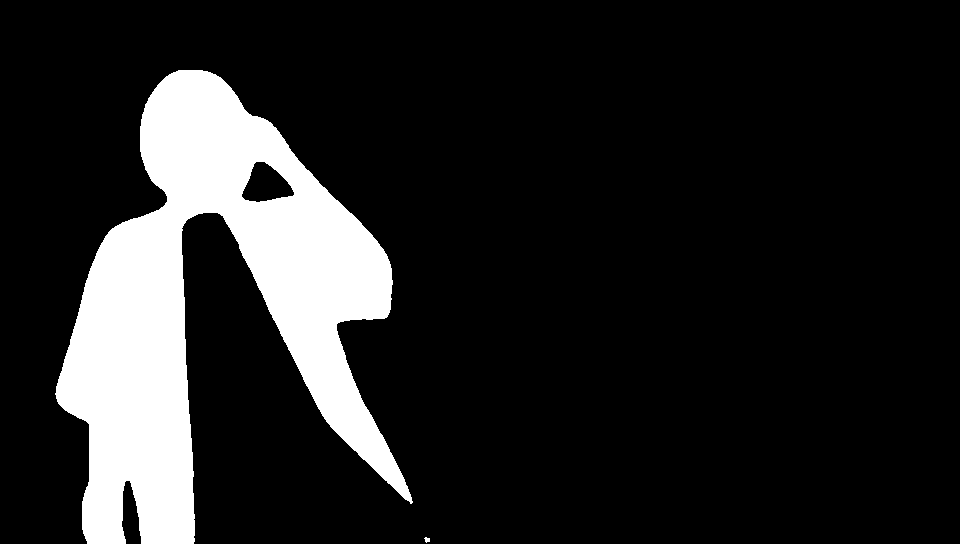}
	\end{subfigure}
	\begin{subfigure}{0.087\textwidth}
		\includegraphics[width=\textwidth]{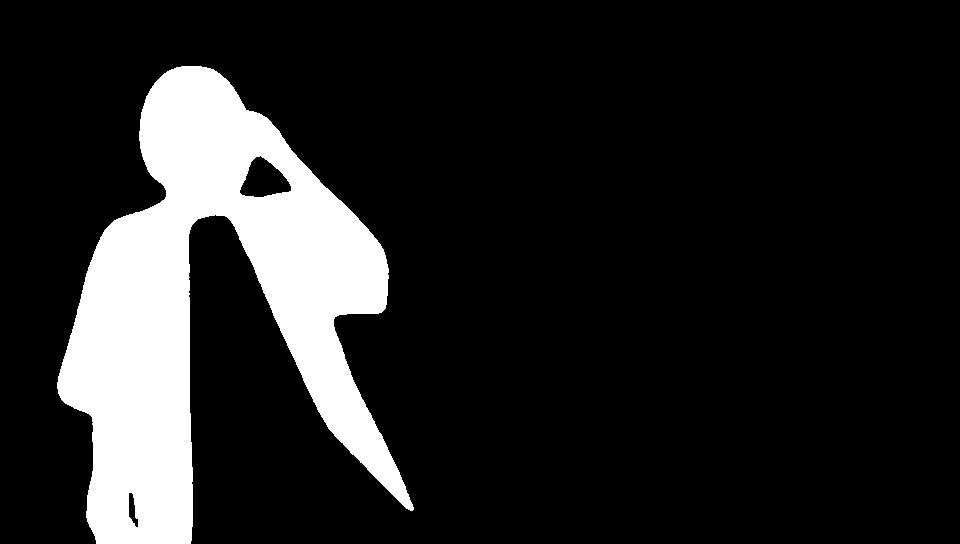}
	\end{subfigure}

	\vspace*{1.3mm}
	\begin{subfigure}{0.087\textwidth}
		\includegraphics[width=\textwidth]{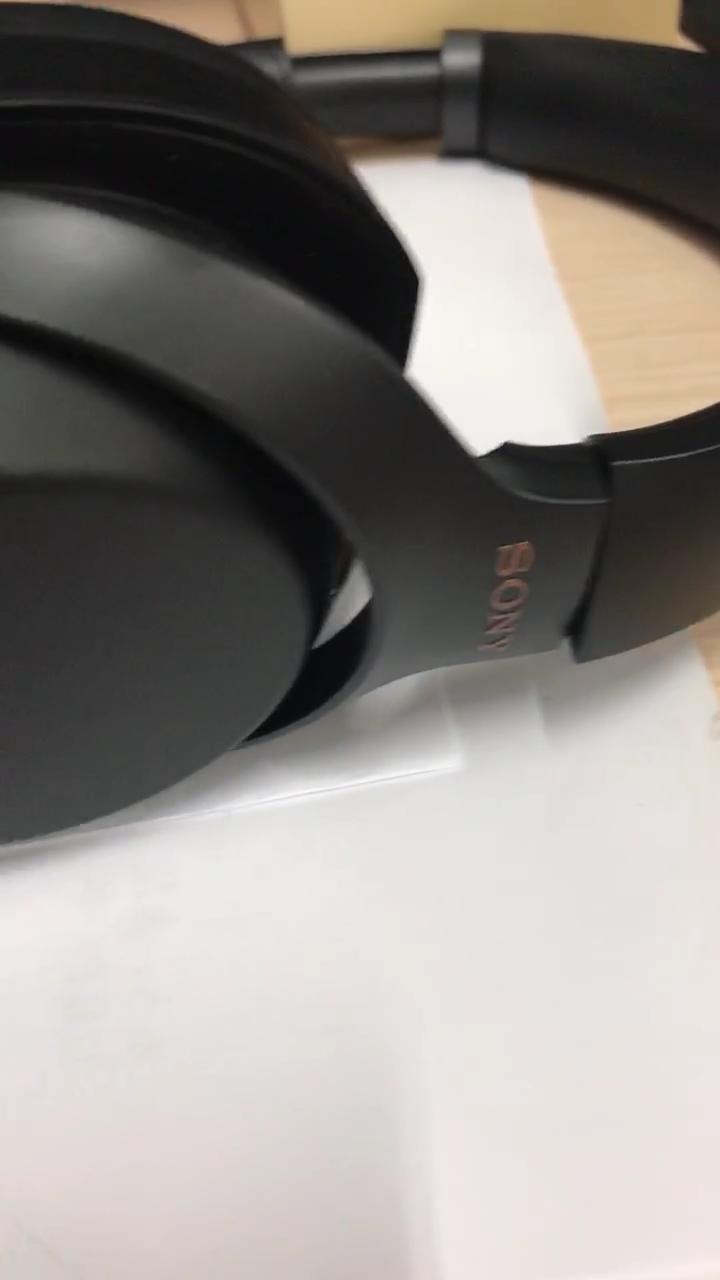}
	\end{subfigure}
	\begin{subfigure}{0.087\textwidth}
		\includegraphics[width=\textwidth]{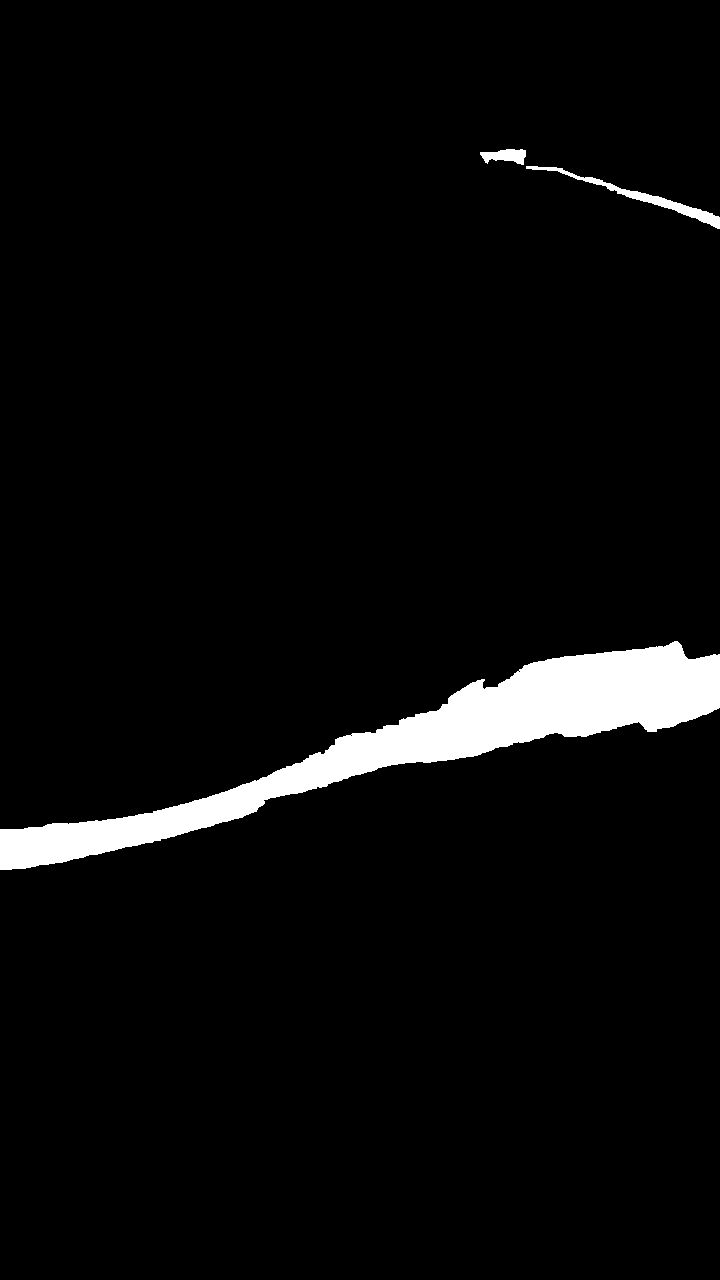}
	\end{subfigure}
	\begin{subfigure}{0.087\textwidth}
		\includegraphics[width=\textwidth]{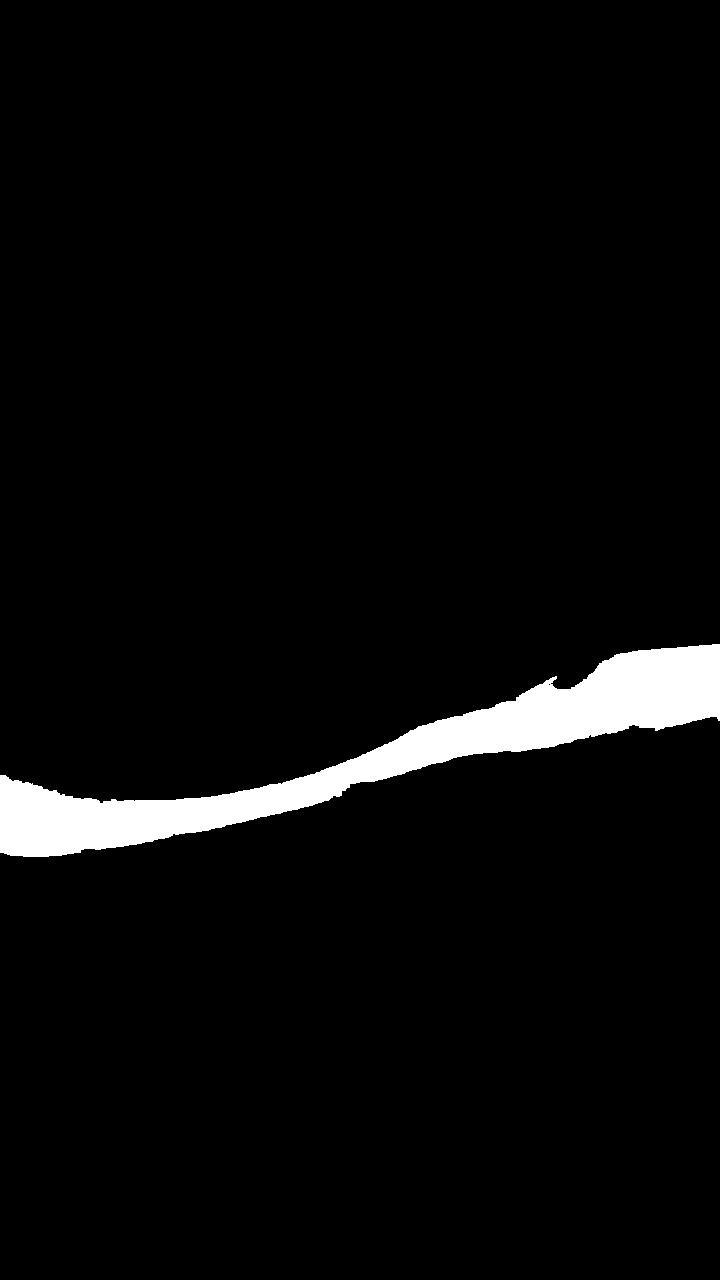}
	\end{subfigure}
	\begin{subfigure}{0.087\textwidth}
		\includegraphics[width=\textwidth]{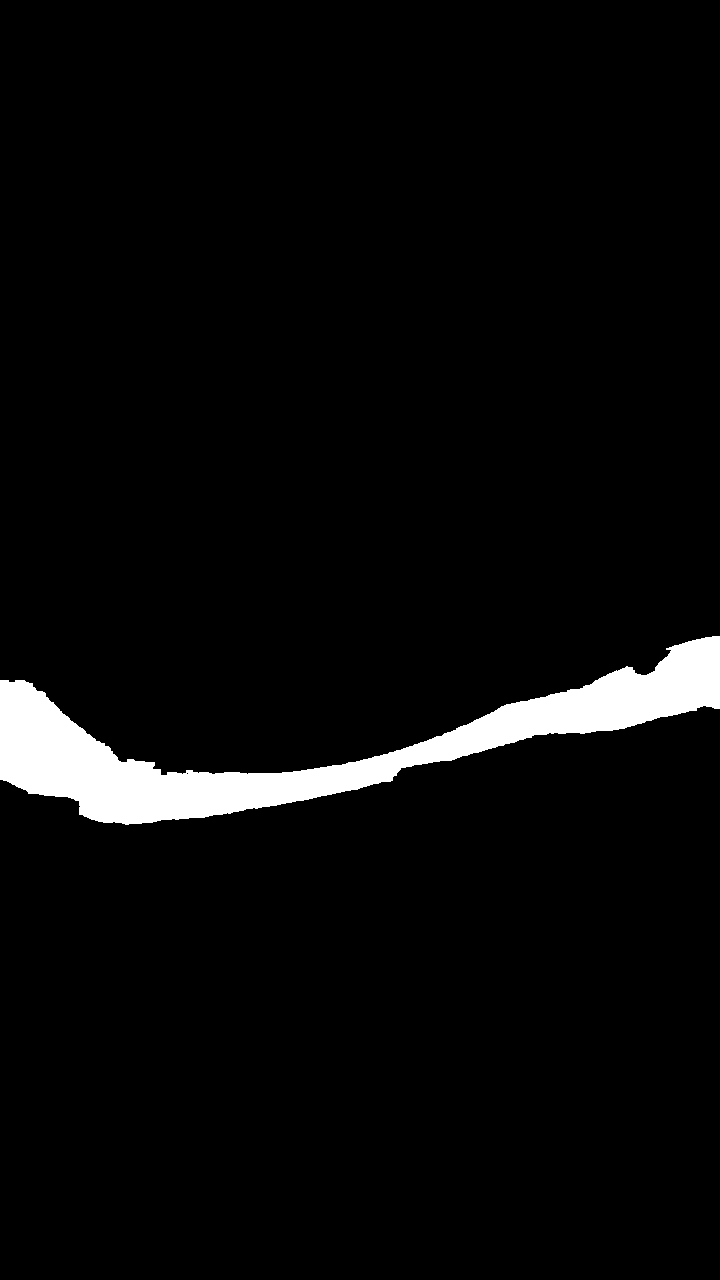}
	\end{subfigure}
	\begin{subfigure}{0.087\textwidth}
		\includegraphics[width=\textwidth]{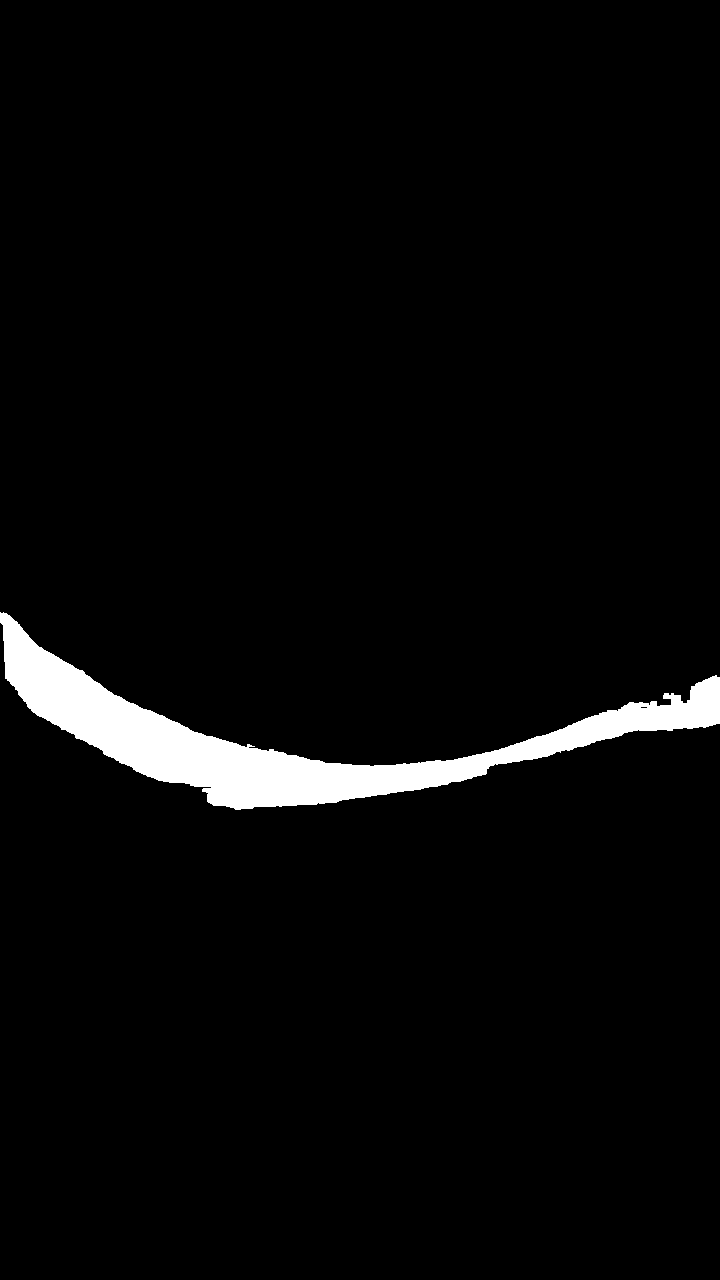}
	\end{subfigure}
	\begin{subfigure}{0.087\textwidth}
		\includegraphics[width=\textwidth]{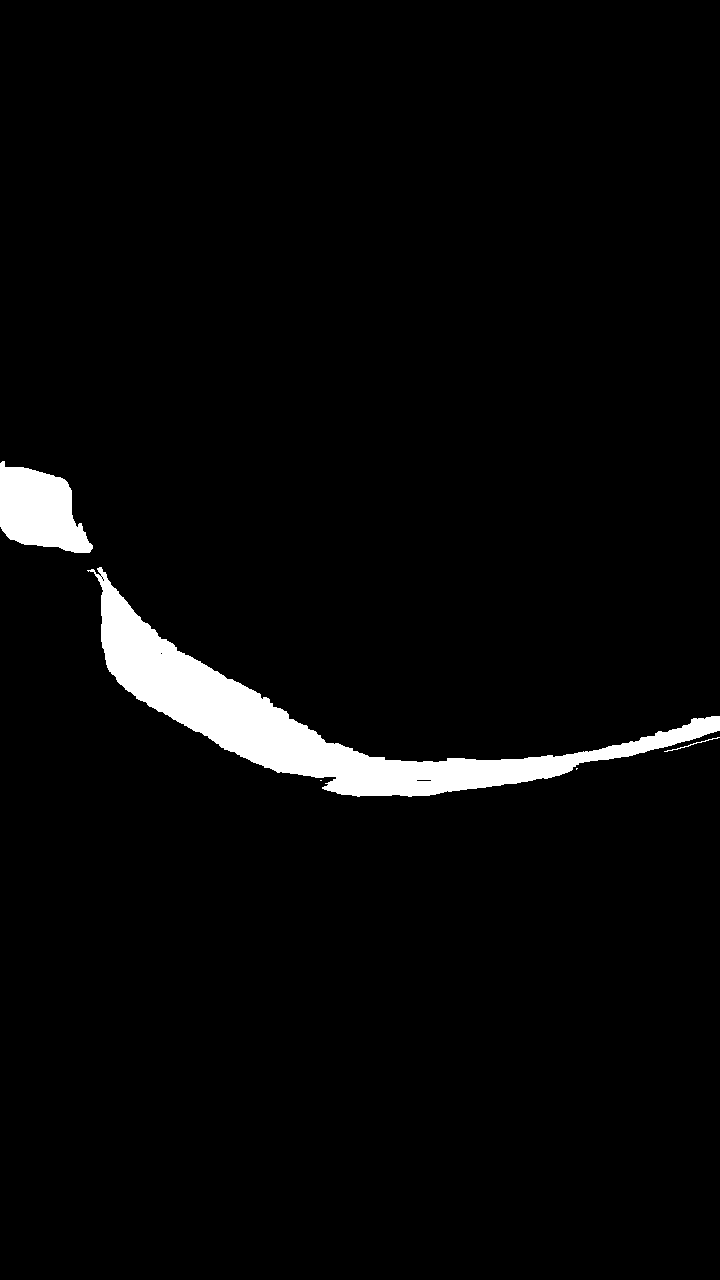}
	\end{subfigure}
	\begin{subfigure}{0.087\textwidth}
		\includegraphics[width=\textwidth]{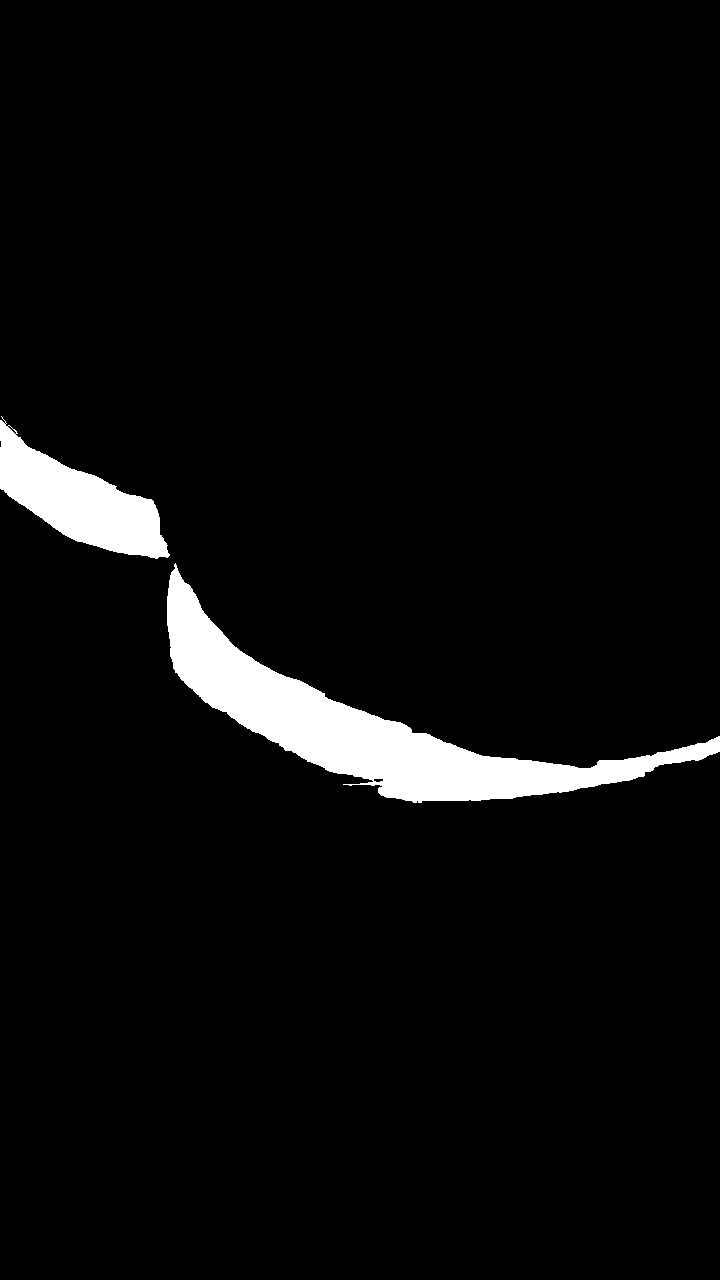}
	\end{subfigure}
	\begin{subfigure}{0.087\textwidth}
		\includegraphics[width=\textwidth]{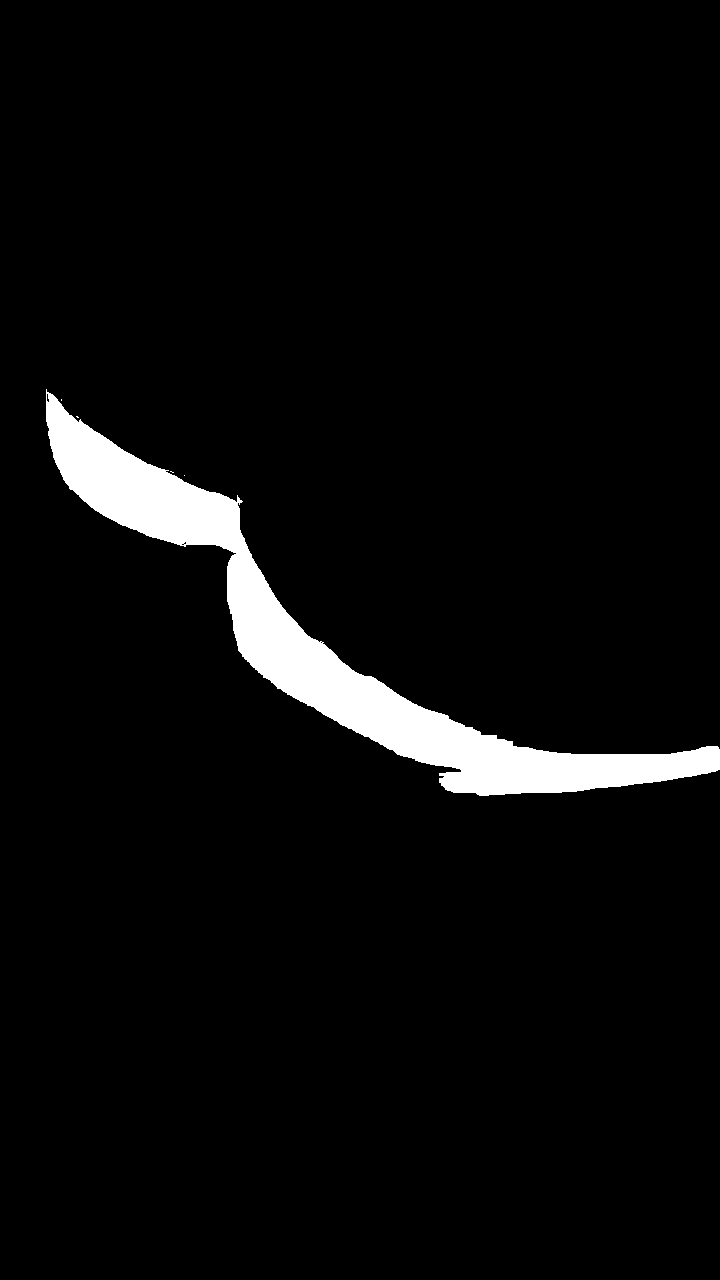}
	\end{subfigure}
	\begin{subfigure}{0.087\textwidth}
		\includegraphics[width=\textwidth]{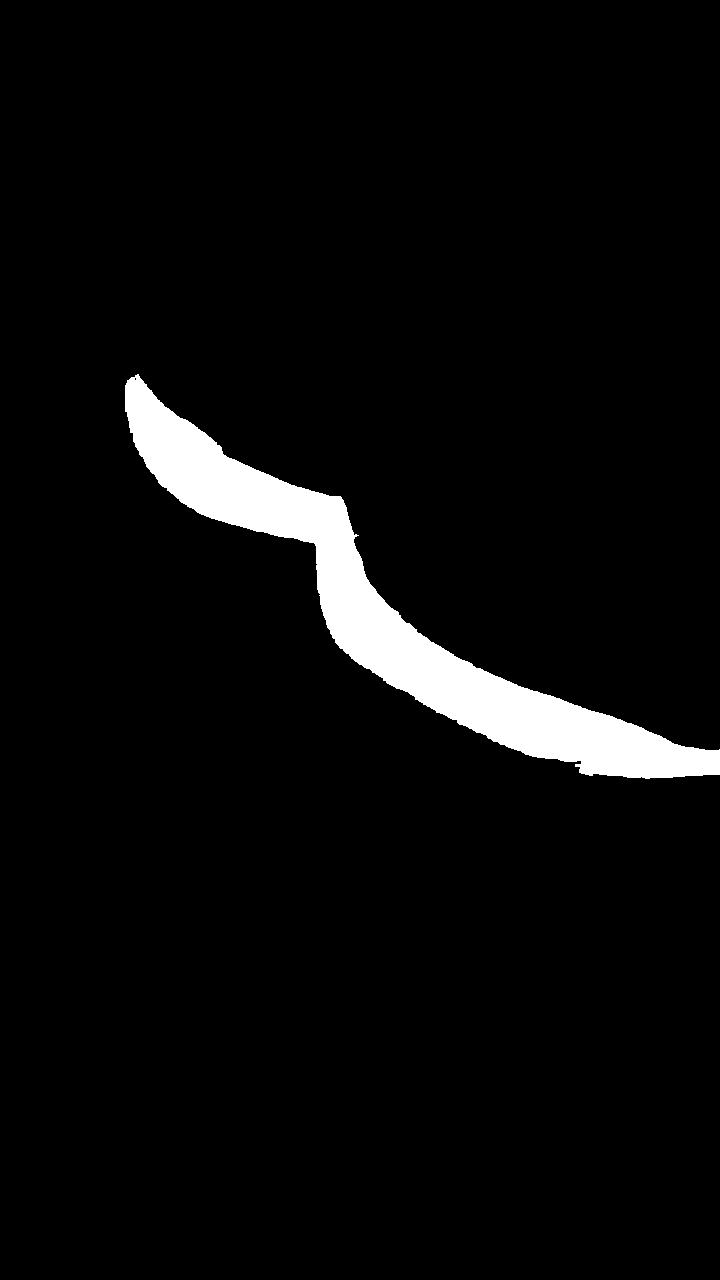}
	\end{subfigure}
	\begin{subfigure}{0.087\textwidth}
		\includegraphics[width=\textwidth]{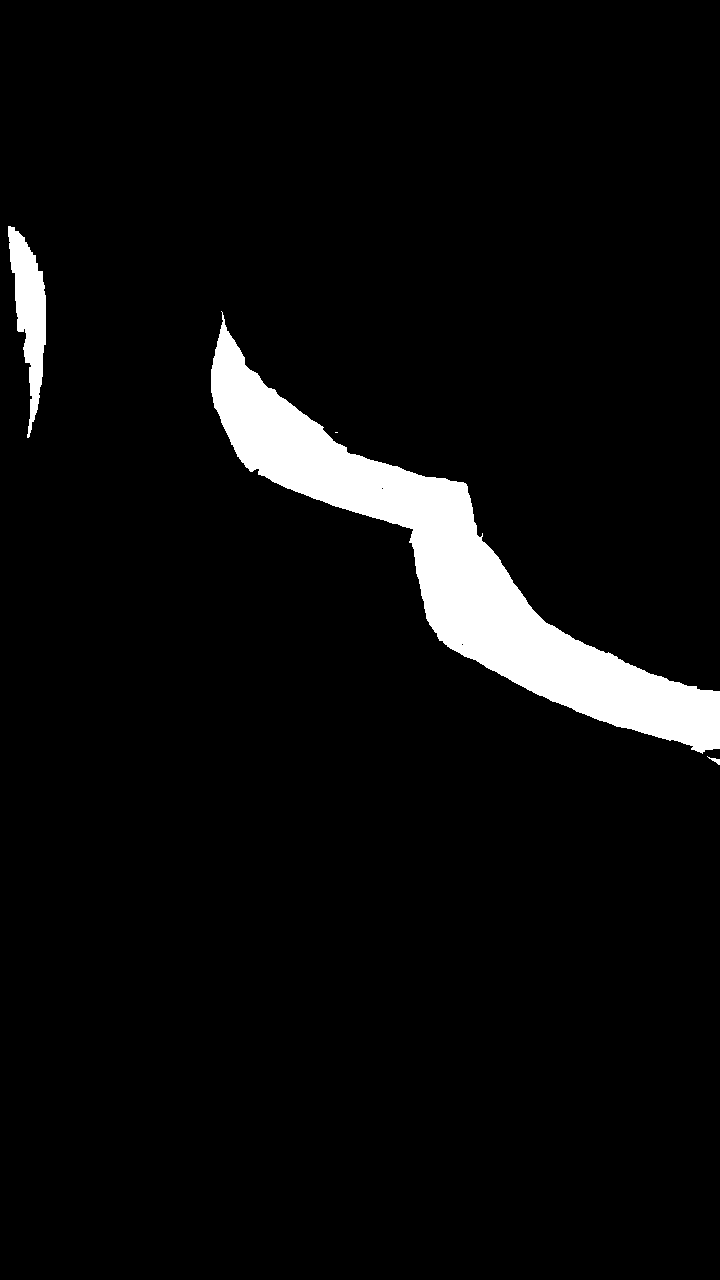}
	\end{subfigure}
	
	\vspace*{1.3mm}
	\begin{subfigure}{0.087\textwidth}
		\includegraphics[width=\textwidth]{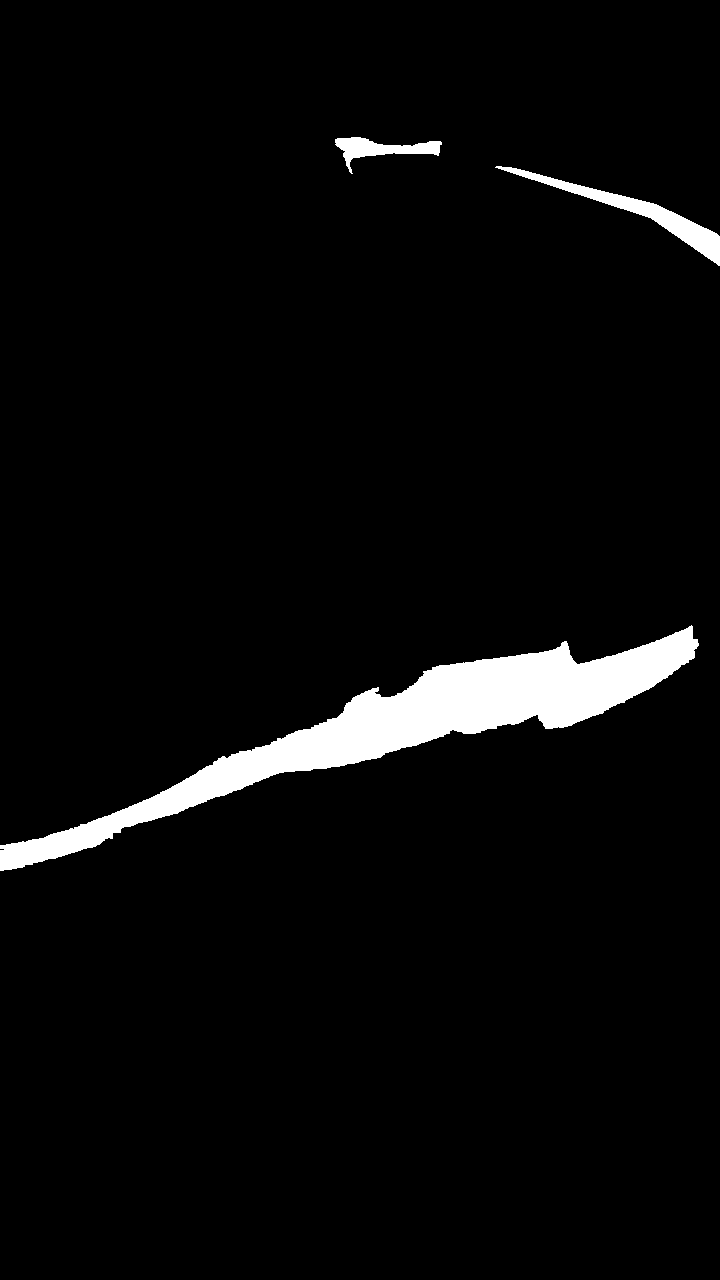}
		\captionsetup{justification=centering}
        \vspace{-5.5mm} \caption{\footnotesize{\\RGB/GT}}
	\end{subfigure}
	\begin{subfigure}{0.087\textwidth}
		\includegraphics[width=\textwidth]{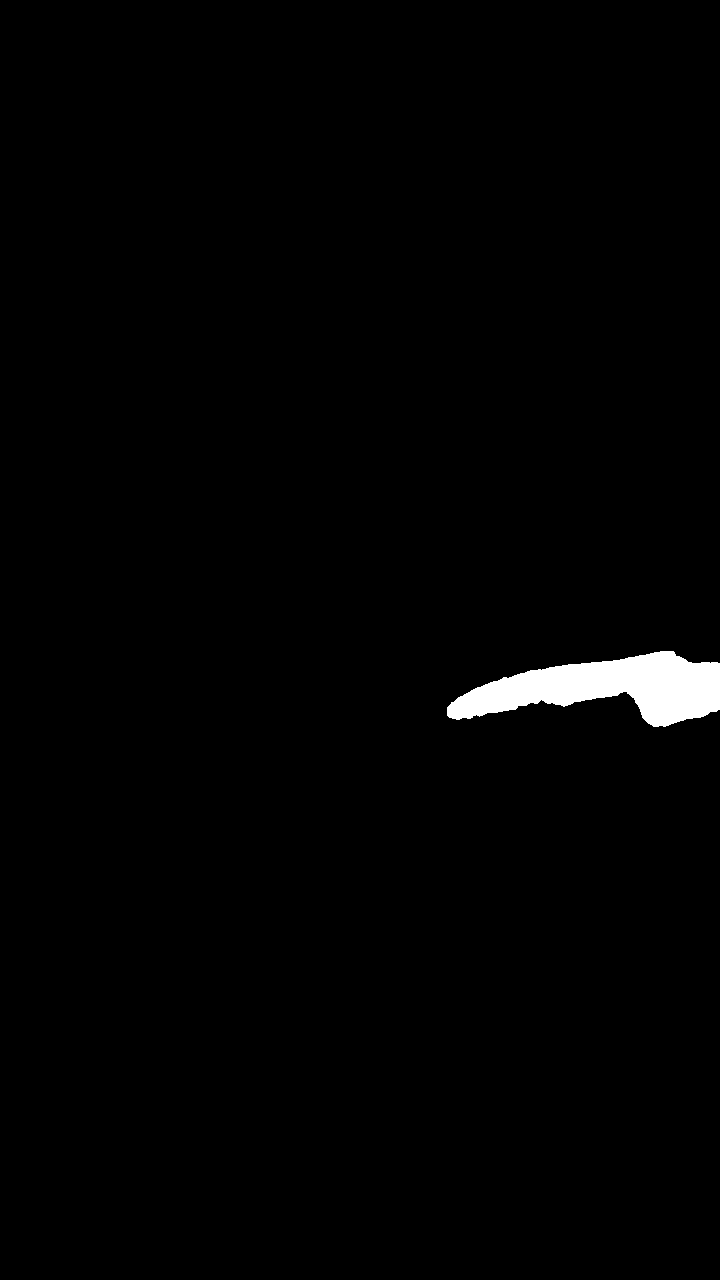}
		\captionsetup{justification=centering}
        \vspace{-5.5mm} \caption{\footnotesize{\\11th}}
	\end{subfigure}
	\begin{subfigure}{0.087\textwidth}
		\includegraphics[width=\textwidth]{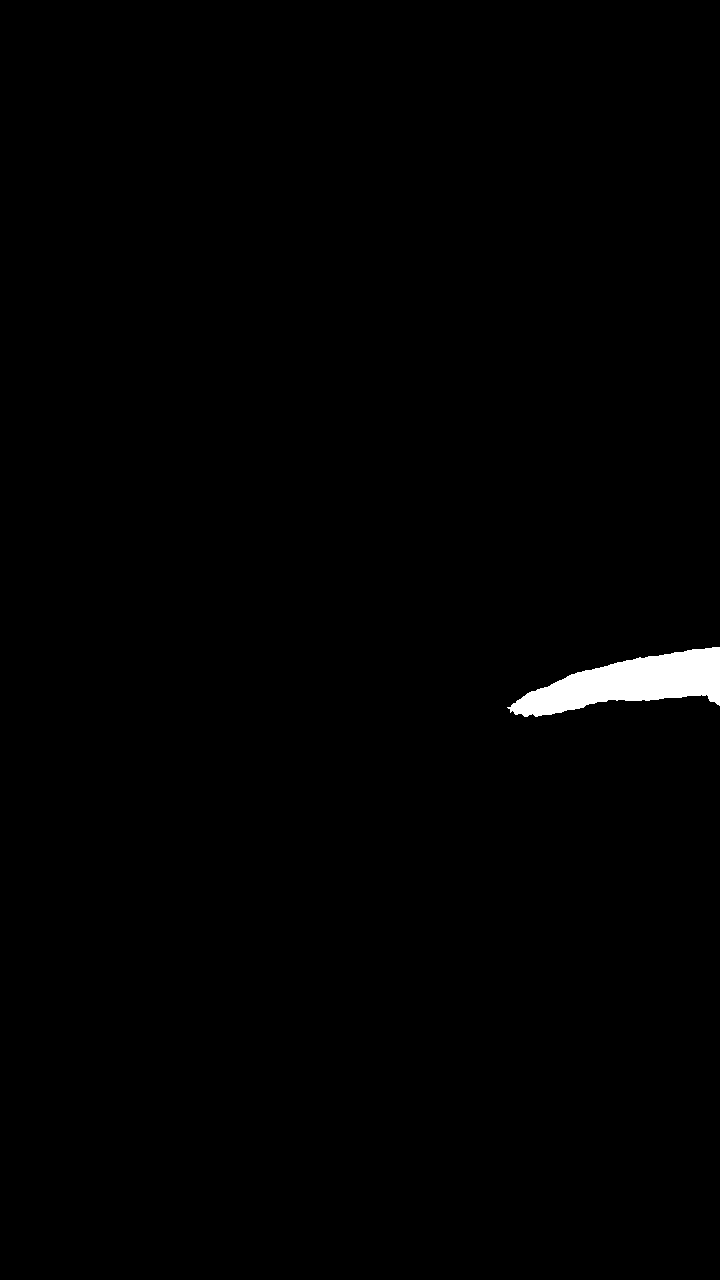}
		\captionsetup{justification=centering}
        \vspace{-5.5mm} \caption{\footnotesize{\\21th}}
	\end{subfigure}
	\begin{subfigure}{0.087\textwidth}
		\includegraphics[width=\textwidth]{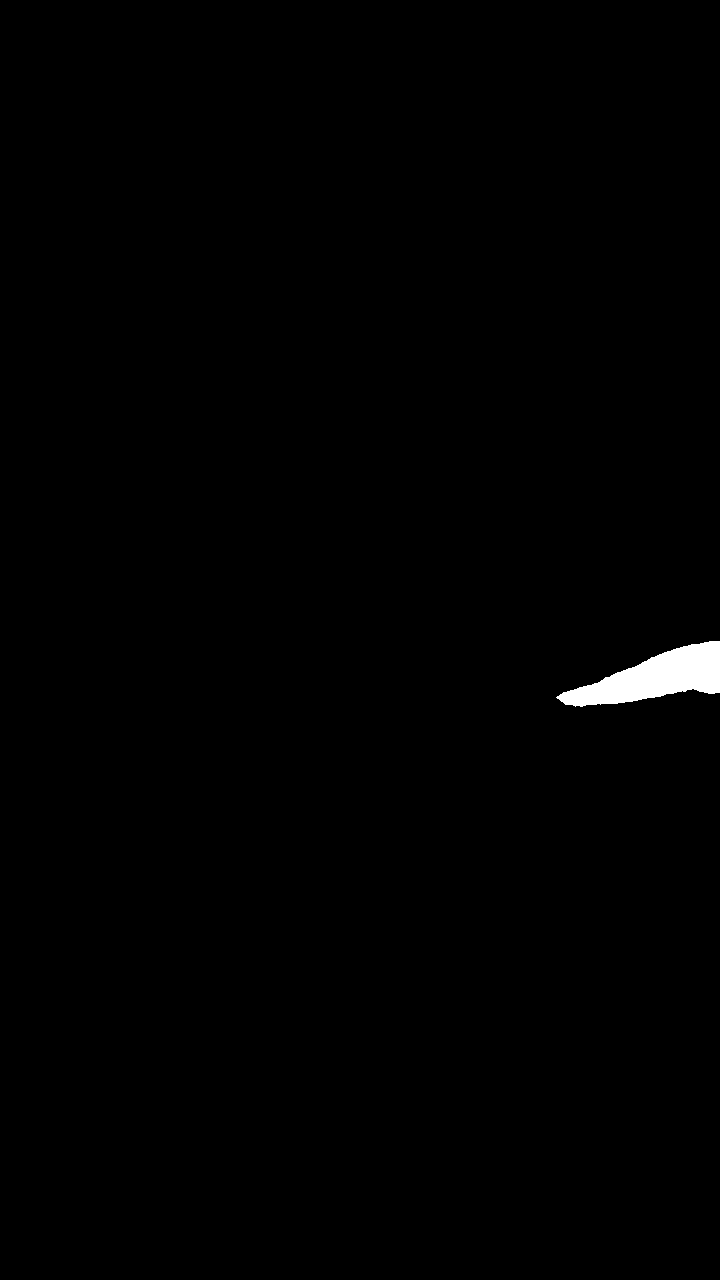}
		\captionsetup{justification=centering}
        \vspace{-5.5mm} \caption{\footnotesize{\\31th}}
	\end{subfigure}
	\begin{subfigure}{0.087\textwidth}
		\includegraphics[width=\textwidth]{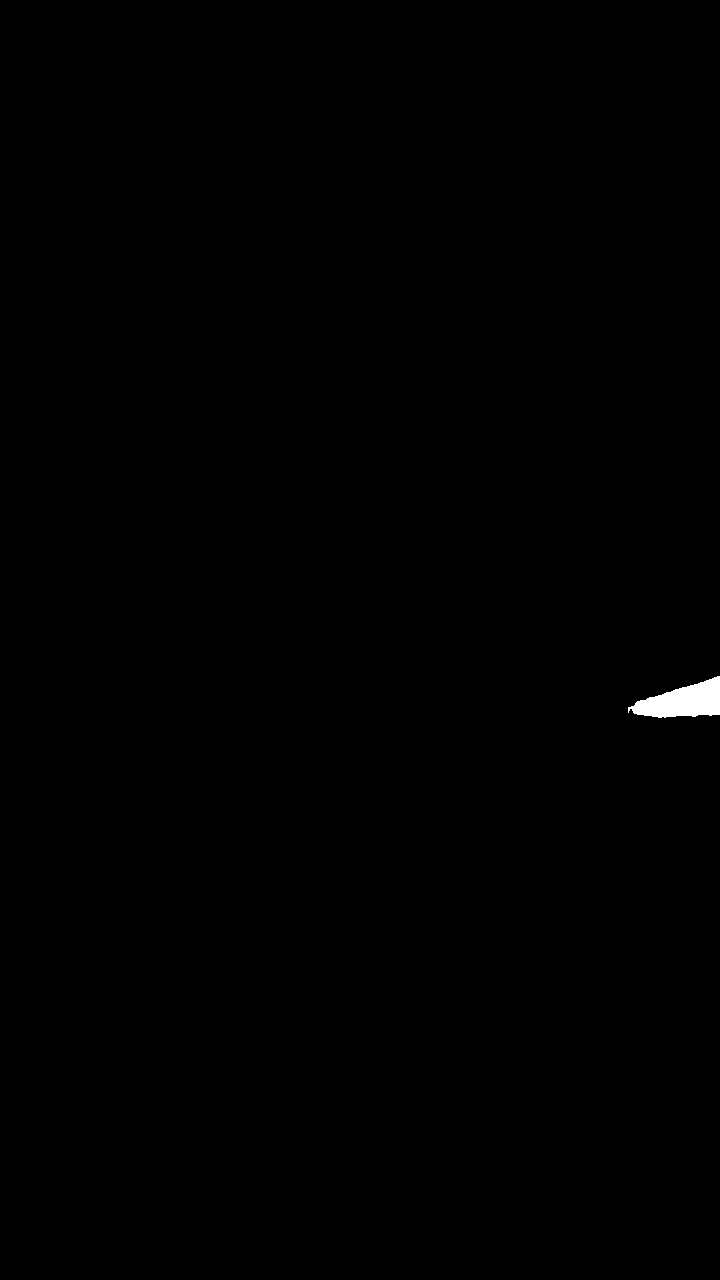}
		\captionsetup{justification=centering}
        \vspace{-5.5mm} \caption{\footnotesize{\\41th}}
	\end{subfigure}
	\begin{subfigure}{0.087\textwidth}
		\includegraphics[width=\textwidth]{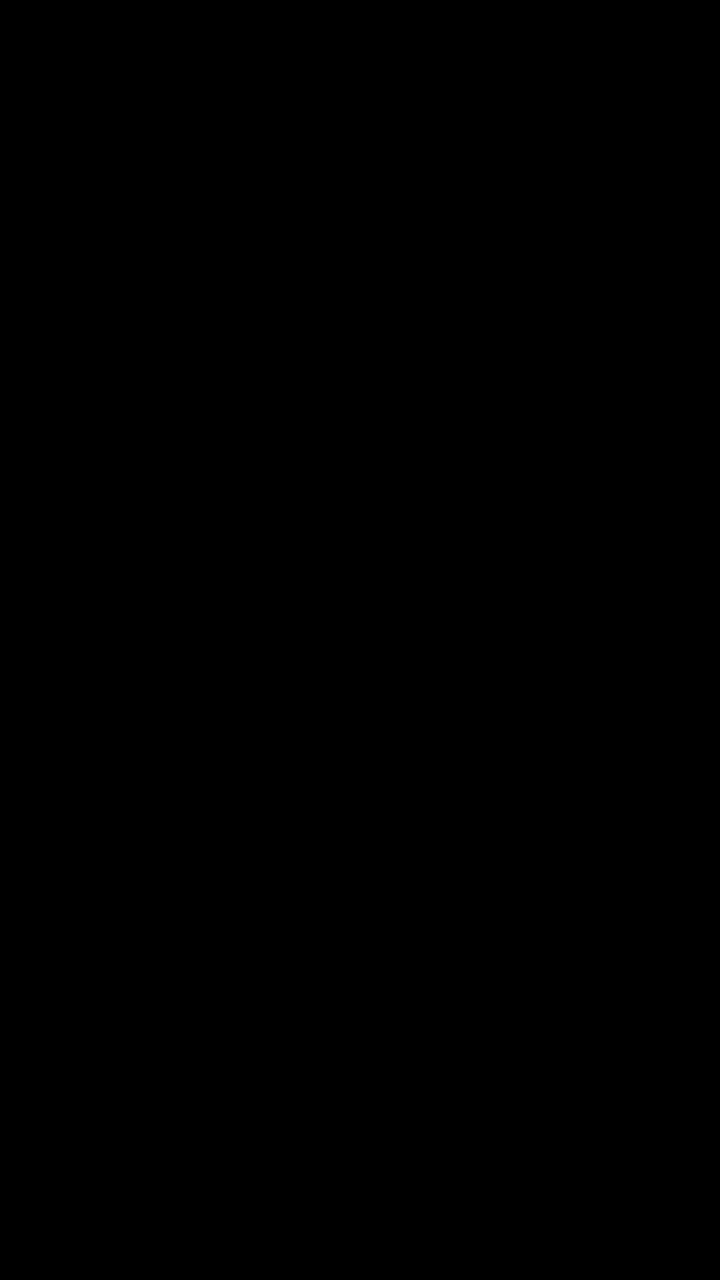}
		\captionsetup{justification=centering}
        \vspace{-5.5mm} \caption{\footnotesize{\\51th}}
	\end{subfigure}
	\begin{subfigure}{0.087\textwidth}
		\includegraphics[width=\textwidth]{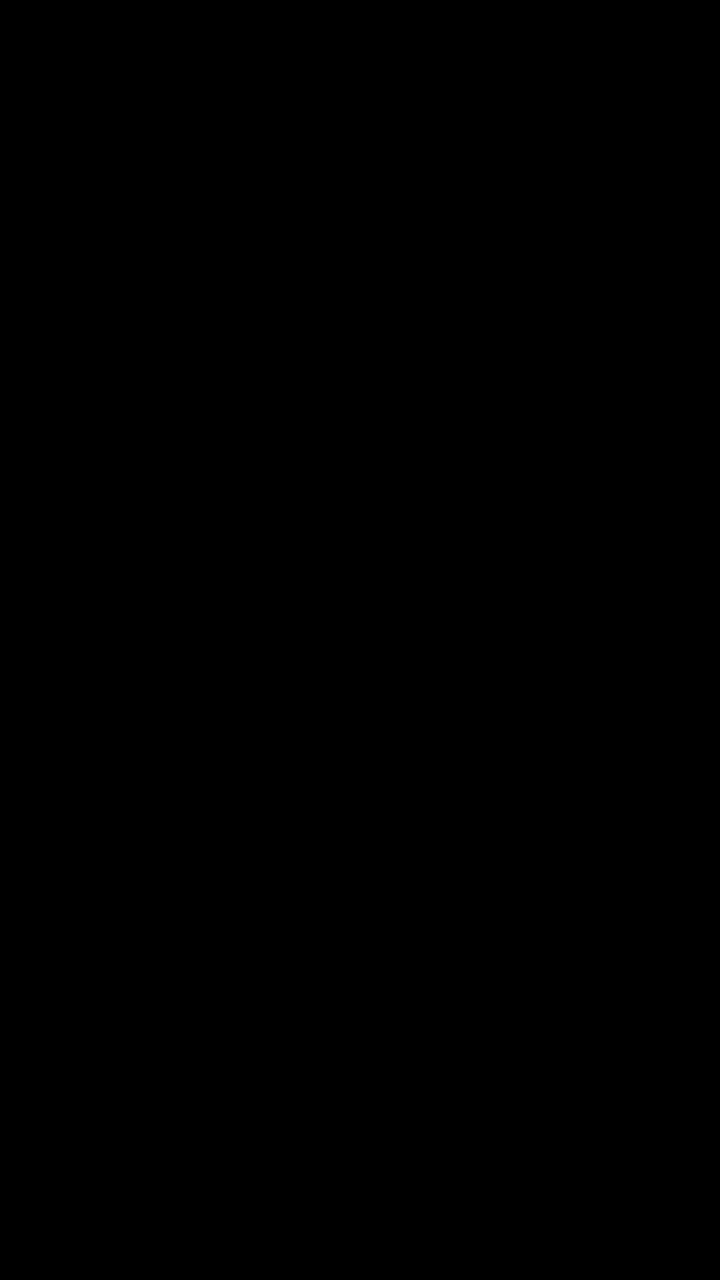}
		\captionsetup{justification=centering}
        \vspace{-5.5mm} \caption{\footnotesize{\\61th}}
	\end{subfigure}
	\begin{subfigure}{0.087\textwidth}
		\includegraphics[width=\textwidth]{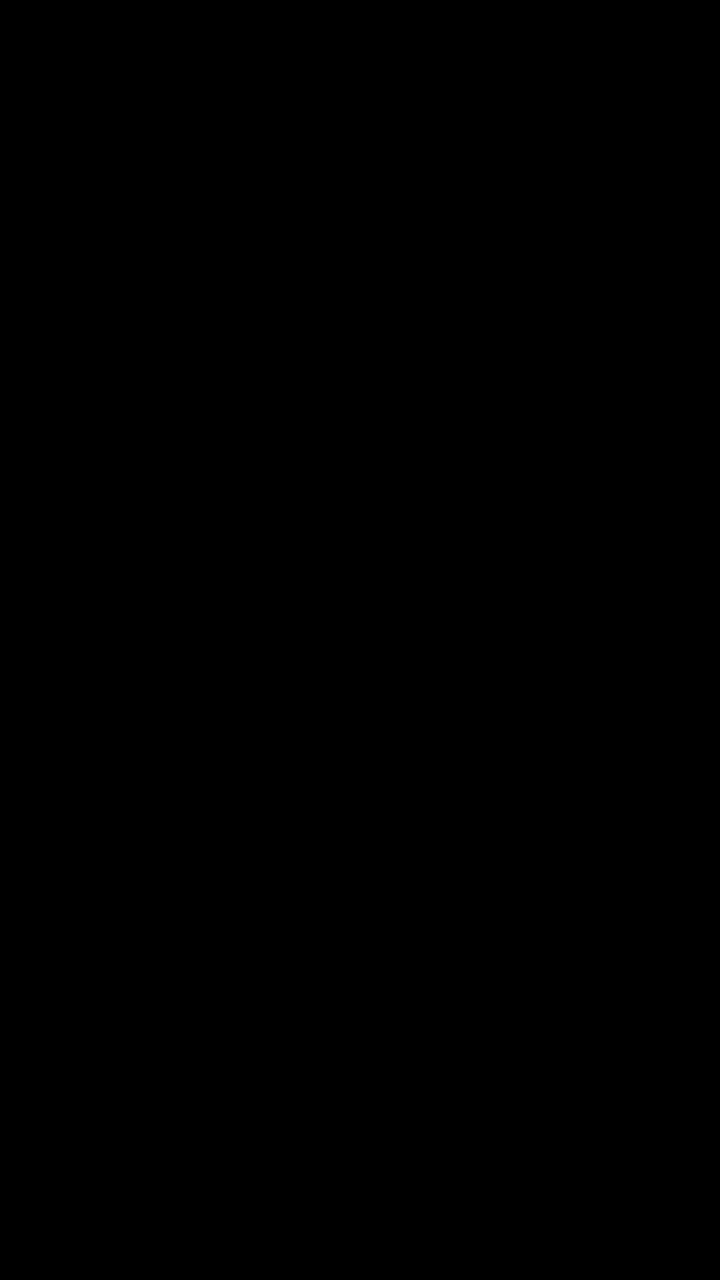}
		\captionsetup{justification=centering}
        \vspace{-5.5mm} \caption{\footnotesize{\\71th}}
	\end{subfigure}
	\begin{subfigure}{0.087\textwidth}
		\includegraphics[width=\textwidth]{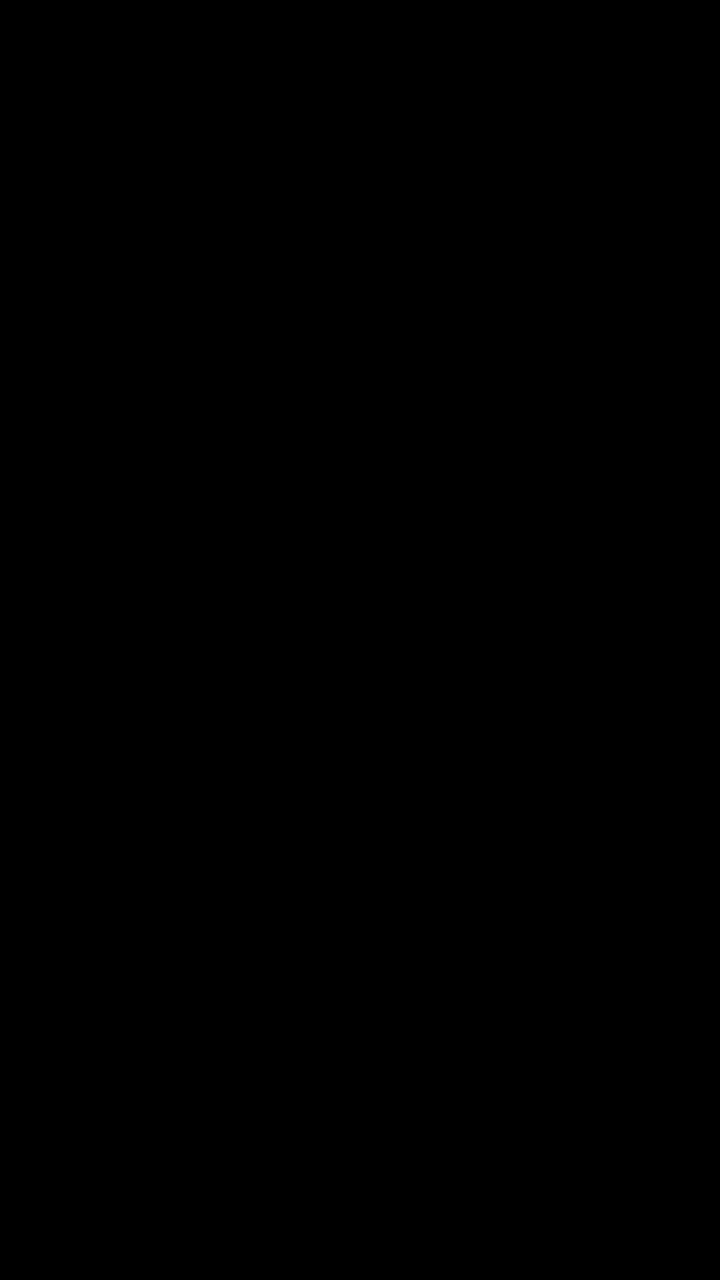}
		\captionsetup{justification=centering}
        \vspace{-5.5mm} \caption{\footnotesize{\\81th}}
	\end{subfigure}
	\begin{subfigure}{0.087\textwidth}
		\includegraphics[width=\textwidth]{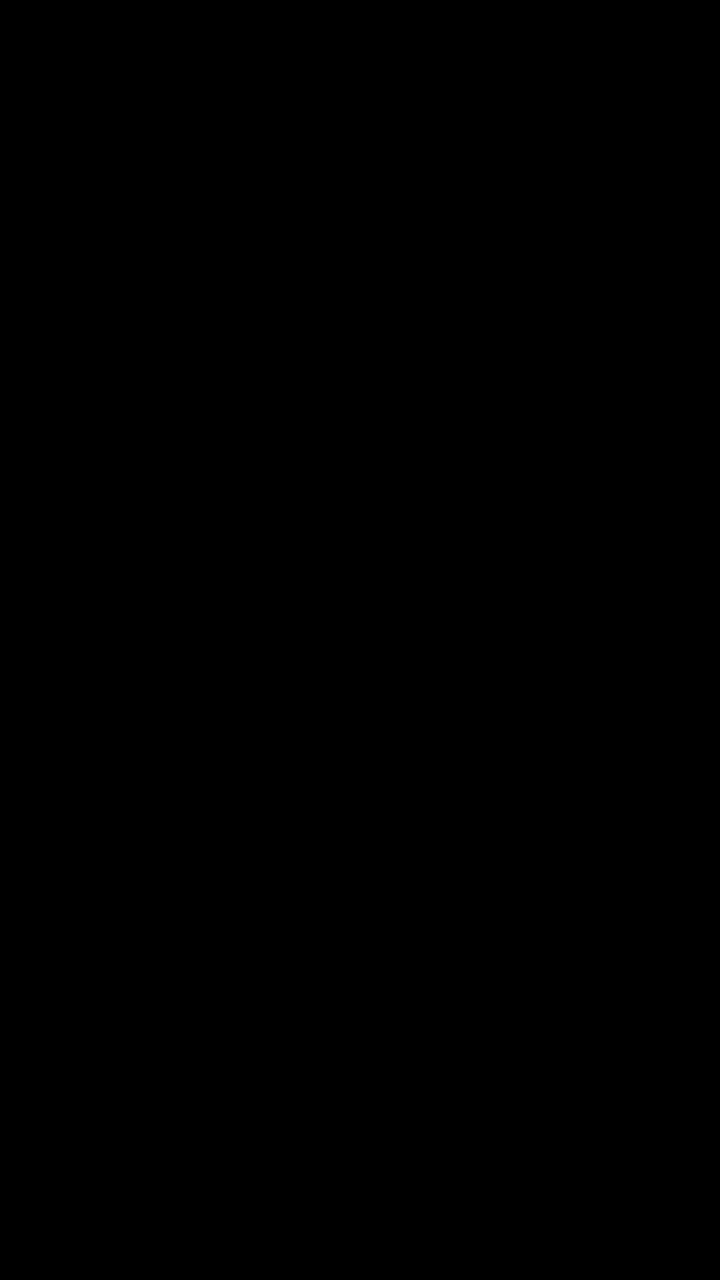}
		\captionsetup{justification=centering}
        \vspace{-5.5mm} \caption{\footnotesize{\\91th}}
	\end{subfigure}

	\caption{Qualitative comparison of the predicted segmentations using point prompts on the Visha dataset. The first images of every two rows represents the rgb and groud truth image of the 1st frame, while the other images shown in the even row and in the odd row are the ground truth and predicted shadow points for the 11th, 21th, 31th, 41th, 51th, 61th, 71th, 81th and 91th, respectively. Best viewed on screen.}
	\label{fig:fig_visha_point}
	
\end{figure*}

\begin{figure*}[!ht]
	\centering
	\vspace*{1.3mm}
	\begin{subfigure}{0.15\textwidth}
		\includegraphics[width=\textwidth]{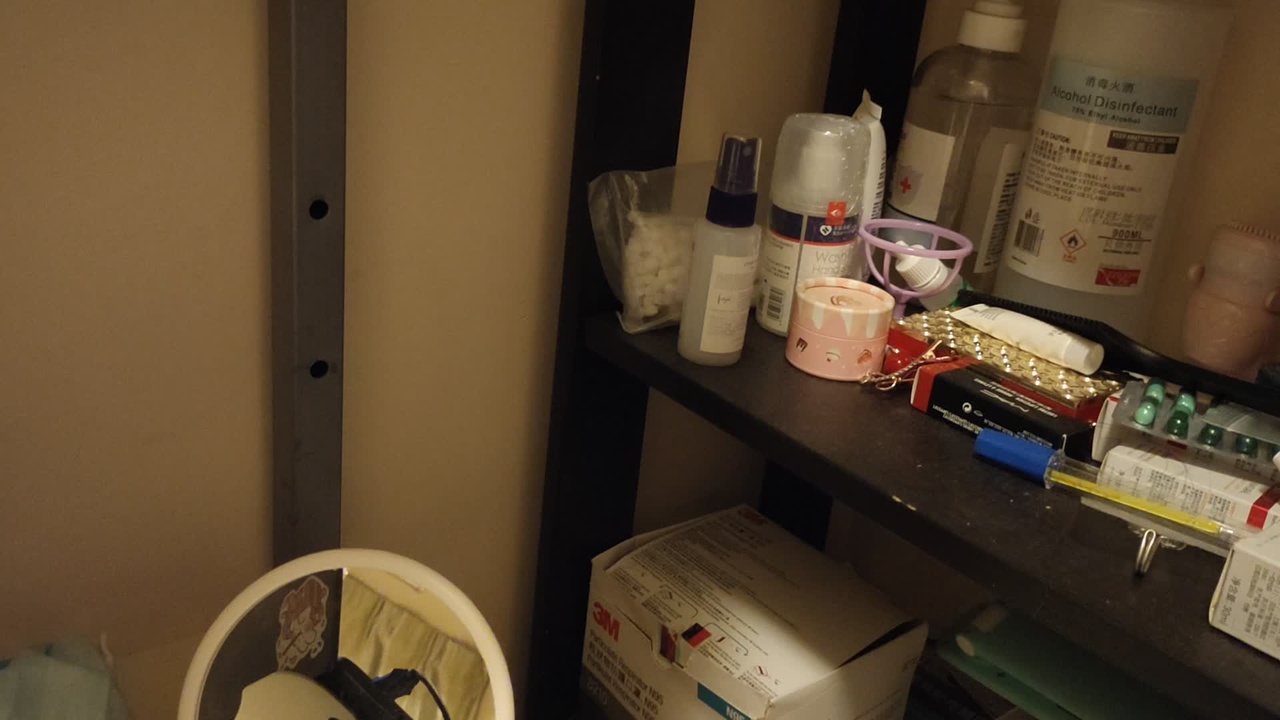}
	\end{subfigure}
	\begin{subfigure}{0.15\textwidth}
		\includegraphics[width=\textwidth]{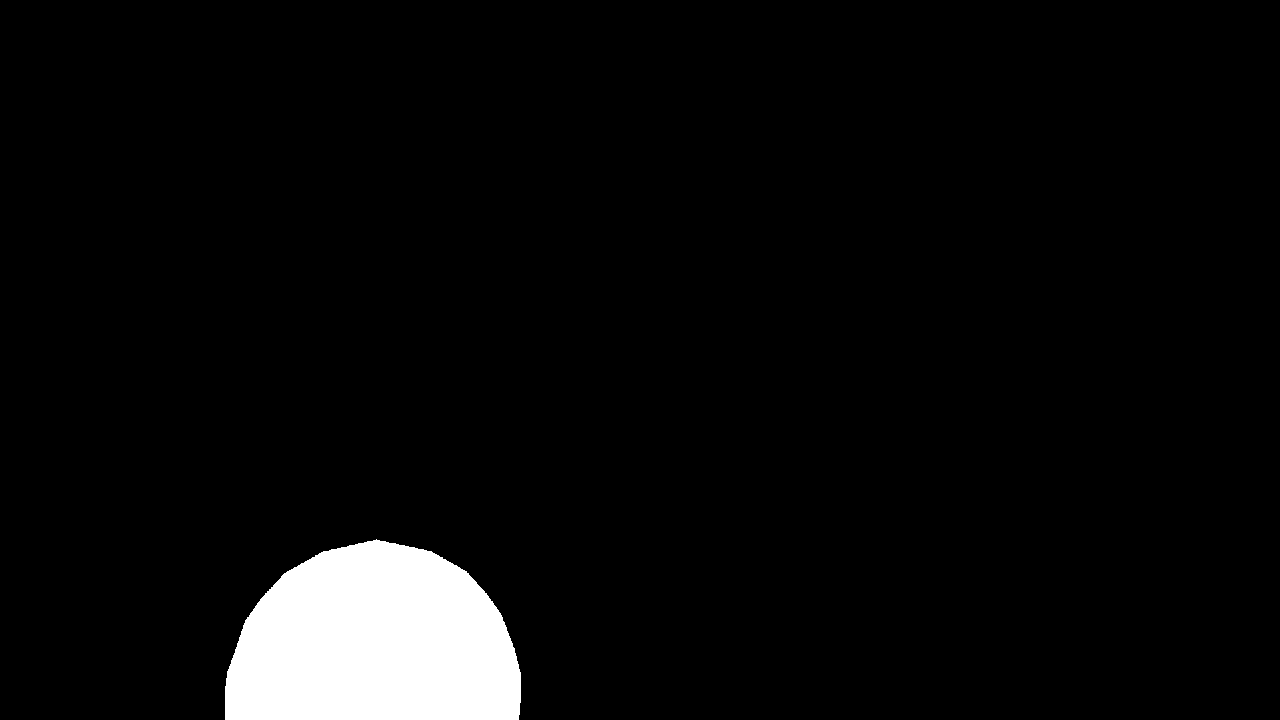}
	\end{subfigure}
	\begin{subfigure}{0.15\textwidth}
		\includegraphics[width=\textwidth]{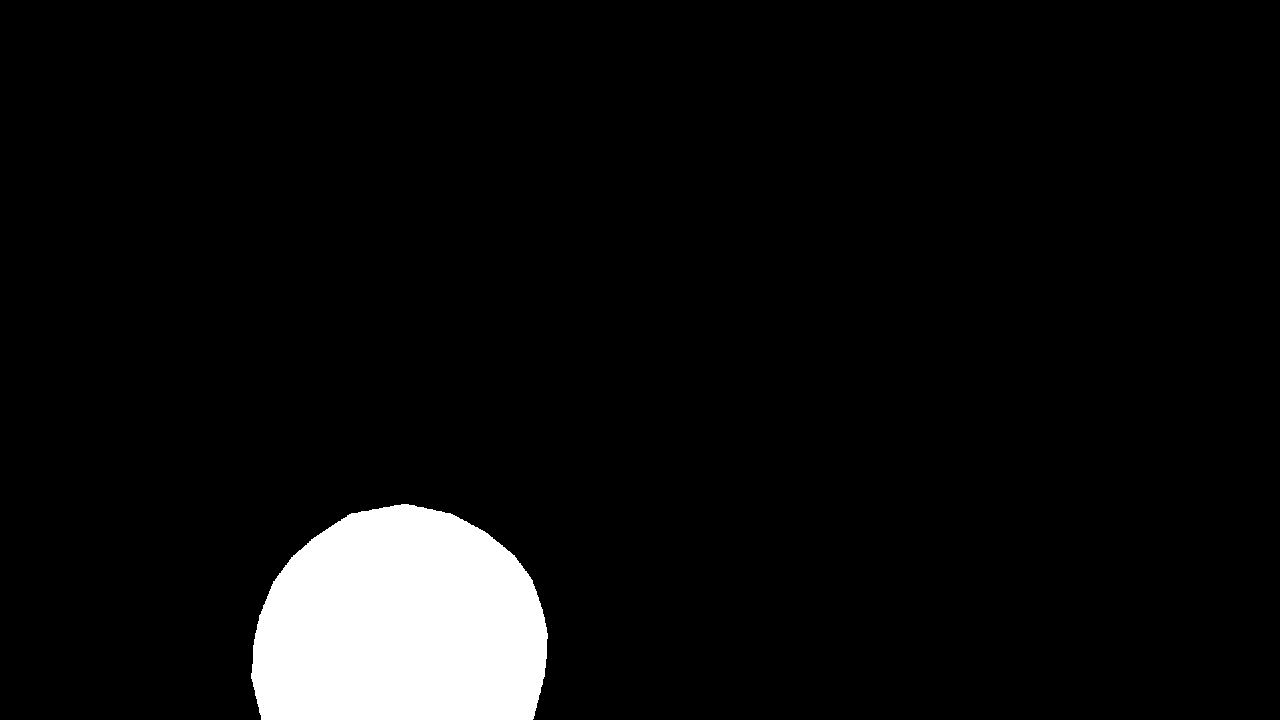}
	\end{subfigure}
	\begin{subfigure}{0.15\textwidth}
		\includegraphics[width=\textwidth]{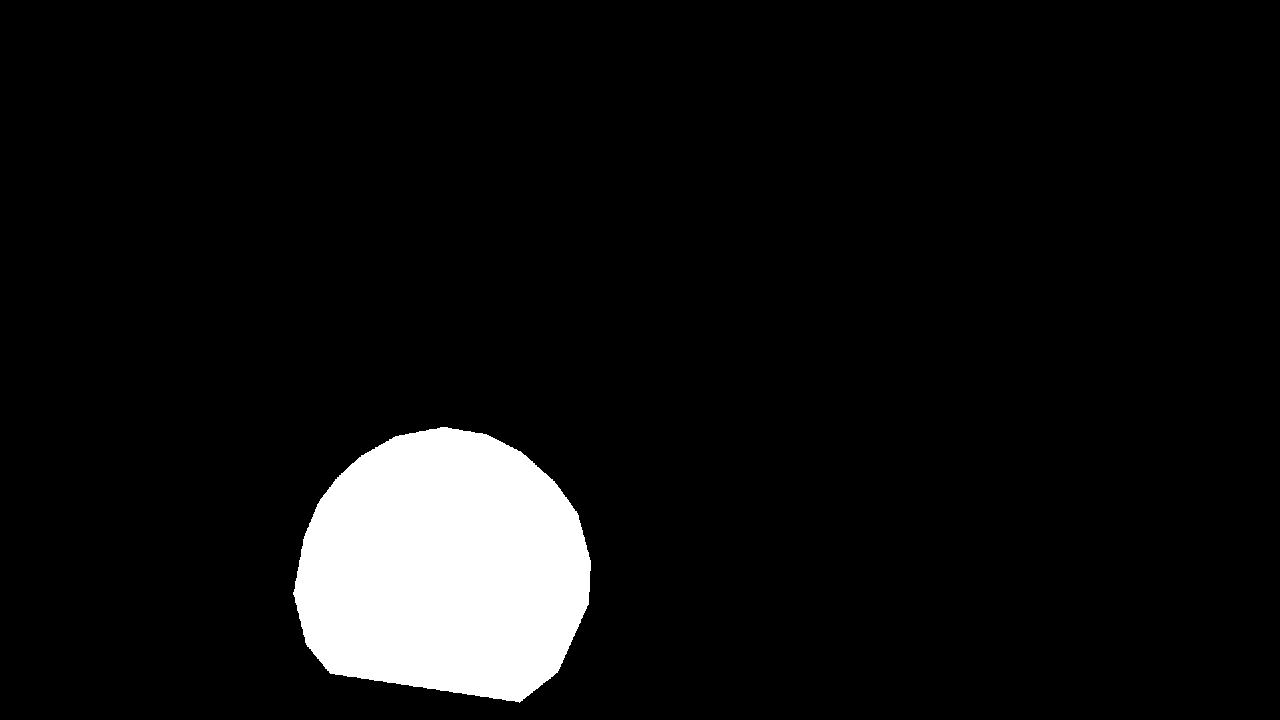}
	\end{subfigure}
	\begin{subfigure}{0.15\textwidth}
		\includegraphics[width=\textwidth]{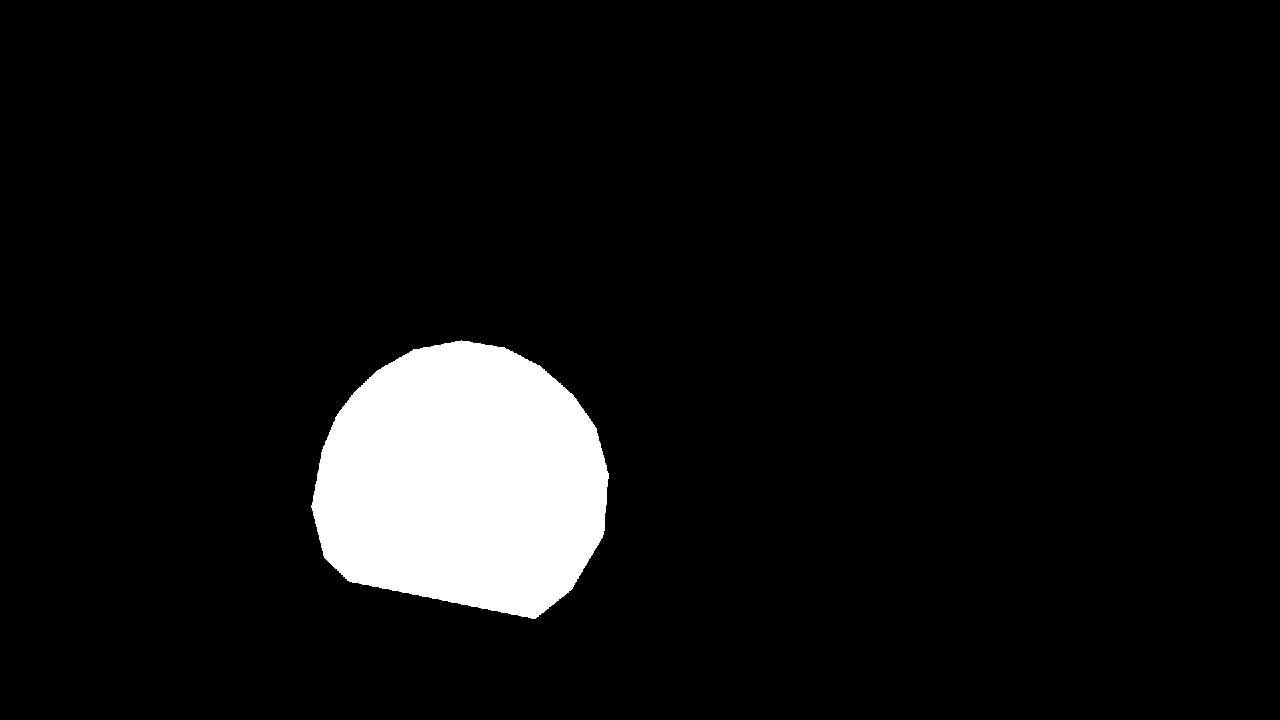}
	\end{subfigure}
	\begin{subfigure}{0.15\textwidth}
		\includegraphics[width=\textwidth]{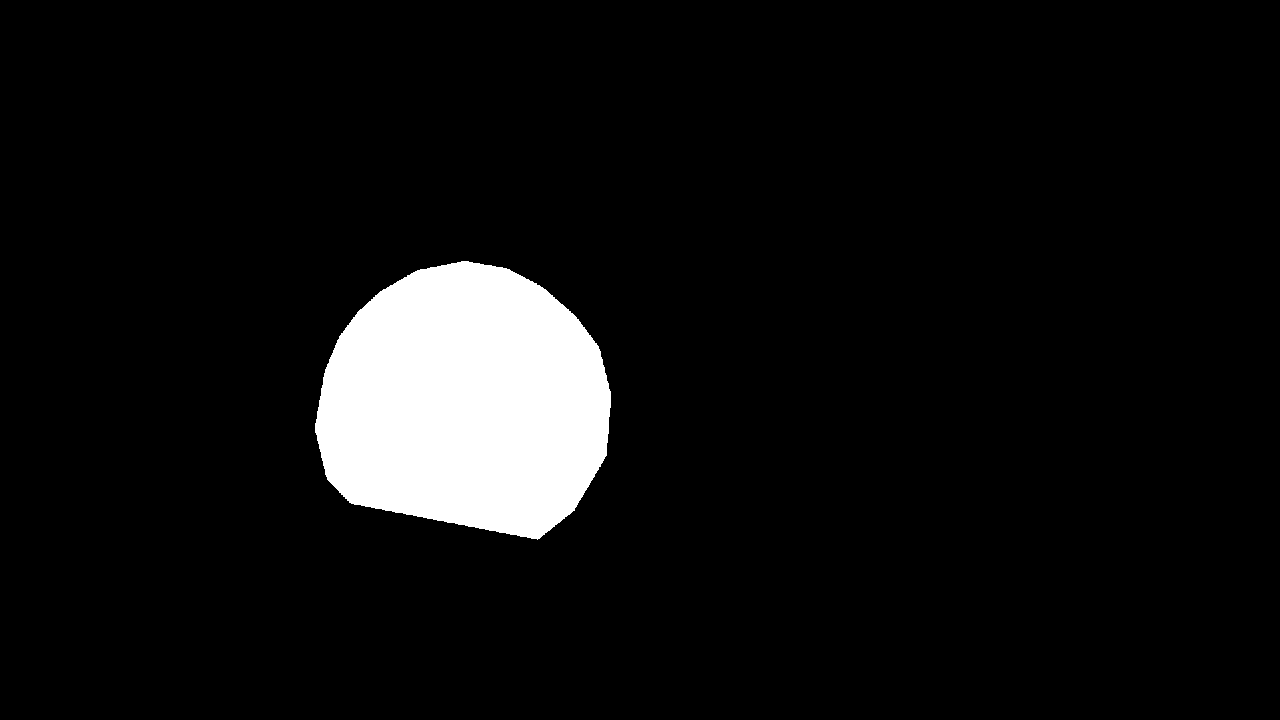}
	\end{subfigure}

	\vspace*{1.3mm}
	\begin{subfigure}{0.15\textwidth}
		\includegraphics[width=\textwidth]{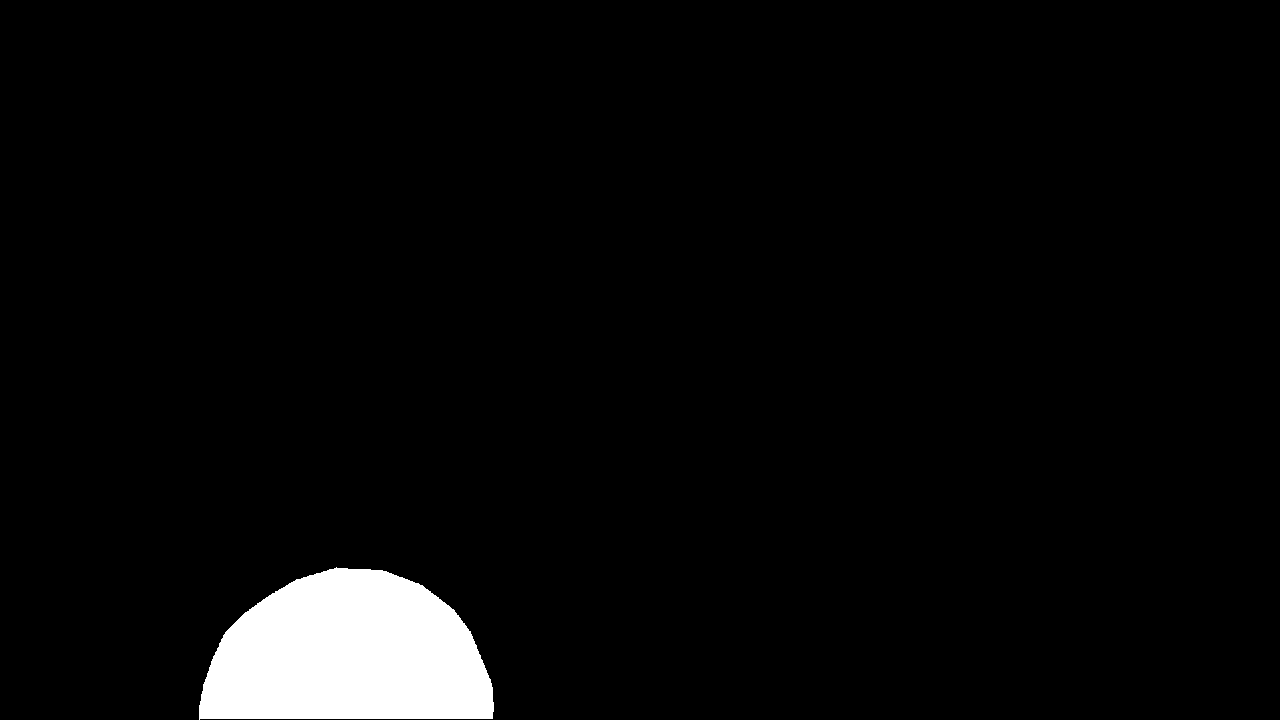}
	\end{subfigure}
	\begin{subfigure}{0.15\textwidth}
		\includegraphics[width=\textwidth]{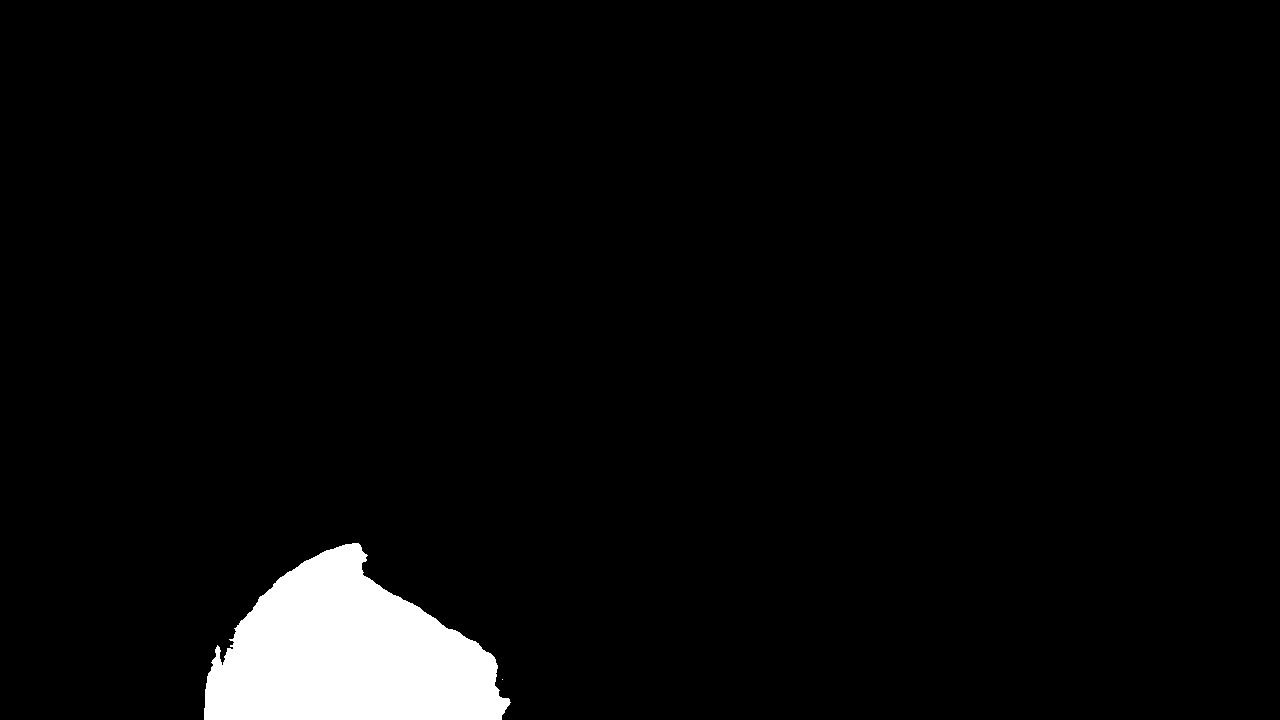}
	\end{subfigure}
	\begin{subfigure}{0.15\textwidth}
		\includegraphics[width=\textwidth]{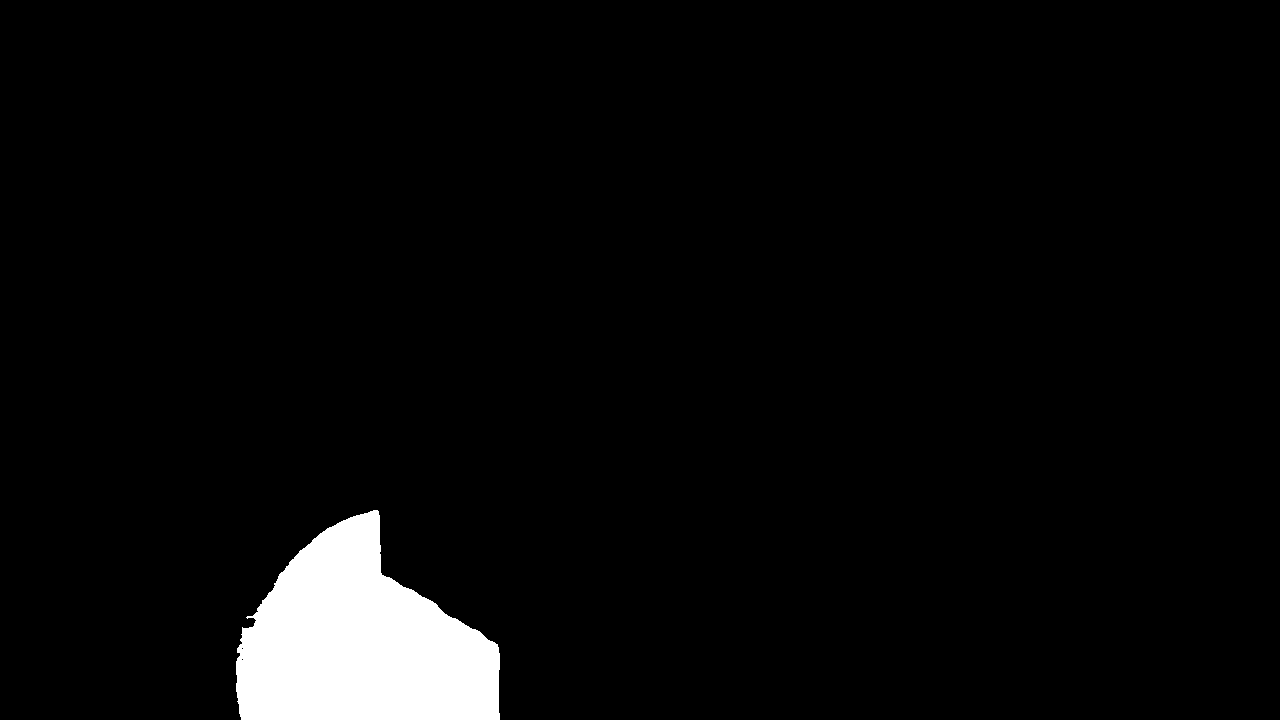}
	\end{subfigure}
	\begin{subfigure}{0.15\textwidth}
		\includegraphics[width=\textwidth]{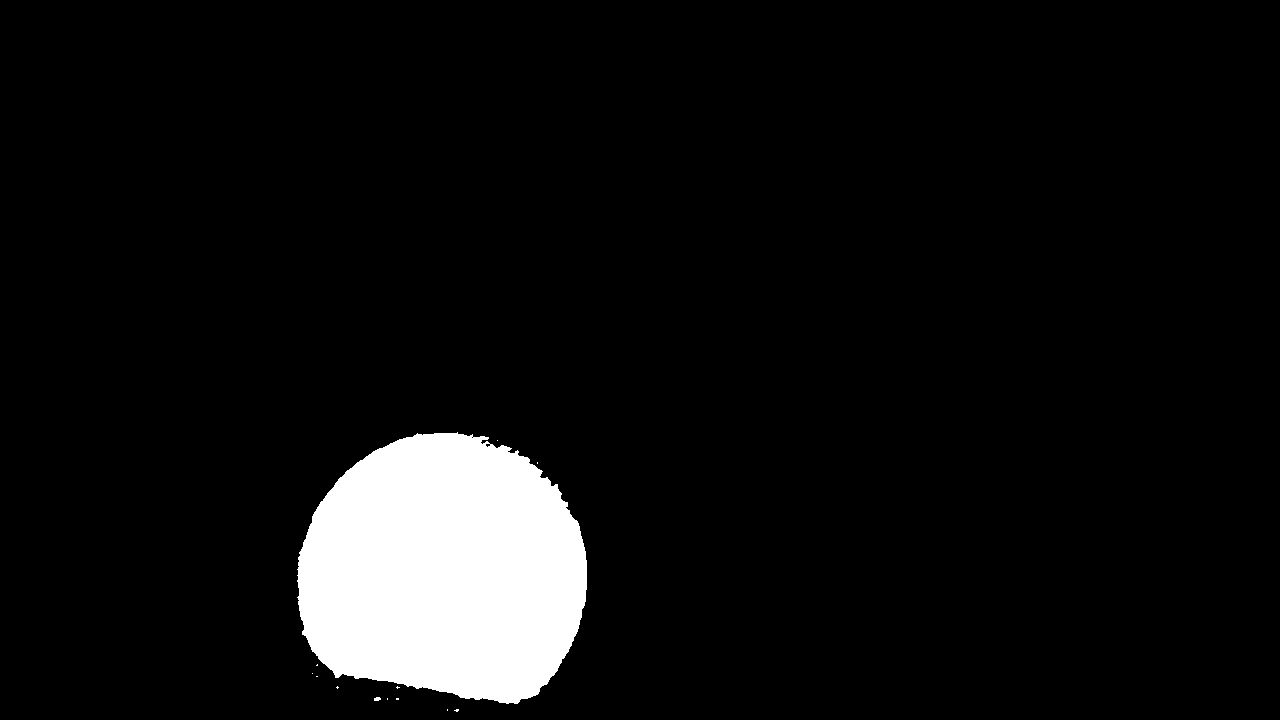}
	\end{subfigure}
	\begin{subfigure}{0.15\textwidth}
		\includegraphics[width=\textwidth]{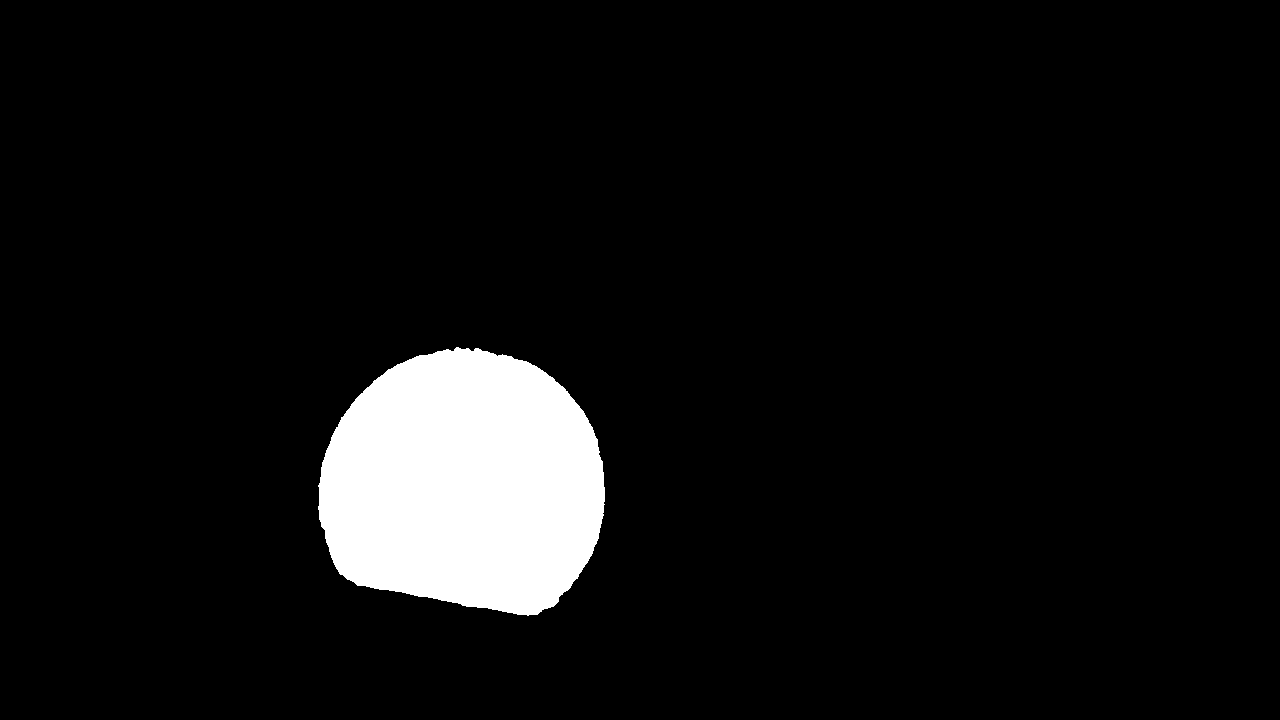}
	\end{subfigure}
	\begin{subfigure}{0.15\textwidth}
		\includegraphics[width=\textwidth]{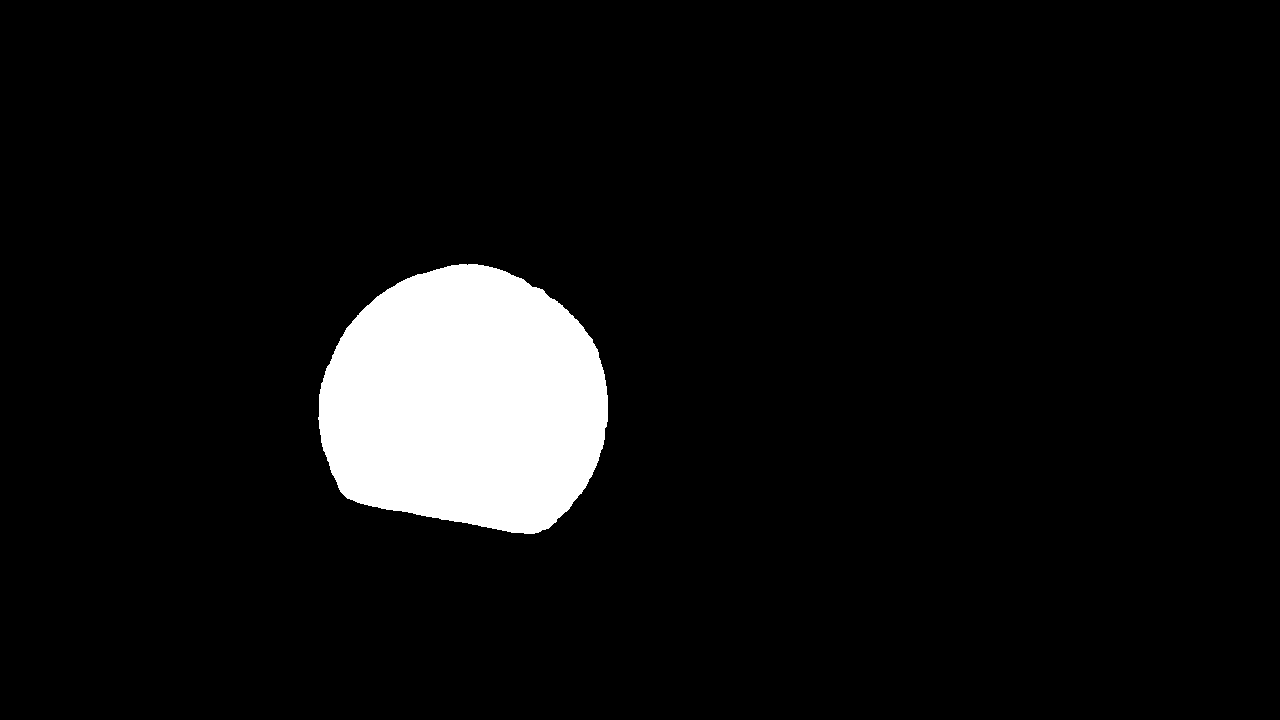}
	\end{subfigure}

	\vspace*{1.3mm}
	\begin{subfigure}{0.15\textwidth}
		\includegraphics[width=\textwidth]{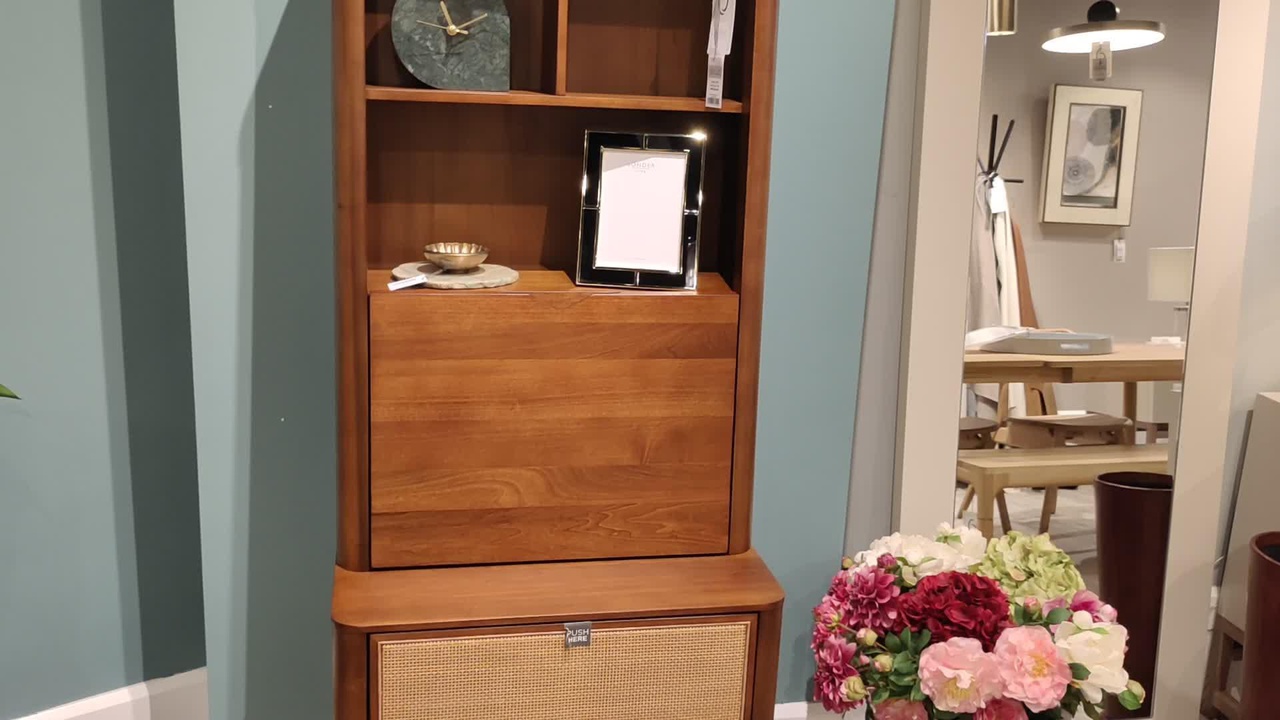}
	\end{subfigure}
	\begin{subfigure}{0.15\textwidth}
		\includegraphics[width=\textwidth]{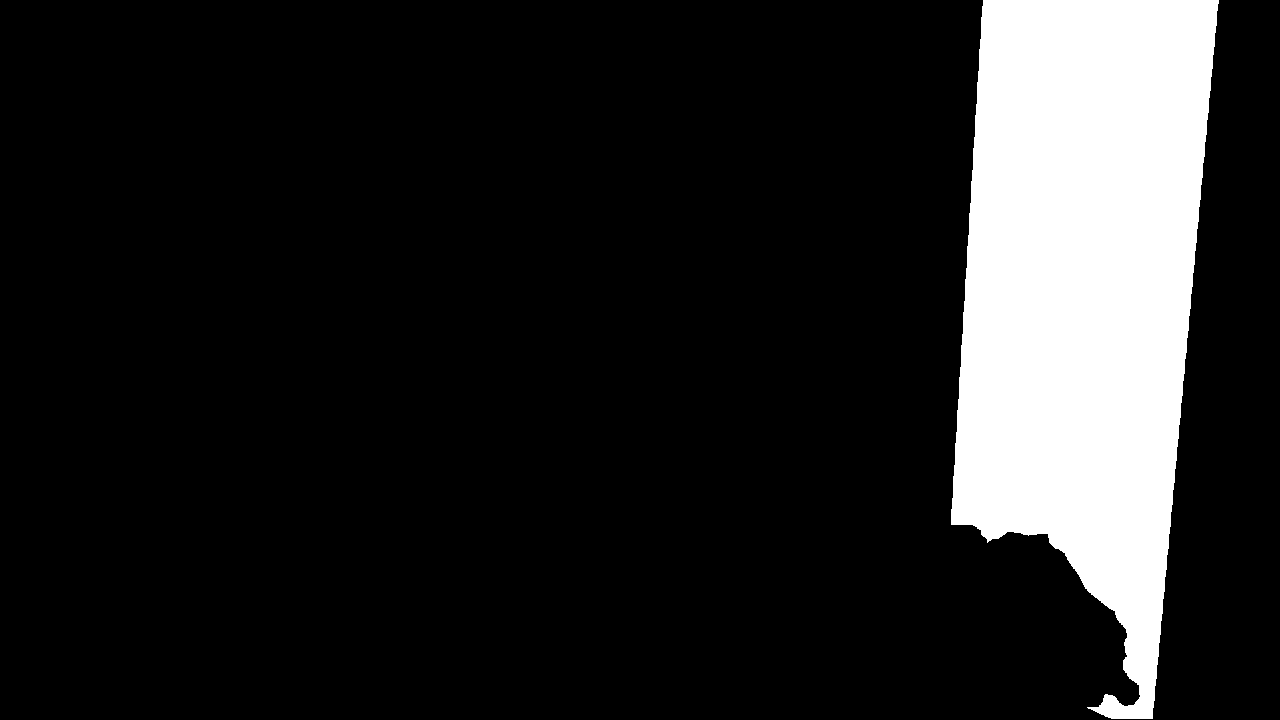}
	\end{subfigure}
	\begin{subfigure}{0.15\textwidth}
		\includegraphics[width=\textwidth]{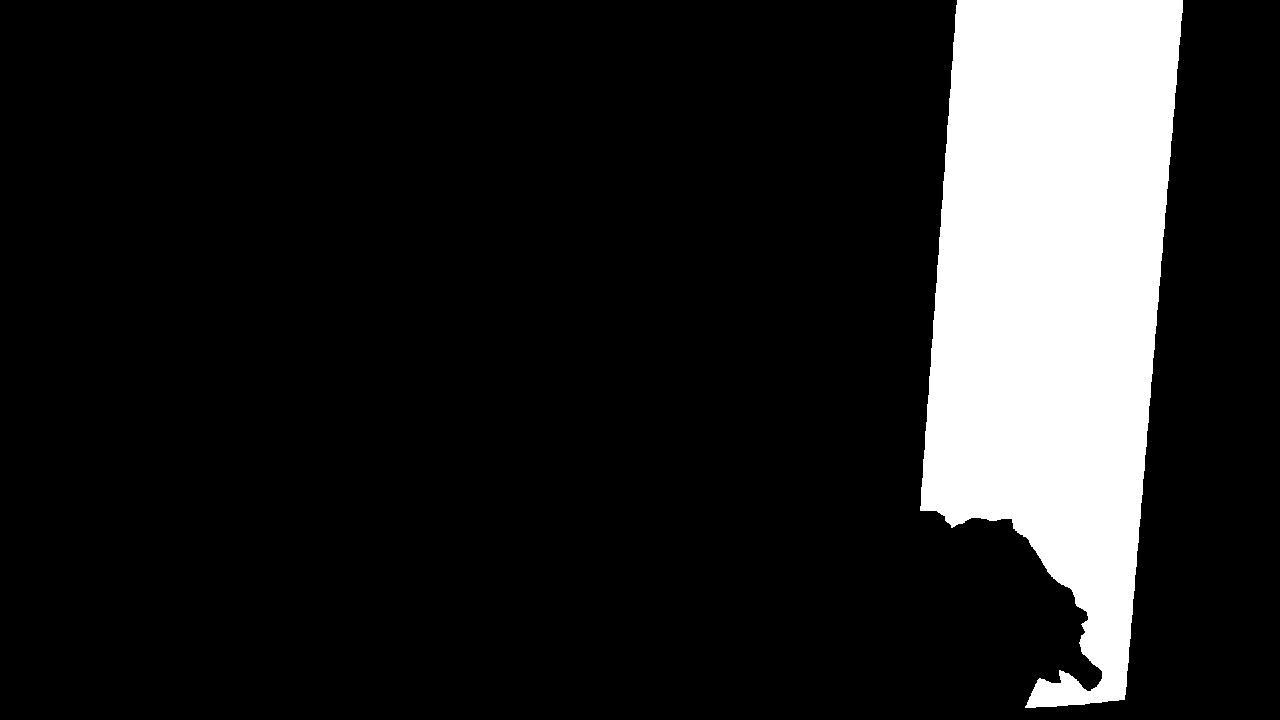}
	\end{subfigure}
	\begin{subfigure}{0.15\textwidth}
		\includegraphics[width=\textwidth]{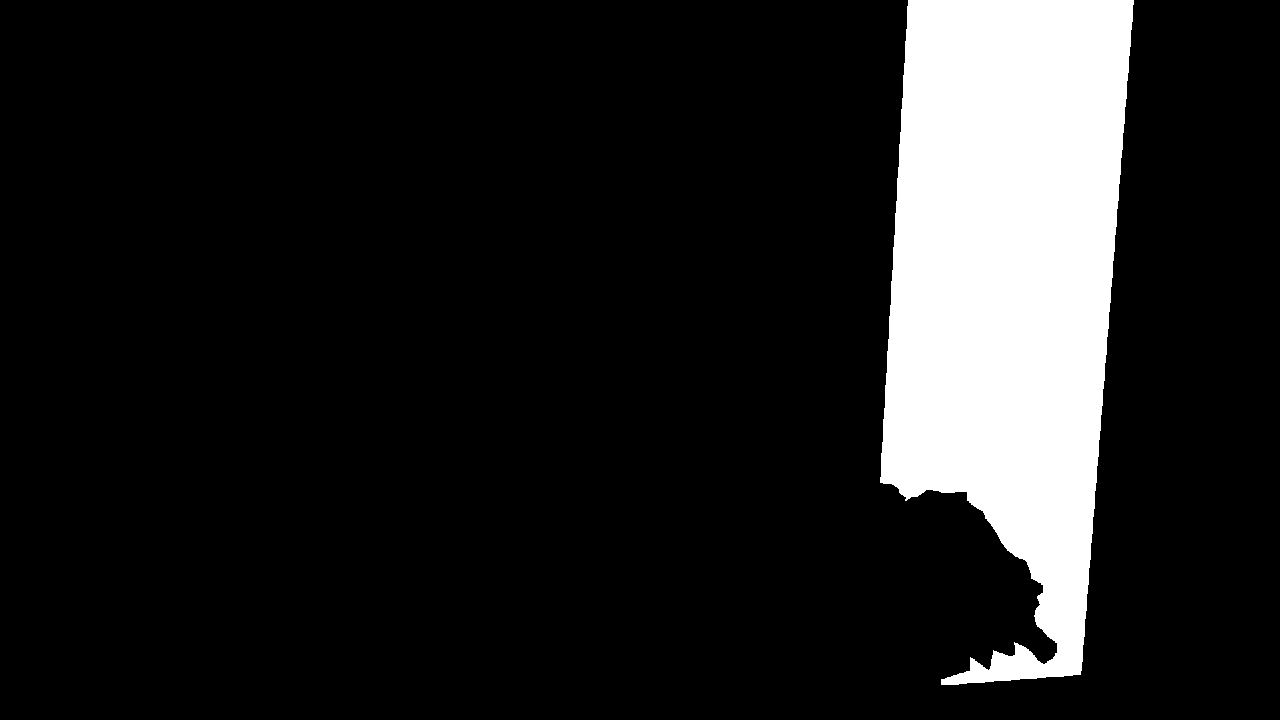}
	\end{subfigure}
	\begin{subfigure}{0.15\textwidth}
		\includegraphics[width=\textwidth]{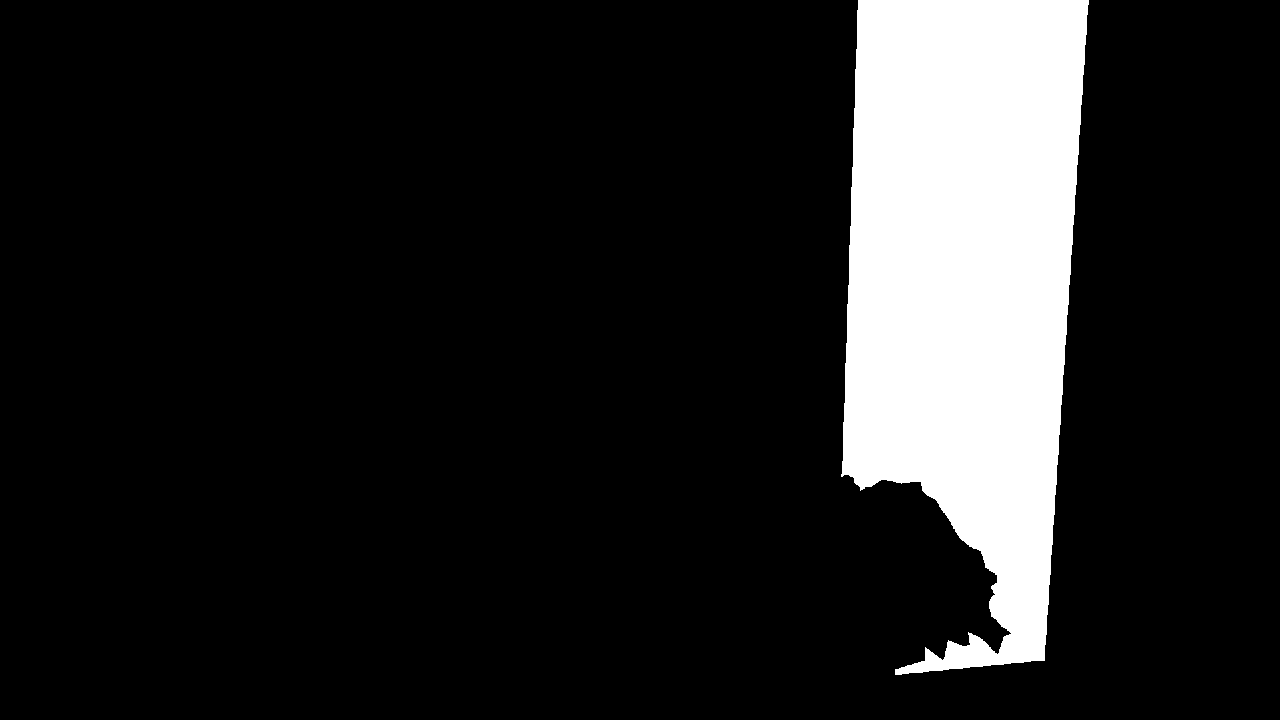}
	\end{subfigure}
	\begin{subfigure}{0.15\textwidth}
		\includegraphics[width=\textwidth]{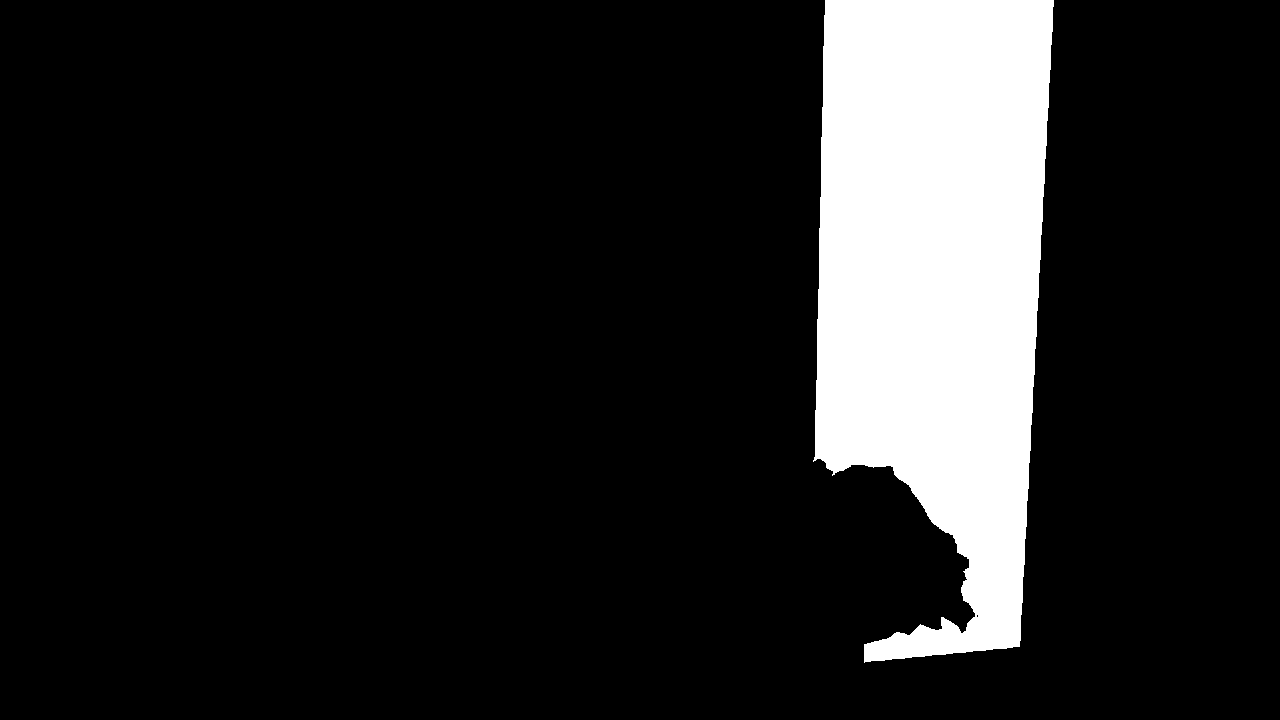}
	\end{subfigure}
	
	\vspace*{1.3mm}
	\begin{subfigure}{0.15\textwidth}
		\includegraphics[width=\textwidth]{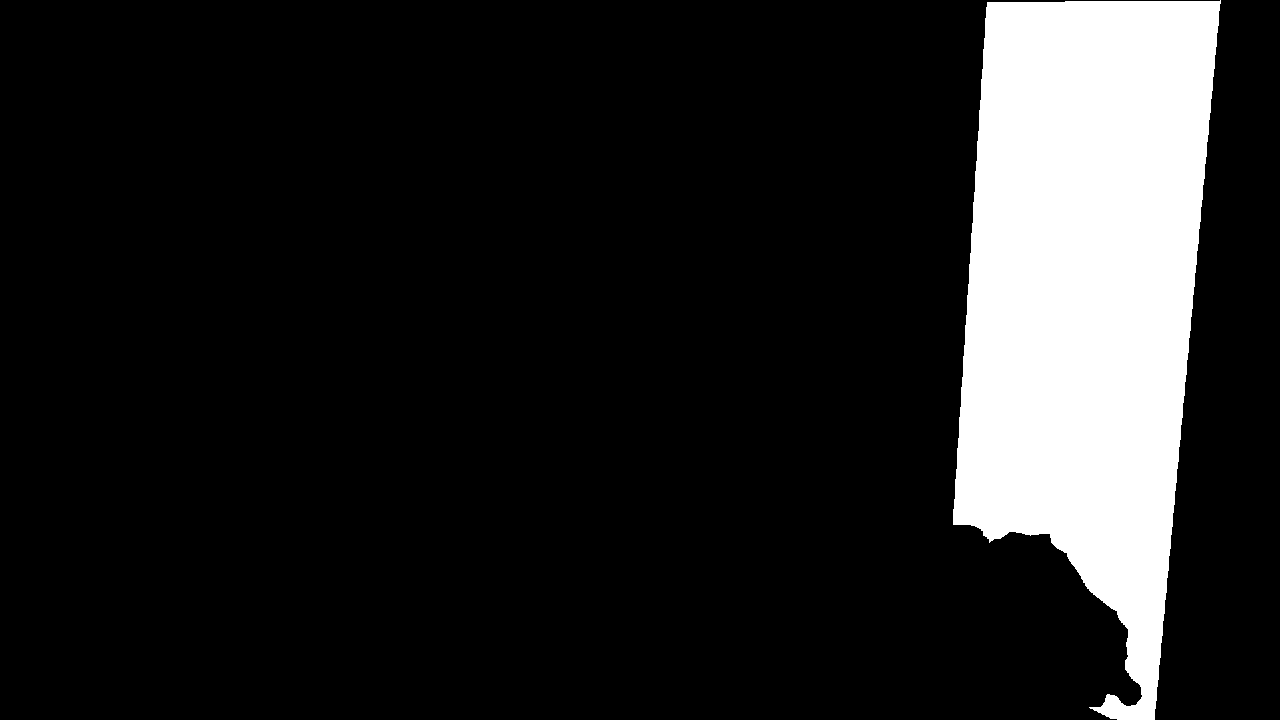}
	\end{subfigure}
	\begin{subfigure}{0.15\textwidth}
		\includegraphics[width=\textwidth]{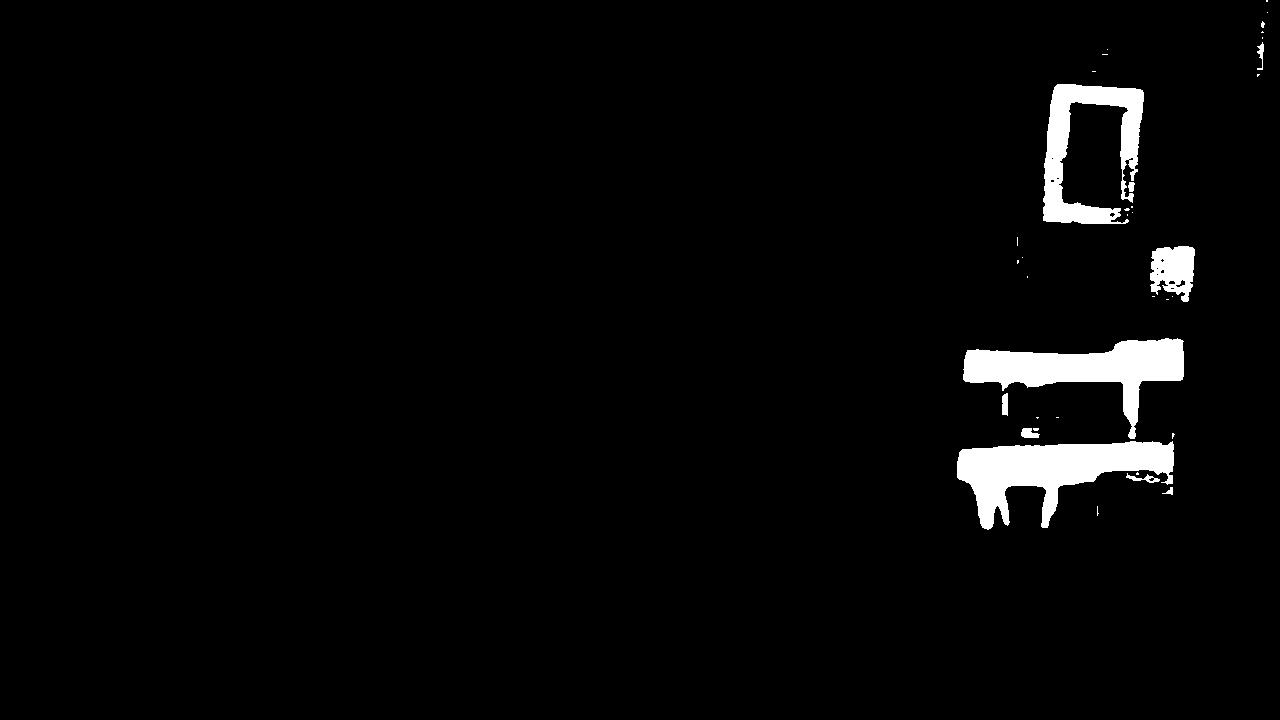}
	\end{subfigure}
	\begin{subfigure}{0.15\textwidth}
		\includegraphics[width=\textwidth]{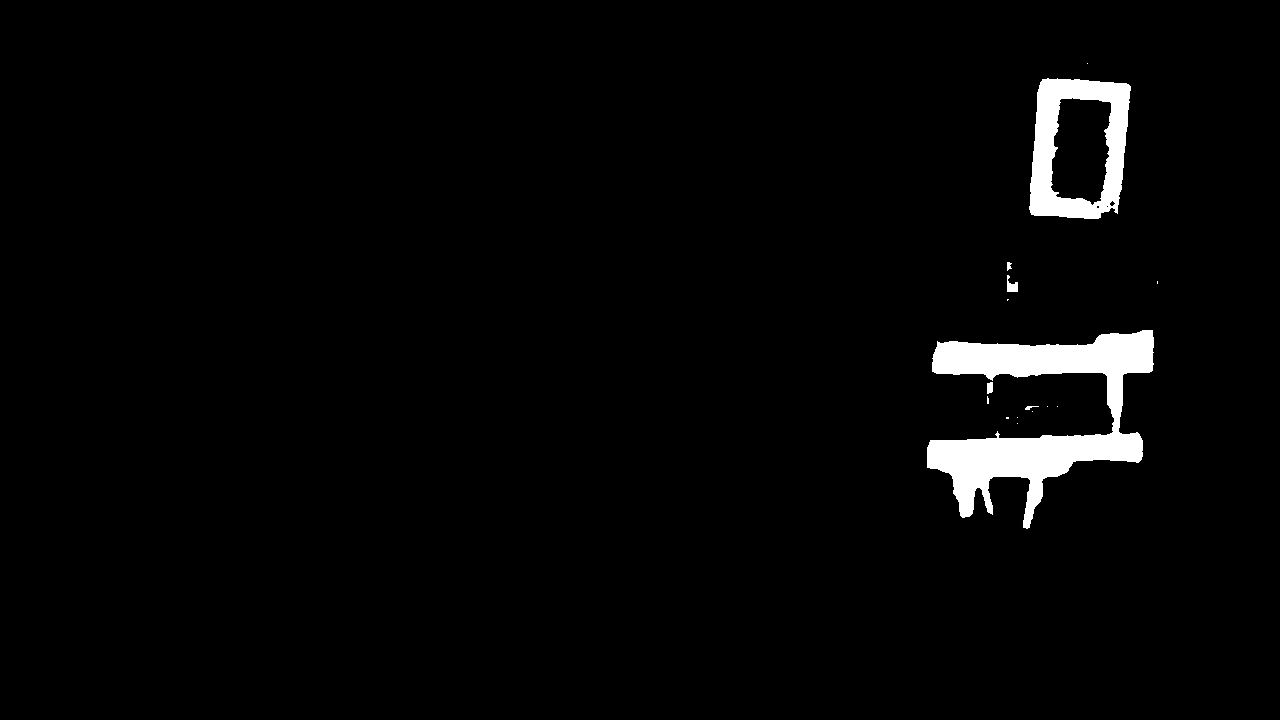}
	\end{subfigure}
	\begin{subfigure}{0.15\textwidth}
		\includegraphics[width=\textwidth]{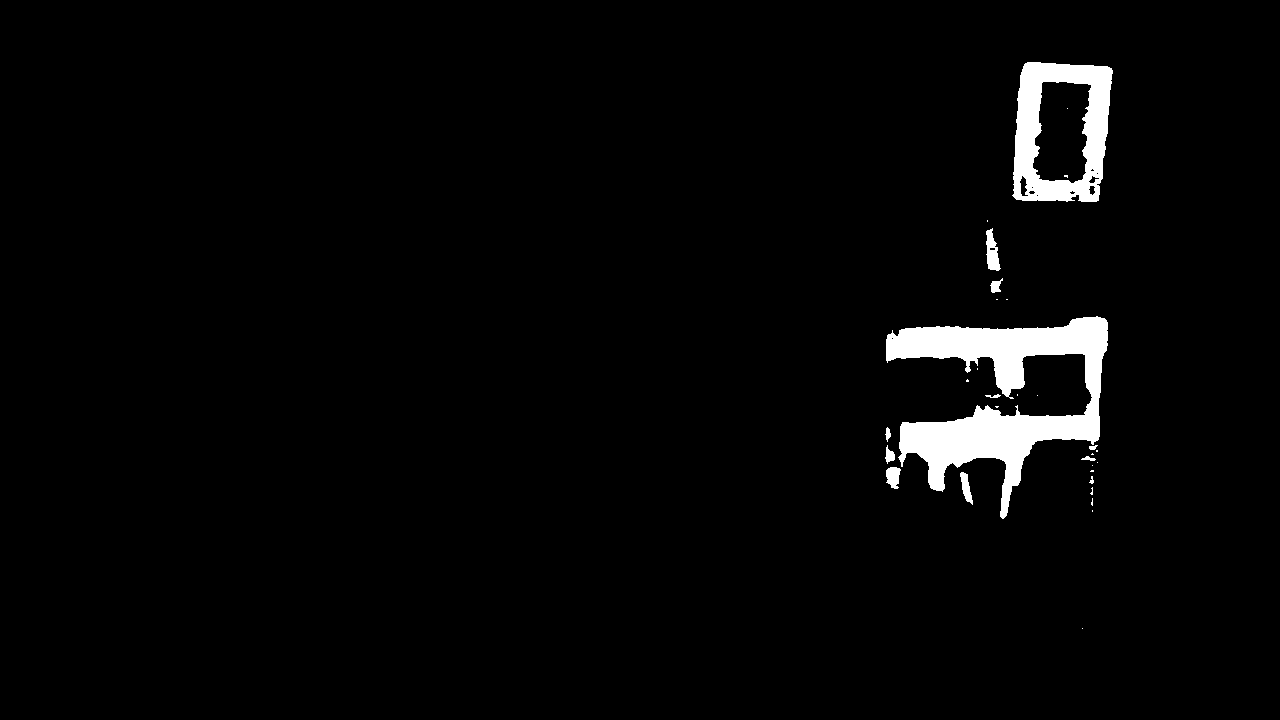}
	\end{subfigure}
	\begin{subfigure}{0.15\textwidth}
		\includegraphics[width=\textwidth]{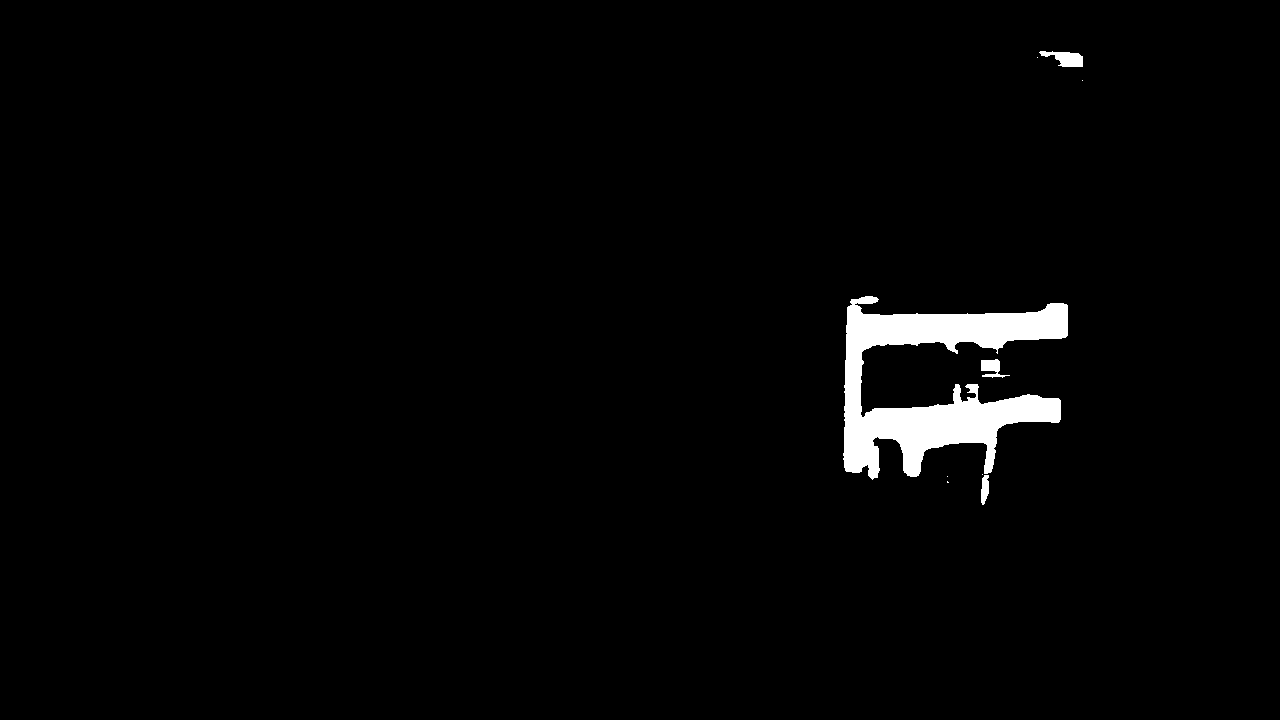}
	\end{subfigure}
	\begin{subfigure}{0.15\textwidth}
		\includegraphics[width=\textwidth]{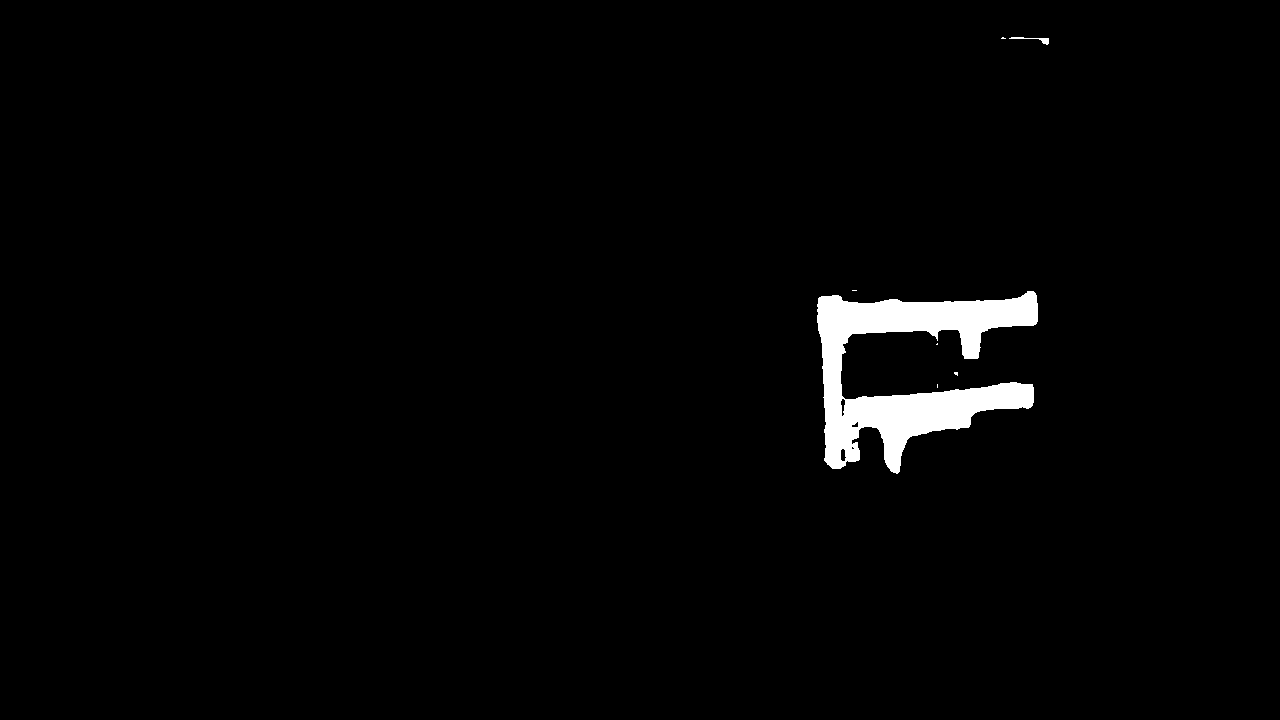}
	\end{subfigure}

	\vspace*{1.3mm}
	\begin{subfigure}{0.15\textwidth}
		\includegraphics[width=\textwidth]{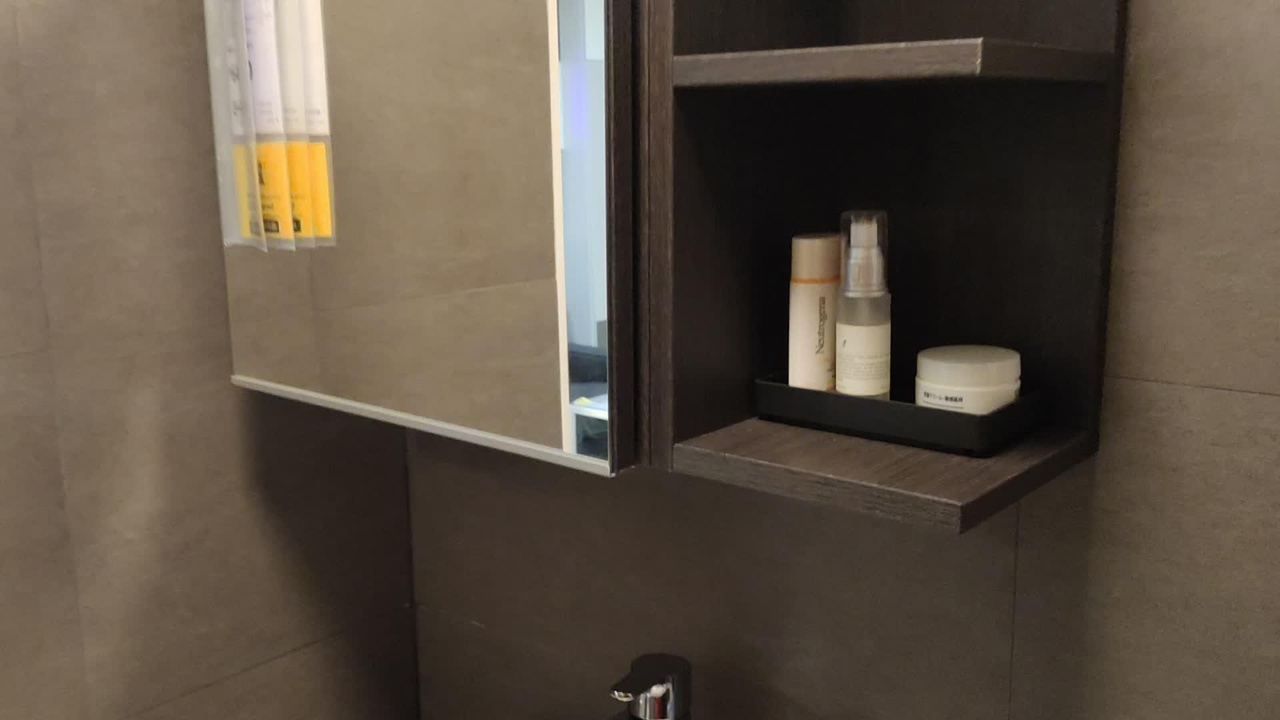}
	\end{subfigure}
	\begin{subfigure}{0.15\textwidth}
		\includegraphics[width=\textwidth]{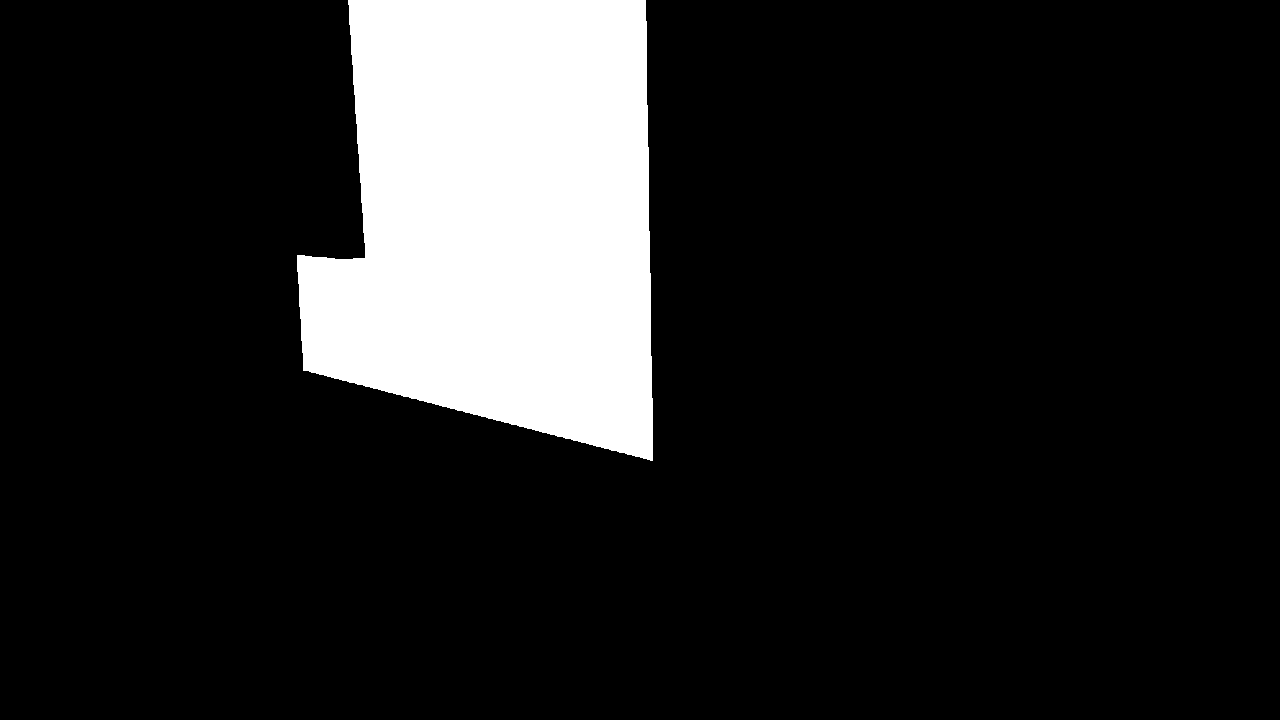}
	\end{subfigure}
	\begin{subfigure}{0.15\textwidth}
		\includegraphics[width=\textwidth]{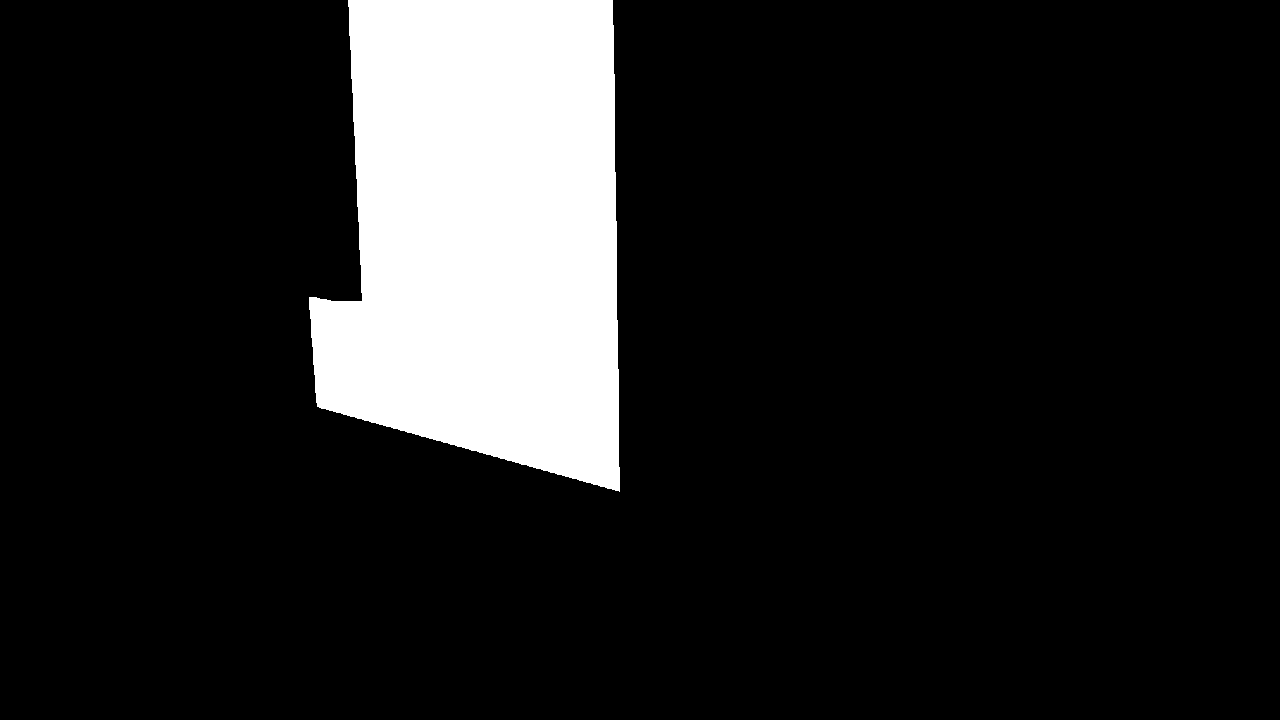}
	\end{subfigure}
	\begin{subfigure}{0.15\textwidth}
		\includegraphics[width=\textwidth]{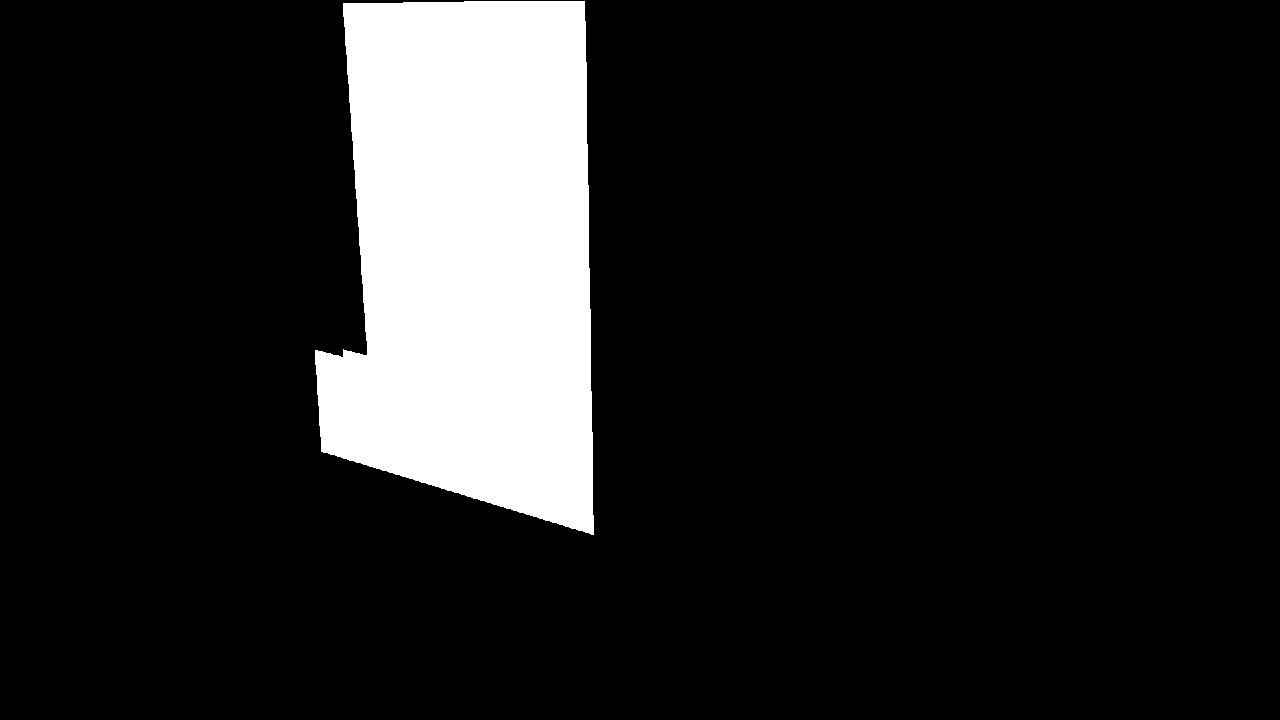}
	\end{subfigure}
	\begin{subfigure}{0.15\textwidth}
		\includegraphics[width=\textwidth]{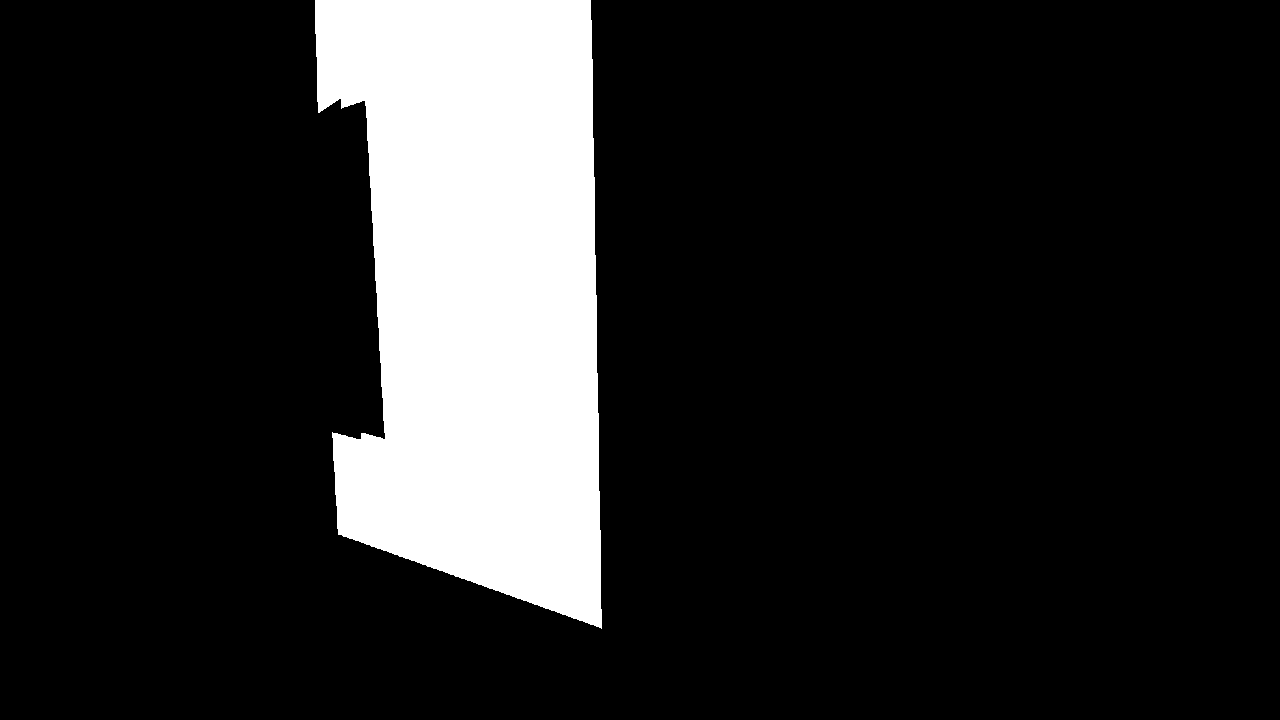}
	\end{subfigure}
	\begin{subfigure}{0.15\textwidth}
		\includegraphics[width=\textwidth]{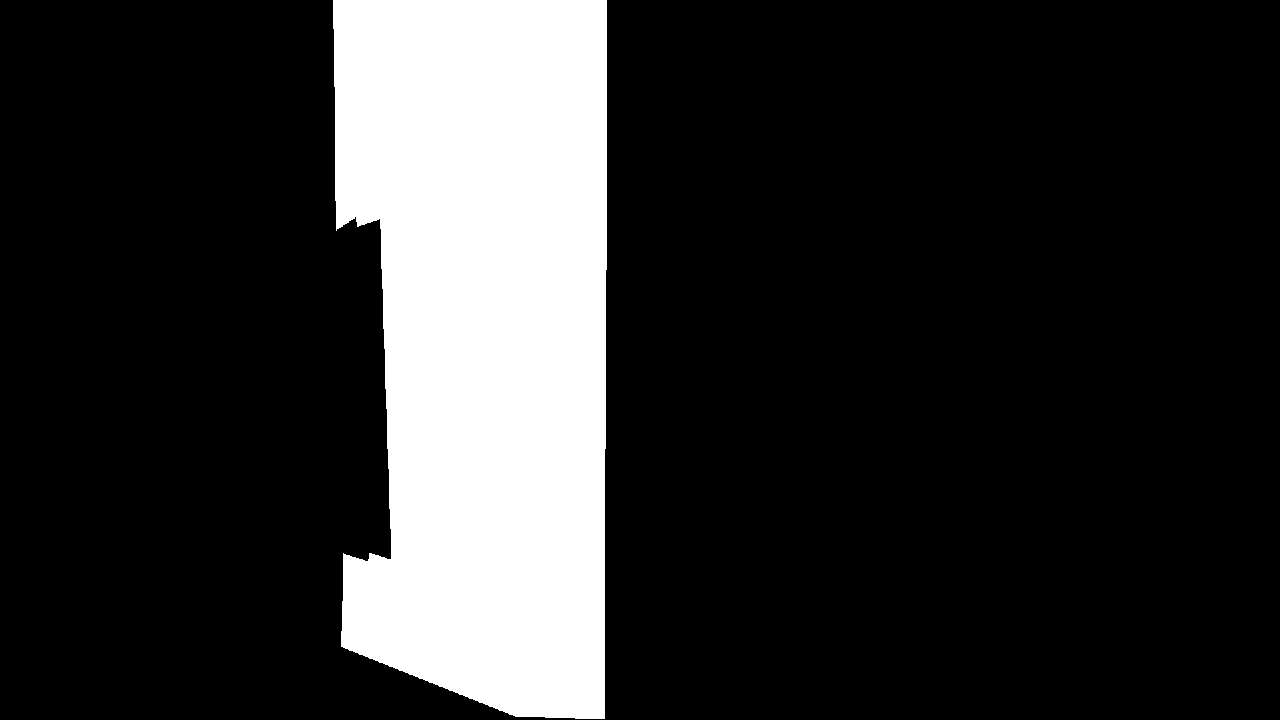}
	\end{subfigure}
	
	\vspace*{1.3mm}
	\begin{subfigure}{0.15\textwidth}
		\includegraphics[width=\textwidth]{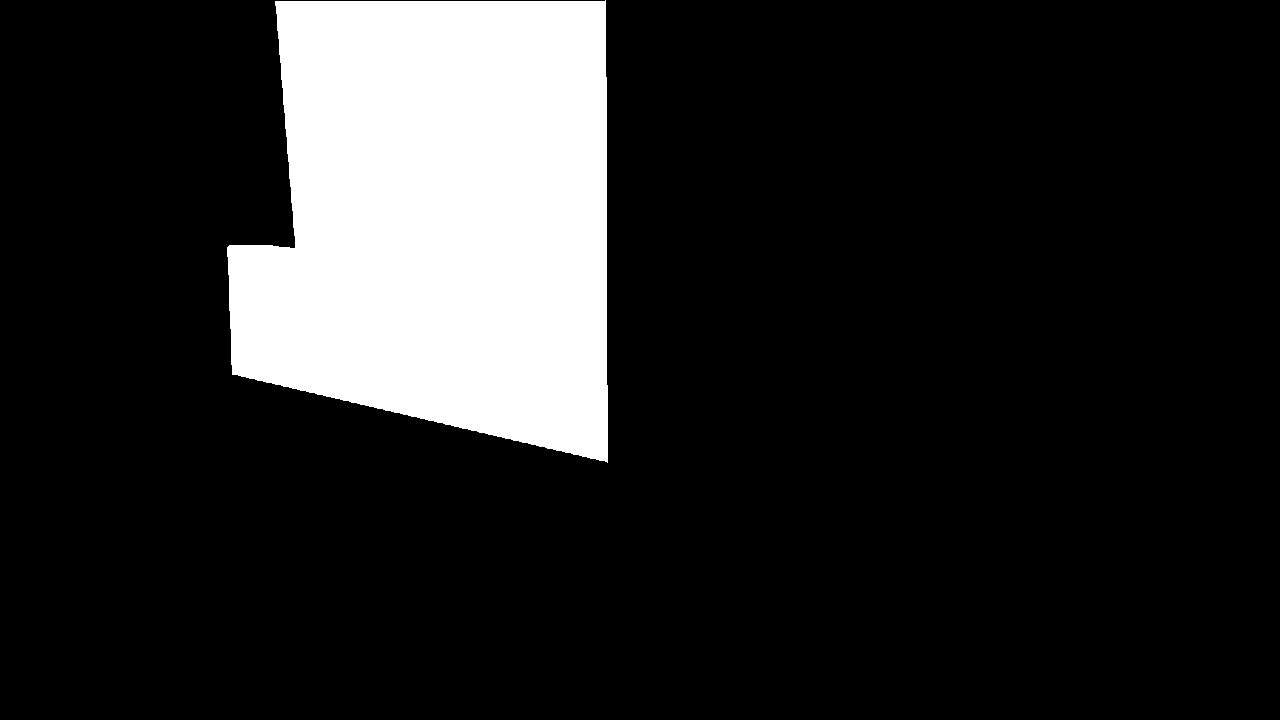}
	\end{subfigure}
	\begin{subfigure}{0.15\textwidth}
		\includegraphics[width=\textwidth]{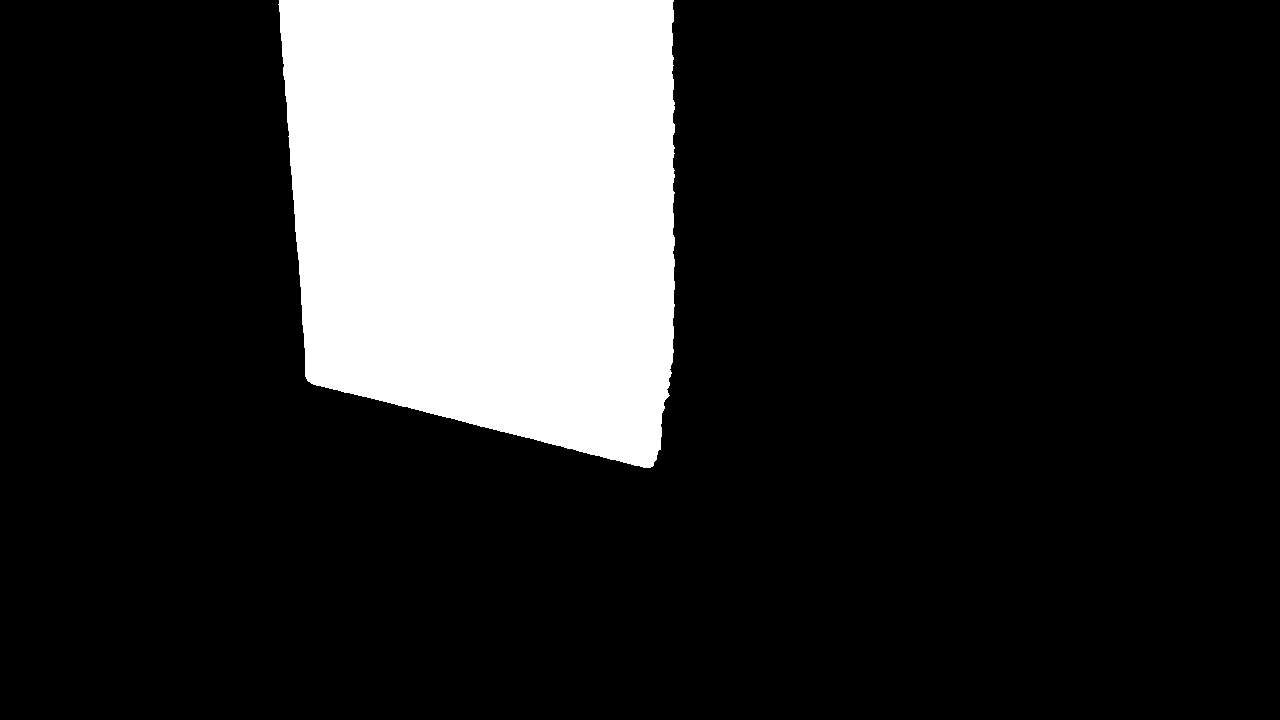}
	\end{subfigure}
	\begin{subfigure}{0.15\textwidth}
		\includegraphics[width=\textwidth]{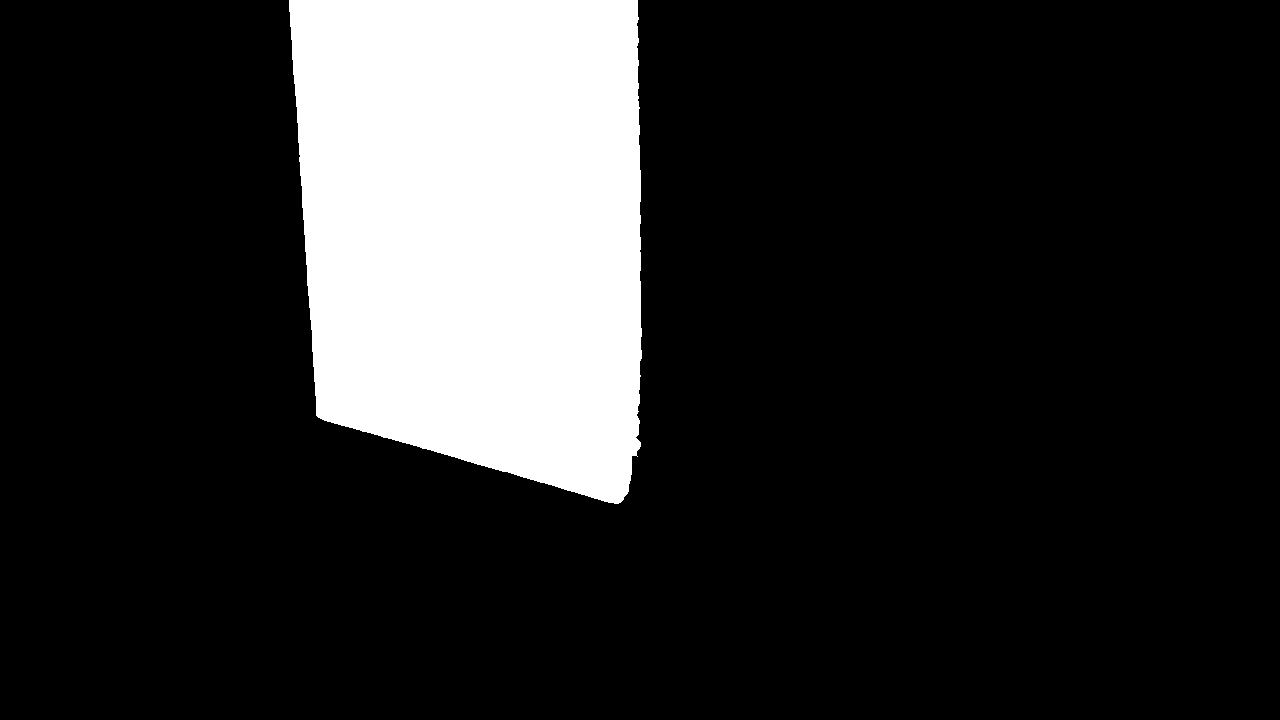}
	\end{subfigure}
	\begin{subfigure}{0.15\textwidth}
		\includegraphics[width=\textwidth]{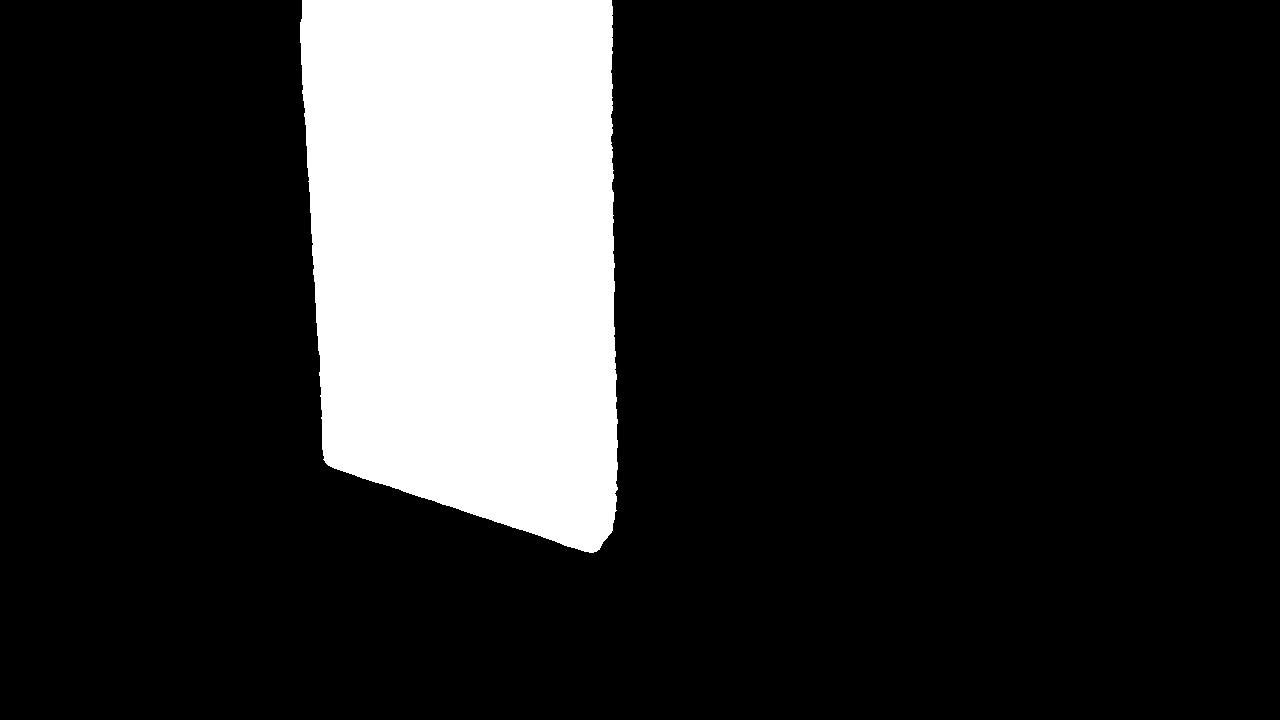}
	\end{subfigure}
	\begin{subfigure}{0.15\textwidth}
		\includegraphics[width=\textwidth]{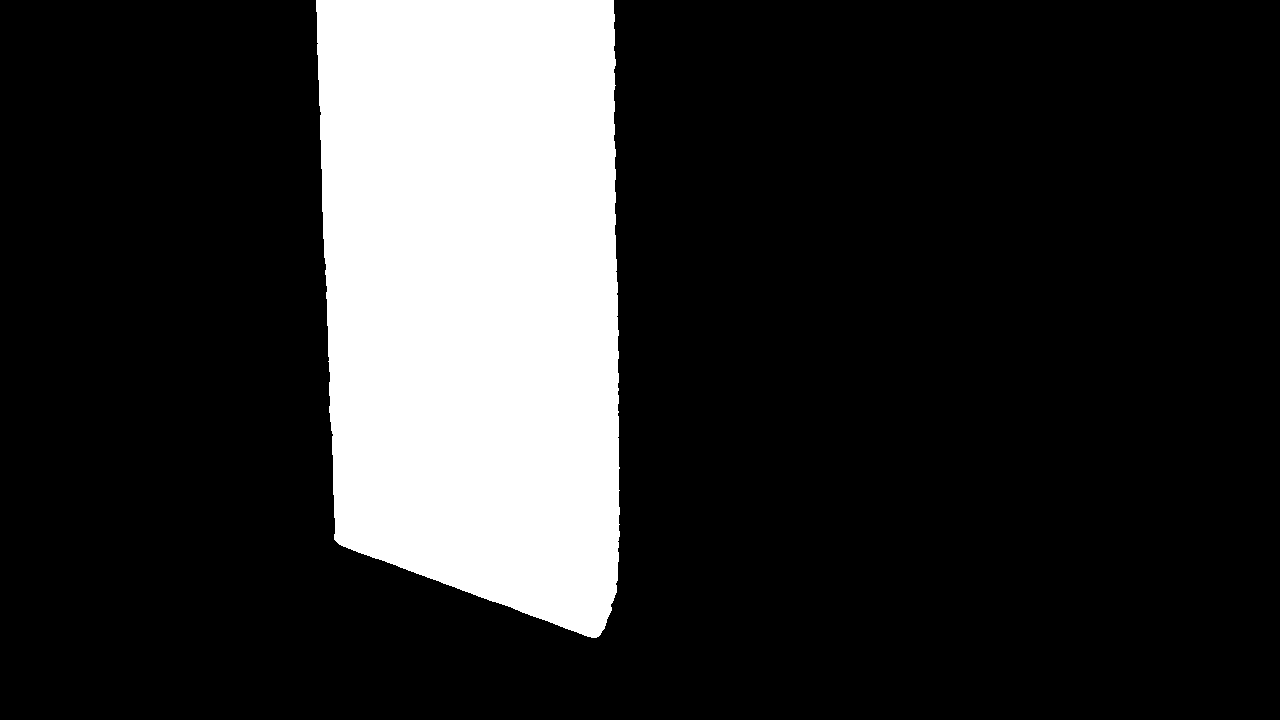}
	\end{subfigure}
	\begin{subfigure}{0.15\textwidth}
		\includegraphics[width=\textwidth]{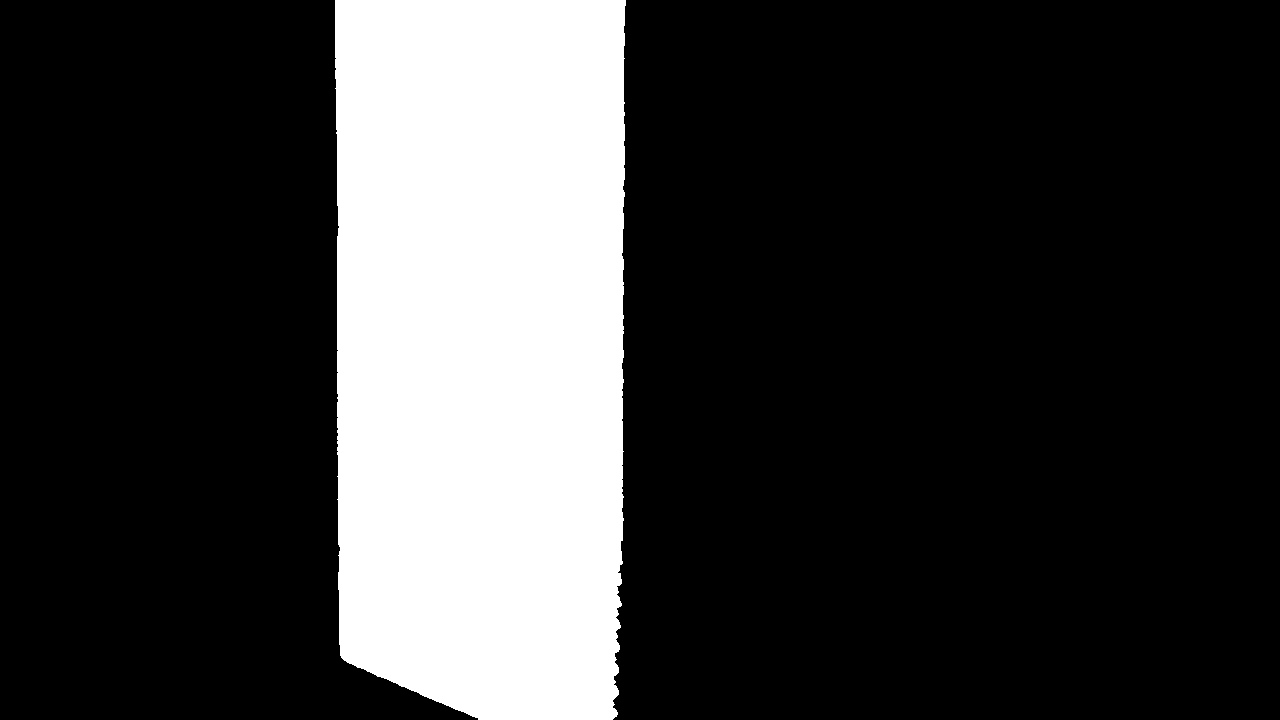}
	\end{subfigure}

	\vspace*{1.3mm}
	\begin{subfigure}{0.15\textwidth}
		\includegraphics[width=\textwidth]{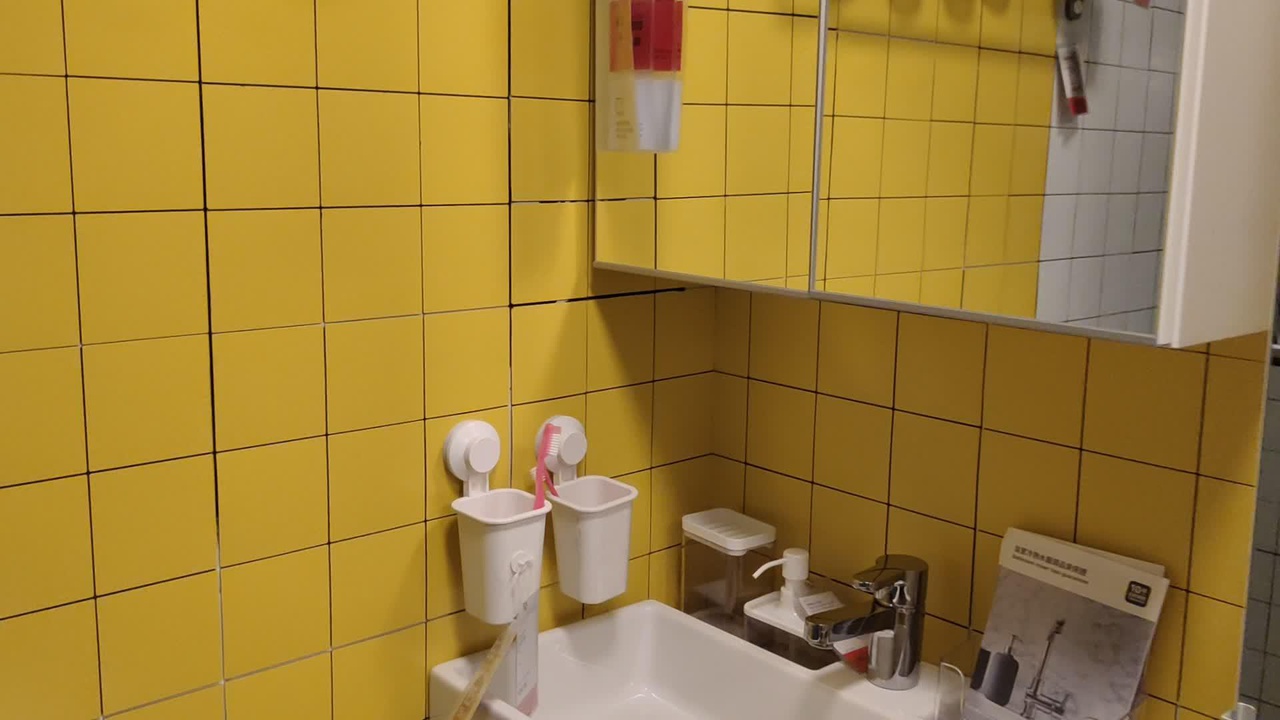}
	\end{subfigure}
	\begin{subfigure}{0.15\textwidth}
		\includegraphics[width=\textwidth]{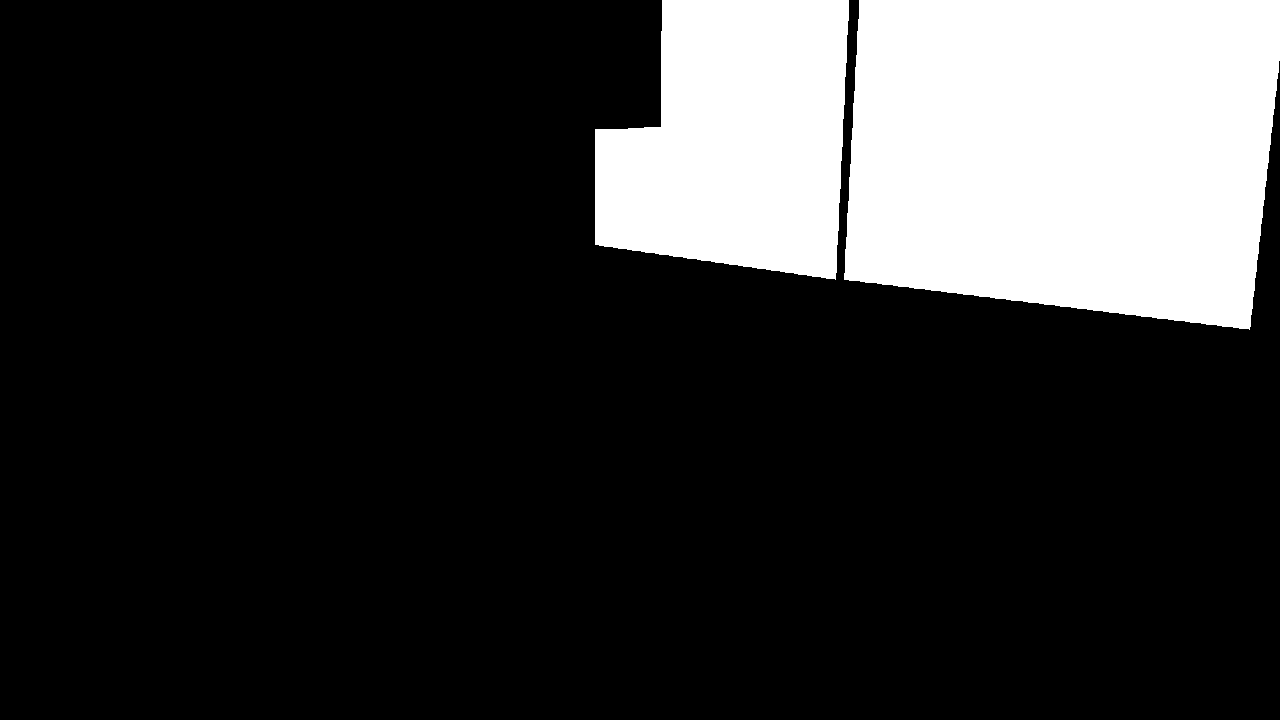}
	\end{subfigure}
	\begin{subfigure}{0.15\textwidth}
		\includegraphics[width=\textwidth]{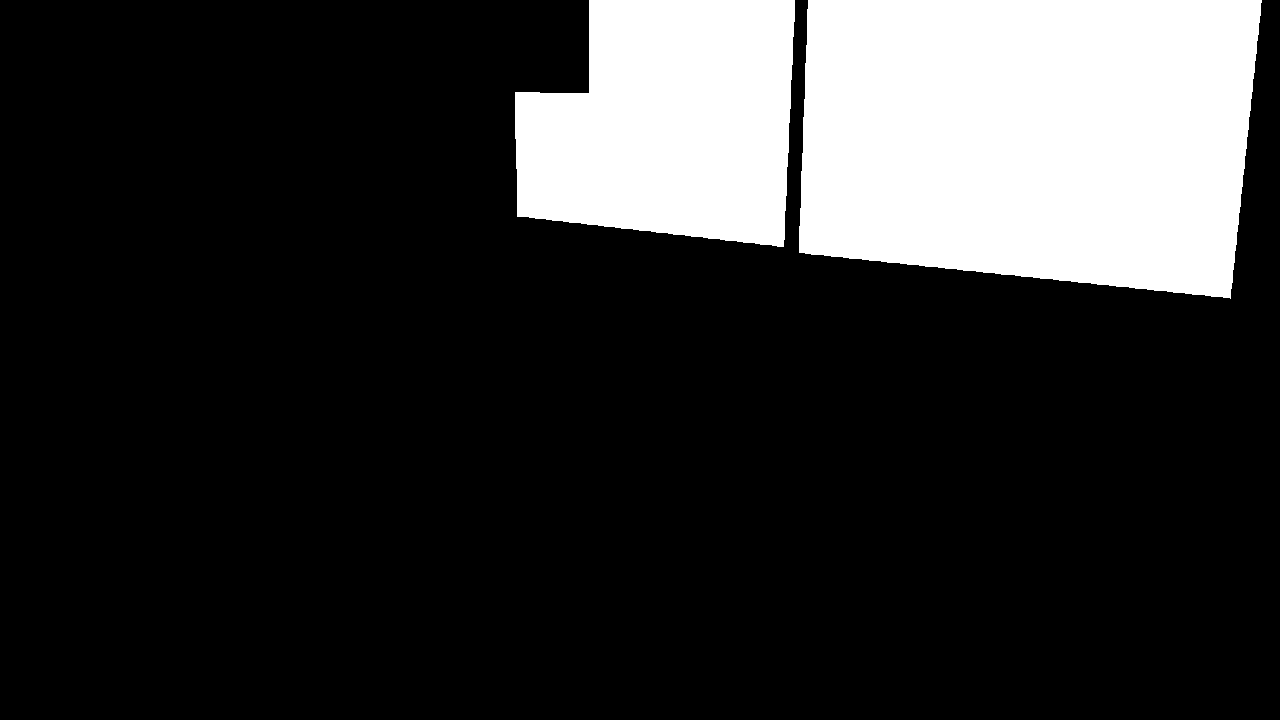}
	\end{subfigure}
	\begin{subfigure}{0.15\textwidth}
		\includegraphics[width=\textwidth]{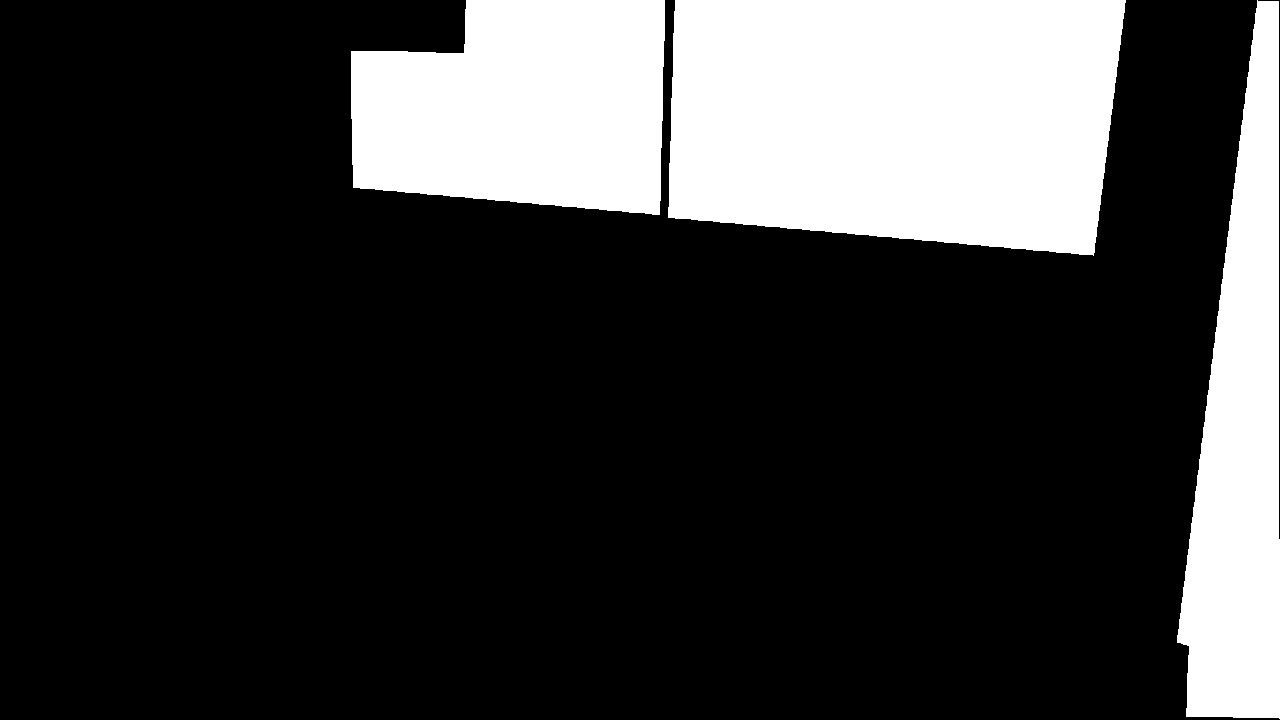}
	\end{subfigure}
	\begin{subfigure}{0.15\textwidth}
		\includegraphics[width=\textwidth]{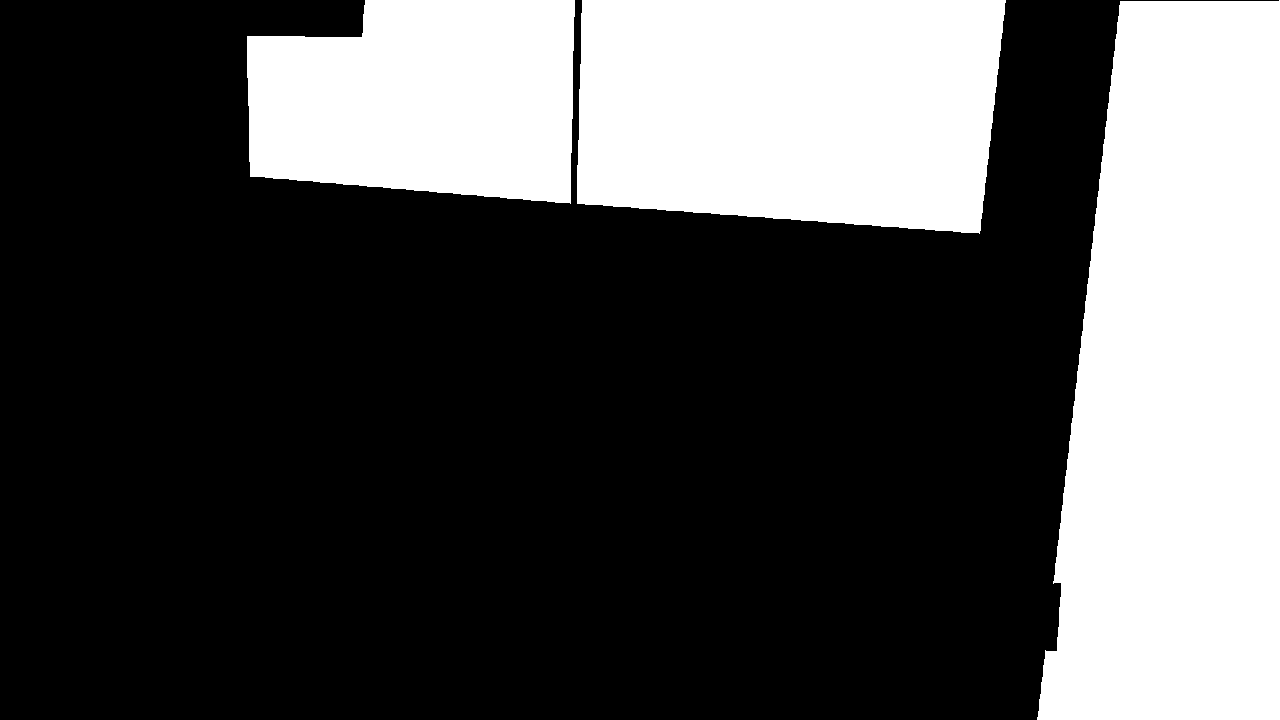}
	\end{subfigure}
	\begin{subfigure}{0.15\textwidth}
		\includegraphics[width=\textwidth]{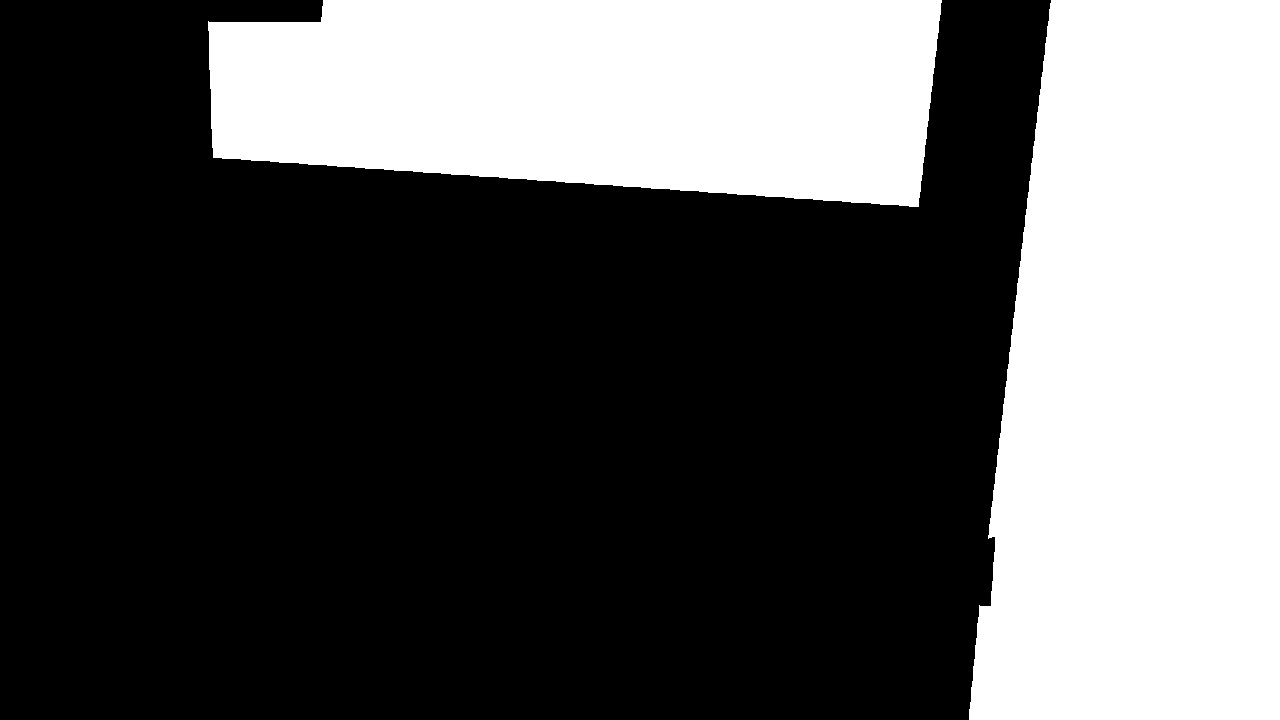}
	\end{subfigure}
	
	\vspace*{1.3mm}
	\begin{subfigure}{0.15\textwidth}
		\includegraphics[width=\textwidth]{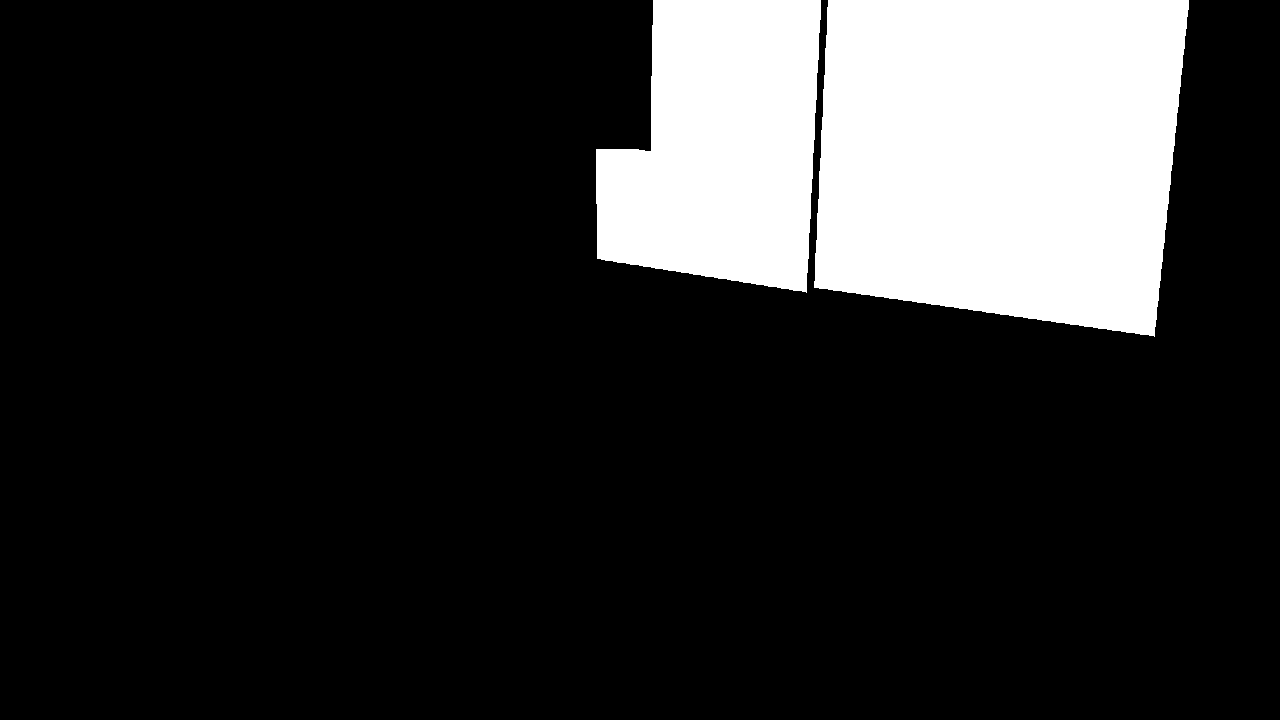}
	\end{subfigure}
	\begin{subfigure}{0.15\textwidth}
		\includegraphics[width=\textwidth]{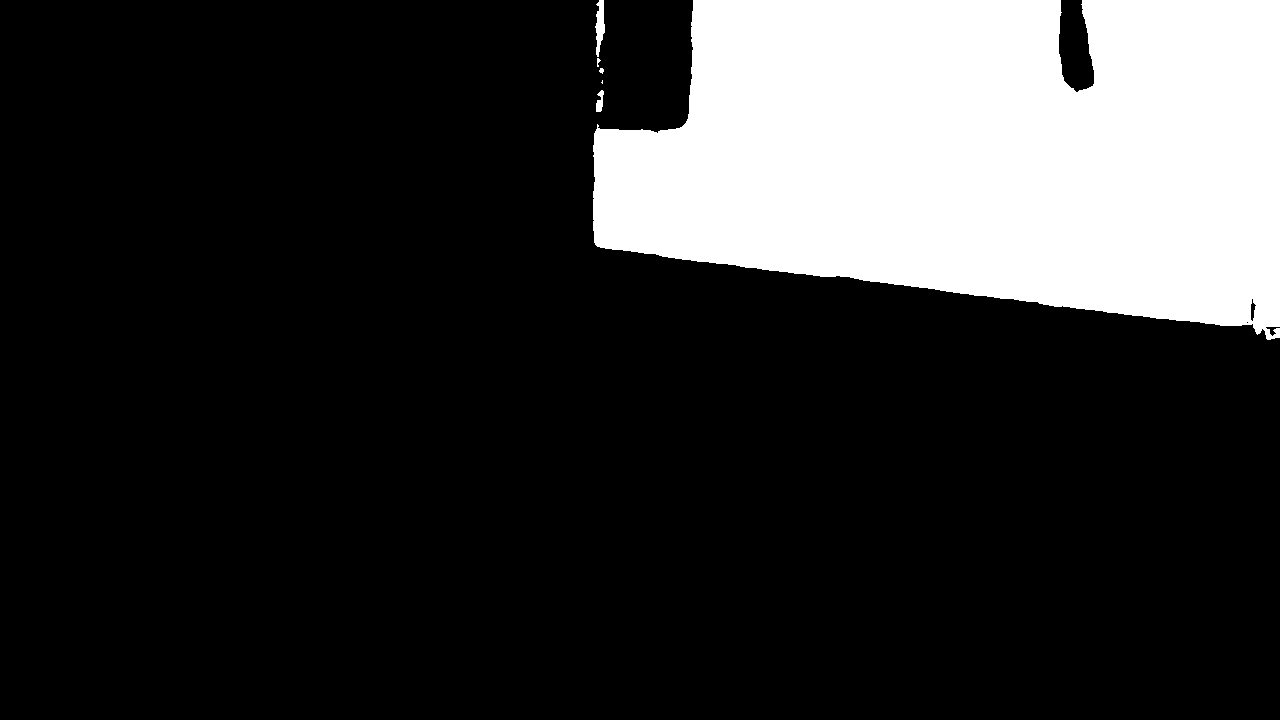}
	\end{subfigure}
	\begin{subfigure}{0.15\textwidth}
		\includegraphics[width=\textwidth]{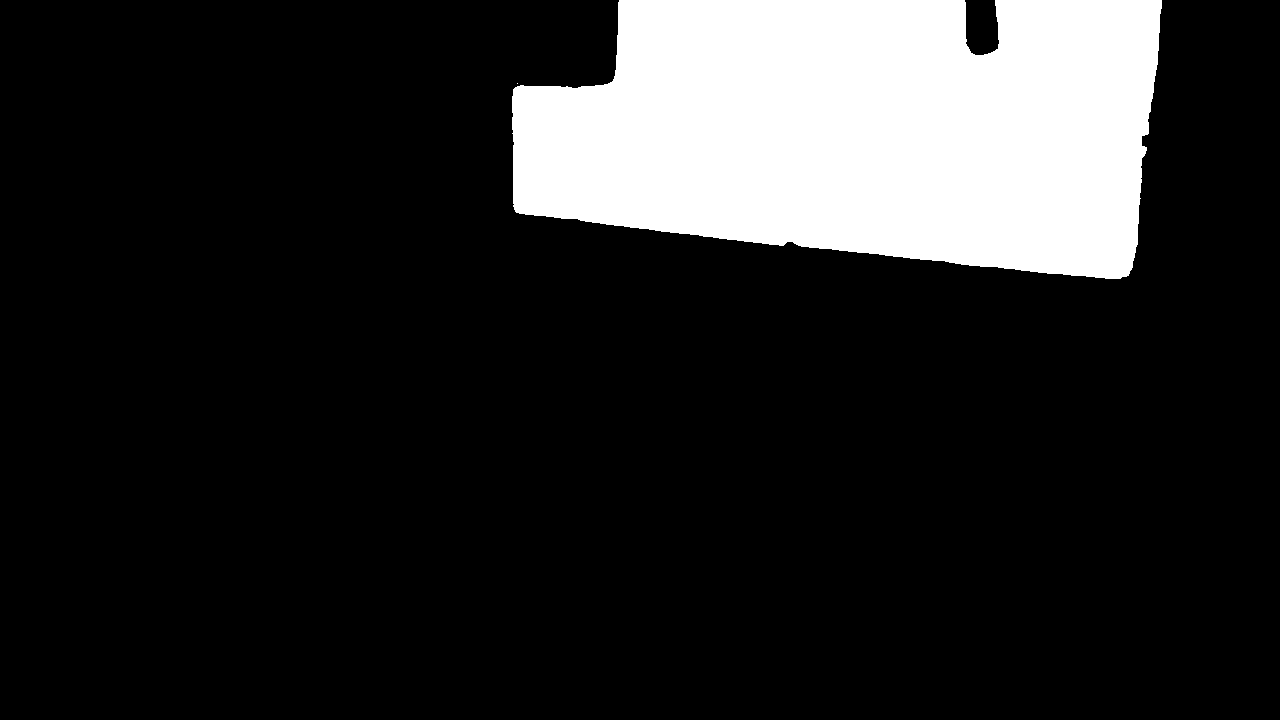}
	\end{subfigure}
	\begin{subfigure}{0.15\textwidth}
		\includegraphics[width=\textwidth]{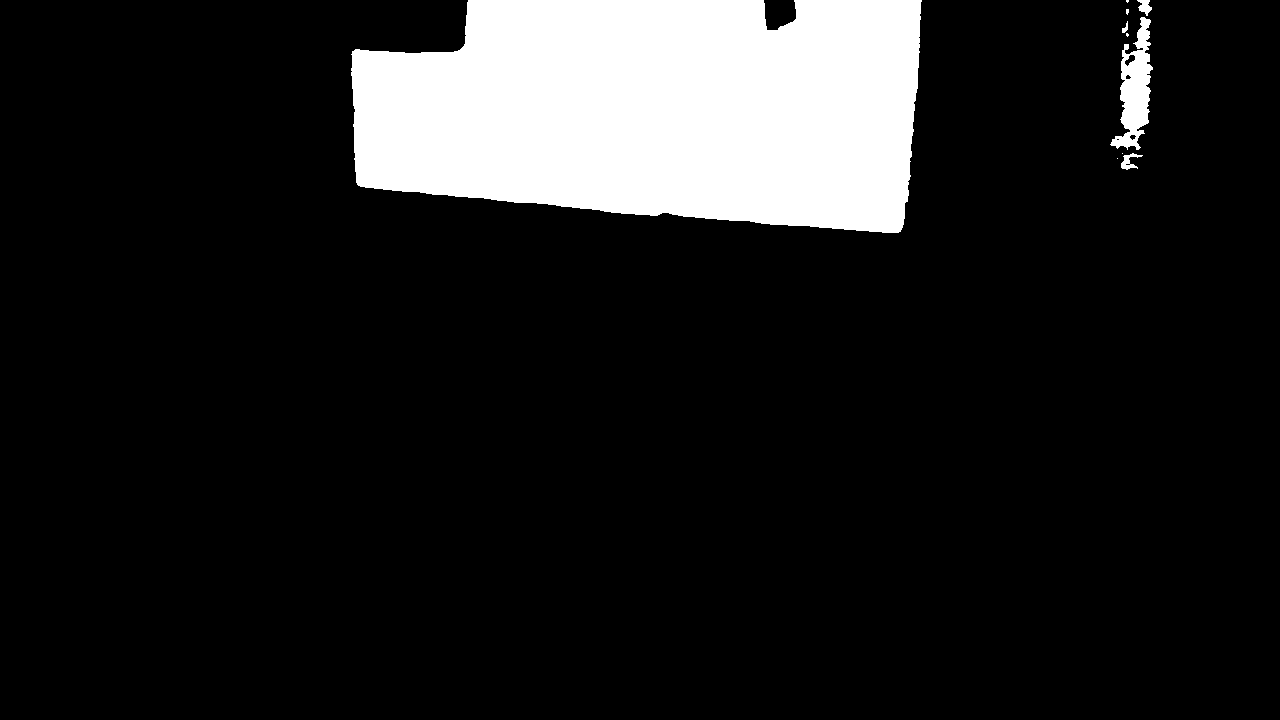}
	\end{subfigure}
	\begin{subfigure}{0.15\textwidth}
		\includegraphics[width=\textwidth]{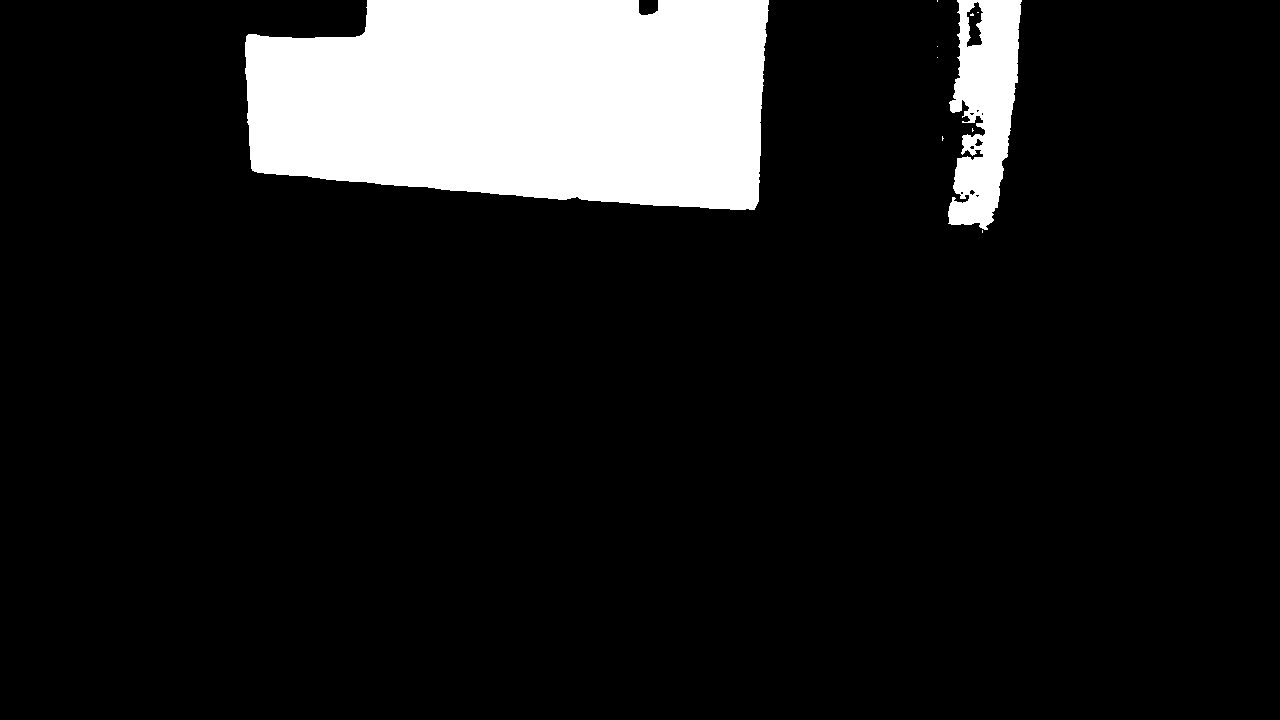}
	\end{subfigure}
	\begin{subfigure}{0.15\textwidth}
		\includegraphics[width=\textwidth]{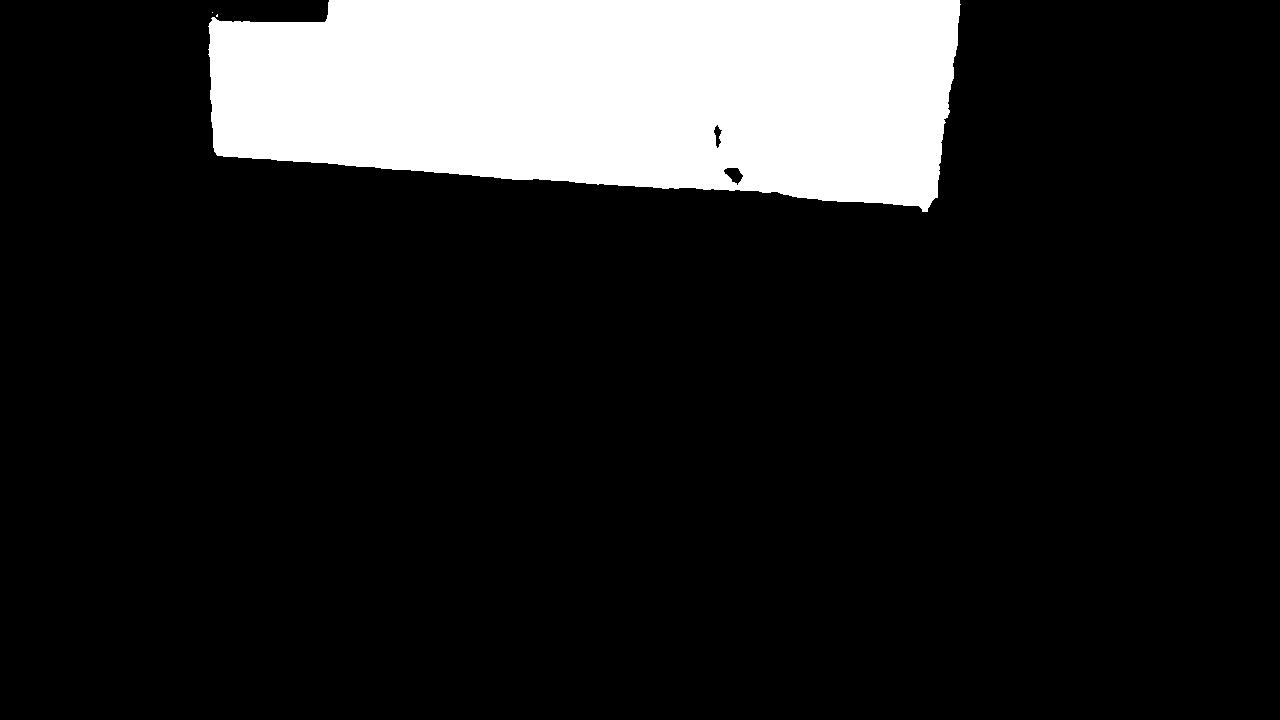}
	\end{subfigure}

	\vspace*{1.3mm}
	\begin{subfigure}{0.15\textwidth}
		\includegraphics[width=\textwidth]{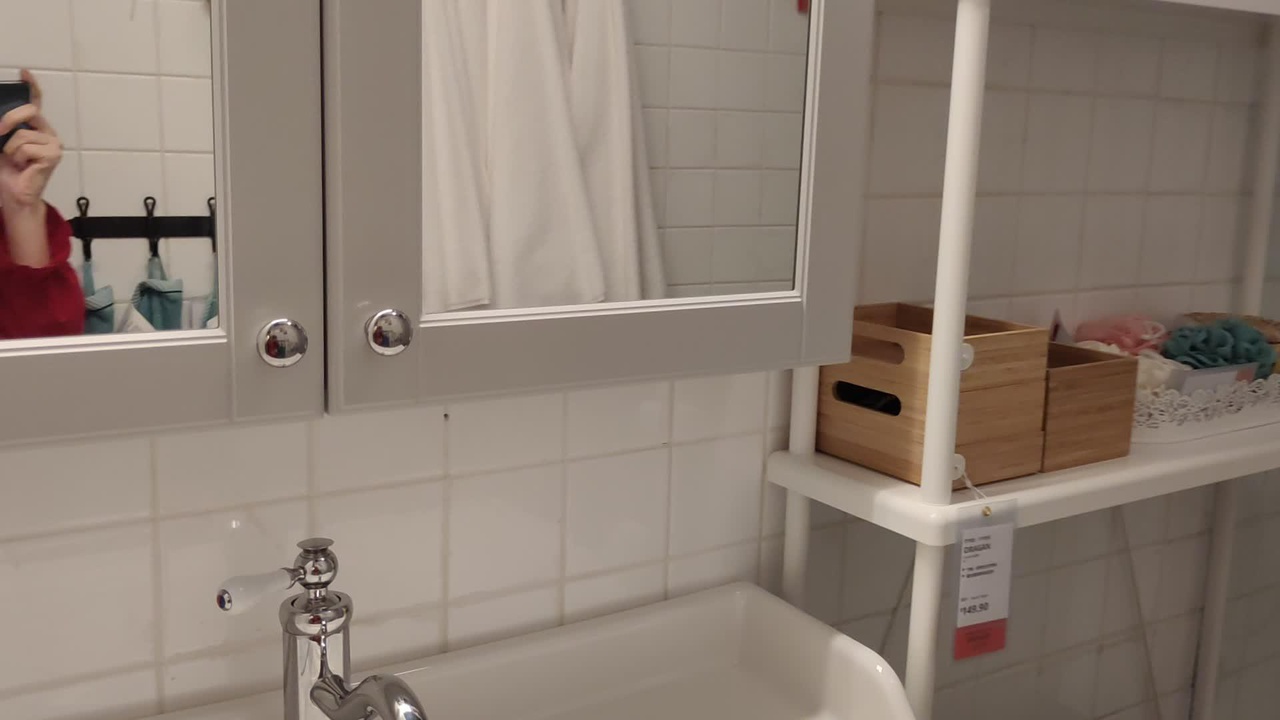}
	\end{subfigure}
	\begin{subfigure}{0.15\textwidth}
		\includegraphics[width=\textwidth]{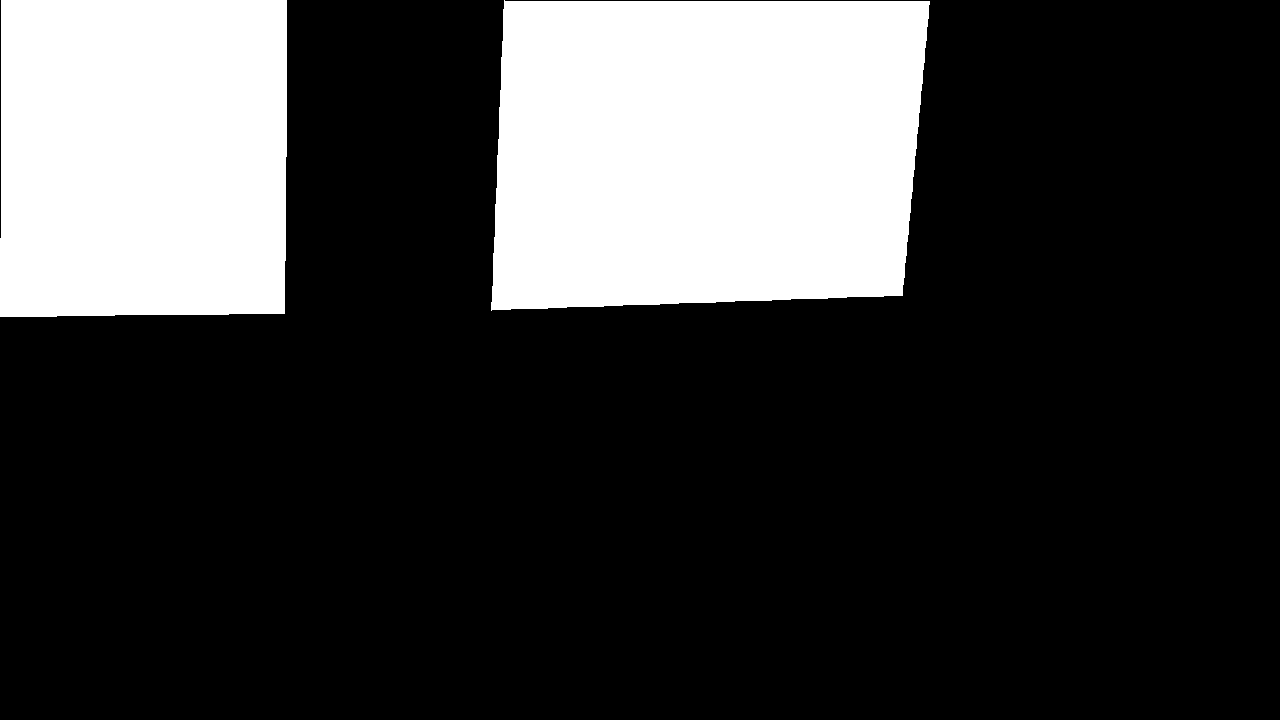}
	\end{subfigure}
	\begin{subfigure}{0.15\textwidth}
		\includegraphics[width=\textwidth]{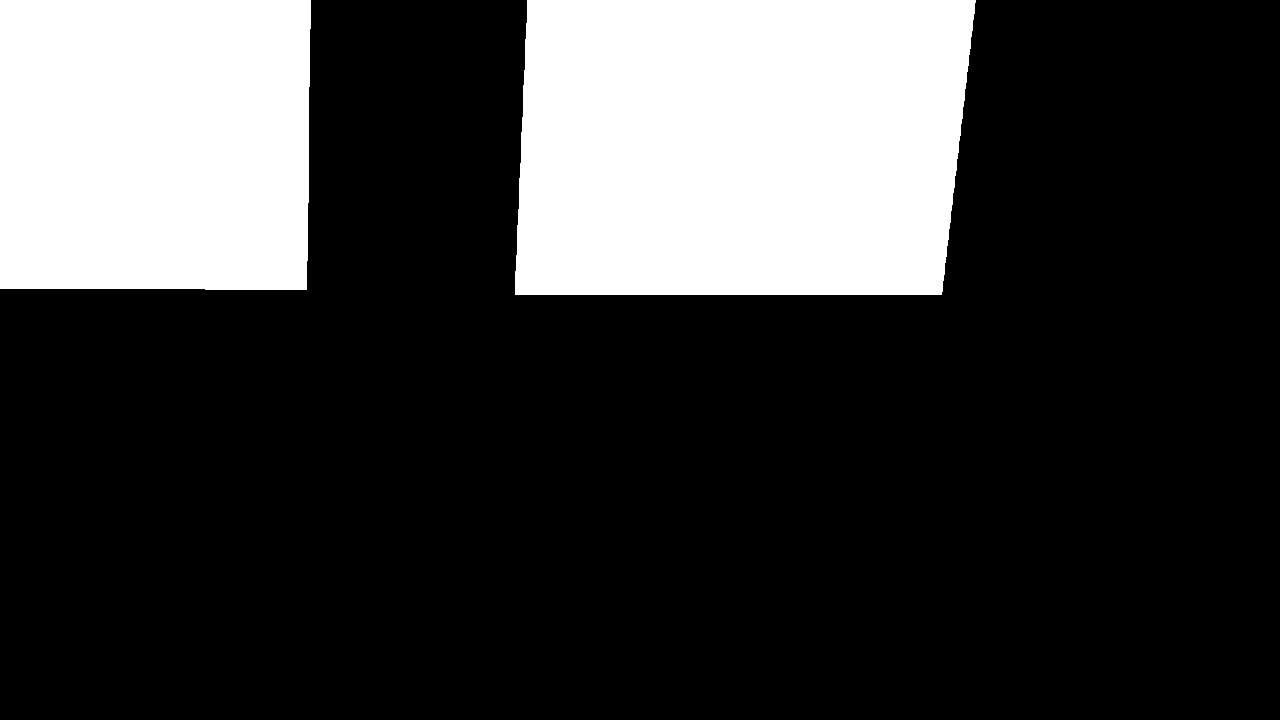}
	\end{subfigure}
	\begin{subfigure}{0.15\textwidth}
		\includegraphics[width=\textwidth]{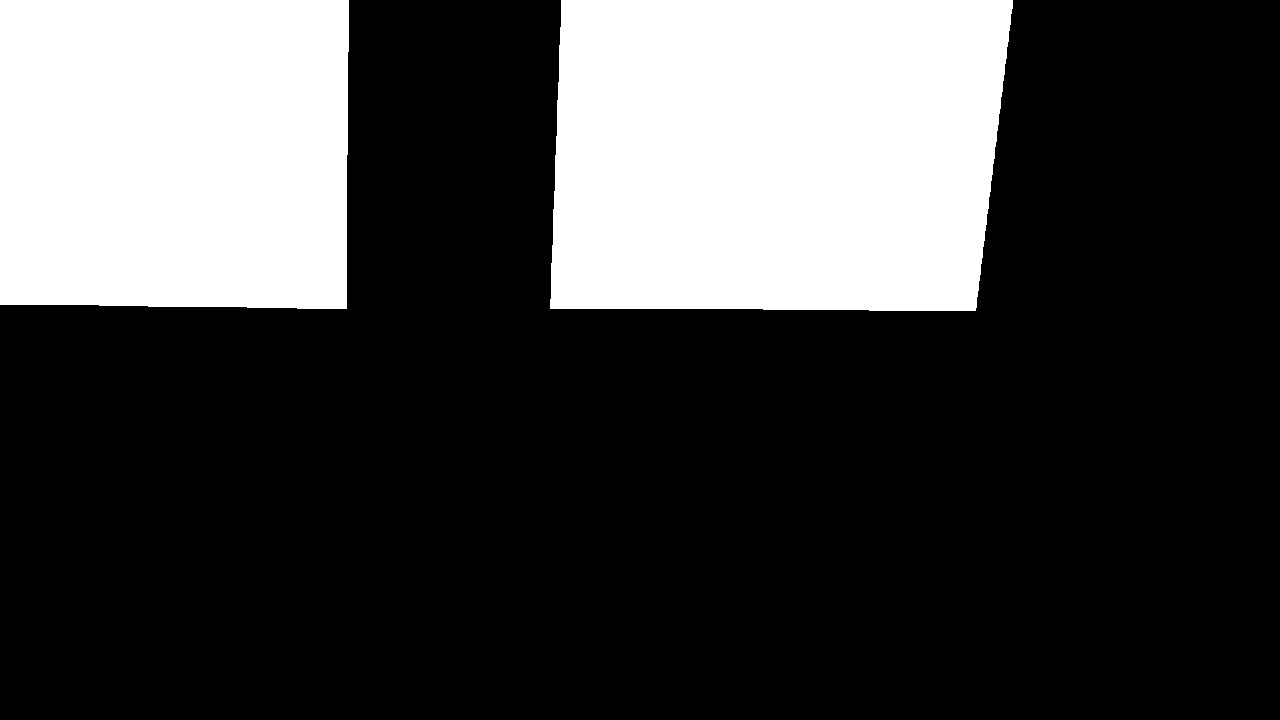}
	\end{subfigure}
	\begin{subfigure}{0.15\textwidth}
		\includegraphics[width=\textwidth]{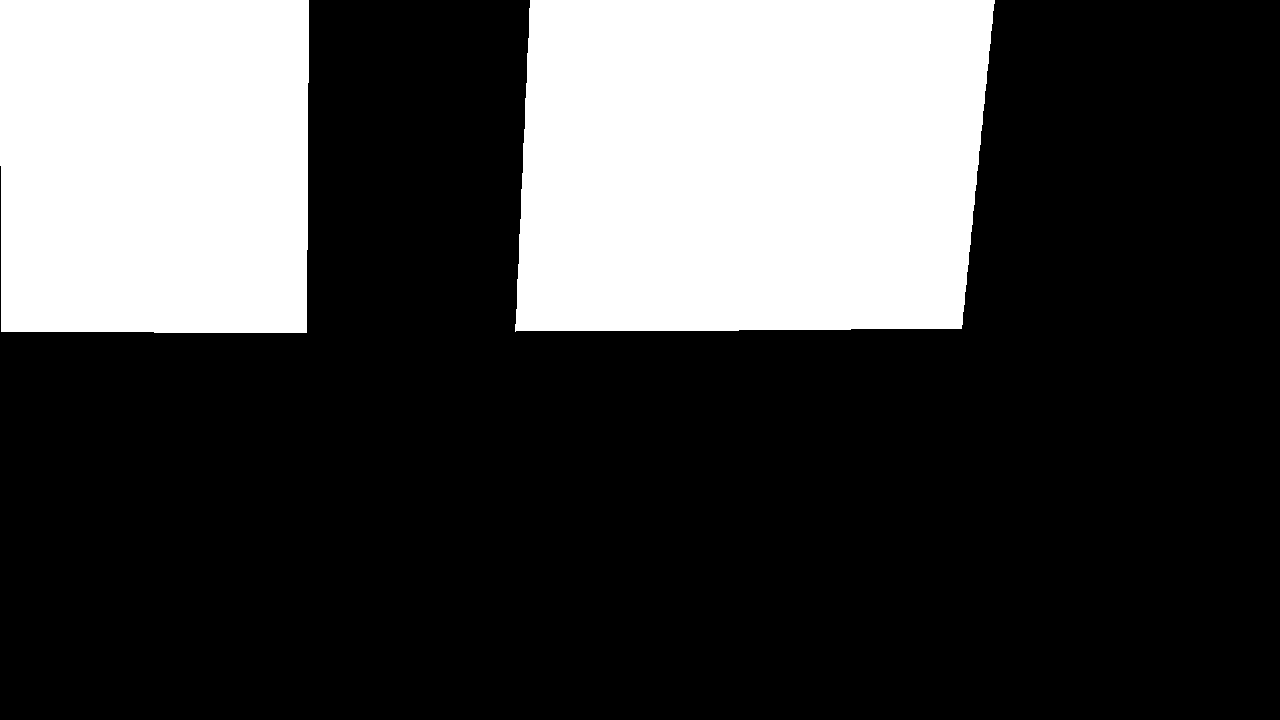}
	\end{subfigure}
	\begin{subfigure}{0.15\textwidth}
		\includegraphics[width=\textwidth]{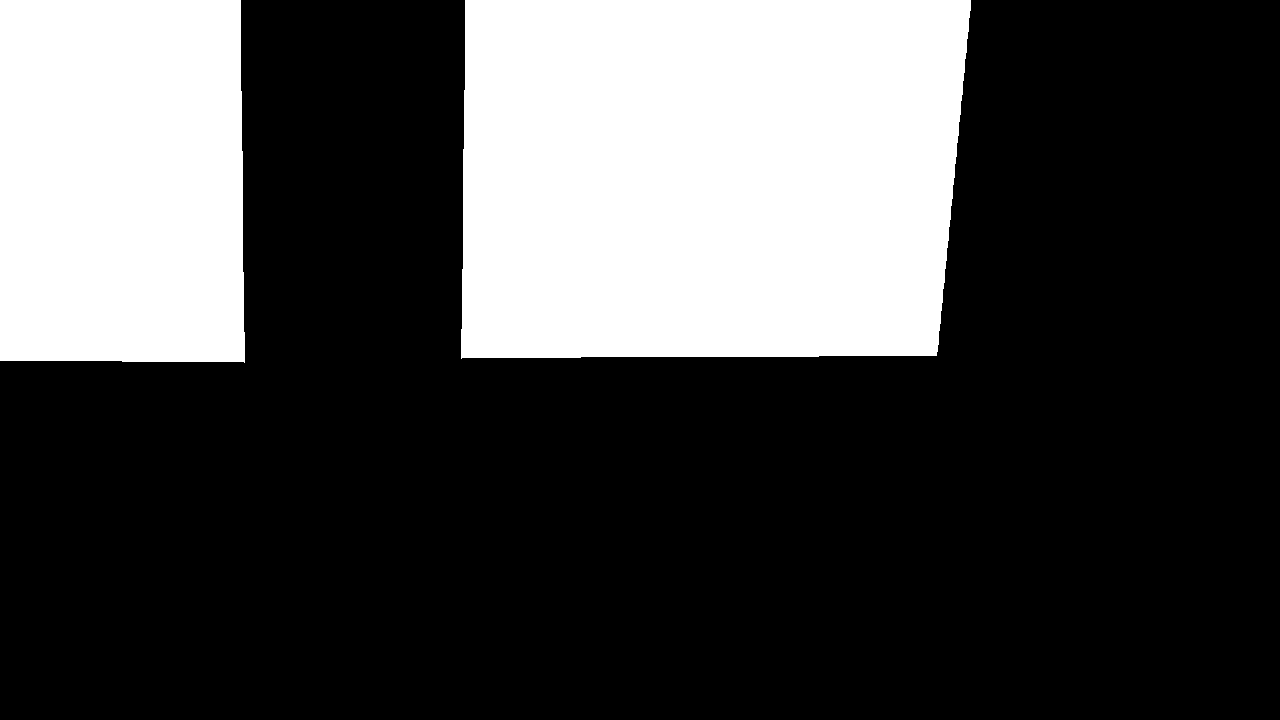}
	\end{subfigure}
	
	\vspace*{1.3mm}
	\begin{subfigure}{0.15\textwidth}
		\includegraphics[width=\textwidth]{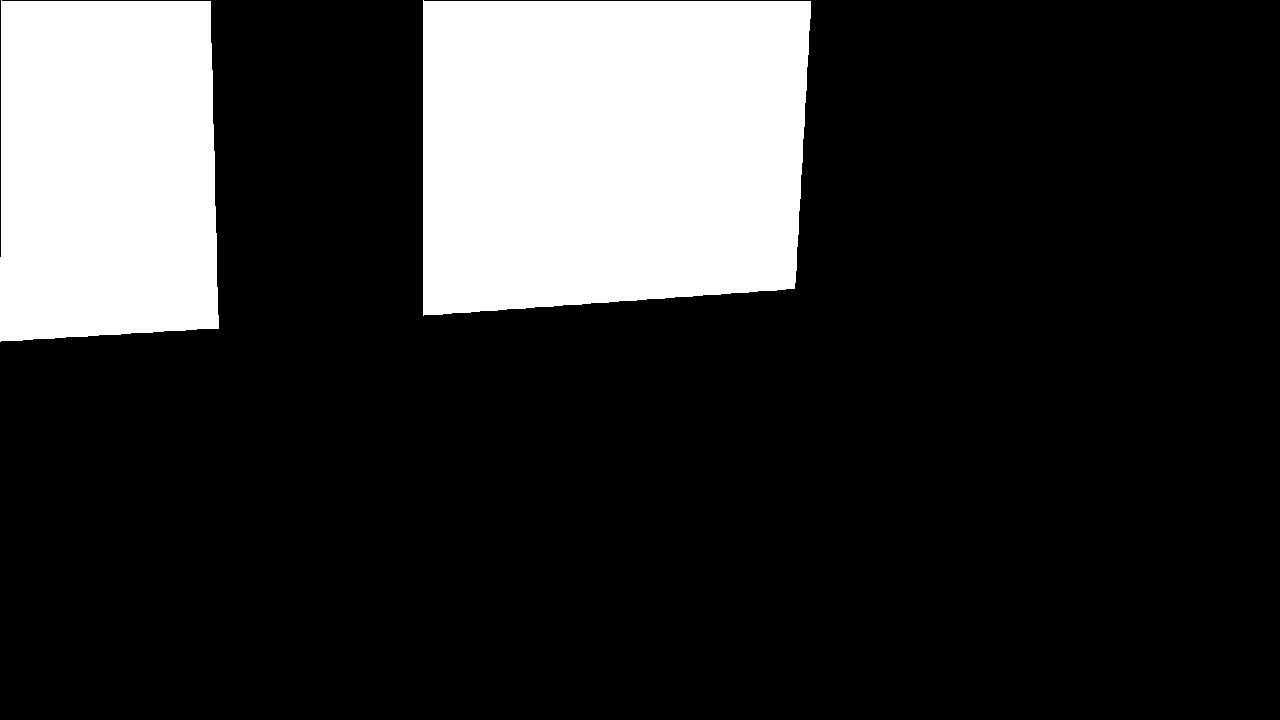}
	\end{subfigure}
	\begin{subfigure}{0.15\textwidth}
		\includegraphics[width=\textwidth]{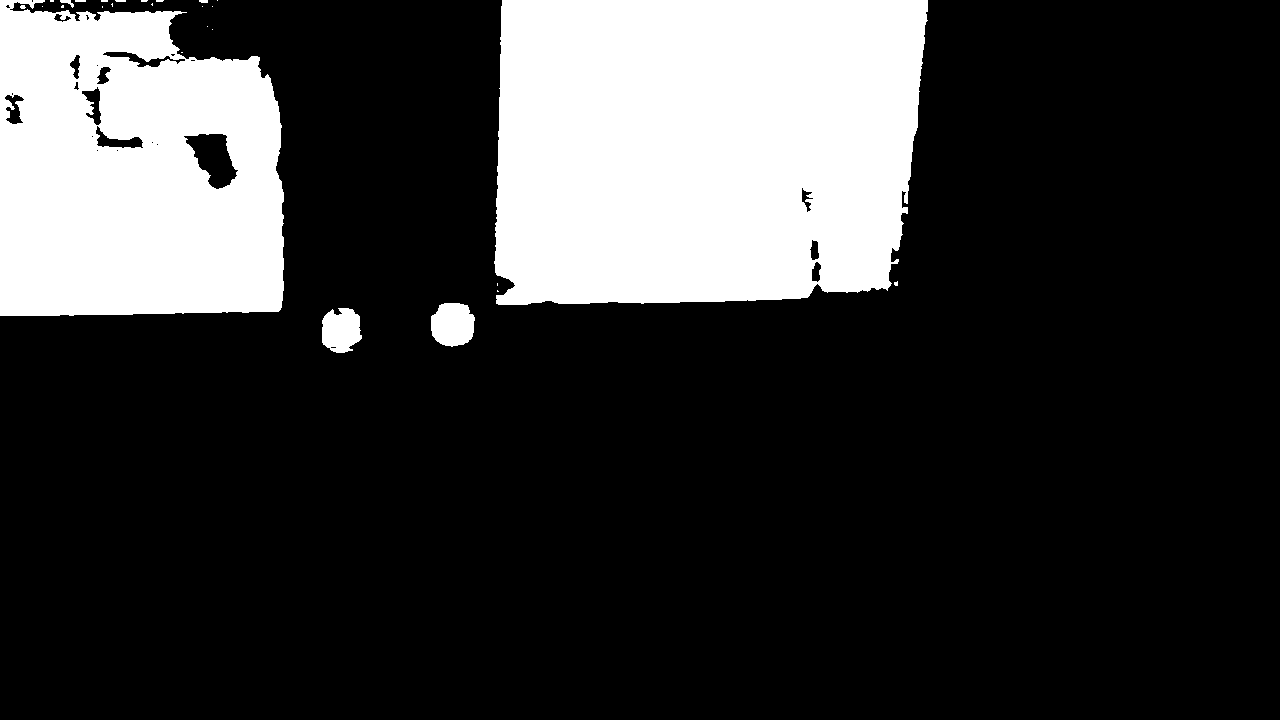}
	\end{subfigure}
	\begin{subfigure}{0.15\textwidth}
		\includegraphics[width=\textwidth]{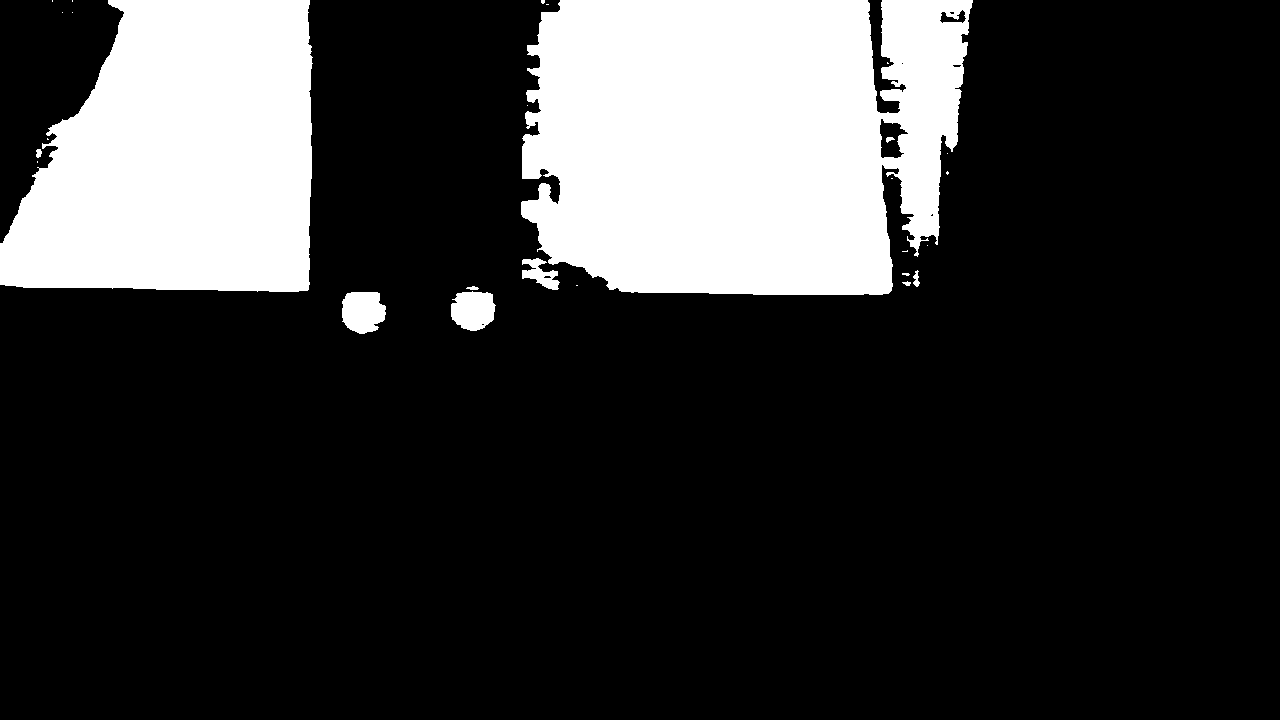}
	\end{subfigure}
	\begin{subfigure}{0.15\textwidth}
		\includegraphics[width=\textwidth]{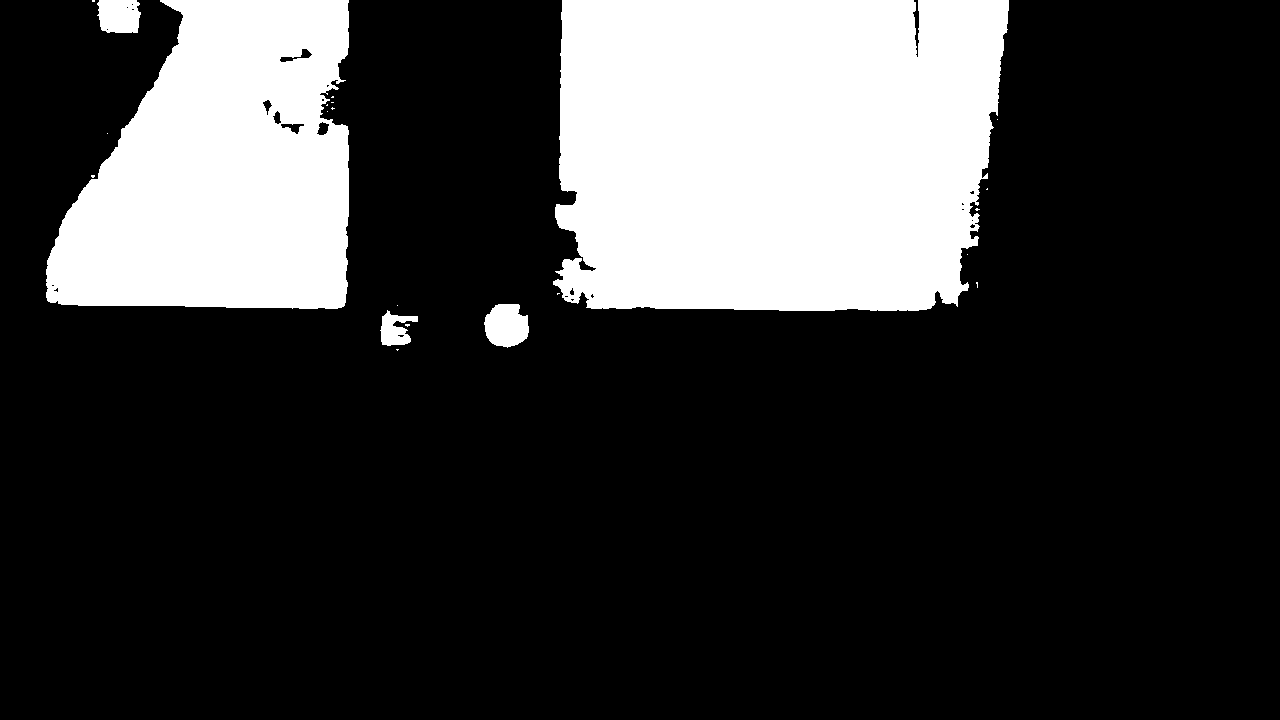}
	\end{subfigure}
	\begin{subfigure}{0.15\textwidth}
		\includegraphics[width=\textwidth]{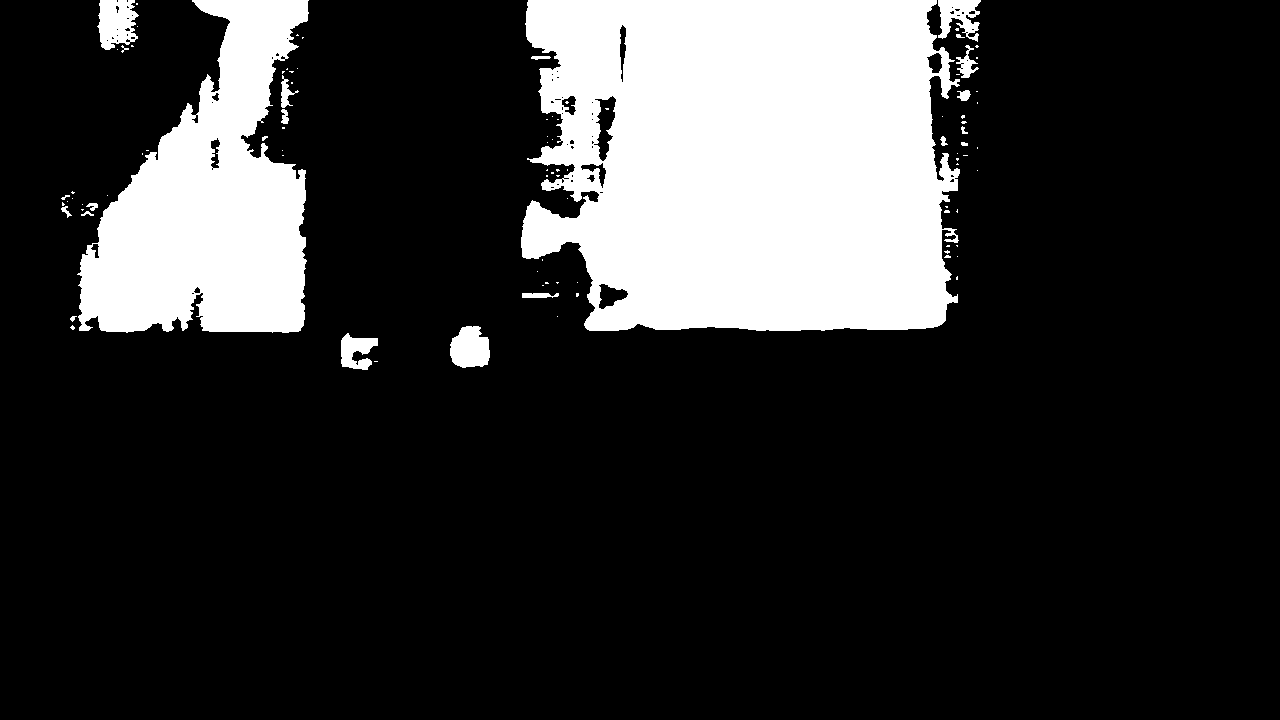}
	\end{subfigure}
	\begin{subfigure}{0.15\textwidth}
		\includegraphics[width=\textwidth]{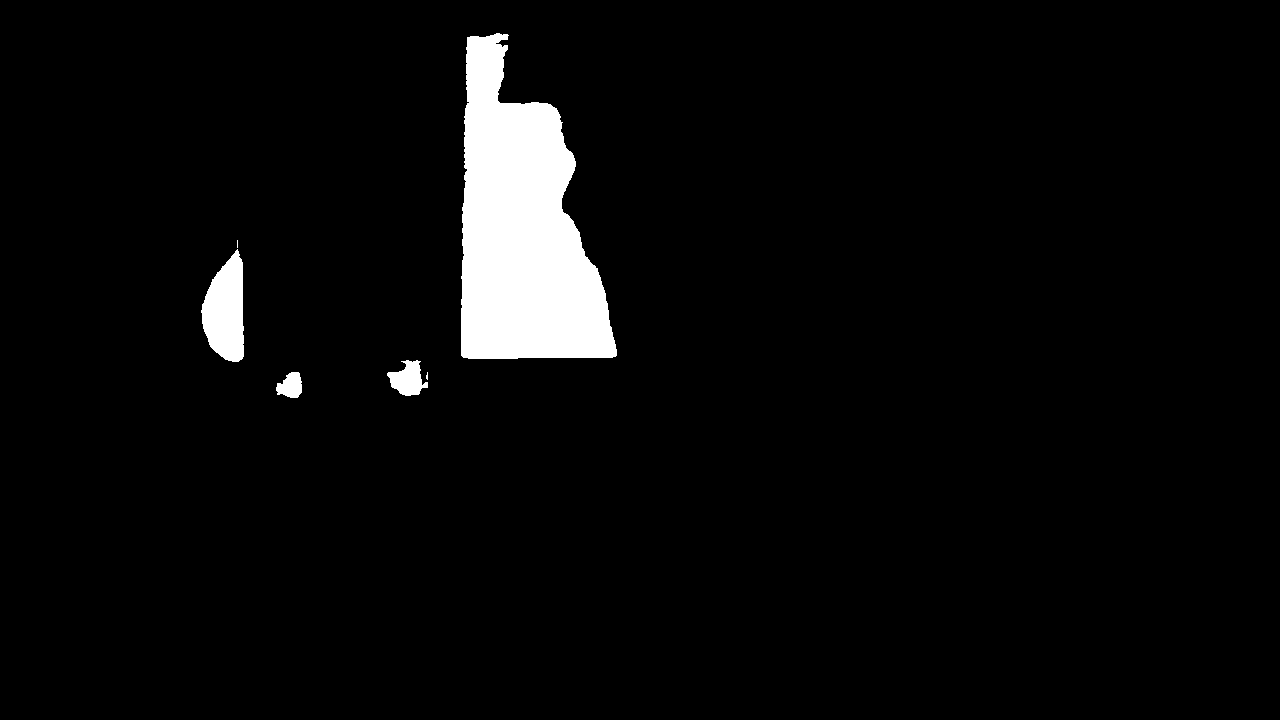}
	\end{subfigure}
	
	\vspace*{1.3mm}
	\begin{subfigure}{0.15\textwidth}
		\includegraphics[width=\textwidth]{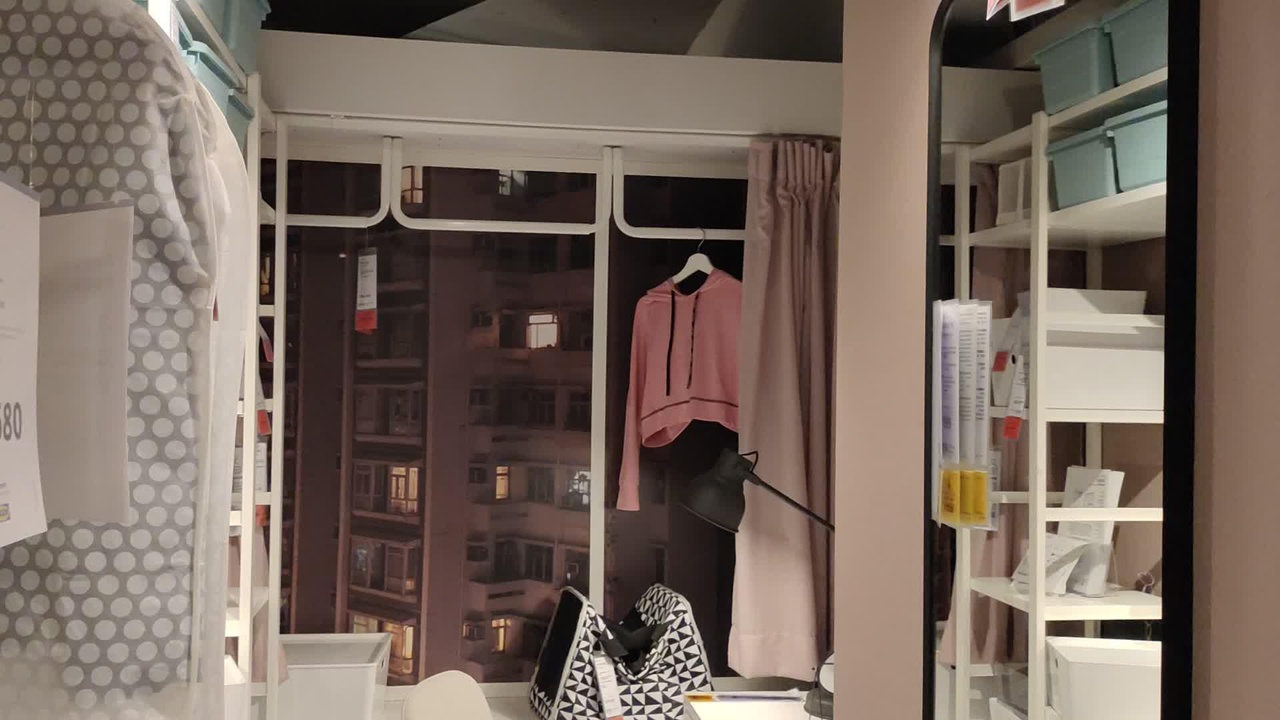}
	\end{subfigure}
	\begin{subfigure}{0.15\textwidth}
		\includegraphics[width=\textwidth]{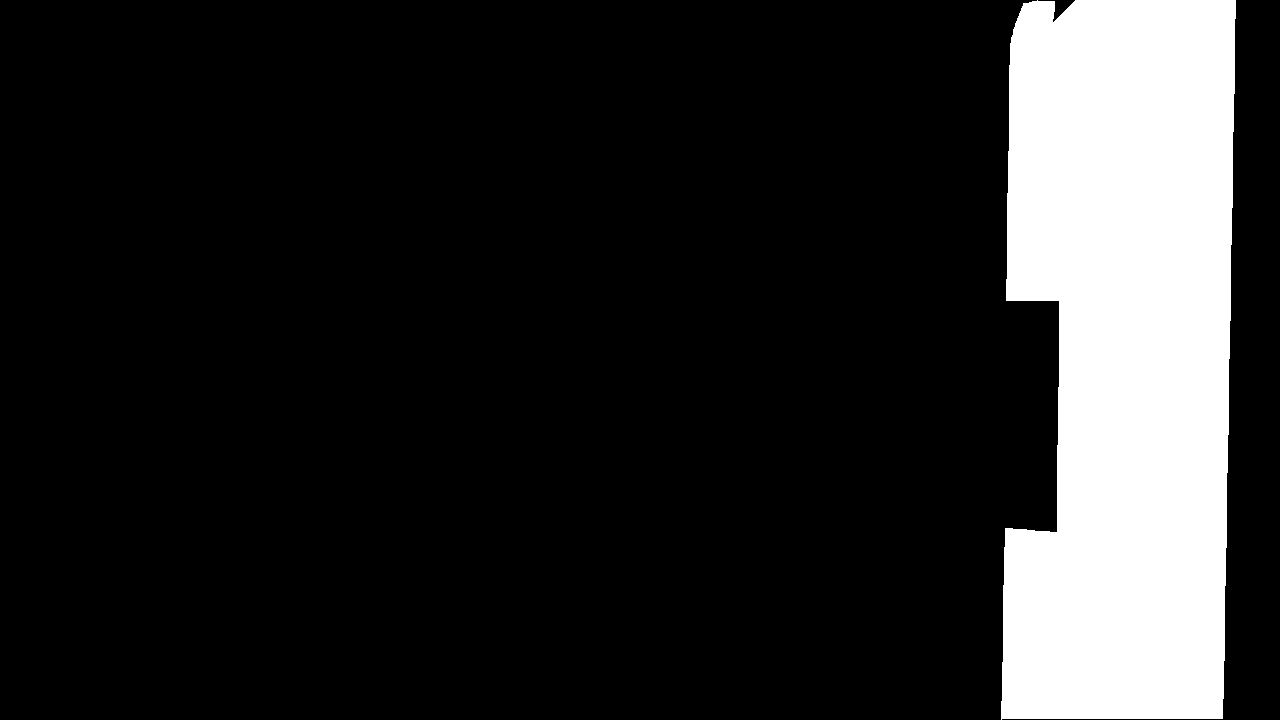}
	\end{subfigure}
	\begin{subfigure}{0.15\textwidth}
		\includegraphics[width=\textwidth]{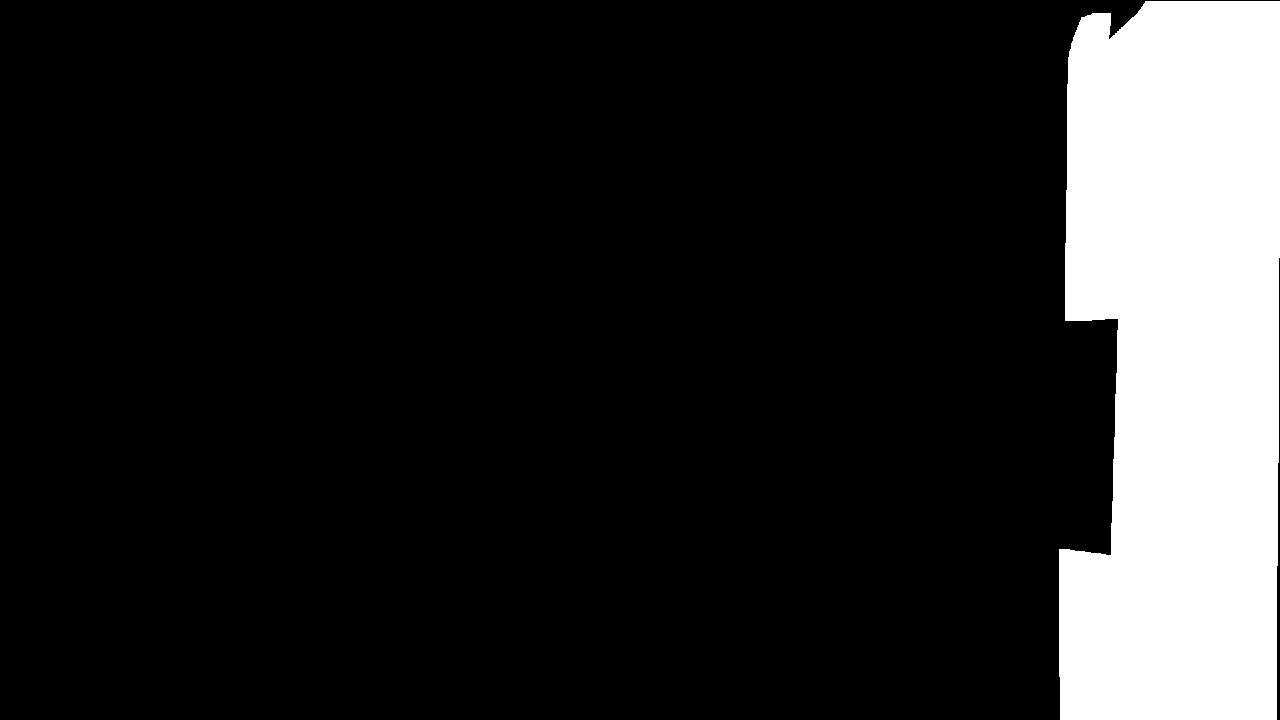}
	\end{subfigure}
	\begin{subfigure}{0.15\textwidth}
		\includegraphics[width=\textwidth]{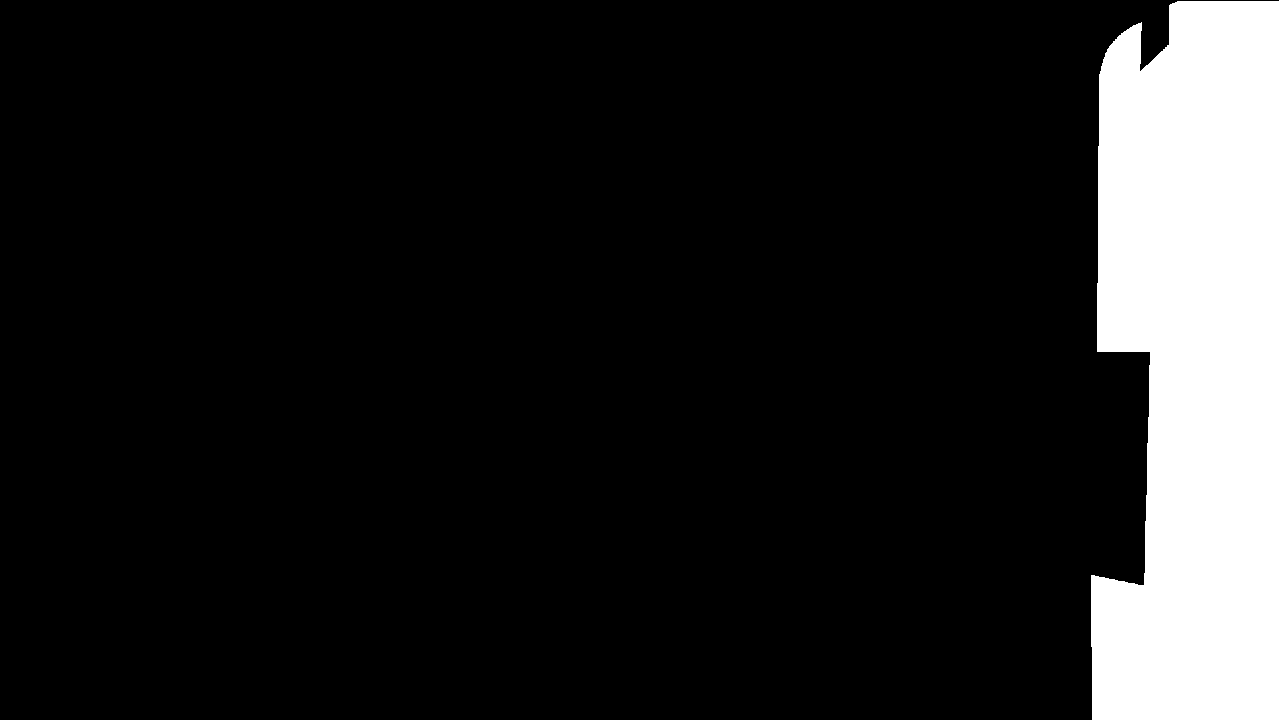}
	\end{subfigure}
	\begin{subfigure}{0.15\textwidth}
		\includegraphics[width=\textwidth]{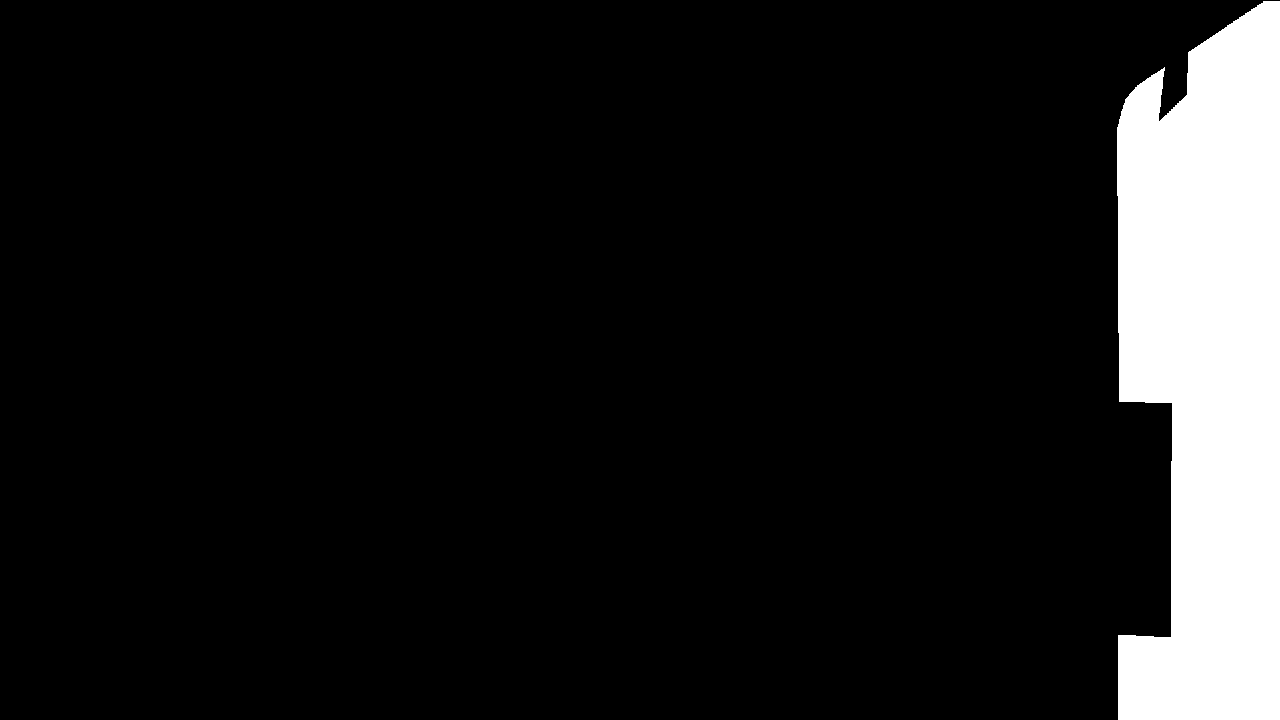}
	\end{subfigure}
	\begin{subfigure}{0.15\textwidth}
		\includegraphics[width=\textwidth]{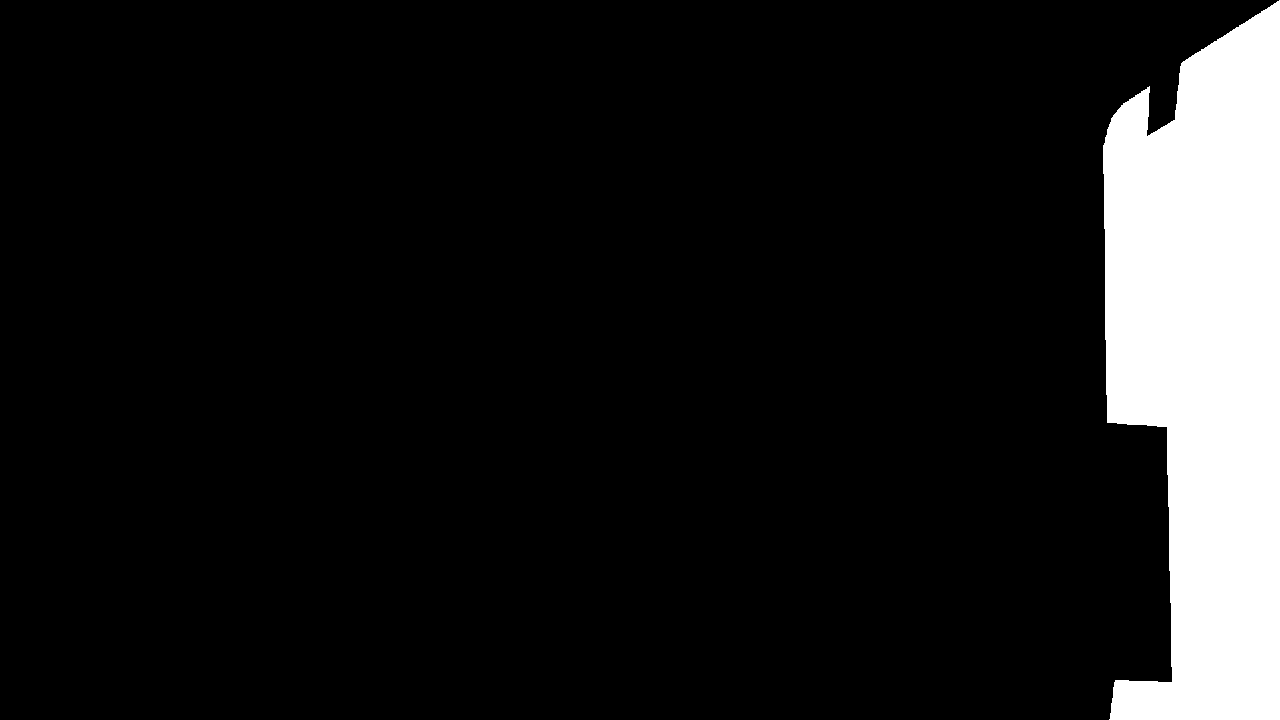}
	\end{subfigure}
	
	\vspace*{1.3mm}
	\begin{subfigure}{0.15\textwidth}
		\includegraphics[width=\textwidth]{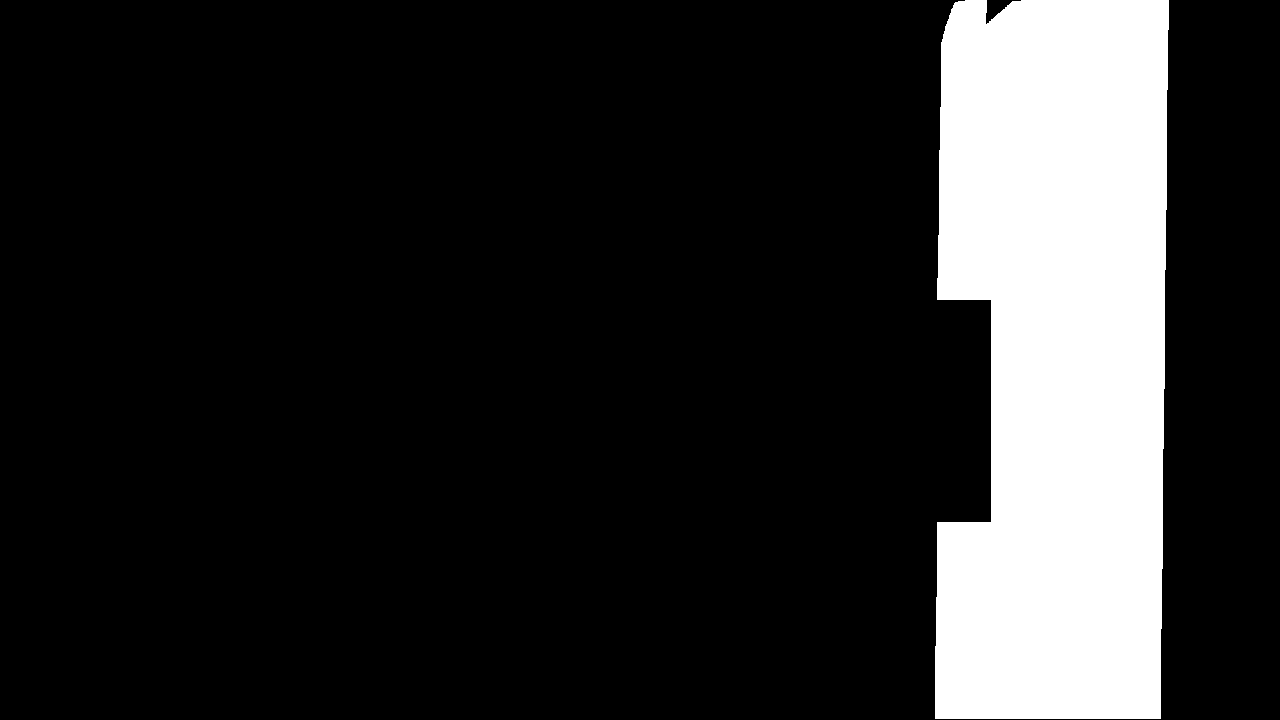}
	\end{subfigure}
	\begin{subfigure}{0.15\textwidth}
		\includegraphics[width=\textwidth]{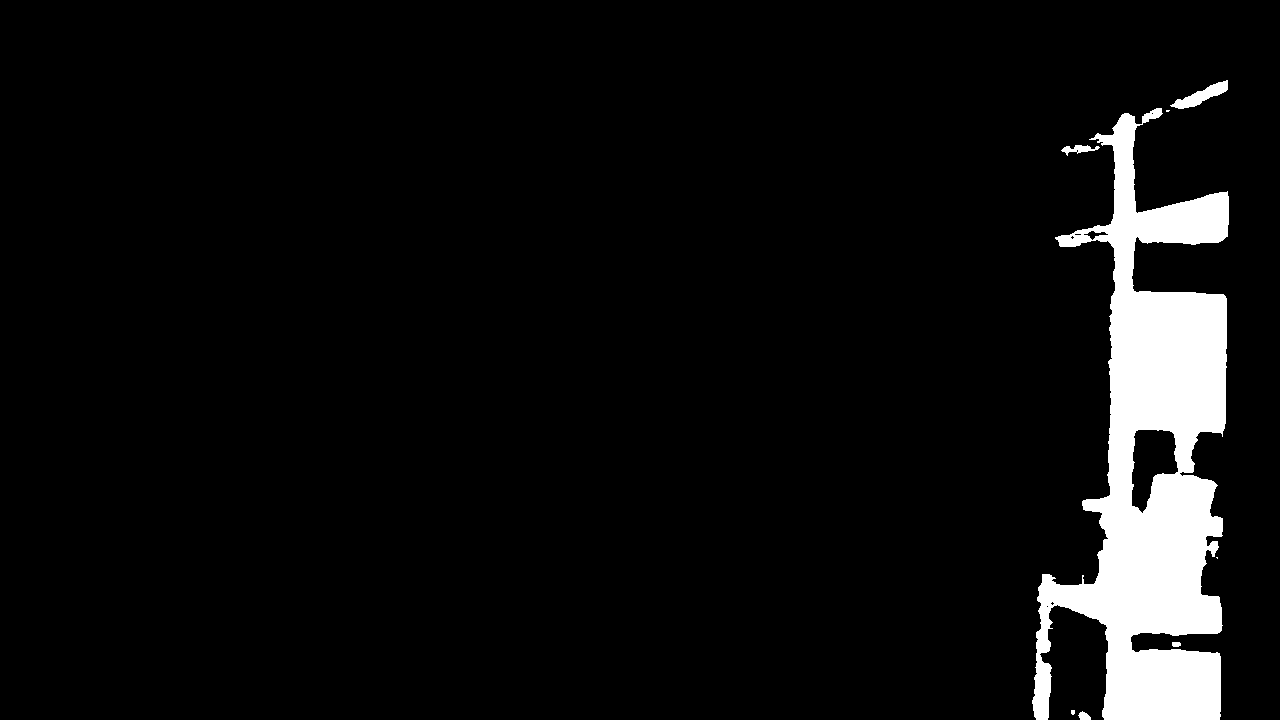}
	\end{subfigure}
	\begin{subfigure}{0.15\textwidth}
		\includegraphics[width=\textwidth]{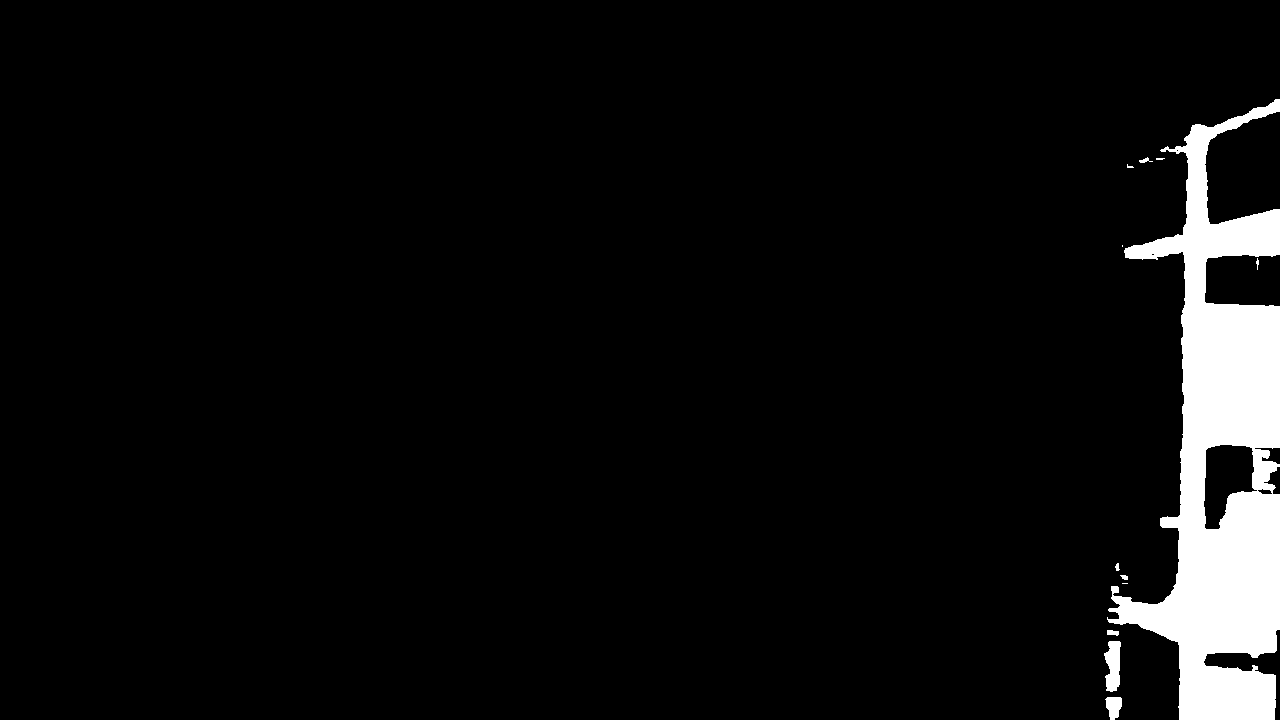}
	\end{subfigure}
	\begin{subfigure}{0.15\textwidth}
		\includegraphics[width=\textwidth]{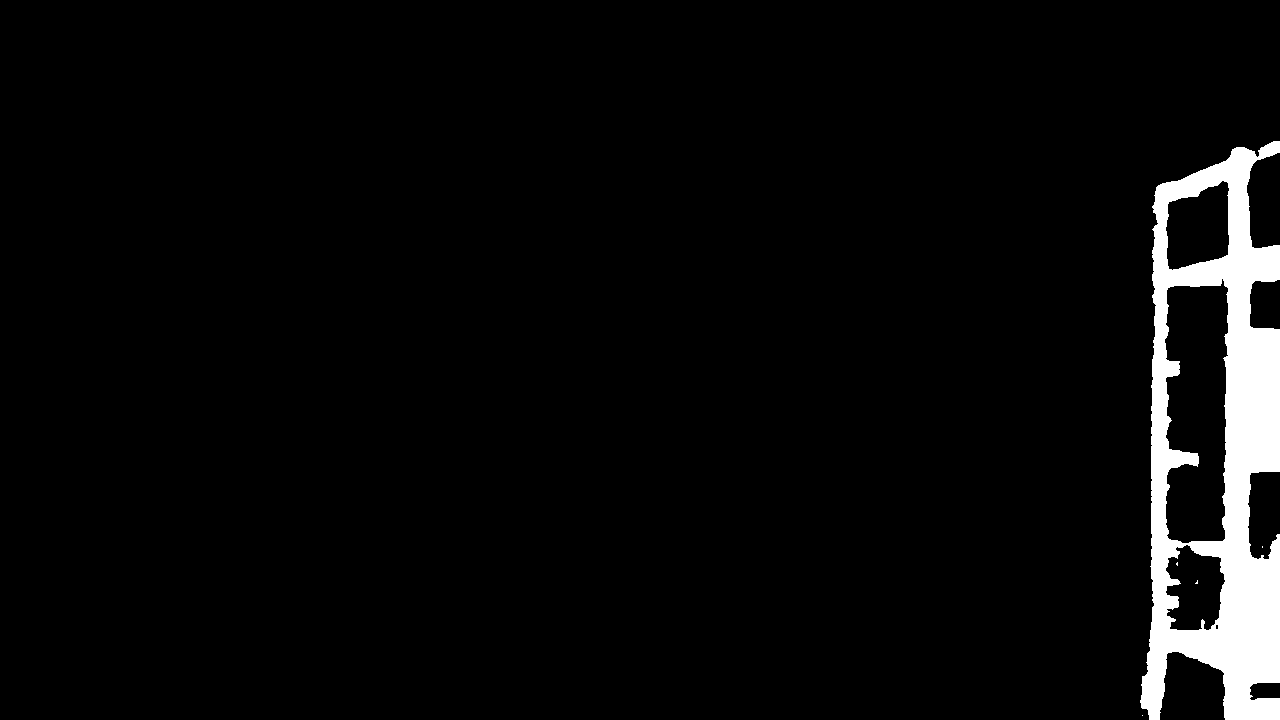}
	\end{subfigure}
	\begin{subfigure}{0.15\textwidth}
		\includegraphics[width=\textwidth]{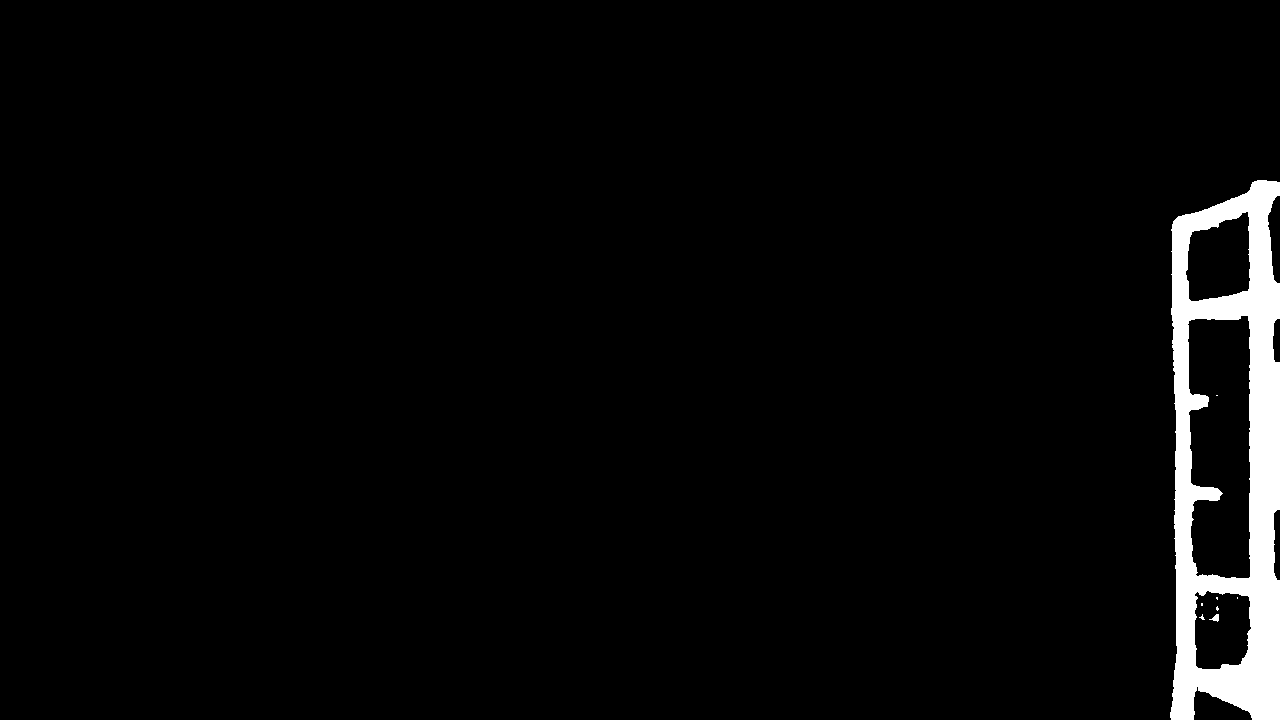}
	\end{subfigure}
	\begin{subfigure}{0.15\textwidth}
		\includegraphics[width=\textwidth]{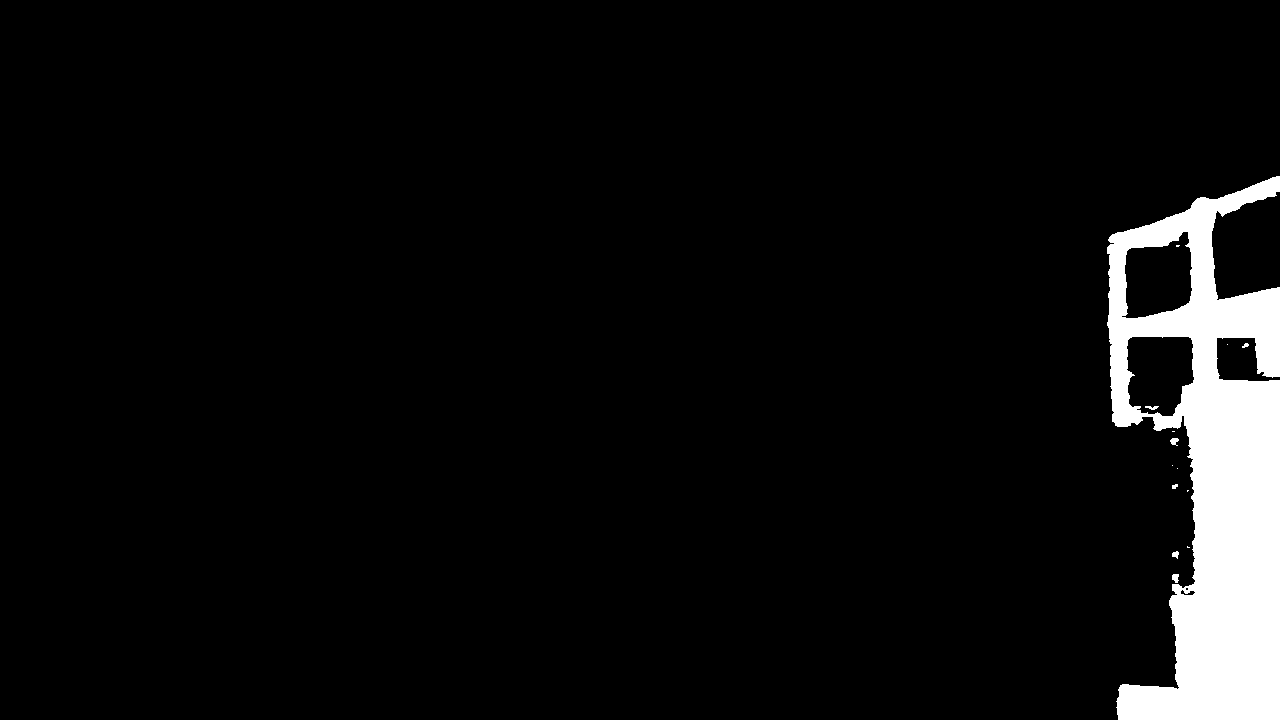}
	\end{subfigure}

	\vspace*{1.3mm}
	\begin{subfigure}{0.15\textwidth}
		\includegraphics[width=\textwidth]{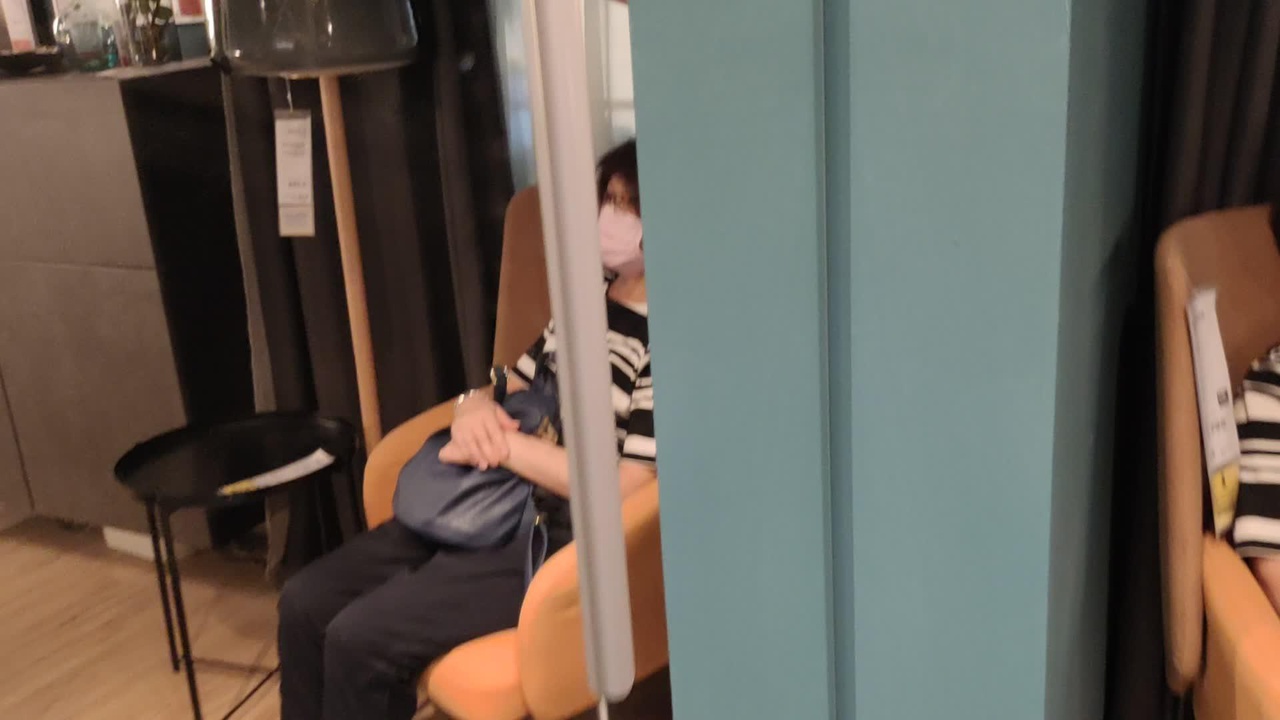}
	\end{subfigure}
	\begin{subfigure}{0.15\textwidth}
		\includegraphics[width=\textwidth]{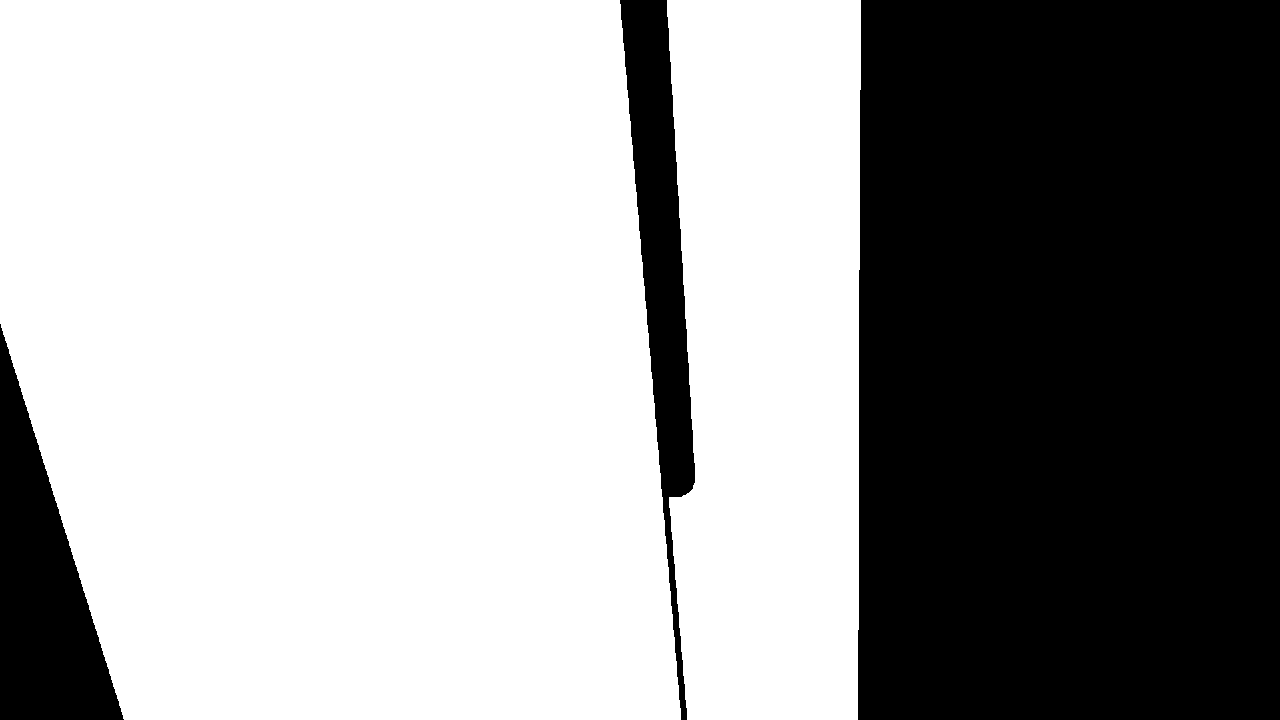}
	\end{subfigure}
	\begin{subfigure}{0.15\textwidth}
		\includegraphics[width=\textwidth]{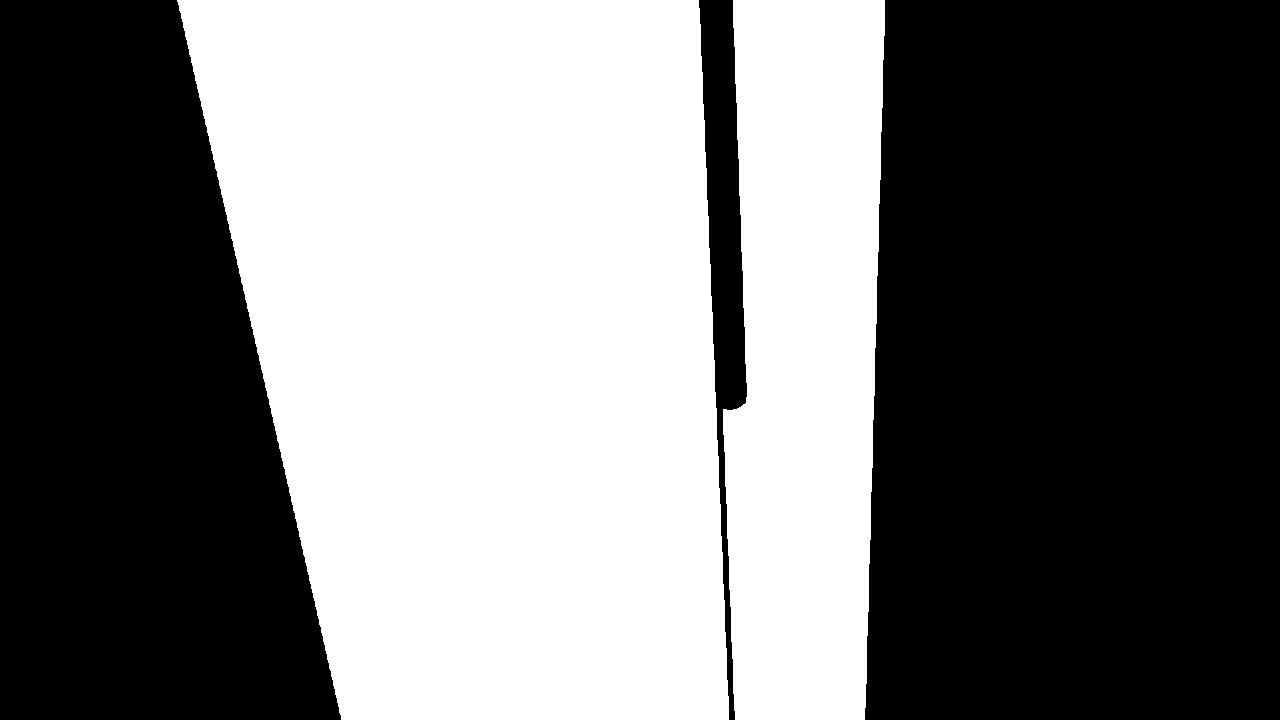}
	\end{subfigure}
	\begin{subfigure}{0.15\textwidth}
		\includegraphics[width=\textwidth]{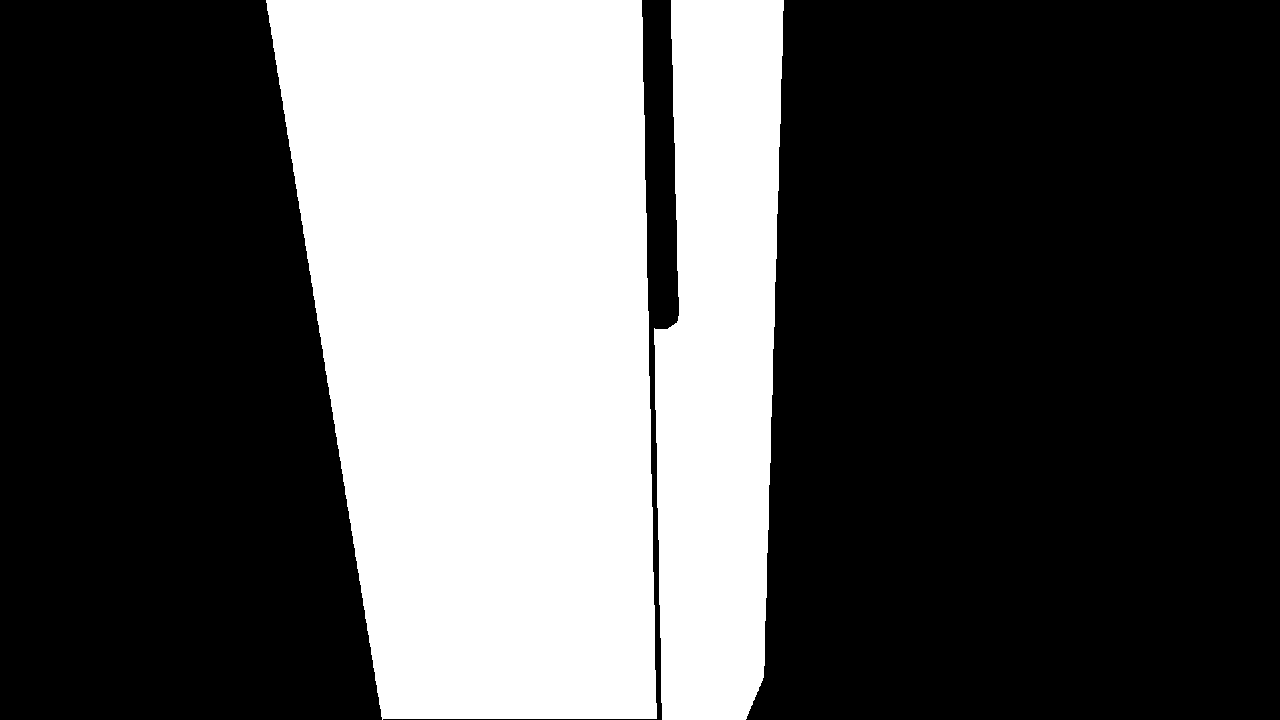}
	\end{subfigure}
	\begin{subfigure}{0.15\textwidth}
		\includegraphics[width=\textwidth]{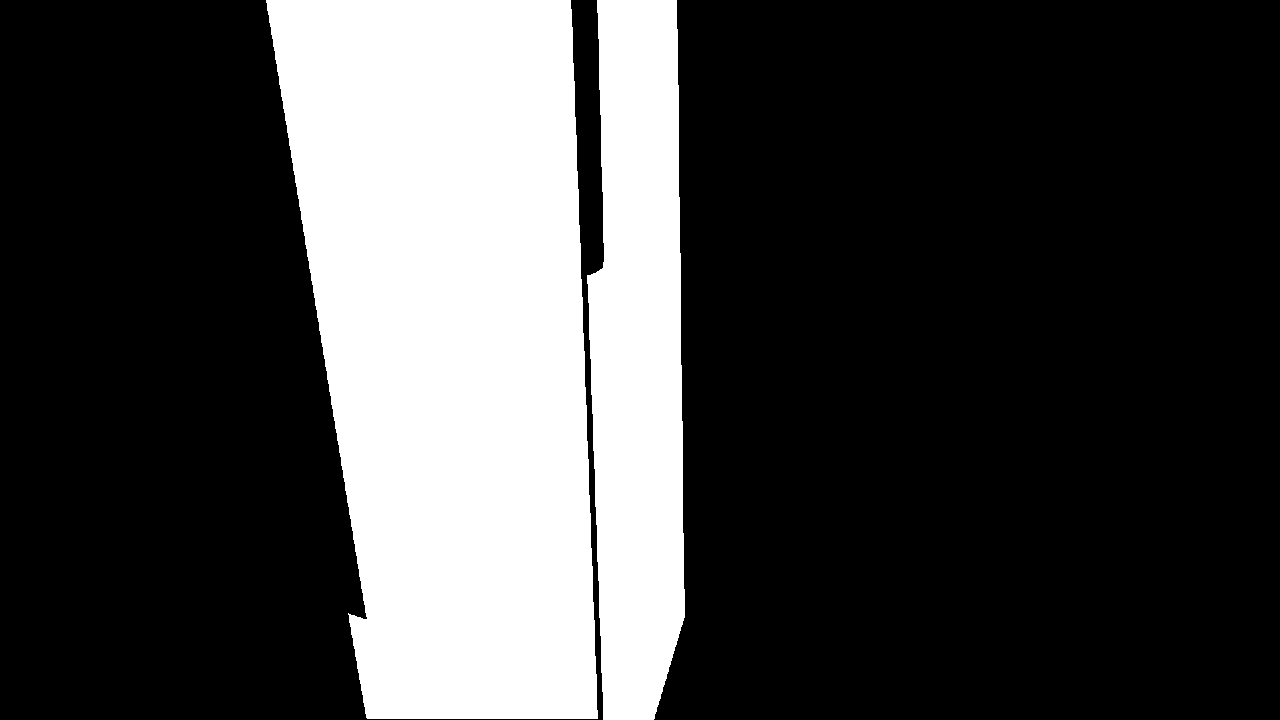}
	\end{subfigure}
	\begin{subfigure}{0.15\textwidth}
		\includegraphics[width=\textwidth]{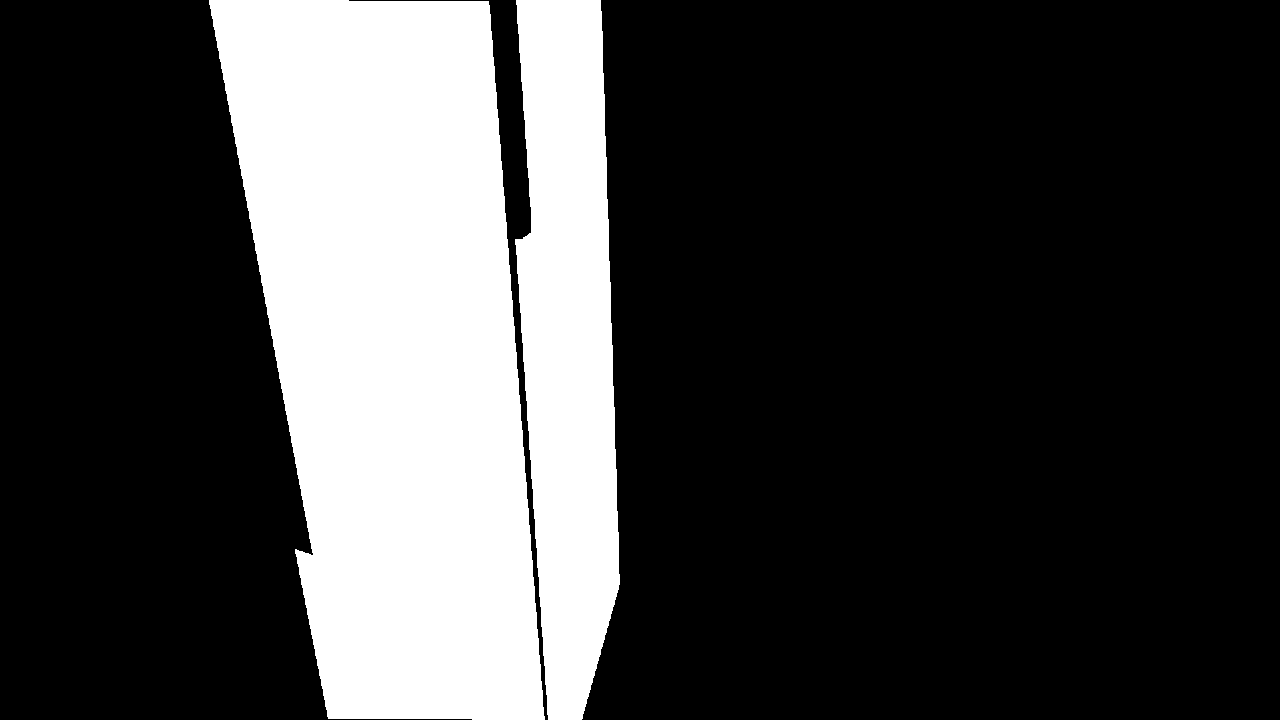}
	\end{subfigure}
	
	\vspace*{1.3mm}
	\begin{subfigure}{0.15\textwidth}
		\includegraphics[width=\textwidth]{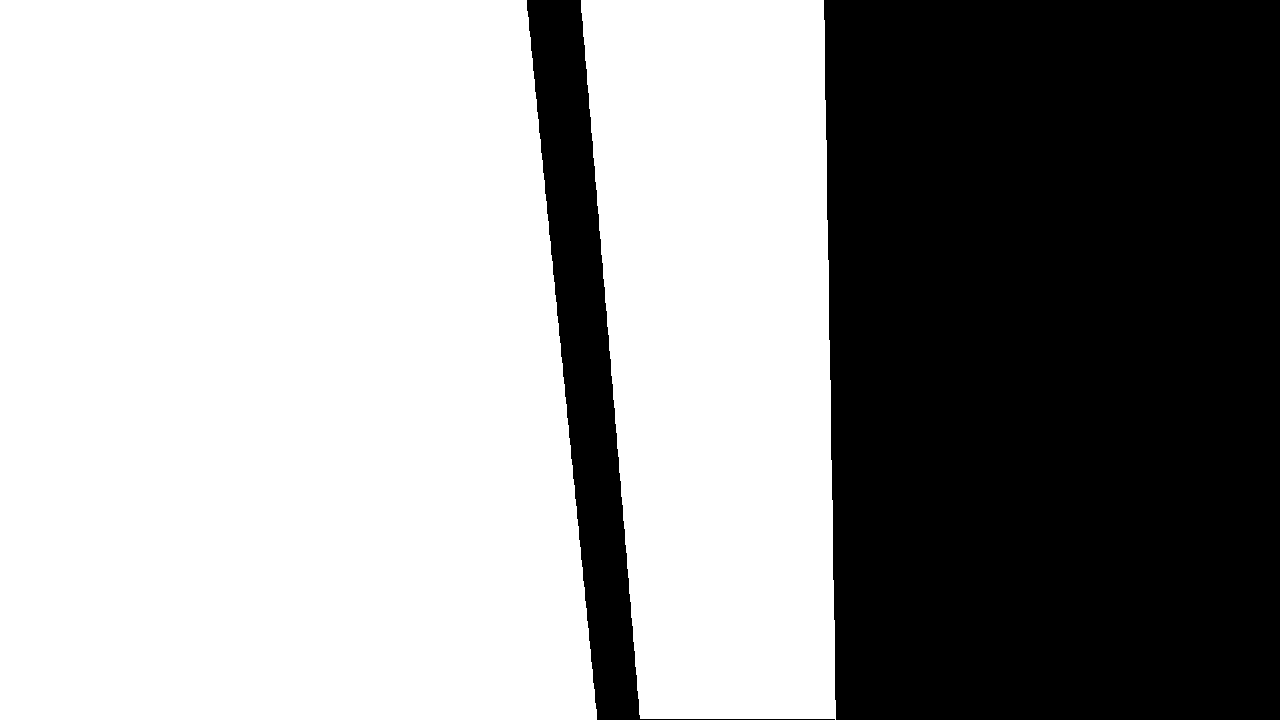}
	\end{subfigure}
	\begin{subfigure}{0.15\textwidth}
		\includegraphics[width=\textwidth]{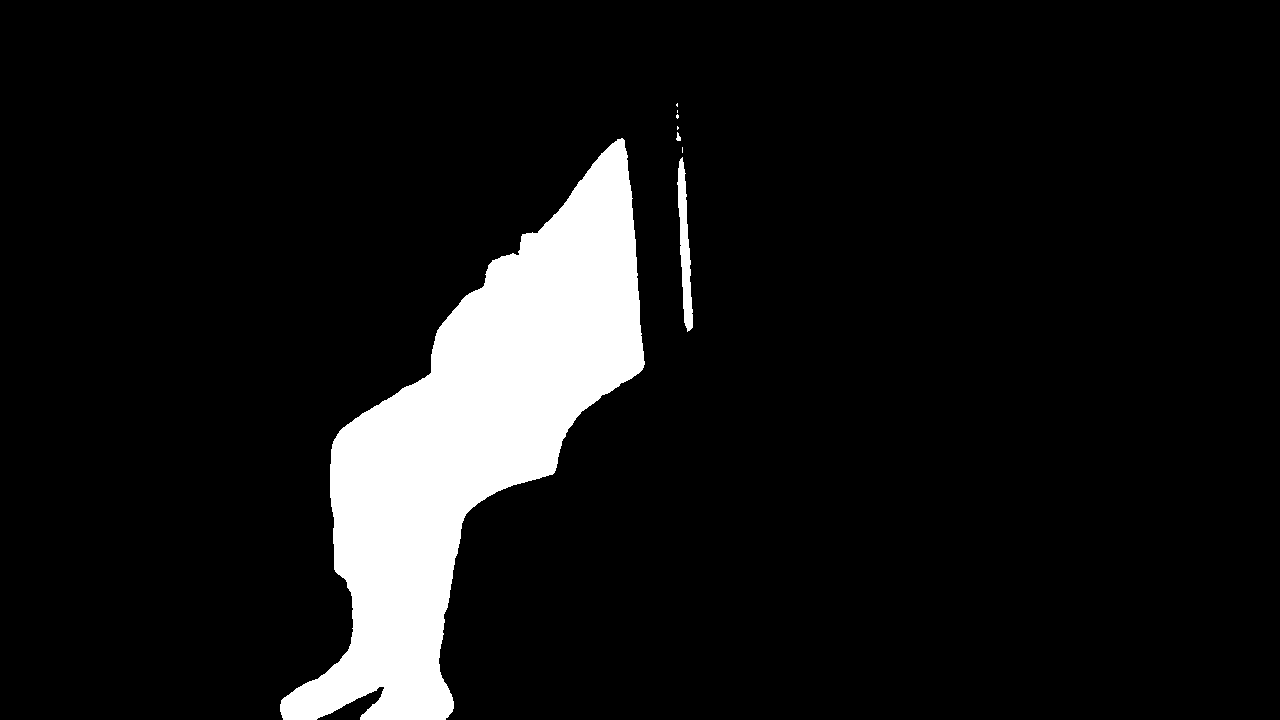}
	\end{subfigure}
	\begin{subfigure}{0.15\textwidth}
		\includegraphics[width=\textwidth]{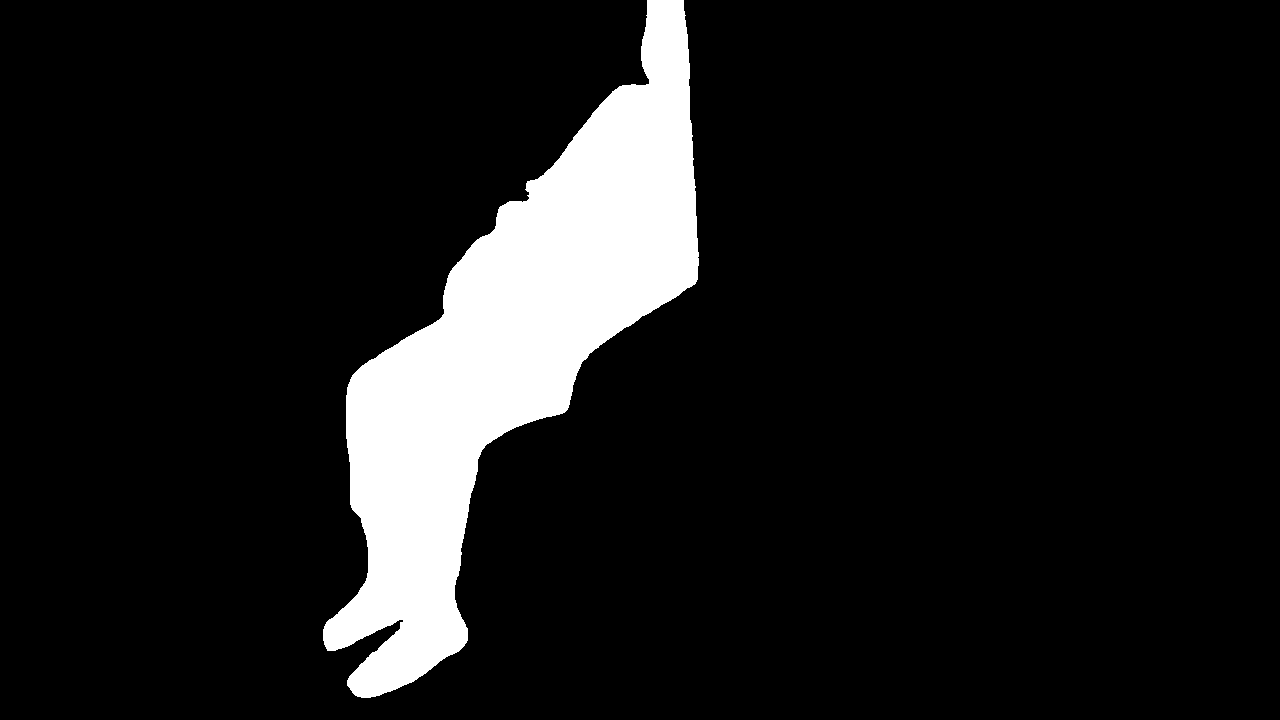}
	\end{subfigure}
	\begin{subfigure}{0.15\textwidth}
		\includegraphics[width=\textwidth]{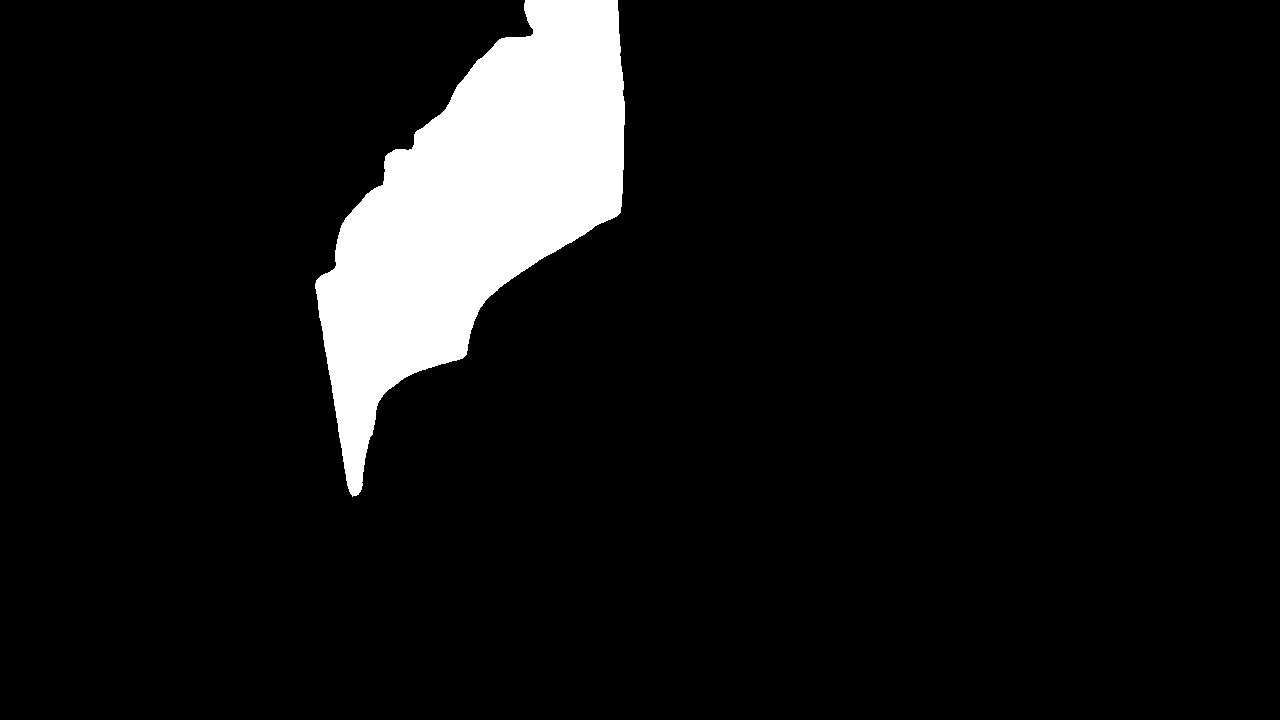}
	\end{subfigure}
	\begin{subfigure}{0.15\textwidth}
		\includegraphics[width=\textwidth]{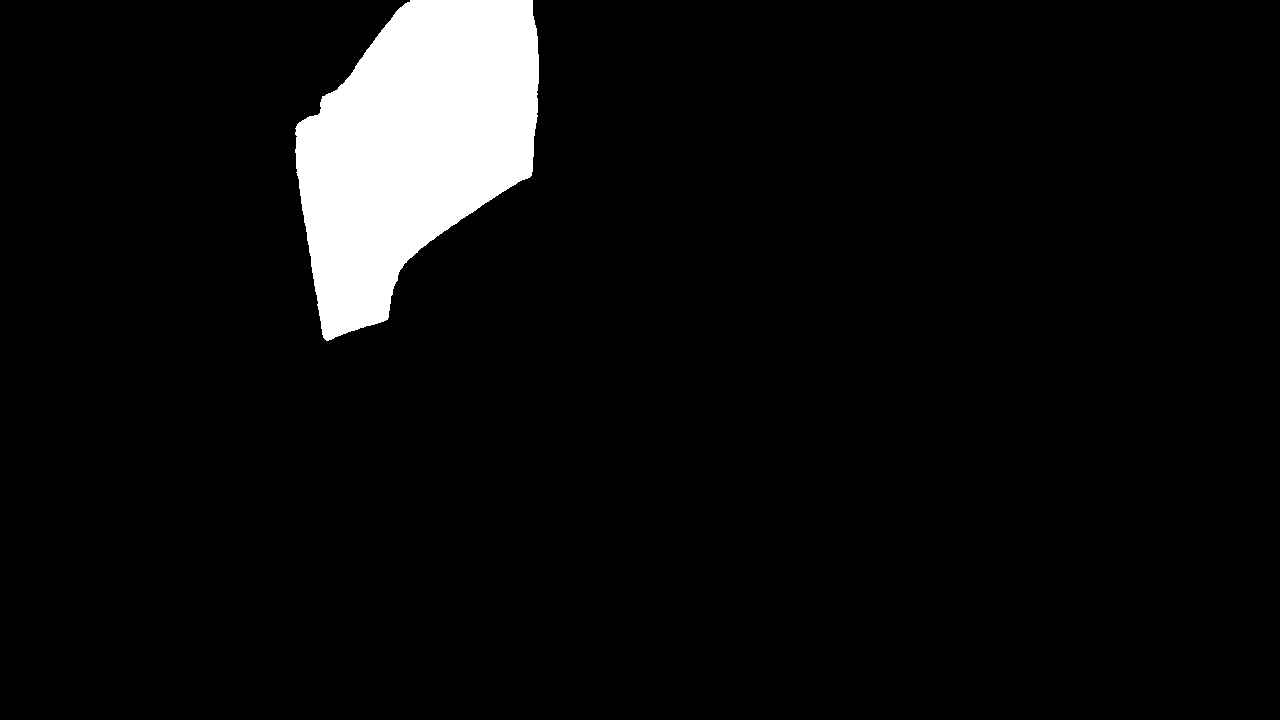}
	\end{subfigure}
	\begin{subfigure}{0.15\textwidth}
		\includegraphics[width=\textwidth]{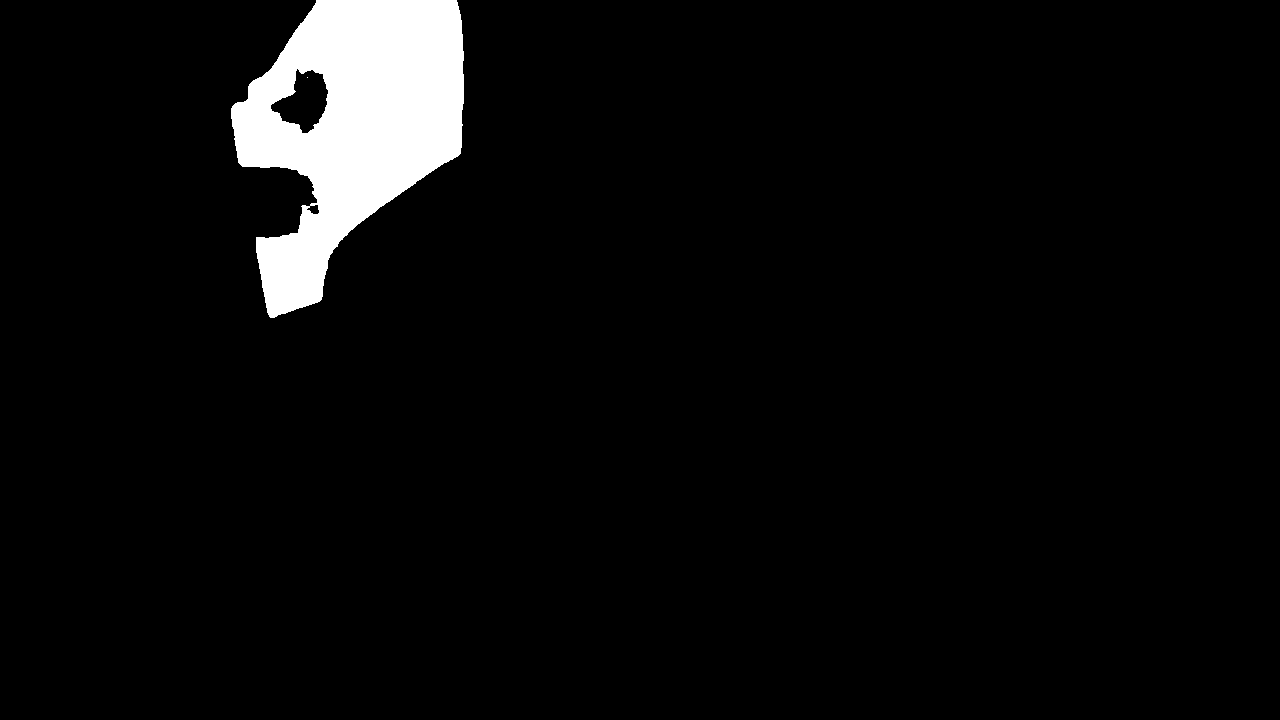}
	\end{subfigure}

	\caption{Qualitative comparison of the predicted segmentations using point prompts on the VMD dataset. The first images of every two rows represents the rgb and groud truth image of the 1st frame, while the other images shown in the even row and in the odd row are the ground truth and predicted shadow points for the 11th, 21th, 31th, 41th, 51th, respectively. Best viewed on screen.}
	\label{fig:fig_vmd_point}
	
\end{figure*}

\subsection{Evaluation Metric}\label{sec:evaluation_metric}
To fully evaluate the peformance of video segmentation, we employ four different metrics:
\begin{itemize}
  \item \textbf{Intersection over Union (IOU)}, also known as the Jaccard Index, is a widely used evaluation metric in computer vision tasks such as object detection and image segmentation, which measures the degree of overlap between the predicted results and the ground truth. IOU can be formulated as:
  \begin{equation}
    IOU=\frac{TP}{TP+FP+FN},
  \end{equation}
  where $TP$, $FP$ and $FN$ are the number of true positive pixels, the number of false positive pixels, the number of false negative pixels, respectively.
  \item \textbf{F1-score} is the harmonic mean of precision and recall. It combines both precision and recall into a single metric, providing a balanced measure of a model's performance. The formula for calculating the F1-score is:
  \begin{equation}
    Precision = \frac{TP}{TP+FP},
  \end{equation}
  \begin{equation}
    Recall = \frac{TP}{TP+FN}, 
  \end{equation}
  \begin{equation}
    F_{1} = 2\times \frac{Precision*Recall}{Precision+Recall} = \frac{2\times TP}{2\times TP+FP+FN},
  \end{equation}
  where $TP$, $FP$ and $FN$ are the number of true positive pixels, the number of false positive pixels, the number of false negative pixels, respectively.
  \item \textbf{Mean absolute error (MAE)} is calculated as the average of the absolute difference between the predicted value ($y_{i}$) and the ground truth value ($\hat{y_{i}}$) for all data points in a dataset. The formula for MAE is given by:
  \begin{equation}
    MAE=\frac{1}{N} \sum\limits_{x=1}\limits^{N}\lvert y_{i}-\hat{y_{i}}\rvert,
  \end{equation}
  \item \textbf{Balance error rate (BER)} is a commonly used metric for shadow detection evaluation. It considers the performance of both shadow prediction and non-shadow prediction. BER can be formulated as:
  \begin{equation}
    BER = \left(1-\frac{1}{2}\times \left(\frac{TP}{N_p} + \frac{TN}{N_n}\right) \right )\times 100
  \end{equation}
  where $TP$, $TN$, $N_p$ and $N_n$ represents the number of true positive pixels, the number of true negative pixels, the number of shadow pixels and the number of non-shadow pixels respectively. For $BER$, the smaller its value, the better the performance.
\end{itemize}
\subsection{Prompt Initialization}\label{sec:prompt_initialization} 
We use two types of prompts: point and mask. For point prompts, we randomly select a predefined number of positive and negative points from the first frame based on the corresponding ground truth mask. Specifically, we randomly choose $N_1$ positive points from the shadow region and $N_2$ negative points from the non-shadow region. For simplicity, we set the number of positive points equal to the number of negative points ($N_1 = N_2$). Although points are explicitly specified as initial prompts, SAM2 will automatically convert them into masks. As a result, there is a risk that the generated masks may not be accurate, which is the primary reason why point prompts perform much worse than mask prompts (for further discussion, see Section~\ref{sec:point_prompt}).

For mask prompts, we directly use the ground truth mask of the first frame as the prompt.

\subsection{Quantitative Evaluation}\label{sec:quantitative_eval}
We present our quantitative evaluation results in Table~\ref{table_quantitative_visha} and Table~\ref{table_quantitative_vmd}. As shown in Table~\ref{table_quantitative_visha}, which demonstrates our results on the ViSha dataset, SAM2 outperforms the state-of-the-art methods in terms of MAE, F1-score, and IoU, except for the BER metric. This holds true whether using SAM-tiny, SAM-small, or SAM-large. When using mask prompts, SAM2 achieves the best performance across all metrics (MAE, F1-score, and IoU). However, when using point prompts, SAM2 performs significantly worse than the state-of-the-art methods, primarily due to the lower accuracy of the generated masks in the first frame (see Figure~\ref{fig:fig_pts}).

As shown in Table~\ref{table_quantitative_vmd}, SAM2 achieves the best performance in all metrics on the VMD dataset when using mask prompts, demonstrating a significant improvement over other methods. Once again, when using point prompts, SAM2's performance is much worse than that of the state-of-the-art methods. Based on these results, we conclude that SAM2 performs better with mask prompts than with point prompts, highlighting its superior ability in modeling both spatial and temporal information for video segmentation.

\subsection{Qualitative Evaluation}\label{sec:qualitative_eval}
Figure~\ref{fig:fig_visha_mask} and Figure~\ref{fig:fig_vmd_mask} show the segmentation masks produced using mask prompts. As illustrated in Figure~\ref{fig:fig_visha_mask}, SAM2 demonstrates satisfactory performance on the ViSha dataset. The generated masks are accurate, and the shadow regions are well segmented. However, as the number of frames increases, the accuracy of the predicted shadow masks declines. A similar phenomenon is observed in the VMD dataset (see the last two rows of Figure~\ref{fig:fig_vmd_mask}).

We also visualize the results using point prompts. As shown in Figure~\ref{fig:fig_visha_point}, the generated masks are highly inaccurate. This issue is also evident in the VMD dataset. SAM2 struggles to generate accurate masks when using point prompts, particularly when the segmented areas contain complex textures.

\subsection{Why Point Prompt Performs Poorly?}\label{sec:point_prompt}
When testing the online demo provided by SAM2, we found that it performs excellently with point prompts for identifying objects in the first frame, particularly when both positive and negative point prompts are used. However, for both video shadow and mirror detection, point prompts perform significantly worse than mask prompts. The primary reason for this is that the generated masks are not accurate enough.

As shown in Figure~\ref{fig:fig_pts}, it is challenging to determine the optimal number of point prompts required to generate a satisfactory mask. For example, in the scene shown in the third row, using 20 or 30 point prompts results in worse performance compared to using just 10 point prompts. A similar phenomenon is observed in the first, second, ninth, and tenth rows. SAM2 struggles to generate accurate masks when using point prompts, especially when the segmented areas contain complex textures. This is the root cause of why point prompts perform much worse than mask prompts.

\subsection{Discussion}\label{sec:discussion}
Based on our experiments, we conclude the following:

\begin{itemize} 
  \item SAM2 struggles to generate accurate masks when using point prompts for the first frames of videos. 
  \item SAM2 performs well when using mask prompts for the first frames of videos. 
  \item SAM2's video segmentation capability is fragile when handling shadow or mirror objects with complex textures when using point prompts. 
\end{itemize}

\section{Conclusion}\label{sec:conclusion}
In this paper, we evaluate SAM2’s performance on two distinct video segmentation tasks: Video Shadow Detection (VSD) and Video Mirror Detection (VMD). We use ground truth point or mask prompts to initialize the first frame and predict corresponding masks for the subsequent frames. Experimental results demonstrate that SAM2 performs satisfactorily on these tasks when using mask prompts, but shows significantly worse performance with point prompts, both quantitatively and qualitatively. In the future, we plan to explore methods for improving SAM2’s performance on these tasks using auto-generated prompts.

\ifCLASSOPTIONcaptionsoff
  \newpage
\fi

{
\bibliographystyle{IEEEtran}
\bibliography{bibliography}
}

\vfill

\end{document}